\documentclass{article}
\pdfoutput=1

\usepackage[utf8]{inputenc} 
\usepackage[T1]{fontenc}    
\PassOptionsToPackage{hyphens}{url} 
\usepackage[hyperfootnotes=false]{hyperref}
\usepackage{booktabs}       
\usepackage{amsfonts}       
\usepackage{amsmath}
\usepackage{nicefrac}       
\usepackage{microtype}      
\usepackage{xcolor}         
\usepackage[noend,ruled]{algorithm2e}

\usepackage{graphicx}
\usepackage{enumerate}
\usepackage{float}
\usepackage{caption}
\usepackage{subcaption}
\usepackage{comment}
\usepackage[export]{adjustbox}
\usepackage{mathtools}
\usepackage[htt]{hyphenat}

\usepackage{authblk}
\usepackage{fullpage}

\usepackage{enumitem}

\usepackage[sort&compress, numbers]{natbib}

\usepackage{tikz}
\usepackage{xcolor}
\usepackage{tcolorbox}

\usepackage{Definitions}
\usepackage{dsfont}
\usepackage{tcolorbox}
\usepackage{array}
\usepackage{xcolor}
\usepackage{colortbl}
\usepackage{multirow}

\newcommand{\Var}{\textnormal{Var}}
\newcommand{\Cov}{\textnormal{Cov}}
\newcommand{\PP}{P}
\newcommand{\norm}[1]{\left\lVert#1\right\rVert}

\usepackage{mathtools}

\newcommand\vertarrowbox[3][0ex]{%
  \begin{array}[t]{@{}c@{}} #2 \\
  \left\uparrow\vcenter{\hrule height #1}\right.\kern-\nulldelimiterspace\\
  \makebox[0pt]{\scriptsize#3}
  \end{array}%
}

\newcommand\numberthis{\addtocounter{equation}{1}\tag{\theequation}}

\newlength{\inlineheight}
\newcommand{\UnnumberedFootnote}[1]{{\def\thefootnote{}\footnote{#1}
\addtocounter{footnote}{-1}}}

\title{Evaluation of Large Language Models \\ via Coupled Token Generation}

\author{Nina {Corvelo Benz}$^{1}$}
\author{Stratis Tsirtsis$^{*,2}$}
\author{Eleni Straitouri$^{3}$}
\author{Ivi Chatzi$^{3}$}
\author{Ander Artola Velasco$^{3}$}
\author{\mbox{Suhas Thejaswi}$^{4}$}
\author{Manuel~Gomez-Rodriguez$^{3}$}

\affil{$^{1}$Max Planck Institute of Biochemistry, Munich, Germany \\
corvelo@biochem.mpg.de}

\affil{$^{2}$Hasso Plattner Institute, Potsdam, Germany \\ stratis.tsirtsis@hpi.de}

\affil{$^{3}$Max Planck Institute for Software Systems, Kaiserslautern, Germany \\
\{estraitouri, ichatzi, avelasco, manuel\}@mpi-sws.org}

\affil{$^{4}$Aalto University, Aalto, Finland \\
suhas.thejaswi@aalto.fi}

\date{}

\begin{document}

\maketitle

\UnnumberedFootnote{$^{*}$The author contributed to this work during his doctoral studies at the Max Planck Institute for Software Systems.}

\begin{abstract}
State of the art large language models rely on randomization to respond to a prompt. As an immediate consequence, a model may respond differently to the same prompt if asked multiple times.
%
In this work, we argue that the evaluation and ranking of large language models should control for the randomization underpinning their functioning.
%
Our starting point is the development of a causal model for coupled autoregressive generation, which allows different large language models to sample responses with the same source of randomness.
%
Building upon our causal model, we first show that, on evaluations based on benchmark datasets, coupled autoregressive generation leads to the same conclusions as vanilla autoregressive generation but using provably fewer samples. However, we further show that, on evaluations based on (human) pairwise comparisons, coupled and vanilla autoregressive generation can surprisingly lead to different rankings when comparing more than two models, even with an infinite amount of samples. This suggests that the apparent advantage of a model over others in existing evaluation protocols may not be genuine but rather confounded by the randomness inherent to the generation process.
%
To illustrate and complement our theoretical results, we conduct experiments with several large language models from the \texttt{Llama}, \texttt{Mistral} and \texttt{Qwen} families.
%
We find that, across multiple benchmark datasets, coupled autoregressive generation requires up to $75$\% fewer samples to reach the same conclusions as vanilla autoregressive generation.
%
Further, we find that the win-rates derived from pairwise comparisons by a strong large language model to prompts from the LMSYS Chatbot Arena platform differ under coupled and vanilla autoregressive generation.
\end{abstract}

\section{Introduction}
\label{sec:intro}
%
One of the most celebrated aspects of state-of-the-art large language models (LLMs) is that they can solve open-ended, complex tasks across many different application domains such as coding, healthcare and scientific discovery~\citep{bubeck2023sparks, mozannar2022reading, haupt2023ai, romera2023mathematical}. 
However, this is crucially what also makes the evaluation and comparison of LLMs very challenging---it is very difficult, if not impossible, to create a single benchmark.
As a consequence, in recent years, there has been a flurry of papers introducing different benchmarks~\citep{bach2022promptsource,wei2022finetuned,talmor2019commonsense,mishra2022cross,chen2021evaluating,liang2023holistic,longpre2023flan,hendryckstest2021,wang2022self,chiang2024chatbot,zheng2023judging,li2023generative,li2023prd}.
In fact, one of the flagship conferences in machine learning has created a separate datasets and benchmarks track!

%
In this context, it is 
surprising that, in comparison, there has been a paucity of work understanding, measuring or controlling for the different sources of uncertainty present in the evaluations and comparisons of LLMs based on these benchmarks~\citep{miller2024adding,madaan2024quantifying,dubey2024llama,saadfalcon2023ares,boyeau2024autoeval,chatzi2024prediction,dorner2024limits,gera2024justrank}.
In our work, we focus on one source of uncertainty that has been particularly overlooked, the uncertainty in the outputs of the LLMs under sampling-based decoding. 

%
Under sampling-based decoding, given an input prompt, LLMs generate a sequence of tokens (\eg, sub-words)
as output using a non-deterministic autoregressive process~\citep{bengio2000neural,radford2019language}. 
At each time step, they first use a neural network to map the prompt and the (partial) sequence of tokens generated so far to a token distribution. 
Then, they use a sampler to draw the next token at random from the token distribution.
Finally, they append the next token to the (partial) sequence of tokens, and continue until a special end-of-sequence token is sampled.
To illustrate why, in the context of LLM evaluation and ranking, one may like to control for the randomization underpinning sampling-based decoding, we will use a stylized example.
%
%
Consider that we are given two LLMs $m$ and $m'$ and we need to evaluate the accuracy of their responses to the prompts of a benchmark dataset. 
However, we are not told that both LLMs are in reality identical copies of each other.
It is easy to see that, due to the randomization of the autoregressive processes used by $m$ and $m'$, there may exist prompts where the (correctness of) their answers differ. 
As a consequence, 
one may need many responses to the prompts of the benchmark dataset to conclude confidently that $m$ and $m'$ achieve the same accuracy. 
In our work, we show both theoretically and empirically that controlling for the randomization of the autoregressive processes underpinning the LLMs under comparison can significantly reduce the number of responses required to reliably compare the performance of LLMs, and this advantage generalizes to non-identical LLMs.

\xhdr{Our contributions}
%
Our key idea is to couple the autoregressive processes underpinning a set of LLMs under comparison, particularly their samplers, by means of sharing the same source of randomness. 
To this end, we treat the sampler of each LLM as a causal mechanism that receives as input the distribution of the next token and the same set of noise values, which determine the sampler'{}s (stochastic) state.
By doing so, at each time step of the generation, we can expect that, if different LLMs map the prompt and the (partial) sequence of tokens generated so far to the same token distribution, they will sample the same next token.
Loosely speaking, in the context of LLM evaluation and ranking, coupled autoregressive generation ensures that no LLM will have better luck than others.
%
More formally, on evaluations based on benchmark datasets, we show that the difference in average performance of each pair of LLMs under comparison is asymptotically the same under coupled and vanilla autoregressive generation, but coupled autoregressive generation provably leads to a reduction in the required sample size.
On evaluations based on 
pairwise comparisons, we show that the win-rates of the LLMs under comparison can be asymptotically different under coupled and vanilla autoregressive generation and, 
surprisingly, the resulting rankings can differ.
This suggests that the apparent advantage of an LLM over others in existing evaluation protocols may not be genuine but rather confounded by the randomness inherent to the generation process.

%
To illustrate and complement our theoretical results, we conduct experiments with several LLMs of the \texttt{Llama}, \texttt{Mistral} and \texttt{Qwen} families.
%
%
We find that, across multiple benchmark datasets, namely, MMLU, GSM8K and HumanEval, coupled autoregressive generation leads to a reduction of up to $75$\% in the required number of samples to reach the same conclusions as vanilla autoregressive generation.
%
%
Further, we find that the win-rates derived from pairwise comparisons by a strong large language model to prompts from the LMSYS Chatbot Arena platform differ under coupled and vanilla autoregressive generation.
An open-source implementation of coupled autoregressive generation is available at \url{https://github.com/Networks-Learning/coupled-llm-evaluation}.

\xhdr{Related work}
Our work builds upon a very recent work 
by~\citet{chatzi2025counterfactual}, which also treats the sampler of an LLM as a causal mechanism.
However, their focus is different to ours; they augment a single LLM with the ability to reason counterfactually about alternatives to its own outputs if individual tokens had been different.
%
Our work also shares technical elements with a recent work by~\citet{ravfogel2024counterfactual}, which develops a causal model to generate counterfactual strings resulting from interventions within (the network of) an LLM. 
However, their work does not study counterfactual generation for the purposes of model evaluation.
%
In this context, it is also worth pointing out that the specific class of causal models used in the aforementioned works and our work, called the Gumbel-max structural causal model~\citep{oberst2019counterfactual}, has also been used to enable counterfactual reasoning in different contexts~\citep{tsirtsis2021counterfactual,noorbakhsh2022counterfactual,benz2022counterfactual}.
%
Lastly, our work broadly relates to the rapidly increasing literature on evaluation and comparison of LLMs~\cite{chang2024asurvey}---for further discussion, we refer the reader to Appendix~\ref{app:further-related-work}.

\section{A Causal Model for Coupled Autoregressive Generation}
\label{sec:scm}
\begin{figure*}
    \centering
    \includegraphics[width=1\linewidth]{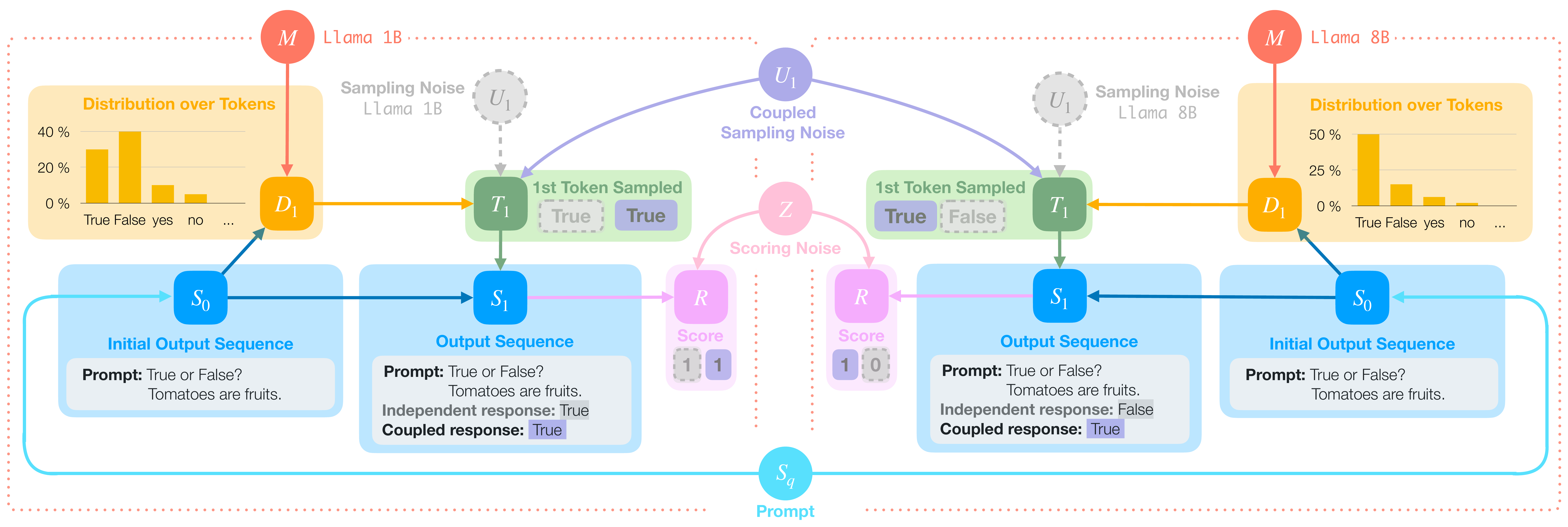}
    \caption{{\bf Example of coupled autoregressive generation for \texttt{Llama 1B} and \texttt{Llama 8B}.} Boxes represent endogenous random variables and circles represent exogenous random variables. The value of each endogenous variable is given by a function of the values of its ancestors in the causal graph, as defined by Eq.~\ref{eq:SCM}. The value of the coupled noise variable $U_1$ (purple) is sampled independently from a given distribution $P_U$, and it determines the stochastic state of the samplers used by both \texttt{Llama 1B} and \texttt{Llama 8B} during the generation of token $T_1$.}
    \label{fig:scm}
\end{figure*}

Let $V$ denote a vocabulary (set) of tokens, including an end-of-sequence token $\bot$, 
$V^*=V\cup V^2 \cup \dots \cup V^K$ be the set of sequences of tokens up to a (context) length $K$, 
and $\varnothing$ be the empty token.
%
An LLM $m \in \Mcal$ takes as input a prompt sequence $s_q \in V^*$ and responds with an output sequence $s \in V^*$, generated using an autoregressive process. 
At each time step $i \in [K]$ of the process, the LLM first takes as input the concatenation of the prompt sequence $s_q$ and the (partial) output sequence $s_{i-1}$, and generates a distribution over tokens $d_{i} \in \Delta(V)$. 
Then, under sampling-based decoding, it samples the next token $t_{i} \sim d_{i}$ from the distribution $d_{i}$ and creates the output sequence $s_{i} = s_{i-1} \circ t_{i}$, where $\circ$ denotes the concatenation of a token or sequence with another sequence.
If $t_{i} = \bot$, it terminates and returns $s = s_i$, otherwise, it continues to the next step $i+1$ in the generation. 
Once the process is completed, the output sequence $s$ is assigned a score $r$, which is subsequently used for model evaluation.

Following~\citet{chatzi2025counterfactual}, we augment the above autoregressive process using a structural causal model (SCM)~\citep{pearl2009causality, peters2017elements}, denoted as $\Ccal$, and use capital and lowercase letters for random variables and their realizations, respectively. 
The SCM $\Ccal$ is defined by the following structural equations:
%
\begin{equation}\label{eq:SCM}
\begin{split}
           S_0&=S_q,
    \quad  T_{i} = \begin{cases}
        f_T( D_{i}, U_{i}) & \text{if} \,\, D_{i} \neq P_\varnothing,\\
        \varnothing & \text{otherwise}
    \end{cases},\\
     D_{i} 
        &=\begin{cases}
        f_D(S_{i-1},M)
        & \text{if} \,\, \texttt{last}(S_{i-1})\neq \bot,\\
        P_\varnothing & \text{otherwise}
    \end{cases},
        \\[2ex] 
    S_{i} &=  
    S_{i-1} \circ T_{i}, 
    \quad S=S_K,
    \text{ and } \quad R=f_R(S, Z).
\end{split}
\end{equation}
In the above equations, $M,S_q,\Ub = (U_i)_{i \in \{1, \ldots, K\}}$, and $Z$ are independent exogenous random variables, 
with $M \sim P_{M}$, $S_q \sim P_{Q}$, $U_i \sim P_{U}$, and $Z \sim P_{Z}$.
Moreover, $f_D$, $f_T$ and $f_R$ are given functions,
$P_\varnothing$ denotes the point mass distribution on $\varnothing$, and $\texttt{last}(S_{i-1})$ denotes the last token of the sequence $S_{i-1}$. 
Here, the function $f_D$ maps an input sequence $S_{i-1}$ to a distribution $D_i$ for the next token using the architecture and network weights of the LLM $M$, 
the function $f_T$ and distribution $P_U$ specify the sampling mechanism that is used to sample the next token from the distribution $D_i$,
and the function $f_R$ and distribution $P_Z$ specify the scoring process by which the score $R$ is assigned to an output sequence $S$ during the evaluation of the LLM $M$.

Throughout the paper, we focus on sampling mechanisms that satisfy counterfactual stability~\citep{oberst2019counterfactual,tsirtsis2021counterfactual,chatzi2025counterfactual}, an intuitive form of consistency between the next token $T_i$, the distribution $D_i$, and the corresponding noise variable $U_i$.
Loosely speaking, counterfactual stability ensures that, if the sampling mechanism samples a specific token $t_i$ from a distribution $d_i$, it would still sample $t_i$ in a counterfactual scenario where $d_i$ is replaced by a distribution $d'_i$ that assigns an even higher probability to $t_i$
%
(refer to Appendix~\ref{app:stability} for a formal definition).
In this context, note that this choice does not restrict the practicality of our approach---the default categorical sampler in \texttt{PyTorch}~\citep{paszke2019pytorch}, one of the most popular libraries used by state of the art LLMs, is an implementation of the Gumbel-Max SCM~\citep{oberst2019counterfactual}, an SCM which satisfies counterfactual stability.
Moreover, since the choice of the sampling mechanism is entirely up to the evaluator, one may be better off using a sampling mechanism that satisfies counterfactual stability in order to reduce the number of samples needed for evaluation, as we show theoretically and experimentally in our work. 
%
That said, evidence from related work~\citep{chatzi2025counterfactual} indicates that,
even if the sampling mechanism does not satisfy counterfactual stability, coupled generation may still be proven useful, but to a lesser extent. 
Refer to Section~\ref{sec:discussion} for more details.

Further, we allow the score $R$ to be observable or unobservable, and its semantic meaning and support of its distribution to vary depending on the evaluation protocol. For example, in multiple-choice questions~\citep{hendrycks2021measuring}, $R\in\{0,1\}$ may represent whether an LLM outputs a correct ($R=1$) or an incorrect ($R=0$) response. In pairwise comparisons~\citep{chiang2024chatbot}, $R\in\RR^+$ may represent the level of user'{}s satisfaction with the response provided by an LLM. 
In this context, the noise variable $Z$ models any potential sources of uncertainty in the scoring process, \eg, uncertainty in users' preferences~\citep{thurstone1927law,bradley1952rank,luce1959individual}.

Building upon the above causal model, we can now formally express what it means to sample (and evaluate) output sequences by different LLMs using the same source of randomness, a process we refer to as \emph{coupled autoregressive generation}.\footnote{We (implicitly) assume that the LLMs share the same vocabulary $V$; refer to Section~\ref{sec:discussion} for a discussion of potential approaches to relax this assumption.}
%
Consider a specific model $m$, a prompt $s_q$, and fixed noise values $\ub$ and $z$. It is easy to see that specifying these values is sufficient to compute the exact value of the output sequence $S$ and its score $R$ using the autoregressive generation and scoring process given by Eq.~\ref{eq:SCM}. 
Then, we can formally express the coupled output sequences by two models $m$ and $m'$ and their corresponding scores $R_m (\ub, s_q, z)$ and $R_{m'} (\ub, s_q, z)$ as the result of \emph{interventions} $do(M=m)$ and $do(M=m')$, respectively, where the $do(\cdot)$ operator forcibly sets the value of $M$ while keeping the prompt $s_q$ and the noise values $\ub$, $z$ fixed~\citep{pearl1994probabilistic}. 
%
%
For an illustration of coupled autoregressive generation against independent autoregressive generation---the vanilla generation approach---refer to Figure~\ref{fig:scm}. 

From a computational perspective, it is important to note that coupled autoregressive generation does not increase the time complexity or memory footprint of the evaluation of LLMs, thus presenting zero overhead compared to the vanilla approach.
Specifically, under an efficient implementation of coupled autoregressive generation, computing the (coupled) responses of a set of models under comparison simply reduces to running their generative processes using the same random seed; this effectively ensures that they all share the same prompt $s_q$ and noise values $\ub$ and $z$.\footnote{In benchmark datasets, $z$ is fixed since the scoring process reduces to deterministically comparing each output against a ground truth. In pairwise comparisons, $z$ is fixed because the scoring process for each pair of outputs under comparison is performed by an end user simultaneously.}
From a causal perspective, we can view these runs as realizations of possible worlds where everything is equal except for the (architecture and network weights of the) LLM.
Or we can also view one of these runs as a realization of the factual world and the 
other runs as realizations of different counterfactual worlds. 
Consequently, this lends support to attribute any difference in the scores $R_m (\ub, s_q, z)$ across models $m \in \Mcal$ to the models' architectures and weights rather than the randomness in their autoregressive generation processes. 
In that context, note that coupled autoregressive generation only modifies the joint distribution of the output of the models under comparison, while keeping their marginal distributions the same as under independent autoregressive generation---it only modifies the \emph{pairing} of the outputs, not the outputs themselves.
Therefore, it does not influence any capabilities of the models, such as their creativity, that are typically associated with the randomness of their outputs~\citep{peeperkorn2024temperature}.

\section{Evaluation based on benchmark datasets}
\label{sec:individual}
In this section, we focus on the evaluation and comparison of LLMs based on benchmark datasets, \eg, multiple-choice questions~\citep{hendrycks2021measuring}, and theoretically investigate under which conditions coupled autoregressive generation requires fewer samples than independent autoregressive generation to estimate the competitive advantage of one LLM over another.

The results we present here apply to real-valued scores in a bounded interval, however, we focus on binary scores (\eg, for benchmark datasets where the output sequences can be classified as correct or incorrect) for ease of exposition.
Given a benchmark dataset characterized by an input prompt distribution $P_Q$, for each prompt $s_q \sim P_Q$,
let $\textsc{c}(s_q) \subset V^*$ denote the set of correct output sequences. Then, the score is given by
    $R_m(\ub, s_q) = \one\left\{S_m(\ub, s_q) \in \textsc{c}(s_q)\right\} \in \{0, 1\}$,
where $S_m(\ub, s_q)$ denotes the output sequence of a model $m$ given a prompt $s_q$ under a realized sequence of noise values $\ub$ and $\one\{\cdot\}$ is the indicator function.

The standard approach to compare the performance of any pair of LLMs $m, m' \in \Mcal$ using a benchmark dataset reduces to estimating the difference in their expected score, \ie,
\begin{align}
    \EE_{\Ub\sim P_{\Ub}, \Ub'\sim P_{\Ub}, S_q\sim P_Q} [&R_{m}(\vertarrowbox{\Ub}{}, S_q) - R_{m'}(\vertarrowbox{\Ub}{}', S_q)], \label{eq:independent-generation-difference} \\[-3.5ex]
    &\text{\small \hspace{2mm} \,\, Independent generation} \nn
\end{align}
where note that we use different noise variables $\Ub$ and $\Ub'$ for each LLM because, in the standard approach, each LLM generates outputs to each query independently (\ie, using independent autoregressive generation).
At first, one may think that, in this context, coupled autoregressive generation will not be helpful. 
Under coupled autoregressive generation, the difference in the expected score adopts the following form:
\begin{align}
\EE_{\Ub\sim P_{\Ub},S_q \sim P_Q} [R_m&(\vertarrowbox{\Ub}{},S_q)-R_{m'}(\vertarrowbox{\Ub}{},S_q)]. \label{eq:coupled-generation-difference} \\[-3.5ex] &\text{\small \,\,Coupled generation} \nn
\end{align}
Therefore, based on the linearity of expectation and the fact that, under independent generation, both $\Ub$ and $\Ub'$ are sampled from the same distribution $P_{\Ub}$, it is easy to see that Eqs.~\ref{eq:independent-generation-difference} and~\ref{eq:coupled-generation-difference} are equivalent. 
However, as we will show next, coupled autoregressive generation allows us to reliably estimate the difference in the two LLMs' scores from finite samples faster.
More formally, we first characterize the relation between the variances of the difference of scores between LLMs as follows (refer to Appendix~\ref{sec:proofs} for all proofs):
%
\begin{proposition}\label{prop:variance}
    For any $m, m' \in \Mcal$, it holds that
    \begin{multline*}
        \textnormal{Var}[R_m(\Ub,S_q)-R_{m'}(\Ub',S_q) ] 
        = \textnormal{Var}[R_m(\Ub,S_q)-R_{m'}(\Ub,S_q) ] 
        + 2\cdot  \Cov[R_m(\Ub,S_q),R_{m'}(\Ub,S_q)]
    \end{multline*}
\end{proposition}
%
This result immediately implies that, if the scores achieved by the LLMs under comparison are positively correlated, \ie, the LLMs tend to generate a (in-)correct output sequence on the same prompts under the same noise values, then the variance of the difference in scores is lower under coupled generation than under independent generation, and thus we can expect a reduction in the sample size required to obtain equivalent estimation errors.
Next, we theoretically analyze two canonical settings in which this condition holds and, in Section~\ref{sec:experiments}, we provide empirical evidence that it also holds in well-known benchmark datasets.

In the first canonical setting, the correct response to each prompt is one of two given single-token sequences and the LLMs $m$ and $m'$ under comparison always output a response that is either of these two sequences.
While this setting may seem restrictive, it is found in real-world scenarios. 
For example, think of evaluation protocols in which the LLMs are explicitly instructed to always output true/false (or one of two options) via their system  prompt.\footnote{Here, our goal is to illustrate that there exist natural conditions under which coupled autoregressive generation is provably beneficial in comparison to independent autoregressive generation. However, in practice, in this canonical setting, one could directly use the LLMs' probabilities for the two tokens in each prompt to estimate the average difference of scores exactly.} 
The following proposition~shows that the variance of the difference in scores is lower under coupled autoregressive generation:
\begin{proposition} \label{prop:var_stability}
    Consider a benchmark dataset such that $\textsc{c}(s_q) \subsetneq \{t_1, t_2\}$ for all $s_q \sim P_Q$, where $t_1$ and $t_2$ are two single-token sequences. Let $m$ and $m'$ be two LLMs that assign positive probability to the sequences $t_1$ and $t_2$ and zero probability to any other sequence.
    If the sampling mechanism defined by $f_T$ and $P_U$ satisfies counterfactual stability, then, it holds that
    \begin{equation}\label{eq:binary_var}
        \textnormal{Var}[R_m(\Ub,S_q)-R_{m'}(\Ub',S_q)] > \textnormal{Var}[R_m(\Ub,S_q)-R_{m'}(\Ub,S_q)]. 
    \end{equation}
\end{proposition}
%

In the second canonical setting, the correct response to each prompt is a single-token sequence, the LLMs $m$ and $m'$ under comparison always output a single-token response, and the sampling mechanism used by the LLMs is given by the Gumbel-Max SCM\footnote{\label{fn:gumbel-max} The Gumbel-Max SCM is defined as $f_T(D_i, U_i) = \argmax_{t\in V} \left\{\log\left(D_{i,t}\right) + U_{i,t}\right\}$, where $U_{i,t} \sim \text{Gumbel}(0,1)$ are i.i.d. noise variables associated with each token~\citep{chatzi2025counterfactual}.}.
Similarly as in the first canonical setting, this second setting is also found in real-world scenarios, particularly taking into account that the default categorical sampler in the library \texttt{PyTorch} implements the Gumbel-Max SCM.
The following proposition shows that, as long as the model $m'$ is \emph{similar enough} to $m$, the variance of the difference in scores is lower under coupled generation:
\begin{proposition}\label{prop:var_gumbel}
    Consider a benchmark dataset such that $|\textsc{c}(s_q)|=1$ for all $s_q \sim P_Q$. Let $m$ be an LLM that assigns positive probability to every single-token sequence and zero probability to any other sequence. 
    If the sampling mechanism defined by $f_T$ and $P_U$ is given by the Gumbel-Max SCM, then, there exists a constant $\varepsilon(m)>0$ such that, for every LLM $m'$ that assigns positive probability to every single-token sequence and zero probability to any other sequence and satisfies $d(m,m')=\sup_{s_q} \norm{f_D(s_q,m)-f_D(s_q, m')}_\infty < \varepsilon(m)$, it holds that
    \begin{equation*}
        \textnormal{Var}[R_m(\Ub,S_q)-R_{m'}(\Ub',S_q)]
        > \textnormal{Var}[R_m(\Ub,S_q)-R_{m'}(\Ub,S_q)].
    \end{equation*}
\end{proposition}
Based on the above proposition, we hypothesize that coupled autoregressive generation will reduce the number of samples required to reliably compare the performance of LLMs whenever these are sufficiently \emph{similar}, \eg, whenever we compare fine-tuned or quantized versions of the same pre-trained LLM.

\section{Evaluation based on pairwise comparisons}
\label{sec:pairwise}
In this section, we focus on the evaluation and comparison of LLMs according to their level of alignment with human preferences, as elicited by pairwise comparisons between outputs of different LLMs to the same prompts.
Such an evaluation protocol has become particularly popular to evaluate and compare LLMs in open-ended, complex tasks in which, in contrast to benchmark datasets, there are no structured ground-truth outputs.
In what follows, we provably show that, perhaps surprisingly, different LLMs may compare differently under coupled autoregressive generation and under independent autoregressive generation. 

One of the standard approaches to evaluate and compare different LLMs according to their level of alignment with human pairwise preferences reduces to estimating the win-rate achieved by each LLM $m$ against any other LLM $m' \neq m$, \ie,
%
%
\begin{align*}\label{eq:independent-generation-win-rates} 
\EE_{\Ub\sim P_{\Ub}, \Ub'\sim P_{\Ub}, S_q\sim P_Q} [&\one\{R_{m}(\vertarrowbox{\Ub}{}, S_q) > R_{m'}(\vertarrowbox{\Ub}{}', S_q)\}]  \raisetag{5ex}\numberthis \\[-4ex]
&\text{\small \hspace{6mm} \,\, Independent generation} \nn
\end{align*}
where $\one\{R_m(\ub, s_q) > R_{m'}(\ub, s_q)\} = 1 \, (0)$ means that, for prompt $s_q$ and realized sequence of noise values $\ub$, the output of $m$ is (not) preferred over the output of $m'$.\footnote{For simplicity, we assume that human preferences are deterministic and thus $R_m(\ub, s_q, z) = R_m(\ub, s_q)$. We lift this assumption in our experiments in Section~\ref{sec:experiments}.}

Here, similarly as in Eq.~\ref{eq:independent-generation-difference} in the evaluation based on benchmark datasets, we use different noise variables $\Ub$ and $\Ub'$ because, in this standard approach, each LLM generates outputs to each prompt independently (\ie, using independent autoregressive generation).
Conversely, under coupled autoregressive generation, the win-rate adopts the following form:
\begin{align} 
\EE_{\Ub\sim P_{\Ub}, S_q\sim P_Q} [&\one\{R_{m}(\vertarrowbox{\Ub}{}, S_q) > R_{m'}(\vertarrowbox{\Ub}{}, S_q)\}] \label{eq:coupled-generation-win-rates} \\[-3.5ex]
&\text{\small \hspace{8mm} \,\, Coupled generation} \nn
\end{align}
However, in contrast with the comparison of the expected difference in scores under independent and coupled autoregressive generation in the evaluation based on benchmark datasets, 
we cannot directly claim that Eqs.~\ref{eq:independent-generation-win-rates} and~\ref{eq:coupled-generation-win-rates} are equivalent because the win-rate is non-linear with respect to $R_{m}(\ub, s_q)$ and $R_{m'}(\ub', s_q)$. 
In what follows, we further analyze the difference between win-rates in a canonical setting similar to one of the canonical settings used in Section~\ref{sec:individual}.

Consider that, for each prompt, the response can only be one of two given single-token sequences, one of these sequences is preferred over the other by the user, and the LLMs under comparison always output one of them as a response. 
Then, we can compute the win-rates achieved by each LLM $m$ against any other LLM $m' \neq m$ under independent and coupled autoregressive generation: 
\begin{proposition}\label{prop:gap_win_rates_stability}
    Given a fixed prompt $s_q \sim P_{Q}$, assume that $f_{R}(s_+)>f_{R}(s_-)$ for $s_+ = s_q \circ t_+$ and $s_-=s_q \circ t_-$, where $t_+$ and $t_-$ are single-token sequences. 
    Further, assume that the LLMs $m$ and $m'$ respond $t_+$ with probability $p_{m}$ and $p_{m'}$, respectively, and $t_-$ with probability $1-p_{m}$ and $1-p_{m'}$, and the sampling mechanism defined by $f_T$ and $P_{U}$ satisfies counterfactual stability. Without loss of generality, assume $p_{m'} > p_{m}$. 
    Then, under
    coupled
    autoregressive generation, we have that
    \begin{align*}     \label{eq:gap_win_rates_stability_coupled}
        & \EE_{\Ub\sim P_{\Ub}}[\one\{R_m(\Ub, s_q)>R_{m'}(\Ub, s_q)\}] = 0, \\ 
        & \EE_{\Ub\sim P_{\Ub}}[\one\{R_m(\Ub, s_q)<R_{m'}(\Ub, s_q)\}] = p_{m'} - p_{m}.
    \end{align*}
    Conversely, under
    independent
    autoregressive generation, we have that
    \begin{equation*}    \label{eq:gap_win_rates_stability_independent}
    \begin{split}
        \EE_{\Ub, \Ub' \sim P_{\Ub}}[\one\{R_m(\Ub, s_q)>R_{m'}(\Ub', s_q)\}] &= p_m (1-p_{m'}), \\
        \EE_{\Ub, \Ub' \sim P_{\Ub}}[\one\{R_m(\Ub, s_q)<R_{m'}(\Ub', s_q)\}] &= p_{m'} (1-p_{m}) 
    \end{split}
    \end{equation*}
\end{proposition}
%
From the above proposition, we can readily conclude that, in general, the win-rates do differ under independent and coupled autoregressive generation. Nevertheless, we may be tempted to conclude that, for ranking LLMs, 
this difference appears inconsequential because, for each fixed prompt $s_q$, we have that
\begin{multline*}
\EE_{\Ub\sim P_{\Ub}}[\one\{R_m(\Ub, s_q)<R_{m'}(\Ub, s_q)\}]  - \EE_{\Ub\sim P_{\Ub}}[\one\{R_m(\Ub, s_q)>R_{m'}(\Ub, s_q)\}] \\ 
\\= \EE_{\Ub, \Ub' \sim P_{\Ub}}[\one\{R_m(\Ub, s_q)<R_{m'}(\Ub', s_q)\}] 
 - \EE_{\Ub, \Ub' \sim P_{\Ub}}[\one\{R_m(\Ub, s_q)>R_{m'}(\Ub', s_q)\}].
\end{multline*}
However, whenever one needs to rank more than two LLMs, the difference in win-rates can be actually consequential---the rankings derived from the win-rates can be different under independent and coupled autoregressive generation, as illustrated by the following simple example.

\begin{figure*}
    \centering
    \subfloat[Score covariance]{
    \label{fig:computer-sc-cov-1B-3B}
    \includegraphics[width=0.3\linewidth]{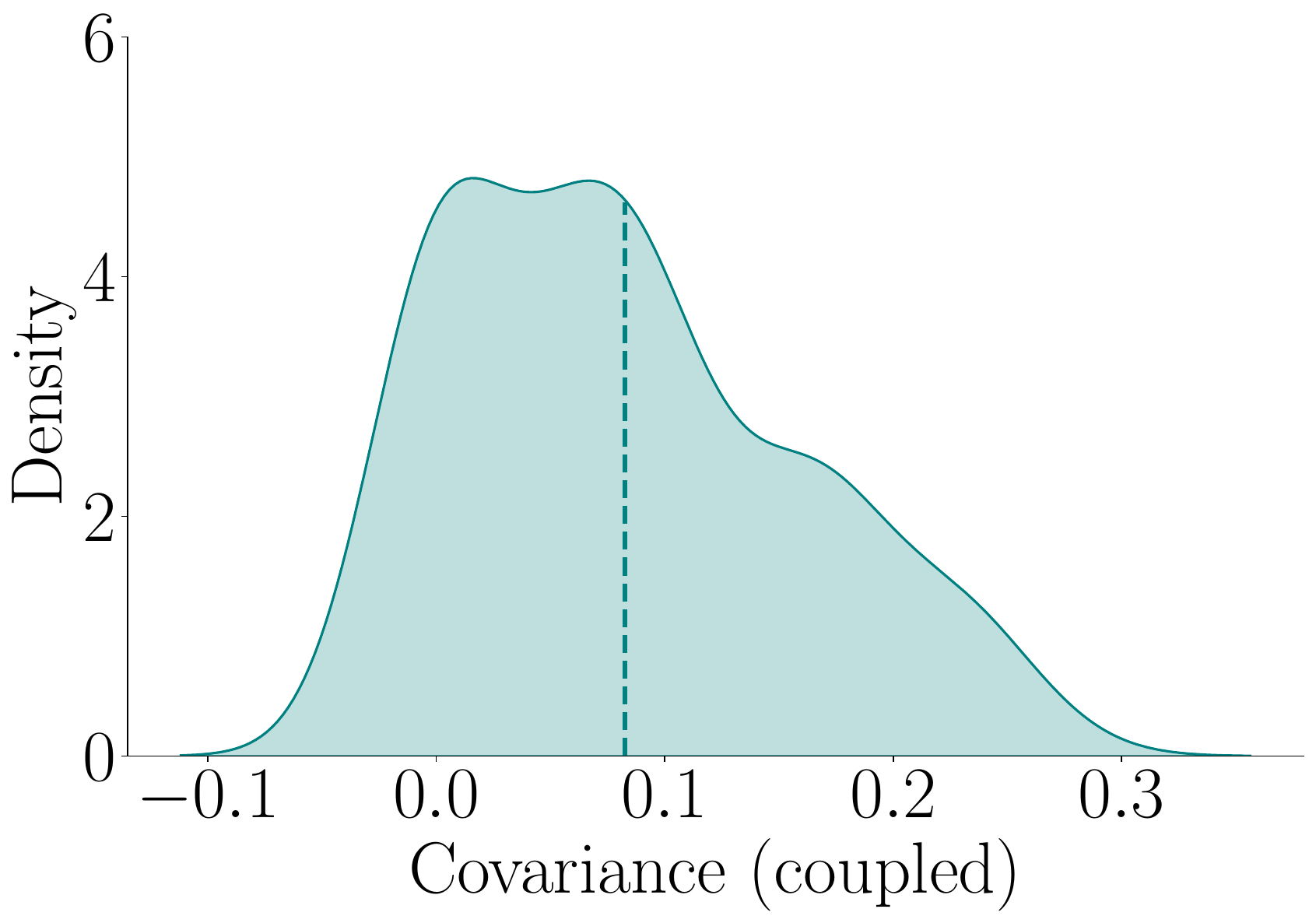}
    }
    \hspace{1mm}
    \subfloat[Variance of the score difference]{
    \label{fig:computer-sc-var-1B-3B}
    \includegraphics[width=0.3\linewidth]{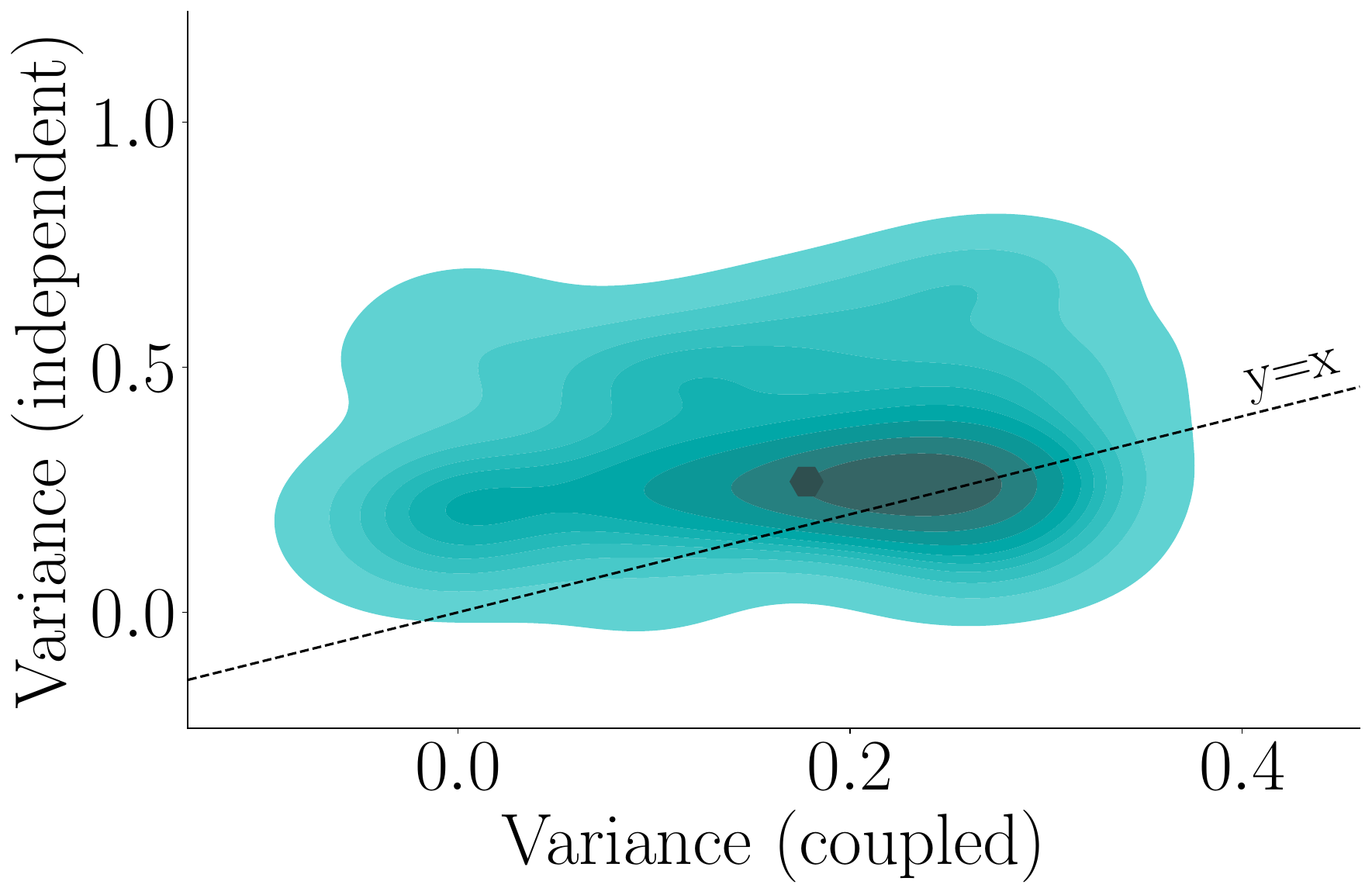}
    }
    \hspace{1mm}
    \subfloat[Estimation error vs. \# samples ]{
    \label{fig:computer-sc-err-1B-3B}
    \includegraphics[width=0.3\linewidth]{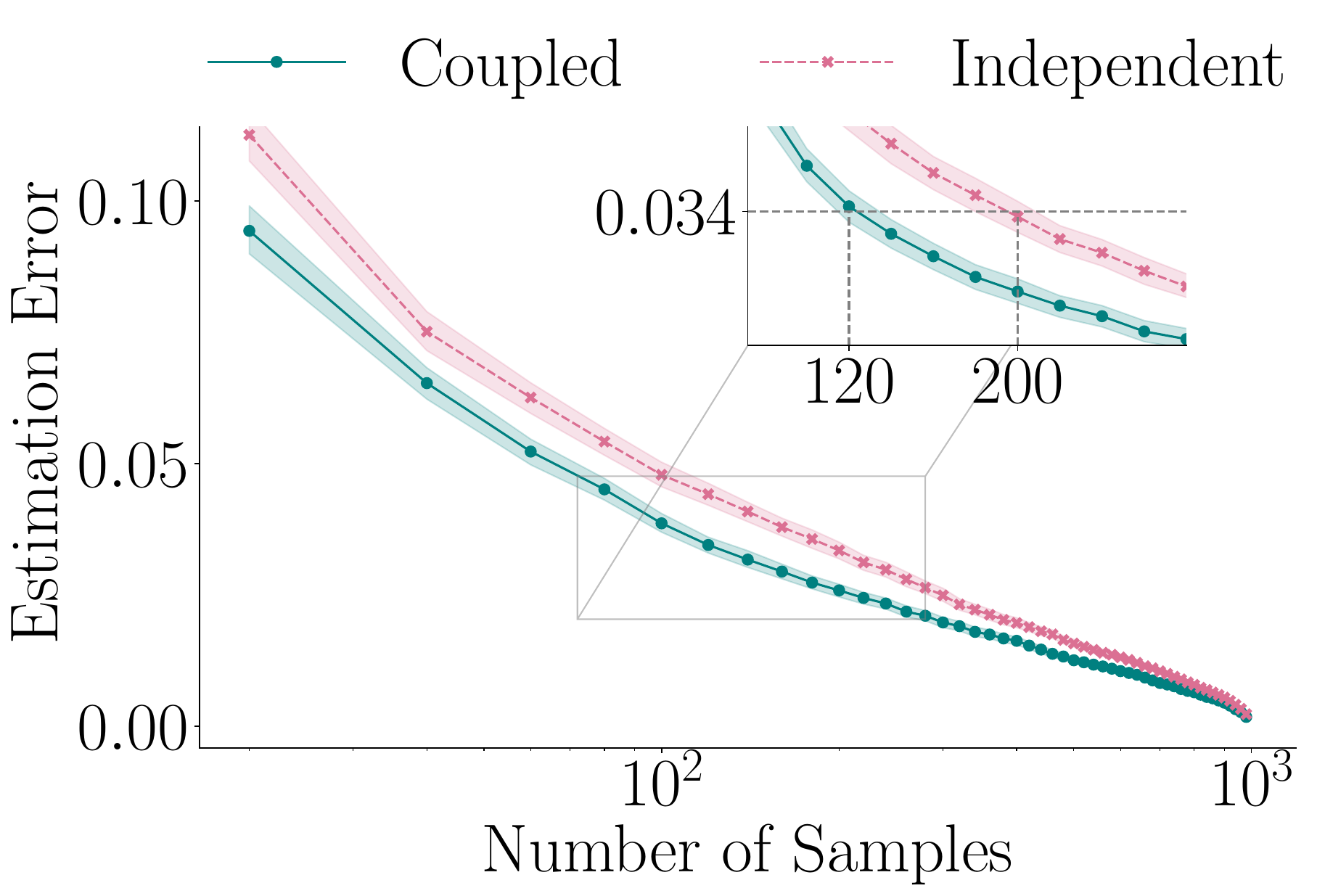}
    }
    \caption{\textbf{Comparison between \texttt{1B} and \texttt{3B} on questions from the knowledge area ``college computer science'' of the MMLU dataset.}
    Panel (a) shows the kernel density estimate (KDE) of the covariance between the scores of the two LLMs on each question under coupled generation; the dashed line corresponds to the average value. Panel (b) shows the KDE of the variance of the difference between the scores of the LLMs on each question under coupled and independent generation; the highlighted point corresponds to the median value. Panel (c) shows the absolute error in the estimation of the expected difference between the scores of the LLMs against the number of samples; for each point on the x-axis, we perform $1{,}000$ sub-samplings and shaded areas correspond to $95\%$ confidence intervals. 
    }
    \label{fig:mmlu-1B-vs-3B-college-cs}
\end{figure*}

Consider we are given three LLMs $m_1$, $m_2$, and $m_3$, and we need to rank them according to the average win-rate they achieve against each other on two input prompts $q$ and $q'$, each with a preferred single-token response out of two single-token responses.
Assume that the probability that the LLMs output the preferred single-token response for $q$ and $q'$ are given by $(m_1\colon 0.4, m_2\colon 0.48, m_3\colon 0.5)$ and $(m_1\colon 1, m_2\colon 0.9, m_3\colon 0.89)$, respectively.
Under independent autoregressive generation, the average win-rates of $m_1$, $m_2$ and $m_3$ are $0.1545$, $0.15675$ and $0.16225$, respectively. Therefore, $m_3$ is ranked at the top, followed by $m_2$, and $m_1$ is ranked last. 
In contrast, under coupled autoregressive generation, the average win-rates of $m_1$, $m_2$ and $m_3$ are $0.0525$, $0.0225$, and $0.03$, respectively, and thus $m_1$ is ranked at the top, followed by $m_3$, and $m_2$ is ranked last.\footnote{Refer to Appendix~\ref{app:example-ranking} for the detailed calculation of the average win-rates.} Crucially, this case illustrates how rankings obtained using coupled and independent autoregressive generation can differ, leading to opposite conclusions regarding the LLMs' performance.
In Appendix~\ref{app:canonical-pairwise-comparisons}, we analyze another canonical setting in which the win-rates also differ under independent and coupled autoregressive generation as long as the LLMs are \emph{sufficiently similar}.


\section{Experiments}
\label{sec:experiments}
In this section, we evaluate three LLMs from the \texttt{Llama} family of different sizes, namely, \texttt{Llama-3.2-\{1B, 3B\}-Instruct} and \texttt{Llama-3.1-8B-Instruct}, as well as three quantized variants of the latter, namely, \texttt{Llama-3.1-8B-Instruct-\{AWQ-INT4, bnb-4bit, bnb-8bit\}}, 
under coupled and independent autoregressive generation using an implementation of the Gumbel-Max SCM as a sampler~\citep{chatzi2025counterfactual}.
In the remainder of the paper, we refer to them as $\texttt{1B}$, $\texttt{3B}$, $\texttt{8B}$, $\texttt{AWQ-INT4}$, $\texttt{bnb-4bit}$, and $\texttt{bnb-8bit}$. 
For implementation details related to the hardware, datasets and models, refer to Appendix~\ref{app:add_exp}.
For additional qualitatively similar results using models from the \texttt{Mistral} and \texttt{Qwen} families, refer to Appendices~\ref{app:qwen} and~\ref{app:mistral}.
\footnote{Although we experiment only with open-weights LLMs, coupled generation is technically applicable to proprietary LLMs as well, since it only requires access to next-token probabilities.
For example, it can be used during internal evaluations or on crowd-sourced benchmarking platforms, where providers may be incentivized to share them with the platform to enable more sample-efficient comparisons.} 

\begin{figure*}[t]
    \centering
    \subfloat[Empirical win-rates of \texttt{bnb-8bit}]{ 
    \label{fig:win_rate}
    \includegraphics[width=0.34\linewidth]{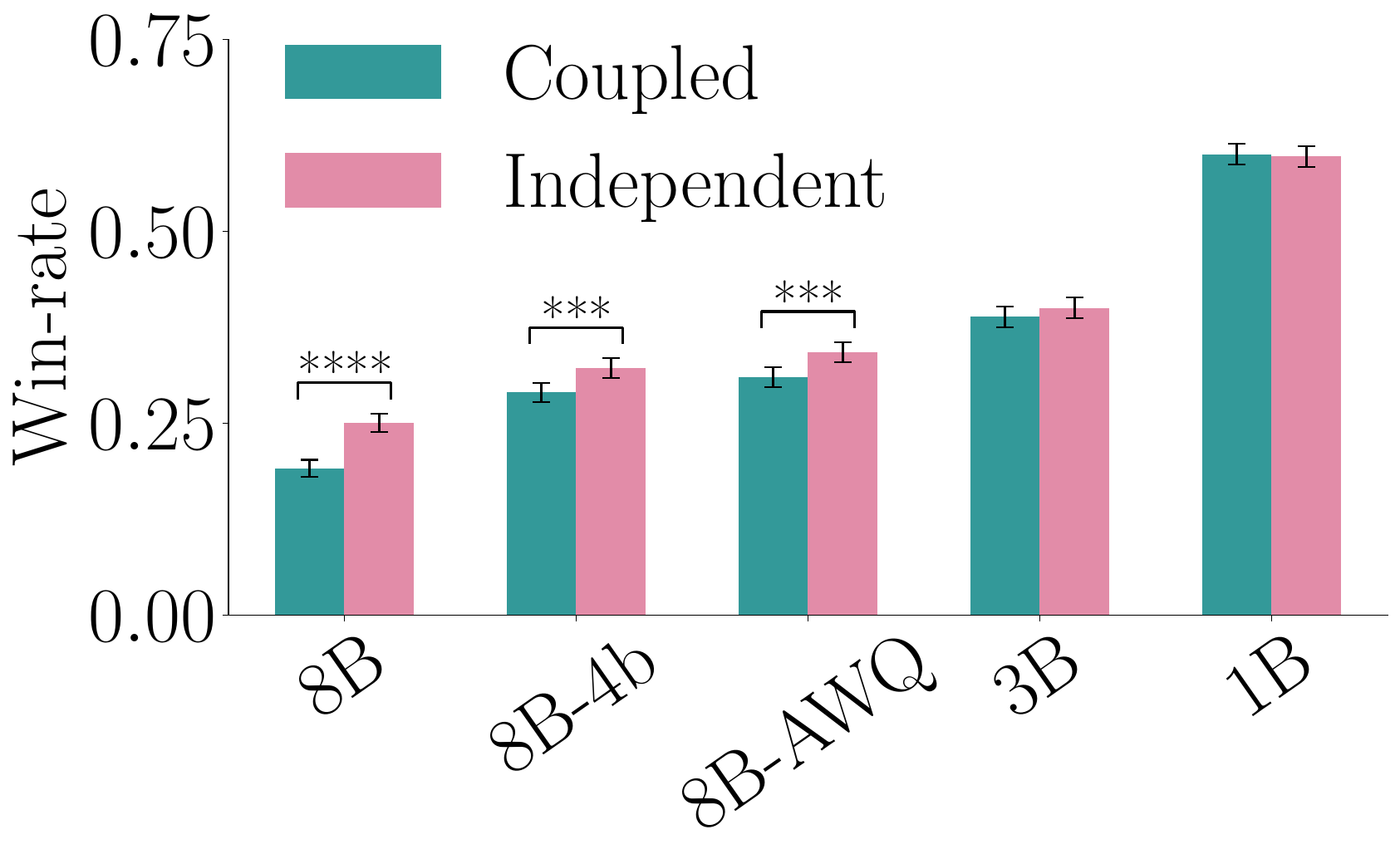}
    }
    \subfloat[Average win-rate and ranking of each LLM]{
    \centering
    \setlength{\tabcolsep}{4pt}
    \footnotesize
        \label{tab:ranking}
        \begin{tabular}{l c c c c}
    \toprule & \multicolumn{2}{c}{Coupled} &  \multicolumn{2}{c}{Independent} \\
    \cmidrule(r{1mm}){2-3} \cmidrule(l{1mm}){4-5}
    {LLM} & Rank & Avg. win-rate 
    & Rank & Avg. win-rate \\ 
    \midrule \texttt{8B} & 1 & 0.3670 $\pm$0.0020 & 1 & 0.3863 $\pm$0.0020 \\ 
    \texttt{bnb-8bit} & 2 & 0.3562 $\pm$0.0020 & 1 & 0.3825 $\pm$0.0020 \\
    \texttt{bnb-4bit} & 3 & 0.3339 $\pm$0.0020 & 3 & 0.3463 $\pm$0.0020 \\
    \texttt{AWQ-INT4} & 4 & 0.3164 $\pm$0.0019 & 4 & 0.3310 $\pm$0.0019 \\
    \texttt{3B} & 5 & 0.2787 $\pm$0.0019 & 5 & 0.2828 $\pm$0.0019 \\
    \texttt{1B} & 6 & 0.1650 $\pm$0.0015 & 6 & 0.1664 $\pm$0.0015 \\ \bottomrule
    \end{tabular}
    }
    \caption{
    \textbf{Evaluation of six LLMs using pairwise comparisons on questions from the LMSYS-Chat-1M dataset.} 
We estimate the empirical win-rate of each LLM against each other using 
pairwise comparisons between the outputs 
to $500$ questions with $10$ (different) random seeds under both coupled and independent generation.
Panel (a) shows the empirical win-rate of $\texttt{bnb-8bit}$ against all other LLMs, where error bars correspond to $95\%$ confidence intervals.
Here, for each pair of empirical win-rates under coupled and independent generation, we conduct a two-tailed z-test, to test the null hypothesis that the empirical win-rates are the same; (\fourstars, \threestars) indicate $p$-values ($<0.0001$, $< 0.001$). We present qualitatively similar results for other LLMs in Appendix~\ref{app:lmsys}.
Panel (b) shows the average win-rate of each LLM across all other LLMs ($\pm$ $95\%$ confidence intervals).
To derive the rankings, for each LLM, we choose the lowest ranking provided by the method of~\citet{chatzi2024prediction}.}
    \label{fig:temp}
\end{figure*}

\subsection{Evaluation on Benchmark Datasets}\label{sec:mmlu}
%
Here, we compare the aforementioned LLMs using the MMLU benchmark dataset~\citep{hendrycks2021measuring}, which comprises $14{,}042$ multiple choice questions covering $52$ knowledge areas.
%
Our goal is to empirically investigate to what extent the theoretical results in Section~\ref{sec:individual} generalize to evaluations based on well known benchmark datasets.\footnote{In practice, for multiple-choice benchmarks, the standard evaluation protocol often uses greedy decoding (\ie, deterministic generation). However, multiple-choice benchmarks offer a simple setting to show quantitatively the effect of coupled autoregressive generation in reducing the amount of samples required for the evaluation of models using sampling-based decoding.}
For additional qualitatively similar results beyond multiple-choice benchmarks using the GSM8K~\citep{cobbe2021gsm8k} and HumanEval~\citep{chen2021evaluating} benchmark datasets, refer to Appendix~\ref{app:llama-gsm8k-humaneval}.

\xhdr{Experimental setup} 
In our experiments, for each multiple choice question in the MMLU benchmark dataset, we provide the question itself together with the available options ($4$ for each question, indexed from A to D) as an input prompt to the LLMs. 
Further, we instruct the LLMs to generate an output sequence comprising only the index of the selected option through a system prompt---refer to Appendix~\ref{app:add_exp} for the exact prompt. 
To evaluate the outputs provided by each LLM, we use a binary score $R \in \{0,1\}$, which indicates whether the LLM output is the (single) correct ($R=1$) or incorrect $(R=0)$ answer of the given options.
To obtain reliable conclusions,
we experiment with each multiple choice question $10$ times, each time using a (different) random seed to generate the noise variables used by the sampler. 
%
Due to space constraints, in what follows, we compare \texttt{1B} and \texttt{3B} on the knowledge area ``college computer science''. In Appendix~\ref{app:llama-mmlu}, we provide qualitatively similar results on other knowledge areas, different sampling parameters, and other LLMs.  

\xhdr{Results} 
Figures~\ref{fig:computer-sc-cov-1B-3B} and~\ref{fig:computer-sc-var-1B-3B} show that the scores of the LLMs are positively correlated under coupled generation and thus the variance of the difference in scores is lower under coupled generation than under independent, in agreement with Proposition~\ref{prop:variance}.
Further, we compute the error in the estimation of the expected difference in scores resulting from using the two approaches as a function of the available sample size. To this end, we first estimate the expected score difference using $1{,}000$ samples and consider this as (a proxy of) the ground truth. Then, we compute the absolute estimation error achieved by independent and coupled generation while sub-sampling the original samples across various sample sizes.
Figure~\ref{fig:computer-sc-err-1B-3B} summarizes the results, which show that, as expected from our theoretical analysis, a lower variance of the difference in scores under coupled generation leads to a reduction in the number of samples required to achieve equivalent error in the estimation of the expected difference between the scores of the LLMs.
Perhaps surprisingly, we find that this reduction can, in practice, be quite large. For example, to achieve an estimation error of $\approx$$0.034$, coupled generation needs $40$\% fewer samples than independent generation. 
In Appendix~\ref{app:llama-mmlu}, we find that this reduction is up to a striking $75\%$ 
for sufficiently similar LLMs in agreement with Proposition~\ref{prop:var_gumbel}.
Finally, it is worth noting that, when the models under comparison are very different, there is essentially zero correlation between their evaluation scores and, hence, independent and coupled generation become effectively equivalent; see, for example, Figure~\ref{fig:human-eval-second-5} in Appendix~\ref{app:llama-gsm8k-humaneval}.

\subsection{Evaluation on Pairwise Comparisons}
\label{sec:lmsys}
%

%
Here, we compare the same LLMs as in the previous section using pairwise comparisons between their outputs by a strong LLM, when prompted with open-ended questions from the LMSYS Chatbot Arena platform~\citep{lmsys2023chatbot}. 
%
Our goal is to investigate to what extent the theoretical results derived in Section~\ref{sec:pairwise} generalize to a commonly used evaluation protocol (\ie, LLM-as-a-judge), and not to what extent the rankings derived using a strong LLM are consistent with human preferences~\citep{wang2023large}.

\xhdr{Experimental setup} %
We experiment with $500$ questions from the LMSYS-Chat-1M dataset~\citep{zheng2024lmsys}. We provide the question itself as an input prompt to the LLMs, and instruct them to generate a concise response as an output through a system prompt.
Further, similarly as elsewhere~\citep{chatzi2024prediction,li2024llms,boyeau2024autoeval,gera2024justrank,li2023generative,zheng2023judging}, we use a strong LLM, namely, \texttt{GPT-4o-2024-11-20}, as a judge. 
More specifically, for each question and pair of outputs provided by two different LLMs, we prompt the judge to respond which of the two outputs it prefers, but allowing the judge to declare a tie---for the exact prompts we use, refer to Appendix~\ref{app:add_exp}.
Given these pairwise comparisons, to evaluate the outputs provided by each LLM, we use the win-rate achieved by each LLM against each other.
To obtain reliable conclusions, similarly as in the previous section, we repeat each experiment $10$ times, each time using a (different) random seed to generate the Gumbel noise variables used by the Gumbel-Max SCM.

\xhdr{Results}
We find that the empirical win-rate of each LLM against any other LLM is generally lower under coupled generation than under independent generation, as shown in Figure~\ref{fig:win_rate} for \texttt{bnb-8bit} and Figure~\ref{fig:lmsys-all-llms} in Appendix~\ref{app:lmsys} for other LLMs.
Moreover, whenever the LLMs under comparison are \textit{sufficiently} similar, the difference between win-rates is statistically significant, suggesting that our theoretical results may generalize beyond the canonical setting discussed in Section~\ref{sec:pairwise}. 
We hypothesize that this is partially due to an increase in the number of ties under coupled autoregressive generation. For example, for 
\texttt{bnb-8bit},
we observe a $24\%$, $11\%$, $15\%$ increase in the number of ties in the pairwise comparisons against
\texttt{8B},
\texttt{bnb-4bit}, and 
\texttt{AWQ-INT4}. 
Remarkably, the difference in empirical win-rates 
leads to differences in the rankings derived from the average win-rates, as shown in Figure~\ref{tab:ranking}. 
Under independent generation, the average win-rates achieved by
\texttt{8B}
and \texttt{bnb-8bit}
are statistically indistinguishable and they are both ranked at the top.
However, under coupled generation, only \texttt{8B} 
is ranked at the top.

\section{Discussion and Limitations}
\label{sec:discussion}
In this section, we discuss several aspects of our work which we believe are important to consider and may serve as a basis for future research. 

\xhdr{Model assumptions}
Our theoretical analysis of coupled autoregressive generation focuses on sampling mechanisms that satisfy counterfactual stability~\citep{oberst2019counterfactual}. Although counterfactual stability has been shown to be a desirable property for causal mechanisms in SCMs and, more specifically, for causal mechanisms used for sampling in LLMs~\citep{chatzi2025counterfactual}, counterfactual stability may not always be appropriate and should be justified by domain specific knowledge~\citep{haugh2023counterfactual}. 
In this context, it is worth mentioning that~\citet{chatzi2025counterfactual} have empirically shown that, both under the counterfactually stable Gumbel-Max sampling mechanism and a non-counterfactually stable mechanism based on inverse transform sampling, controlling for the randomness in two autoregressive processes with sufficiently similar next-token distributions yields more similar outputs.
Based on this evidence, we also expect that, even if a sampling scheme does not satisfy counterfactual stability, coupled generation may still reduce variance or alter model rankings as long as the LLMs under comparison are not very different, but to a lesser extent. 
It would be interesting to understand the sensitivity of coupled autoregressive generation to the specific choice of the Gumbel-Max SCM as well as extending our analysis to sampling mechanisms satisfying properties other than counterfactual stability~\citep{vlontzos2023estimating}. 
%
%

\xhdr{Practical considerations}
Our experimental results and theoretical analysis suggest that coupled autoregressive generation is most advantageous over independent autoregressive generation whenever the LLMs under comparison are sufficiently close in terms of their next-token distributions.
%
It would be important to identify which parts of the LLM development pipeline (\eg, the LLMs' architectures, training data, or fine-tuning process) lead, in practice, to sufficiently small changes in the next-token distributions for coupled autoregressive generation to be most beneficial.

Throughout our paper, we have focused on comparisons of models sharing the same token vocabulary, which include important practical use cases. 
For example, during the rapid and continual development of a large language model, one may like to compare different model versions whose performance differences are not easily predictable, such as those resulting from small architectural changes, data preprocessing, and supervised fine-tuning. 
Nevertheless, there are also many practical use cases where one may want to compare models using different token vocabularies, for which our methodology could, in principle, be extended leveraging very recent methods in the field of tokenization~\citep{vieira2024language,zheng2025broken}.
More specifically, 
one could transform two models with different token vocabularies into equivalent character-level models over a shared vocabulary of tokens corresponding exclusively to characters, and then apply coupled generation on the derived character-level models.
However, this would bring additional challenges. 
For example, to compute the next-character distributions, the above methods use approximations to marginalize over all possible suffixes of a string and thus the relative performance of the original models might not be preserved.

\xhdr{Evaluation}
We have experimented with (i) a single dataset of prompts for pairwise comparisons (\ie, LMSYS Chatbot Arena), where we have focused on win-rate as an evaluation metric, and (ii) we have used a strong LLM as a judge (\ie, \texttt{GPT-4o-2024-11-20}), which may introduce biases and lead to rankings that are inconsistent with (the distribution of) human preferences~\citep{wang2023large}.
To better understand the benefits of coupled autoregressive generation, it would be important to experiment with additional datasets, pairwise comparisons made by humans, and additional evaluation metrics based on, \eg, the Elo rating system~\citep{askell2021general,dettmers2024qlora,bai2022training,wu2023chatarena,lin2023llm} and the B-T model~\citep{chiang2024chatbot,boyeau2024autoeval}.

\section{Conclusions}
\label{sec:conclusions}
We have introduced a causal model of coupled autoregressive generation that enables the evaluation and comparison of different LLMs under the same source of randomness.
We have theoretically and empirically shown that, in evaluations based on benchmark datasets, coupled autoregressive generation can reduce the number of samples required to reliably compare the performance of LLMs and, in evaluations based on pairwise comparisons, it can lead to different and, perhaps more intuitive, rankings of LLMs in comparison with independent autoregressive generation. 

\vspace{2mm}
\subsubsection*{Acknowledgements}
Gomez-Rodriguez acknowledges support from the European Research Council (ERC) under the European Union'{}s Horizon 2020 research and innovation programme (grant agreements No. 945719 and 101169607).
Tsirtsis acknowledges support from the Alexander von Humboldt Foundation in the framework of the Alexander von Humboldt Professorship (Humboldt Professor of Technology and Regulation awarded to Sandra Wachter) endowed by the Federal Ministry of Education and Research via the Hasso Plattner Institute.
Straitouri acknowledges support from a Google PhD Fellowship. 
{ 
\small
\bibliographystyle{unsrtnat}
\bibliography{counterfactual-llm-evaluation}

@article{dubey2024llama,
  author       = {Abhimanyu Dubey et al.},
  title        = {The {Llama} 3 Herd of Models},
  journal      = {arXiv preprint arXiv:2407.21783},
  year         = {2024}
}

@inproceedings{chatzi2025counterfactual,
  author    = {Ivi Chatzi and Nina L Corvelo Benz and Eleni Straitouri and Stratis Tsirtsis and Manuel Gomez{-}Rodriguez},
  title     = {Counterfactual Token Generation in Large Language Models},
  booktitle = {Proceedings of the 4th Conference on Causal Learning and Reasoning},
  series    = {Proceedings of Machine Learning Research},
  volume    = {275},
  pages     = {1291--1315},
  publisher = {PMLR},
  year      = {2025}
}

@book{pearl2009causality,
  title={Causality},
  author={Pearl, Judea},
  year={2009},
  publisher={Cambridge university press}
}

@book{peters2017elements,
  title={Elements of causal inference: foundations and learning algorithms},
  author={Peters, Jonas and Janzing, Dominik and Sch{\"o}lkopf, Bernhard},
  year={2017},
  publisher={The MIT Press}
}

@inproceedings{chiang2024chatbot,
author = {Chiang, Wei-Lin and Zheng, Lianmin and Sheng, Ying and Angelopoulos, Anastasios N. and Li, Tianle and Li, Dacheng and Zhu, Banghua and Zhang, Hao and Jordan, Michael I. and Gonzalez, Joseph E. and Stoica, Ion},
title = {Chatbot arena: an open platform for evaluating {LLMs} by human preference},
year = {2025},
booktitle = {Proceedings of the International Conference on Machine Learning}
}

@article{bradley1952rank,
  title={Rank analysis of incomplete block designs: I. The method of paired comparisons},
  author={Bradley, Ralph Allan and Terry, Milton E},
  journal={Biometrika},
  volume={39},
  number={3/4},
  pages={324--345},
  year={1952},
  publisher={JSTOR}
}

@book{luce1959individual,
  title={Individual choice behavior},
  author={Luce, R Duncan},
  volume={4},
  year={1959},
  publisher={Wiley New York}
}

@article{thurstone1927law,
	author = {Louis L Thurstone},
	doi = {10.1037/h0070288},
	journal = {Psychological Review},
	number = {4},
	pages = {273--286},
	title = {A Law of Comparative Judgment},
	volume = {34},
	year = {1927}
}

@inproceedings{pearl1994probabilistic,
  title={A probabilistic calculus of actions},
  author={Pearl, Judea},
  booktitle={Proceedings of the Annual Conference on Uncertainty in Artificial Intelligence},
  year={1994}
}

@article{cobbe2021gsm8k,
  title={Training Verifiers to Solve Math Word Problems},
  author={Cobbe, Karl and Kosaraju, Vineet and Bavarian, Mohammad and Chen, Mark and Jun, Heewoo and Kaiser, Lukasz and Plappert, Matthias and Tworek, Jerry and Hilton, Jacob and Nakano, Reiichiro and Hesse, Christopher and Schulman, John},
  journal={arXiv preprint arXiv:2110.14168},
  year={2021}
}

@inproceedings{hendrycks2021measuring,
  author       = {Dan Hendrycks and
                  Collin Burns and
                  Steven Basart and
                  Andy Zou and
                  Mantas Mazeika and
                  Dawn Song and
                  Jacob Steinhardt},
  title        = {Measuring Massive Multitask Language Understanding},
  booktitle    = {Proceedings of the International Conference on Learning Representations},
  year         = {2021}
}

@inproceedings{oberst2019counterfactual,
  title={Counterfactual off-policy evaluation with {Gumbel-Max} structural causal models},
  author={Oberst, Michael and Sontag, David},
  booktitle={Proceedings of the International Conference on Machine Learning},
  pages={4881--4890},
  year={2019},
  organization={PMLR}
}

@inproceedings{tsirtsis2021counterfactual,
  title={Counterfactual explanations in sequential decision making under uncertainty},
  author={Tsirtsis, Stratis and De, Abir and Rodriguez, Manuel},
  booktitle={Advances in Neural Information Processing Systems},
  year={2021}
}

@inproceedings{noorbakhsh2022counterfactual,
  title={Counterfactual temporal point processes},
  author={Noorbakhsh, Kimia and Gomez-Rodriguez, Manuel},
  booktitle={Advances in Neural Information Processing Systems},
  year={2022}
}

@inproceedings{benz2022counterfactual,
  title={Counterfactual inference of second opinions},
  author={Benz, Nina L Corvelo and Gomez-Rodriguez, Manuel Gomez},
  booktitle={Proceedings of the Annual Conference on Uncertainty in Artificial Intelligence},
  year={2022}
}

@article{ravfogel2024counterfactual,
  title={Counterfactual Generation from Language Models},
  author={Ravfogel, Shauli and Svete, Anej and Sn{\ae}bjarnarson, V{\'e}steinn and Cotterell, Ryan},
  journal={arXiv preprint arXiv:2411.07180},
  year={2024}
}

@article{miller2024adding,
  title={Adding Error Bars to Evals: A Statistical Approach to Language Model Evaluations},
  author={Miller, Evan},
  journal={arXiv preprint arXiv:2411.00640},
  year={2024}
}

@article{paszke2019pytorch,
  title={Pytorch: An imperative style, high-performance deep learning library},
  author={Paszke, Adam and Gross, Sam and Massa, Francisco and Lerer, Adam and Bradbury, James and Chanan, Gregory and Killeen, Trevor and Lin, Zeming and Gimelshein, Natalia and Antiga, Luca and others},
  journal={Advances in Neural Information Processing Systems},
  volume={32},
  year={2019}
}

@inproceedings{haugh2023counterfactual,
  title={Counterfactual Analysis in Dynamic Latent State Models},
  author={Haugh, Martin B and Singal, Raghav},
  booktitle={Proceedings of the International Conference on Machine Learning},
  year={2023}
}

@article{chang2024asurvey,
author = {Chang, Yupeng and Wang, Xu and Wang, Jindong and Wu, Yuan and Yang, Linyi and Zhu, Kaijie and Chen, Hao and Yi, Xiaoyuan and Wang, Cunxiang and Wang, Yidong and Ye, Wei and Zhang, Yue and Chang, Yi and Yu, Philip S. and Yang, Qiang and Xie, Xing},
title = {{A Survey on Evaluation of Large Language Models}},
year = {2024},
journal = {ACM Transactions on Intelligent Systems and Technology},
publisher = {ACM}
}

@inproceedings{chatzi2024prediction,
  title={Prediction-Powered Ranking of Large Language Models},
  author={Chatzi, Ivi and Straitouri, Eleni and Thejaswi, Suhas and Rodriguez, Manuel Gomez},
  booktitle={Advances in Neural Information Processing Systems},
  year={2024}
}

@inproceedings{hendryckstest2021,
  title={{Measuring Massive Multitask Language Understanding}},
  author={Dan Hendrycks and Collin Burns and Steven Basart and Andy Zou and Mantas Mazeika and Dawn Song and Jacob Steinhardt},
  year={2021},
  booktitle={Proceedings of the International Conference on Learning Representations},
  publisher={ICLR}
}

@inproceedings{wang2022self,
    title = "{{{Self-Instruct}: Aligning Language Models with Self-Generated Instructions}}",
    author = "Wang, Yizhong  and
      Kordi, Yeganeh  and
      Mishra, Swaroop  and
      Liu, Alisa  and
      Smith, Noah A.  and
      Khashabi, Daniel  and
      Hajishirzi, Hannaneh",
    booktitle = "Proceedings of the Association for Computational Linguistics (Volume 1: Long Papers)",
    year = "2023",
    publisher = "ACL",
    pages = "13484--13508"
}

@inproceedings{ouyang2022training,
title={{Training Language Models to Follow Instructions with Human Feedback}},
author={Long Ouyang and Jeffrey Wu and Xu Jiang and Diogo Almeida and Carroll Wainwright and Pamela Mishkin and Chong Zhang and Sandhini Agarwal and Katarina Slama and Alex Gray and John Schulman and Jacob Hilton and Fraser Kelton and Luke Miller and Maddie Simens and Amanda Askell and Peter Welinder and Paul Christiano and Jan Leike and Ryan Lowe},
booktitle={Advances in Neural Information Processing Systems},
year={2022},
pages = {27730--27744},
publisher = {Curran Associates, Inc.}
}

@article{wang2023aligning,
  title={{Aligning Large Language Models with Human: A Survey}},
  author={Wang, Yufei and Zhong, Wanjun and Li, Liangyou and Mi, Fei and Zeng, Xingshan and Huang, Wenyong and Shang, Lifeng and Jiang, Xin and Liu, Qun},
  journal={arXiv preprint arXiv:2307.12966},
  year={2023}
}

@misc{taori2023stanford,
  title={{Stanford {Alpaca}: An instruction-following {LLaMA} model}},
  author={Taori, Rohan and Gulrajani, Ishaan and Zhang, Tianyi and Dubois, Yann and Li, Xuechen and Guestrin, Carlos and Liang, Percy and Hashimoto, Tatsunori B},
  year={2023},
  howpublished = {\url{https://github.com/tatsu-lab/stanford\_alpaca}} ,
  note = {Online; accessed 08 Aug 2025}
}

@inproceedings{zheng2023judging,
  title={{Judging {LLM}-as-a-{J}udge with {MT-B}ench and {Chatbot Arena}}}, 
  author={Lianmin Zheng and Wei-Lin Chiang and Ying Sheng and Siyuan Zhuang and Zhanghao Wu and Yonghao Zhuang and Zi Lin and Zhuohan Li and Dacheng Li and Eric P. Xing and Hao Zhang and Joseph E. Gonzalez and Ion Stoica},
  booktitle={Advances in Neural Information Processing Systems, data track},
  year={2023},
  pages={46595--46623},
  publisher = {Curran Associates, Inc.}
}

@inproceedings{li2023generative,
  title={{Generative Judge for Evaluating Alignment}},
  author={Li, Junlong and Sun, Shichao and Yuan, Weizhe and Fan, Run-Ze and Zhao, Hai and Liu, Pengfei},
  booktitle={Proceedings of the International Conference on Learning Representations},
  year={2024},
  publisher={ICLR}
}

@article{li2023prd,
  author    = {Ruosen Li and Teerth Patel and Xinya Du},
  title     = {{PRD: Peer Rank and Discussion Improve Large Language Model based Evaluations}},
  journal   = {{Transactions on Machine Learning Research}},
  year      = {2024},
  publisher = {ML Research Press},
}

@inproceedings{boubdir2023elo,
title={Elo Uncovered: Robustness and Best Practices in Language Model Evaluation},
author={Meriem Boubdir and Edward Kim and Beyza Ermis and Sara Hooker and Marzieh Fadaee},
booktitle={Advances in Neural Information Processing Systems},
year={2024}
}

@article{singhal2023large,
  title = {{Large Language Models Encode Clinical Knowledge}},
  volume = {620},
  number = {7972},
  journal = {Nature},
  publisher = {Nature Publishing Group},
  author = {Singhal,  Karan and Azizi,  Shekoofeh and Tu,  Tao and Mahdavi,  S. Sara and Wei,  Jason and Chung,  Hyung Won and Scales,  Nathan and Tanwani,  Ajay and Cole-Lewis,  Heather and Pfohl,  Stephen and Payne,  Perry and Seneviratne,  Martin and Gamble,  Paul and Kelly,  Chris and Babiker,  Abubakr and Sch\"{a}rli,  Nathanael and Chowdhery,  Aakanksha and Mansfield,  Philip and Demner-Fushman,  Dina and Ag\"{u}era y Arcas,  Blaise and Webster,  Dale and Corrado,  Greg S. and Matias,  Yossi and Chou,  Katherine and Gottweis,  Juraj and Tomasev,  Nenad and Liu,  Yun and Rajkomar,  Alvin and Barral,  Joelle and Semturs,  Christopher and Karthikesalingam,  Alan and Natarajan,  Vivek},
  year = {2023},
  month = jul,
  pages = {172–180}
}

@article{askell2021general,
      title={{A General Language Assistant as a Laboratory for Alignment}}, 
      author={Amanda Askell and Yuntao Bai and Anna Chen and Dawn Drain and Deep Ganguli and Tom Henighan and Andy Jones and Nicholas Joseph and Ben Mann and Nova DasSarma and Nelson Elhage and Zac Hatfield-Dodds and Danny Hernandez and Jackson Kernion and Kamal Ndousse and Catherine Olsson and Dario Amodei and Tom Brown and Jack Clark and Sam McCandlish and Chris Olah and Jared Kaplan},
      year={2021},
      journal={arXiv preprint arXiv:2112.00861}
}

@inproceedings{dettmers2024qlora,
  title={{QLoRA: Efficient Finetuning of Quantized {LLMs}}},
  author={Dettmers, Tim and Pagnoni, Artidoro and Holtzman, Ari and Zettlemoyer, Luke},
  booktitle={Advances in Neural Information Processing Systems},
  year={2024},
  pages = {10088--10115},
  publisher = {Curran Associates, Inc.}
}

@article{bai2022training,
  title={{Training a Helpful and Harmless Assistant with Reinforcement Learning from Human Feedback}},
  author={Yuntao Bai and Andy Jones and Kamal Ndousse and Amanda Askell and Anna Chen and Nova DasSarma and Dawn Drain and Stanislav Fort and Deep Ganguli and Tom Henighan and Nicholas Joseph and Saurav Kadavath and Jackson Kernion and Tom Conerly and Sheer El-Showk and Nelson Elhage and Zac Hatfield-Dodds and Danny Hernandez and Tristan Hume and Scott Johnston and Shauna Kravec and Liane Lovitt and Neel Nanda and Catherine Olsson and Dario Amodei and Tom Brown and Jack Clark and Sam McCandlish and Chris Olah and Ben Mann and Jared Kaplan},
  journal={arXiv preprint arXiv:2204.05862},
  year={2022}
}

@software{wu2023chatarena,
  author = {Yuxiang Wu and Zhengyao Jiang and Akbir Khan and Yao Fu and Laura Ruis and Edward Grefenstette and Tim Rocktäschel},
  title = {{ChatArena: Multi-Agent Language Game Environments for Large Language Models}},
  year = {2023},
  publisher = {GitHub},
  journal = {GitHub repository},
  version = {0.1},
  howpublished = {\url{https://github.com/chatarena/chatarena}},
}

@inproceedings{lin2023llm,
    title = "{{LLM}-Eval: Unified Multi-Dimensional Automatic Evaluation for Open-Domain Conversations with Large Language Models}",
    author = "Lin, Yen-Ting  and Chen, Yun-Nung",
    booktitle = "Proceedings of the Workshop on {NLP} for Conversational {AI}",
    month = jul,
    year = "2023",
    publisher = "ACL",
    pages = "47--58",
}

@article{boyeau2024autoeval,
  title={{AutoEval Done Right: Using Synthetic Data for Model Evaluation}},
  author={Boyeau, Pierre and Angelopoulos, Anastasios N and Yosef, Nir and Malik, Jitendra and Jordan, Michael I},
  journal={arXiv preprint arXiv:2403.07008},
  year={2024}
}

@inproceedings{wang2023large,
  author    = {Peiyi Wang and Lei Li and Liang Chen and Zefan Cai and Dawei Zhu and Binghuai Lin and Yunbo Cao and Lingpeng Kong and Qi Liu and Tianyu Liu and Zhifang Sui},
  title     = {Large Language Models are not Fair Evaluators},
  booktitle = {Proceedings of the Association for Computational Linguistics},
  pages     = {9440--9450},
  publisher = {ACL},
  year      = {2024},
  doi       = {10.18653/v1/2024.acl-long.511}
}

@misc{chiang2023vicuna,
  title={{Vicuna: An Open-Source Chatbot Impressing {GPT}-4 with 90\%* {ChatGPT} Quality}},
  author={Chiang, Wei-Lin and Li, Zhuohan and Lin, Zi and Sheng, Ying and Wu, Zhanghao and Zhang, Hao and Zheng, Lianmin and Zhuang, Siyuan and Zhuang, Yonghao and Gonzalez, Joseph E. and Stoica, Ion and Xing, Eric P.},
  howpublished={\url{https://vicuna.lmsys.org}},
  note={Online; accessed 21 May 2024},
  year={2023}
}

@inproceedings{saadfalcon2023ares,
  author    = {Jon Saad-Falcon and Omar Khattab and Christopher Potts and Matei Zaharia},
  title     = {{ARES: An Automated Evaluation Framework for Retrieval-Augmented Generation Systems}},
  booktitle = {Proceedings of the Conference of the North {A}merican Chapter of the Association for Computational Linguistics: Human Language Technologies (Volume 1: Long Papers)},
  pages     = {338--354},
  year      = {2024},
  publisher = {ACL}
}

@inproceedings{bach2022promptsource,
    title = "{{P}rompt{S}ource: An Integrated Development Environment and Repository for Natural Language Prompts}",
    author = "Bach, Stephen  and
      Sanh, Victor  and
      Yong, Zheng Xin  and
      Webson, Albert  and
      Raffel, Colin  and
      Nayak, Nihal V.  and
      Sharma, Abheesht  and
      Kim, Taewoon  and
      Bari, M Saiful  and
      Fevry, Thibault  and
      Alyafeai, Zaid  and
      Dey, Manan  and
      Santilli, Andrea  and
      Sun, Zhiqing  and
      Ben-david, Srulik  and
      Xu, Canwen  and
      Chhablani, Gunjan  and
      Wang, Han  and
      Fries, Jason  and
      Al-shaibani, Maged  and
      Sharma, Shanya  and
      Thakker, Urmish  and
      Almubarak, Khalid  and
      Tang, Xiangru  and
      Radev, Dragomir  and
      Jiang, Mike Tian-jian  and
      Rush, Alexander",
    booktitle = "Proceedings of the Association for Computational Linguistics: System Demonstrations",
     month = may,
    year = "2022",
    publisher = "ACL",
    pages = "93--104",
}

@inproceedings{wei2022finetuned,
    title={{Finetuned Language Models are Zero-Shot Learners}},
    author={Jason Wei and Maarten Bosma and Vincent Zhao and Kelvin Guu and Adams Wei Yu and Brian Lester and Nan Du and Andrew M. Dai and Quoc V Le},
    booktitle={Proceedings of the International Conference on Learning Representations},
    year={2022},
    publisher={ICLR}
}

@inproceedings{talmor2019commonsense,
    title = "{{C}ommonsense{QA}: A Question Answering Challenge Targeting Commonsense Knowledge}",
    author = "Talmor, Alon  and
      Herzig, Jonathan  and
      Lourie, Nicholas  and
      Berant, Jonathan",
    booktitle = "Proceedings of the North {A}merican Chapter of the Association for Computational Linguistics: Human Language Technologies, Volume 1 (Long and Short Papers)",
    year = "2019",
    publisher = "ACL",
    pages = "4149--4158",
}

@inproceedings{mishra2022cross,
    title = "{Cross-Task Generalization via Natural Language Crowdsourcing Instructions}",
    author = "Mishra, Swaroop  and
      Khashabi, Daniel  and
      Baral, Chitta  and
      Hajishirzi, Hannaneh",
    booktitle = "Proceedings of the Association for Computational Linguistics (Volume 1: Long Papers)",
    month = may,
    year = "2022",
    address = "Dublin, Ireland",
    publisher = "ACL",
    pages = "3470--3487",
}

@article{chen2021evaluating,
  title={{Evaluating Large Language Models Trained on Code}},
  author={Mark Chen and Jerry Tworek and Heewoo Jun and Qiming Yuan and Henrique Ponde de Oliveira Pinto and Jared Kaplan and Harri Edwards and Yuri Burda and Nicholas Joseph and Greg Brockman and Alex Ray and Raul Puri and Gretchen Krueger and Michael Petrov and Heidy Khlaaf and Girish Sastry and Pamela Mishkin and Brooke Chan and Scott Gray and Nick Ryder and Mikhail Pavlov and Alethea Power and Lukasz Kaiser and Mohammad Bavarian and Clemens Winter and Philippe Tillet and Felipe Petroski Such and Dave Cummings and Matthias Plappert and Fotios Chantzis and Elizabeth Barnes and Ariel Herbert-Voss and William Hebgen Guss and Alex Nichol and Alex Paino and Nikolas Tezak and Jie Tang and Igor Babuschkin and Suchir Balaji and Shantanu Jain and William Saunders and Christopher Hesse and Andrew N. Carr and Jan Leike and Josh Achiam and Vedant Misra and Evan Morikawa and Alec Radford and Matthew Knight and Miles Brundage and Mira Murati and Katie Mayer and Peter Welinder and Bob McGrew and Dario Amodei and Sam McCandlish and Ilya Sutskever and Wojciech Zaremba},
  journal={arXiv preprint arXiv:2107.03374},
  year={2021}
}

@article{liang2023holistic,
title={{Holistic Evaluation of Language Models}},
author={Percy Liang and Rishi Bommasani and Tony Lee and Dimitris Tsipras and Dilara Soylu and Michihiro Yasunaga and Yian Zhang and Deepak Narayanan and Yuhuai Wu and Ananya Kumar and Benjamin Newman and Binhang Yuan and Bobby Yan and Ce Zhang and Christian Alexander Cosgrove and Christopher D Manning and Christopher Re and Diana Acosta-Navas and Drew Arad Hudson and Eric Zelikman and Esin Durmus and Faisal Ladhak and Frieda Rong and Hongyu Ren and Huaxiu Yao and Jue WANG and Keshav Santhanam and Laurel Orr and Lucia Zheng and Mert Yuksekgonul and Mirac Suzgun and Nathan Kim and Neel Guha and Niladri S. Chatterji and Omar Khattab and Peter Henderson and Qian Huang and Ryan Andrew Chi and Sang Michael Xie and Shibani Santurkar and Surya Ganguli and Tatsunori Hashimoto and Thomas Icard and Tianyi Zhang and Vishrav Chaudhary and William Wang and Xuechen Li and Yifan Mai and Yuhui Zhang and Yuta Koreeda},
journal={Transactions on Machine Learning Research},
year={2023},
publisher={JMLR}
}

@inproceedings{longpre2023flan,
  title = 	 {The Flan Collection: Designing Data and Methods for Effective Instruction Tuning},
  author =       {Longpre, Shayne and Hou, Le and Vu, Tu and Webson, Albert and Chung, Hyung Won and Tay, Yi and Zhou, Denny and Le, Quoc V and Zoph, Barret and Wei, Jason and Roberts, Adam},
  booktitle = {Proceedings of the International Conference on Machine Learning},
  pages = 	 {22631--22648},
  year = 	 {2023},
  month =	 {Jul},
  publisher = {PMLR}
}

@book{elo1966uscf,
  title={{The USCF Rating System: Its Development, Theory, and Applications}},
  author={Elo, Arpad E.},
  year={1966},
  publisher={United States Chess Federation}
}

@inproceedings{bertrand2023limitations,
  title={{On the Limitations of the Elo, Real-World Games are Transitive, Not Additive}},
  author={Bertrand, Quentin and Czarnecki, Wojciech Marian and Gidel, Gauthier},
  booktitle={Proceedings of the International Conference on Artificial Intelligence and Statistics},
  pages={2905--2921},
  year={2023},
  organization={PMLR}
}

@article{dorner2024limits,
  title = {Limits to Scalable Evaluation at the Frontier: {LLM} as Judge Won’t Beat Twice the Data},
  author = {Dorner, Florian E. and Nastl, Vivian Y. and Hardt, Moritz},
  journal = {arXiv preprint arXiv:2410.13341},
  year = {2024}
}

@article{li2024llms,
  title={{LLMs-as-Judges}: A Comprehensive Survey on {LLM}-based Evaluation Methods},
  author={Li, Haitao and Dong, Qian and Chen, Junjie and Su, Huixue and Zhou, Yujia and Ai, Qingyao and Ye, Ziyi and Liu, Yiqun},
  journal={arXiv preprint arXiv:2412.05579},
  year={2024}
}

@article{gera2024justrank,
  title={JuStRank: Benchmarking {LLM} Judges for System Ranking},
  author={Gera, Ariel and Boni, Odellia and Perlitz, Yotam and Bar-Haim, Roy and Eden, Lilach and Yehudai, Asaf},
  journal={arXiv preprint arXiv:2412.09569},
  year={2024}
}

@article{madaan2024quantifying,
  title={Quantifying Variance in Evaluation Benchmarks},
  author={Madaan, Lovish and Singh, Aaditya K and Schaeffer, Rylan and Poulton, Andrew and Koyejo, Sanmi and Stenetorp, Pontus and Narang, Sharan and Hupkes, Dieuwke},
  journal={arXiv preprint arXiv:2406.10229},
  year={2024}
}

@article{bengio2000neural,
  title={A neural probabilistic language model},
  author={Bengio, Yoshua and Ducharme, R{\'e}jean and Vincent, Pascal},
  journal={Advances in Neural Information Processing Systems},
  volume={13},
  year={2000}
}

@article{radford2019language,
  title={Language models are unsupervised multitask learners},
  author={Radford, Alec and Wu, Jeffrey and Child, Rewon and Luan, David and Amodei, Dario and Sutskever, Ilya and others},
  journal={OpenAI blog},
  volume={1},
  number={8},
  pages={9},
  year={2019}
}

@article{bubeck2023sparks,
  title={{Sparks of Artificial General Intelligence: Early experiments with {GPT}-4}},
  author={S{\'e}bastien Bubeck and Varun Chandrasekaran and Ronen Eldan and Johannes Gehrke and Eric Horvitz and Ece Kamar and Peter Lee and Yin Tat Lee and Yuanzhi Li and Scott Lundberg and Harsha Nori and Hamid Palangi and Marco Tulio Ribeiro and Yi Zhang},
  journal={arXiv preprint arXiv:2303.12712},
  year={2023}
}

@inproceedings{mozannar2022reading,
  title={{Reading Between the Lines: Modeling User Behavior and Costs in AI-Assisted Programming}},
  author={Mozannar, Hussein and Bansal, Gagan and Fourney, Adam and Horvitz, Eric},
  booktitle={Proceedings of the Conference on Human Factors in Computing Systems},
  year={2024},
  publisher = {Association for Computing Machinery}
}

@article{haupt2023ai,
  title={{AI-Generated Medical Advice—GPT and Beyond}},
  author={Haupt, Claudia E and Marks, Mason},
  journal={Journal of American Medical Association},
  volume={329},
  number={16},
  pages={1349--1350},
  year={2023},
  publisher={American Medical Association}
}

@article{romera2023mathematical,
  title = {{Mathematical Discoveries from Program Search with Large Language Models}},
  volume = {625},
  number = {7995},
  journal = {Nature},
  publisher = {Springer},
  author = {Romera-Paredes,  Bernardino and Barekatain,  Mohammadamin and Novikov,  Alexander and Balog,  Matej and Kumar,  M. Pawan and Dupont,  Emilien and Ruiz,  Francisco J. R. and Ellenberg,  Jordan S. and Wang,  Pengming and Fawzi,  Omar and Kohli,  Pushmeet and Fawzi,  Alhussein},
  year = {2023},
  pages = {468–475}
}

@article{vlontzos2023estimating,
  title={Estimating categorical counterfactuals via deep twin networks},
  author={Vlontzos, Athanasios and Kainz, Bernhard and Gilligan-Lee, Ciar{\'a}n M},
  journal={Nature Machine Intelligence},
  volume={5},
  number={2},
  pages={159--168},
  year={2023},
  publisher={Nature Publishing Group UK London}
}

@ARTICLE{9729603,
  author={Huijben, Iris A. M. and Kool, Wouter and Paulus, Max B. and van Sloun, Ruud J. G.},
  journal={IEEE Transactions on Pattern Analysis and Machine Intelligence}, 
  title={A Review of the {Gumbel-Max} Trick and its Extensions for Discrete Stochasticity in Machine Learning}, 
  year={2023},
  volume={45},
  number={2},
  pages={1353-1371},
  doi={10.1109/TPAMI.2022.3157042}
}

@misc{lmsys2023chatbot,
    author = "{LMSYS}",
    title = {{Chatbot Arena: Benchmarking {LLMs} in the Wild with Elo Ratings}},
    howpublished = {\url{https://lmsys.org/}},
    note = {Online; accessed 21 May 2024} ,
    year=2023,
}

@misc{bitsnbytes2024bits,
    author = "{Bits and Bytes Foundation}",
    title = {Bits and bytes quantisation library},
    howpublished = {\url{https://huggingface.co/docs/bitsandbytes/main/en/index}},
    note = {Online; accessed 08 Aug 2025} ,
    year=2024,
}

@misc{zheng2024lmsys,
      title={LMSYS-Chat-1M: A Large-Scale Real-World {LLM} Conversation Dataset}, 
      author={Lianmin Zheng and Wei-Lin Chiang and Ying Sheng and Tianle Li and Siyuan Zhuang and Zhanghao Wu and Yonghao Zhuang and Zhuohan Li and Zi Lin and Eric P. Xing and Joseph E. Gonzalez and Ion Stoica and Hao Zhang},
      year={2024},
      eprint={2309.11998},
      archivePrefix={arXiv},
      primaryClass={cs.CL}
}

@inproceedings{vieira2024language,
  title={From language models over tokens to language models over characters},
  author={Vieira, Tim and LeBrun, Ben and Giulianelli, Mario and Gastaldi, Juan Luis and DuSell, Brian and Terilla, John and O'Donnell, Timothy J and Cotterell, Ryan},
  booktitle={Proceedings of the International Conference on Machine Learning},
  year={2025}
}

@article{zheng2025broken,
  title={Broken Tokens? Your Language Model can Secretly Handle Non-Canonical Tokenizations},
  author={Zheng, Brian Siyuan and Liu, Alisa and Ahia, Orevaoghene and Hayase, Jonathan and Choi, Yejin and Smith, Noah A},
  journal={arXiv preprint arXiv:2506.19004},
  year={2025}
}

@article{peeperkorn2024temperature,
  title={Is temperature the creativity parameter of large language models?},
  author={Peeperkorn, Max and Kouwenhoven, Tom and Brown, Dan and Jordanous, Anna},
  journal={arXiv preprint arXiv:2405.00492},
  year={2024}
}
}

\clearpage
\newpage

\appendix

\vspace{-8mm}

\section{FURTHER RELATED WORK} \label{app:further-related-work}
Our work also builds upon the rapidly increasing literature on evaluation and comparison of LLMs~\citep{chang2024asurvey}. 
Within this literature, LLMs are evaluated and compared using: 
(i) benchmark datasets with manually hand-crafted inputs and ground-truth outputs~\citep{bach2022promptsource,wei2022finetuned,talmor2019commonsense,mishra2022cross,chen2021evaluating,liang2023holistic,longpre2023flan} and (ii) the level of alignment with human preferences, as elicited by means of pairwise comparisons~\citep{wang2023aligning,ouyang2022training,taori2023stanford,zheng2023judging,li2023generative,li2023prd,boubdir2023elo,singhal2023large,chiang2024chatbot}.
However, it has become 
clear that oftentimes rankings derived from benchmark datasets do not match those derived from human preferences~\citep{zheng2023judging,li2023generative,li2023prd,chiang2023vicuna,chiang2024chatbot}.
Within the literature on ranking LLMs from pairwise comparisons, most studies use the Elo rating system~\citep{askell2021general,dettmers2024qlora,bai2022training,wu2023chatarena,lin2023llm}, originally introduced for chess tournaments~\citep{elo1966uscf}. 
However, Elo-based rankings are sensitive to the order of pairwise comparisons, as newer comparisons have more weight than older ones, 
which leads to unstable rankings~\citep{boubdir2023elo}.
To address this limitation, several studies have instead used the Bradley-Terry (B-T) model~\citep{chiang2024chatbot,boyeau2024autoeval}, which weighs pairwise comparisons equally regardless of their order. 
Nevertheless, both the Elo rating system and the B-T model have faced criticism, 
as pairwise comparisons often fail to satisfy the fundamental axiom of transitivity, 
upon which both approaches rely~\citep{boubdir2023elo,bertrand2023limitations}. 
Recently, several studies have used the win-rate~\citep{zheng2023judging,chiang2024chatbot,boyeau2024autoeval}, which weighs comparisons equally regardless of their order and 
does not require the transitivity assumption.
In our work, we focus on win-rates. However, 
it may be possible to extend our theoretical and empirical results to rankings based on Elo ratings and the B-T model.


\section{FORMAL DEFINITION OF COUNTERFACTUAL STABILITY}\label{app:stability}

Counterfactual stability is a desirable property of SCMs~\citep{oberst2019counterfactual} that has previously been used in the context of autoregressive generation of LLMs~\citep{chatzi2025counterfactual}. In the following, we provide its formal definition along with a simple example to explain the intuition behind it. Throughout this section, $\PP^{\Ccal \,;\, do(\cdot)}$ denotes the probability of the interventional distribution entailed by an SCM $\Ccal$ under an intervention $do(\cdot)$.
Moreover, $\PP^{\Ccal \,|\, \star \,;\, do(\cdot)}$ denotes the probability of the counterfactual distribution entailed by an SCM $\Ccal$ under an intervention $do(\cdot)$ given that an observed event $\star$ has already occurred.

\begin{definition}
    A sampling mechanism defined by $f_T$ and $P_U$ satisfies counterfactual stability if for all LLMs $m,m'\in \Mcal$, $i\in\{1,2,\ldots,K\}$ and tokens $t_1,t_2 \in V$ with $t_1\neq t_2$, the condition 
    \begin{equation}\label{eq:stability_condition}
        \frac{\PP^{\Ccal \,;\, do(M=m')}[T_i=t_1 \given D_i]}{\PP^{\Ccal \,;\, do(M=m)}[T_i=t_1 \given D_i]} \geq \frac{\PP^{\Ccal \,;\, do(M=m')}[T_i=t_2 \given D_i]}{\PP^{\Ccal \,;\, do(M=m)}[T_i=t_2 \given D_i]}
    \end{equation}
    implies that $\PP^{\Ccal \,|\, D_i,M=m,T_i=t_1 \,;\, do(M=m')}[T_i=t_2]=0$.
\end{definition}

The property of counterfactual stability has an intuitive interpretation that can be best understood via a simple example. Assume that the vocabulary contains $2$ tokens ``A'' and ``B'' and, using LLM $m$, the next-token distribution at a time step $i$ assigns values $0.6$, $0.4$ to the two tokens, respectively. Moreover, the realized noise value $\ub_i$ is such that the token ``A'' is sampled.
Now, consider that, while keeping the noise value $\ub_i$ fixed, we change the LLM to $m'$, resulting in a next-token distribution that assigns values $0.7$, $0.3$ to the two tokens, respectively. Counterfactual stability ensures that, since the noise value $\ub_i$ led to ``A'' being sampled under $m$ at $0.6$ to $0.4$ odds, the same value
cannot lead to ``B'' being sampled under $m'$ where its relative odds are lower (\ie, $0.3$ to $0.7$).

\section{ADDITIONAL THEORETICAL RESULTS}\label{app:canonical}

\subsection{Evaluation based on Pairwise Comparison}
\label{app:canonical-pairwise-comparisons}

In this Section, we consider another canonical setting in which, for each prompt, the response can be one of any single-token sequences, and each of the sequences may provide a different level of user'{}s satisfaction (\ie, achieve a different score). 
Further, the LLMs under comparison always output one of them as a response and the sampling mechanism used by the LLMs is given by the Gumbel-Max SCM as defined in Footnote~\ref{fn:gumbel-max} in Section~\ref{sec:individual}.
The following proposition shows that the number of ties between an LLM $m$ and any other \emph{sufficiently similar} LLM $m' \neq m$ are higher under coupled autoregressive generation than under independent autoregressive generation:
\begin{proposition}\label{prop:gumbel_ties}
    Given a fixed prompt $s_q \sim P_{Q}$, 
    assume, without loss of generality, that $f_R(s_q \circ t_1) \geq f_R(s_q \circ t_2) \geq \ldots \geq f_R(s_q \circ t_{|V|})$. 
    Let $m$ be an LLM that assigns positive probability to every single-token sequence and zero probability to any other sequence. 
    If the sampling mechanism defined by $f_T$ and $P_U$ is given by the Gumbel-Max SCM, then, there exists a constant $\varepsilon(m)>0$ such that, for every LLM $m'$ that assigns positive probability to every single-token sequence and zero probability to any other sequence and satisfies $d(m,m')=\sup_{s_q} \norm{f_D(s_q,m)-f_D(s_q, m')}_\infty < \varepsilon(m)$, it holds that
    \begin{equation*}
    \EE_{\Ub \sim P_{\Ub}}[\one\{R_m(\Ub, s_q)=R_{m'}(\Ub, s_q)\}] > \EE_{\Ub, \Ub' \sim P_{\Ub}}[\one\{R_m(\Ub, s_q)=R_{m'}(\Ub', s_q)\}].
    \end{equation*}
\end{proposition}
%
The above proposition implies that the win-rates under independent and coupled autoregressive generation are different and, similarly as in the first canonical setting, rankings derived from the win-rates may differ under independent and coupled autoregressive generation. We investigate this further in our experiments in Section~\ref{sec:experiments}.

\section{PROOFS} \label{sec:proofs}

\subsection{Proof of Proposition~\ref{prop:variance}}
 We can rewrite the variance of the difference in scores under independent generation in terms of the variance of the difference in scores under coupled generation as follows:
    \begin{align*}
        \Var[R_m(\Ub,S_q)-R_{m'}(\Ub',S_q)] ] 
        &= \Var[R_m(\Ub,S_q)-R_{m'}(\Ub,S_q) + R_{m'}(\Ub,S_q) -R_{m'}(\Ub',S_q)] ] \nonumber
        \\ &= \Var[R_m(\Ub,S_q)-R_{m'}(\Ub,S_q)]  + \Var[R_{m'}(\Ub,S_q)-R_{m'}(\Ub',S_q)] \nn
            \\ & \qquad \quad + 2 \cdot \Cov[R_m(\Ub,S_q)-R_{m'}(\Ub,S_q),
        R_{m'}(\Ub,S_q)-R_{m'}(\Ub',S_q) ]. 
    \end{align*}
For the variance of the difference in scores for the same LLM under independent noise values, we have that
    \begin{align*}
        \Var[R_{m'}(\Ub,S_q)-R_{m'}(\Ub',S_q)] 
        & \overset{(a)}{=} \EE[(R_{m'}(\Ub,S_q)-R_{m'}(\Ub',S_q))^2] 
        -
         \EE[R_{m'}(\Ub,S_q)-R_{m'}(\Ub',S_q)]^2 
        \\& \overset{(b)}{=} \EE[R_{m'}(\Ub,S_q)^2
        -2\cdot R_{m'}(\Ub,S_q)R_{m'}(\Ub',S_q) + R_{m'}(\Ub',S_q)^2] 
        \\& \overset{(c)}{=} 2 \cdot \EE[R_{m'}(\Ub,S_q)^2] - 2 \cdot \EE[R_{m'}(\Ub,S_q)R_{m'}(\Ub',S_q)], 
    \end{align*}
where (a) holds by the definition of variance, (b) is due to the subtraction term being $0$, and (c) is due to the linearity of expectation.
Further, for the covariance of the difference in scores under independent generation and the difference in scores under coupled generation, we have that
\begingroup
\allowdisplaybreaks
\begin{align*}
    \Cov[R_m&(\Ub,S_q)-R_{m'}(\Ub,S_q), R_{m'}(\Ub,S_q)-R_{m'}(\Ub',S_q)]\nonumber
    \\ 
    &\overset{(a)}{=} \EE[\left(R_m(\Ub,S_q)-R_{m'}(\Ub,S_q)\right) \cdot \left( R_{m'}(\Ub,S_q)-R_{m'}(\Ub',S_q) \right)] 
    \\ &\qquad \quad - \EE[R_m(\Ub,S_q)-R_{m'}(\Ub,S_q)]\cdot \EE[R_{m'}(\Ub,S_q)-R_{m'}(\Ub',S_q)]
    \\ 
    &\overset{(b)}{=} \EE[R_m(\Ub,S_q)R_{m'}(\Ub,S_q)] - \EE[R_m(\Ub,S_q)R_{m'}(\Ub',S_q)]
    \\ &\qquad \quad - \EE[R_{m'}(\Ub,S_q)R_{m'}(\Ub,S_q)] + \EE[R_{m'}(\Ub,S_q)R_{m'}(\Ub',S_q)] 
    \\ 
    & \overset{(c)}{=} \Cov[R_m(\Ub,S_q),R_{m'}(\Ub,S_q)] - \EE[R_{m'}(\Ub,S_q)^2] 
    + \EE[R_{m'}(\Ub,S_q)R_{m'}(\Ub',S_q)]] 
\end{align*}
\endgroup
where (a) and (c) hold by the definition of covariance and (b) is due to the last term being zero and by the expansion of the first term.

Putting all the above results together, it follows that
\begin{align*}
    \Var[R_m(\Ub,S_q)-R_{m'}(\Ub',S_q)] ] 
        & = \Var[R_m(\Ub,S_q)-R_{m'}(\Ub,S_q)] 
        + 2\cdot \Cov\left[R_m(\Ub,S_q),R_{m'}(\Ub,S_q)\right] \\
         &  \qquad \quad- 2\cdot \EE[R_{m'}(\Ub,S_q)^2] + 2\cdot \EE[R_{m'}(\Ub,S_q)R_{m'}(\Ub',S_q)] 
         \\
        & \qquad \quad
         + 2 \cdot \EE[R_{m'}(\Ub,S_q)^2] - 2 \cdot \EE[R_{m'}(\Ub,S_q)R_{m'}(\Ub',S_q)]
        \\&= \Var[R_m(\Ub,S_q)-R_{m'}(\Ub,S_q)] + 2\cdot \Cov[R_m(\Ub,S_q),R_{m'}(\Ub,S_q)] 
\end{align*}
which concludes the proof.

\subsection{Proof of Proposition~\ref{prop:var_stability}}

Due to Proposition~\ref{prop:variance}, to show that Eq.~\ref{eq:binary_var} holds, it suffices to show that the covariance between the scores of the different LLMs under coupled generation is non-negative, \ie, $\Cov[R_m(\Ub,S_q),R_{m'}(\Ub,S_q)]> 0$.

To this end, we first rewrite the covariance as
\begin{align}
    \Cov[R_m&(\Ub,S_q),R_{m'}(\Ub,S_q)] \\ &= \PP[R_m(\Ub,S_q)=1, R_{m'}(\Ub,S_q)=1] \nn -\PP[R_m(\Ub,S_q)=1]\cdot \PP[R_{m'}(\Ub,S_q)=1] \nn \\ 
    &=\sum_{s_q} \PP[S_q=s_q] \cdot \left( \PP[R_m(\Ub,s_q)=1, R_{m'}(\Ub,s_q)=1] \right.  - \left. \PP[R_m(\Ub,s_q)=1]\cdot \PP[R_{m'}(\Ub,s_q)=1] \right) \label{eq:covariance_rewrite}
\end{align}
Next, we note that the event $R_m(\Ub,s_q)=1$ is equivalent to LLM $m$ sampling the ground truth token for prompt $s_q$. 
Without loss of generality, assume $t_1$ is the ground truth token, \ie, $\textsc{c}(s_q)=t_1$. 
Then, since only tokens $\{t_1, t_2\}$ have positive probability under $m$ and $m'$, it must hold that either (i) one LLM assigns a greater probability to $t_1$ and the other LLM assigns a greater probability to $t_2$, 
or (ii) both LLMs assign the same probabilities. 
Further, since the sampling mechanism defined by $f_T$ and $P_U$ satisfies counterfactual stability, we have that the condition in Eq.~\ref{eq:stability_condition} holds in both (i) and (ii) and, under coupled generation, the LLM with greater (or equal) probability for $t_1$ will always sample $t_1$ when the LLM with lower (or equal) probability does. 
This implies that
\begin{equation}
    \PP[R_m(\Ub,s_q)=1, R_{m'}(\Ub,s_q)=1] = \min\{ \PP[R_m(\Ub,s_q)=1], \PP[R_{m'}(\Ub,s_q)=1]\}
\end{equation}
Finally, since it holds that
\begin{equation}
    \min\{ \PP[R_m(\Ub,s_q)=1], \PP[R_{m'}(\Ub,s_q)=1]\} > \PP[R_m(\Ub,s_q)=1] \PP[R_{m'}(\Ub,s_q)=1]
\end{equation}
because $\PP[R_m(\Ub,s_q)=1] \in (0,1)$ and $\PP[R_{m'}(\Ub,s_q)=1]\in(0,1)$ by assumption,
we can conclude from Eq.~\ref{eq:covariance_rewrite} that
    \begin{equation}
        \begin{split}
            &\Cov[R_m(\Ub,S_q),R_{m'}(\Ub,S_q)] > 0.
        \end{split}
    \end{equation}
%
    
\subsection{Proof of Proposition~\ref{prop:var_gumbel}}
Using Proposition~\ref{prop:variance}, we have that
    \begin{align*}
            \Cov[R_m&(\Ub,S_q),R_{m'}(\Ub,S_q)] \\ 
            &= \EE[R_m(\Ub,S_q)\cdot R_{m'}(\Ub,S_q)]-\EE[R_m(\Ub,S_q)]\cdot  \EE[R_{m'}(\Ub,S_q)]\\
            &= \underbrace{\PP[R_m(\Ub,S_q)=1, R_{m'}(\Ub,S_q)=1
            ]}_{(i)} -\underbrace{\PP[R_m(\Ub,S_q)=1) \cdot \PP[R_{m'}(\Ub,S_q)=1]}_{(ii)}.
    \end{align*}
In the remainder of the proof, we will bound each term (i) and (ii) separately and, since $|\textsc{c}(s_q)|=1$ for all $s_q \sim P_{Q}$, assume without loss of generality that the correct token is single-token sequence $t_1$.

To bound the term (ii), first note that, using the definition of the Gumbel-Max SCM, we have that, for each $k \in \{2, \ldots, |V|\}$, it holds that
\begin{align*}
            R_{m}(\Ub,s_q)=1 &\iff U_1 + \log ( [f_D(s_q,m)]_{t_1} ) \geq U_k + \log ( [f_D(s_q, m)]_{t_k} ), \\
            R_{m'}(\Ub,s_q)=1 &\iff U_1 + \log ( [f_D(s_q,m')]_{t_1} ) \geq U_k + \log ( [f_D(s_q, m')]_{t_k} ).
\end{align*}
Next, let $\varepsilon^*>0$ be an arbitrary constant that we will determine later such that
\begin{equation} \label{eq:bound}
    |\log ( [f_D(S_q,m)]_{t_k}) -\log ([f_D(S_q,m')]_{t_k}) |\leq \varepsilon^*,
\end{equation}
and note that since, by assumption, $D_{t_k} > 0$ for all $k \in \{1, \ldots, |V|\}$, any bound on the absolute difference of log-probabilities $|\log ( [f_D(S_q,m)]_{t_k}) -\log ([f_D(S_q,m')]_{t_k}) |$ uniformly implies a bound on the difference of probabilities $|[f_D(S_q,m)]_{t_k} -[f_D(S_q,m')]_{t_k} |$ and vice versa. 
For simplicity, we prove the result in the log-domain.

Now, using the bound defined by Eq.~\ref{eq:bound}, we have that
\begin{multline*}
        \bigcap_{k\neq1} \left\{ U_1 + \log ( [f_D(S_q, m')]_{t_1} ) \geq U_k + \log ( [f_D(S_q, m')]_{t_k} ) \right\}\\
            \subset \bigcap_{k\neq1} \left\{U_1 + \log ( [f_D(S_q, m)]_{t_1} )+\varepsilon^* \geq U_k + \log ( [f_D(S_q, m)]_{t_k} )-\varepsilon^* \right\},
\end{multline*}
and we can then bound the term (ii) as follows:
    \begin{align*}
            \PP[R_m(\Ub,S_q)&=1] \cdot \PP[R_{m'}(\Ub,S_q)=1] \\
            &=
            \PP[\cap_{k\neq1} \{U_1 + \log ( [f_D(S_q,m)]_{t_1} ) \geq U_k + \log ( [f_D(S_q,m)]_{t_k} )\}]\\
            & \qquad \times \PP[\cap_{k\neq1}\{ U_1 + \log ( [f_D(S_q,m')]_{t_1} ) \geq U_k + \log ( [f_D(S_q,m')]_{t_k} ) \}]\\
            &\leq \PP[\cap_{k\neq1} \{U_1 + \log ( [f_D(S_q,m)]_{t_1} ) \geq U_k + \log ( [f_D(S_q,m)]_{t_k} )\}]\\
            & \qquad \times \PP[\cap_{k\neq1} \{U_1 + \log ( [f_D(S_q,m)]_{t_1} )+\varepsilon^* \geq U_k + \log ( [f_D(S_q,m)]_{t_k} )-\varepsilon^*\}].
    \end{align*}
To bound the term (i), first note that, using the bound defined by Eq.~\ref{eq:bound}, we have that
    \begin{multline*}
        \bigcap_{k\neq1} \left\{U_1 + \log ( [f_D(S_q,m')]_{t_1} ) \geq U_k + \log ( [f_D(S_q,m')]_{t_k} ) \right\}\\
         \supset \bigcap_{k\neq1} \left\{U_1 + \log ( [f_D(S_q,m)]_{t_1} )-\varepsilon^* \geq U_k + \log ( [f_D(S_q,m)]_{t_k} )+\varepsilon^* \right\}.
    \end{multline*} 
Thus, we can bound the term (i) as follows:
    \begin{align*}
        \PP[R_m(\Ub&,S_q)=1, R_{m'}(\Ub,S_q)=1] \\=&\PP\Big[\cap_{k\neq1} \{U_1 + \log ( [f_D(S_q,m)]_{t_1} ) \geq U_k + \log ([f_D(S_q,m)]_{t_k} )\}\\
            & \qquad \,\,\, \cap \{ U_1 + \log ( [f_D(S_q,m')]_{t_1} )  \geq U_k + \log ( [f_D(S_q,m')]_{t_k} ) \}\Big] \\
             \geq & \PP\Big[\cap_{k\neq1} \{U_1 + \log ( [f_D(S_q,m)]_{t_1} ) \geq U_k + \log ( [f_D(S_q,m)]_{t_k} )\}\\
            & \qquad \,\,\, \cap \{ U_1 + \log ( [f_D(S_q,m)]_{t_1} ) \geq U_k + \log ( [f_D(S_q,m)]_{t_k} ) +2\varepsilon^* \}\Big] \\
            \overset{(a)}{=} &\sum_{s_q} \PP[S_q=s_q]\cdot \PP[\cap_{k\neq1} \{ U_1 + \log ( [f_D(S_q,m)]_{t_1} )  \geq U_k + \log ( [f_D(S_q,m)]_{t_k} ) +2\varepsilon^* \}],
    \end{align*}
where (a) follows from the fact that
\begin{multline*}
      \left\{ U_1 + \log ( [f_D(S_q,m)]_{t_1} ) \geq U_k + \log ( [f_D(S_q,m)]_{t_k} ) +2\varepsilon^* \right\} \\
      \subset \left\{ U_1 + \log ( [f_D(S_q,m)]_{t_1} ) \geq U_k + \log ( [f_D(S_q,m)]_{t_k} ) \right\}.
\end{multline*}
Now, note that, for $k \in \{2, \ldots, |V|\}$, the variable $X_k \equiv U_1-U_k \sim \text{Logistic}(0,1)$ (for $k=1$, define $X_k \equiv 0$). 
Therefore, we can rewrite the bound for (i) as
\begin{multline*}
      \PP[R_m(\Ub,S_q)=1, R_{m'}(\Ub,S_q)=1]\\
      \geq\sum_{s_q} \PP[S_q=s_q] \cdot \prod_{k\neq1}\cdot \PP[ \{ X_k \geq \log ( [f_D(S_q,m)]_{t_k} )-\log ( [f_D(S_q,m)]_{t_1} ) +2\varepsilon^* \}]
\end{multline*}
and we can rewrite the bound for (ii) as
\begin{multline*}
    \PP[R_m(\Ub,S_q)=1] \PP[R_{m'}(\Ub,S_q)=1]\leq \\
    \sum_{s_q}\PP[S_q=s_q]\cdot 
            \left\{
            \prod_{k\neq1}
            \PP[ \{ X_k \geq \log ( [f_D(S_q,m)]_{t_k} )-\log ( [f_D(S_q,m)]_{t_k} ) - 2\varepsilon^* \} ]
            \right\}\\
            \times \PP[\cap_{k\neq1} \{X_k \geq \log ( [f_D(S_q,m)]_{t_k} )-\log ( [f_D(S_q,m)]_{t_k} )\}].
\end{multline*}
As a consequence, to prove that $\PP[R_m(\Ub,S_q)=1, R_{m'}(\Ub,S_q)=1] > \PP[R_m(\Ub,S_q)=1] \PP[R_{m'}(\Ub,S_q)=1]$, it suffices to show that
    \begin{multline}\label{eq:condition-covariance-binary-scores}
            \sum_{s_q}P[S_q=s_q] \prod_{k\neq1}\cdot \PP[ \{ X_k \geq \log ( [f_D(S_q,m)]_{t_k} )-\log ( [f_D(S_q,m)]_{t_1} ) +2\varepsilon^* \}]\\
            > \sum_{s_q}\PP[S_q=s_q] \prod_{k\neq1}\cdot \PP[ \{ X_k \geq \log ( [f_D(S_q,m)]_{t_k} )-\log ( [f_D(S_q,m)]_{t_1} ) - 2\varepsilon^* \}]
            \\
            \times \PP[\cap_{k\neq1} \{X_k \geq \log ( [f_D(S_q,m)]_{t_k} )-\log ( [f_D(S_q,m)]_{t_1} )\}]
    \end{multline}
To do so, note that Eq.~\ref{eq:condition-covariance-binary-scores} holds trivially for $\varepsilon^*=0$ since
\begin{equation*}
      \PP[\cap_{k\neq1} \{X_k \geq \log ( [f_D(S_q,m)]_{t_k} )-\log ( [f_D(S_q,m)]_{t_1} )\}] < 1,
\end{equation*}
which is a fixed term independent of $m'$. Since all terms in Eq.~\ref{eq:condition-covariance-binary-scores} are continuous in $\varepsilon^*$, there exists $\varepsilon^*(m) >0 $, possibly dependent of $m$ but independent of $m'$, such that Eq.~\ref{eq:condition-covariance-binary-scores} holds if
\begin{equation*}
    \sup_{s_q} \norm{\log(f_D(s_q,m))-\log(f_D(s_q, m'))}_\infty < \varepsilon^*(m).
\end{equation*}
Since by assumption $D_t>0$ for all $t\in V$, there exists $\varepsilon(m)>0$ in probability space such that Eq.~\ref{eq:condition-covariance-binary-scores} holds if
\begin{equation*}
    \sup_{s_q} \norm{f_D(s_q,m)-f_D(s_q, m')}_\infty < \varepsilon(m).
\end{equation*}
This concludes the proof.

\subsection{Proof of Proposition~\ref{prop:gap_win_rates_stability}}

%
Under coupled autoregressive generation, if the LLM $m$ samples the preferred token $t_+$, then the LLM $m'$ must also sample $t_{+}$ because $t_{+}$ is more likely under $m'$ than under $m$ and the sampling mechanism defined by $f_T$ and $P_U$ satisfies counterfactual stability. 
This implies that the win-rate achieved by $m$ against $m'$ is
    \begin{equation}\label{eq:prop_greater_shared}
    \begin{split}
            \EE_{\Ub\sim P_{\Ub}} [\one\{R_m(\Ub,s_q)>R_{m'}(\Ub,s_q)\}] &=
            \PP[f_T(f_D(s_q, m),\Ub)= t_+, f_T(f_D(s_q, m'),\Ub)= t_-]\\&=0 
    \end{split}
    \end{equation}
and that
    \begin{equation}\label{eq:prop_equal_probs+}
            \PP[f_T(f_D(s_q, m),\Ub)= t_+, f_T(f_D(s_q, m'),\Ub)= t_+]= \PP[f_T(f_D(s_q, m),\Ub)= t_+] =p_m.
    \end{equation}
Using the same reasoning, if the LLM $m'$ samples the non-preferred token $t_{-}$, then, $m$ must also sample $t_{-}$ because $t_{-}$ is more likely under $m$ than under $m'$. This implies that
    \begin{equation}\label{eq:prop_equal_probs-}
            \PP[f_T(f_D(s_q, m),\Ub)= t_-, f_T(f_D(s_q, m'),\Ub)= t_-] = \PP[f_T(f_D(s_q, m'),\Ub)= t_-] = 1-p_{m'}
    \end{equation} 
Then, from Eq.~\ref{eq:prop_equal_probs+} and Eq.~\ref{eq:prop_equal_probs-}, we can conclude that
    \begin{equation}\label{eq:prop_equal_shared}
        \begin{aligned}
            \EE_{\Ub\sim P_{\Ub}} [\one\{R_m(\Ub,s_q)=R_{m'}(\Ub,s_q)\}] = p_m + (1-p_{m'})
        \end{aligned}
    \end{equation}
Finally, from Eq.~\ref{eq:prop_greater_shared} and Eq.~\ref{eq:prop_equal_shared}, we can conclude that the win-rate achieved by $m'$ against $m$ is
\begin{align*}
    \EE_{\Ub\sim P_{\Ub}} &[\one\{R_m(\Ub,s_q)<R_{m'}(\Ub,s_q)\}] 
        \\&= 1 - \EE_{\Ub\sim P_{\Ub}} [\one\{R_m(\Ub,s_q)>R_{m'}(\Ub,s_q)\}] -\EE_{\Ub\sim P_{\Ub}} [\one\{R_m(\Ub,s_q)=R_{m'}(\Ub,s_q)\}] 
        \\&= p_{m'}-p_m.
\end{align*}

%
Under independent autoregressive generation, the LLMs $m$ and $m'$ sample tokens independently from each other, \ie, $f_T(f_D(s_q, m),\Ub) \perp f_T(f_D(s_q, m'),\Ub')$. 
Thus, we can factorize all joint probabilities when computing the win-rates and obtain
\begin{align*}\nonumber
        \EE_{\Ub, \Ub'\sim P_{\Ub}} [\one\{R_m(\Ub,s_q)>R_{m'}(\Ub',s_q)\}] 
        &=\PP[f_T(f_D(s_q, m),\Ub)= t_+]  \cdot \PP[f_T(f_D(s_q, m'),\Ub')= t_-] \\ &=
        p_m \cdot (1-p_{m'})
\end{align*}
and
\begin{equation}\nonumber
    \begin{split}
        \EE_{\Ub, \Ub'\sim P_{\Ub}} & [\one\{R_m(\Ub,s_q)<R_{m'}(\Ub',s_q)\}] 
         = p_{m'} \cdot (1-p_m).
    \end{split}
\end{equation}

\subsection{Proof of Proposition~\ref{prop:gumbel_ties}}

We follow the notations and technique of Proposition~\ref{prop:var_gumbel}.
Fix query $s_q$ and consider first the case of independent autoregressive generation. Since each LLM can only assign a non-zero probability to single-token sequences, we have:
\begin{equation*}
    \begin{split}
        \PP[R_m(\Ub, s_q)=R_{m'}(\Ub', s_q)]&=\sum_{k=1}^{|V|}\PP[f_T(f_D(s_q,m),\Ub)=t_k] \cdot \PP[f_T(f_D(s_q,m'),\Ub)=t_k]\\
        &<\sum_{k=1}^{|V|}\PP[f_T(f_D(s_q,m),\Ub)=t_k],
    \end{split}
\end{equation*}
In the case of coupled autoregressive generation, since
\begin{equation*}
    \PP\left[\{f_T(f_D(s_q,m),\Ub)=t_k\} \cap \{f_T(f_D(s_q,m),\Ub)=t_j \}\right]=0,\; i\neq j,
\end{equation*}
we obtain:
\begin{equation*}
    \begin{split}
        \PP[R_m&(\Ub, s_q)=R_{m'}(\Ub, s_q)]\\
        &=\PP\left[\cup_i \{ f_T(f_D(s_q,m),\Ub)=t_k, f_T(f_D(s_q,m'),\Ub)=t_k \} \right]\\
        &=\sum_k\PP[\{f_T(f_D(s_q,m),\Ub)=t_k, f_T(f_D(s_q,m'),\Ub)=t_k \}]\\
        &=\sum_k \PP[f_T(f_D(s_q,m),\Ub)=t_k]\cdot \PP[f_T(f_D(s_q,m'),\Ub)=t_k|f_T(f_D(s_q,m),\Ub)=t_k].\\
    \end{split}
\end{equation*}

We now follow~\citet{9729603} and expand the posterior Gumbels, $\PP[f_T(f_D(s_q,m'),\Ub)=t_k \mid f_T(f_D(s_q,m),\Ub)=t_k]$, as truncated Gumbel distributions. In particular, we leverage the fact that
\begin{equation}\label{eq:max of gumbels}
    \max_{t\in V}\{U_t+\log([f_D(s_q,\bullet)]_{t})\} \sim \text{Gumbel}(0,1),
\end{equation}
and that a Gumbel distribution, with parameter $\log (\theta)$, truncated at $b\sim \text{Gumbel}(0,1)$  can be sampled as
\begin{equation}\label{eq: truncated gumbel}
    -\log(\exp(-b)-\log(\eta)/\theta),\; \eta \sim U(0,1).
\end{equation}
Furthermore, by assumption, $D_{t_k} > 0$ for all $k \in \{1, \ldots, |V|\}$, so that any bound on the absolute difference of log-probabilities $|\log ( [f_D(s_q,m)]_{t_k}) -\log ([f_D(s_q,m')]_{t_k}) |$ uniformly implies a bound on the difference of probabilities $|[f_D(s_q,m)]_{t_k} -[f_D(s_q,m')]_{t_k} |$ and vice versa. Using the bound 
\begin{equation*}
    |\log([f_D(s_q,m)]_{t_k})-\log([f_D(s_q,m')]_{t_k})|\leq \varepsilon^*
\end{equation*}
and the Gumbel properties in Eq.~\ref{eq:max of gumbels} and Eq.~\ref{eq: truncated gumbel}, we obtain:
\begin{align}
        &\PP[R_m(\Ub, s_q)=R_{m'}(\Ub, s_q)] \nn \\
        &=\sum_k \PP[f_T(f_D(s_q,m),\Ub)=t_k] \nn \\
            & \qquad \quad \times \PP\Bigg[
            \bigcap_k \Big\{
            \log( [f_D(s_1, m')]_{t_k}) - \log([f_D(s_1, m)]_{t_k})-\log(-\log (\eta_k)) \nn \\
            & \qquad \qquad \qquad \qquad \geq \log([f_D(s_1, m')]_{t_j}) - \log([f_D(s_1, m)]_{t_j}) -\log(-\log(\eta_k) -\log(\eta_j)/[f_D(s_1,m')]_{t_j})
            \Big\}
            \Bigg] \nn \\
            &\geq
            \sum_k \PP[f_T(f_D(s_q,m),\Ub)=t_k] \nn \\
            &\qquad \quad \times\PP\left[
            \cap_k \{
            -\log(-\log (\eta_k)) \geq -2\varepsilon^* -\log(-\log(\eta_k) -\log(\eta_j)/[f_D(s_1,m')]_{t_j})
            \}
            \right] \label{eq:final-bound}
\end{align}
where $\eta_k\sim \text{U}(0,1)$ are independently distributed uniform random variables.
Now, note that the claim holds for $\varepsilon^*=0$ since, in that case, we have that
\begin{equation*}
   \PP\left[
        \bigcap_k \left\{
        -\log(-\log (\eta_k)) \geq -\log(-\log(\eta_k) -\log(\eta_k)/[f_D(s_1,m')]_{t_k})
        \right\} \right]=1,
\end{equation*}
using that $x\mapsto -\log(x)$ is strictly decreasing. 
Since all terms in Eq.~\ref{eq:final-bound} are continuous in $\varepsilon^*$, there exists $\varepsilon^*(m) >0 $, possibly dependent on $m$ but independent of $m'$, such that 
\begin{equation}\label{eq:prop5 claim prob}
    \PP[R_m(\Ub, s_q)=R_{m'}(\Ub, s_q)] > \PP[R_m(\Ub, s_q)=R_{m'}(\Ub', s_q)]
\end{equation}
holds if
\begin{equation*}
    \sup_{s_q} \norm{\log(f_D(s_q,m))-\log(f_D(s_q, m'))}_\infty < \varepsilon^*(m).
\end{equation*}
Since by assumption $D_t>0$ for all $t\in V$, there exists $\varepsilon(m)>0$ in probability space such that Eq.~\ref{eq:prop5 claim prob} holds if

\begin{equation*}
    \sup_{s_q} \norm{f_D(s_q,m)-f_D(s_q, m')}_\infty < \varepsilon(m).
\end{equation*}
This concludes the proof.

\subsection{Calculation of average win-rates in the example used in Sections~\ref{sec:intro} and~\ref{sec:pairwise}} \label{app:example-ranking}
In this section, we provide detailed calculations of the win-rates for the example in Sections~\ref{sec:intro} and~\ref{sec:pairwise}. Recall that in this example, we are given three LLMs $m_1$, $m_2$ and $m_3$, and we need to rank them according to their ability to answer correctly two types of input prompts, $q$ and $q'$, picked uniformly at random.
We assume that the true probability that each LLM answers correctly each type of input prompt is given by:
\begin{table}[h]
    \centering
    \begin{tabular}{lccc}
        \toprule
         & $m_1$ & $m_2$ & $m_3$ \\
        \midrule
        $q$           & $p_1=0.4$             & $p_2=0.48$           & $p_3=0.5$             \\
        $q'$           & $p'_1=1$           & $p'_2=0.9$         & $p'_3=0.89$              \\
        \bottomrule
 \end{tabular}
 \vspace{-8mm}
 \caption*{}
 \end{table}

Using Proposition~\ref{prop:gap_win_rates_stability}, 
the win-rates under independent autoregressive generation are given, for each LLM $m_k$, by:
\begin{equation}
       \frac{1}{2}\sum_{j\neq k}\EE_{\Ub, \Ub' \sim P_{\Ub}, S_q\sim P_Q}[\one\{R_{m_k}(\Ub, S_q)>R_{m_j}(\Ub', S_q)\}] =\frac{\sum_{j\neq k}p_k(1-p_j) + \sum_{j\neq k}p'_k(1-p'_j)}{4}.
\end{equation}
Substituting the numerical values we obtain:

 \begin{equation}
    \begin{split}
       \frac{1}{2}\sum_{j\neq 1}\EE_{\Ub, \Ub' \sim P_{\Ub}, S_q\sim P_Q}[\one\{R_{m_1}(\Ub, S_q)>R_{m_j}(\Ub', S_q)\}] & =0.1545,\\
       \frac{1}{2}\sum_{j\neq 2}\EE_{\Ub, \Ub' \sim P_{\Ub}, S_q\sim P_Q}[\one\{R_{m_2}(\Ub, S_q)>R_{m_j}(\Ub', S_q)\}] & =0.15675,\\
       \frac{1}{2}\sum_{j\neq 3}\EE_{\Ub, \Ub' \sim P_{\Ub}, S_q\sim P_Q}[\one\{R_{m_3}(\Ub, S_q)>R_{m_j}(\Ub', S_q)\}] & =0.16225\\
    \end{split}
\end{equation}

Similarly, using Proposition~\ref{prop:gap_win_rates_stability}, the win-rates using coupled autoregressive generation can be written, for each LLM $m_k$, as:
\begin{equation}
       \frac{1}{2}\sum_{j\neq k}\EE_{\Ub \sim P_{\Ub}, S_q\sim P_Q}[\one\{R_{m_k}(\Ub, S_q)>R_{m_j}(\Ub, S_q)\}]  =\frac{\sum_{j\neq k}(p_k-p_j)_{+} + \sum_{j\neq k}(p'_k-p'_j)_{+}}{4},
\end{equation}
where $(\bullet)_{+}=\max(0, \bullet)$ denotes the positive part. Substituting the numerical values we obtain:

 \begin{align*}
       \frac{1}{2}\sum_{j\neq 1}\EE_{\Ub \sim P_{\Ub}, S_q\sim P_Q}[\one\{R_{m_1}(\Ub, S_q)>R_{m_j}(\Ub, S_q)\}] & = 
       0.0525, \\
       \frac{1}{2}\sum_{j\neq 2}\EE_{\Ub \sim P_{\Ub}, S_q\sim P_Q}[\one\{R_{m_2}(\Ub, S_q)>R_{m_j}(\Ub, S_q)\}] &=
       0.0225, \\
       \frac{1}{2}\sum_{j\neq 3}\EE_{\Ub \sim P_{\Ub}, S_q\sim P_Q}[\one\{R_{m_3}(\Ub, S_q)>R_{m_j}(\Ub, S_q)\}] &=
       0.03.
\end{align*}

\clearpage
\newpage

\section{ADDITIONAL EXPERIMENTAL DETAILS} \label{app:add_exp}
\vspace{-2mm}

\xhdr{Hardware setup} Our experiments are executed on a compute server equipped with 2 $\times$ Intel Xeon Gold 5317 CPU, $1{,}024$ GB main memory, and $2$ $\times$ A100 Nvidia Tesla GPU ($80$ GB, Ampere Architecture). In each experiment a single Nvidia A100 GPU is used.

\xhdr{Datasets} 
As benchmark datasets, we use (i) the Measuring Massive Multitask Language Understanding dataset (MMLU)~\citep{hendrycks2021measuring}, (ii) the validation data split of the Grade School Math 8K dataset (GSM8K)~\citep{cobbe2021gsm8k}, and (iii) the HumanEval dataset~\citep{chen2021evaluating}. MMLU  consists of $14{,}042$ questions covering $52$ diverse knowledge areas with each question offering four possible choices indexed from A to D, and a ground-truth answer. %
The validation data split of the GSM8K dataset comprises $1{,}319$ grade school math problems along with their full solution as a string. 
The HumanEval dataset includes $164$ programming problems with a function signature, docstring, body, and several unit tests. %
For pairwise comparison tasks, we use the first $500$ questions from the LMSYS-Chat-1M dataset~\citep{zheng2024lmsys}.

\xhdr{Models} In our experiments with the \texttt{Llama} family of models, we use \texttt{Llama-3.1-8B-Instruct}, its quantized variants \texttt{Llama-3.1-8B-Instruct-\{AWQ-INT4, bnb-4bit, bnb-8bit\}} and \texttt{Llama-3.2-\{1B, 3B\}-Instruct} models.
In our experiments with the \texttt{Qwen} family of models, we use \texttt{Qwen2.5-7B-Instruct}, its quantized variants \texttt{Qwen2.5-7B-Instruct-\{AWQ-INT4, bnb-4bit, bnb-8bit\}}, \texttt{Qwen2.5-3B-Instruct}, a distilled variant \texttt{DistilQwen2.5-3B-Instruct}, \texttt{Qwen2.5-1B-Instruct} and \texttt{Qwen3-8B}.
In our experiments with the \texttt{Mistral} family of models, we use \texttt{Mistral-7B-Instruct-v0.3}, its quantized variants \texttt{Mistral-7B-Instruct-v0.3-\{bnb-4bit, bnb-8bit\}} and \texttt{Mistral-7B-Instruct-\{v0.1,v0.2\}}.
The models are obtained from \texttt{Hugging Face}, and the quantized \texttt{bnb-4bit, bnb-8bit} LLM variants are built using the \texttt{bitsandbytes} library~\citep{bitsnbytes2024bits}.
The token vocabulary of \texttt{Mistral-7B-Instruct-v0.3} contains $768$ additional tokens compared to \texttt{Mistral-7B-Instruct-\{v0.1,v0.2\}}, and \texttt{Qwen3-8B}'s vocabulary contains $4$ additional tokens compared to the rest of the models in the \texttt{Qwen} family. Such differences correspond to additional control tokens used to, for example, allow models to access external information retrieval tools, and are therefore neither sampled nor relevant in our experiments. In our experiments, within each model family, we sample from the larger vocabulary and set the probability of tokens that are not included in the smaller vocabulary to zero when using the respective models. 

\begin{figure}[h]
\centering
\begin{tabular}{c c c}
    \multicolumn{3}{c}{temperature = 0.5}\\
    \includegraphics[width=0.23\linewidth]{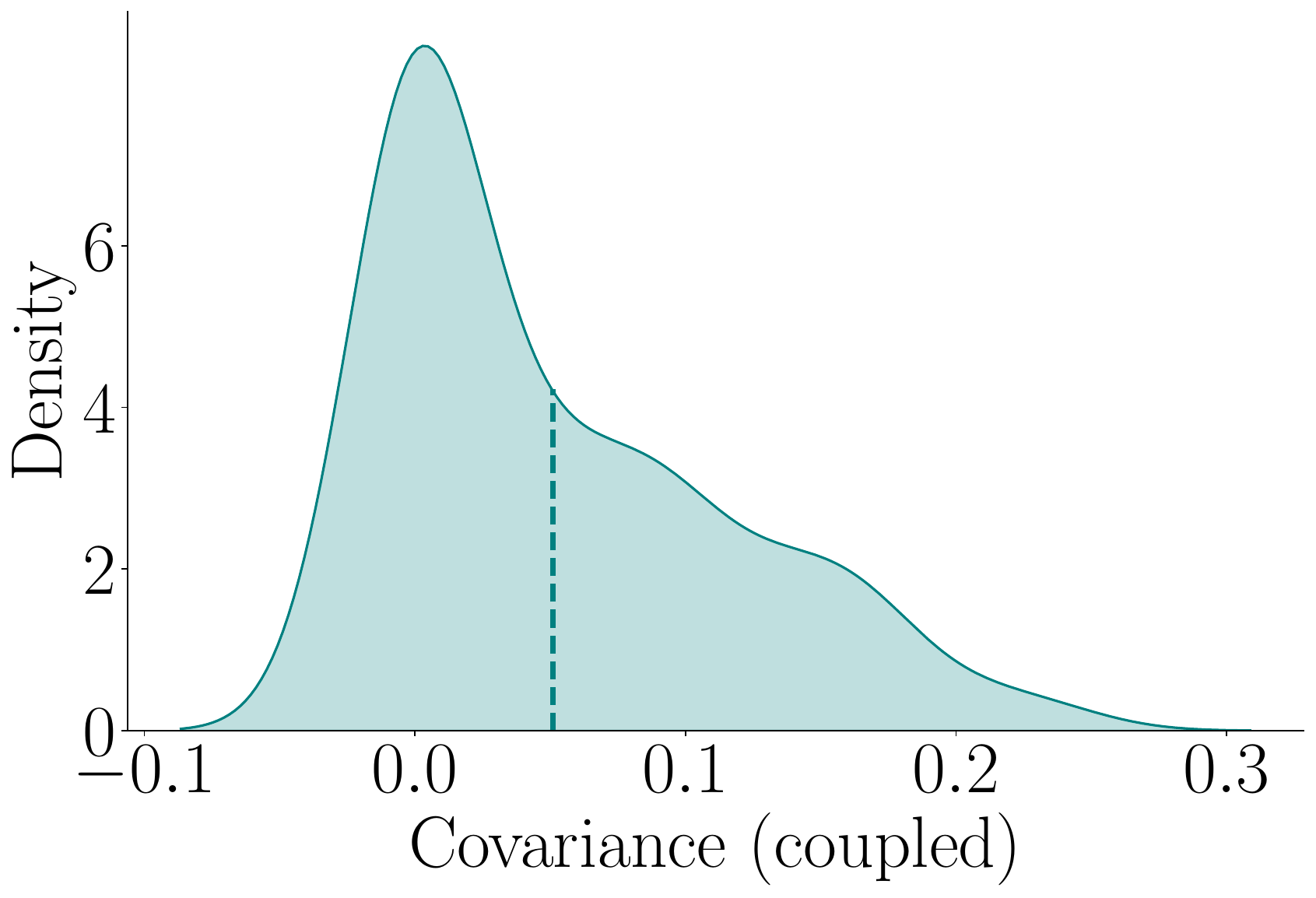} &
    \includegraphics[width=0.23\linewidth]{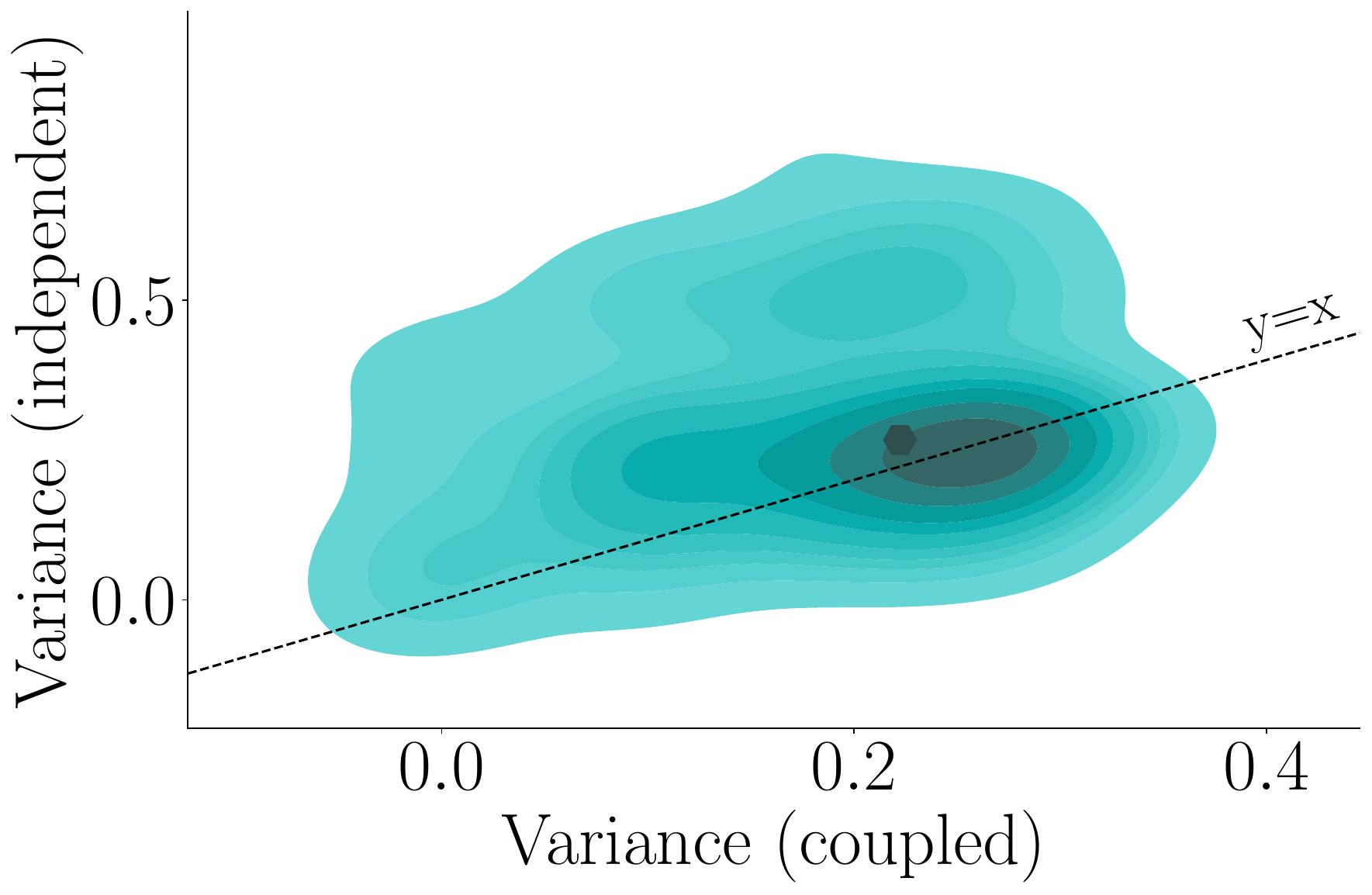} &
    \includegraphics[width=0.23\linewidth]{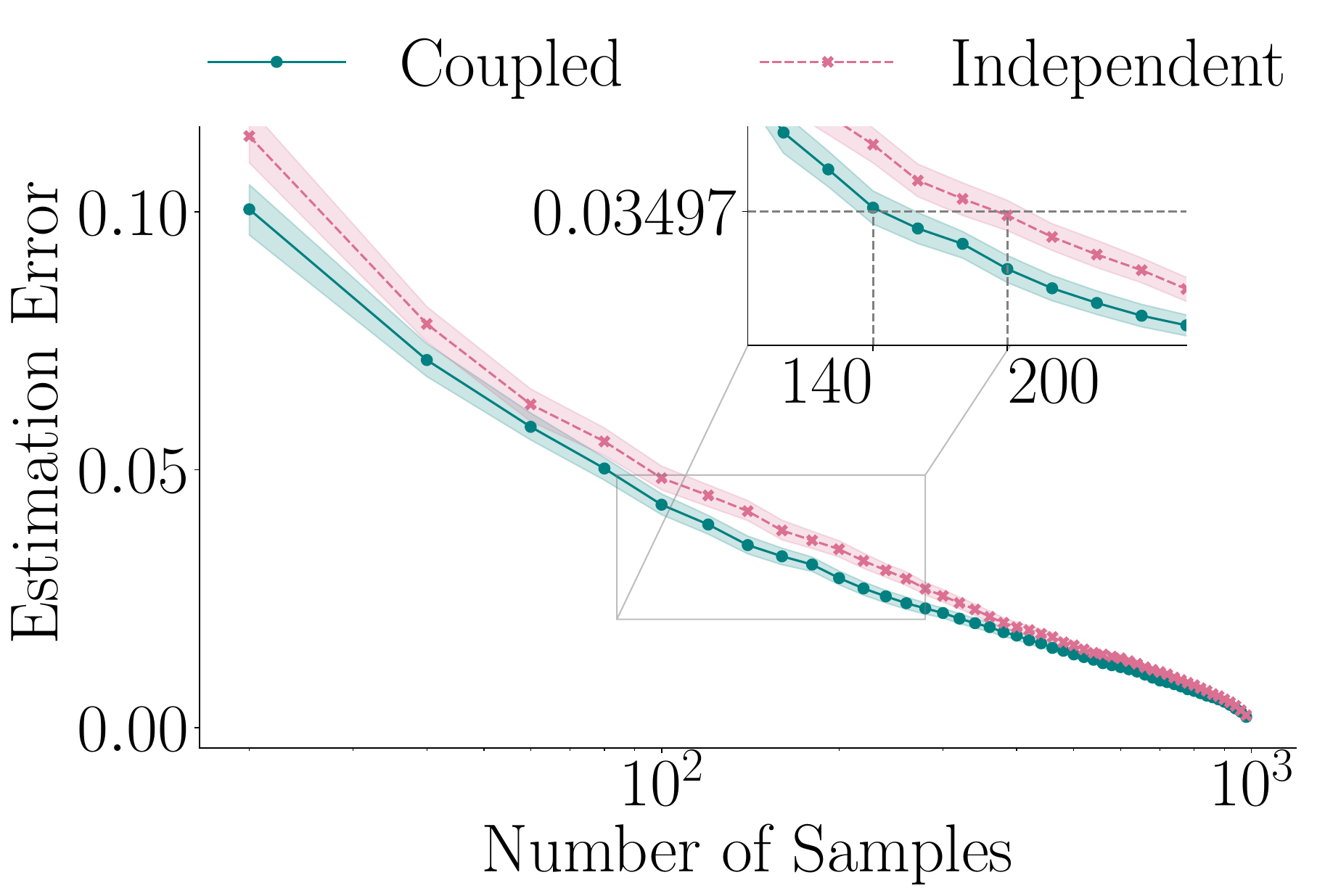} \\ \\
    \multicolumn{3}{c}{temperature = 0.9}\\
     \includegraphics[width=0.23\linewidth]{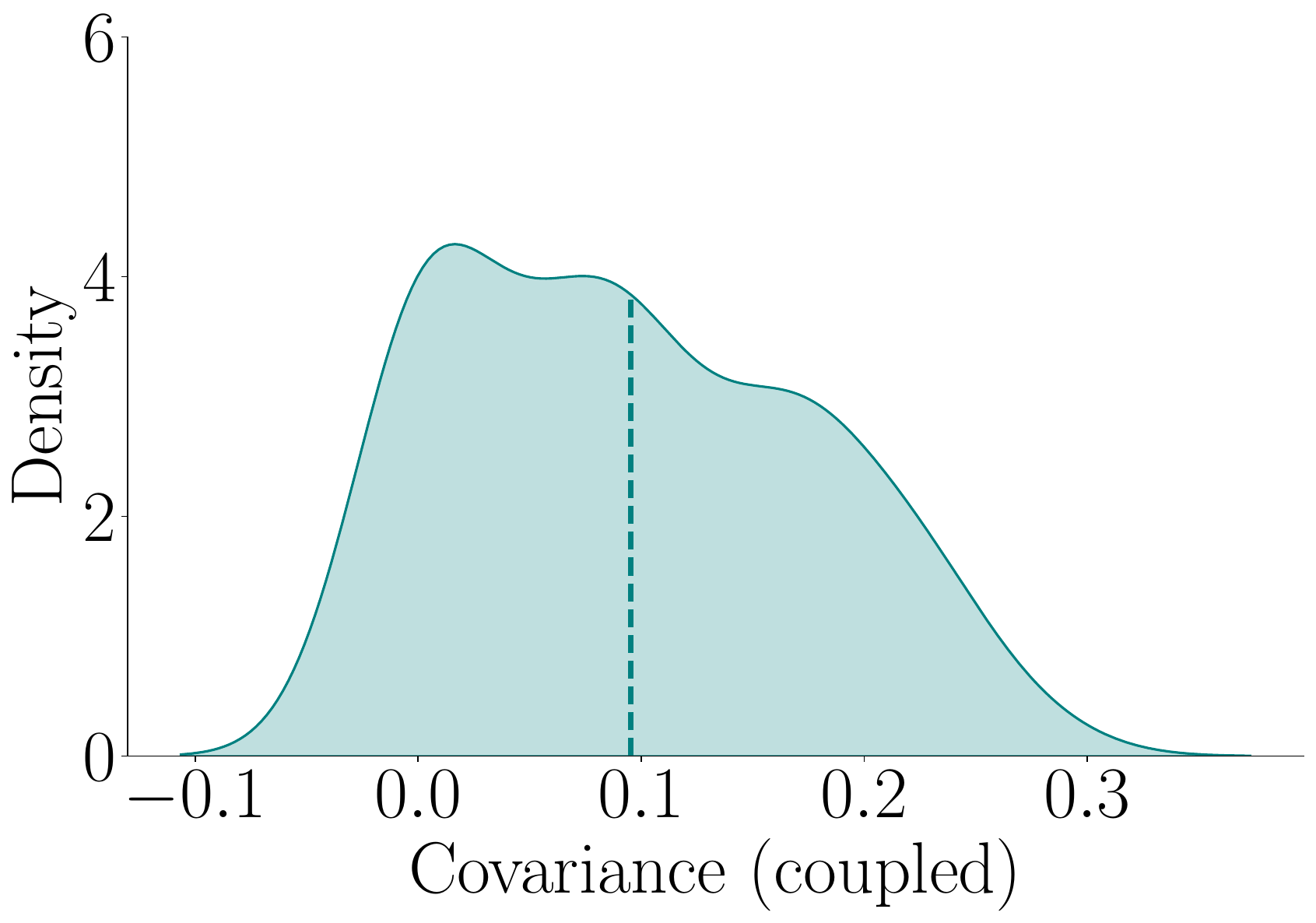} &
    \includegraphics[width=0.23\linewidth]{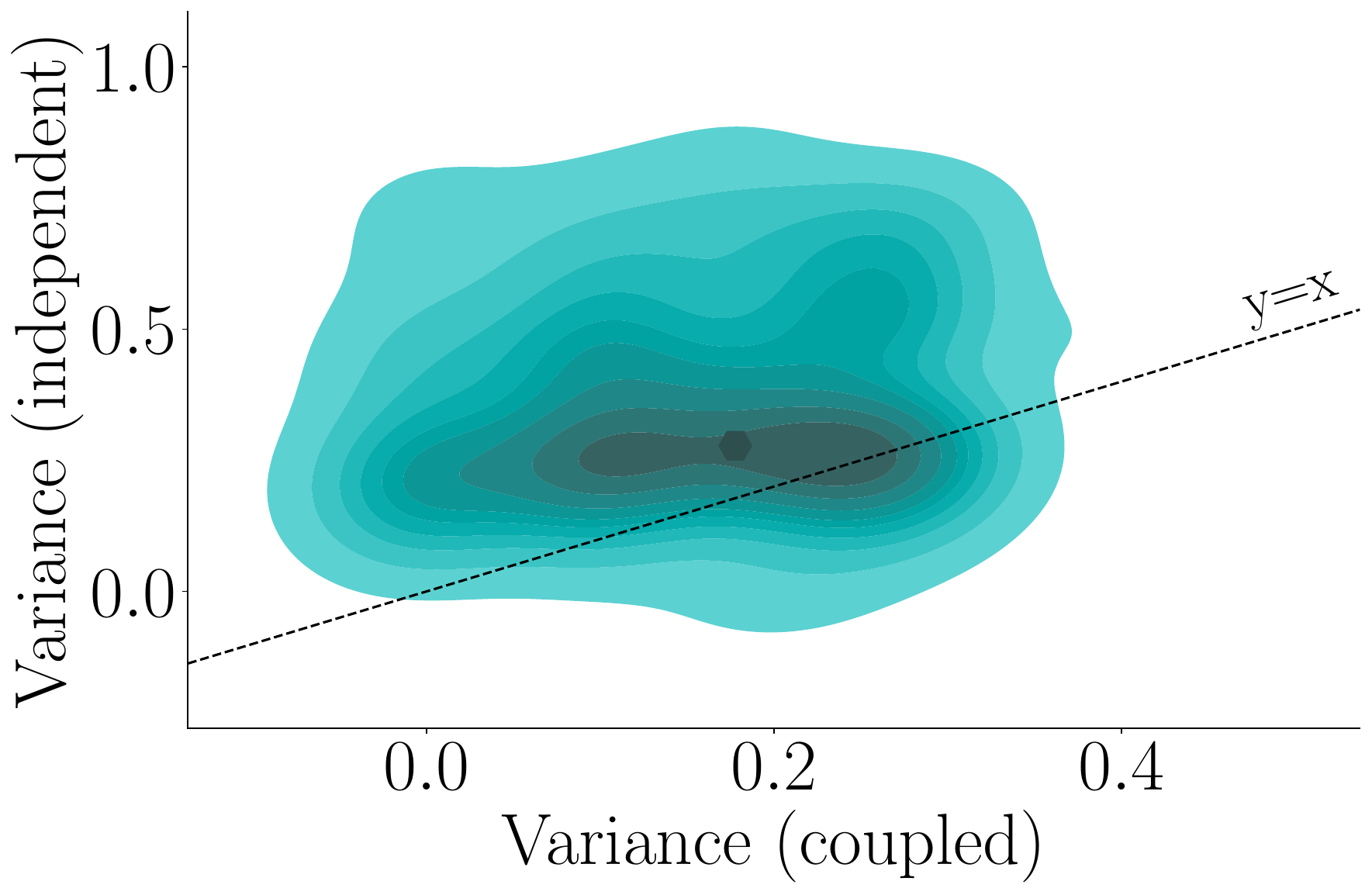} &
    \includegraphics[width=0.23\linewidth]{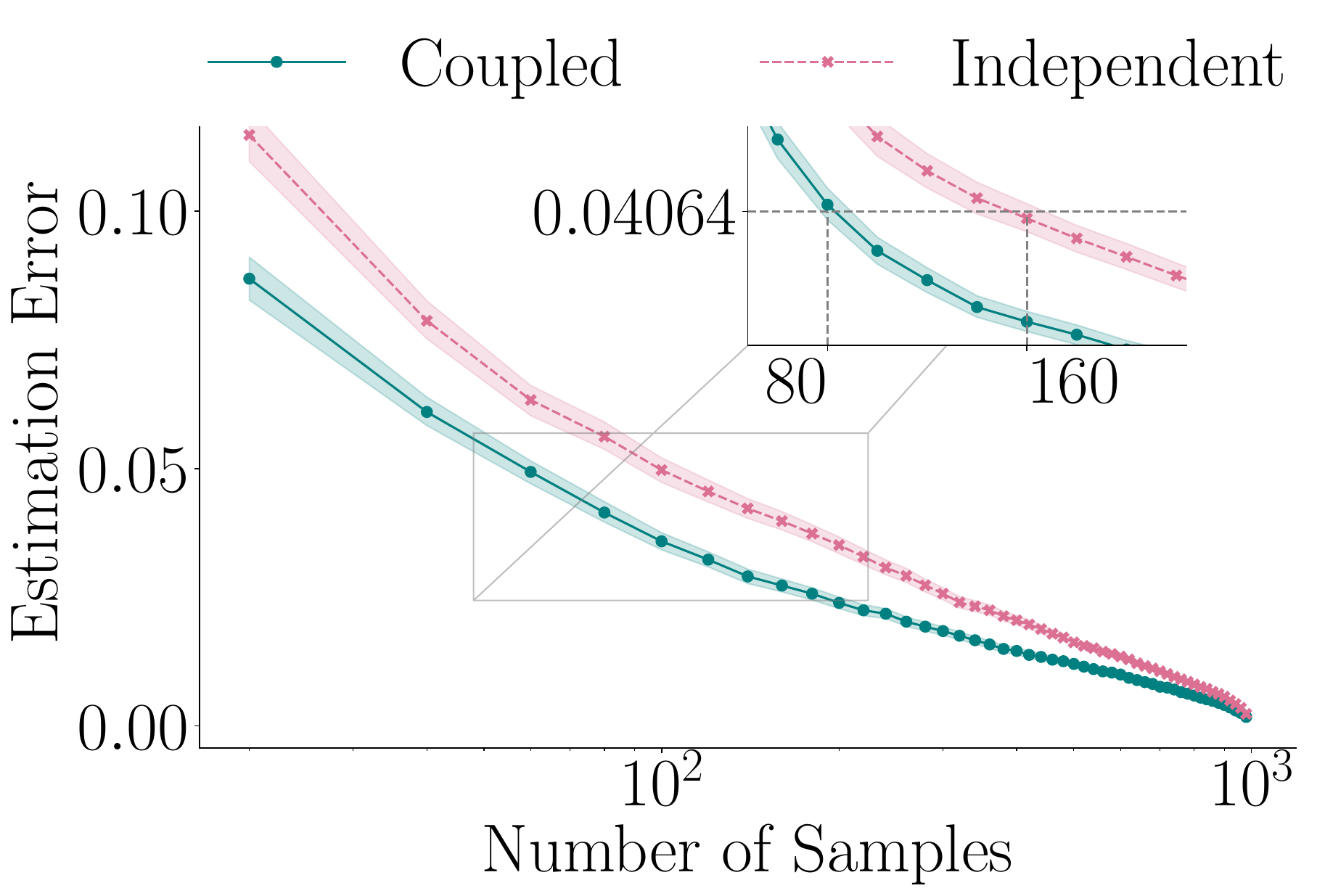} \\ \\ 
    \multicolumn{3}{c}{temperature = 0.7, top-p = 0.9}\\
     \includegraphics[width=0.23\linewidth]{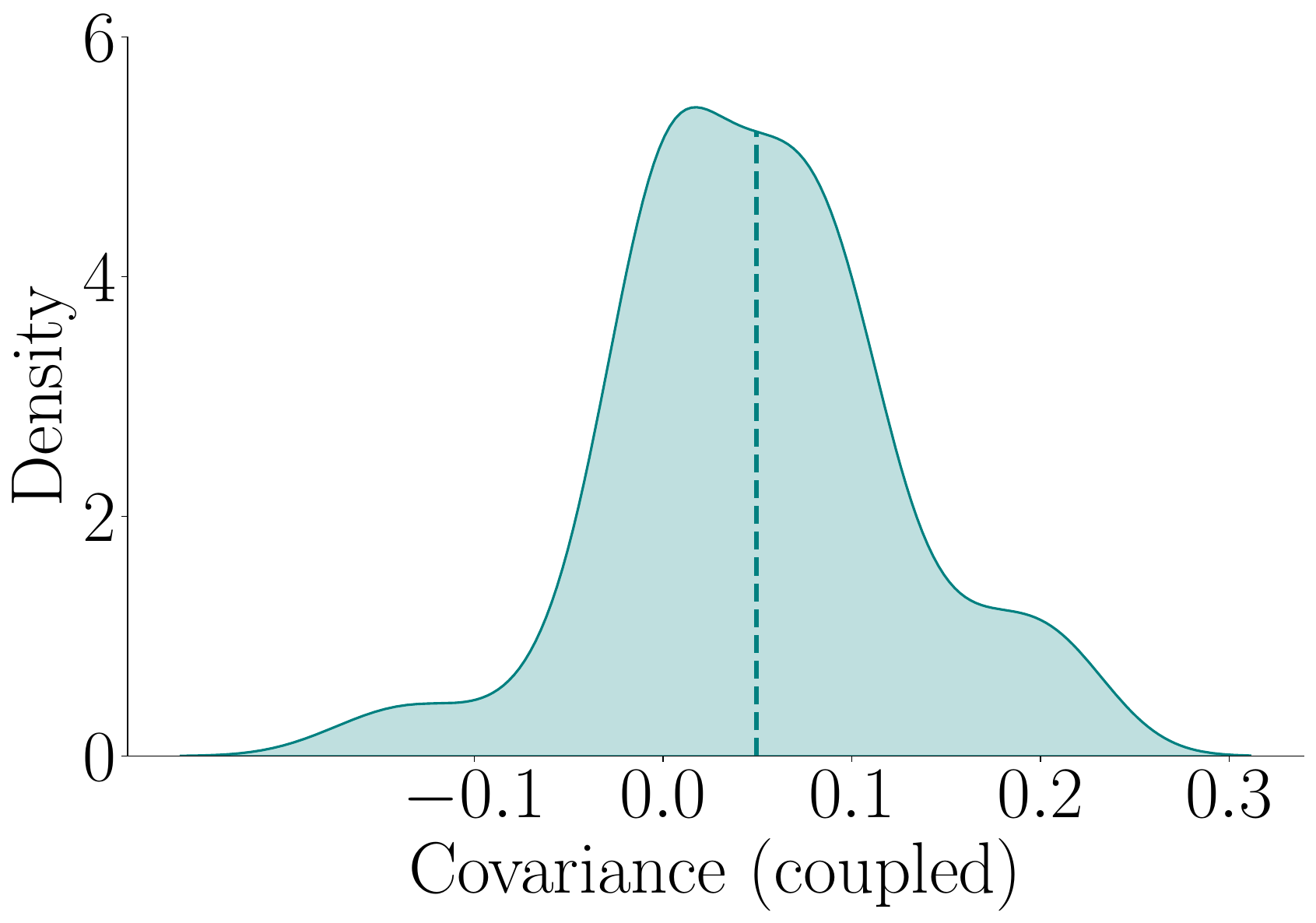} &
    \includegraphics[width=0.23\linewidth]{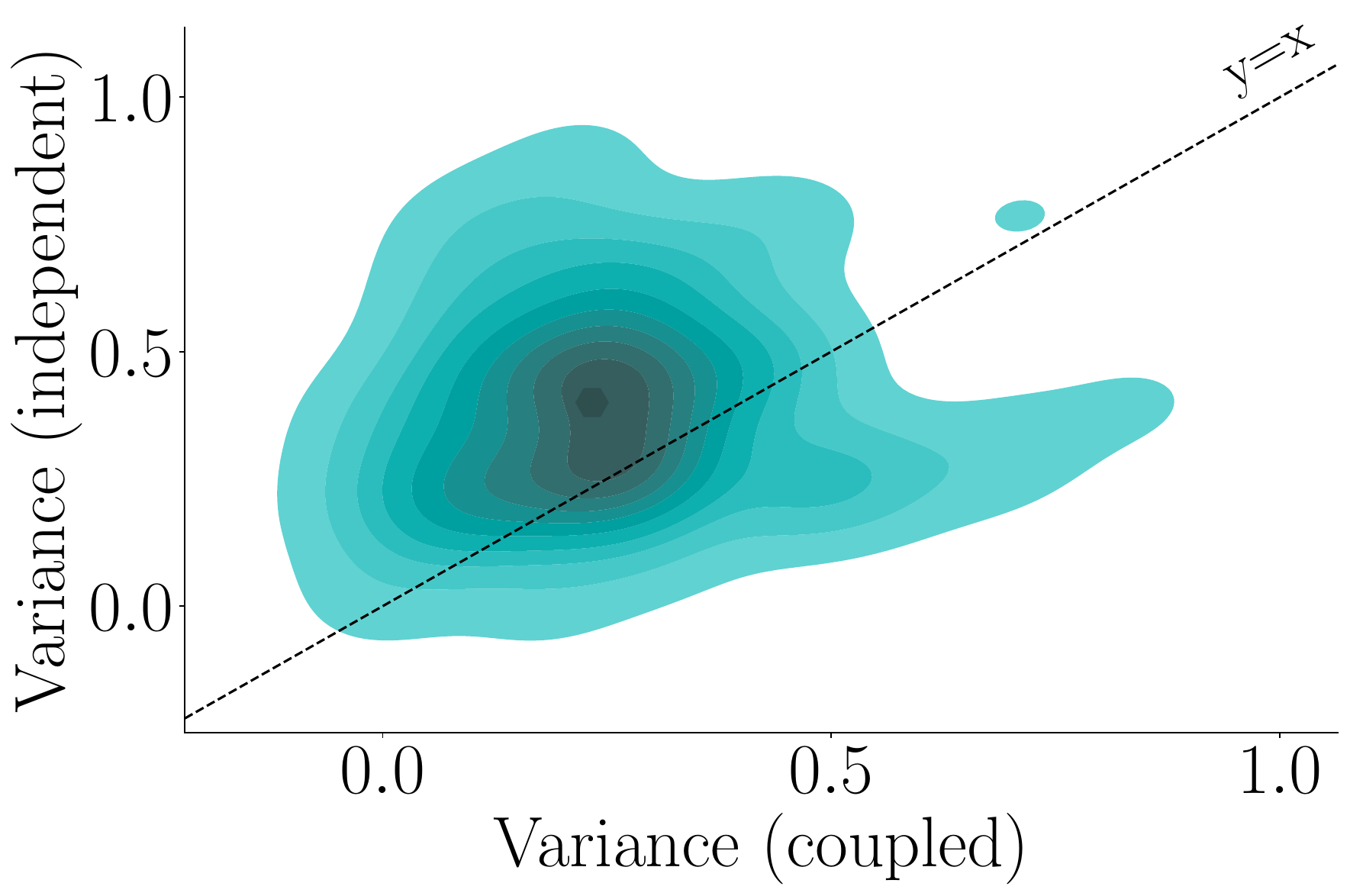} &
    \includegraphics[width=0.23\linewidth]{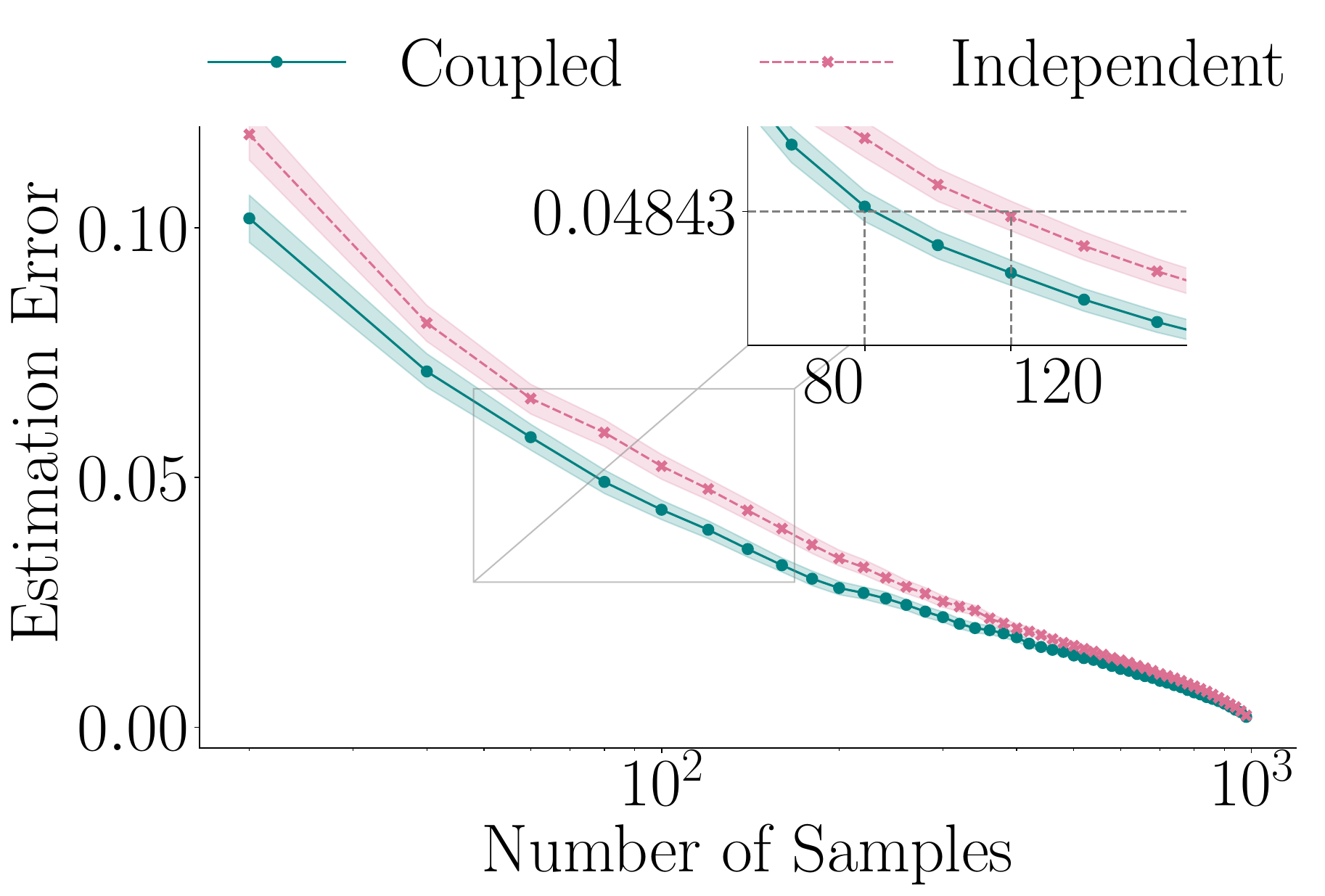} \\ \\

    (a) Score covariance & (b) Variance of the score difference & (c) Estimation error vs. \# samples \\ 
    
\end{tabular}
    \caption{\textbf{Comparison Comparison between \texttt{1B} and \texttt{3B} from the \texttt{Llama} family on multiple-choice questions from the ``college computer science'' knowledge area of the MMLU dataset under different sampling parameters.}
    Panels in column (a) show the kernel density estimate (KDE) of the covariance between the scores of the two LLMs on each question under coupled generation; the dashed lines correspond to average values. Panels in column (b) show the KDE of the variance of the difference between the scores of the LLMs on each question under coupled and independent generation; the highlighted points correspond to median values. Panels in column (c) show the absolute error in the estimation of the expected difference between the scores of the LLMs against the number of samples; for each point on the x-axis, we perform $1{,}000$ sub-samplings and shaded areas correspond to $95\%$ confidence intervals.}
    \label{fig:mmlu-temperature-top-p}
\end{figure}

\xhdr{Parameters}
For the benchmark tasks, we set the temperature to $0.7$ unless specified otherwise (Figure~\ref{fig:mmlu-temperature-top-p}), and, for the pairwise comparison tasks, we set it to $1$.
When generating with \texttt{Qwen3-8B}, we disable thinking mode.

\xhdr{Prompts} To instruct LLMs for generating output, we use the system prompt in Table~\ref{app:sys_prompt_mmlu} for the MMLU dataset, the system prompt in Table~\ref{app:sys_prompt_gsm8k} for the GSM8K data, and the system prompt in Table~\ref{app:sys_prompt_lmsys} for the LMSYS-Chat-1M dataset. For the HumanEval dataset we use no system prompt, but rather the (input) prompt and response prefixes shown in Table~\ref{app:usr-prefix-human-eval}.\footnote{We chose the prefixes following the evaluation details on the HumanEval dataset for \texttt{8B}  available in \url{https://huggingface.co/datasets/meta-llama/Llama-3.1-8B-Instruct-evals} }
Further, to perform pairwise comparisons of outputs of different LLMs, we use the system prompt in Table~\ref{app:sys_prompt_gpt4}, which is adapted from~\citet{chiang2024chatbot}, to prompt the strong LLM.

\xhdr{Licenses} 
The MMLU, GSM8K and HumanEval datasets are licensed under MIT, and the LMSYS-Chat-1M dataset is licensed under the LMSYS-Chat-1M Dataset License Agreement.\footnote{\url{https://huggingface.co/datasets/lmsys/lmsys-chat-1m}}
The \texttt{Llama-3.1-8B-Instruct} model, its quantized version \texttt{Llama-3.1-8B-Instruct-AWQ-INT4} and the prompt used for the HumanEval dataset are licensed under the LLAMA 3.1 COMMUNITY LICENSE AGREEMENT.\footnote{\url{https://www.llama.com/llama3_1/license/}}.
The \texttt{Llama-3.2-\{1B, 3B\}-Instruct} models are licensed under the LLAMA 3.2 COMMUNITY LICENSE AGREEMENT.\footnote{\url{https://www.llama.com/llama3_2/license/}}.
The models in the \texttt{Qwen} and \texttt{Mistral} families are licensed under Apache $2.0$.

\begin{table}[h]
\centering
\begin{tcolorbox}[
    colframe=white,      
    colback=gray!14,     
    boxrule=0.5mm,       
    arc=4mm,             
    left=3mm,            
    right=3mm,           
    top=3mm,             
    bottom=3mm           
]
\begin{tabular}{ m{0.97\textwidth} }
    \rowcolor{gray!14}
    \textbf{System:} You will be given multiple choice questions. Please reply with a single character `A', `B', `C', or `D' only. DO NOT explain your reply.
\end{tabular}
\end{tcolorbox}
\caption{System prompt used for the MMLU dataset.}
\label{app:sys_prompt_mmlu}
\end{table}

\vspace{-3mm}

\begin{table}[h]
\centering
\begin{tcolorbox}[
    colframe=white,      
    colback=gray!14,     
    boxrule=0.5mm,       
    arc=4mm,             
    left=3mm,            
    right=3mm,           
    top=3mm,             
    bottom=3mm           
]
\begin{tabular}{ m{0.97\textwidth} }
    \rowcolor{gray!14}
    \textbf{System:} You will be given a mathematical problem. Please reply only with a single number as your answer.
\end{tabular}
\end{tcolorbox}
\caption{System prompt used for the GSM8K dataset.}
\label{app:sys_prompt_gsm8k}
\end{table}

\vspace{-3mm}

\begin{table}[h]
\centering
\begin{tcolorbox}[
    colframe=white,      
    colback=gray!14,     
    boxrule=0.5mm,       
    arc=4mm,             
    left=3mm,            
    right=3mm,           
    top=3mm,             
    bottom=3mm           
]
\begin{tabular}{ m{0.97\textwidth} }
    \rowcolor{gray!14}
    \textbf{System:} Keep your responses short and to the point.
\end{tabular}
\end{tcolorbox}
\caption{System prompt used for the LMSYS Chatbot Arena dataset.}
\label{app:sys_prompt_lmsys}
\end{table}
\vspace{-3mm}
\begin{table}[th]
\centering
\begin{tcolorbox}[
    colframe=white,      
    colback=gray!14,     
    boxrule=0.5mm,       
    arc=4mm,             
    left=3mm,            
    right=3mm,           
    top=3mm,             
    bottom=3mm           
]
\begin{tabular}{ m{0.97\textwidth} }
    \rowcolor{gray!14}
    \textbf{Prompt Prefix:} \\ \\ 
    Write a solution to the following problem and make sure that it passes the tests:\\
    ```python \\ \\
    \textbf{Response Prefix:} \\ \\
    Here is the completed function: \\
    ```python\\ \\
 
\end{tabular}
\end{tcolorbox}
\caption{Prompt and response prefix for the HumanEval dataset.}
\label{app:usr-prefix-human-eval}
\end{table}

\vspace{-3mm}

\begin{table}[th]
\centering
\begin{tcolorbox}[
    colframe=white,      
    colback=gray!14,     
    boxrule=0.5mm,       
    arc=4mm,             
    left=3mm,            
    right=3mm,           
    top=3mm,             
    bottom=3mm           
]
\begin{tabular}{ m{0.97\textwidth} }
\rowcolor{gray!14} 
    \textbf{System:} Please act as an impartial judge and evaluate the quality of the responses provided by two AI assistants to the user prompt displayed below. Your job is to evaluate which assistant's answer is better. When evaluating the assistants' answers, compare both assistants' answers. You must identify and correct any mistakes or inaccurate information. Then consider if the assistant's answers are helpful, relevant, and concise. Helpful means the answer correctly responds to the prompt or follows the instructions. Note when user prompt has any ambiguity or more than one interpretation, it is more helpful and appropriate to ask for clarifications or more information from the user than providing an answer based on assumptions. Relevant means all parts of the response closely connect or are appropriate to what is being asked. Concise means the response is clear and not verbose or excessive. Then consider the creativity and novelty of the assistant's answers when needed. Finally, identify any missing important information in the assistants' answers that would be beneficial to include when responding to the user prompt. do not provide any justification or explanation for your response. You must output only one of the following choices as your final verdict: 

    \vspace{2mm}
    `A' if the response of assistant A is better
    
    \vspace{1mm}
    `B' if the response of assistant B is better
    
    \vspace{1mm}
    `Tie' if the responses are tied
\end{tabular}
\end{tcolorbox}
\caption{\label{app:system_prompt}System prompt used for obtaining pairwise preferences using \texttt{GPT-4o-2024-11-20} as the judge.}
\label{app:sys_prompt_gpt4}
\end{table}

\clearpage
\newpage

\section{ADDITIONAL EXPERIMENTS WITH LLMS IN THE \texttt{Llama} FAMILY}
\label{app:llama}
\subsection{MMLU Dataset}
\label{app:llama-mmlu}
\begin{figure}[!!h]
\centering
\begin{tabular}{c c c}
    \multicolumn{3}{c}{\texttt{1B} vs. \texttt{3B}}\\
    \includegraphics[width=0.23\linewidth]{./figures/mmlu/college_computer_science/covariance/cov_Llama-3.2-3B-Instruct_Llama-3.2-1B-Instruct_kde.pdf} &
    \includegraphics[width=0.23\linewidth]{./figures/mmlu/college_computer_science/variance/var_Llama-3.2-3B-Instruct_Llama-3.2-1B-Instruct_kde.pdf} &
    \includegraphics[width=0.23\linewidth]{./figures/mmlu/college_computer_science/error/error_Llama-3.2-3B-Instruct_Llama-3.2-1B-Instruct.pdf} \\ \\
%
    \multicolumn{3}{c}{\texttt{1B} vs. \texttt{8B}}\\
    \includegraphics[width=0.23\linewidth]{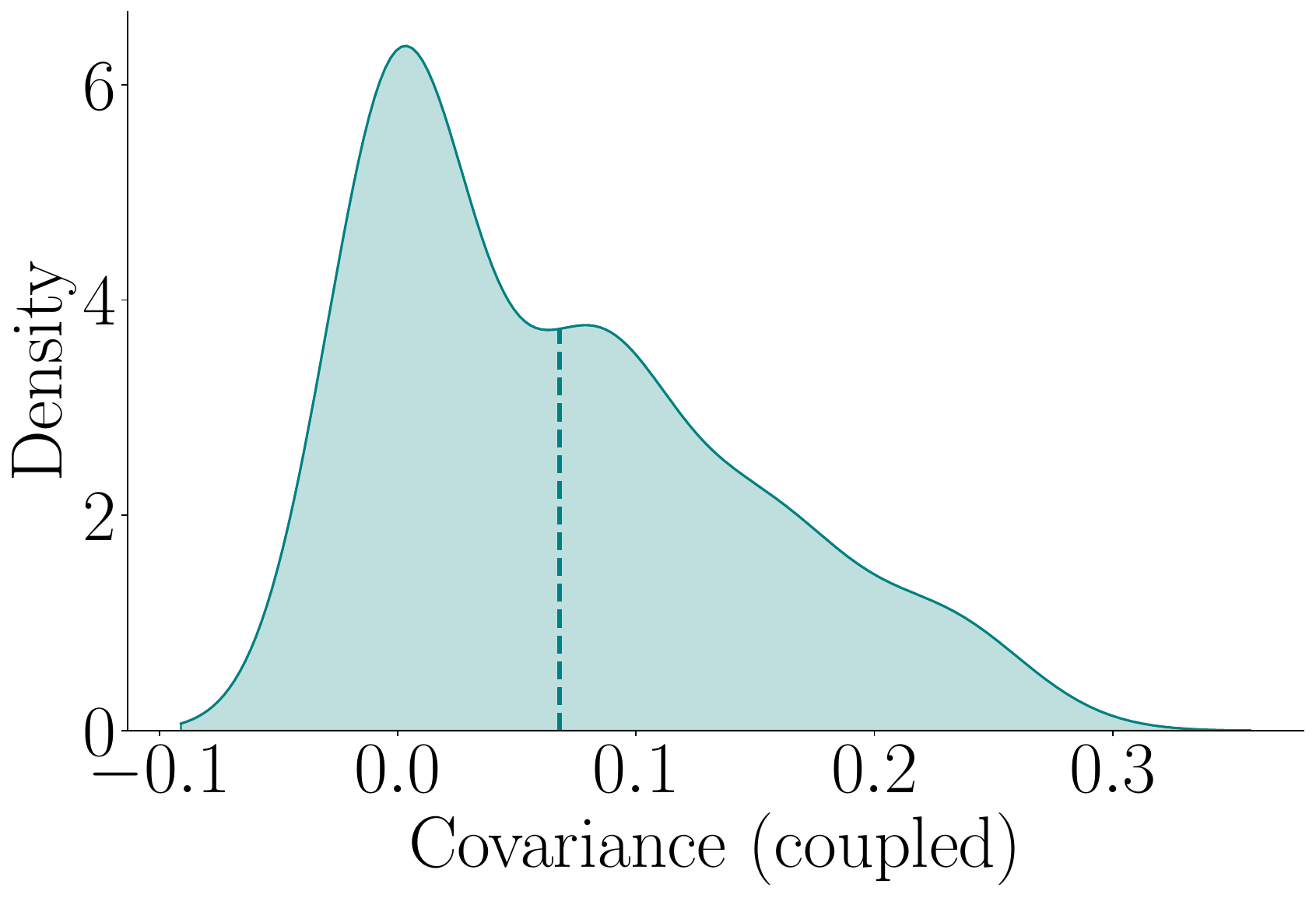} &
    \includegraphics[width=0.23\linewidth]{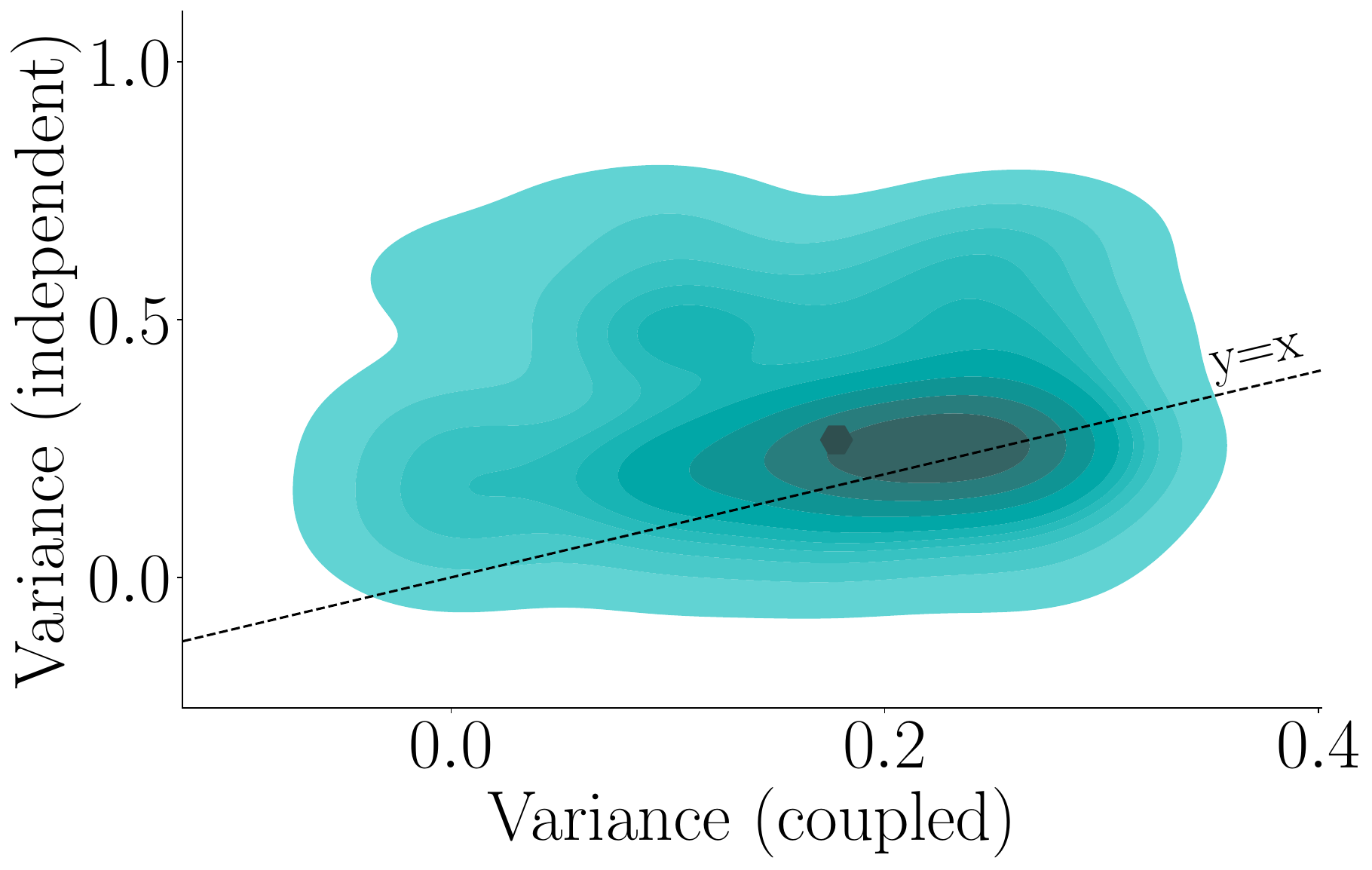} &
    \includegraphics[width=0.23\linewidth]{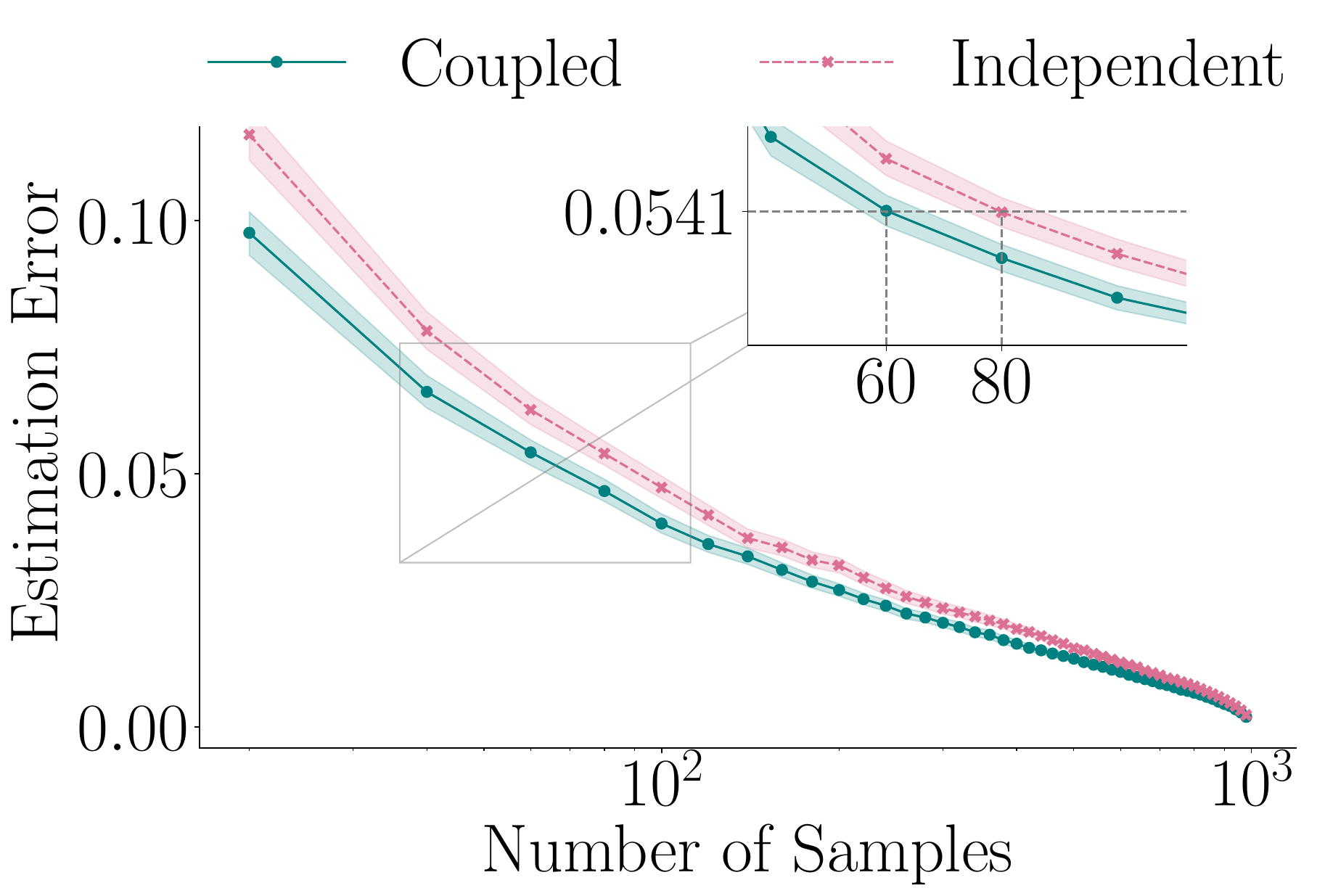} \\ \\
%
    \multicolumn{3}{c}{\texttt{3B} vs. \texttt{8B}}\\
    \includegraphics[width=0.23\linewidth]{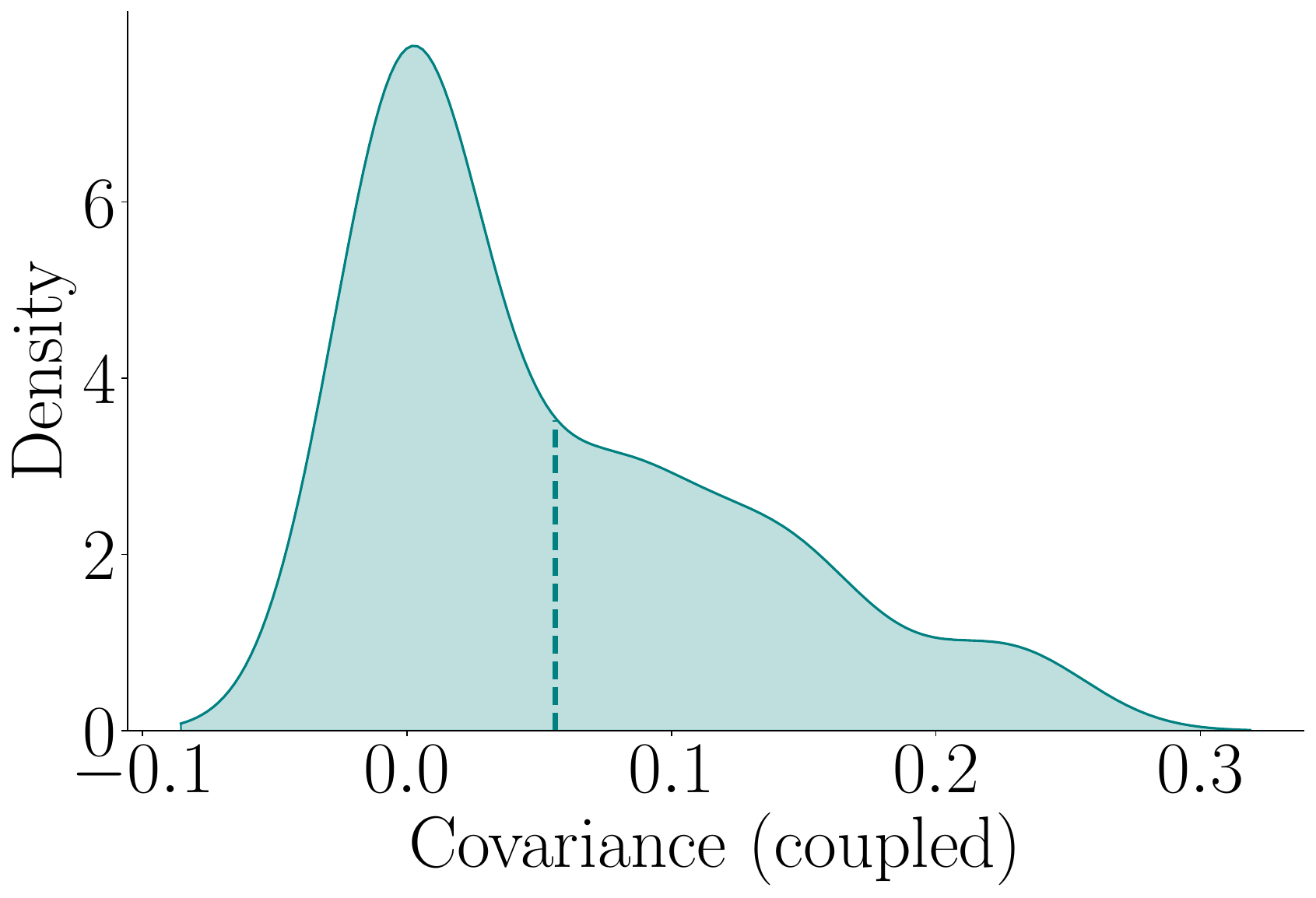} &
    \includegraphics[width=0.23\linewidth]{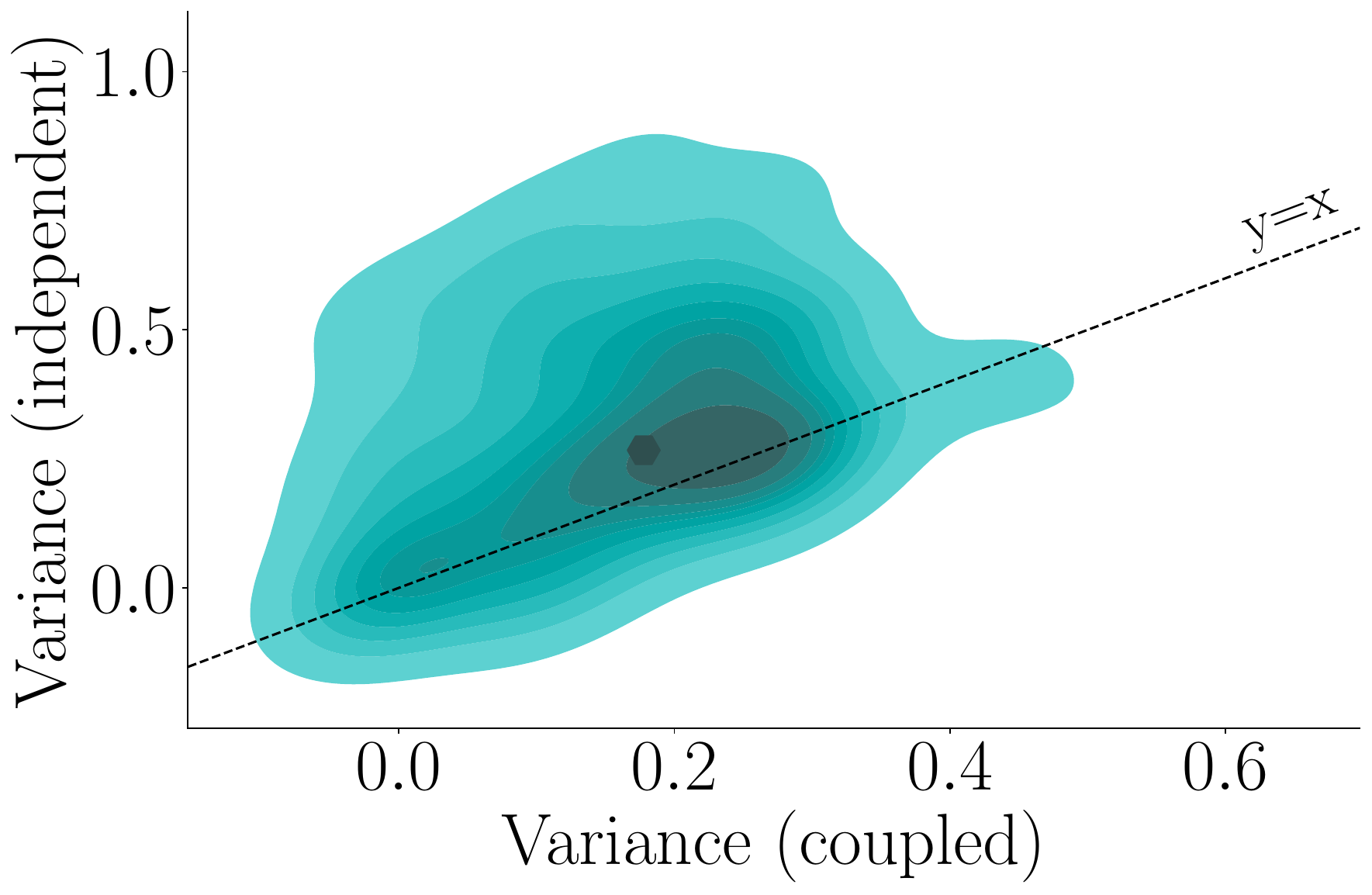} &
    \includegraphics[width=0.23\linewidth]{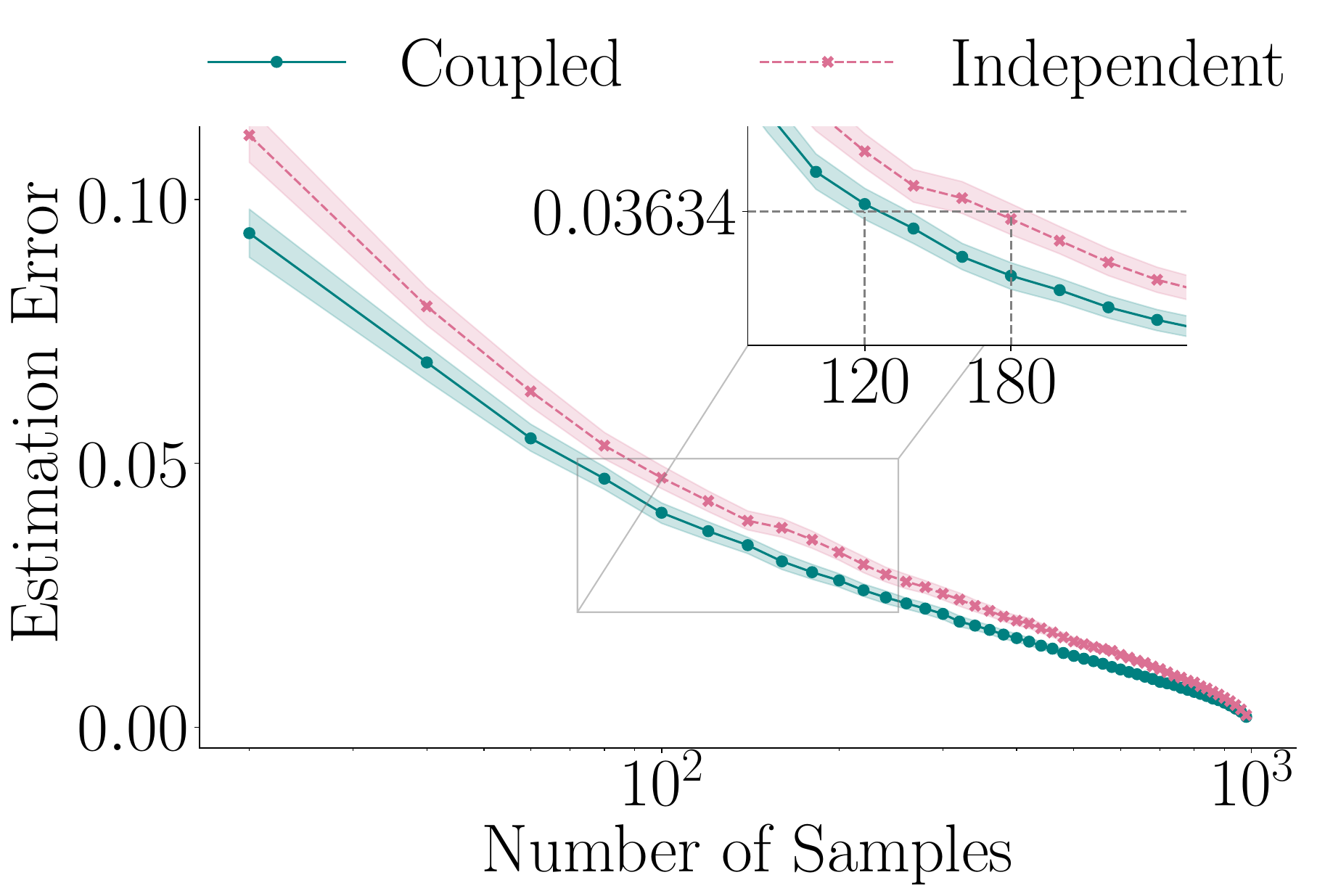} \\ \\
%
 \multicolumn{3}{c}{\texttt{8B} vs. \texttt{bnb-8bit}}\\
    \includegraphics[width=0.23\linewidth]{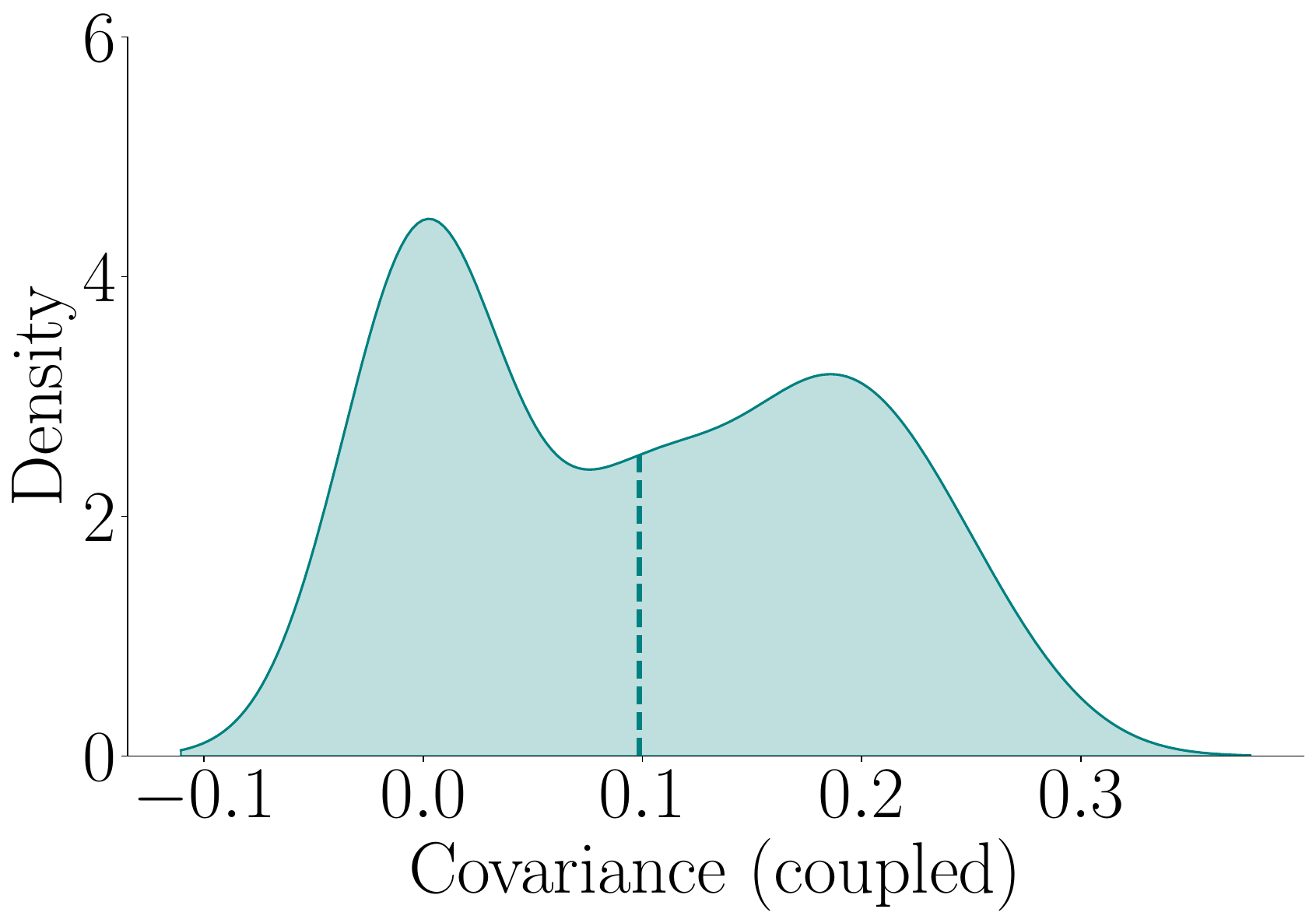} &
    \includegraphics[width=0.23\linewidth]{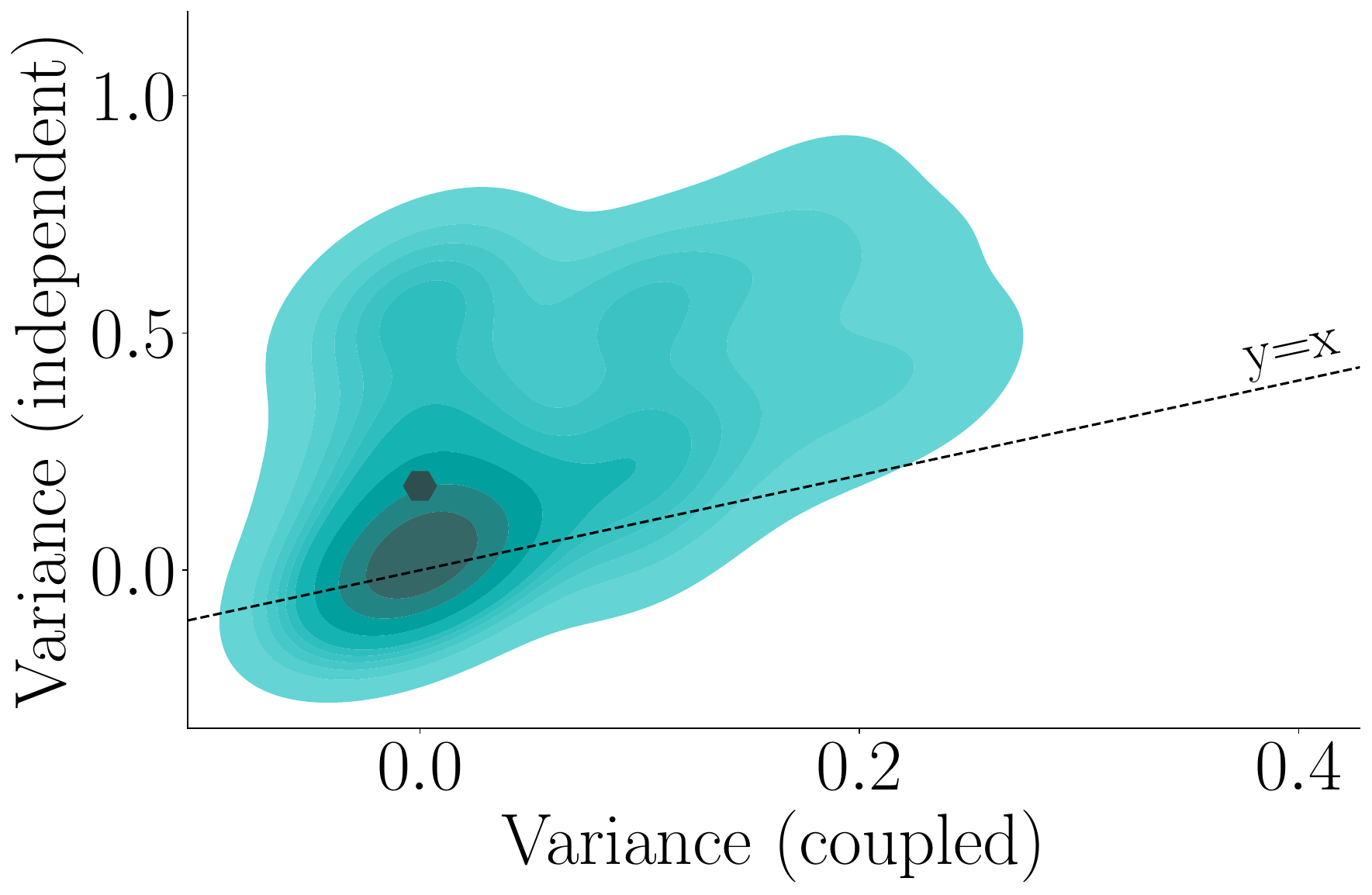} &
    \includegraphics[width=0.23\linewidth]{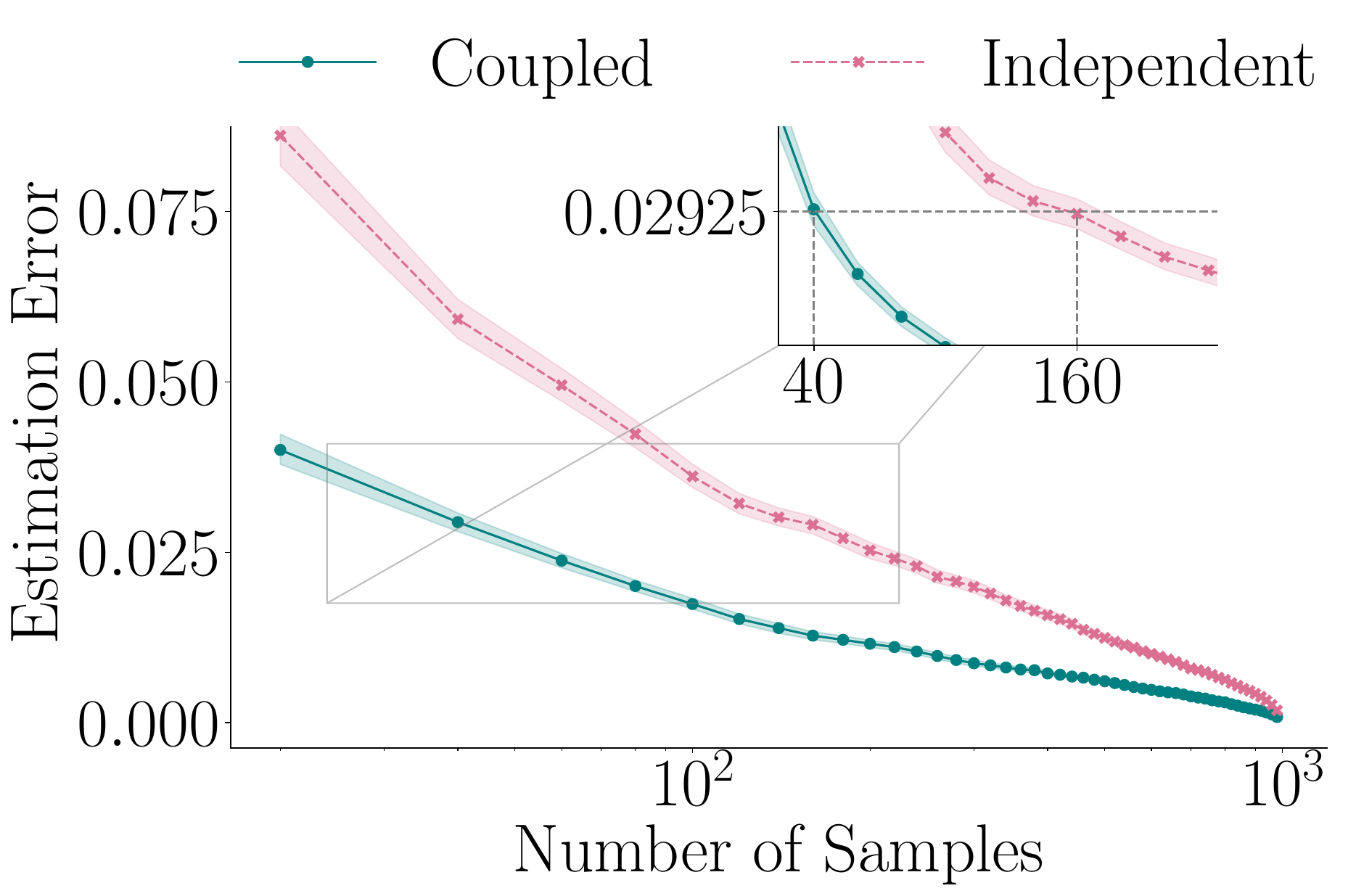} \\ \\
    (a) Score covariance & (b) Variance of the score difference & (c) Estimation error vs. \# samples \\ 
\end{tabular}
    \caption{\textbf{Comparison between four pairs of LLMs in the \texttt{Llama} family on multiple-choice questions from the ``college computer science'' knowledge area of the MMLU dataset.}
    Panels in column (a) show the kernel density estimate (KDE) of the covariance between the scores of the two LLMs on each question under coupled generation; the dashed lines correspond to average values. Panels in column (b) show the KDE of the variance of the difference between the scores of the LLMs on each question under coupled and independent generation; the highlighted points correspond to median values. Panels in column (c) show the absolute error in the estimation of the expected difference between the scores of the LLMs against the number of samples; for each point on the x-axis, we perform $1{,}000$ sub-samplings and shaded areas correspond to $95\%$ confidence intervals.}
    \label{fig:mmlu-first-4}
\end{figure}
\vspace{-0.2cm}
\begin{figure}[h]
\centering
\begin{tabular}{c c c}
    \multicolumn{3}{c}{\texttt{8B} vs. \texttt{bnb-4bit}}\\
    \includegraphics[width=0.23\linewidth]{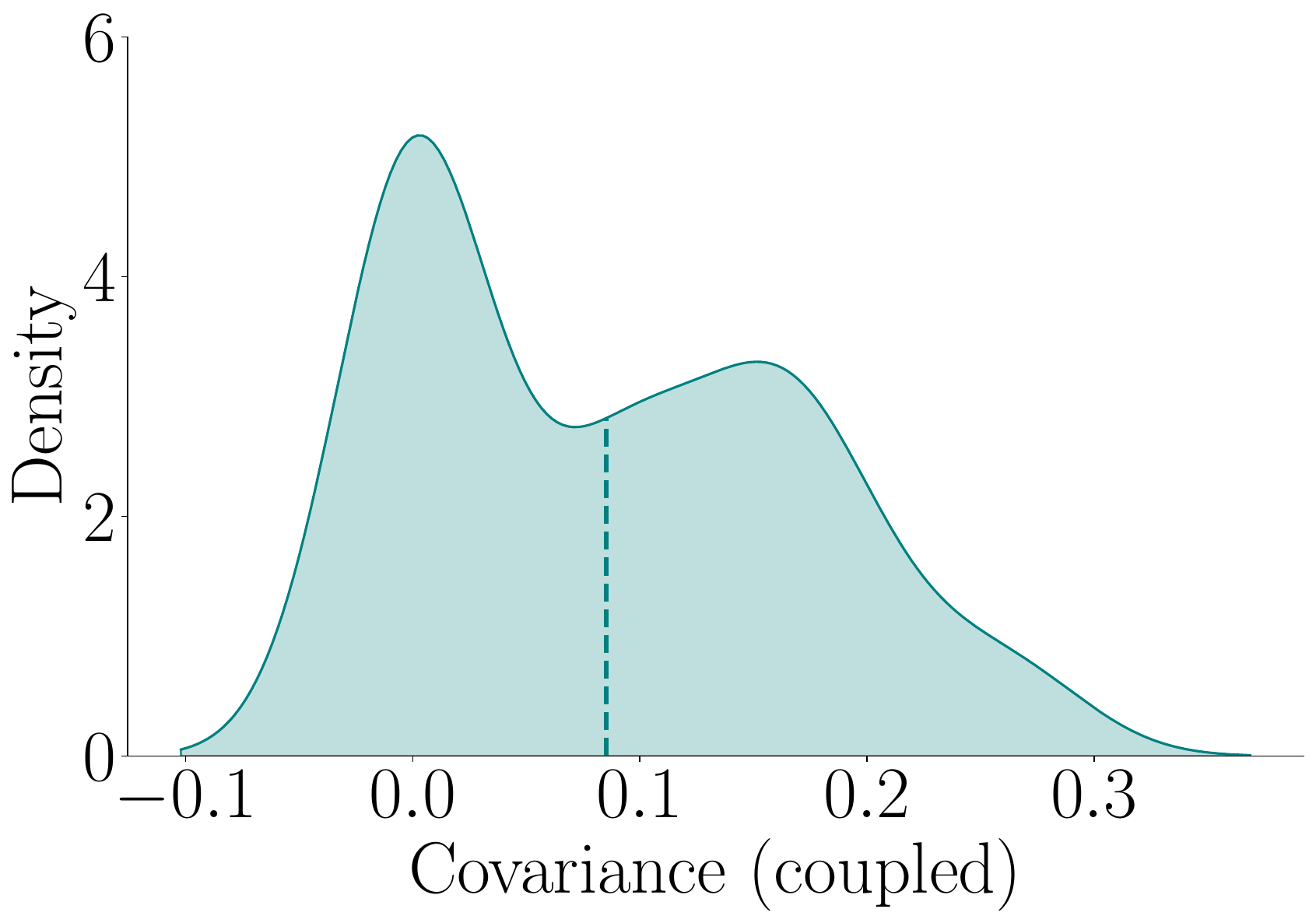} &
    \includegraphics[width=0.23\linewidth]{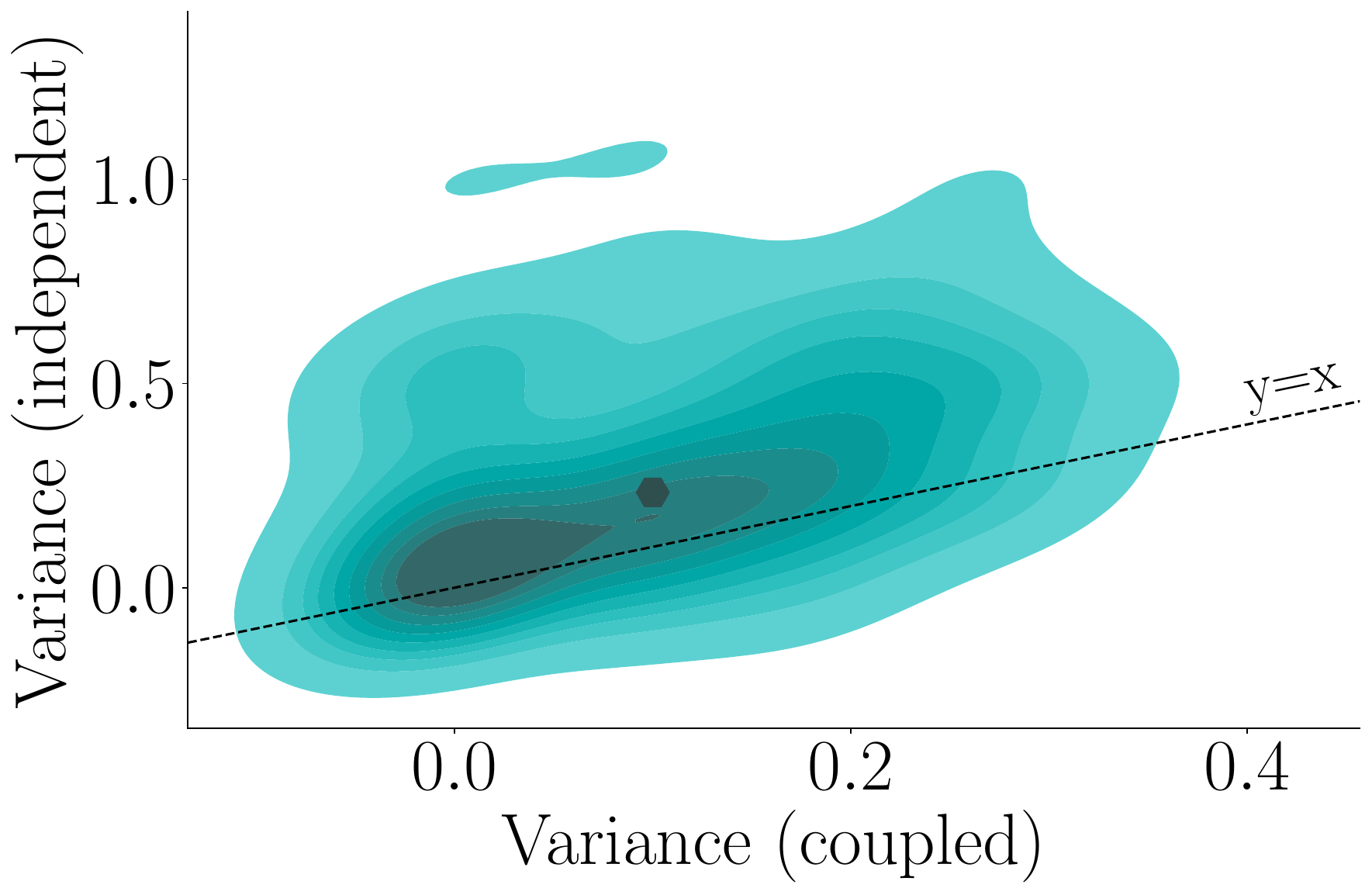} &
    \includegraphics[width=0.23\linewidth]{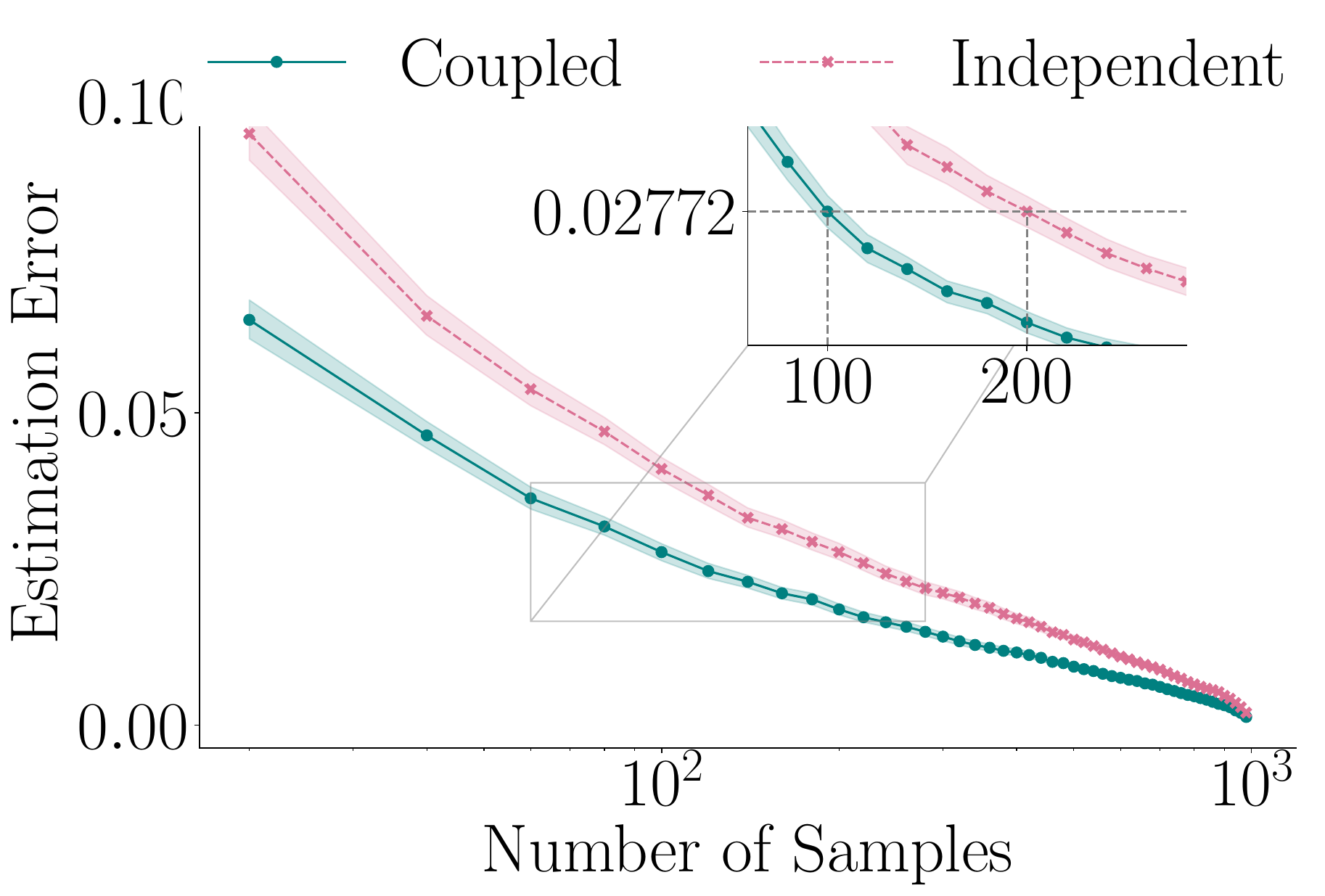} \\ \\
    \multicolumn{3}{c}{\texttt{8B} vs. \texttt{AWQ-INT4}}\\
    \includegraphics[width=0.23\linewidth]{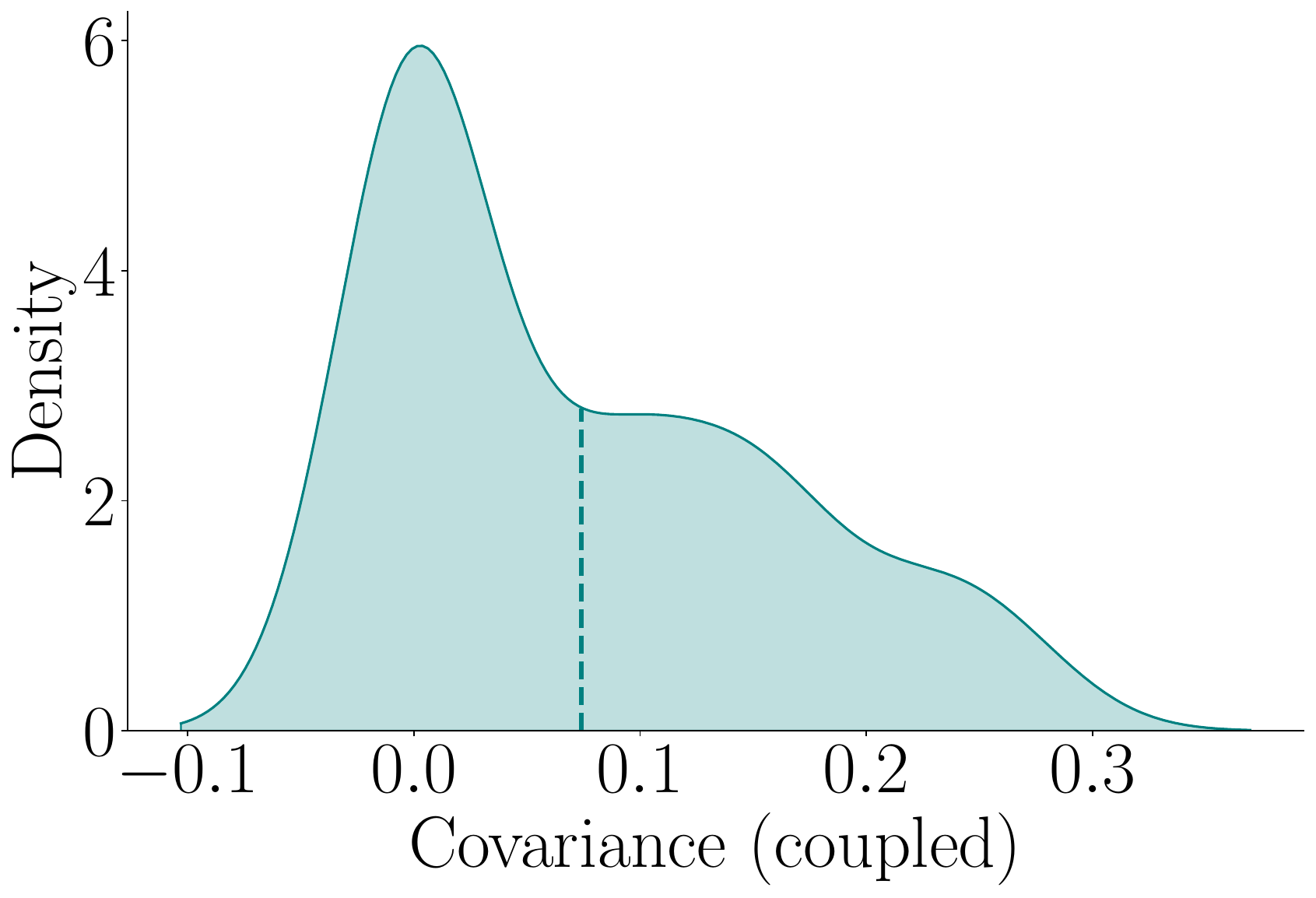} &
    \includegraphics[width=0.23\linewidth]{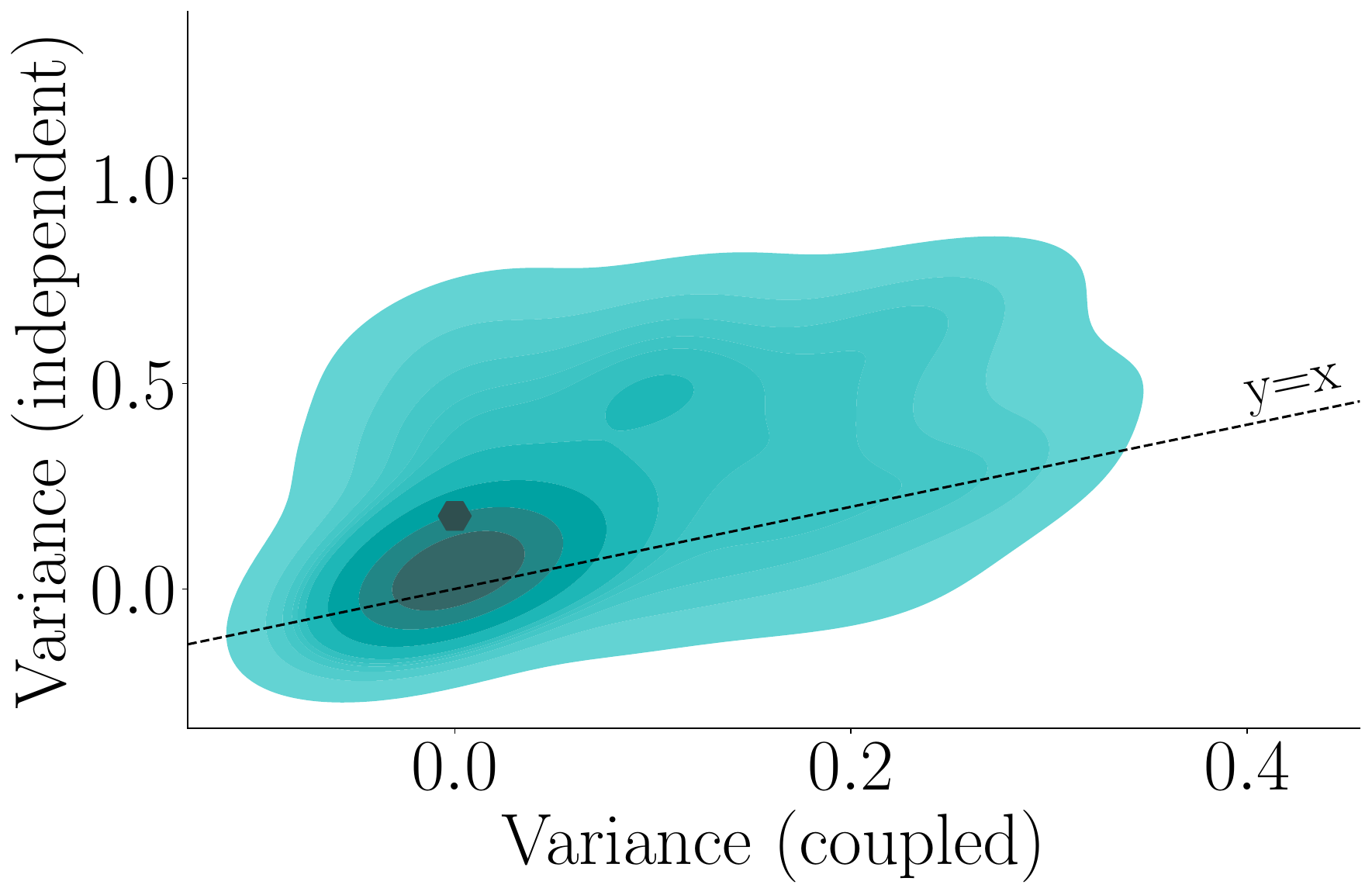} &
    \includegraphics[width=0.23\linewidth]{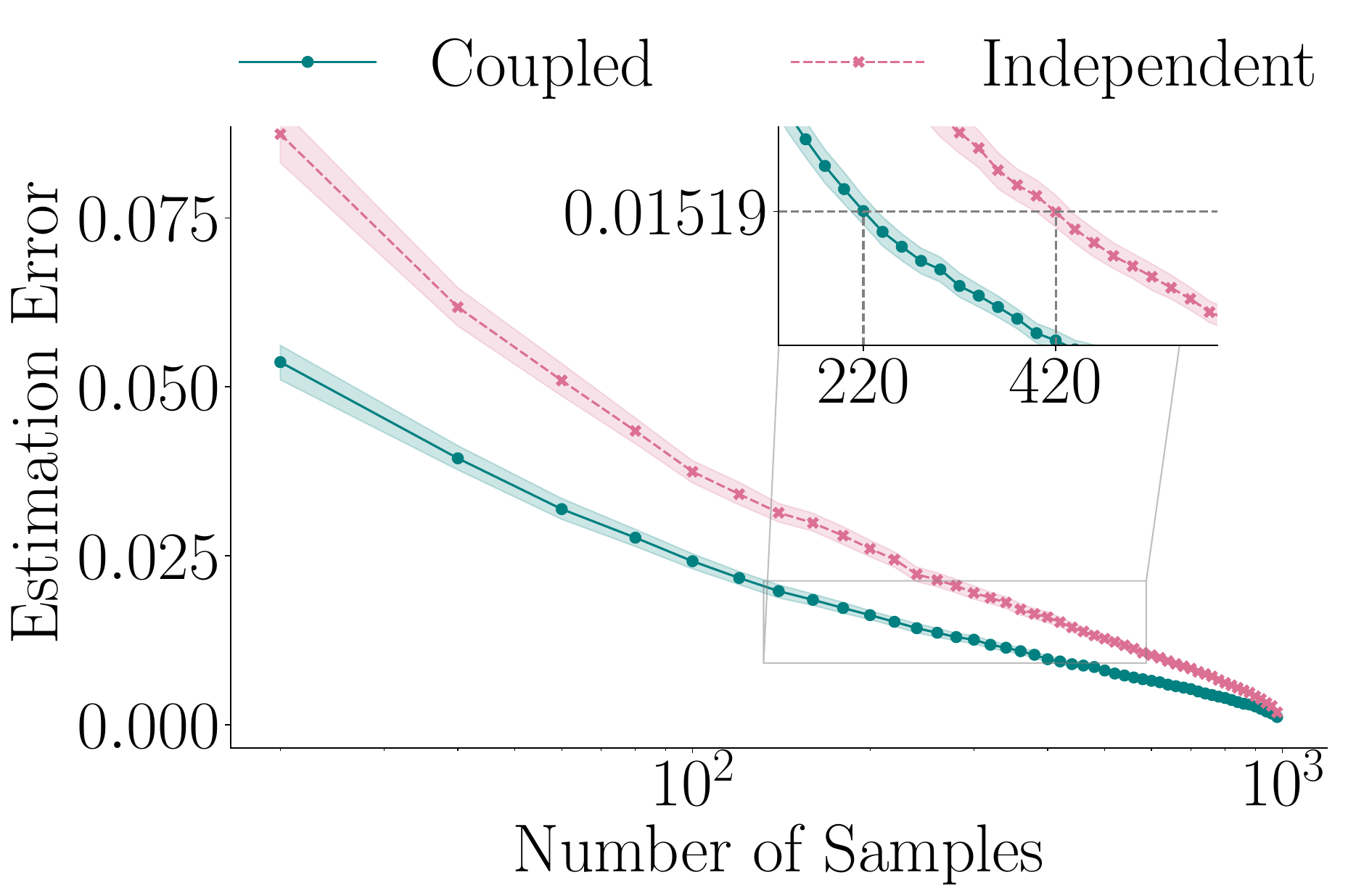} \\ \\
    \multicolumn{3}{c}{\texttt{3B} vs. \texttt{bnb-8bit}}\\
    \includegraphics[width=0.23\linewidth]{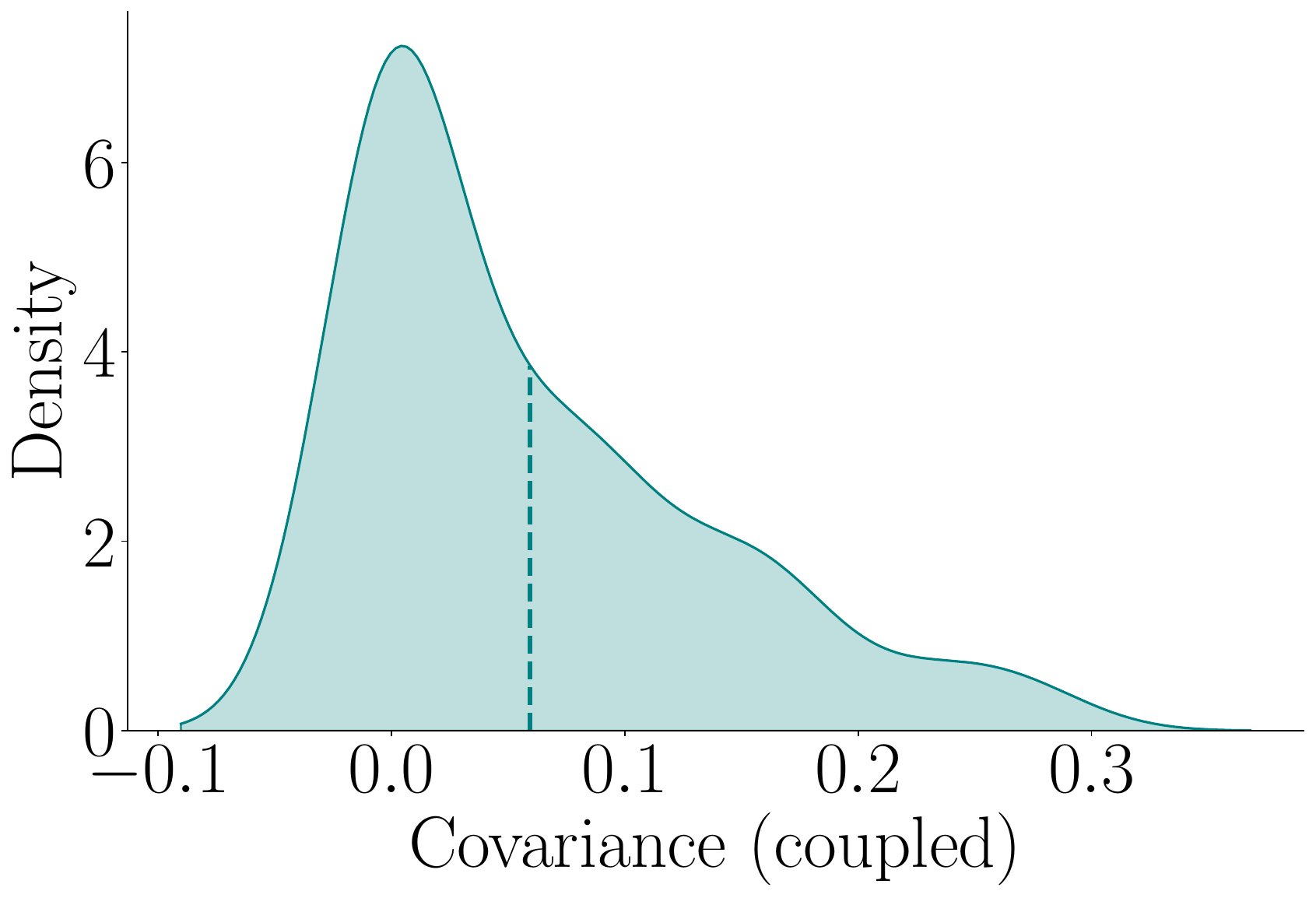} &
    \includegraphics[width=0.23\linewidth]{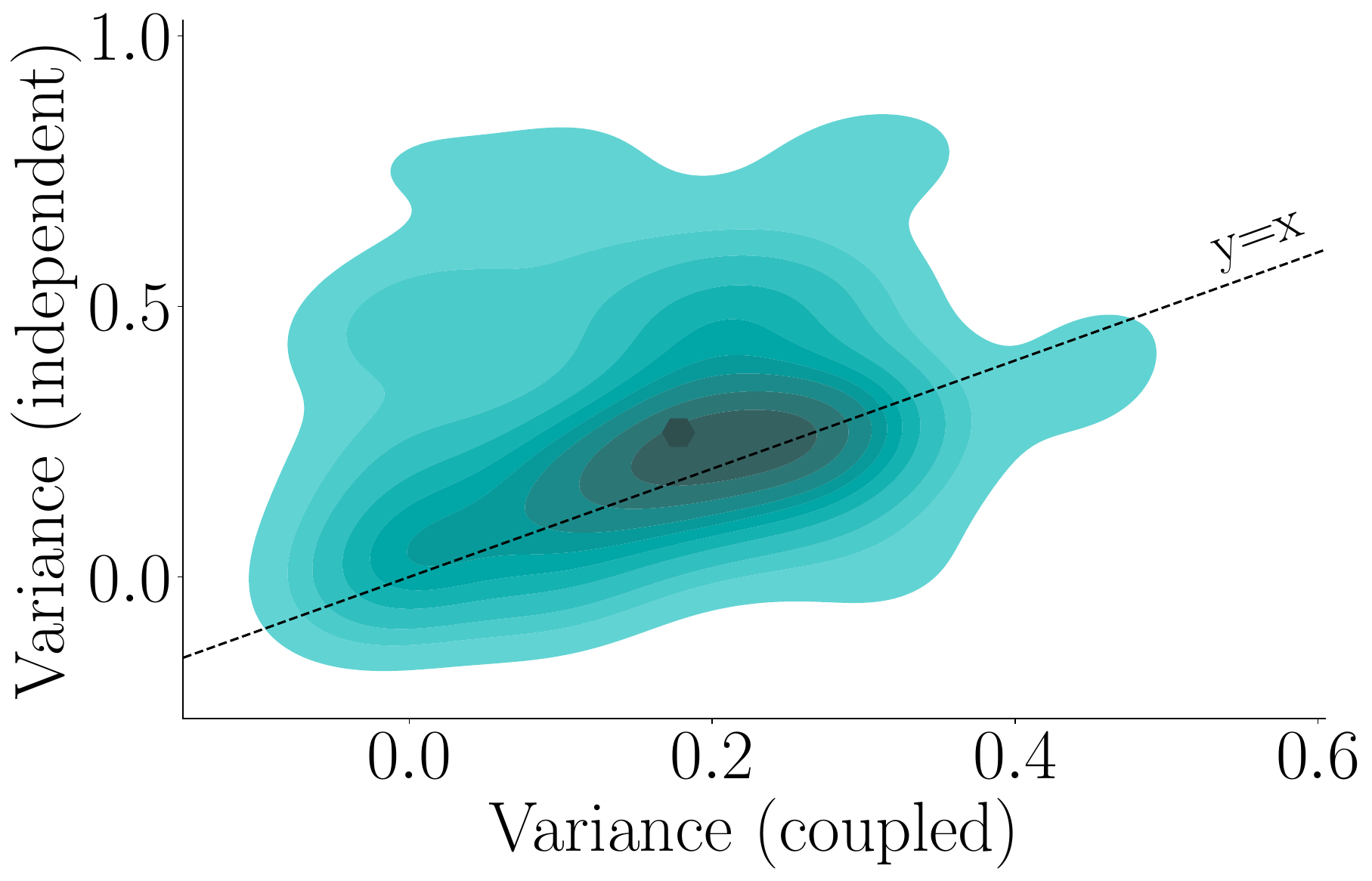} &
    \includegraphics[width=0.23\linewidth]{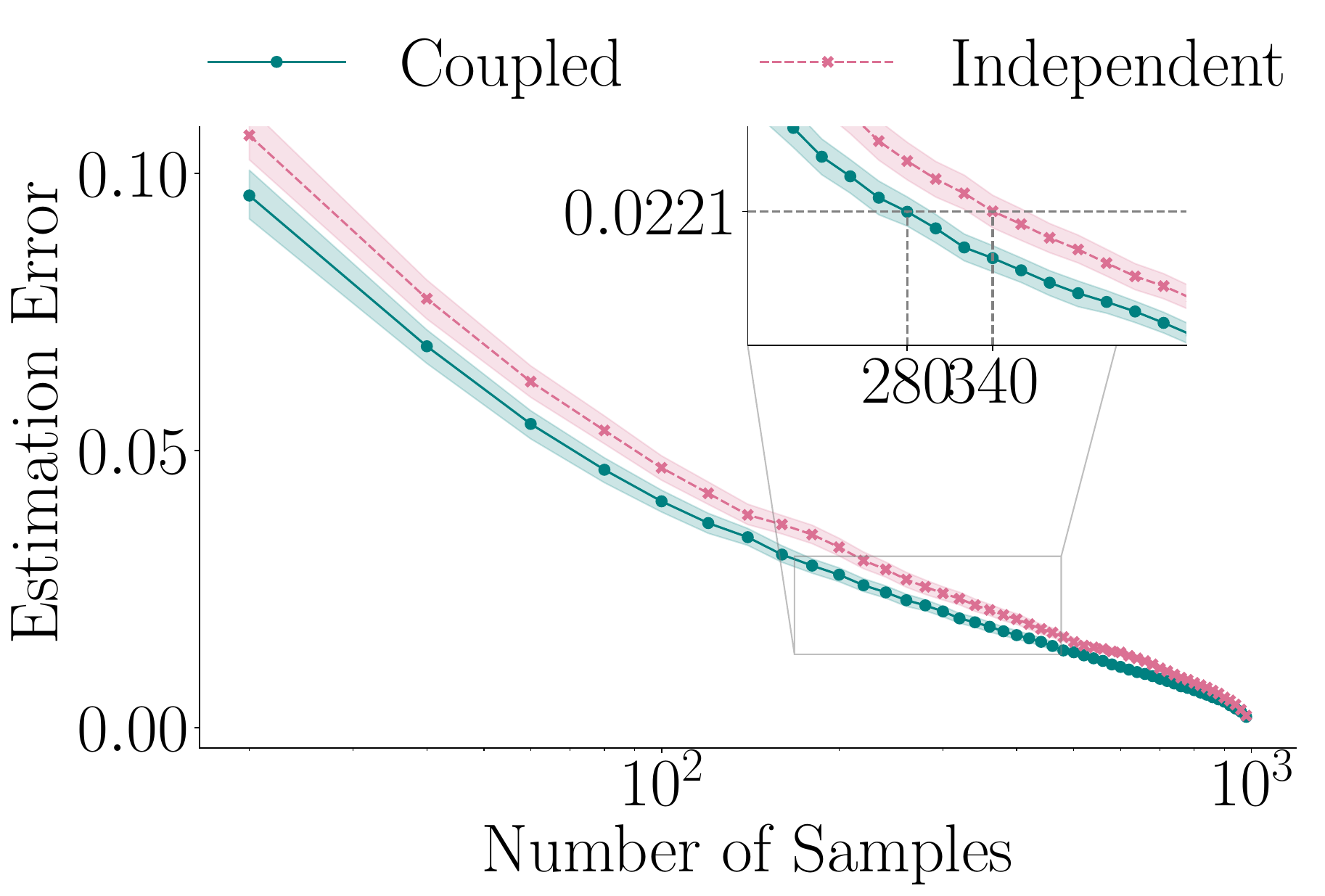} \\ \\
 \multicolumn{3}{c}{\texttt{3B} vs. \texttt{bnb-4bit}}\\
    \includegraphics[width=0.23\linewidth]{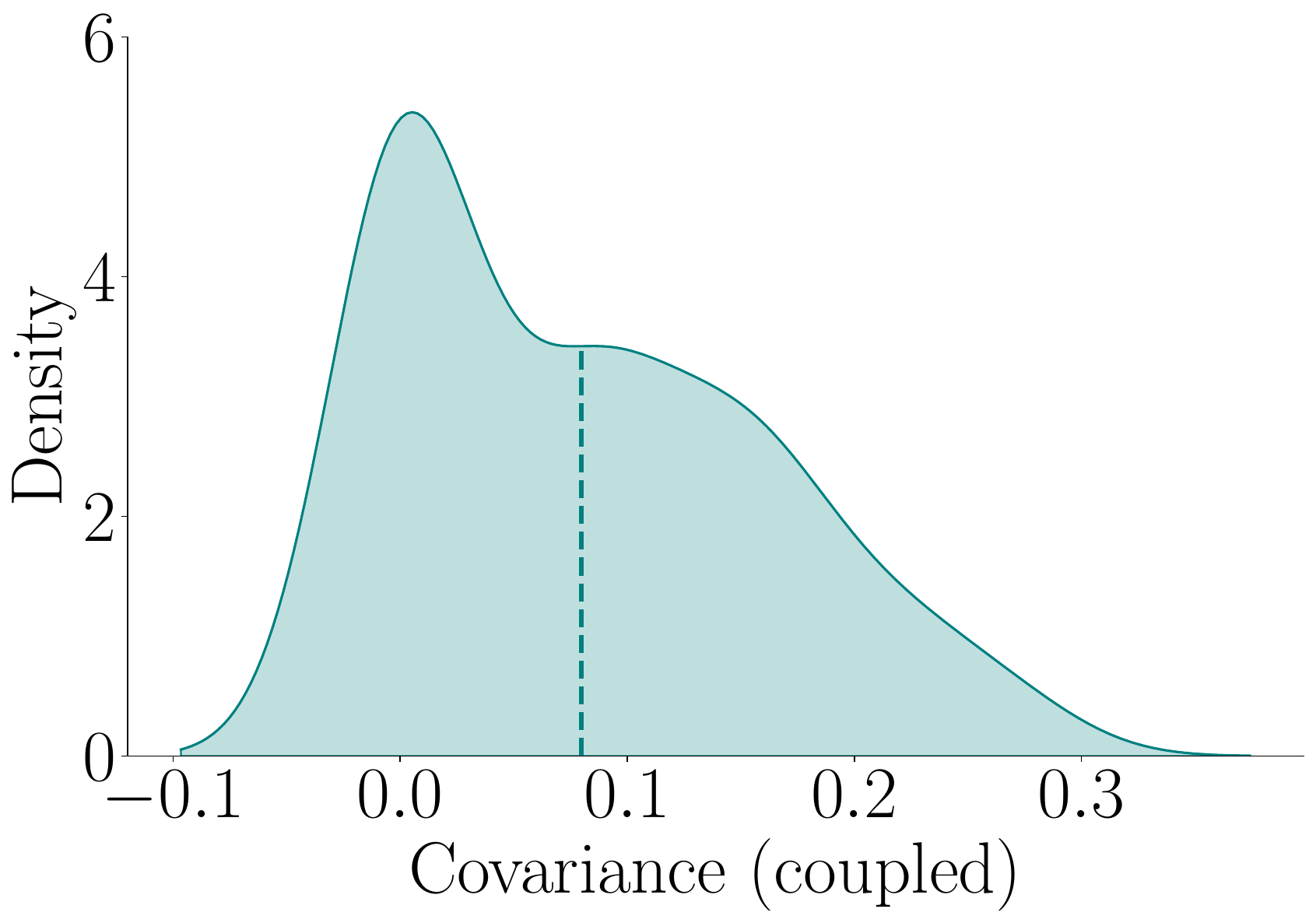} &
    \includegraphics[width=0.23\linewidth]{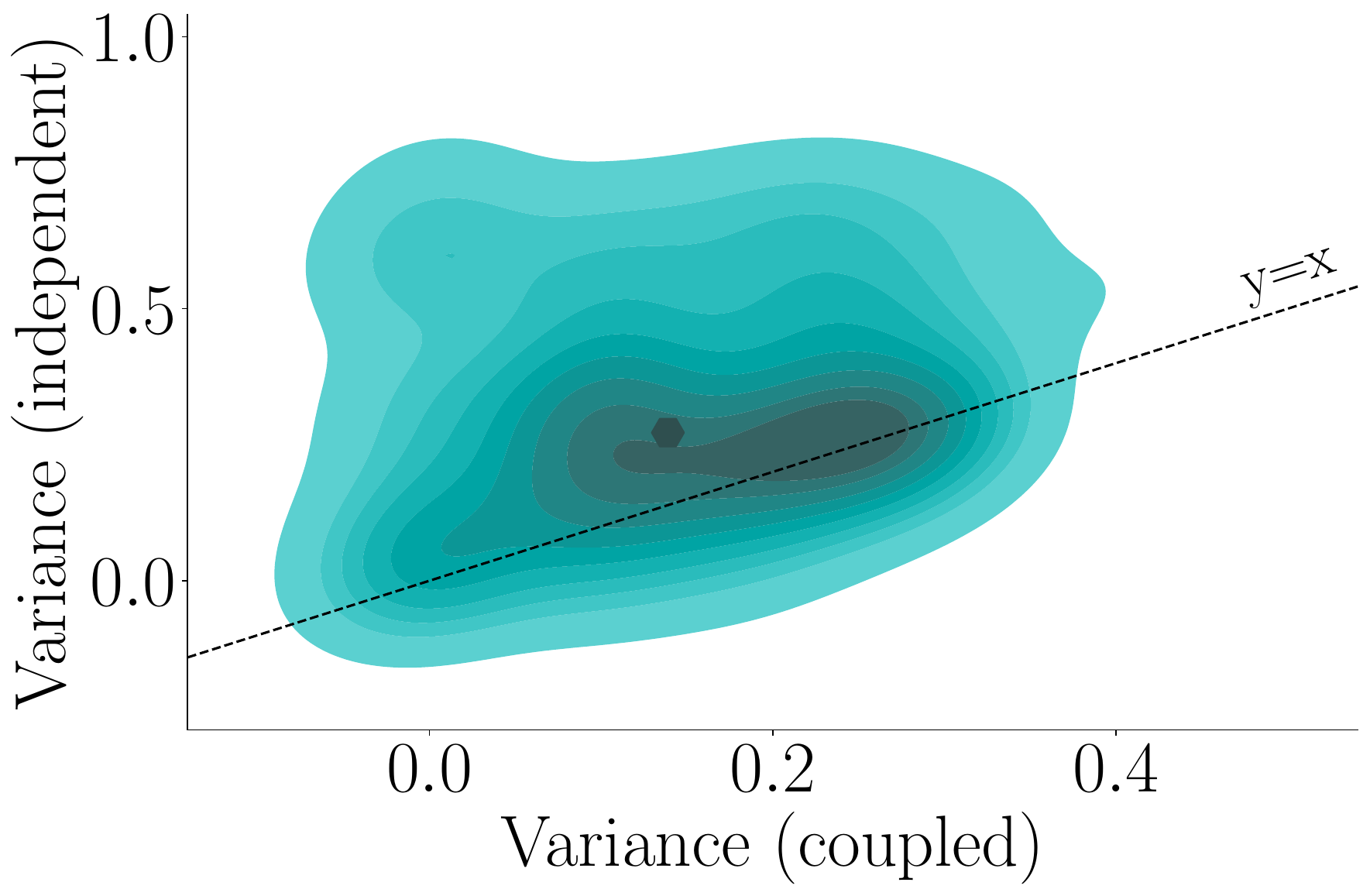} &
    \includegraphics[width=0.23\linewidth]{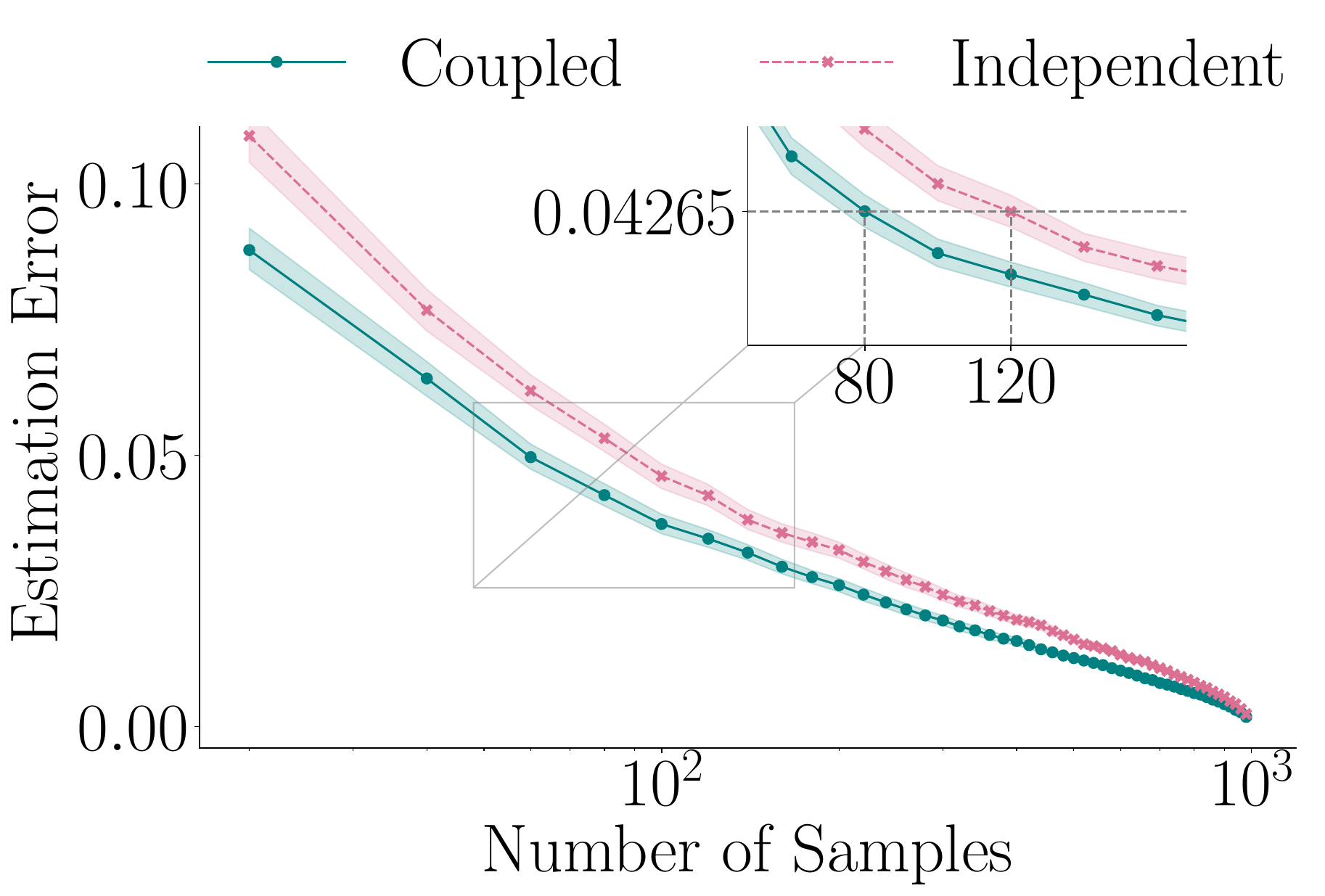} \\ \\

    (a) Score covariance & (b) Variance of the score difference & (c) Estimation error vs. \# samples \\ 
    
\end{tabular}
    \caption{\textbf{Comparison between four pairs of LLMs in the \texttt{Llama} family on multiple-choice questions from the ``college computer science'' knowledge area of the MMLU dataset.}
    Panels in column (a) show the kernel density estimate (KDE) of the covariance between the scores of the two LLMs on each question under coupled generation; the dashed lines correspond to average values. Panels in column (b) show the KDE of the variance of the difference between the scores of the LLMs on each question under coupled and independent generation; the highlighted points correspond to median values. Panels in column (c) show the absolute error in the estimation of the expected difference between the scores of the LLMs against the number of samples; for each point on the x-axis, we perform $1{,}000$ sub-samplings and shaded areas correspond to $95\%$ confidence intervals.}
    \label{fig:mmlu-second-4}
\end{figure}
\begin{figure}[h]
\centering
\begin{tabular}{c c c}
    \multicolumn{3}{c}{\texttt{3B} vs. \texttt{AWQ-INT4}}\\
    \includegraphics[width=0.23\linewidth]{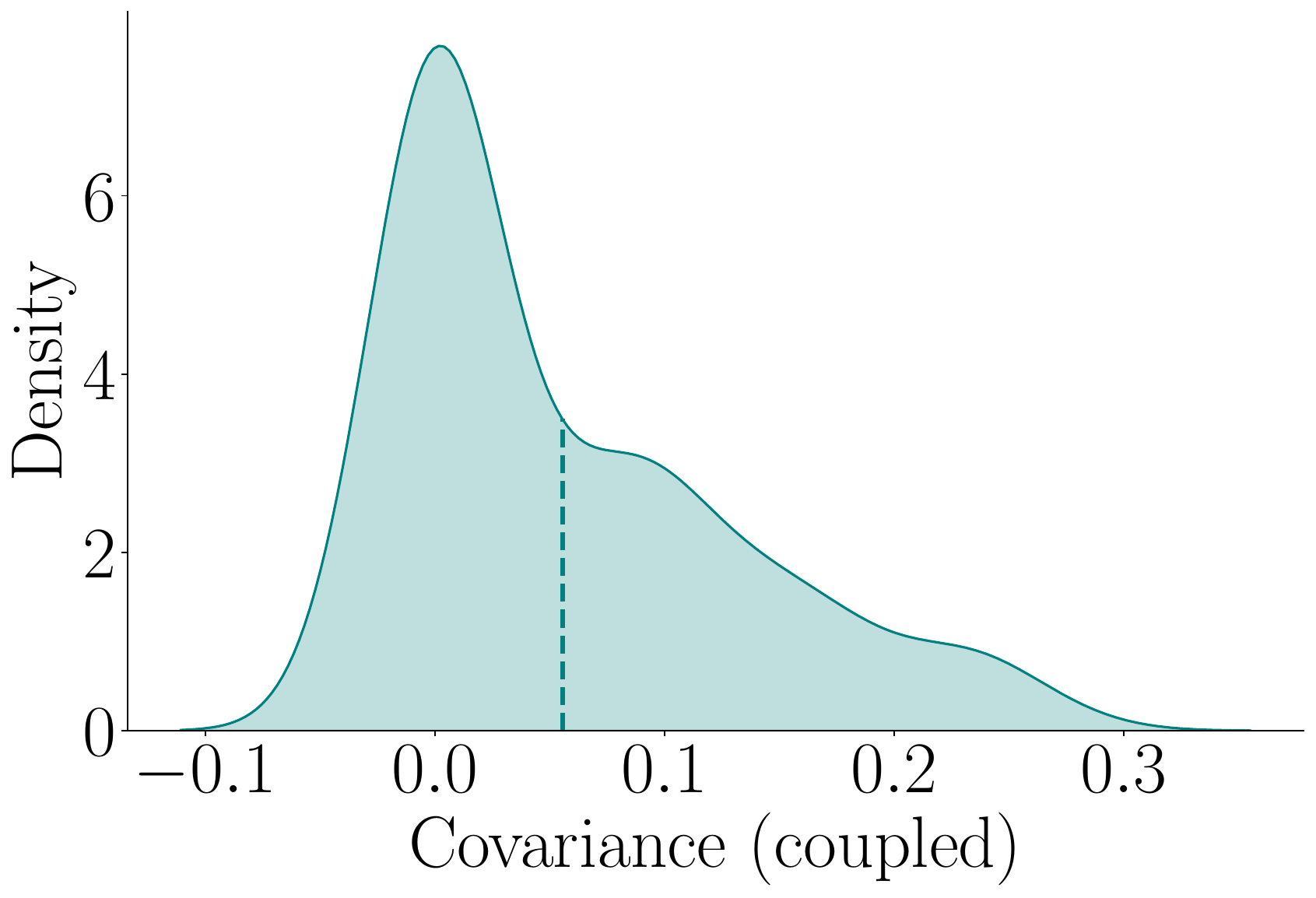} &
    \includegraphics[width=0.23\linewidth]{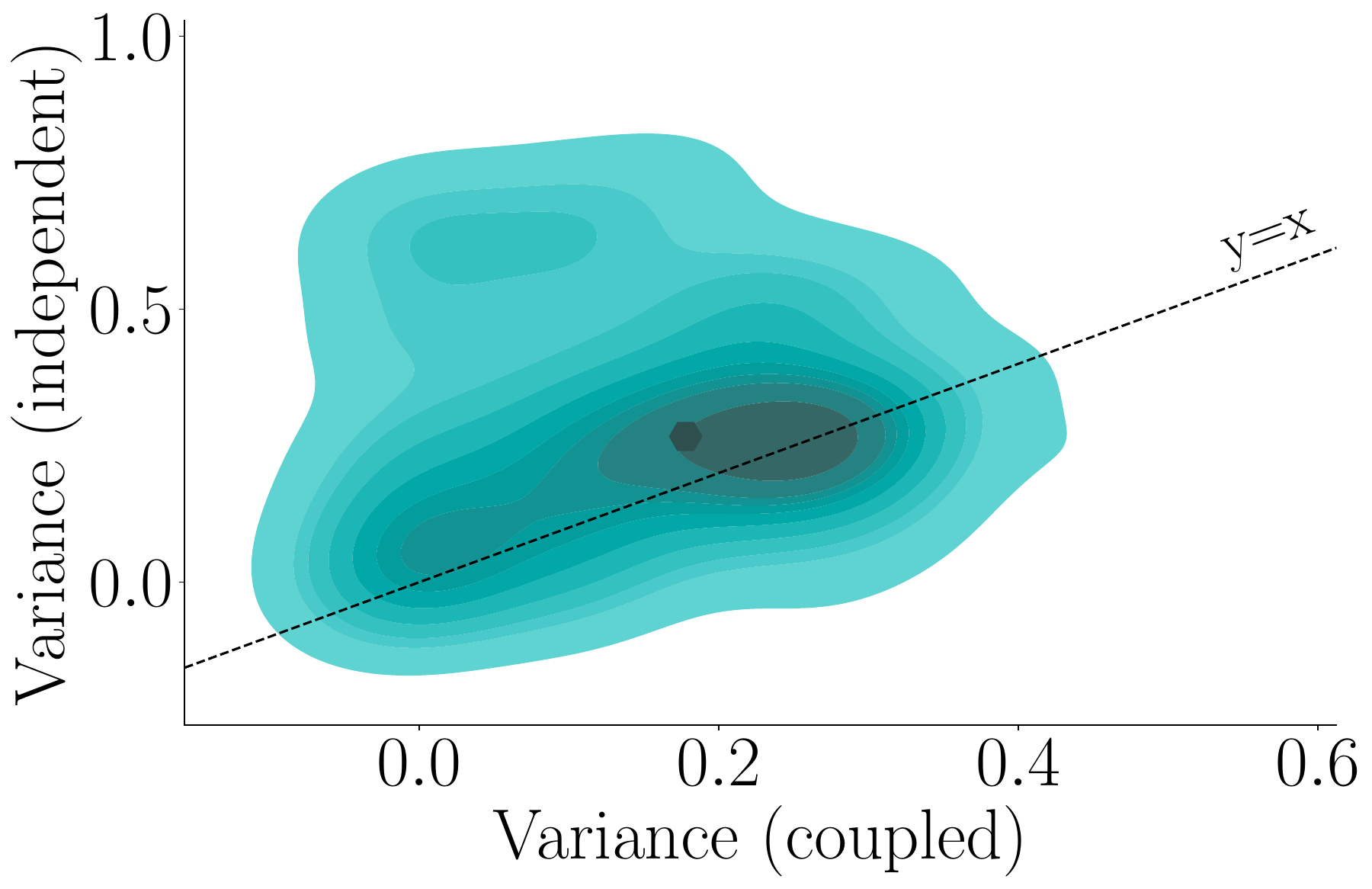} &
    \includegraphics[width=0.23\linewidth]{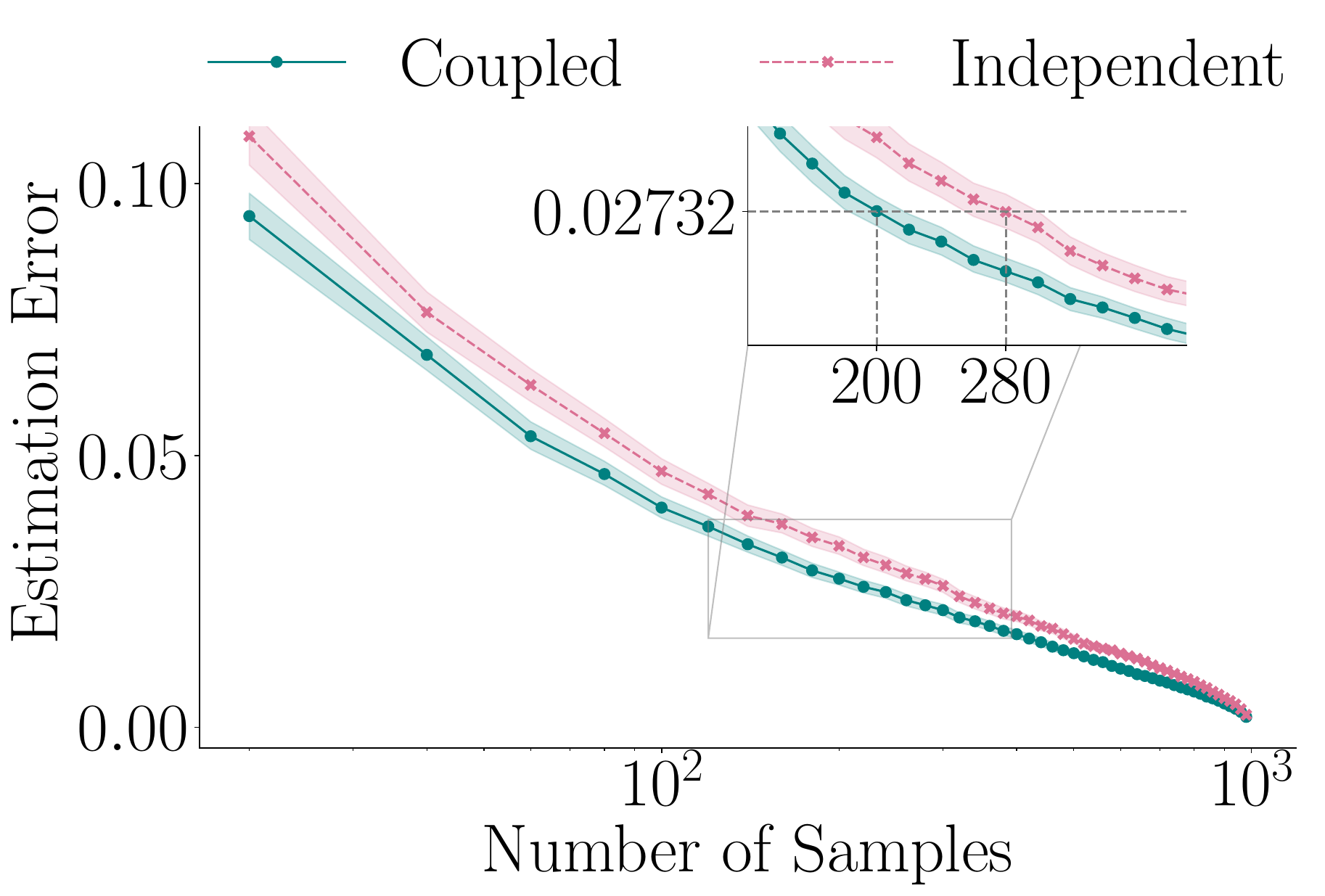} \\ \\
    \multicolumn{3}{c}{\texttt{1B} vs. \texttt{bnb-8bit}}\\
    \includegraphics[width=0.23\linewidth]{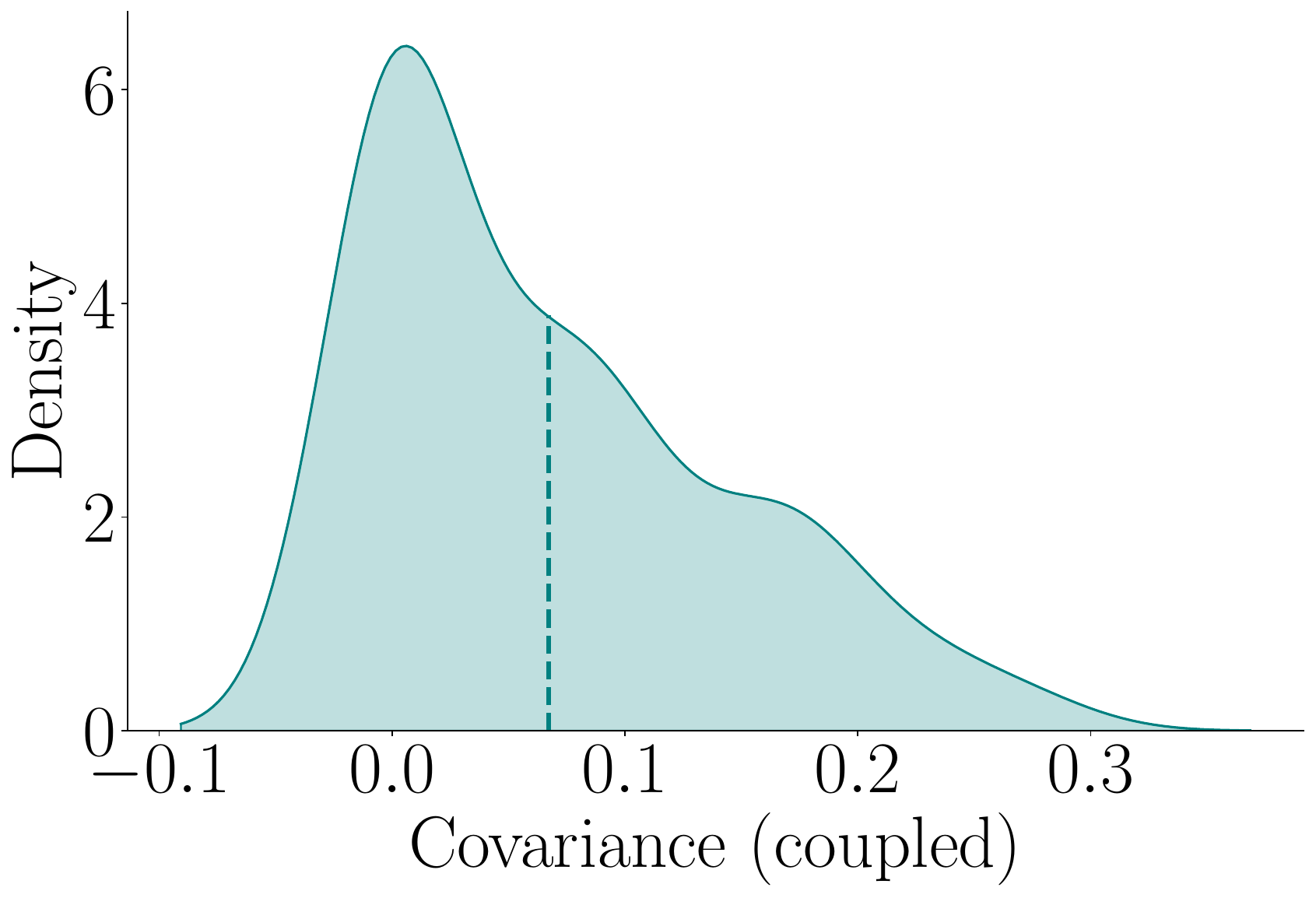} &
    \includegraphics[width=0.23\linewidth]{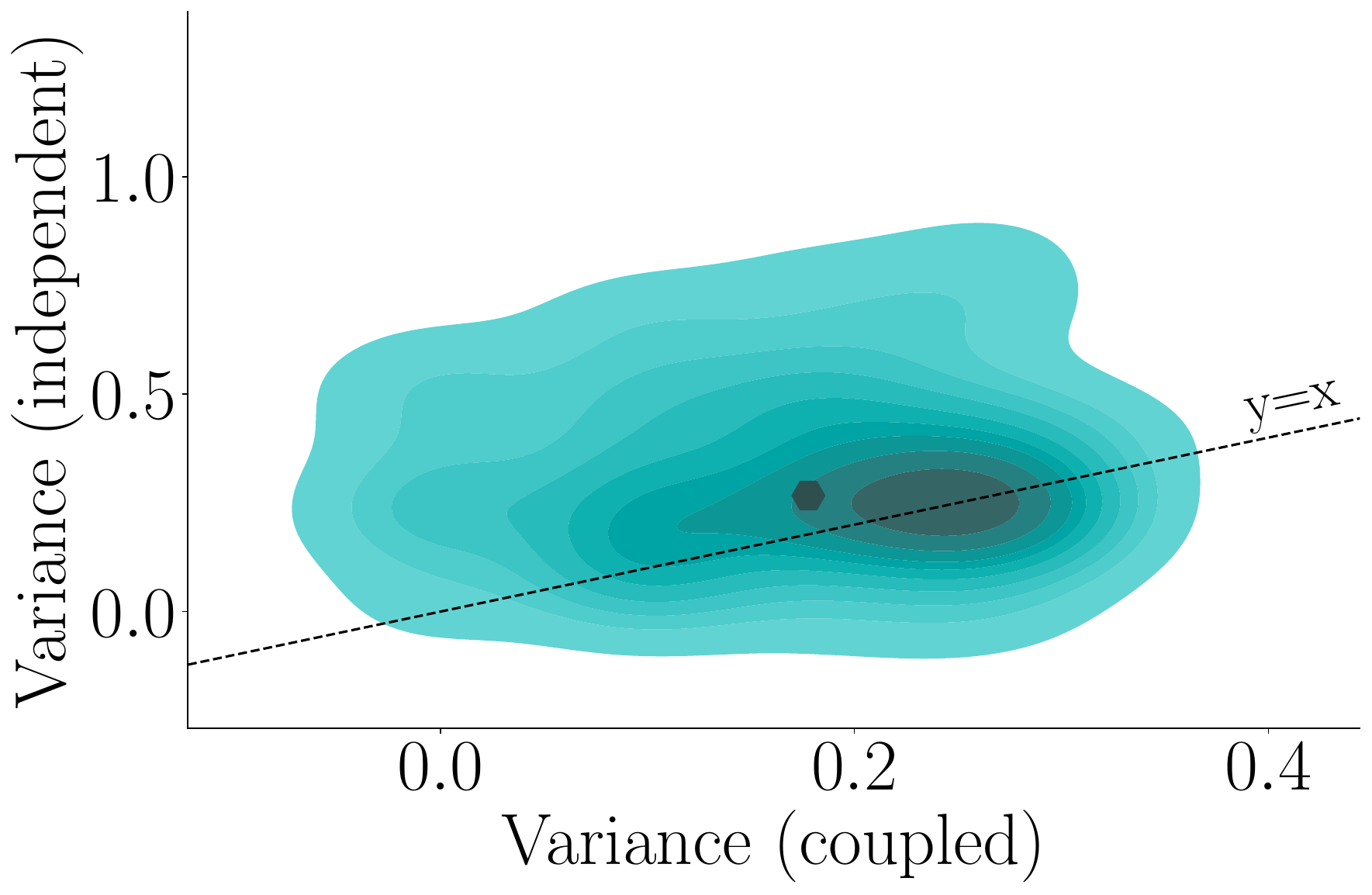} &
    \includegraphics[width=0.23\linewidth]{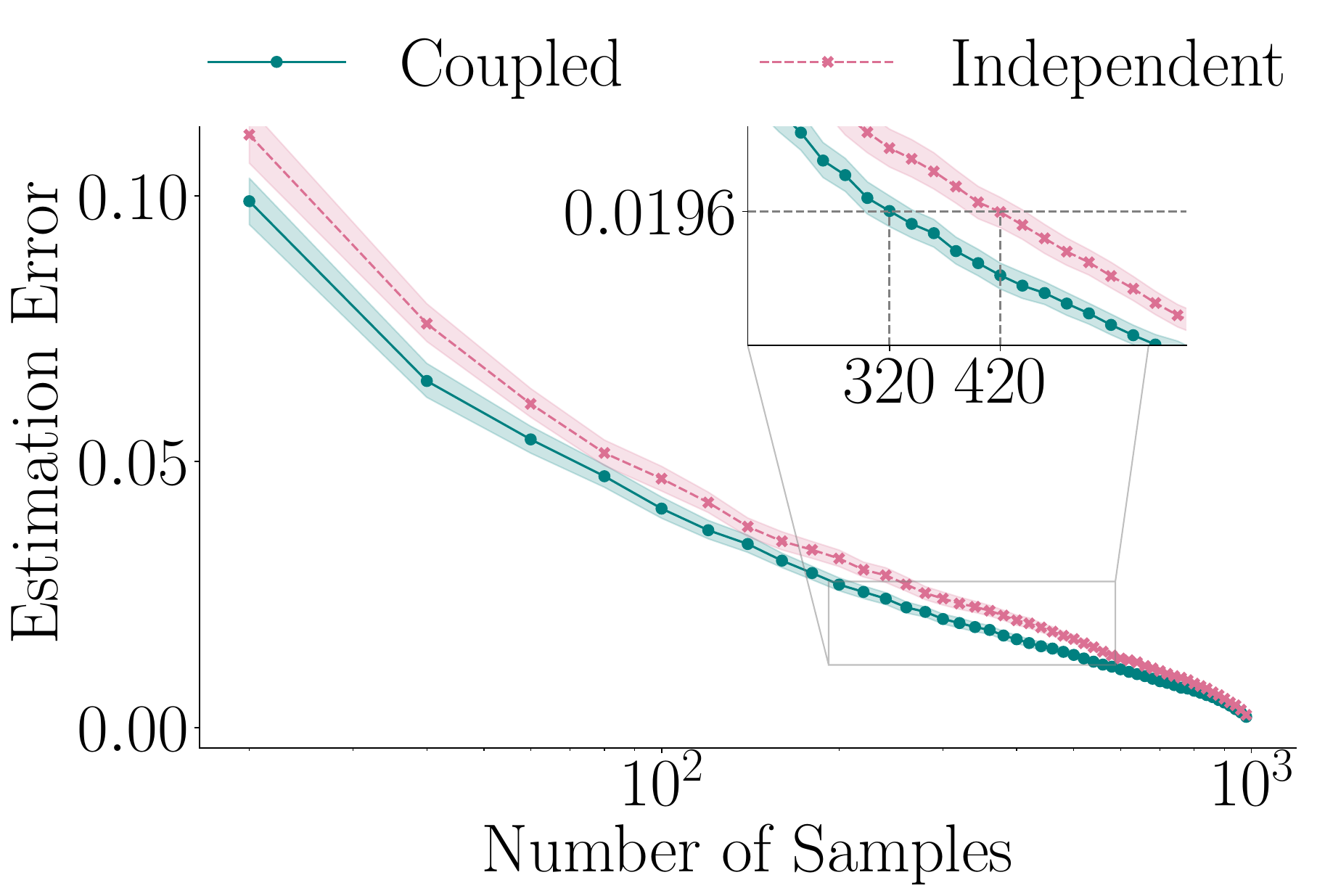} \\ \\
    \multicolumn{3}{c}{\texttt{1B} vs. \texttt{bnb-4bit}}\\
    \includegraphics[width=0.23\linewidth]{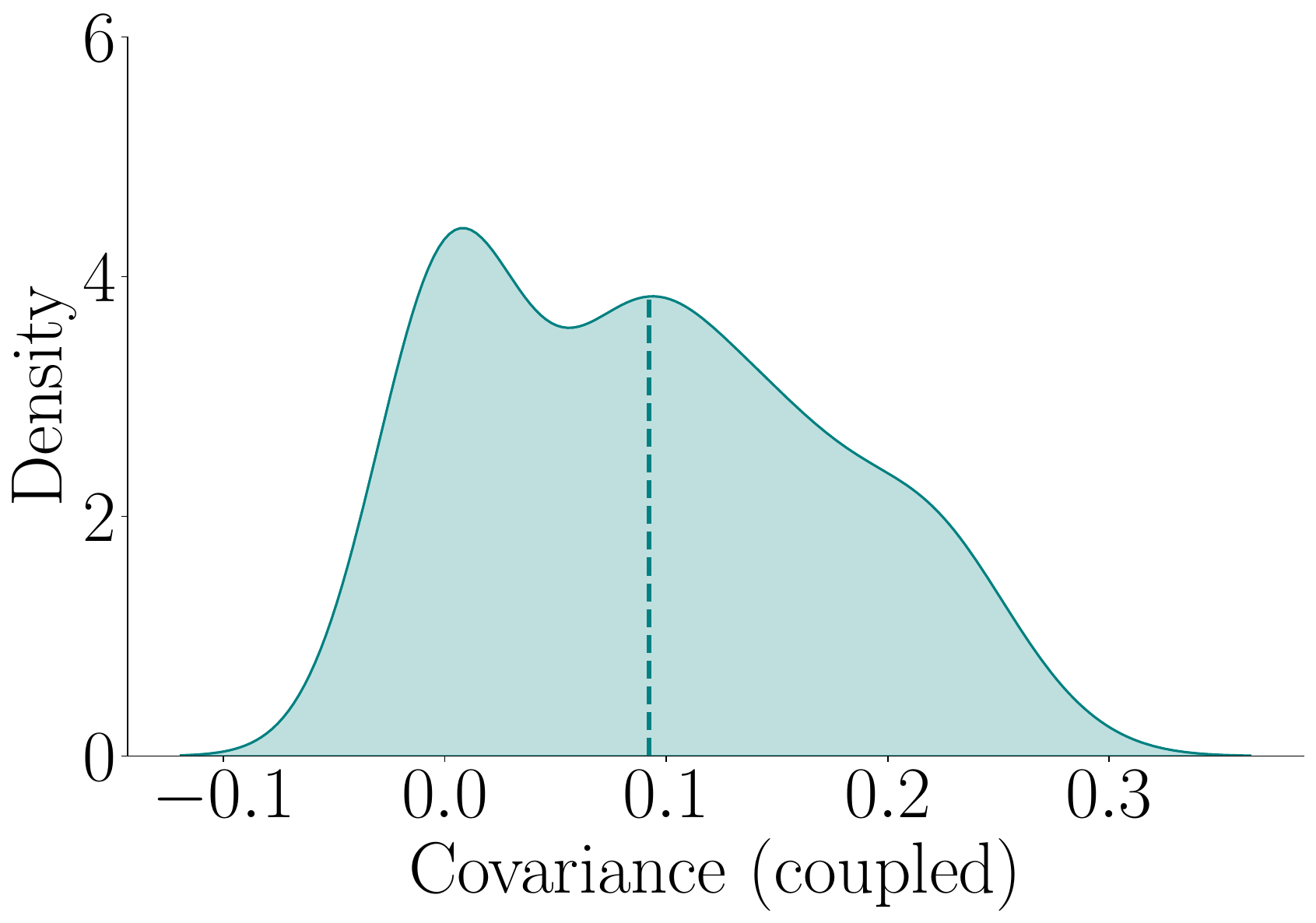} &
    \includegraphics[width=0.23\linewidth]{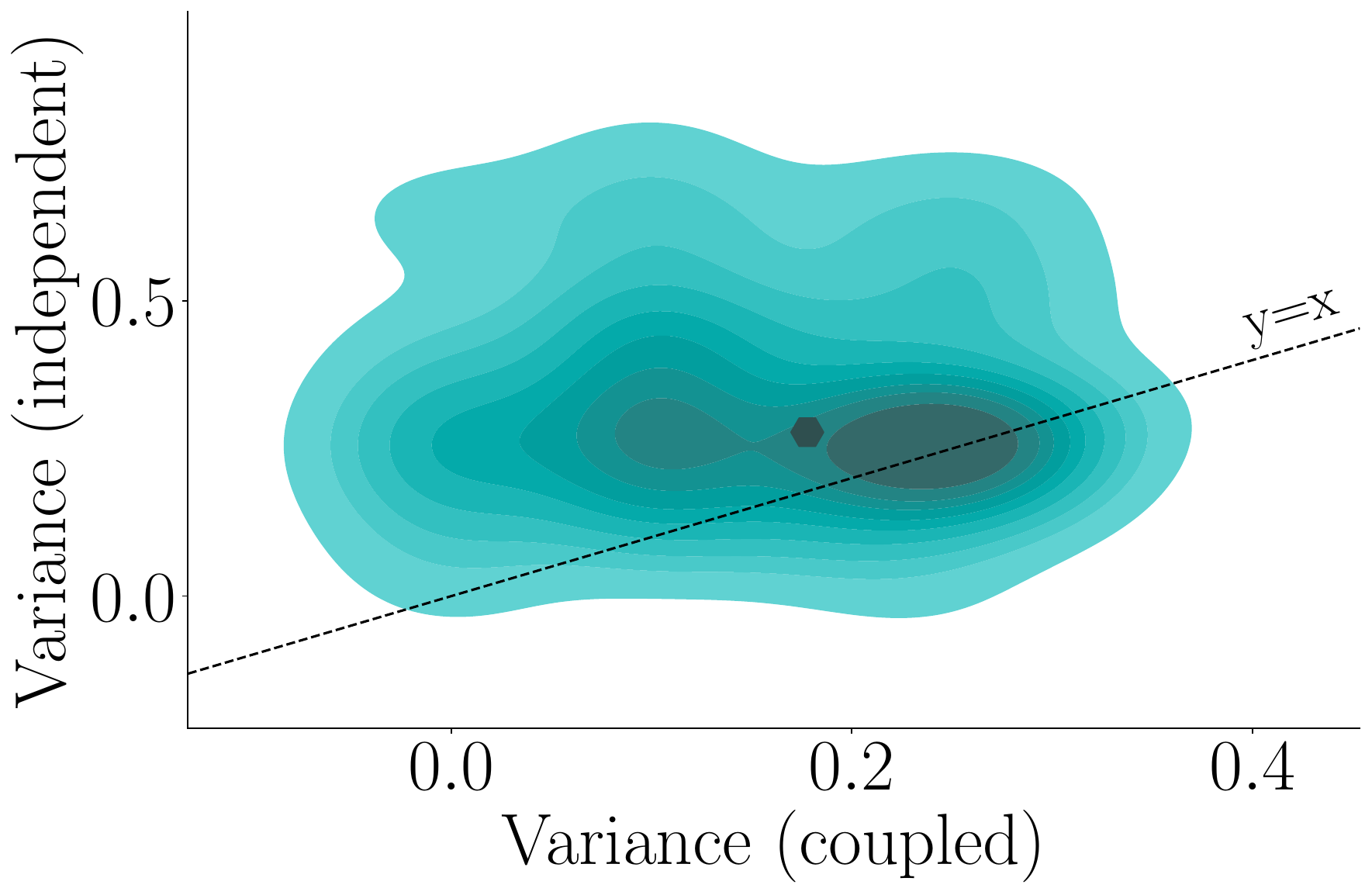} &
    \includegraphics[width=0.23\linewidth]{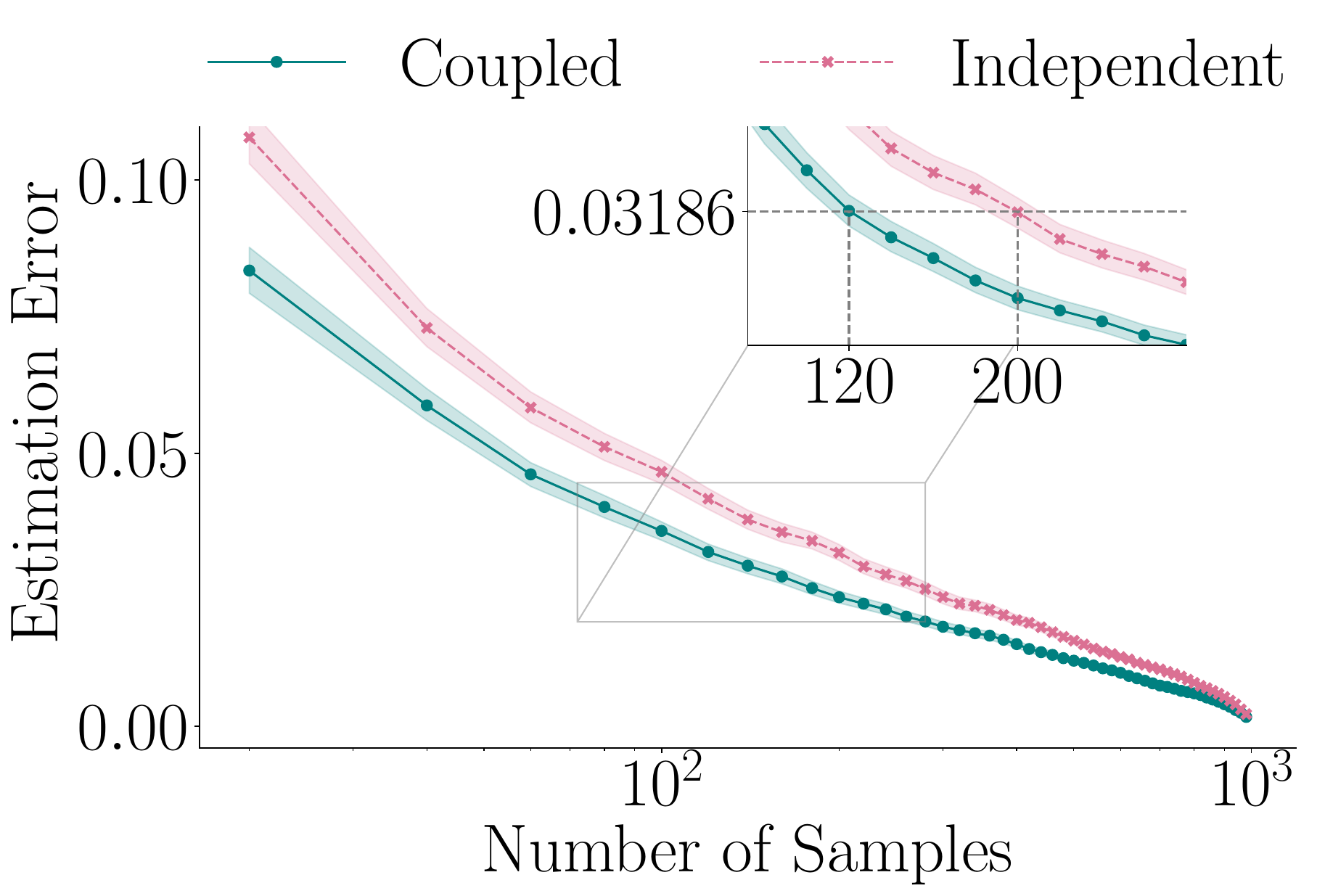} \\ \\
 \multicolumn{3}{c}{\texttt{1B} vs. \texttt{AWQ-INT4}}\\
    \includegraphics[width=0.23\linewidth]{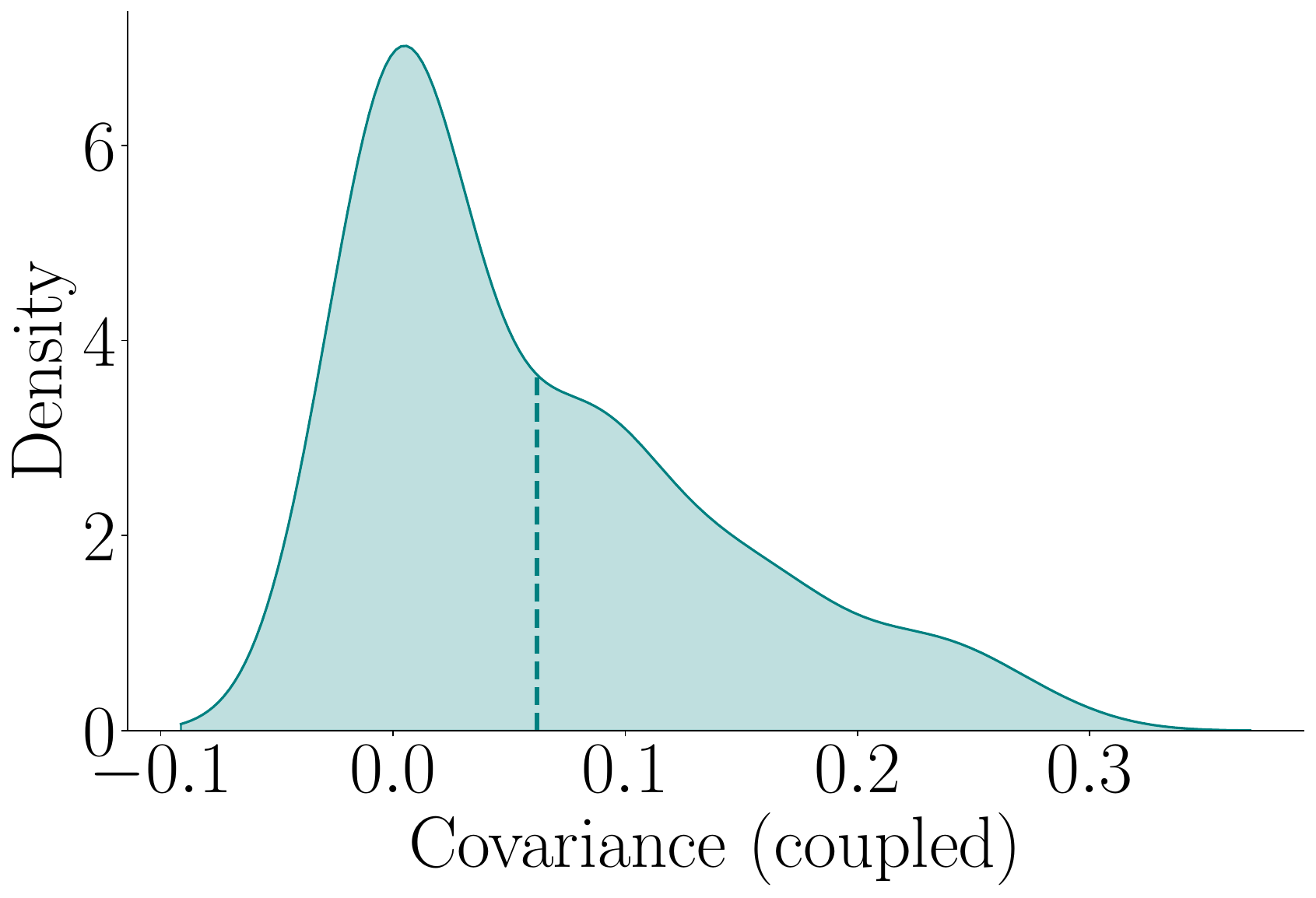} &
    \includegraphics[width=0.23\linewidth]{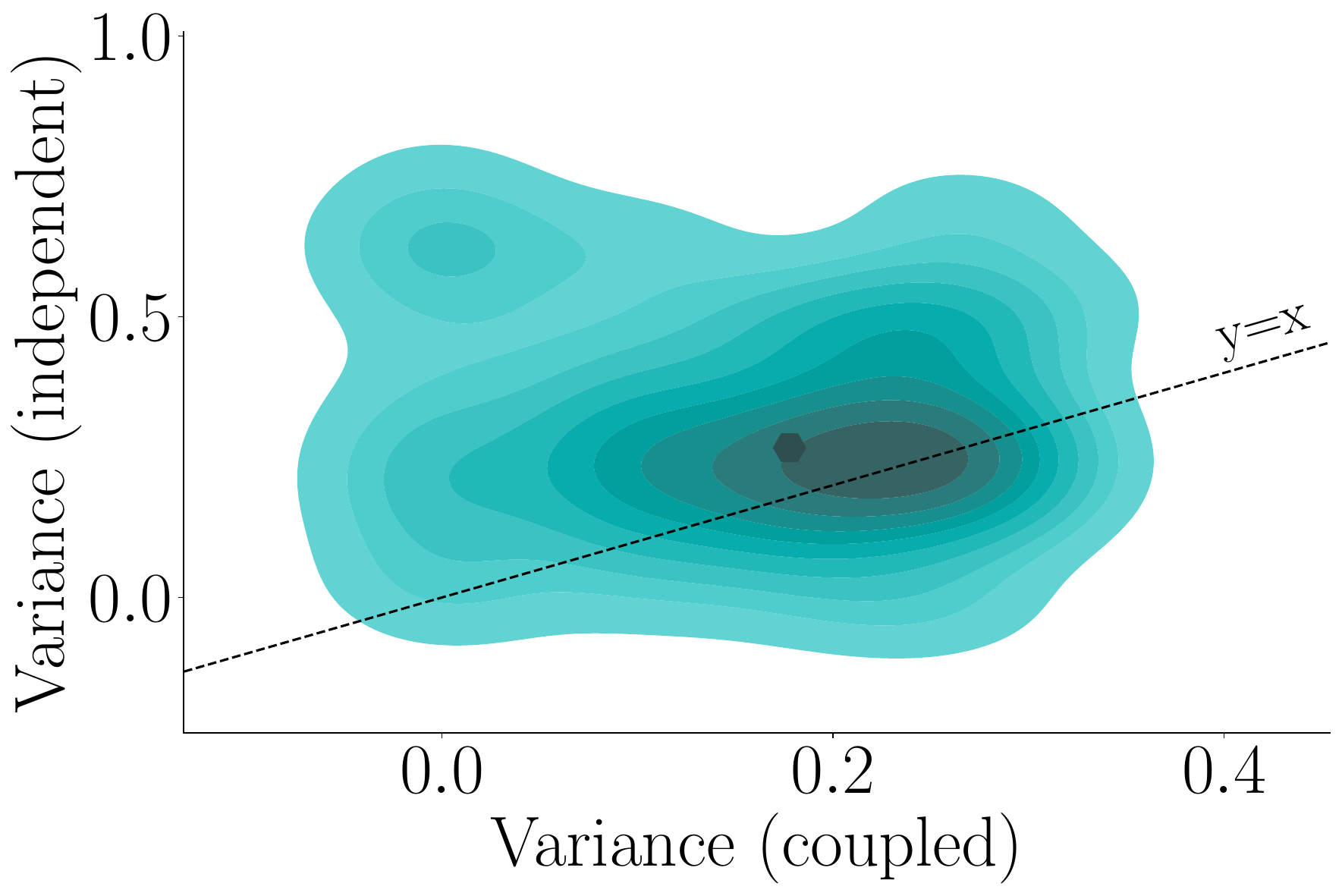} &
    \includegraphics[width=0.23\linewidth]{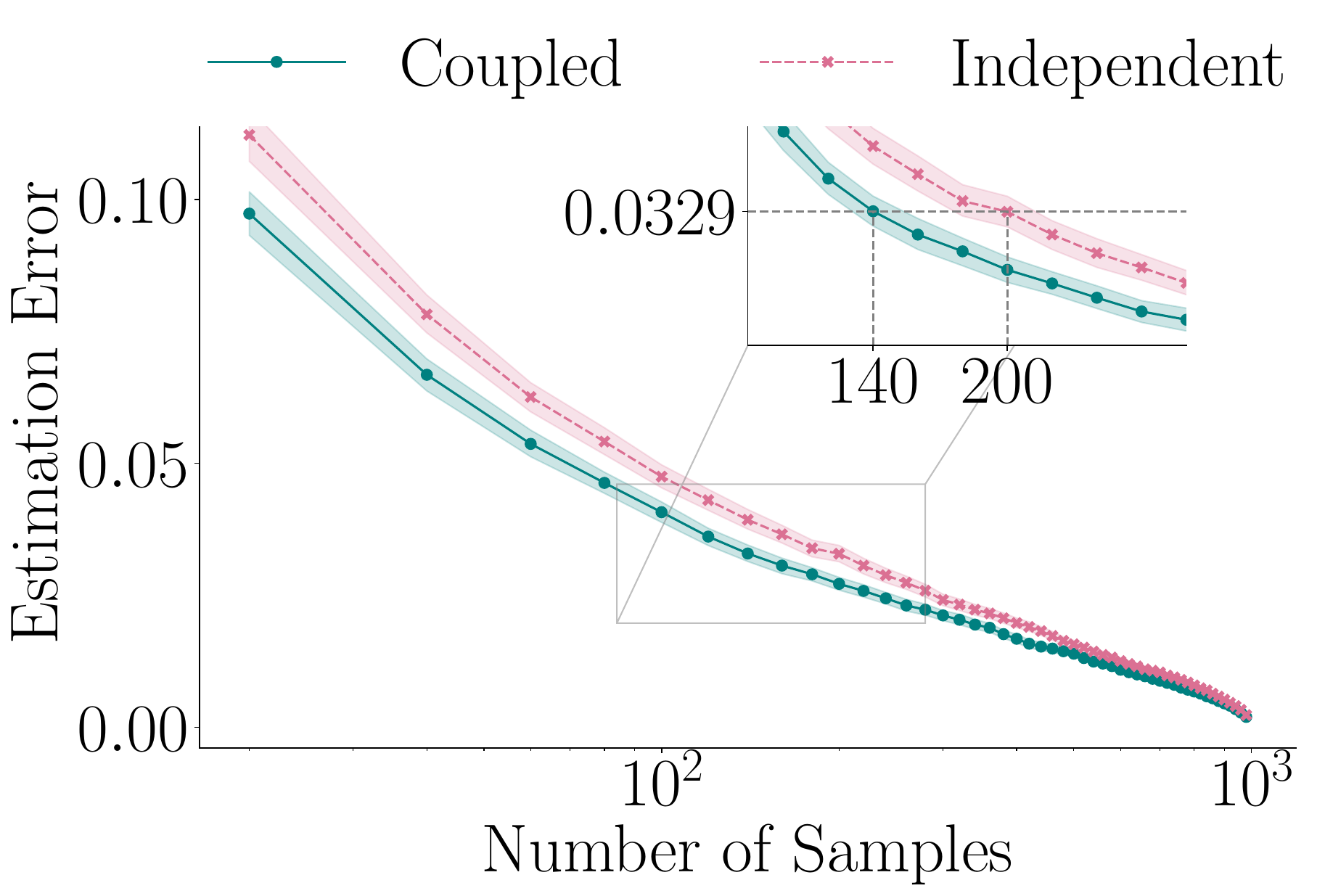} \\ \\

    (a) Score covariance & (b) Variance of the score difference & (c) Estimation error vs. \# samples \\ 
    
\end{tabular}
    \caption{\textbf{Comparison between four pairs of LLMs in the \texttt{Llama} family on multiple-choice questions from the ``college computer science'' knowledge area of the MMLU dataset.}
    Panels in column (a) show the kernel density estimate (KDE) of the covariance between the scores of the two LLMs on each question under coupled generation; the dashed lines correspond to average values. Panels in column (b) show the KDE of the variance of the difference between the scores of the LLMs on each question under coupled and independent generation; the highlighted points correspond to median values. Panels in column (c) show the absolute error in the estimation of the expected difference between the scores of the LLMs against the number of samples; for each point on the x-axis, we perform $1{,}000$ sub-samplings and shaded areas correspond to $95\%$ confidence intervals.}
    \label{fig:mmlu-third-4}
\end{figure}
\begin{figure}[h]
\centering
\begin{tabular}{c c c}
    \multicolumn{3}{c}{\texttt{bnb-8bit} vs. \texttt{bnb-4bit}}\\
    \includegraphics[width=0.23\linewidth]{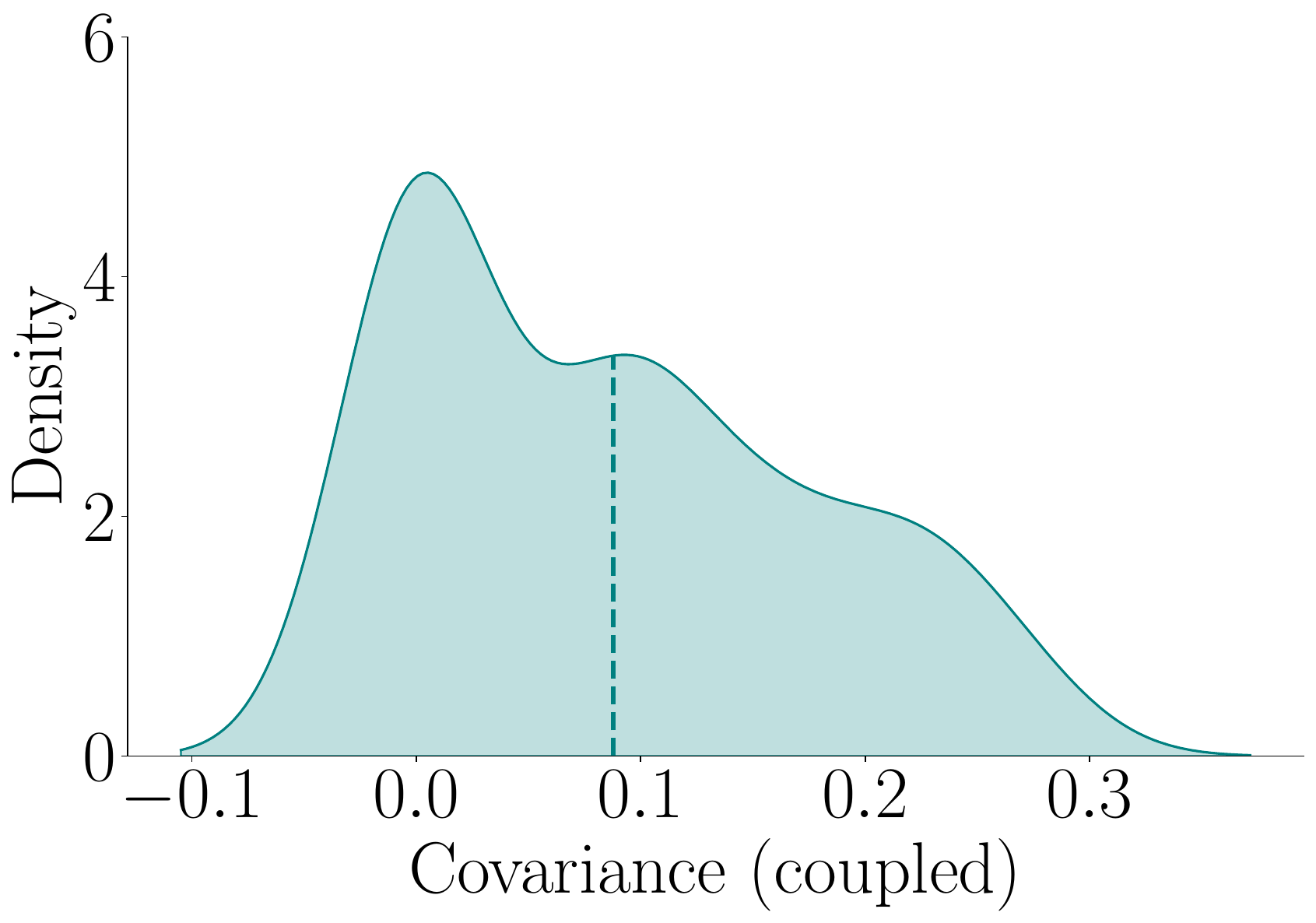} &
    \includegraphics[width=0.23\linewidth]{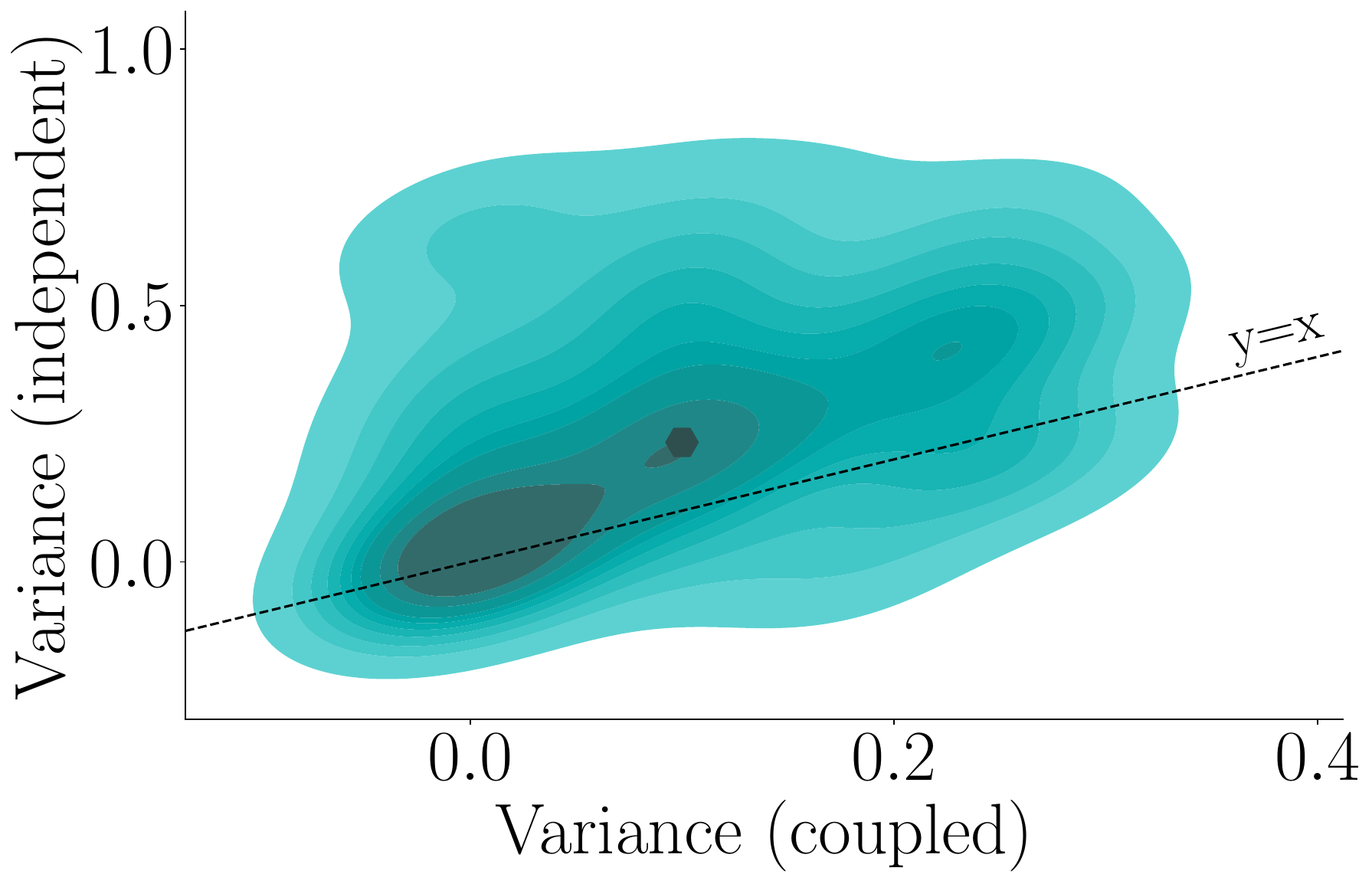} &
    \includegraphics[width=0.23\linewidth]{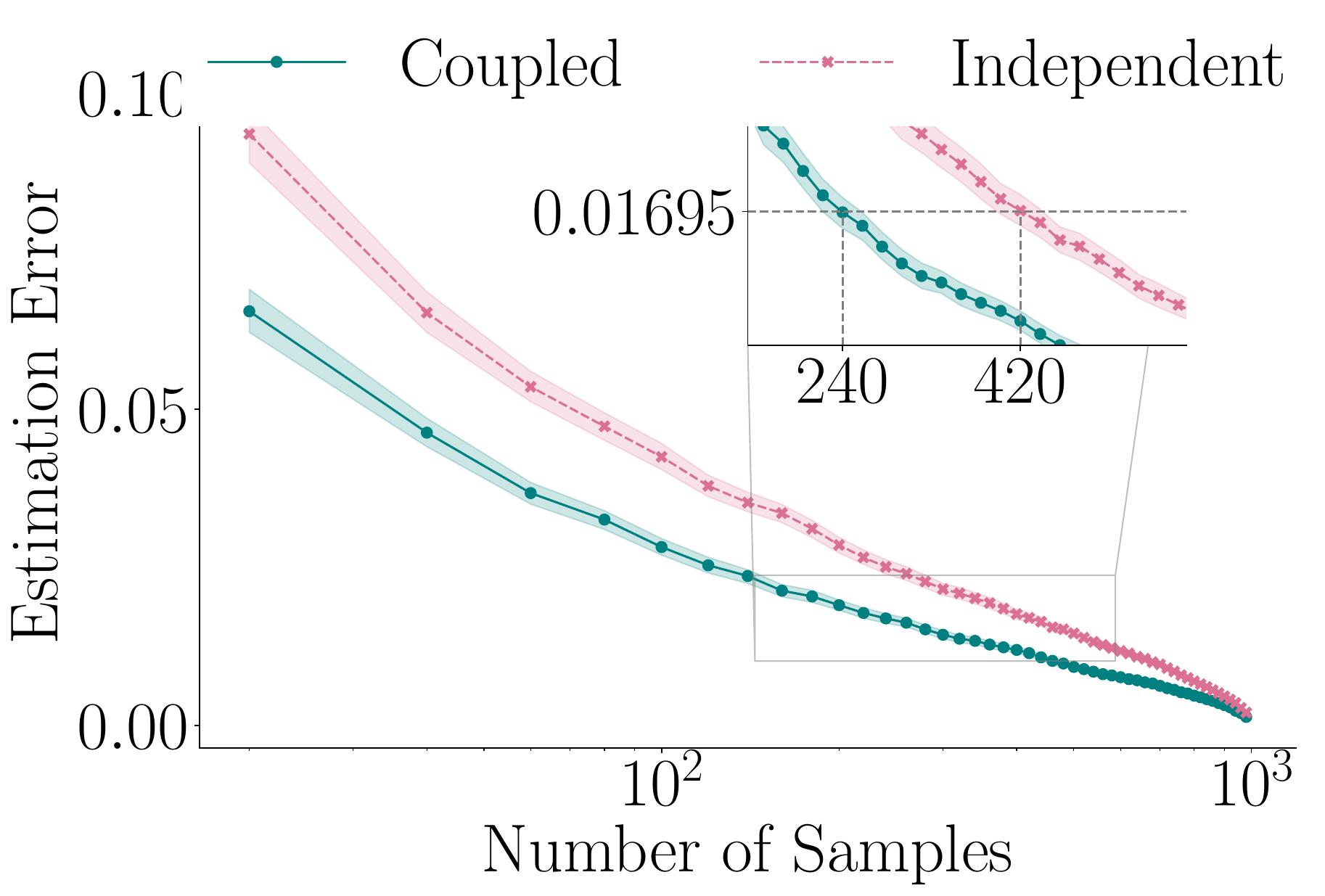} \\ \\
    \multicolumn{3}{c}{\texttt{bnb-8bit} vs. \texttt{AWQ-INT4}}\\
    \includegraphics[width=0.23\linewidth]{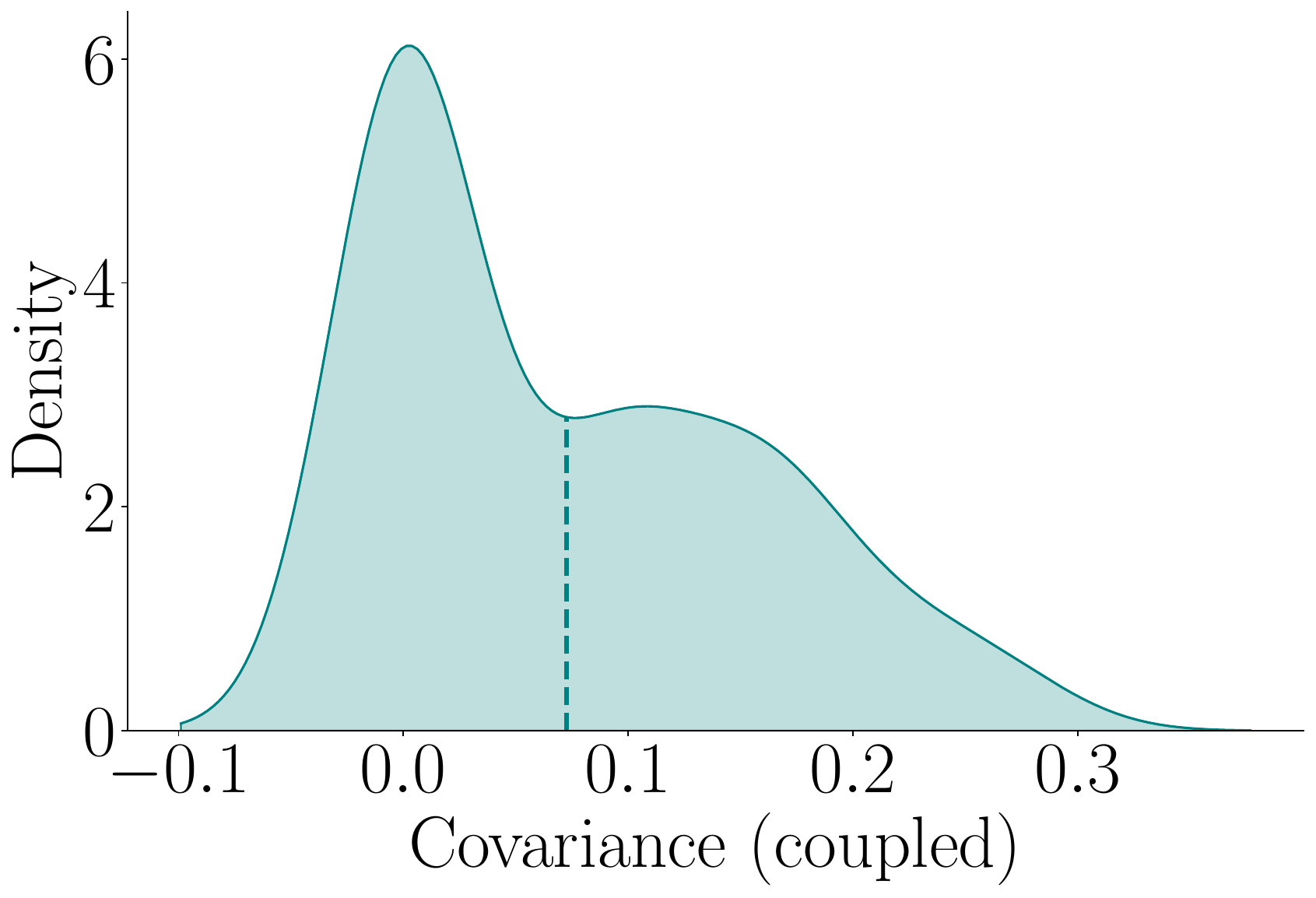} &
    \includegraphics[width=0.23\linewidth]{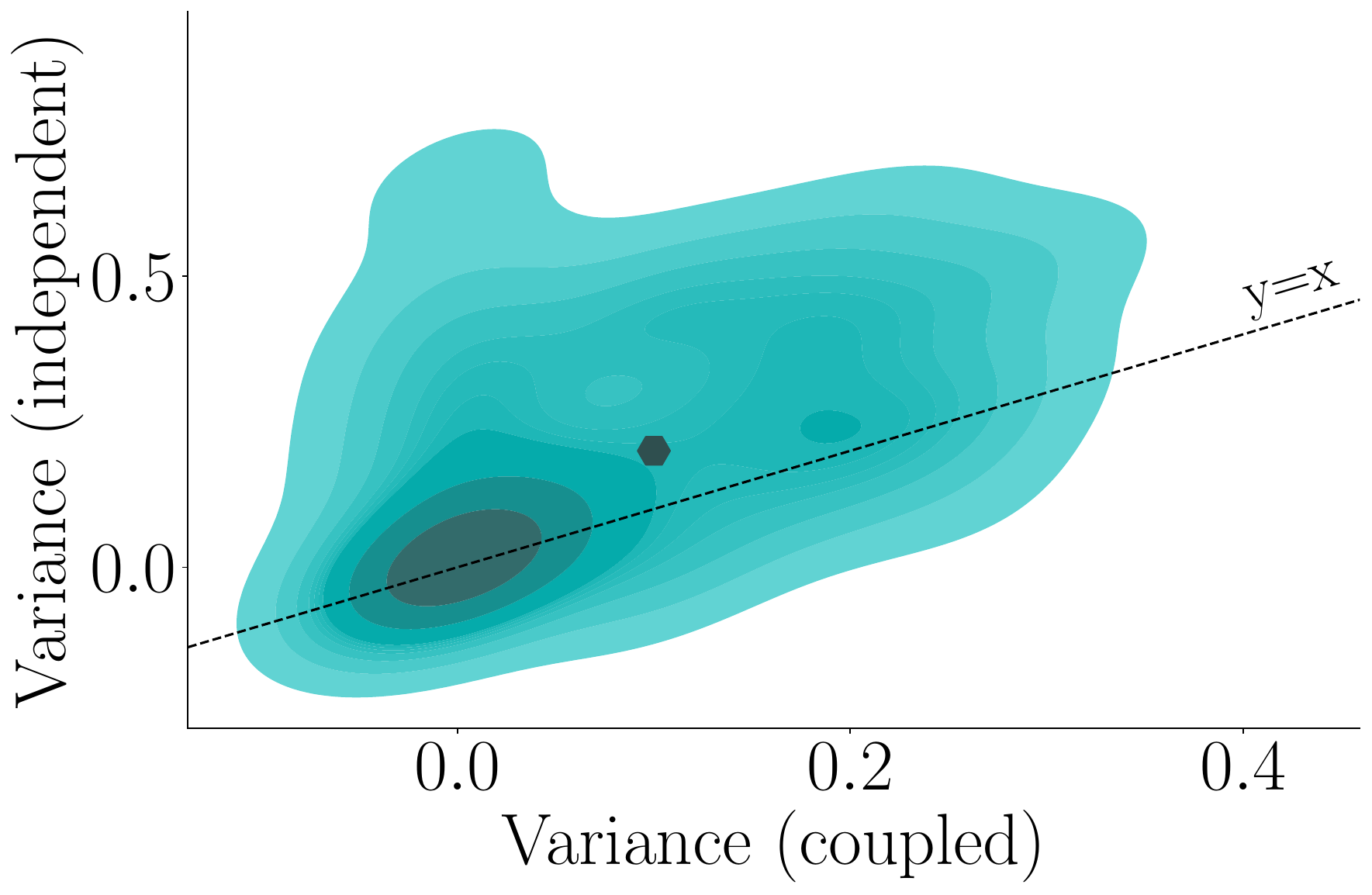} &
    \includegraphics[width=0.23\linewidth]{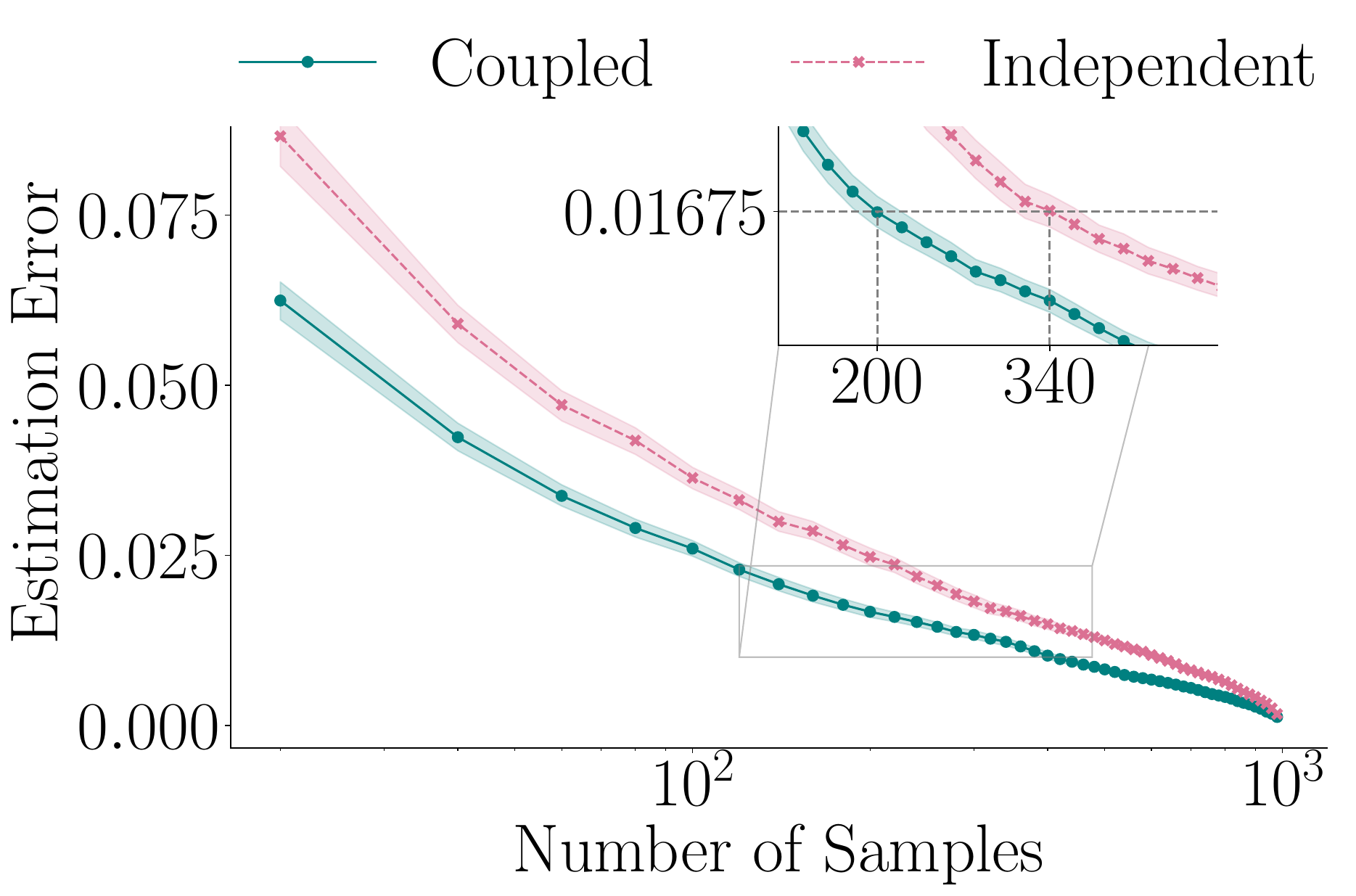} \\ \\
    \multicolumn{3}{c}{\texttt{bnb-4bit} vs. \texttt{AWQ-INT4}}\\
    \includegraphics[width=0.23\linewidth]{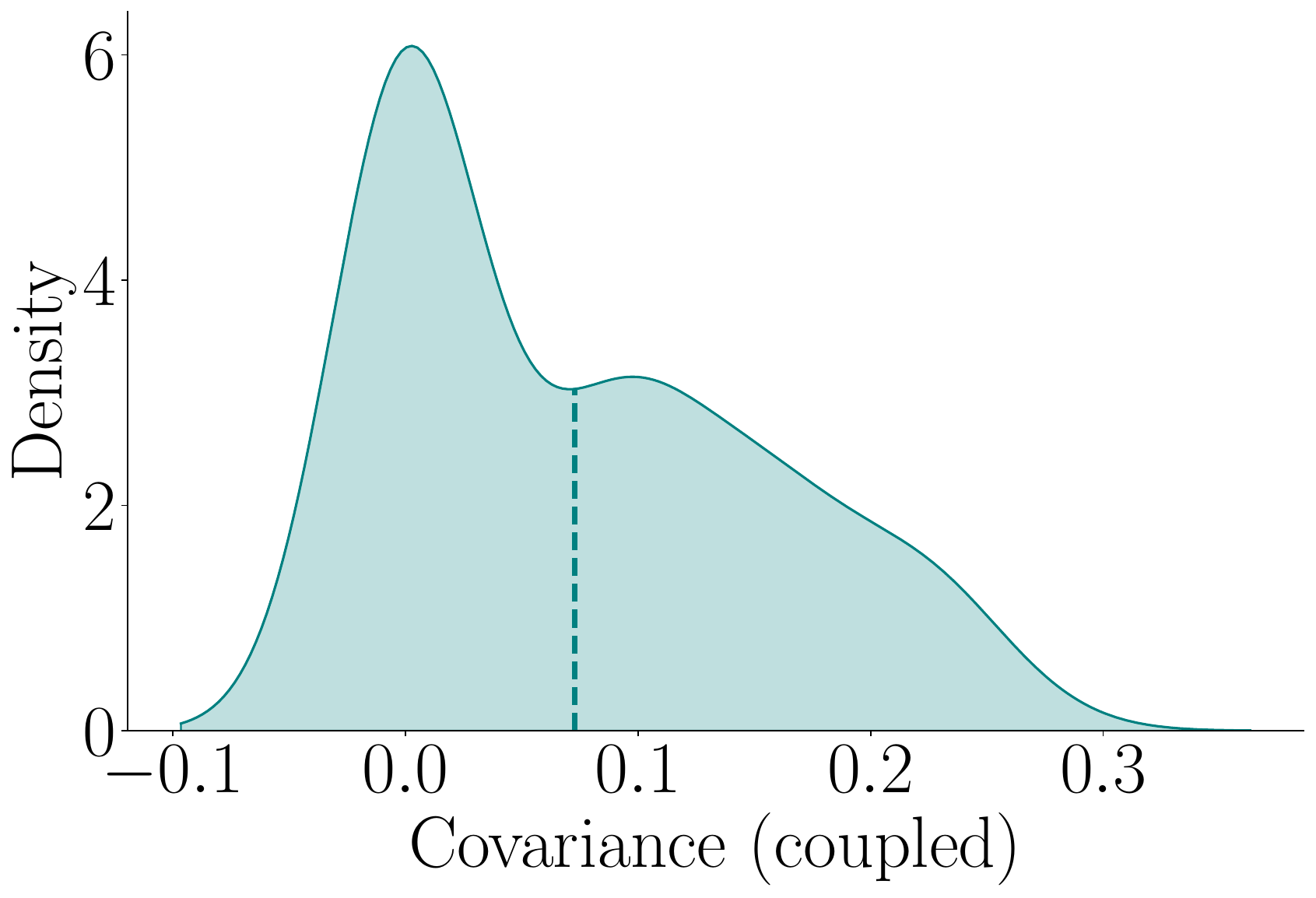} &
    \includegraphics[width=0.23\linewidth]{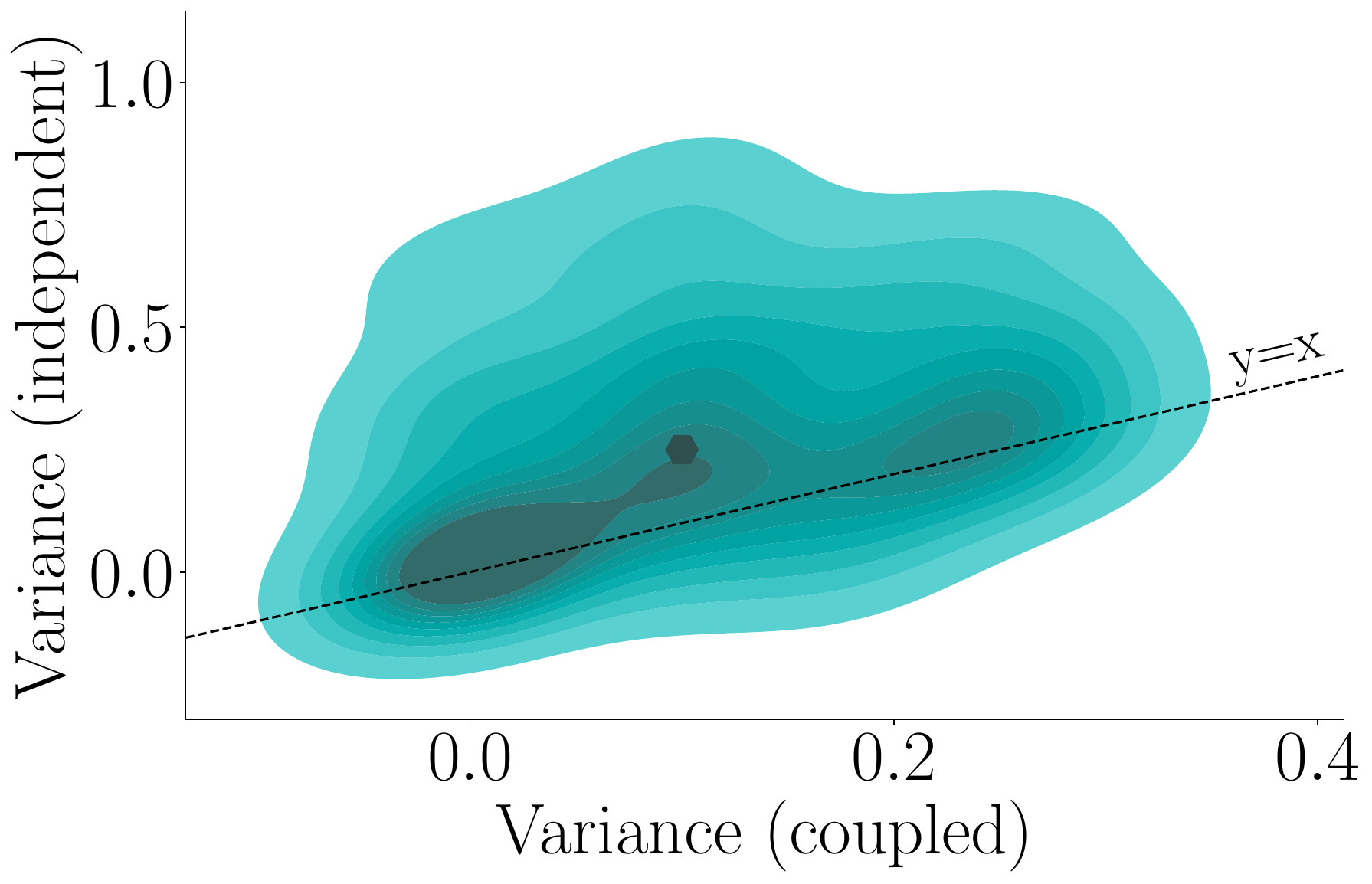} &
    \includegraphics[width=0.23\linewidth]{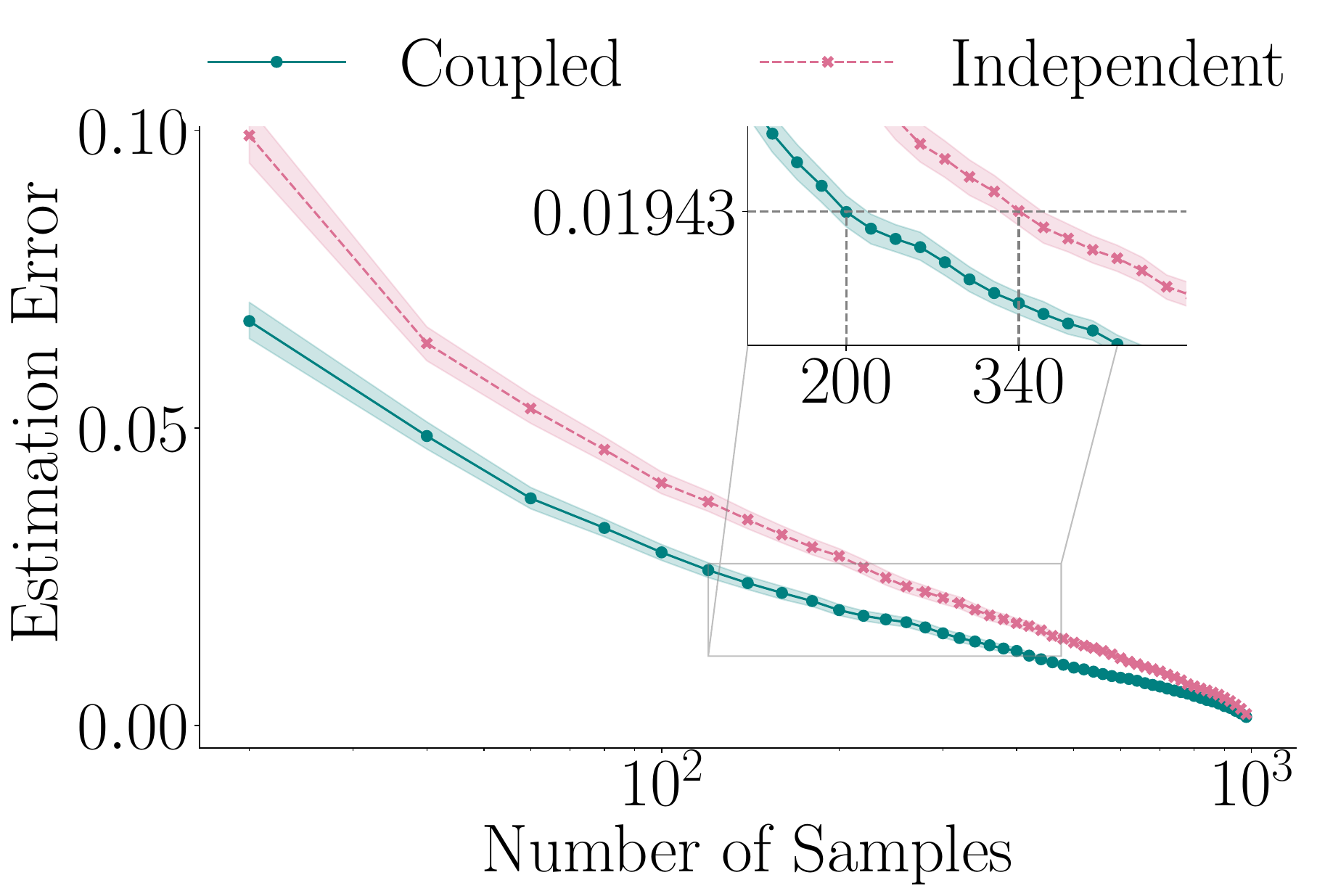} \\ \\

    (a) Score covariance & (b) Variance of the score difference & (c) Estimation error vs. \# samples \\ 
    
\end{tabular}
    \caption{\textbf{Comparison between three pairs of LLMs in the \texttt{Llama} family on multiple-choice questions from the ``college computer science'' knowledge area of the MMLU dataset.}
    Panels in column (a) show the kernel density estimate (KDE) of the covariance between the scores of the two LLMs on each question under coupled generation; the dashed lines correspond to average values. Panels in column (b) show the KDE of the variance of the difference between the scores of the LLMs on each question under coupled and independent generation; the highlighted points correspond to median values. Panels in column (c) show the absolute error in the estimation of the expected difference between the scores of the LLMs against the number of samples; for each point on the x-axis, we perform $1{,}000$ sub-samplings and shaded areas correspond to $95\%$ confidence intervals.}
    \label{fig:mmlu-last-3}
\end{figure}

\begin{figure}[h]
\centering
\begin{tabular}{c c c}
    \multicolumn{3}{c}{\textbf{College chemistry}} \\
    \includegraphics[width=0.23\linewidth]{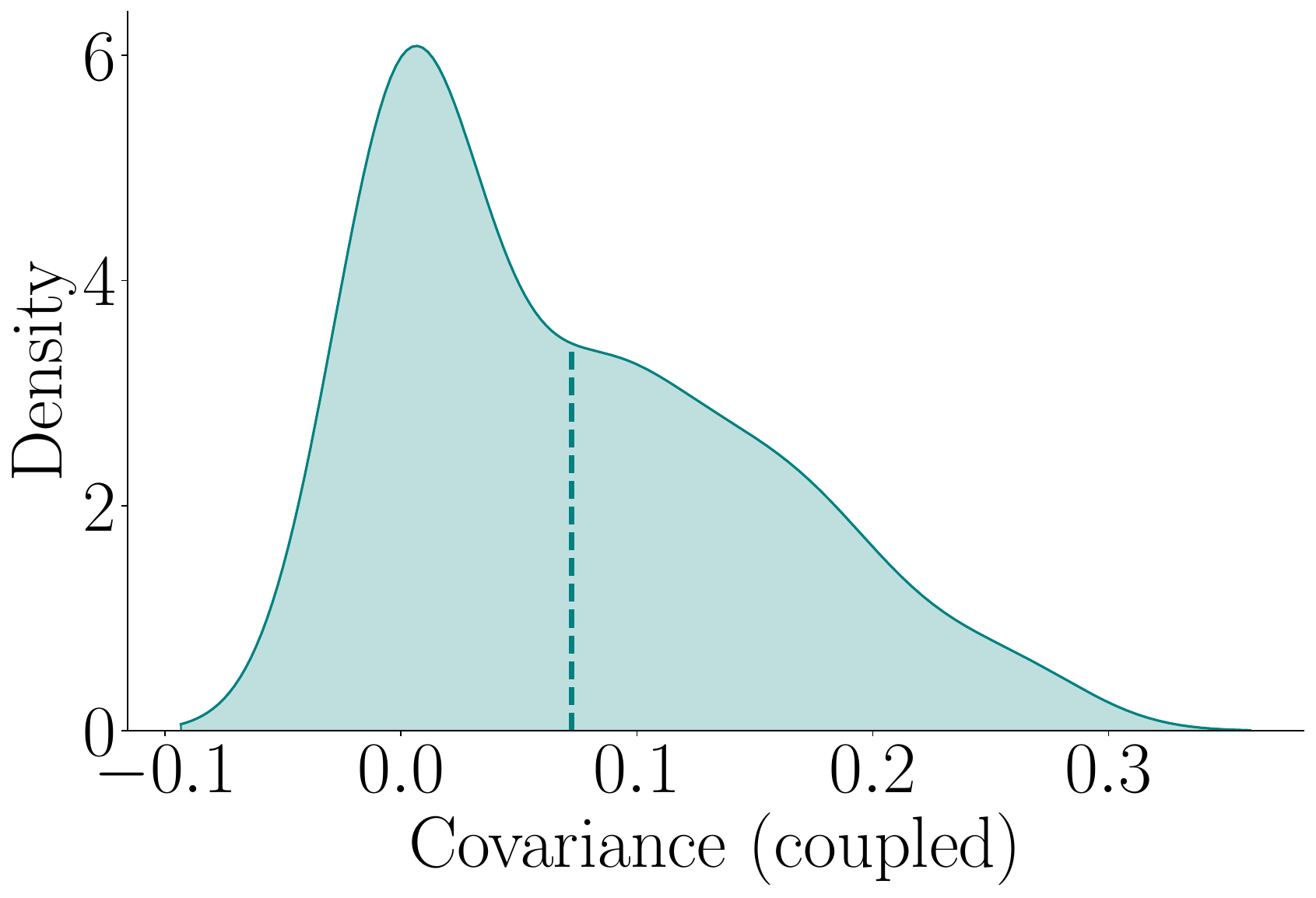} &
    \includegraphics[width=0.23\linewidth]{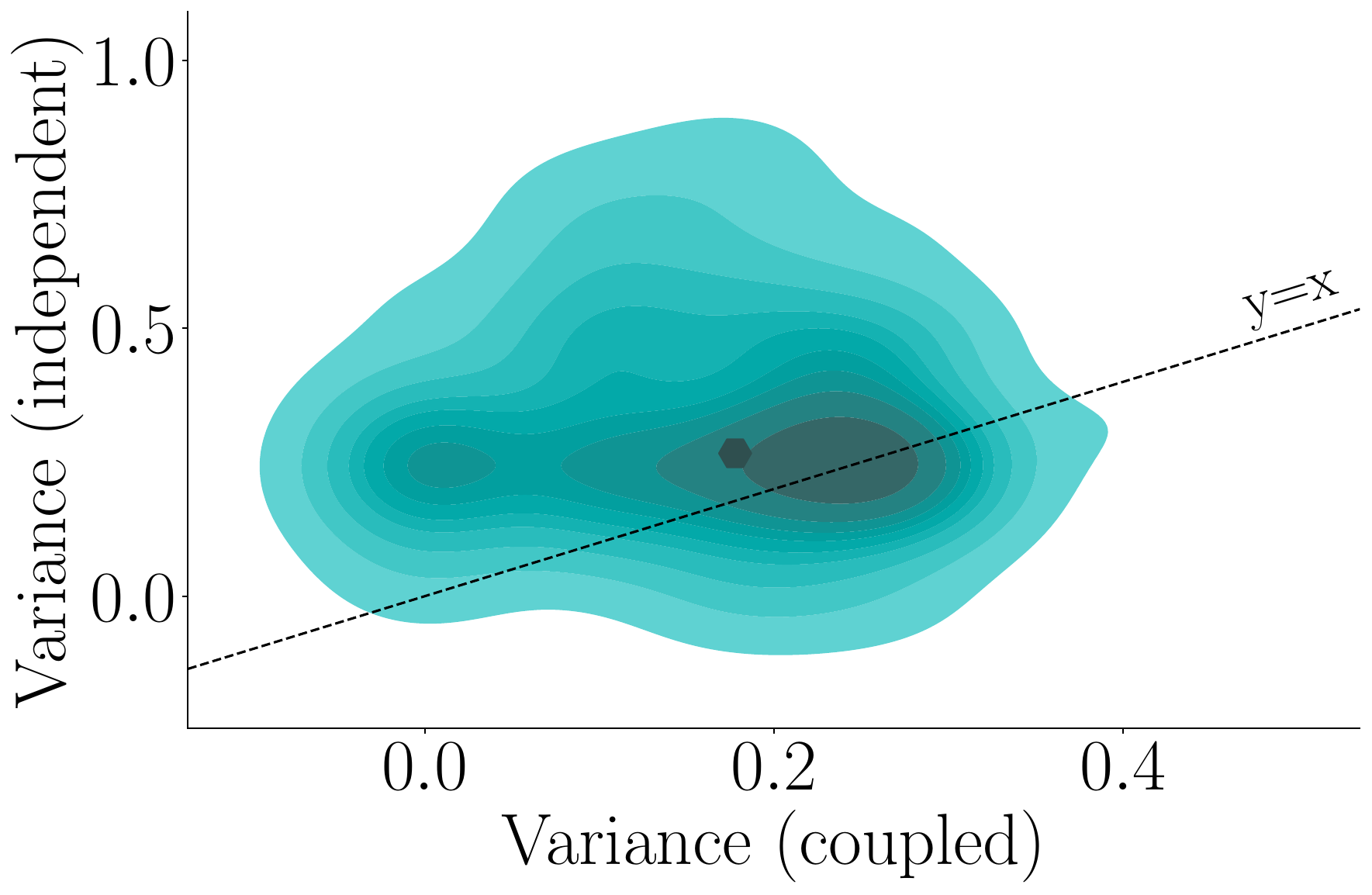} &
    \includegraphics[width=0.23\linewidth]{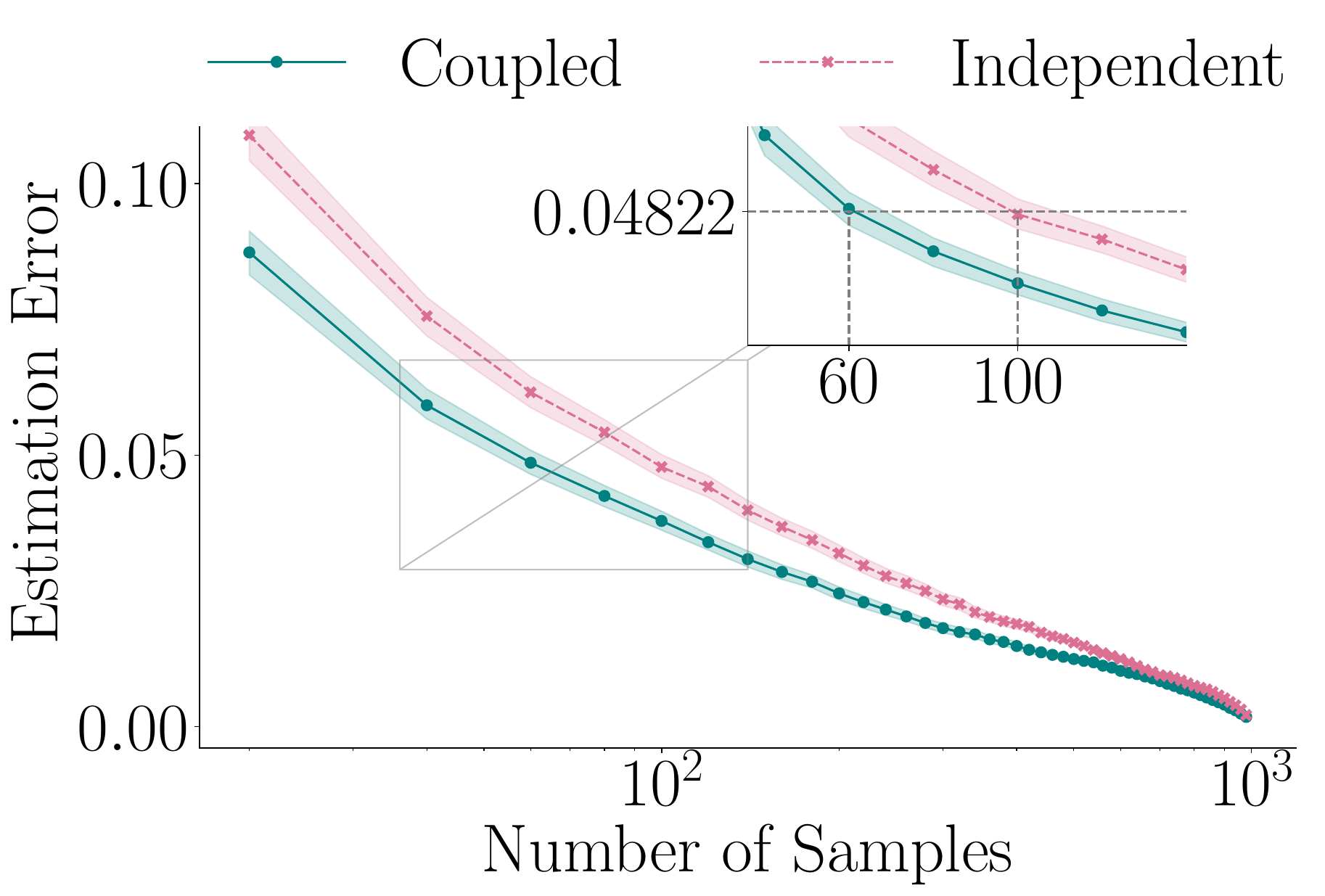} \\
    \multicolumn{3}{c}{\textbf{Professional accounting}} \\
    \includegraphics[width=0.23\linewidth]{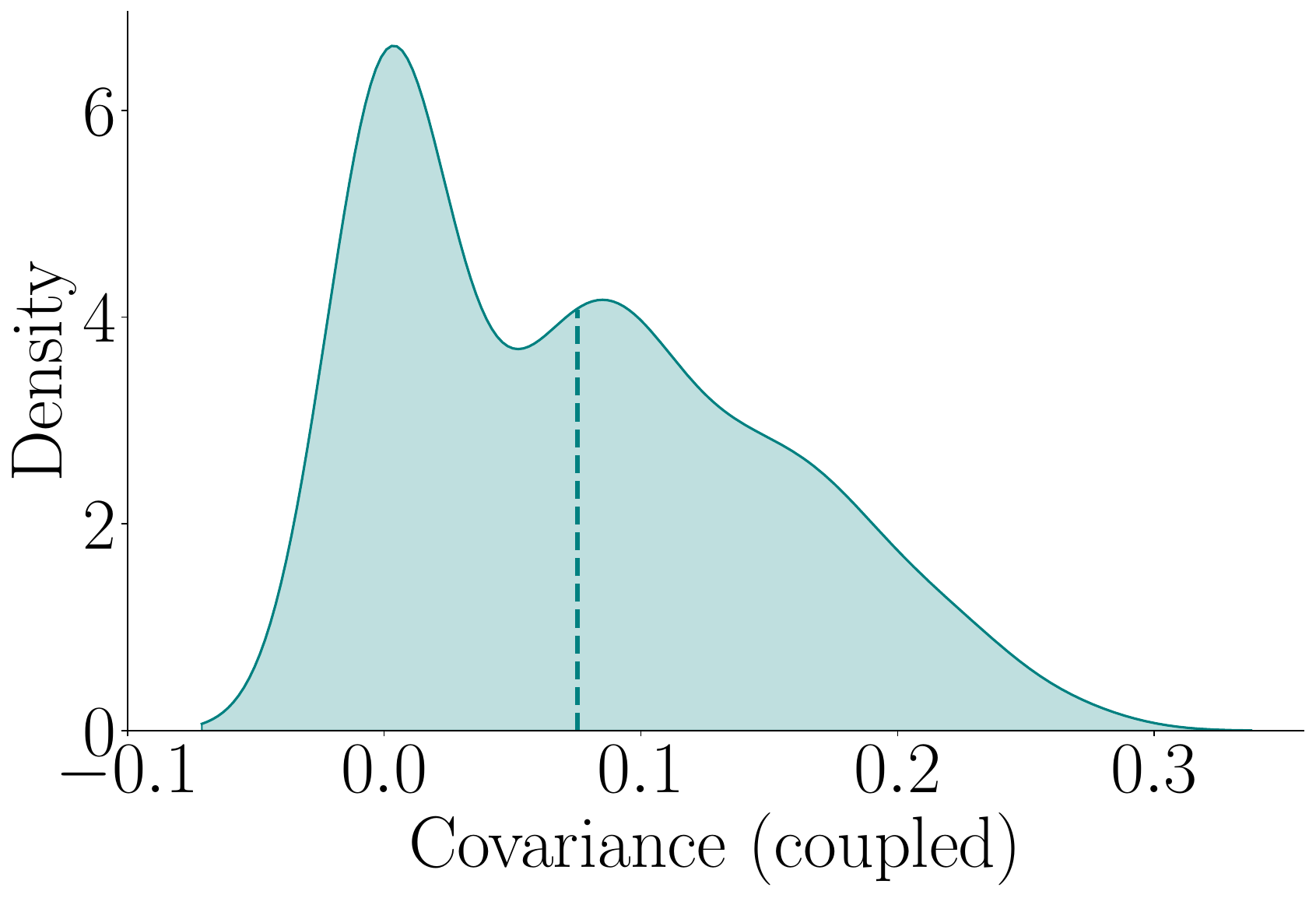} &
    \includegraphics[width=0.23\linewidth]{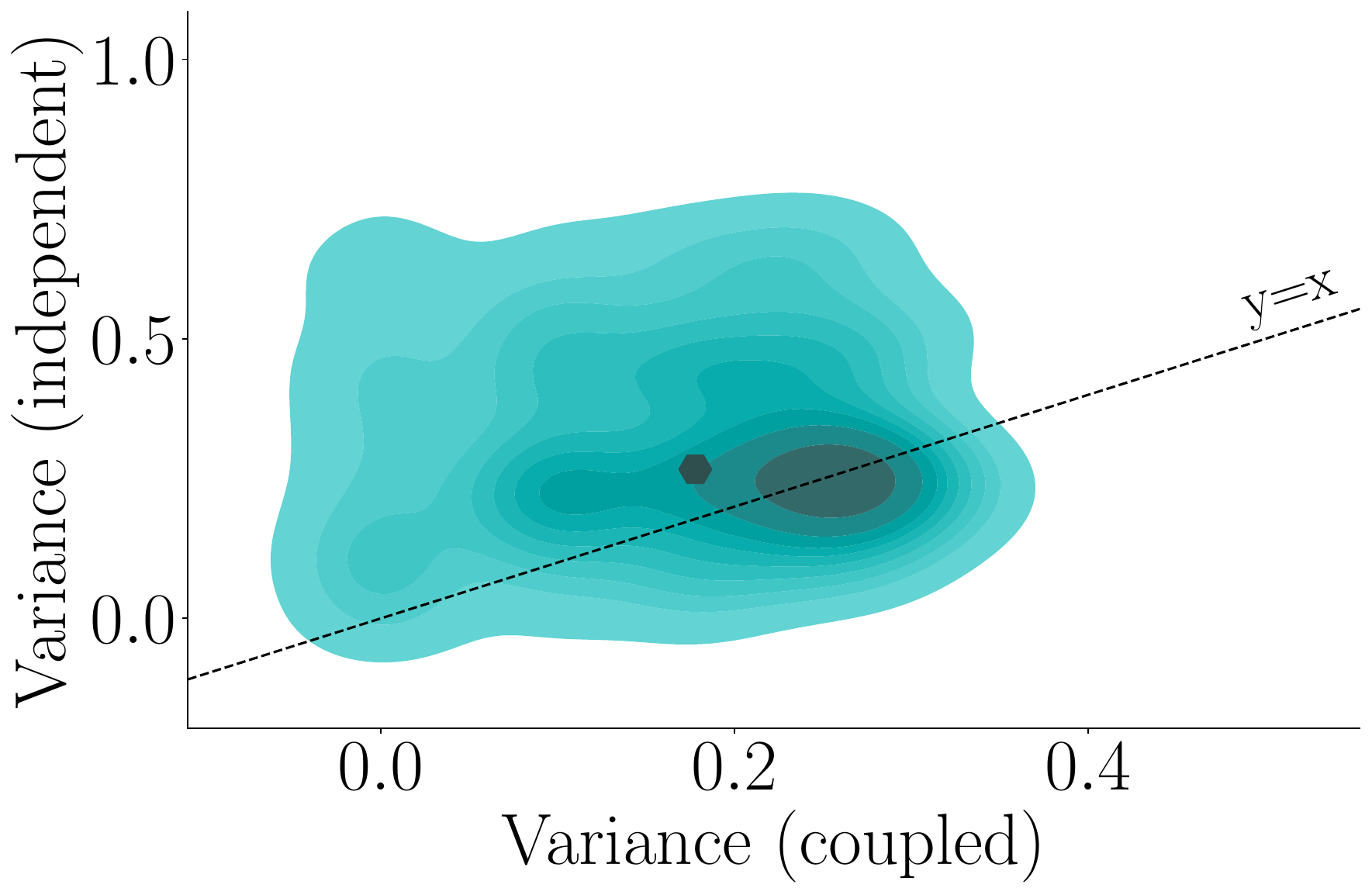} &
    \includegraphics[width=0.23\linewidth]{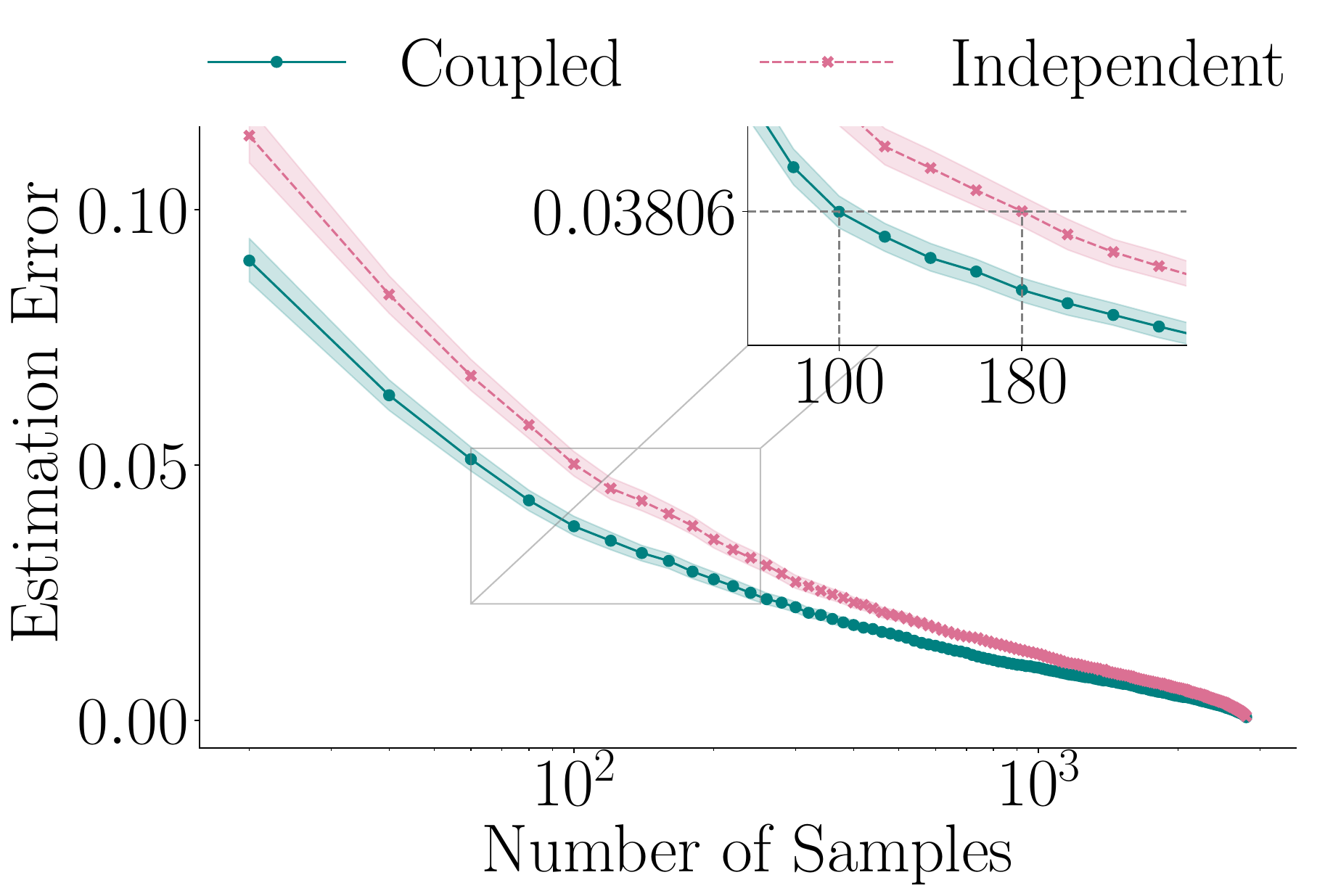} \\
    \multicolumn{3}{c}{\textbf{Professional law}}\\
    \includegraphics[width=0.23\linewidth]{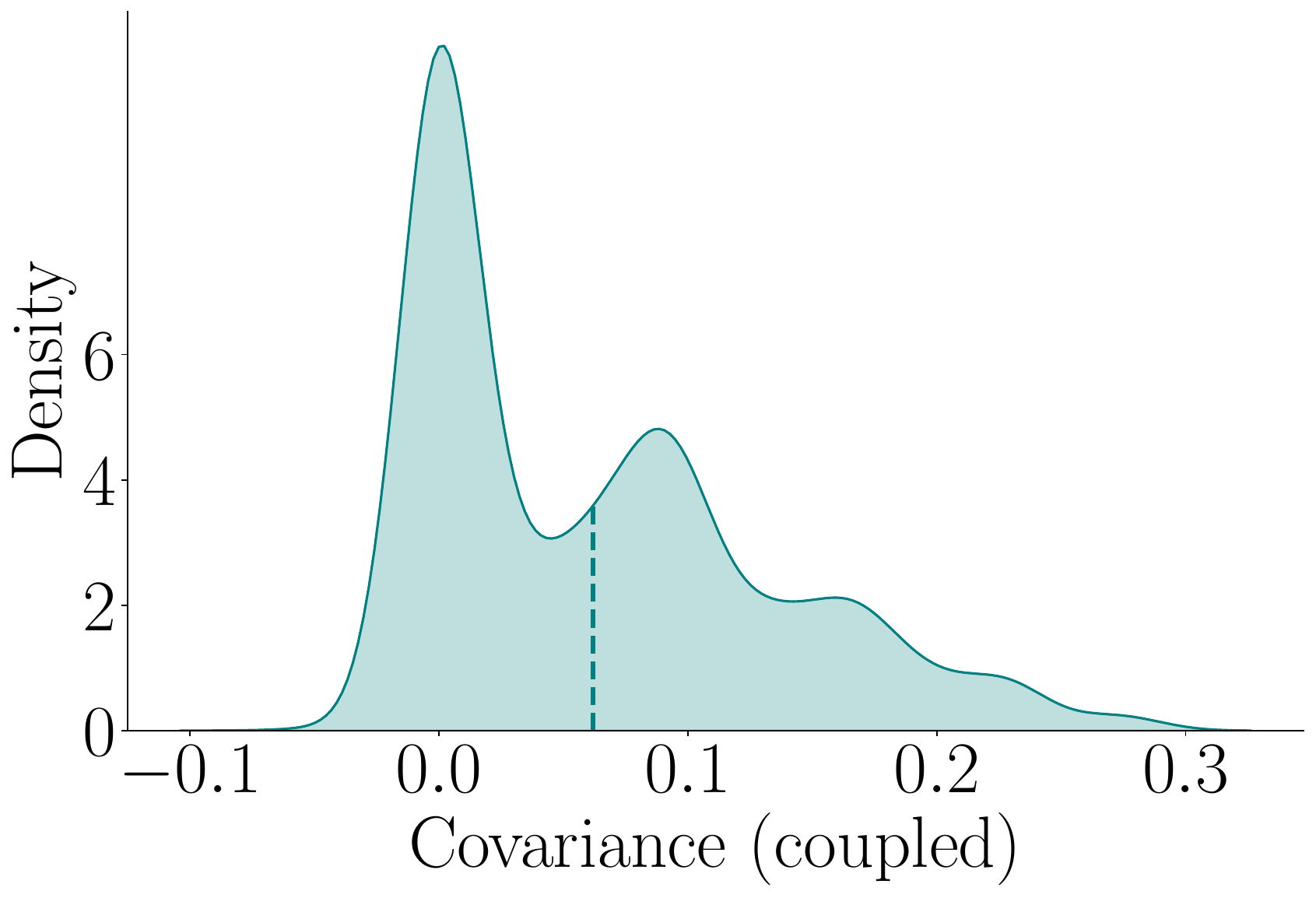} &
    \includegraphics[width=0.23\linewidth]{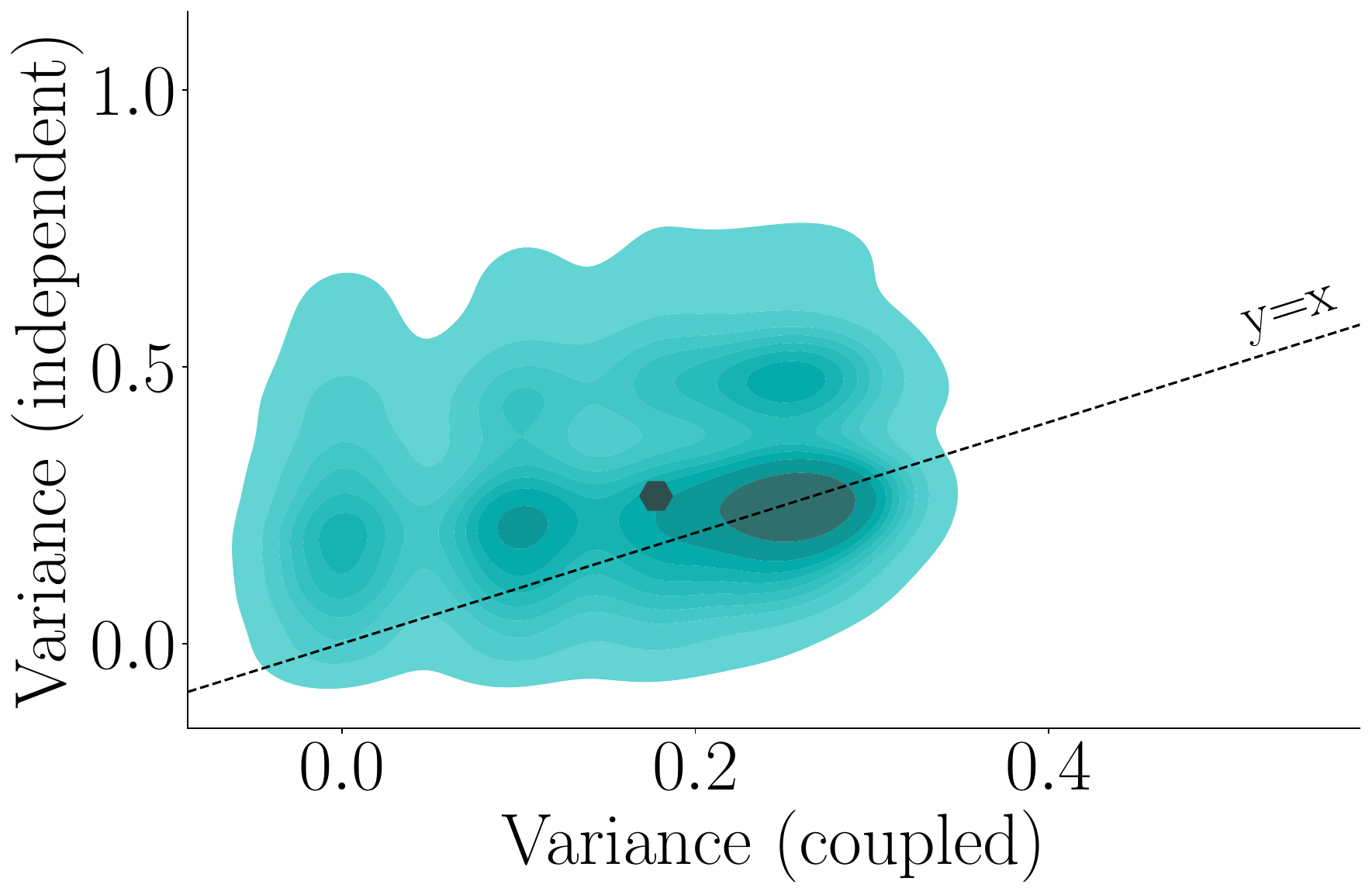} &
    \includegraphics[width=0.23\linewidth]{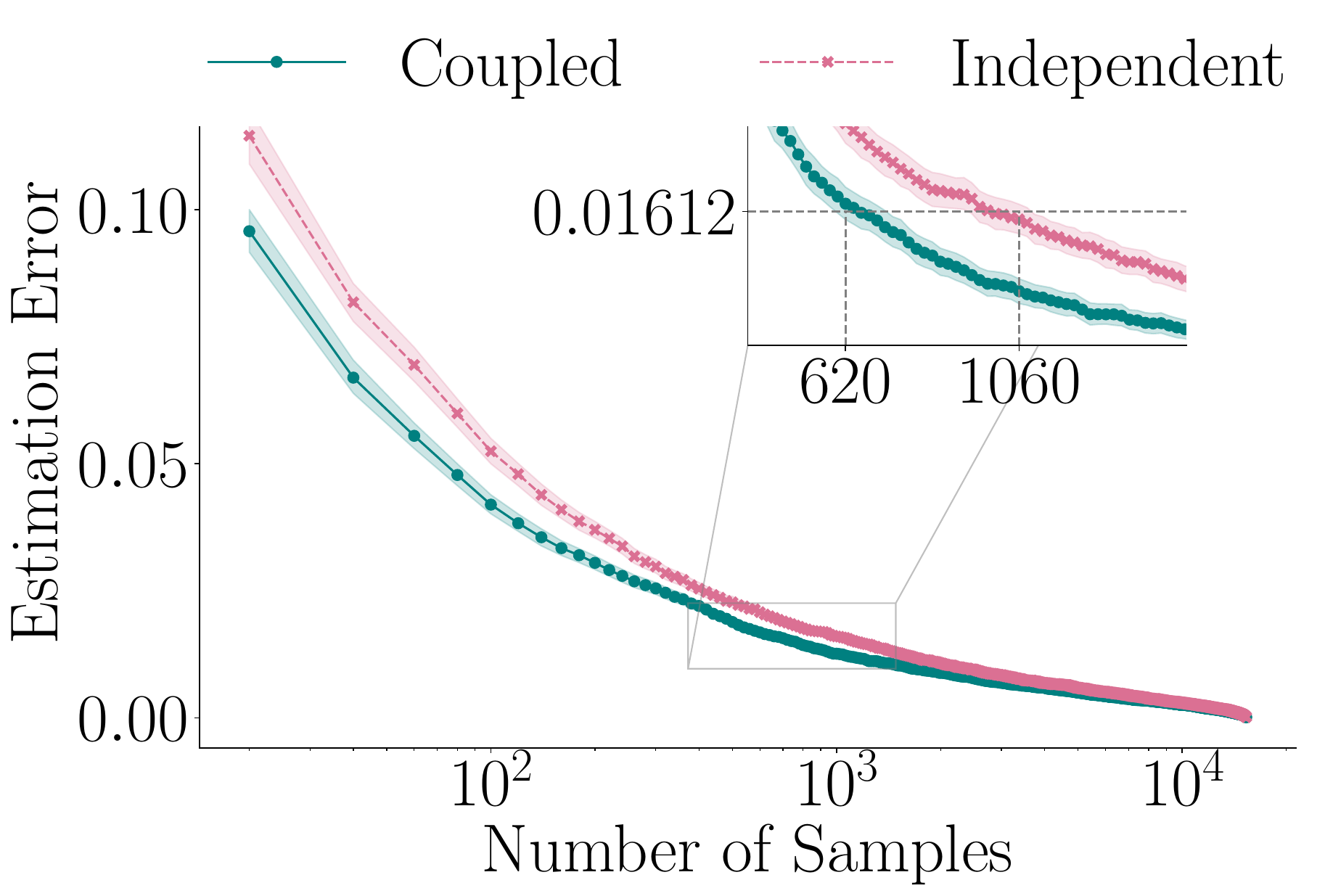} \\ \\
    \multicolumn{3}{c}{\textbf{Professional medicine}} \\   
    \includegraphics[width=0.23\linewidth]{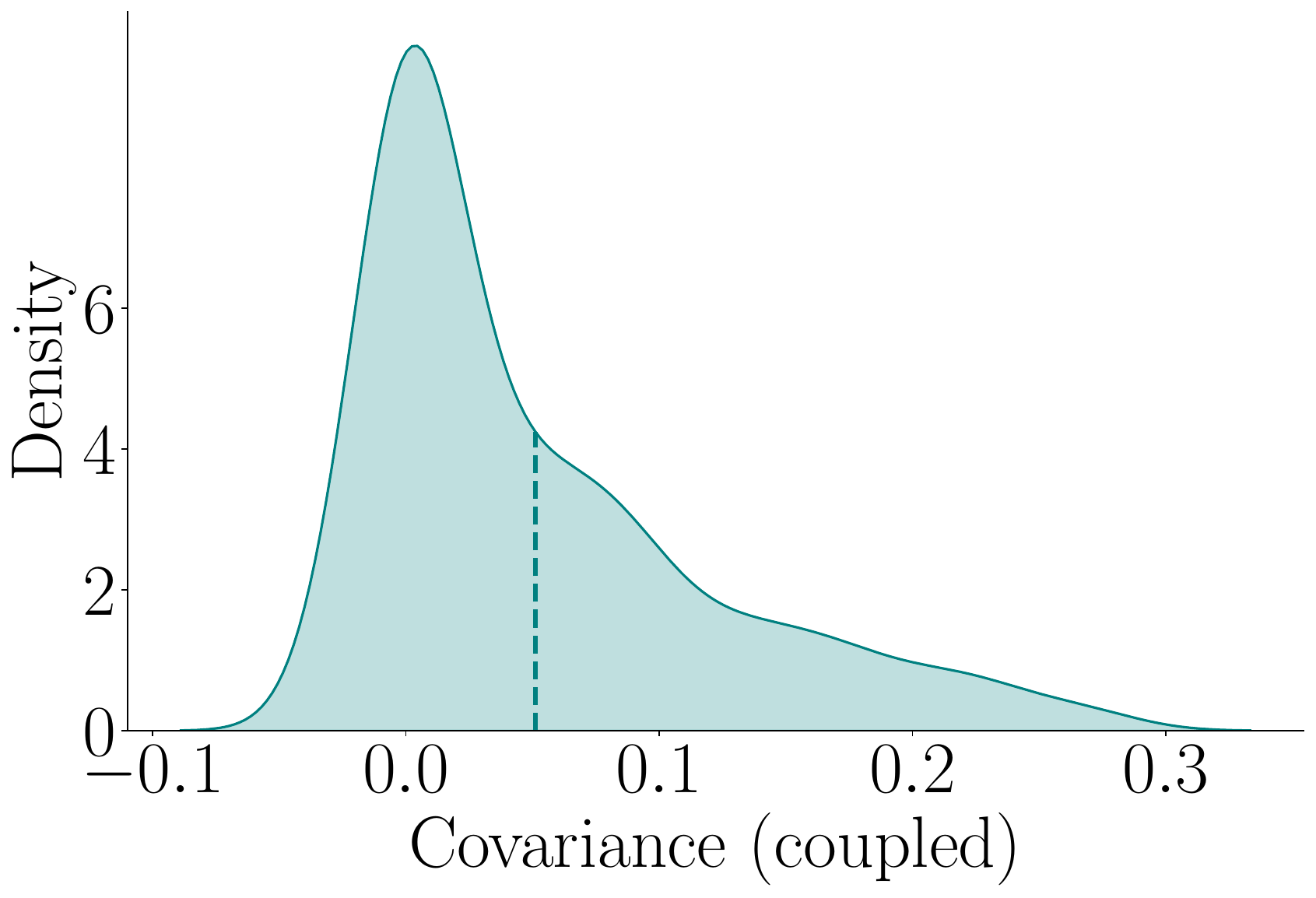} &
    \includegraphics[width=0.23\linewidth]{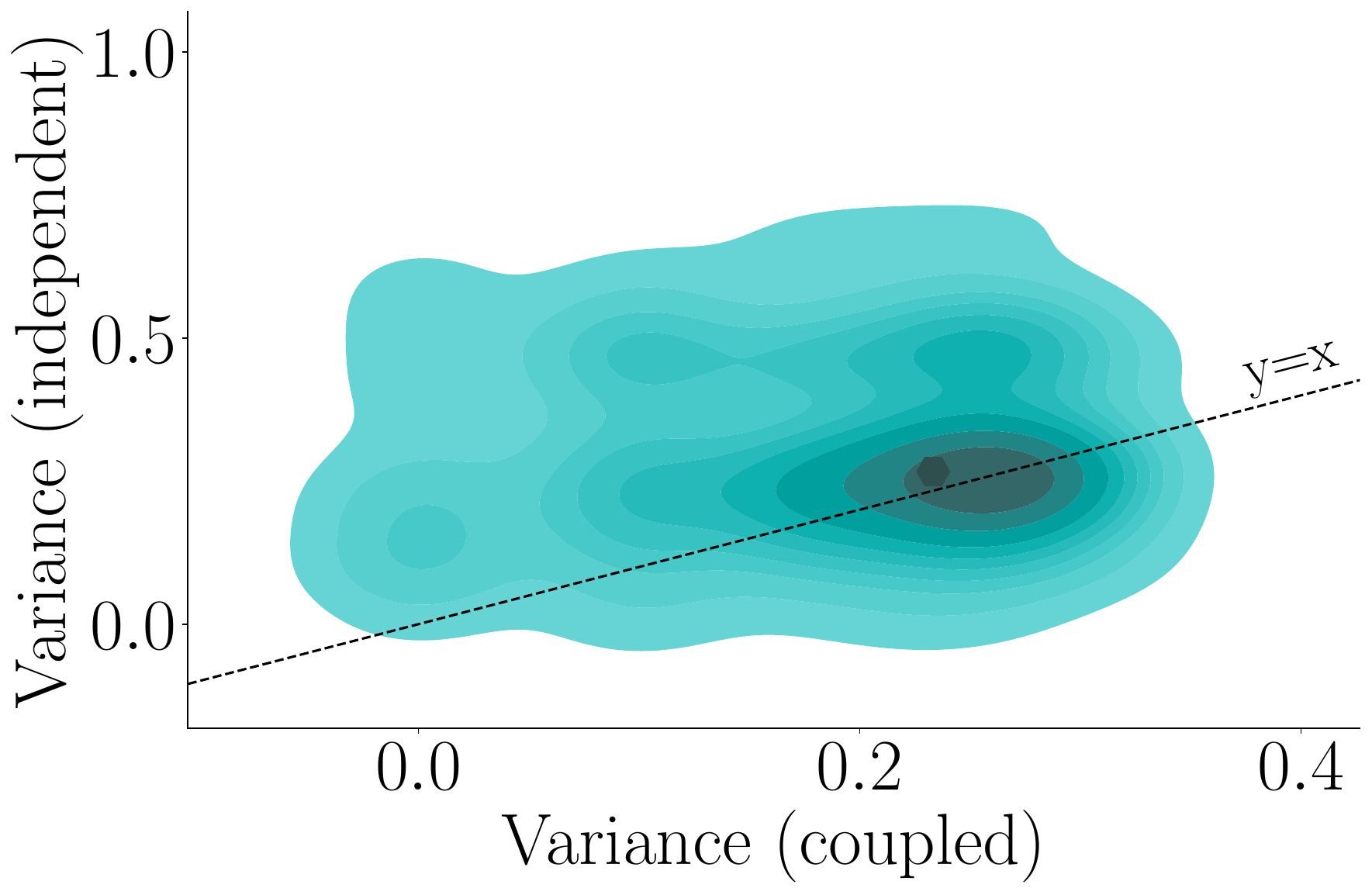} &
    \includegraphics[width=0.23\linewidth]{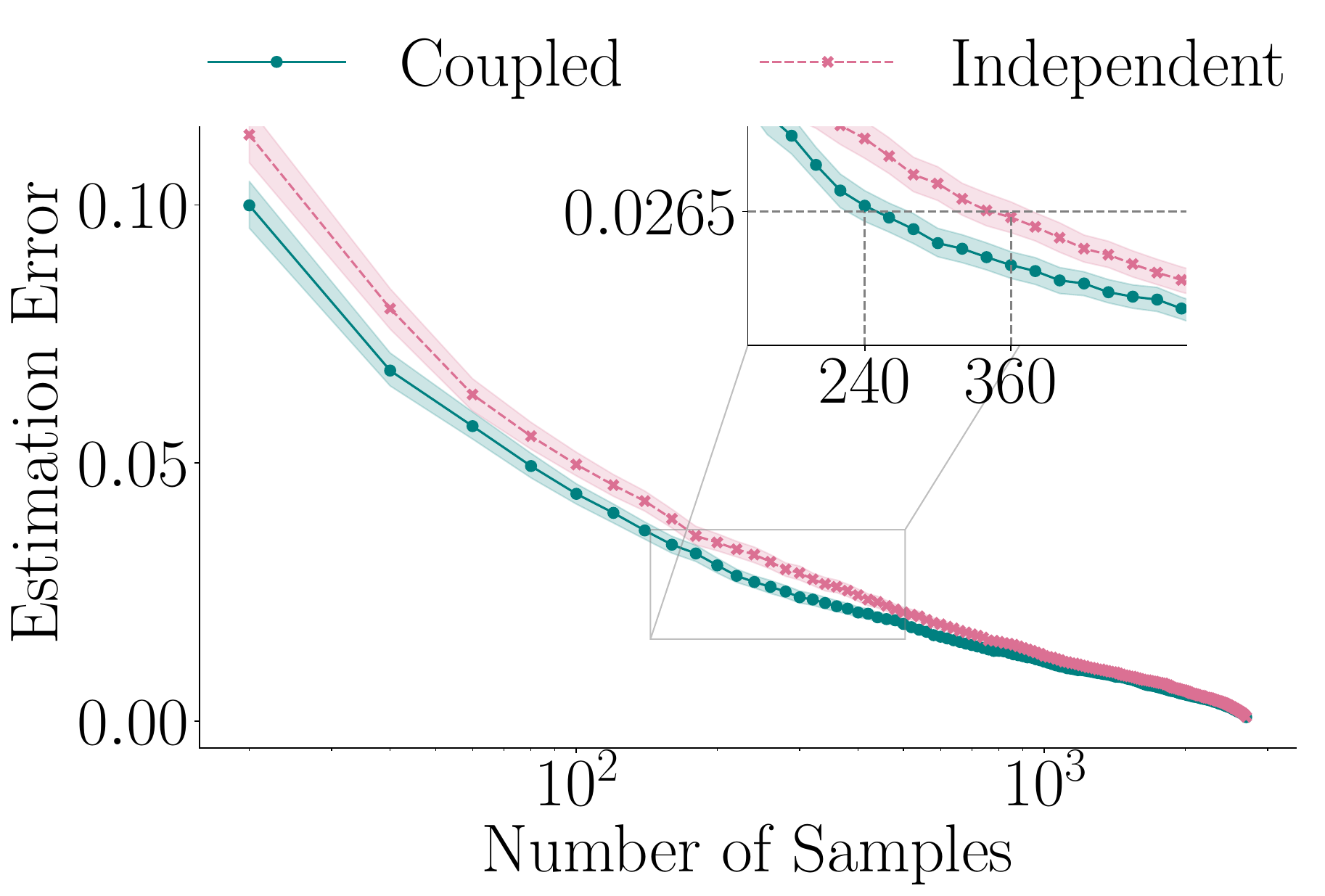} \\
    (a) Score covariance & (b) Variance of the score difference & (c) Estimation error vs. \# samples \\ 
\end{tabular}
    \caption{\textbf{Comparison between \texttt{1B} and \texttt{3B} from the \texttt{Llama} family on multiple-choice questions from four knowledge areas of the MMLU dataset.}
    Panels in column (a) show the kernel density estimate (KDE) of the covariance between the scores of the two LLMs on each question under coupled generation; the dashed lines correspond to average values. Panels in column (b) show the KDE of the variance of the difference between the scores of the LLMs on each question under coupled and independent generation; the highlighted points correspond to median values. Panels in column (c) show the absolute error in the estimation of the expected difference between the scores of the LLMs against the number of samples; for each point on the x-axis, we perform $1{,}000$ sub-samplings and shaded areas correspond to $95\%$ confidence intervals. We observe qualitatively similar results for other knowledge areas.}
    \label{fig:mmlu-1B-vs-3B-areas}
\end{figure}

\clearpage
\newpage
\subsection{GSM8K and HumanEval Datasets}
\label{app:llama-gsm8k-humaneval}

\xhdr{Experimental Setup for GSM8K} 
 We provide each math problem in the GSM8K validation set as an input prompt to the LLMs. Further, we instruct the LLMs through a system prompt to reply with a single numerical value---see Table~\ref{app:sys_prompt_gsm8k} for the exact prompt. 
To evaluate the outputs provided by each LLM, we use a binary score $R \in \{0,1\}$, which indicates whether the last numerical value in the LLM output is the correct ($R=1$) or incorrect $(R=0)$ solution to the math problem.
%
To obtain reliable conclusions,
we experiment with each math problem $10$ times, each time using a (different) random seed to generate the noise variables used by the sampler. 
%
In what follows, we compare all pairs of LLMs described in Section~\ref{sec:experiments}.

\xhdr{Experimental Setup for HumanEval}  
 We prepend the input prefix in Table~\ref{app:usr-prefix-human-eval} to each programming problem in HumanEval dataset and provide it as an input to the LLMs.
 Further, we append each programming problem to the response prefix in Table~\ref{app:usr-prefix-human-eval}
 and provide it as an output prefix, so that the LLMs generate responses that continue the given output prefix (instead of generating a new response).
To evaluate the outputs provided by each LLM, we use 
a binary score $R \in \{0,1\}$, which indicates whether the program given in each LLM output is the correct ($R=1$) or incorrect $(R=0)$ solution to the programming problem.
To parse and evaluate the program in each LLM output, we use the evaluation harness by~\citet{chen2021evaluating}.
To obtain reliable conclusions,
we experiment with each programming problem $10$ times, each time using a (different) random seed to generate the noise variables used by the sampler. 
%
In what follows, we compare all pairs of LLMs described in Section~\ref{sec:experiments}.

\begin{figure}[ht]
\centering
\begin{tabular}{c c c}
    \multicolumn{3}{c}{\texttt{1B} vs. \texttt{3B}}\\
    \includegraphics[width=0.23\linewidth]{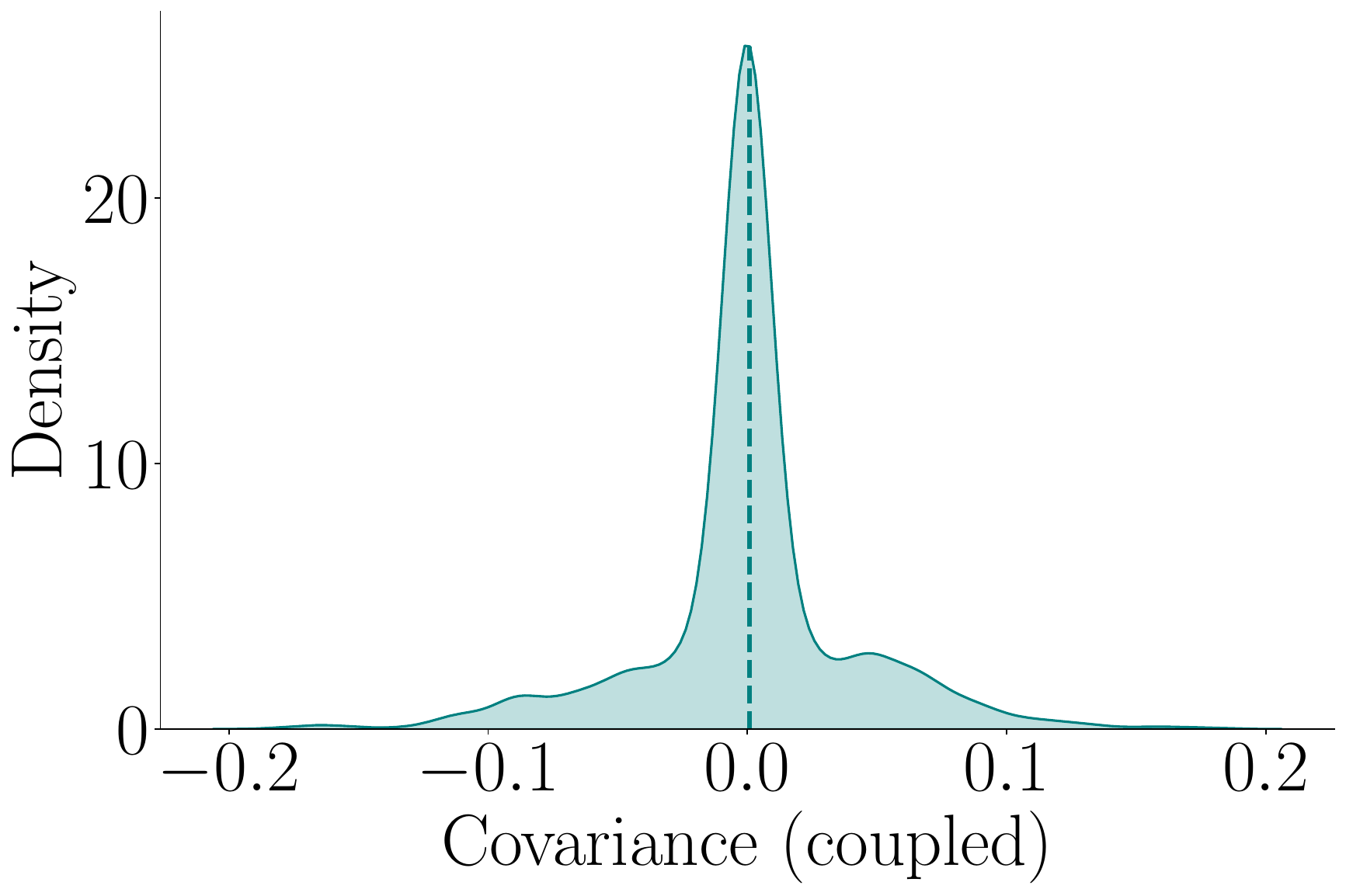} &
    \includegraphics[width=0.23\linewidth]{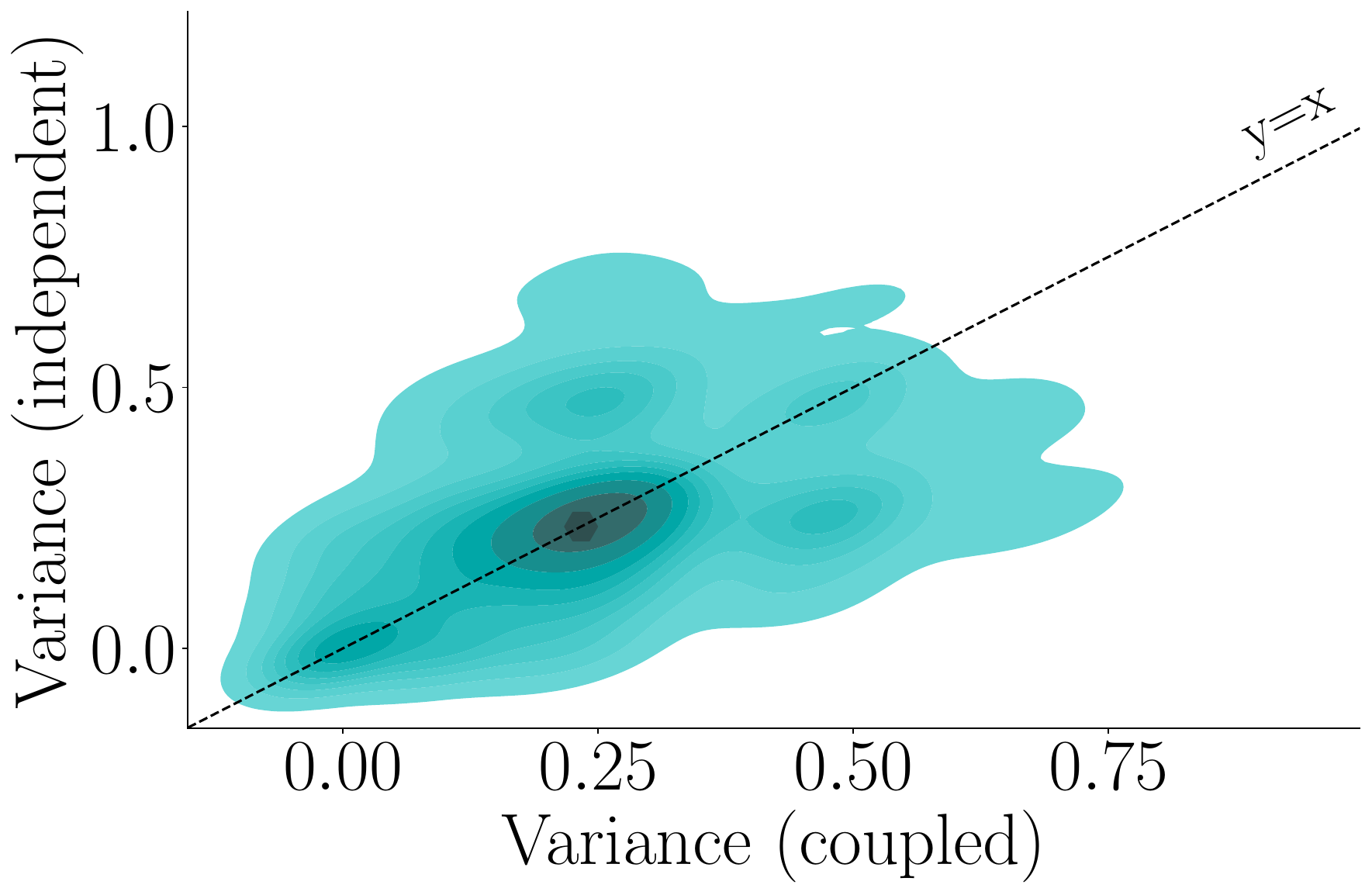} &
    \includegraphics[width=0.23\linewidth]{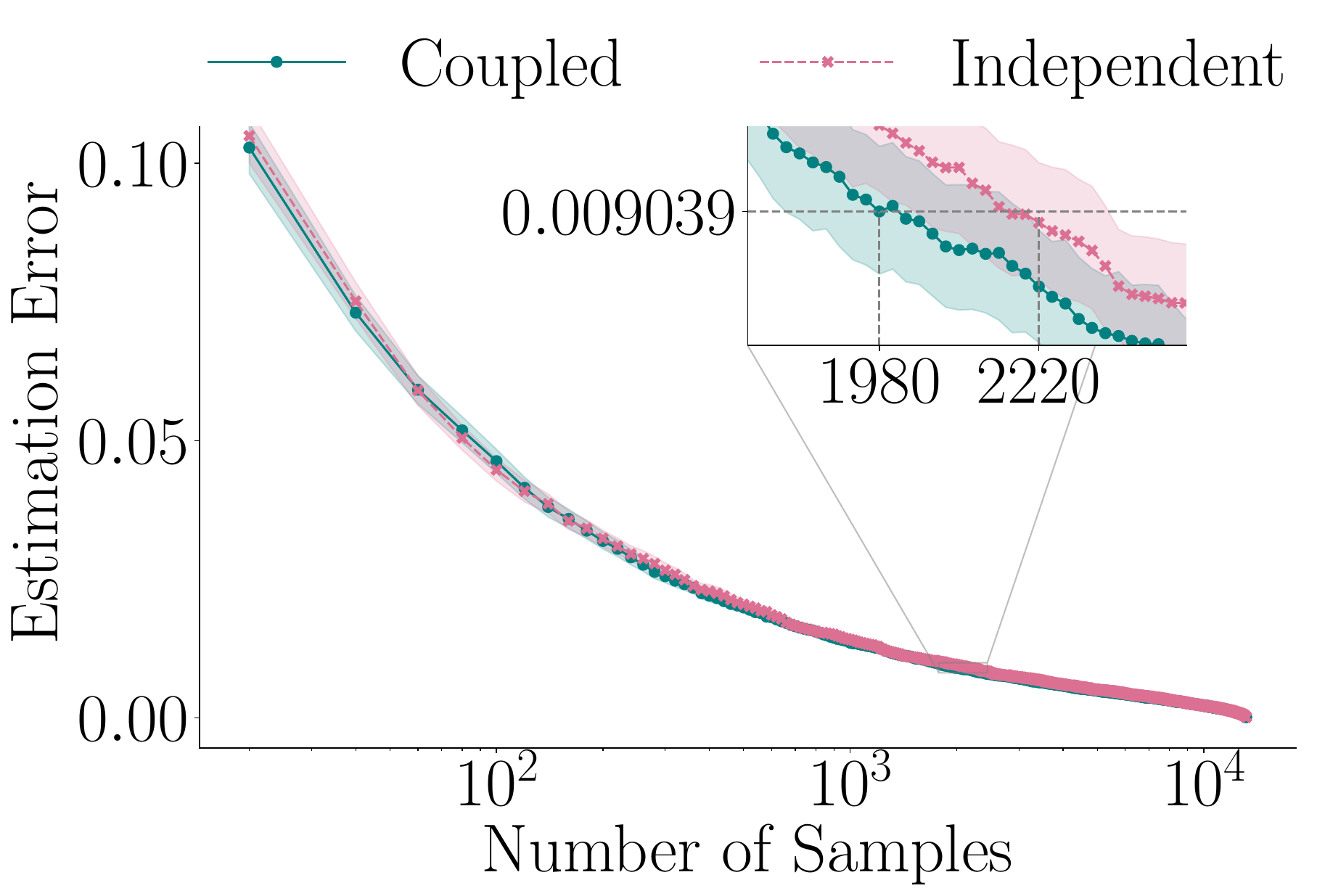} \\ \\
    \multicolumn{3}{c}{\texttt{1B} vs. \texttt{8B}}\\
    \includegraphics[width=0.23\linewidth]{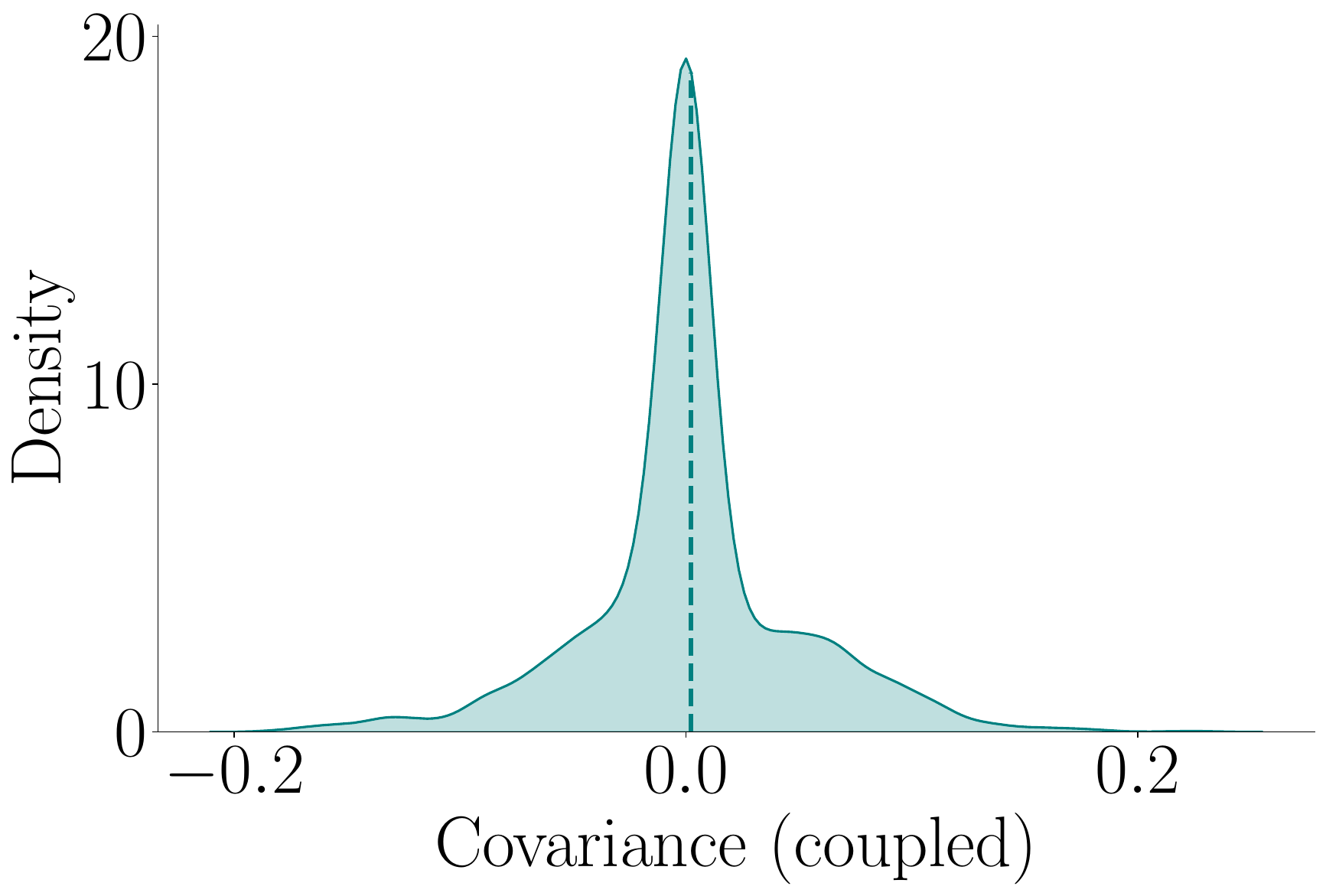} &
    \includegraphics[width=0.23\linewidth]{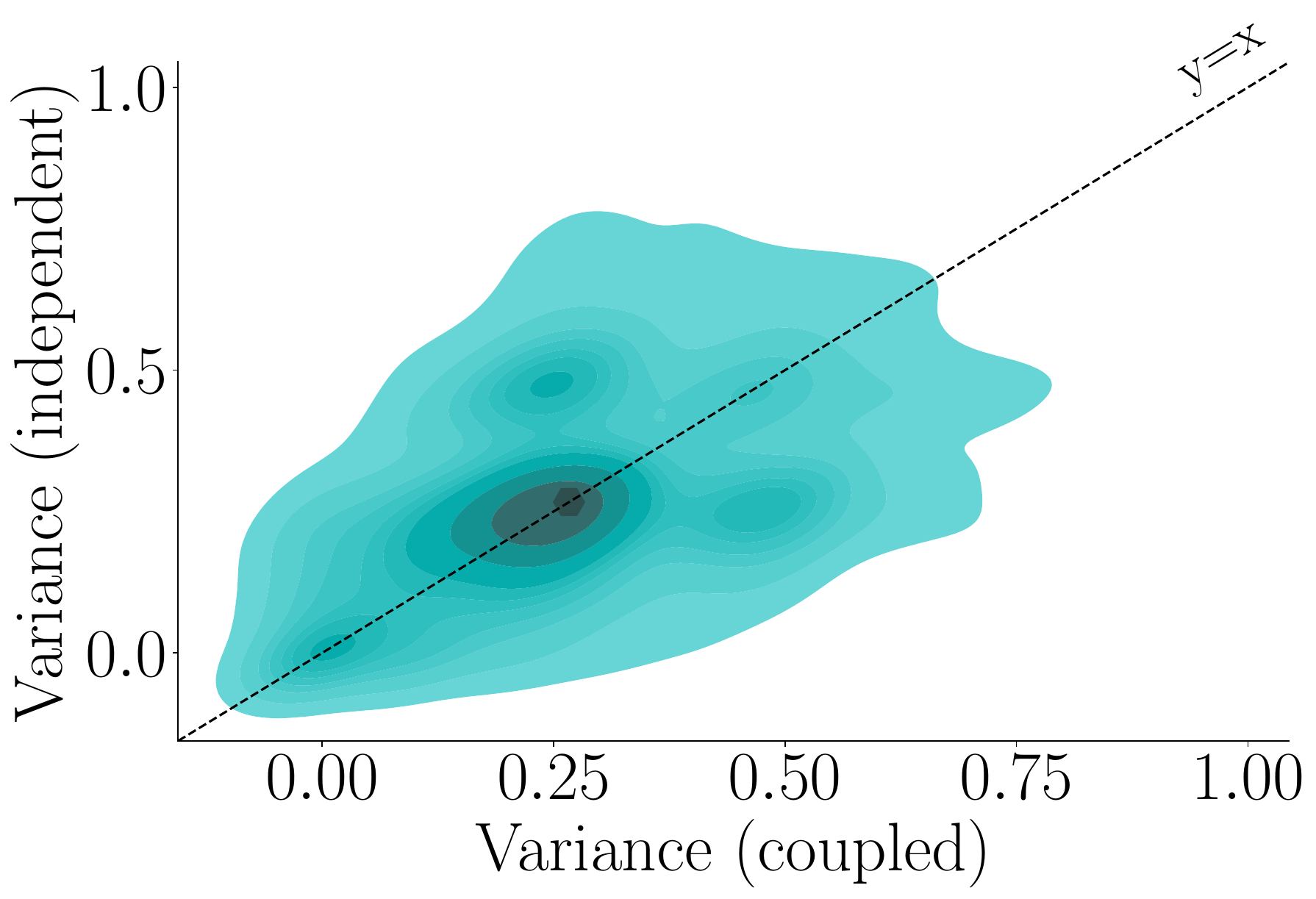} &
    \includegraphics[width=0.23\linewidth]{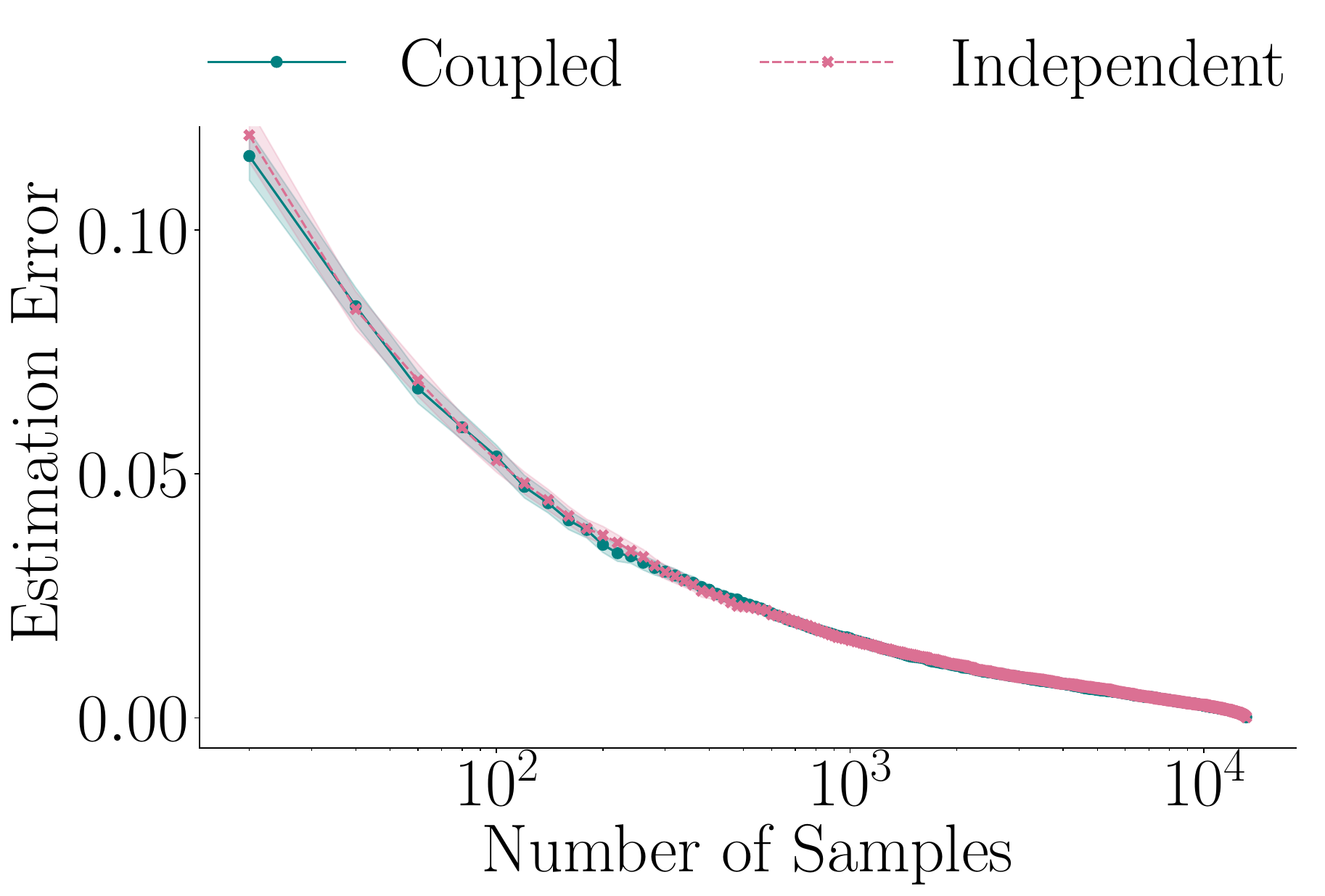} \\ \\
    \multicolumn{3}{c}{\texttt{3B} vs. \texttt{8B}}\\
    \includegraphics[width=0.23\linewidth]{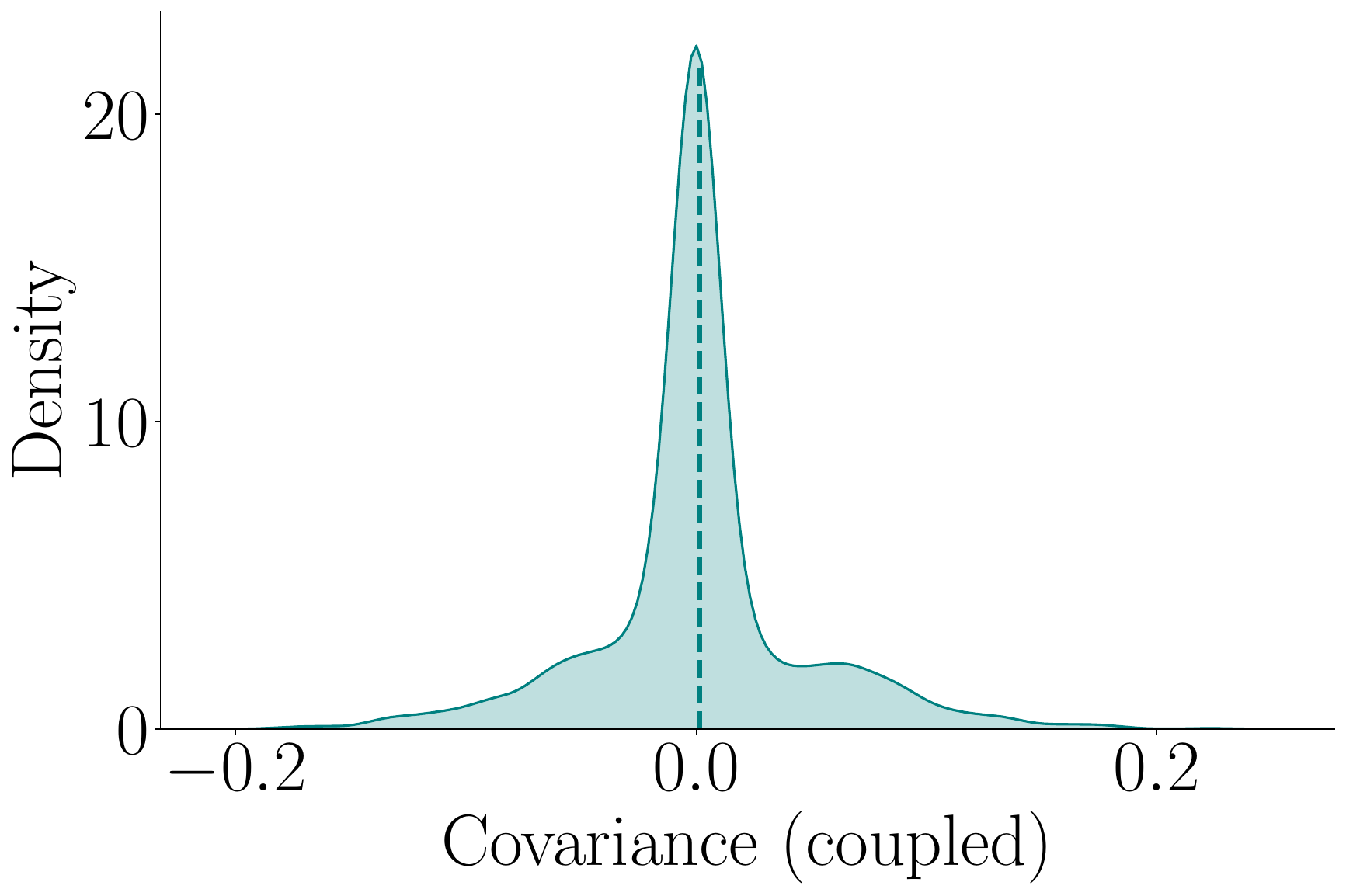} &
    \includegraphics[width=0.23\linewidth]{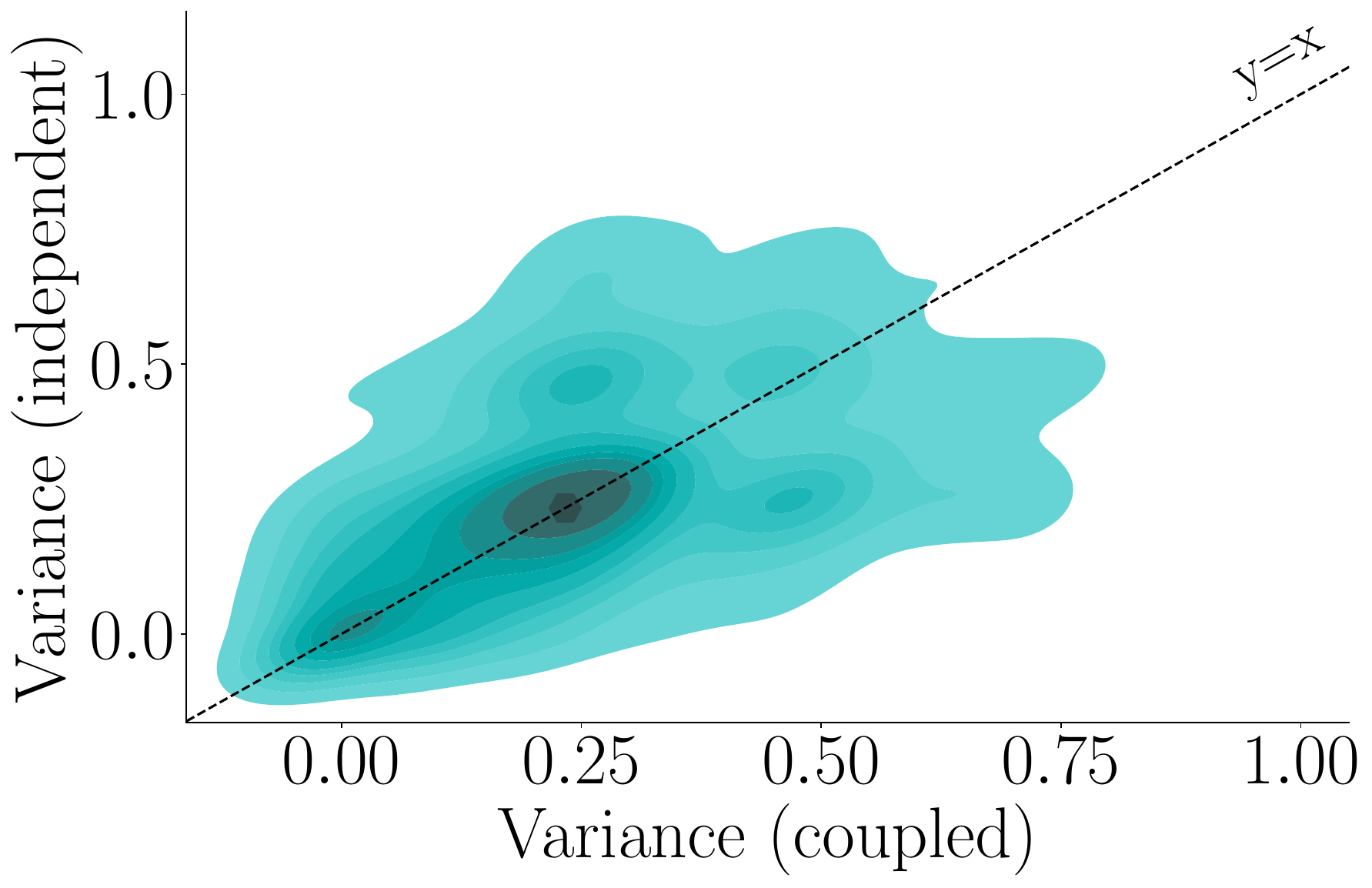} &
    \includegraphics[width=0.23\linewidth]{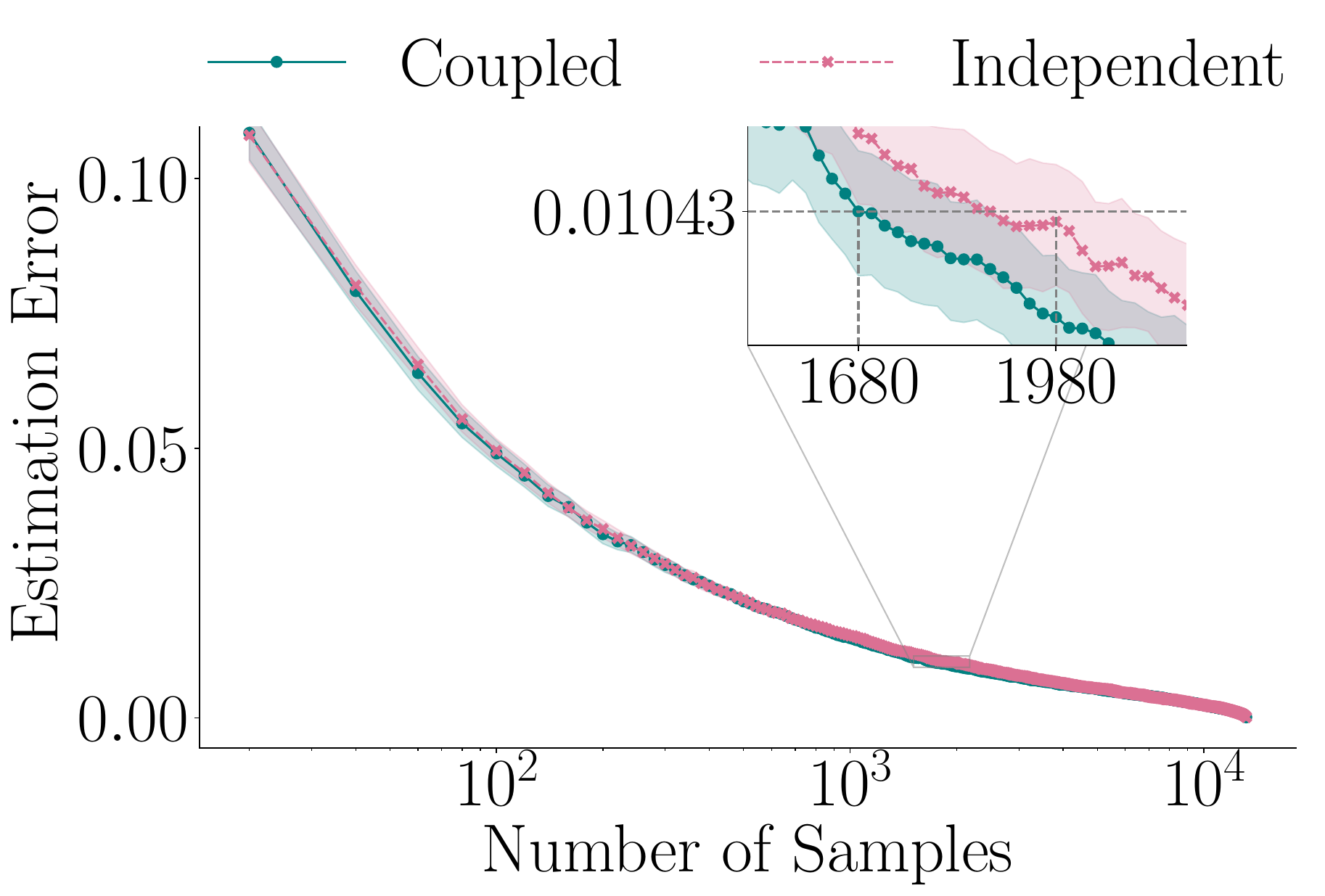} \\ \\
    \multicolumn{3}{c}{\texttt{8B} vs. \texttt{bnb-8bit}}\\
    \includegraphics[width=0.23\linewidth]{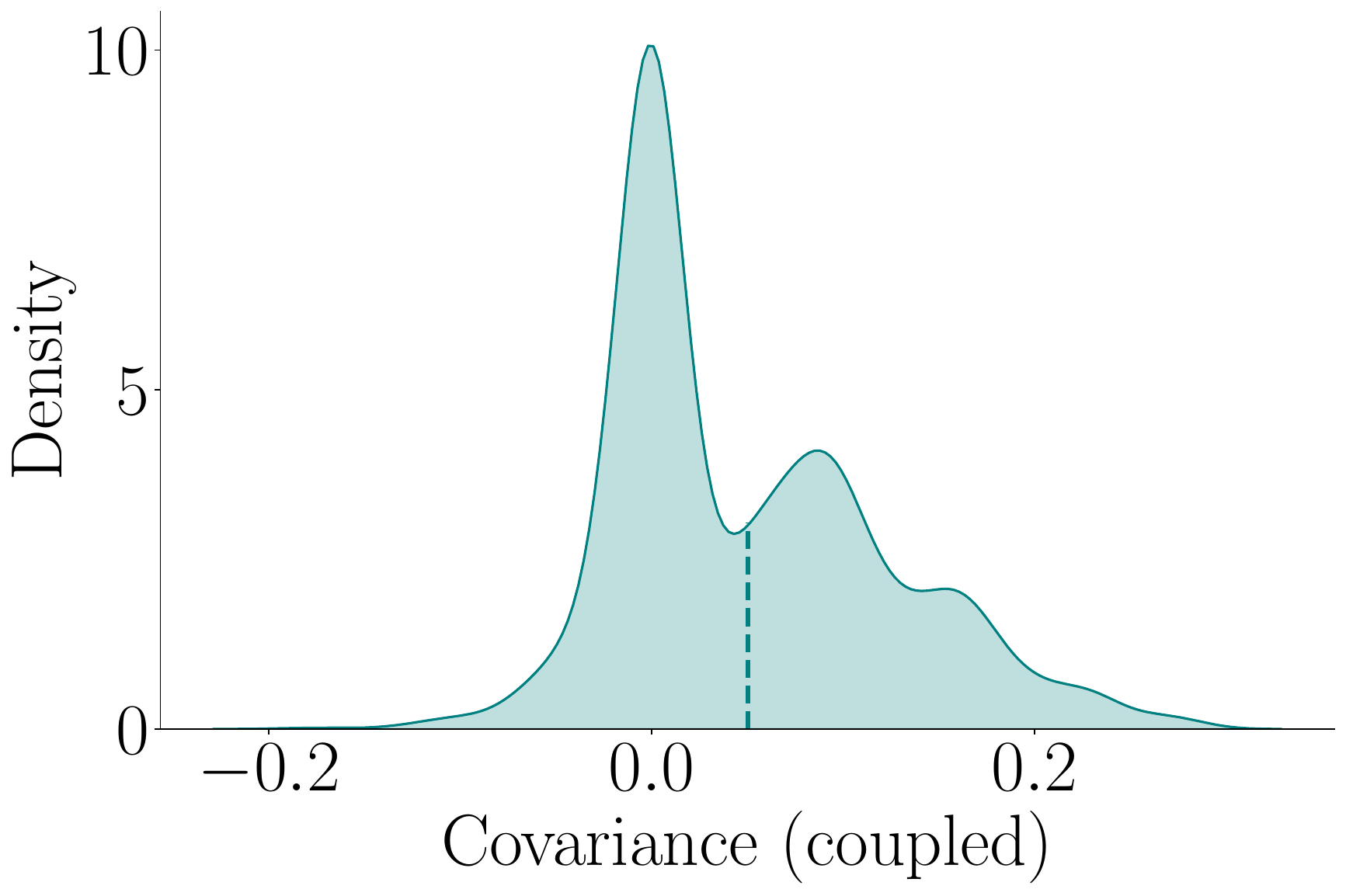} &
    \includegraphics[width=0.23\linewidth]{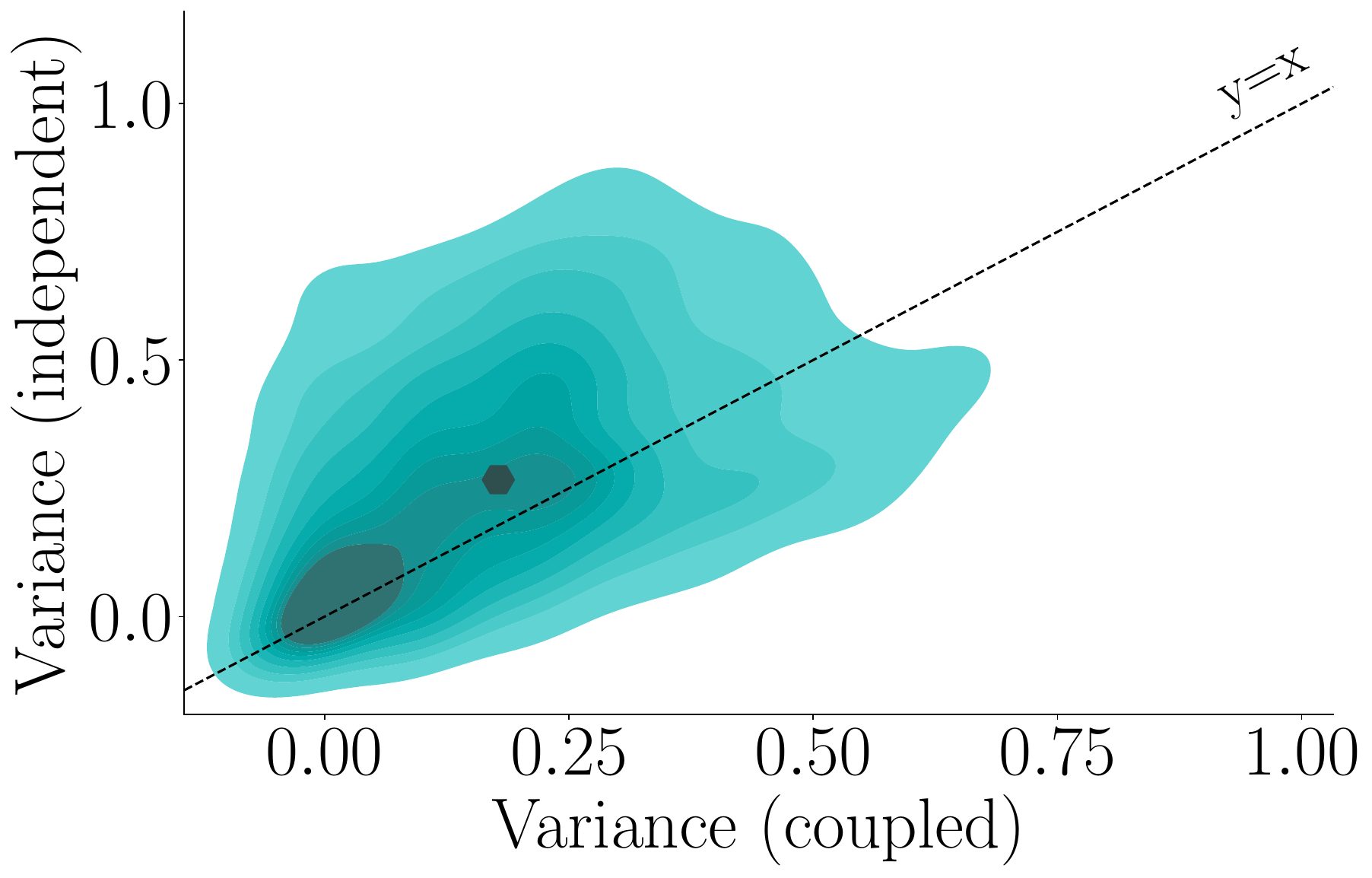} &
    \includegraphics[width=0.23\linewidth]{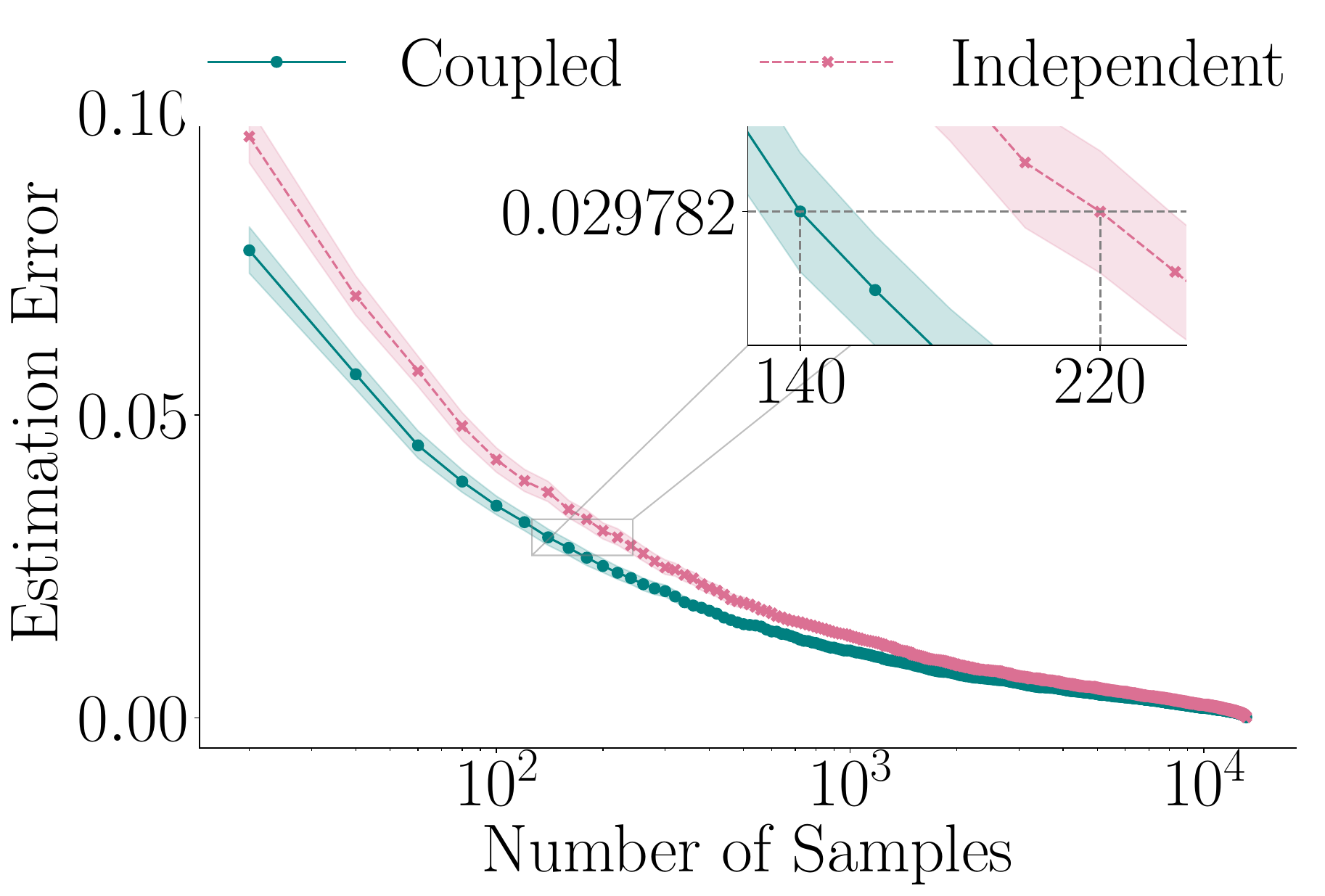} \\ \\

    \multicolumn{3}{c}{\texttt{8B} vs. \texttt{bnb-4bit}}\\
    \includegraphics[width=0.23\linewidth]{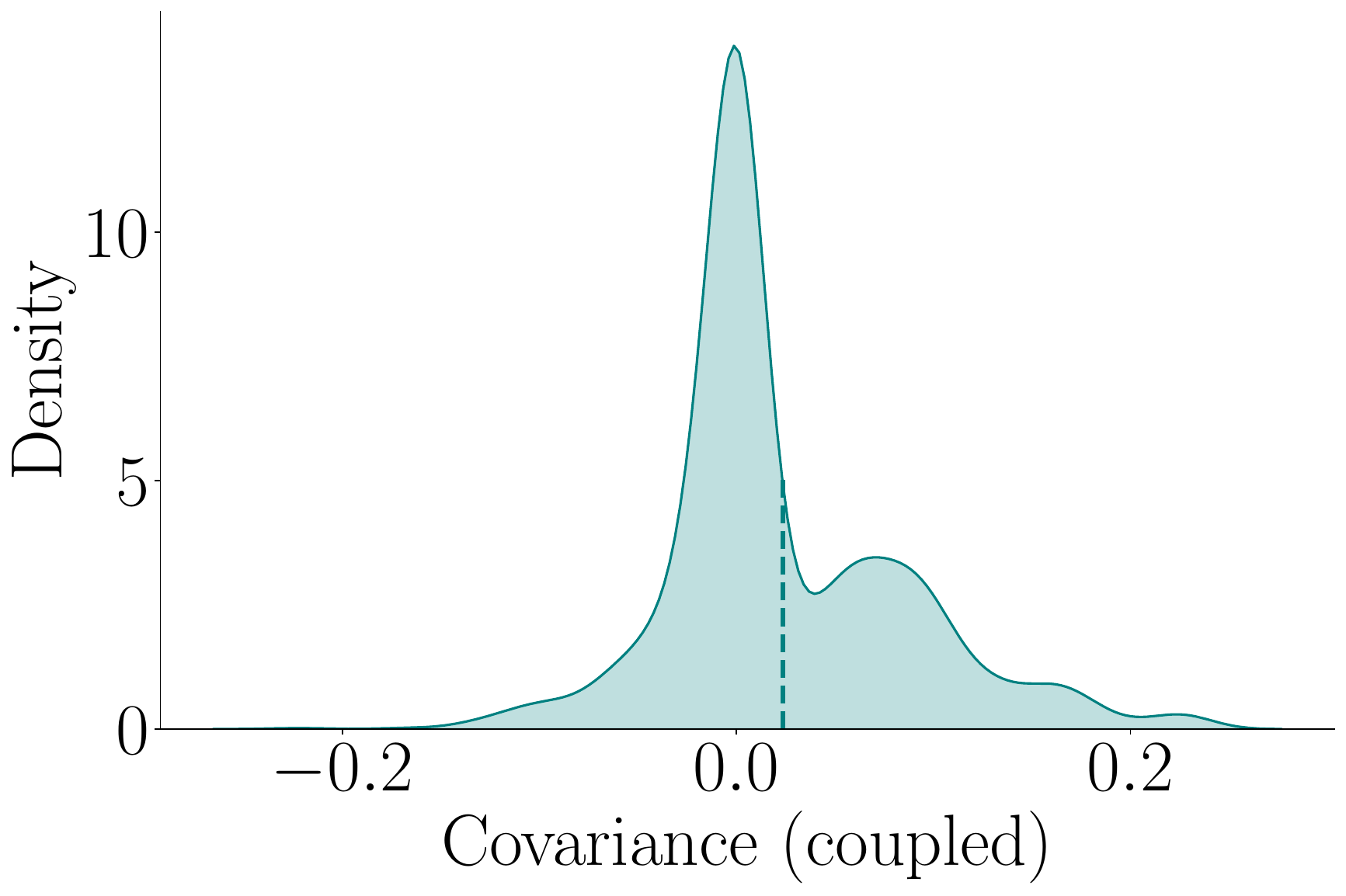}&
    \includegraphics[width=0.23\linewidth]{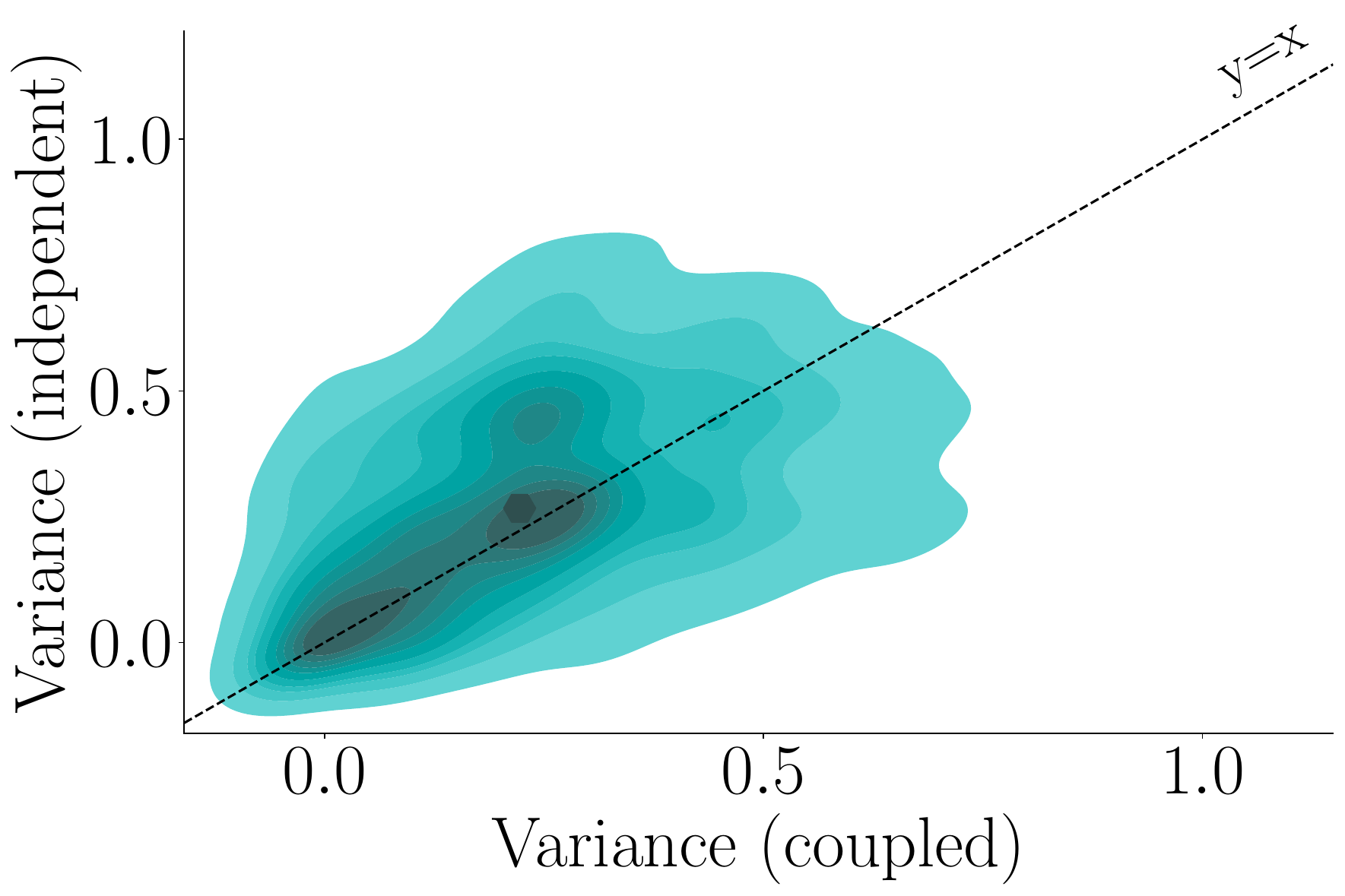} &
    \includegraphics[width=0.23\linewidth]{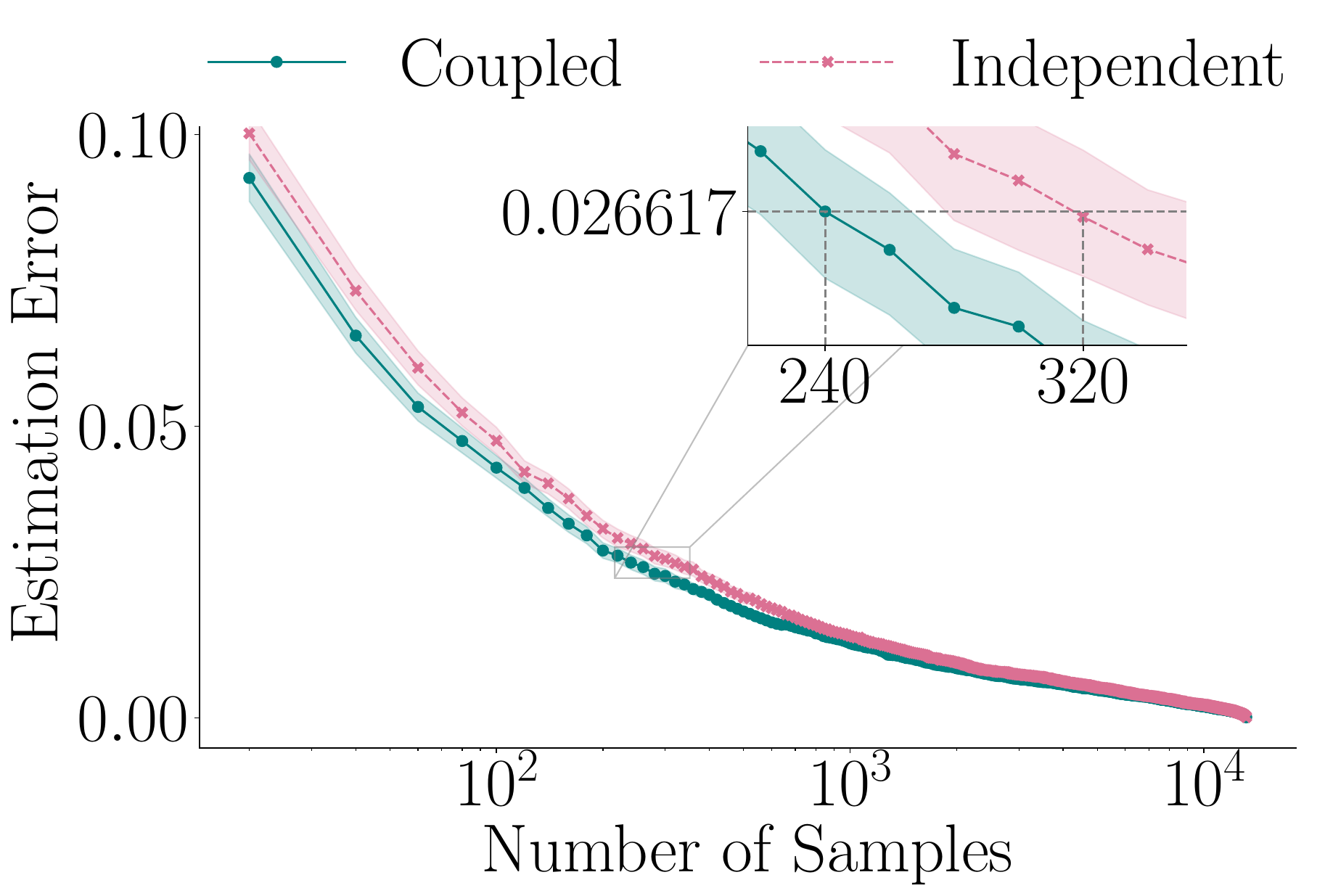} \\ \\
    (a) Score covariance & (b) Variance of the score difference & (c) Estimation error vs. \# samples \\ 
\end{tabular}
    \caption{\textbf{Comparison between several pairs of LLMs in the \texttt{Llama} family on math questions from the GSM8K dataset.}
    Panels in column (a) show the kernel density estimate (KDE) of the covariance between the scores of the two LLMs on each problem under coupled generation; the dashed lines correspond to average values. Panels in column (b) show the KDE of the variance of the difference between the scores of the LLMs on each question under coupled and independent generation; the highlighted points correspond to median values. Panels in column (c) show the absolute error in the estimation of the expected difference between the scores of the LLMs against the number of samples; for each point on the x-axis, we perform $1{,}000$ sub-samplings and shaded areas correspond to $95\%$ confidence intervals.}
    \label{fig:gsm8k-first-5}
\end{figure}

\begin{figure}[ht]
\centering
\begin{tabular}{c c c}
    \multicolumn{3}{c}{\texttt{8B} vs. \texttt{AWQ-INT4}}\\
    \includegraphics[width=0.23\linewidth]{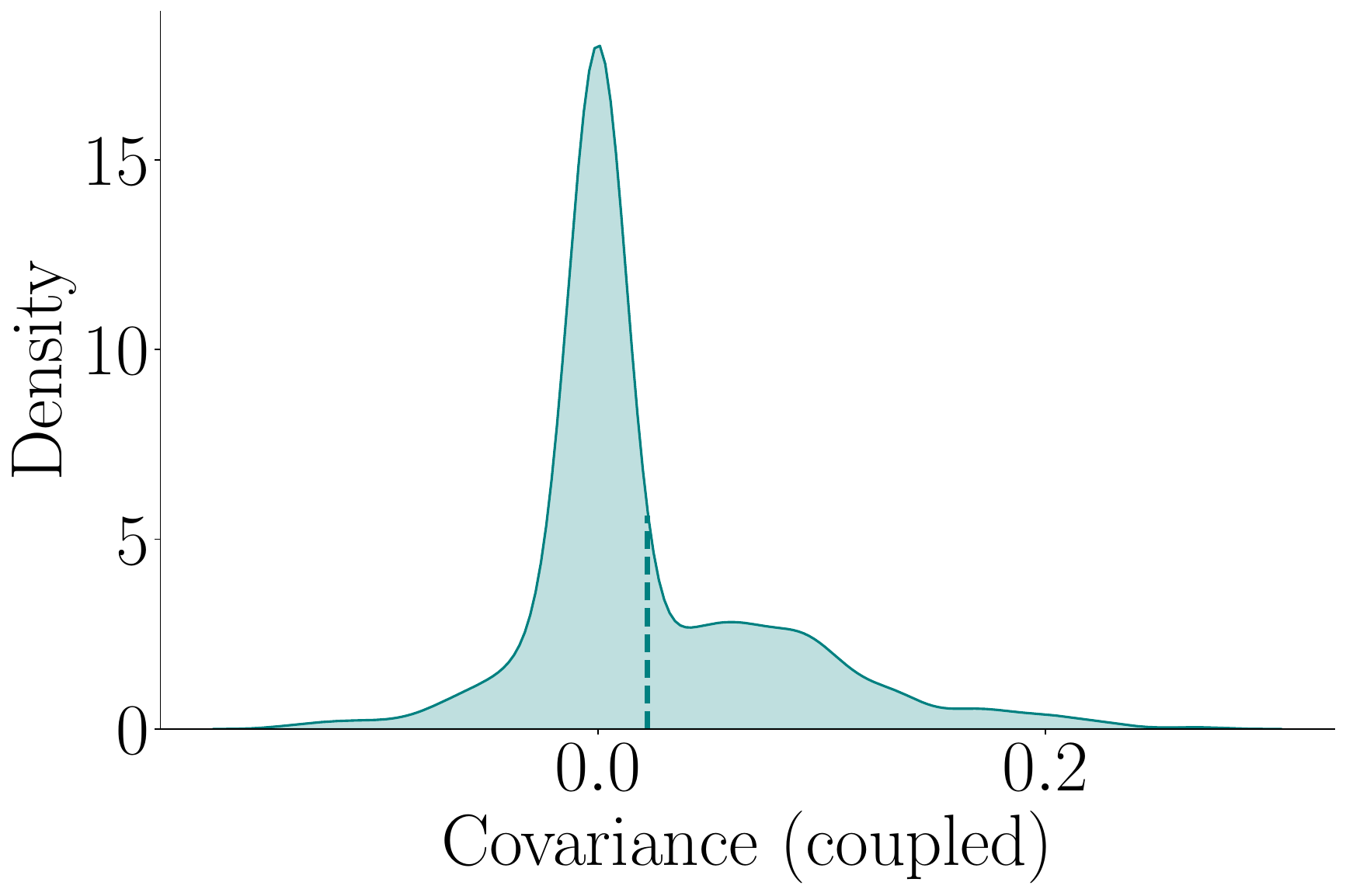} &
    \includegraphics[width=0.23\linewidth]{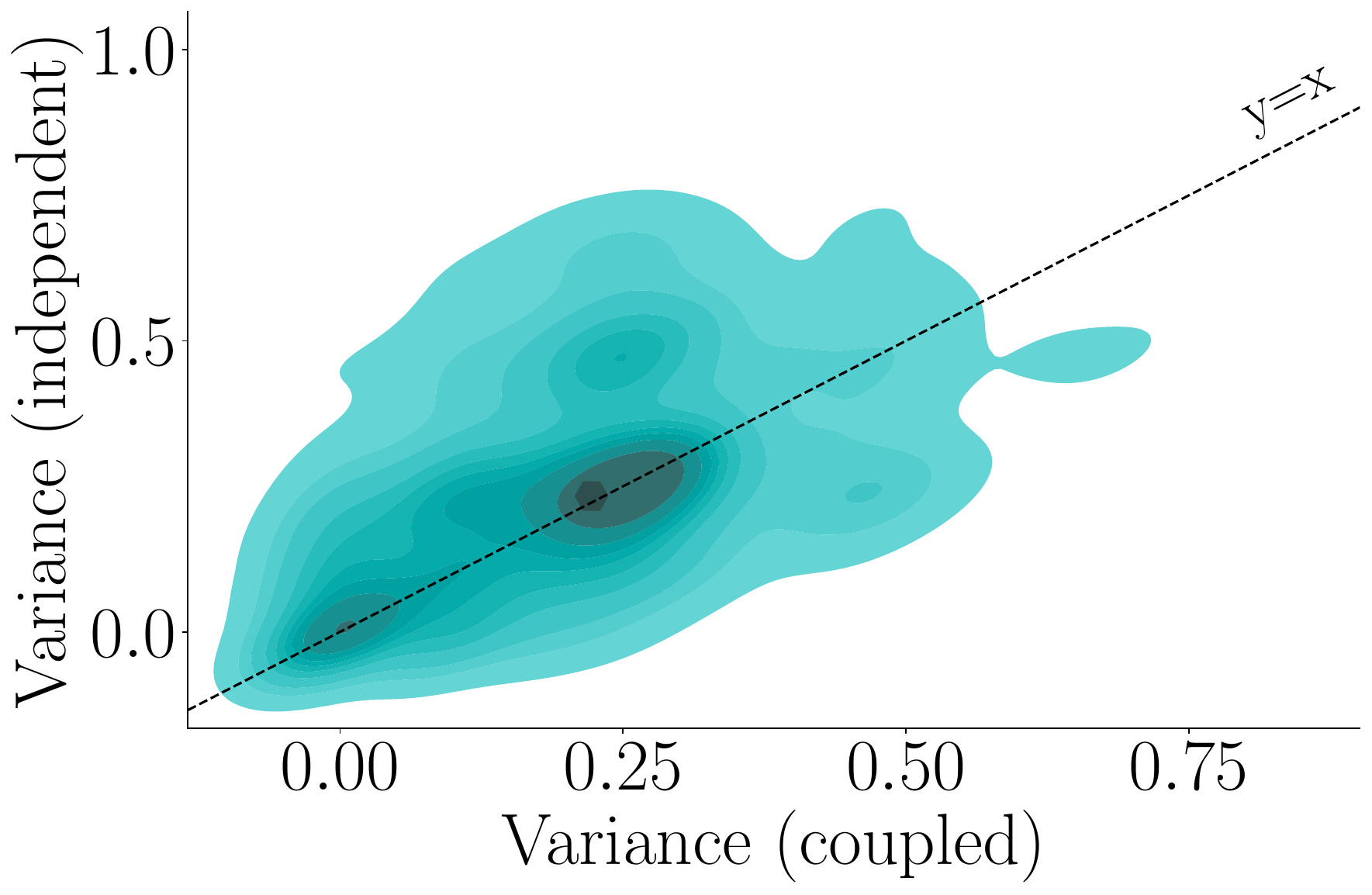} &
    \includegraphics[width=0.23\linewidth]{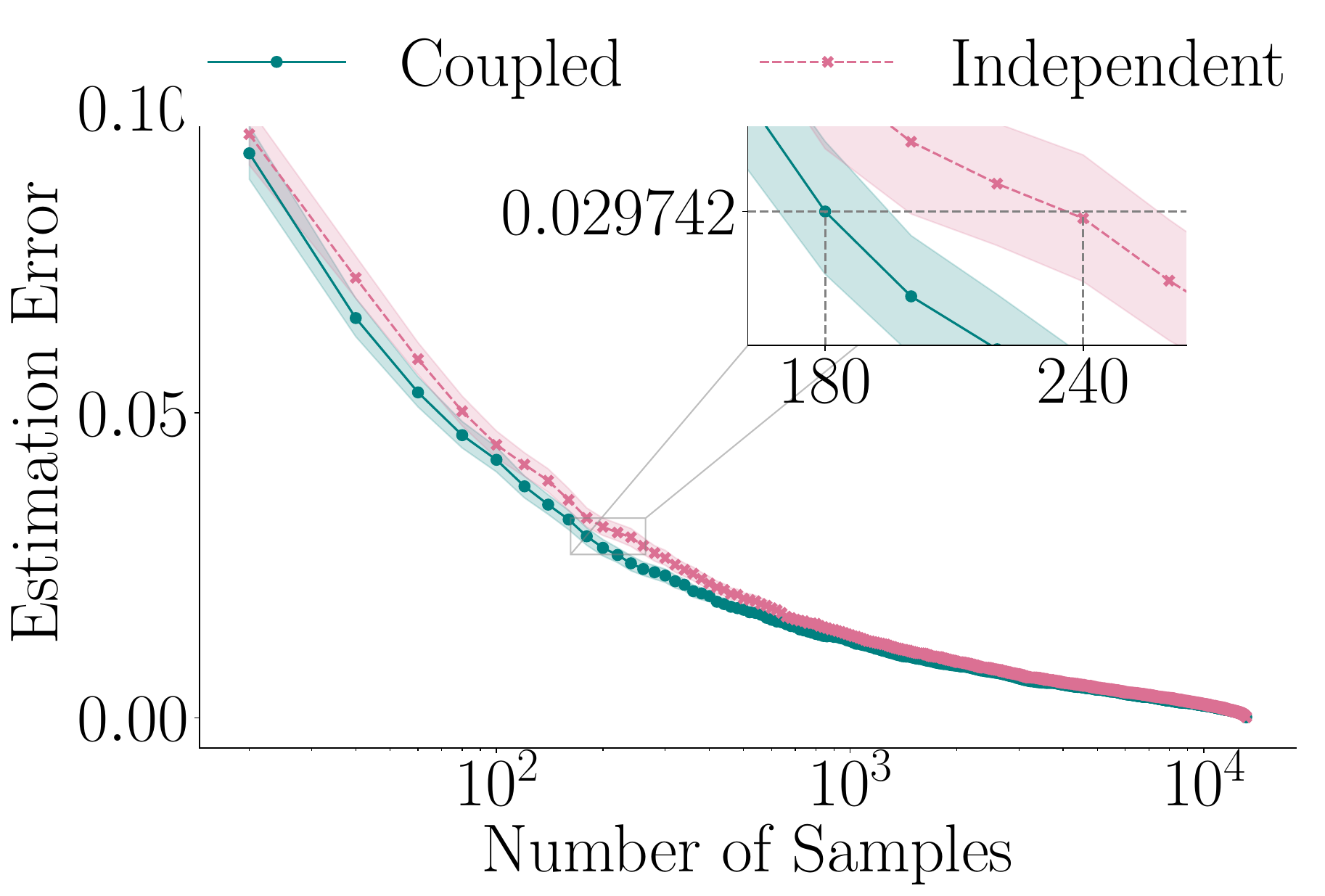} \\ \\
    \multicolumn{3}{c}{\texttt{3B} vs. \texttt{bnb-8bit}}\\
    \includegraphics[width=0.23\linewidth]{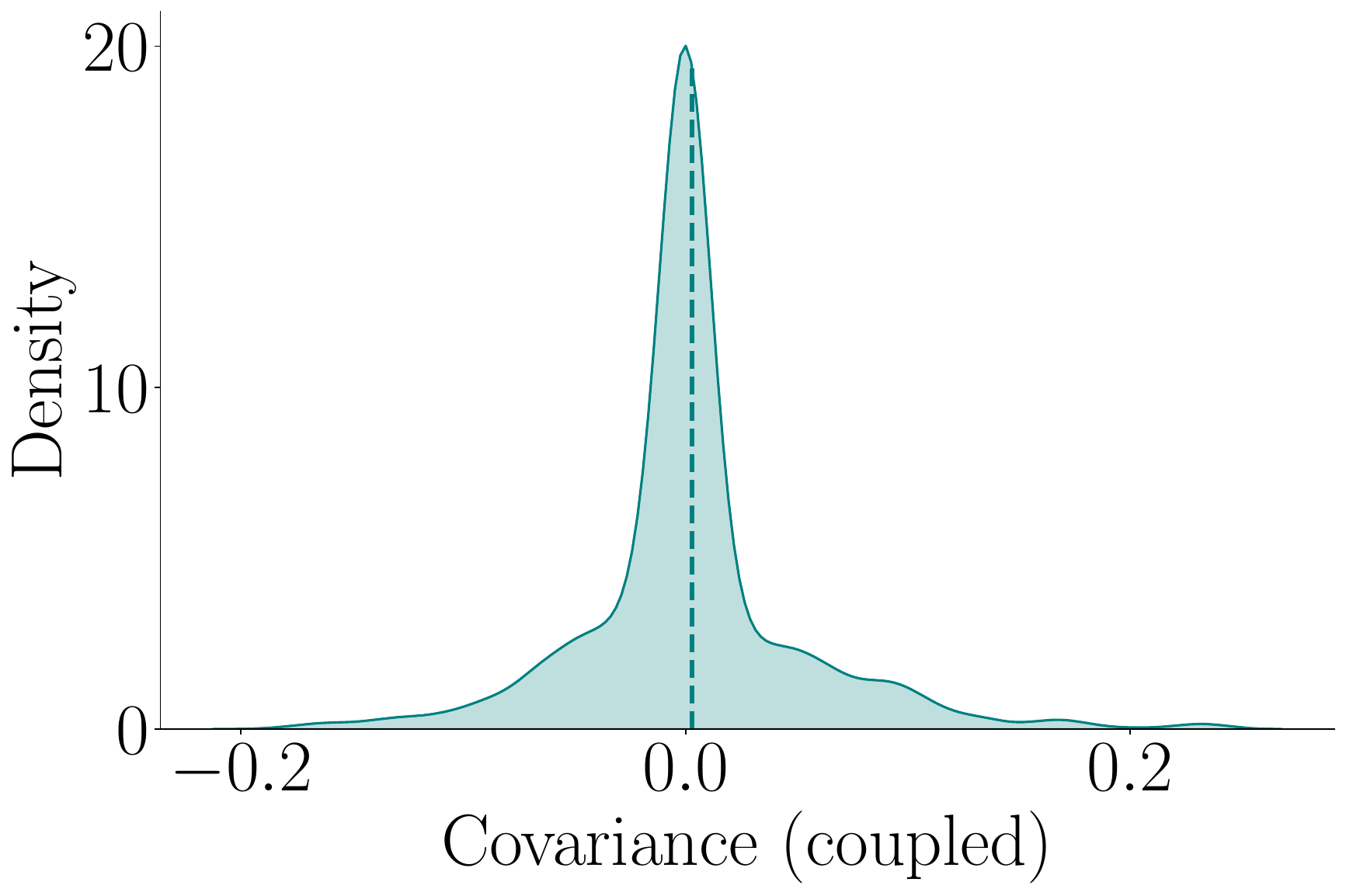} &
    \includegraphics[width=0.23\linewidth]{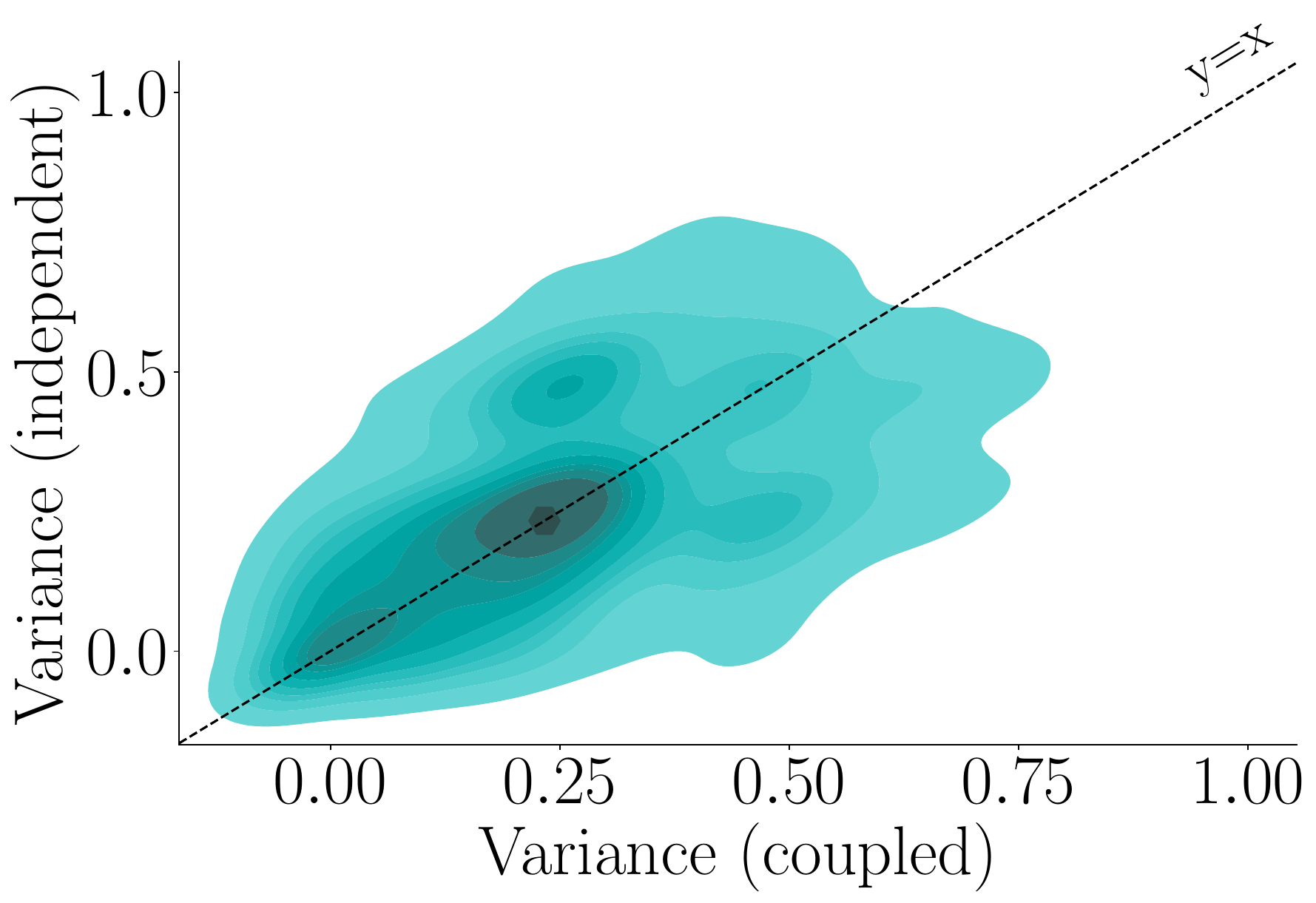} &
    \includegraphics[width=0.23\linewidth]{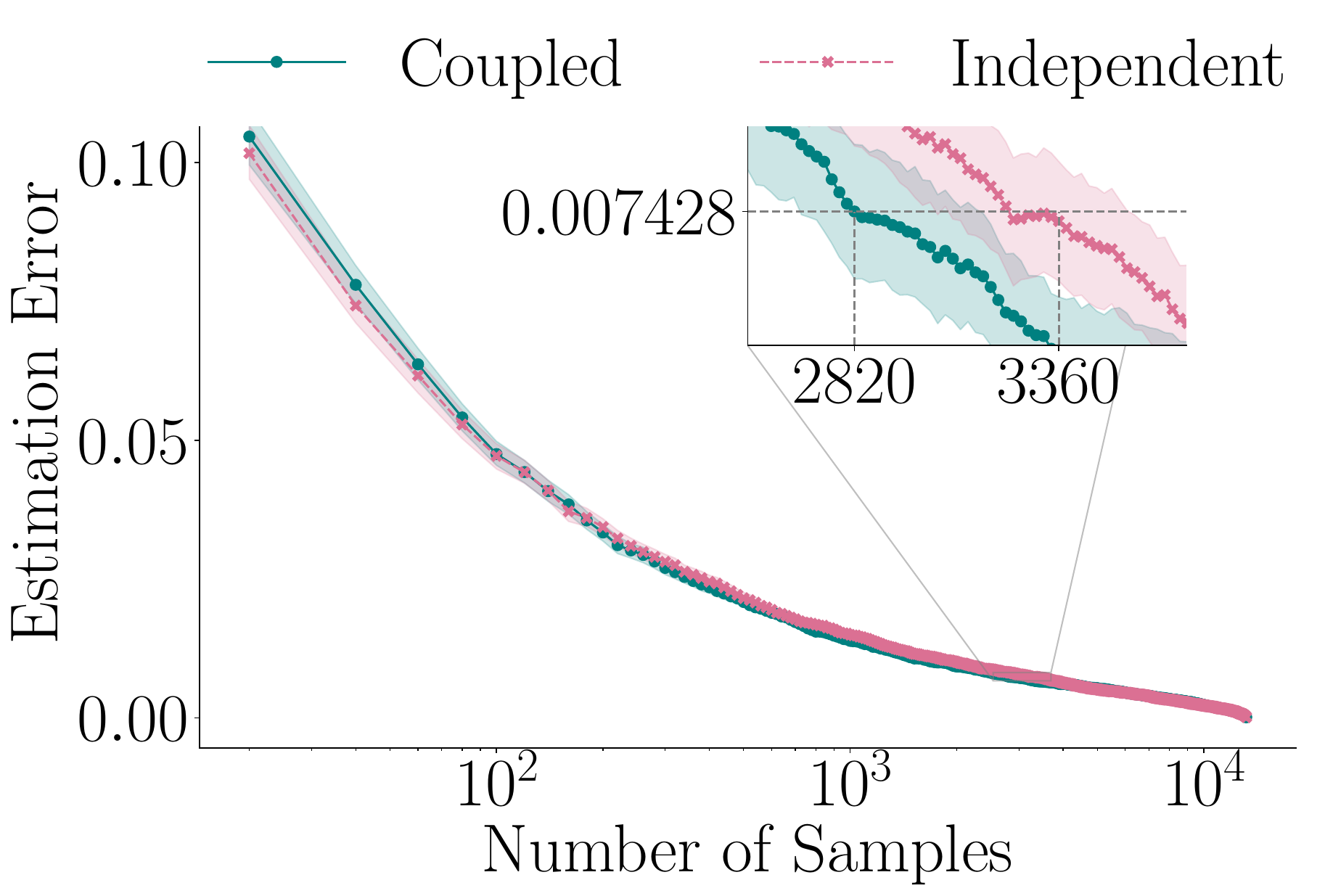} \\ \\
    \multicolumn{3}{c}{\texttt{3B} vs. \texttt{bnb-4bit}}\\
    \includegraphics[width=0.23\linewidth]{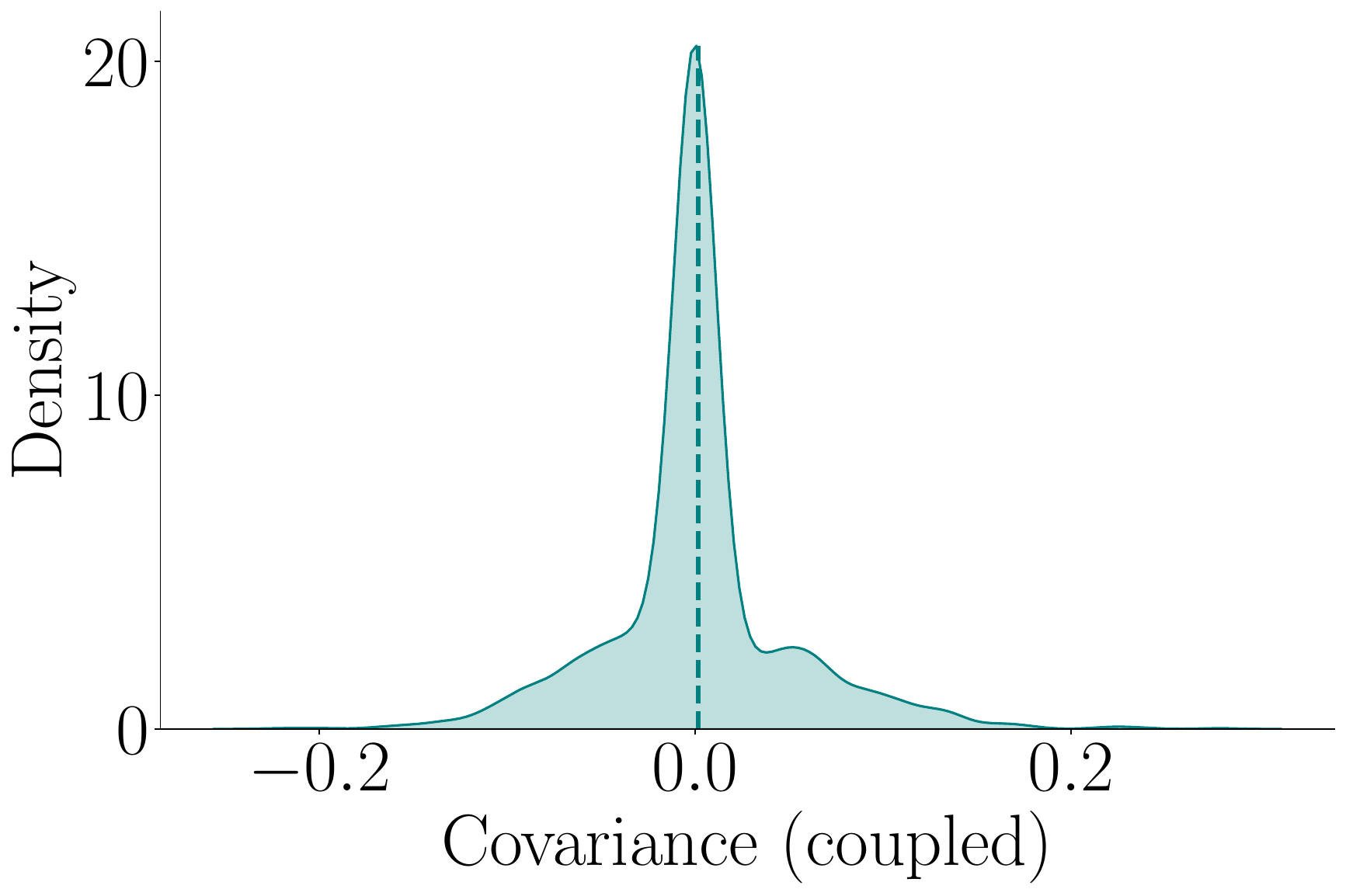} &
    \includegraphics[width=0.23\linewidth]{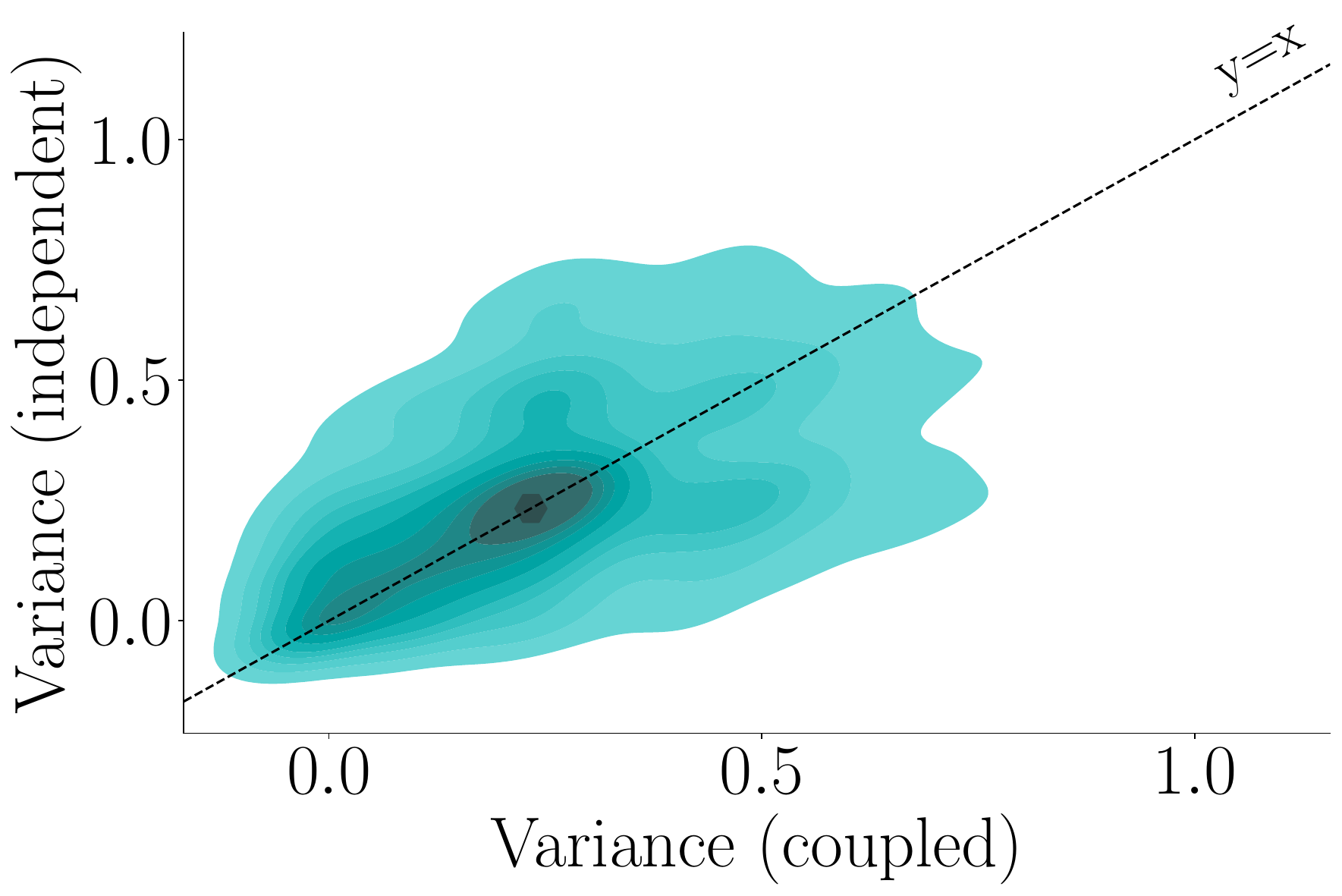} &
    \includegraphics[width=0.23\linewidth]{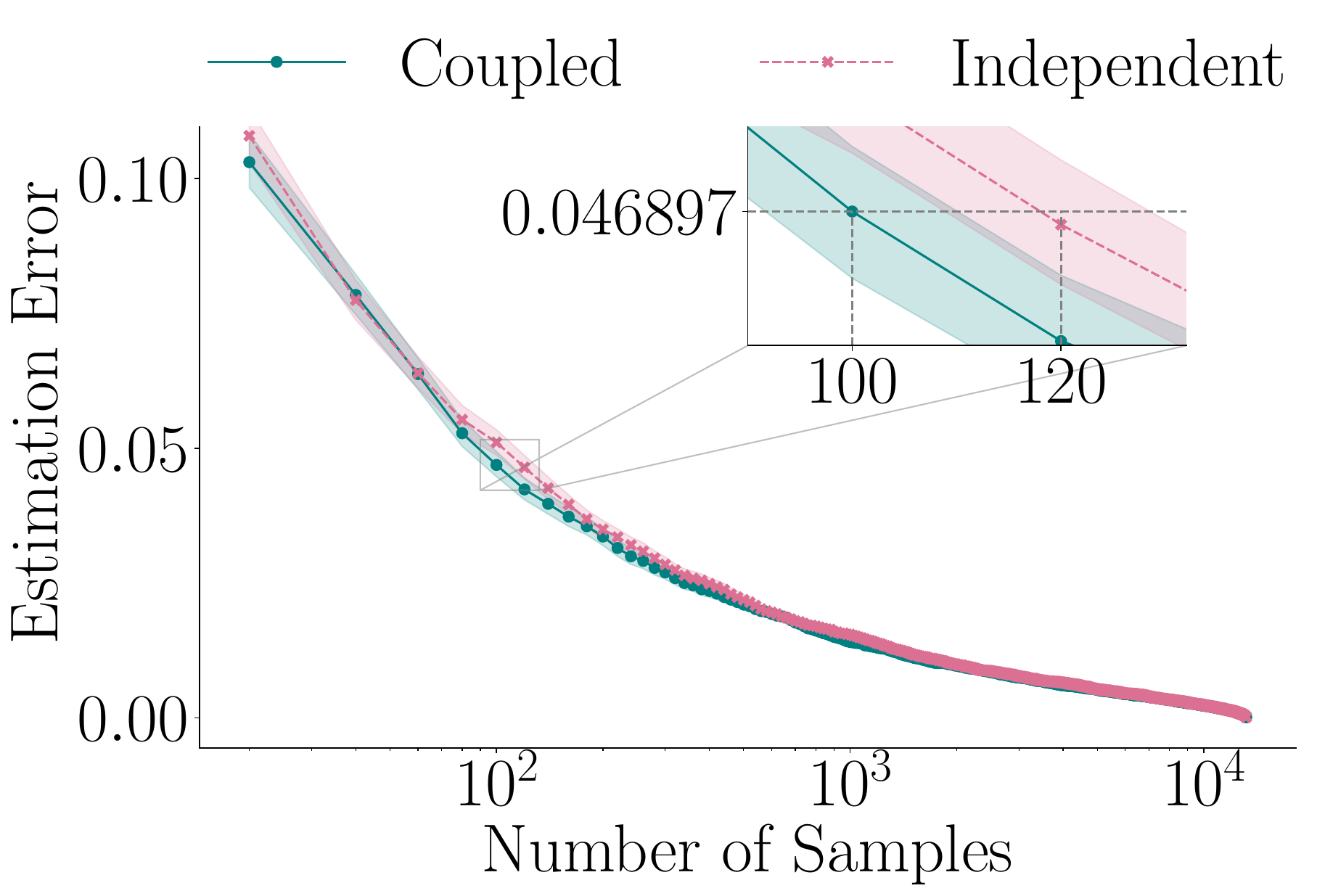} \\ \\
    \multicolumn{3}{c}{\texttt{3B} vs. \texttt{AWQ-INT4}}\\
    \includegraphics[width=0.23\linewidth]{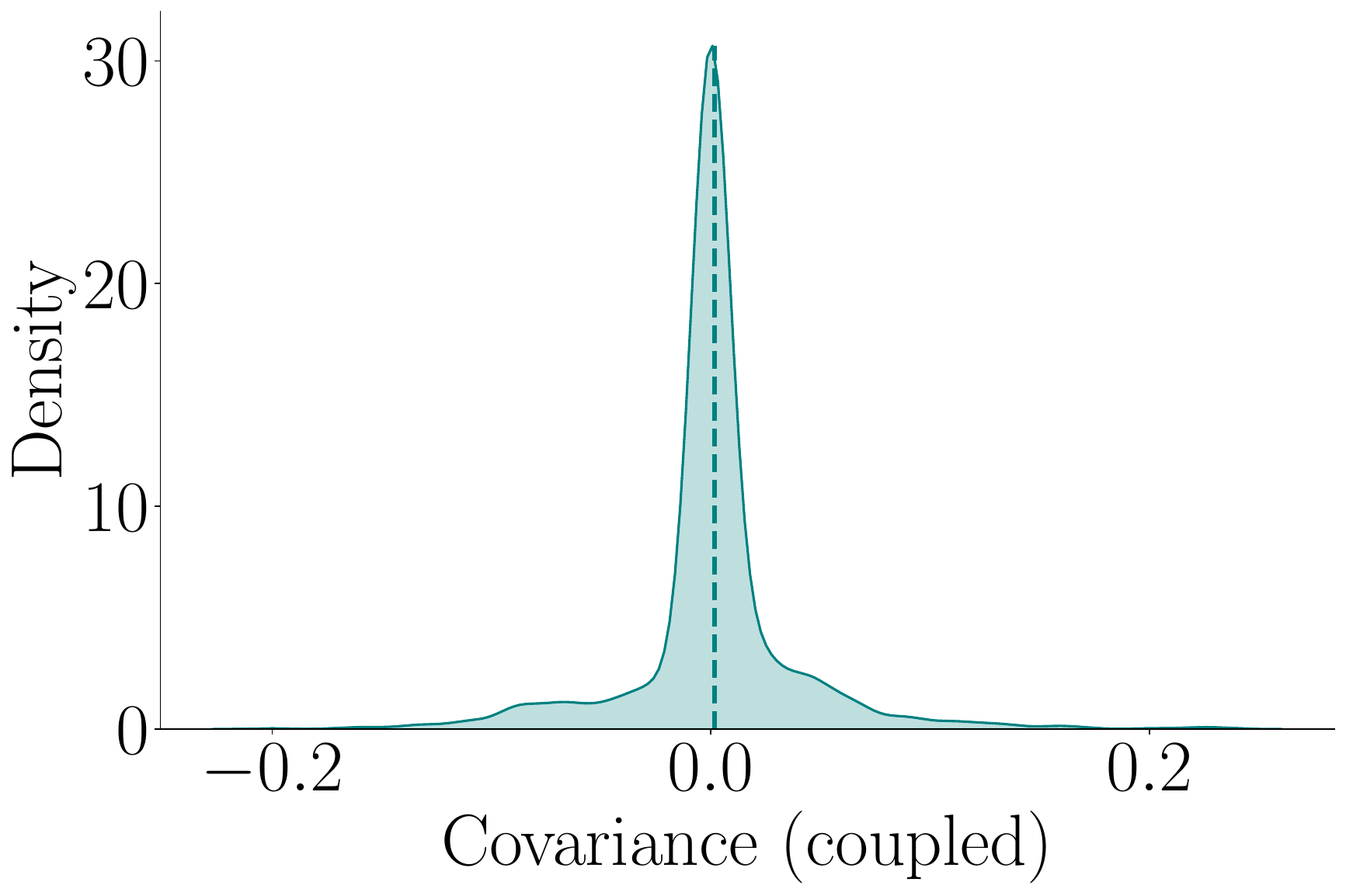} &
    \includegraphics[width=0.23\linewidth]{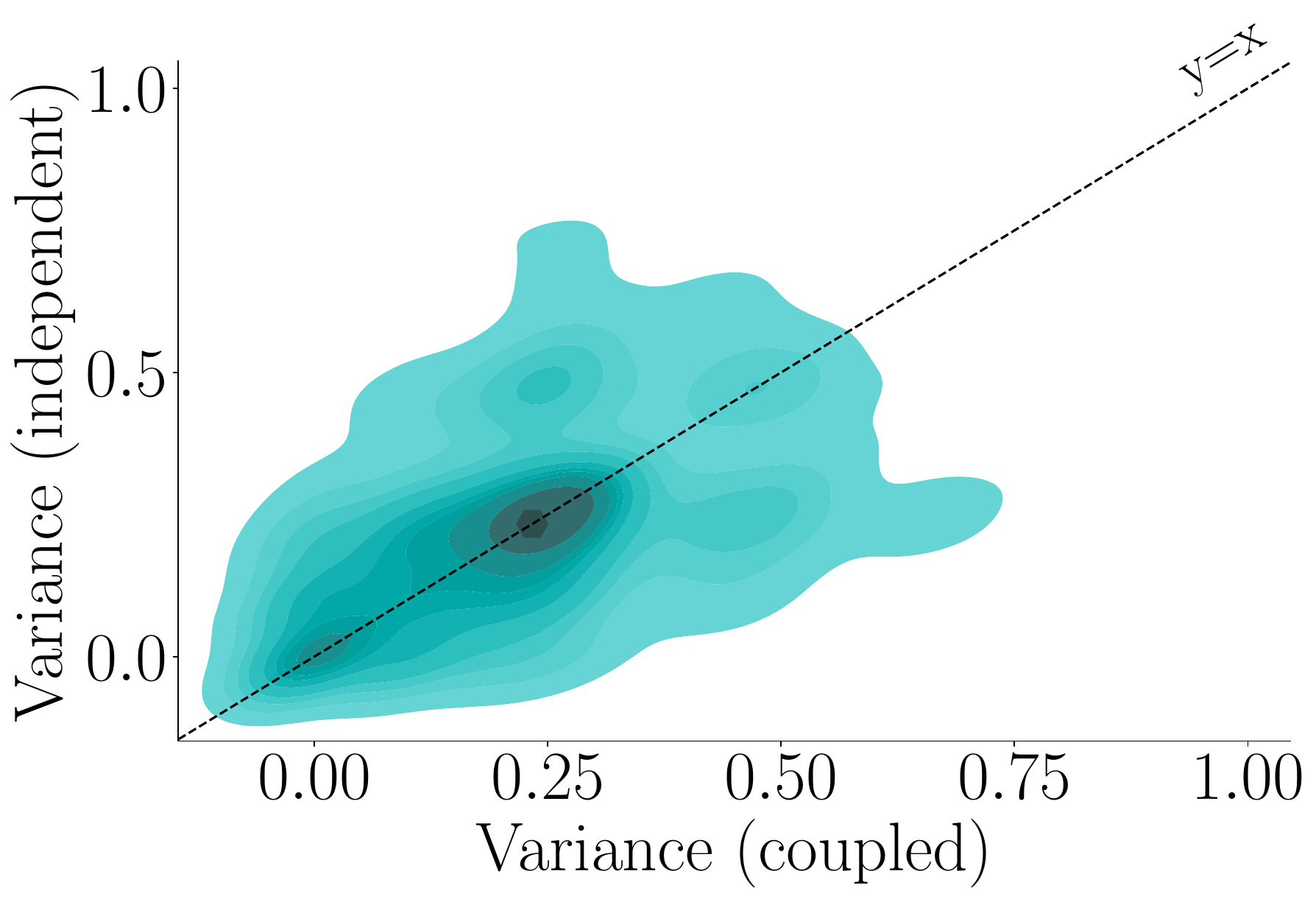} &
    \includegraphics[width=0.23\linewidth]{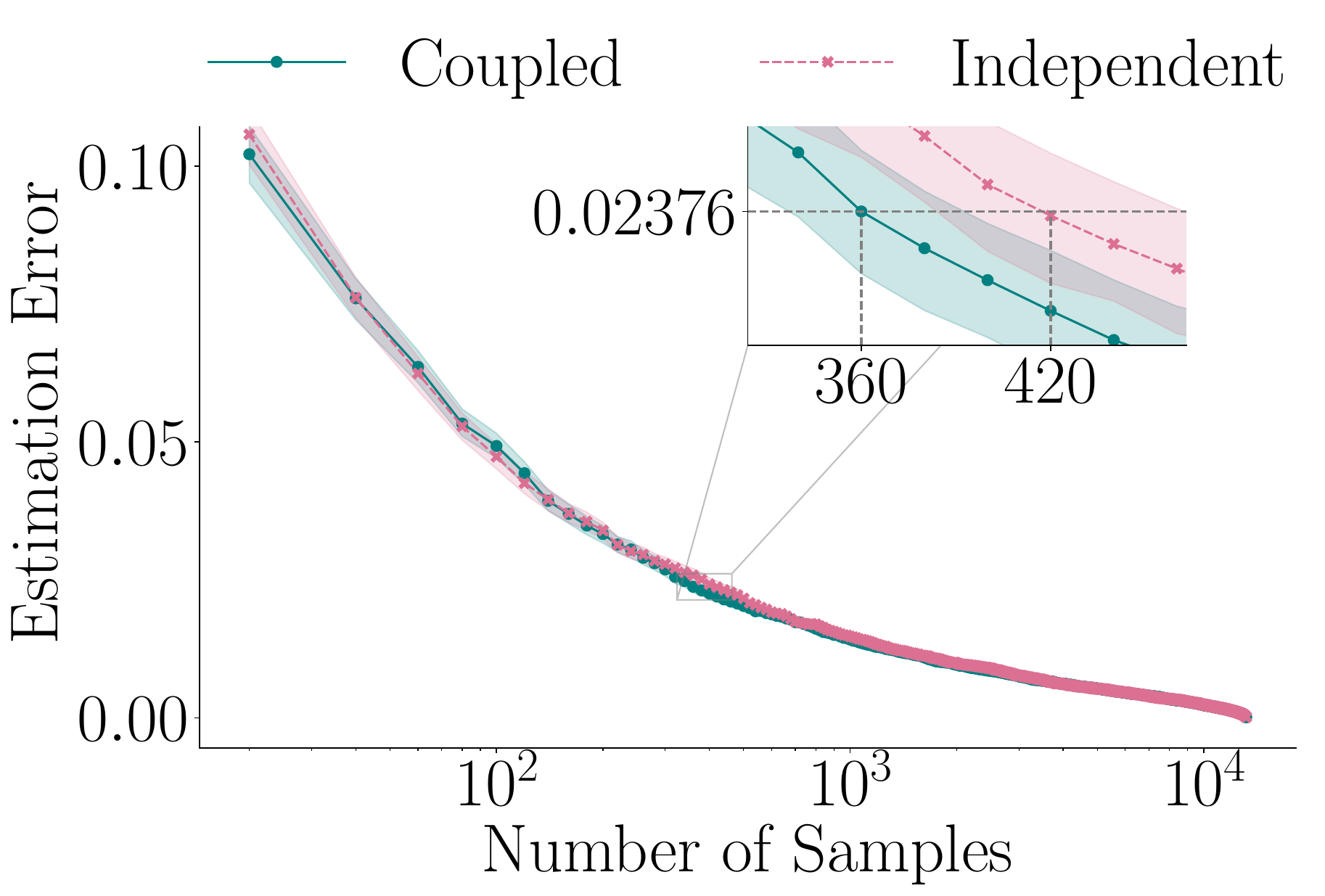} \\ \\

    \multicolumn{3}{c}{\texttt{1B} vs. \texttt{bnb-8bit}}\\
    \includegraphics[width=0.23\linewidth]{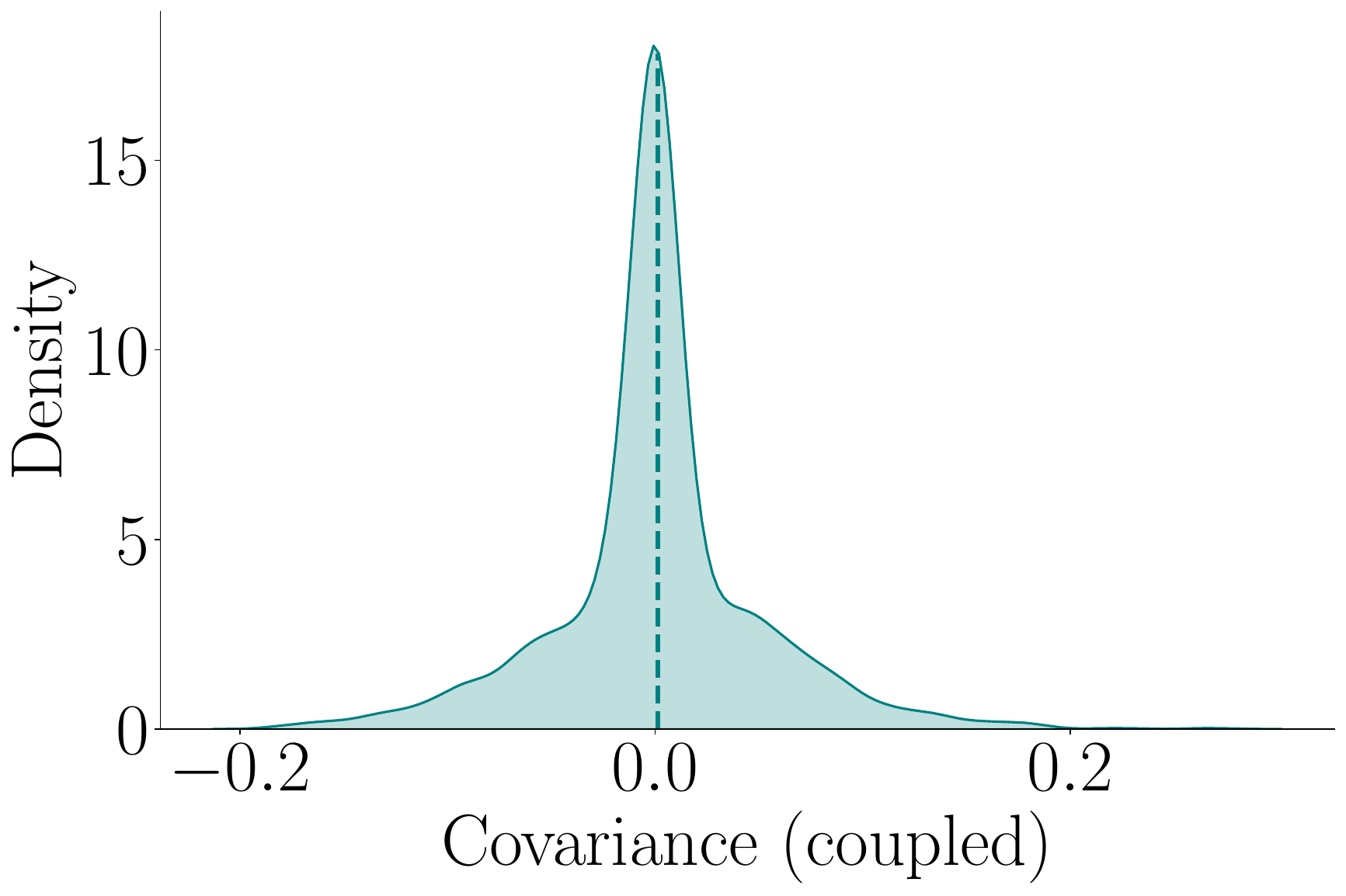} &
    \includegraphics[width=0.23\linewidth]{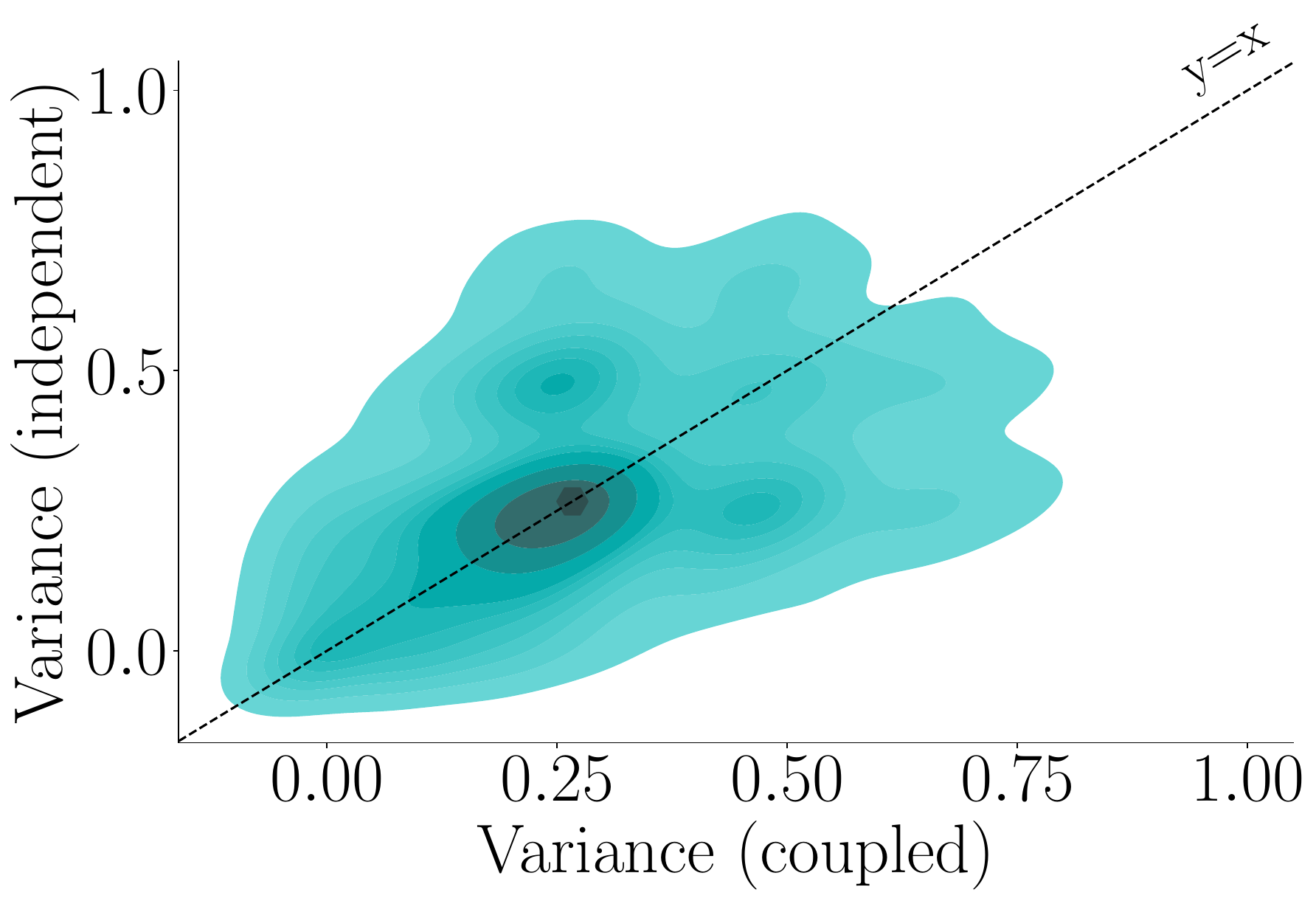} &
    \includegraphics[width=0.23\linewidth]{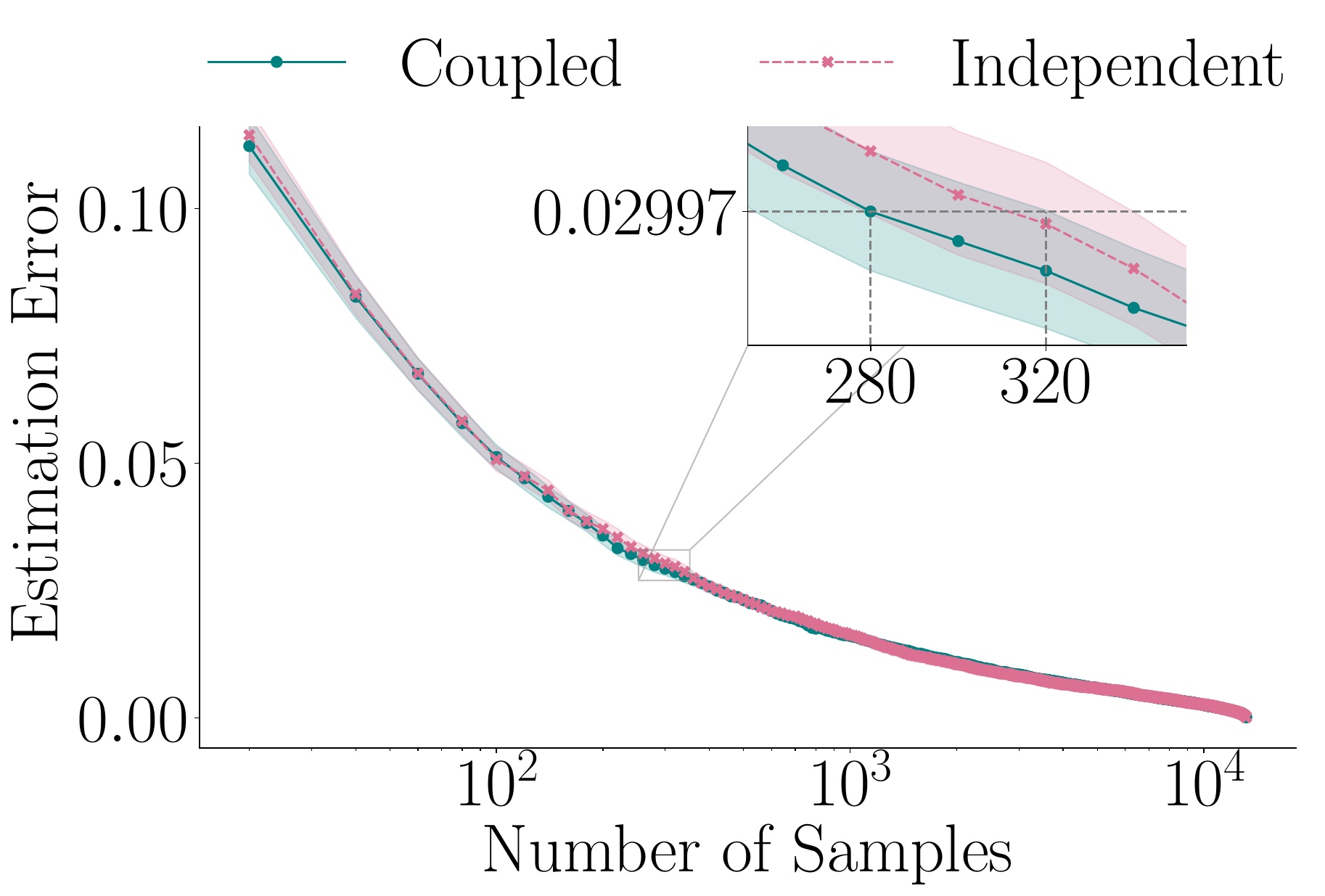} \\ \\
    (a) Score covariance & (b) Variance of the score difference & (c) Estimation error vs. \# samples \\

\end{tabular}
    \caption{\textbf{Comparison between several pairs of LLMs in the \texttt{Llama} family on math questions from the GSM8K dataset.}
    Panels in column (a) show the kernel density estimate (KDE) of the covariance between the scores of the two LLMs on each problem under coupled generation; the dashed lines correspond to average values. Panels in column (b) show the KDE of the variance of the difference between the scores of the LLMs on each question under coupled and independent generation; the highlighted points correspond to median values. Panels in column (c) show the absolute error in the estimation of the expected difference between the scores of the LLMs against the number of samples; for each point on the x-axis, we perform $1{,}000$ sub-samplings and shaded areas correspond to $95\%$ confidence intervals.}
    \label{fig:gsm8k-second-5}
\end{figure}

\begin{figure}[ht]
\centering
\begin{tabular}{c c c}
    \multicolumn{3}{c}{\texttt{1B} vs. \texttt{bnb-4bit}}\\
    \includegraphics[width=0.23\linewidth]{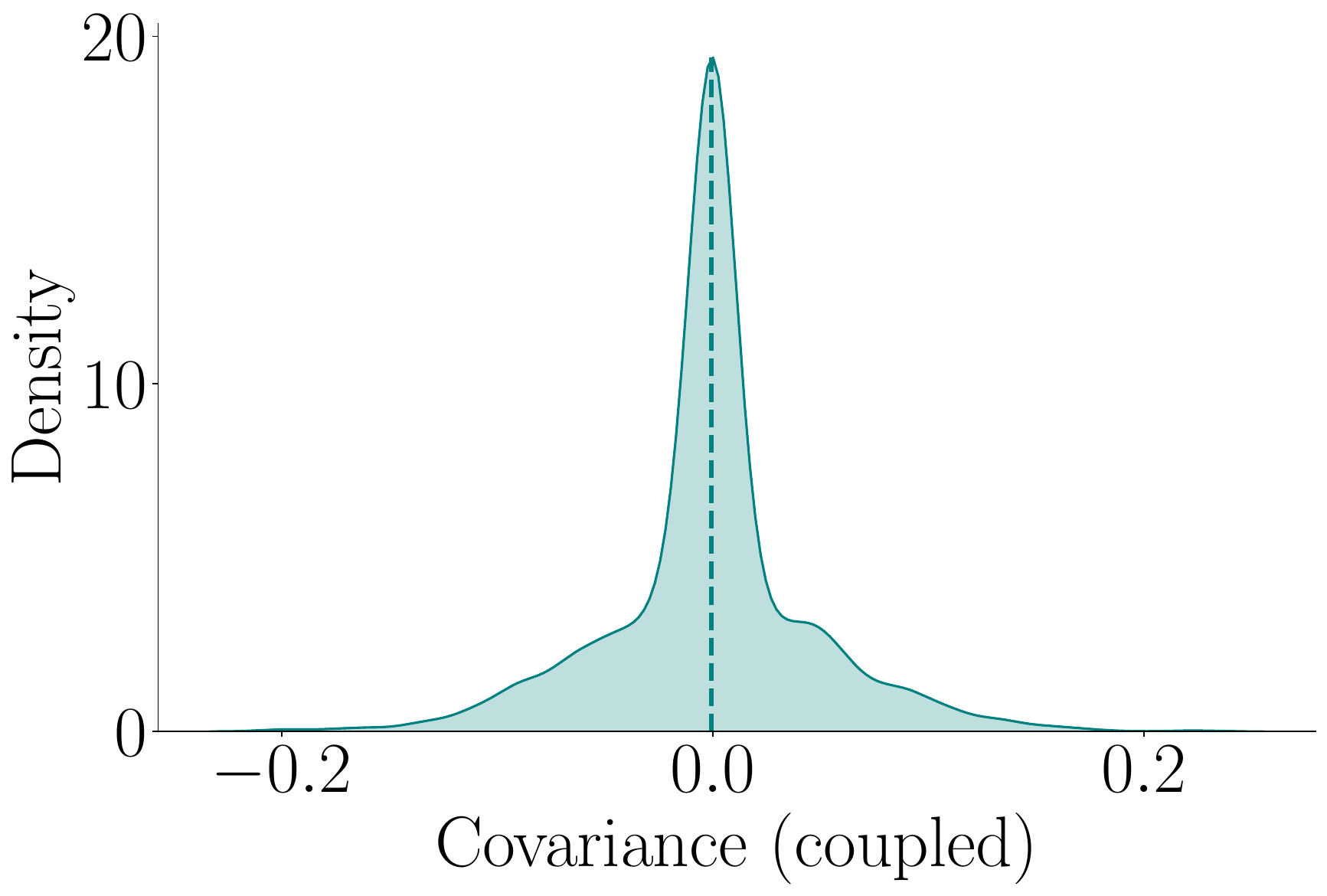} &
    \includegraphics[width=0.23\linewidth]{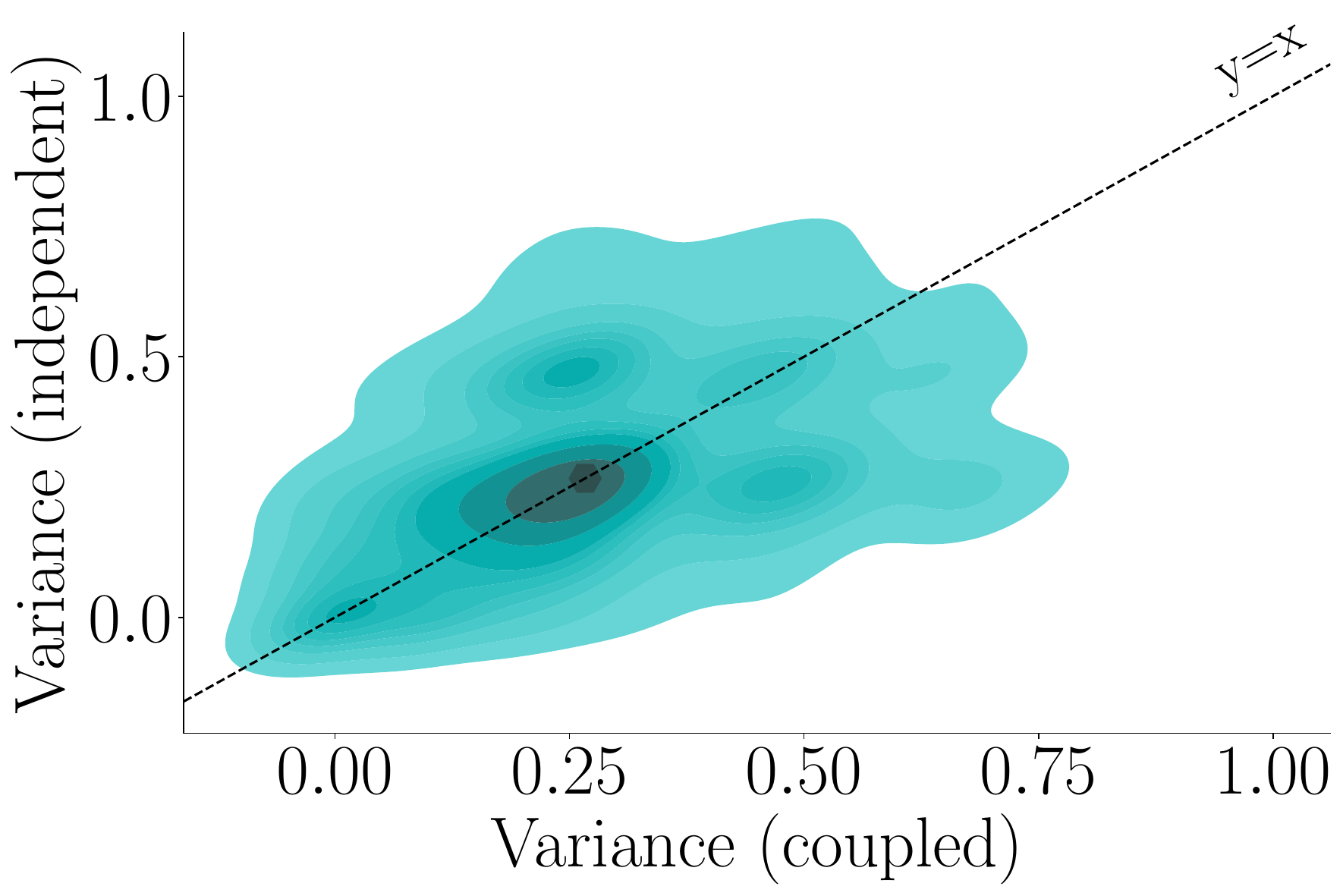} &
    \includegraphics[width=0.23\linewidth]{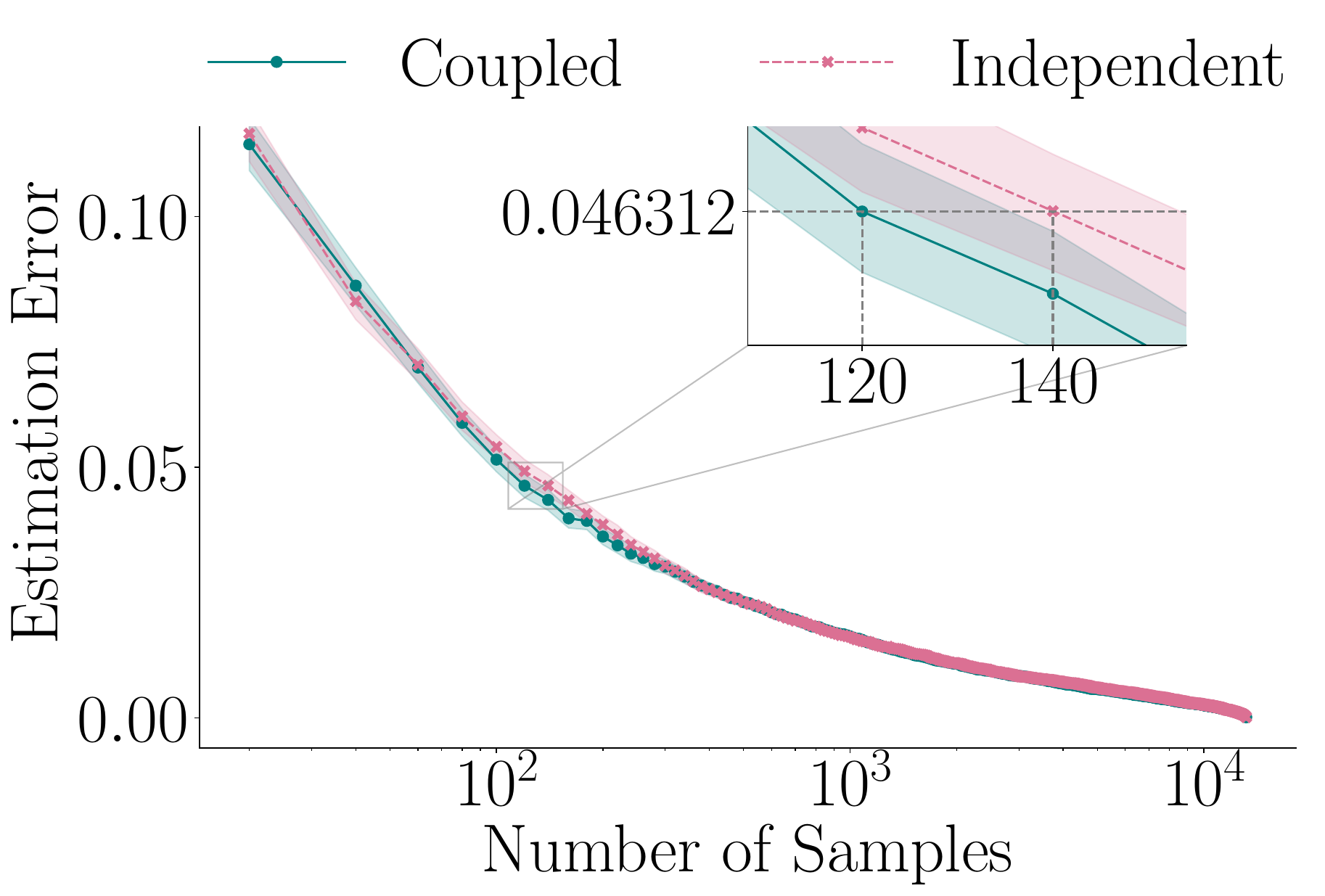} \\ \\
    \multicolumn{3}{c}{\texttt{1B} vs. \texttt{AWQ-INT4}}\\
    \includegraphics[width=0.23\linewidth]{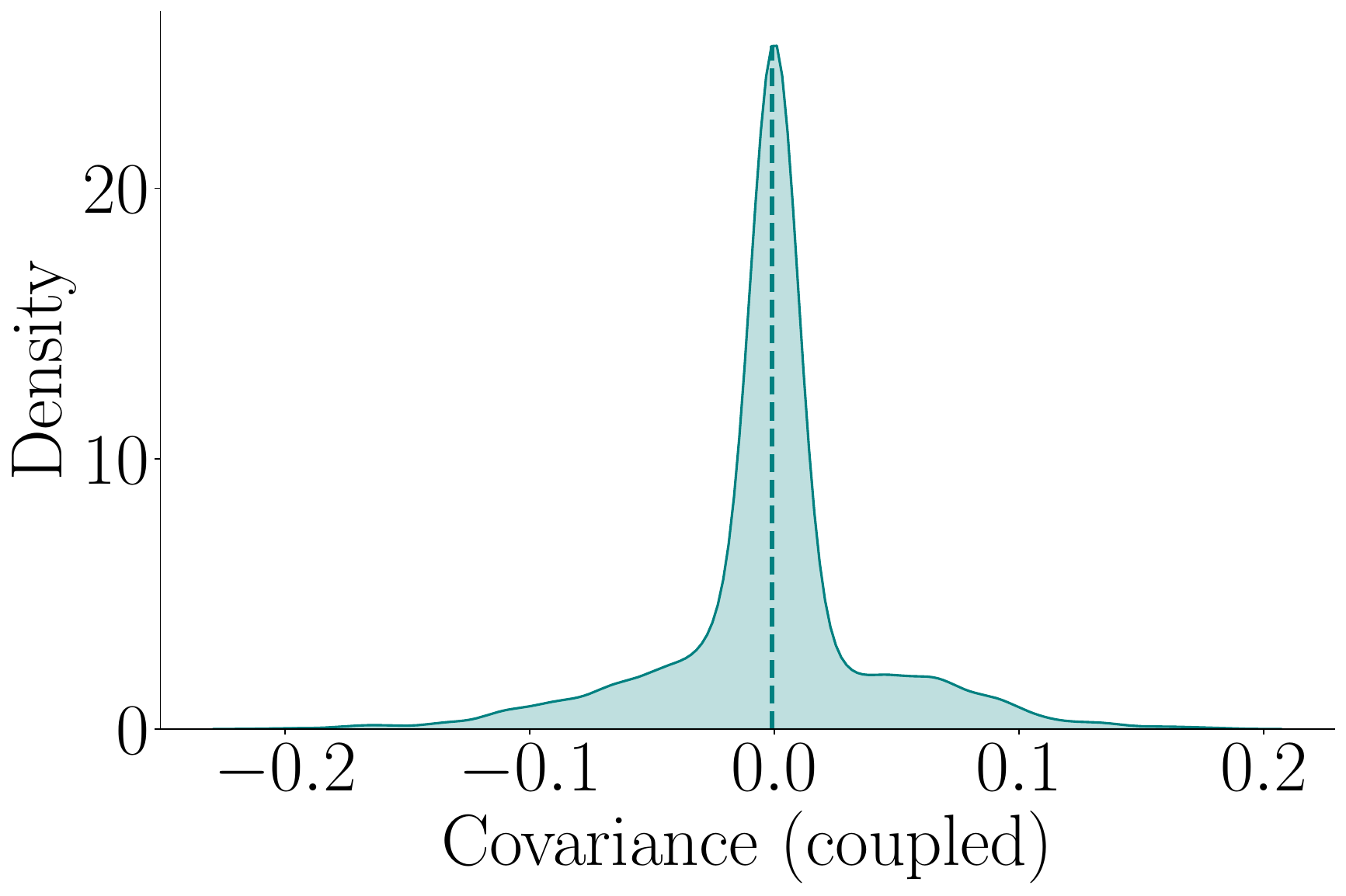} &
    \includegraphics[width=0.23\linewidth]{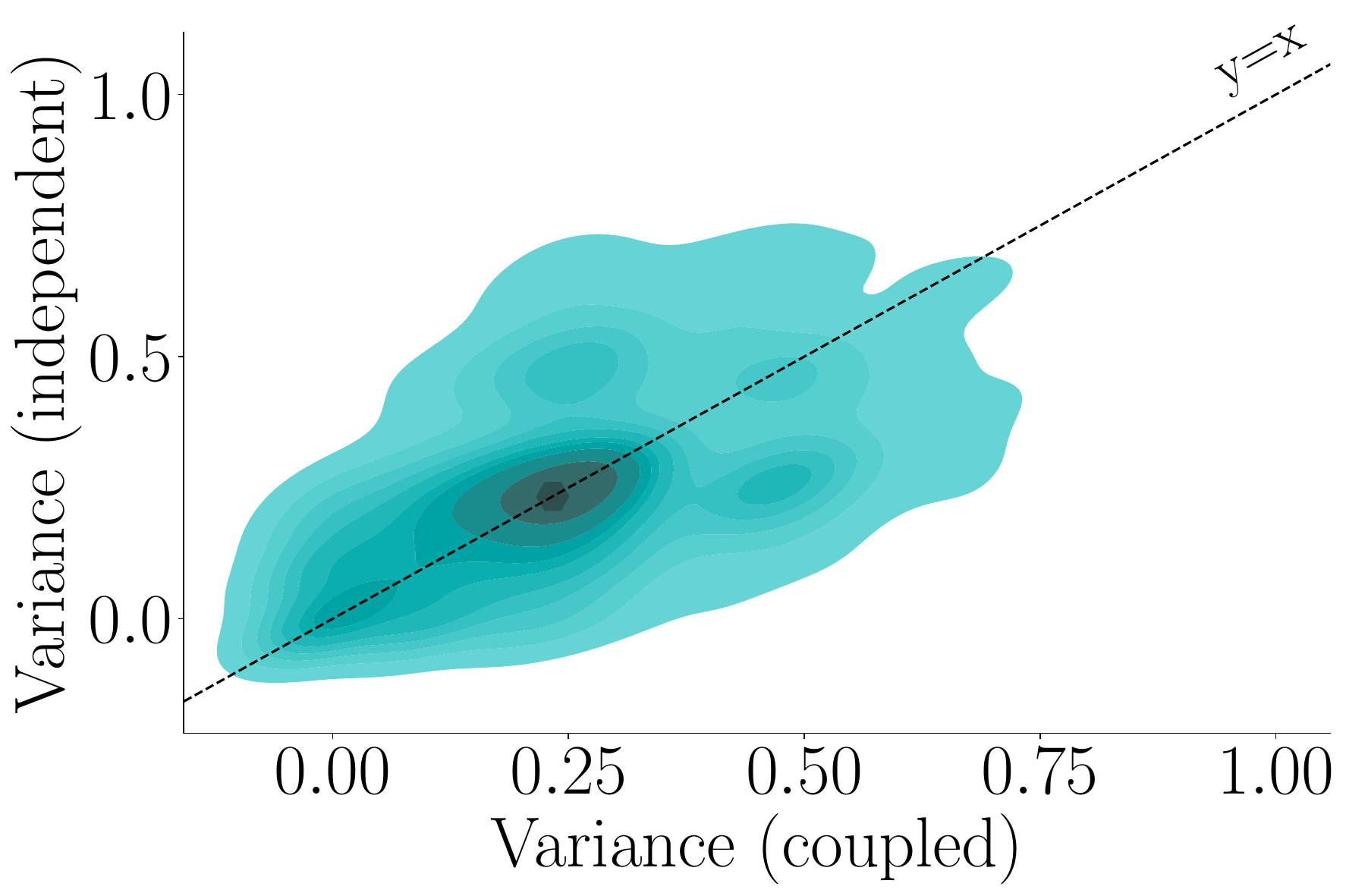} &
    \includegraphics[width=0.23\linewidth]{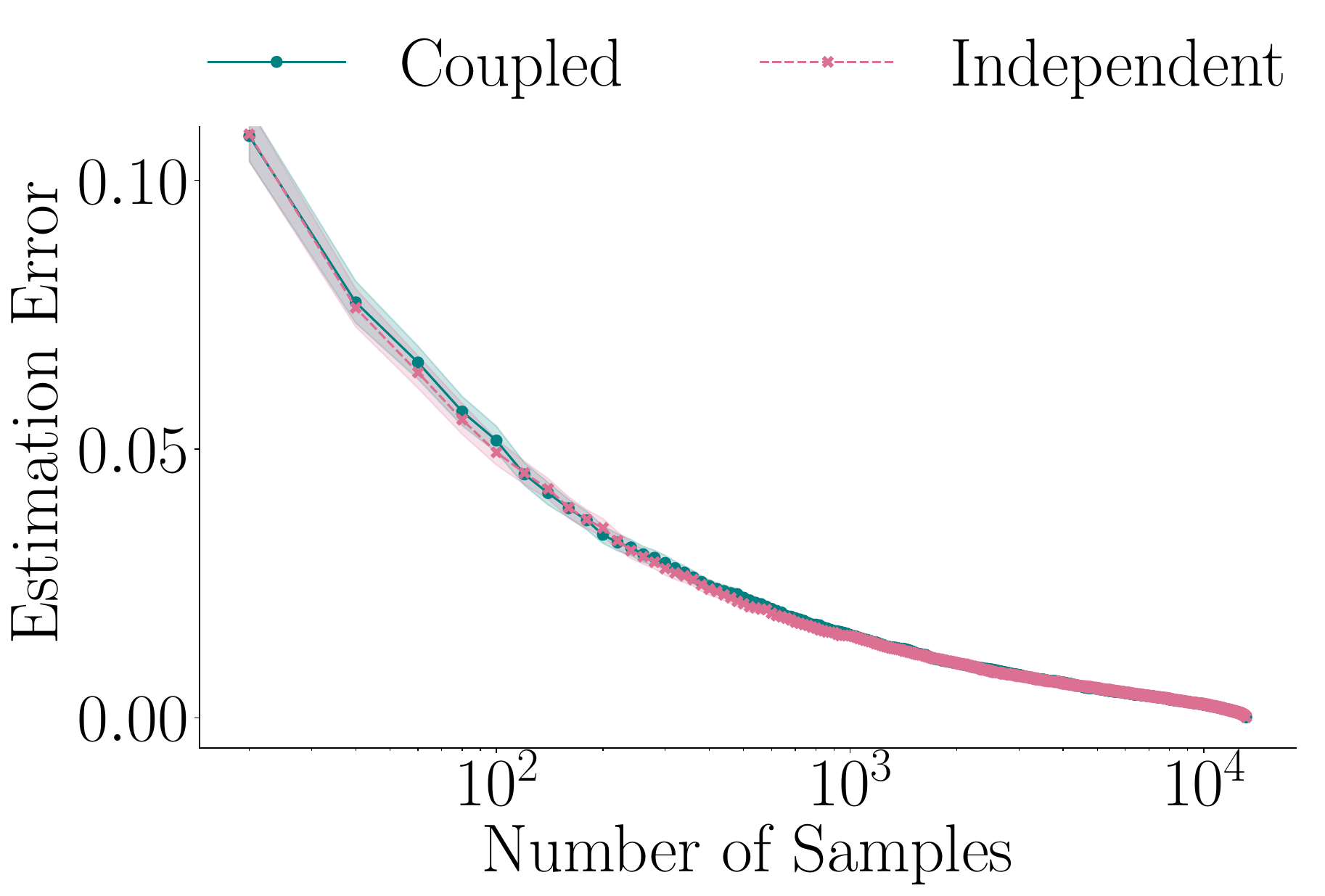} \\ \\
    \multicolumn{3}{c}{\texttt{bnb-8bit} vs. \texttt{bnb-4bit}}\\
    \includegraphics[width=0.23\linewidth]{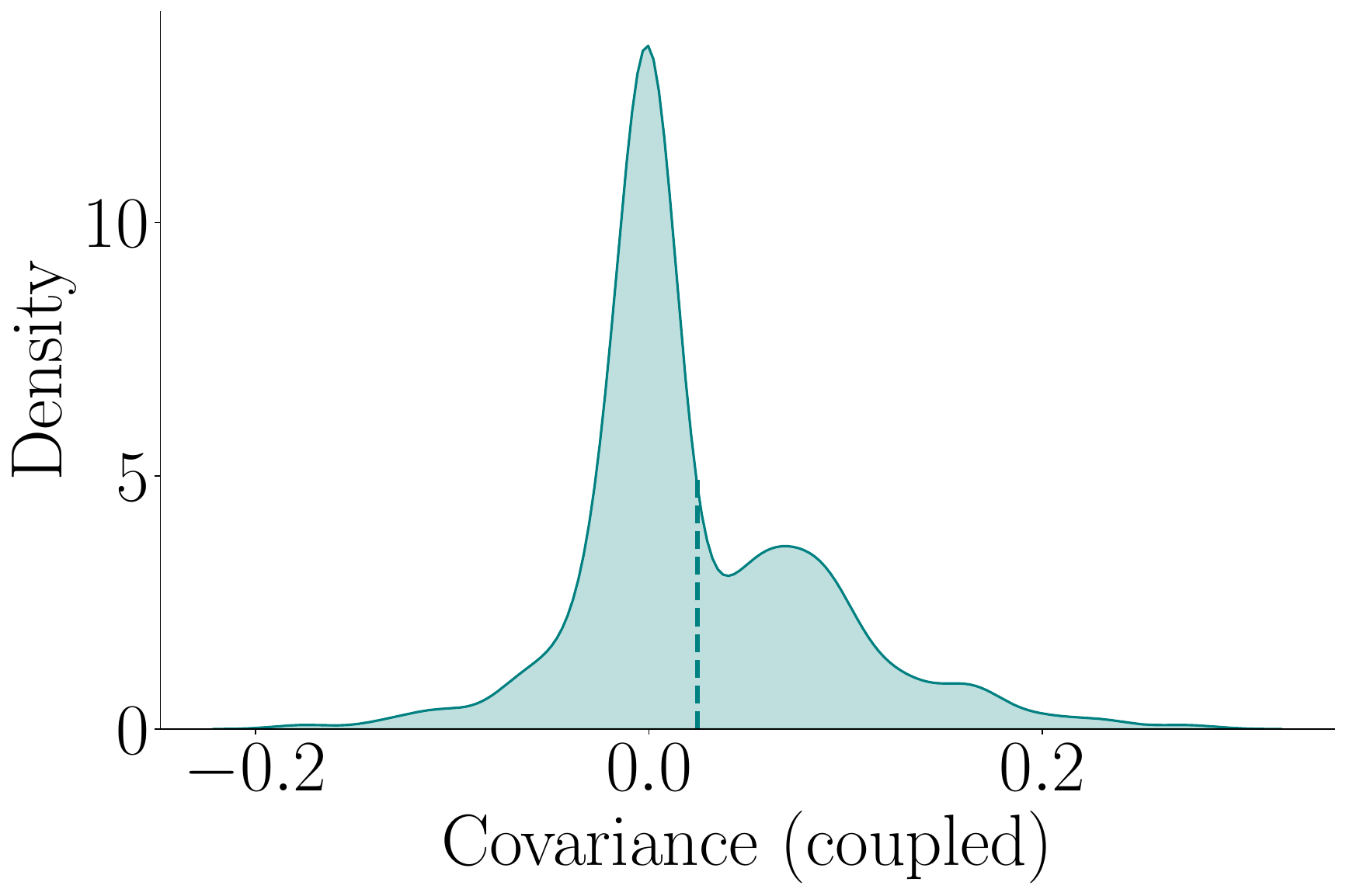} &
    \includegraphics[width=0.23\linewidth]{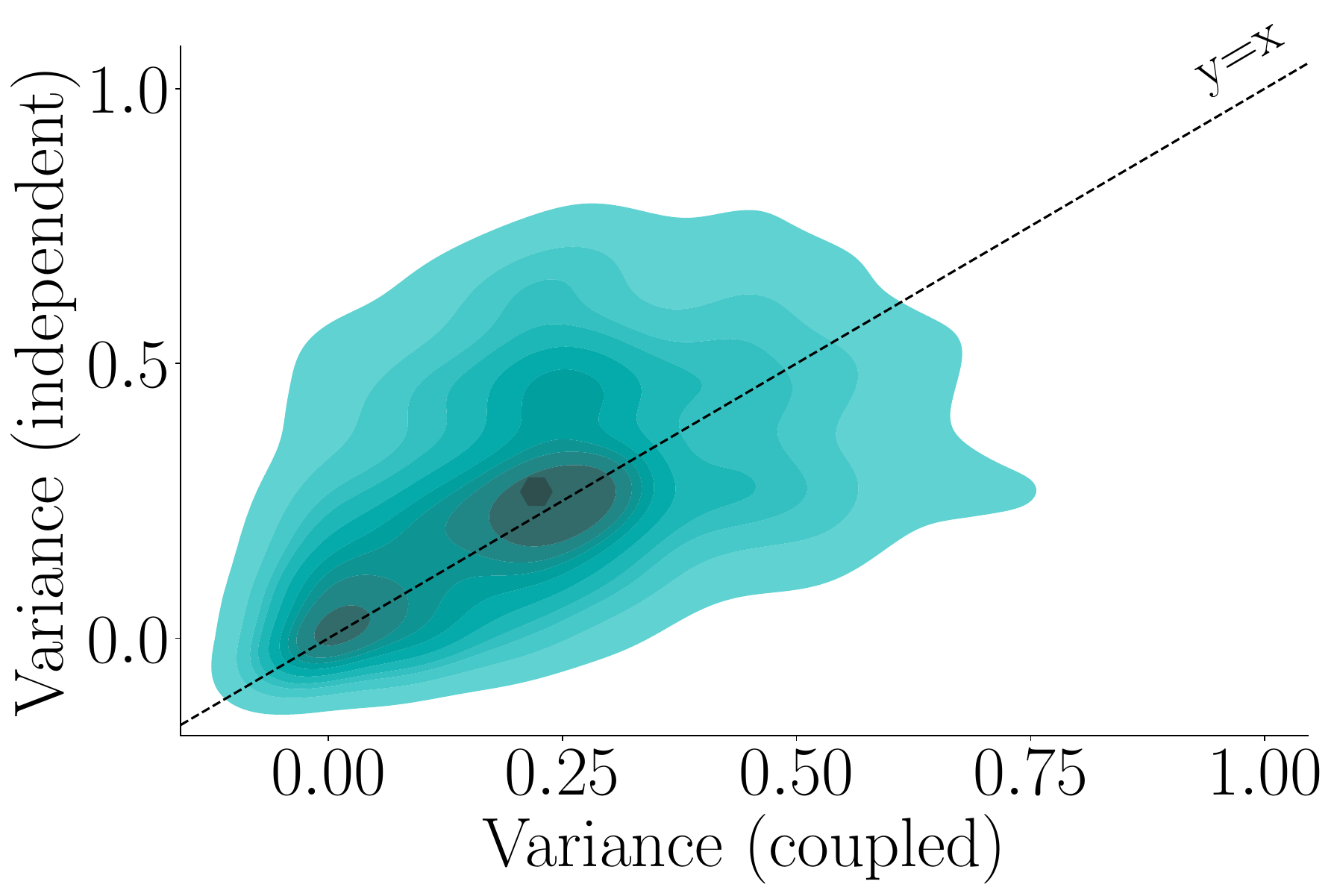} &
    \includegraphics[width=0.23\linewidth]{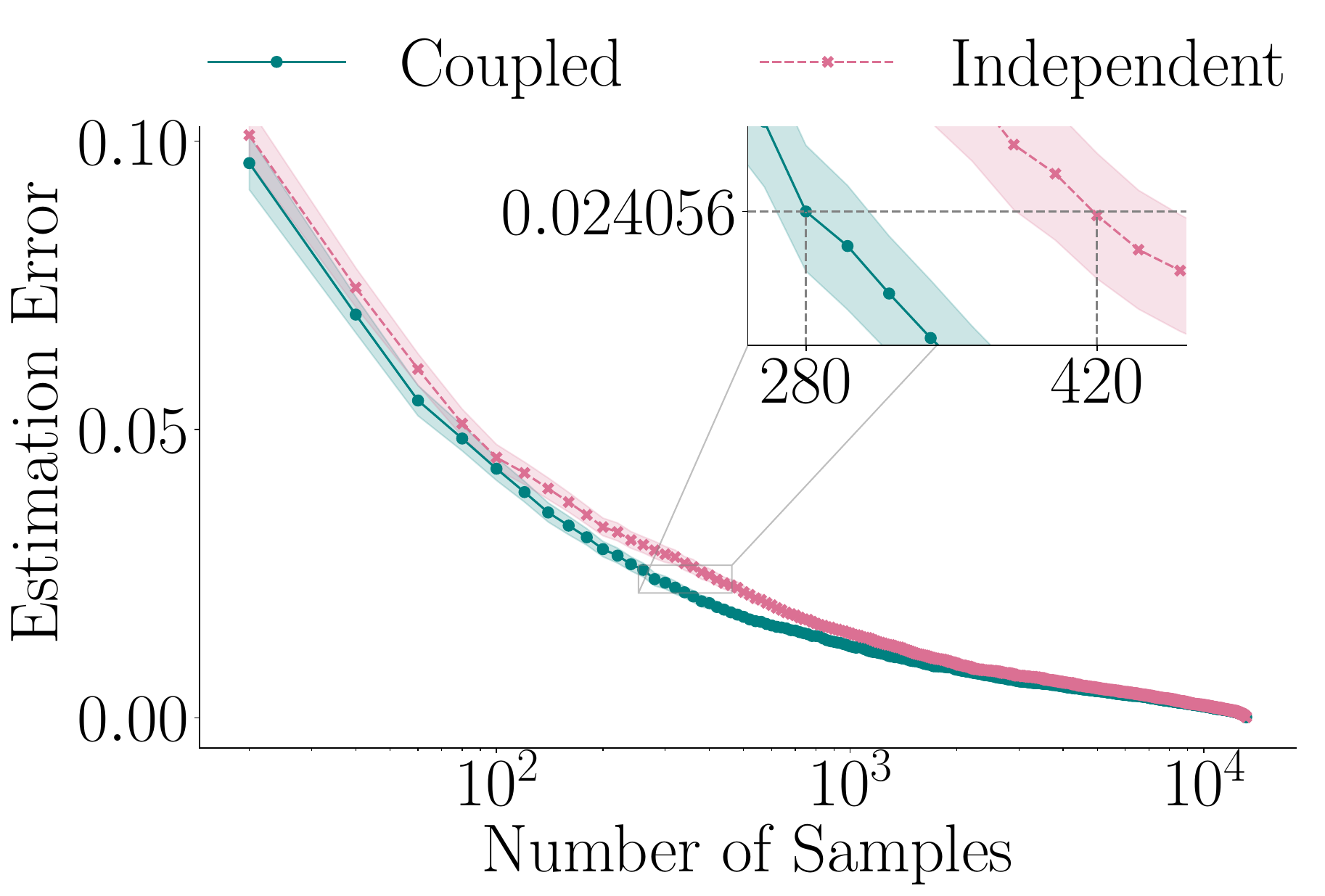} \\ \\
    \multicolumn{3}{c}{\texttt{bnb-8bit} vs. \texttt{AWQ-INT4}}\\
    \includegraphics[width=0.23\linewidth]{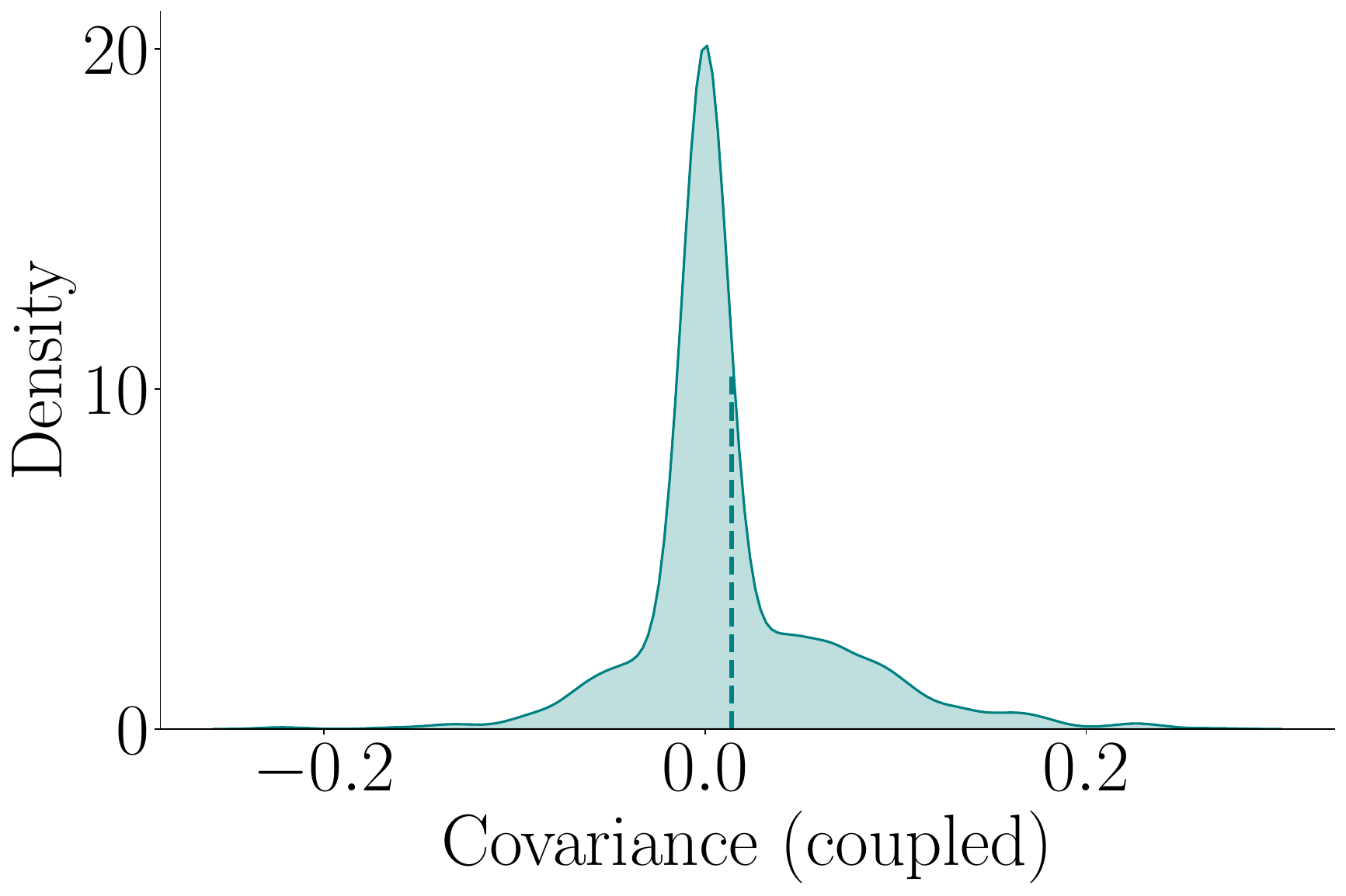} &
    \includegraphics[width=0.23\linewidth]{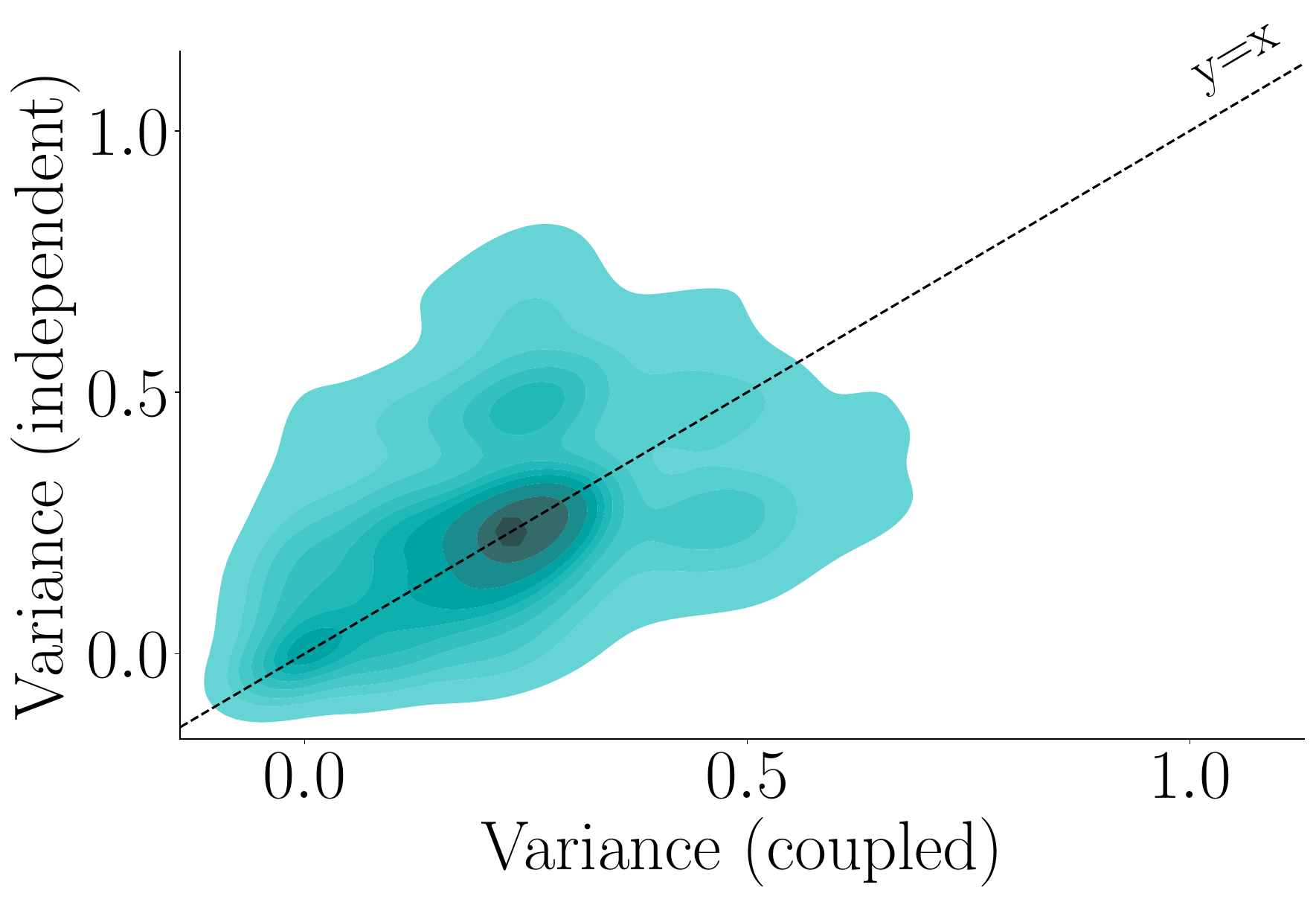} &
    \includegraphics[width=0.23\linewidth]{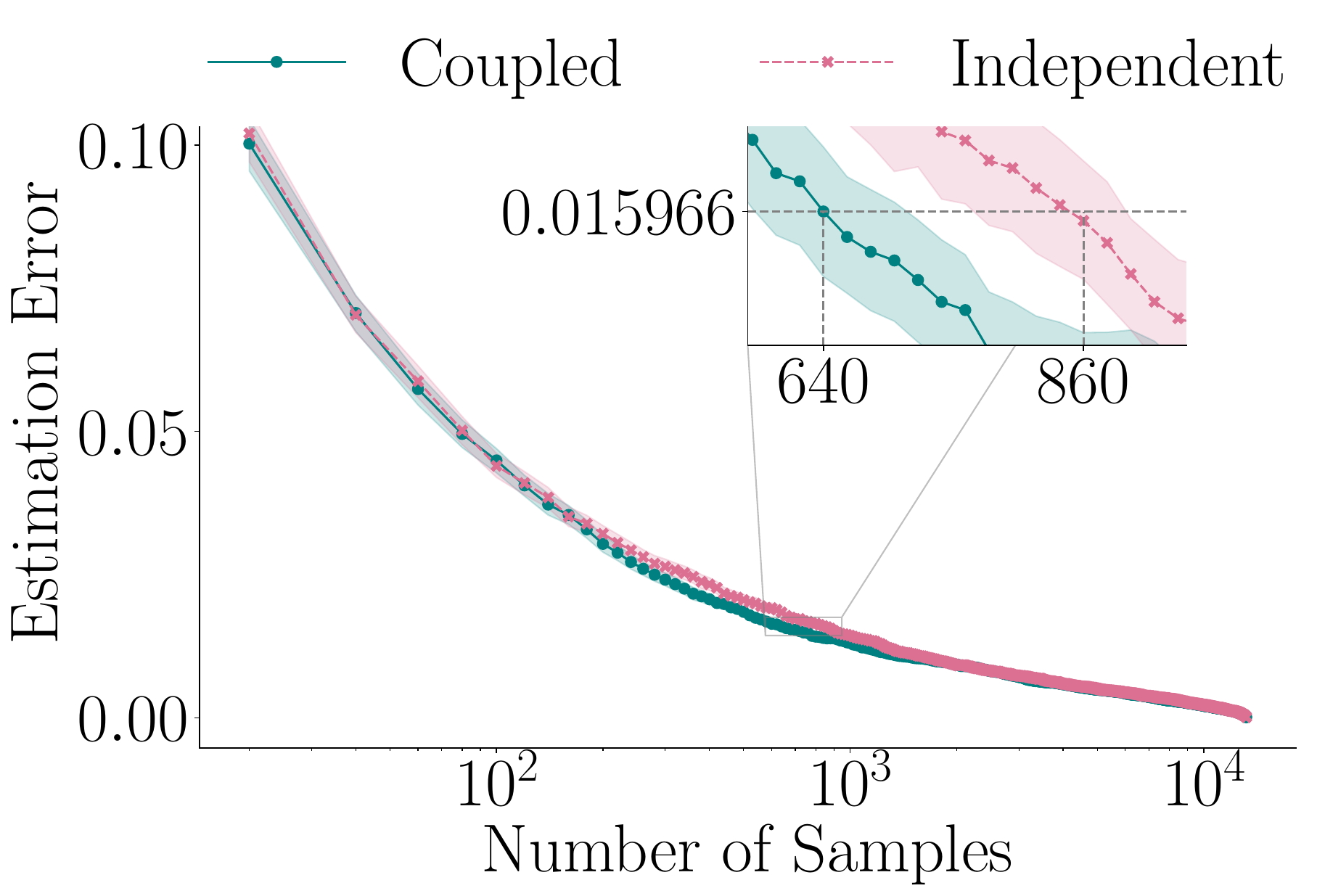} \\ \\

    \multicolumn{3}{c}{\texttt{bnb-4bit} vs. \texttt{AWQ-INT4}}\\
    \includegraphics[width=0.23\linewidth]{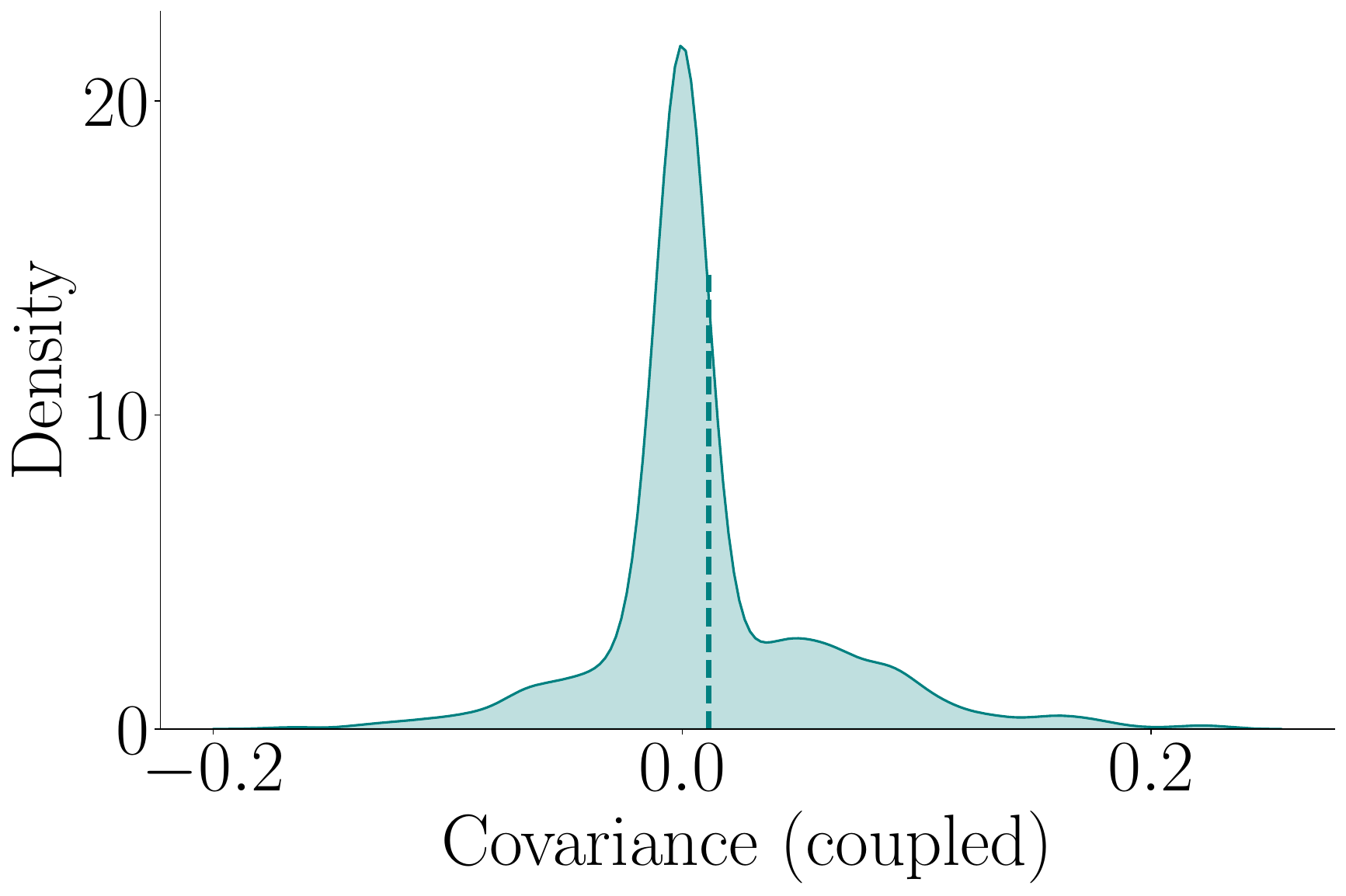} &
    \includegraphics[width=0.23\linewidth]{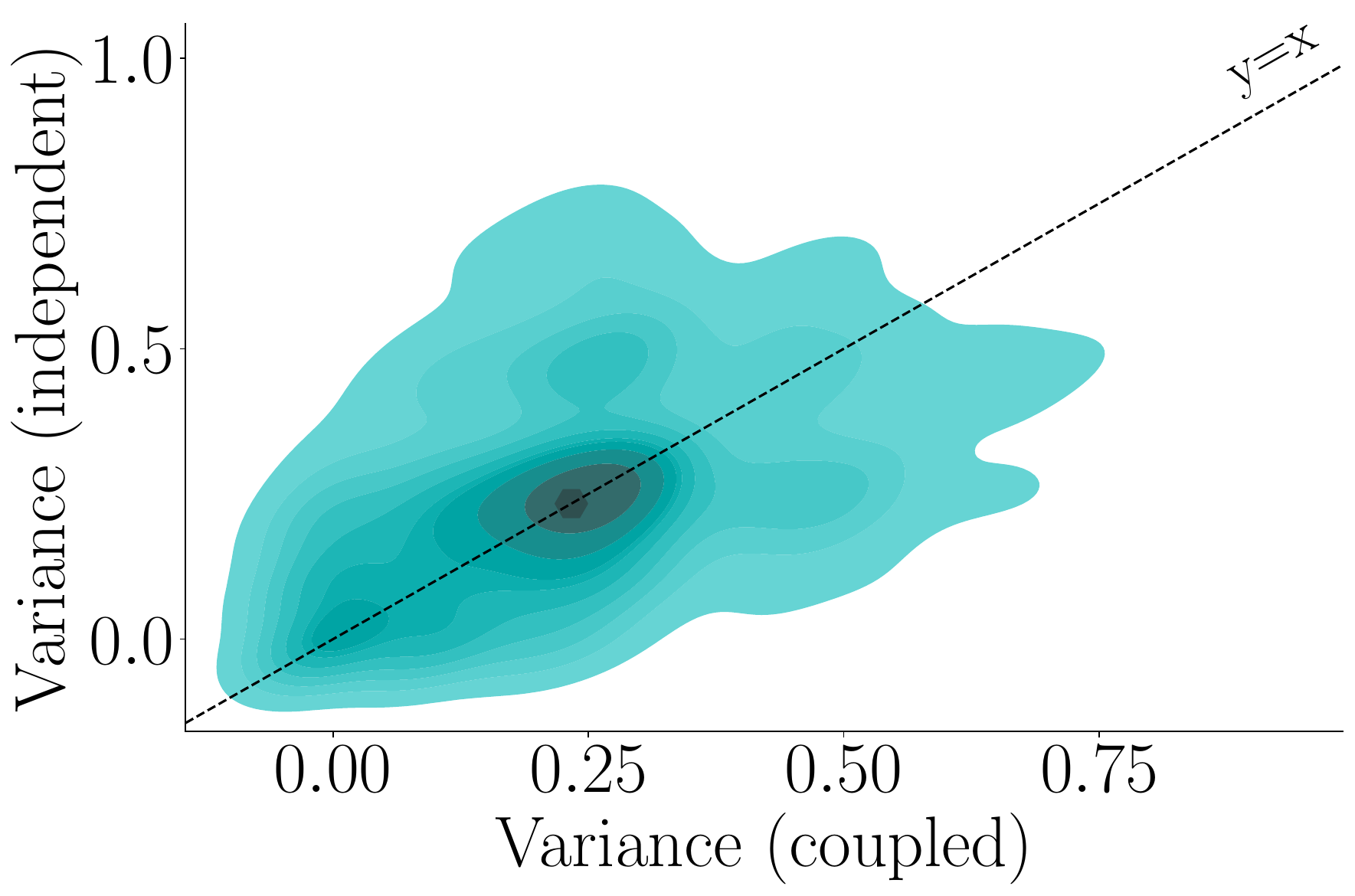}&
    \includegraphics[width=0.23\linewidth]{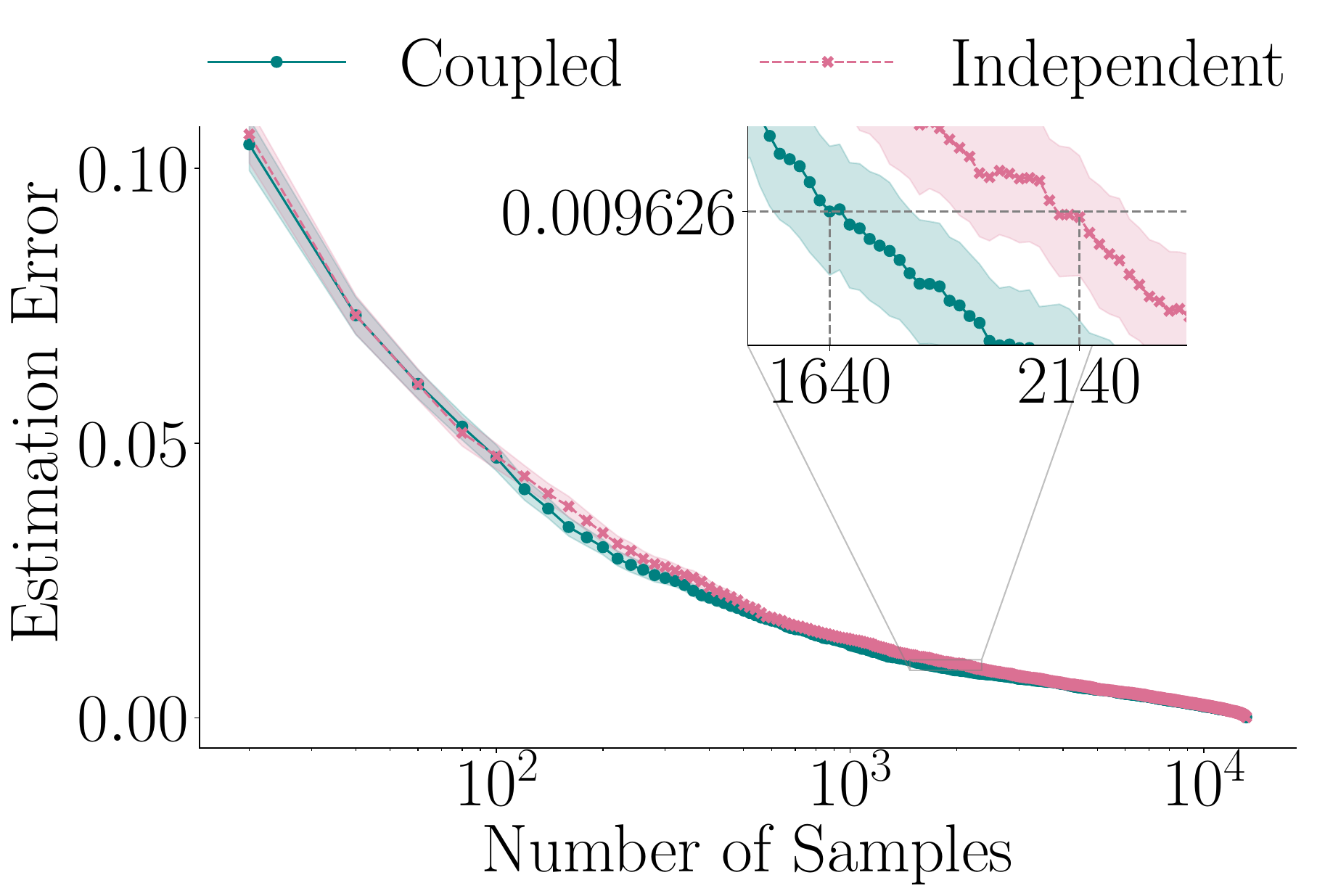} \\ \\
    (a) Score covariance & (b) Variance of the score difference & (c) Estimation error vs. \# samples \\

\end{tabular}
    \caption{\textbf{Comparison between several pairs of LLMs in the \texttt{Llama} family on math questions from the GSM8K dataset.}
    Panels in column (a) show the kernel density estimate (KDE) of the covariance between the scores of the two LLMs on each problem under coupled generation; the dashed lines correspond to average values. Panels in column (b) show the KDE of the variance of the difference between the scores of the LLMs on each question under coupled and independent generation; the highlighted points correspond to median values. Panels in column (c) show the absolute error in the estimation of the expected difference between the scores of the LLMs against the number of samples; for each point on the x-axis, we perform $1{,}000$ sub-samplings and shaded areas correspond to $95\%$ confidence intervals.}
    \label{fig:gsm8k-third-5}
\end{figure}

\begin{figure}[ht]
\centering
\begin{tabular}{c c c}
    \multicolumn{3}{c}{\texttt{1B} vs. \texttt{3B}}\\
    \includegraphics[width=0.23\linewidth]{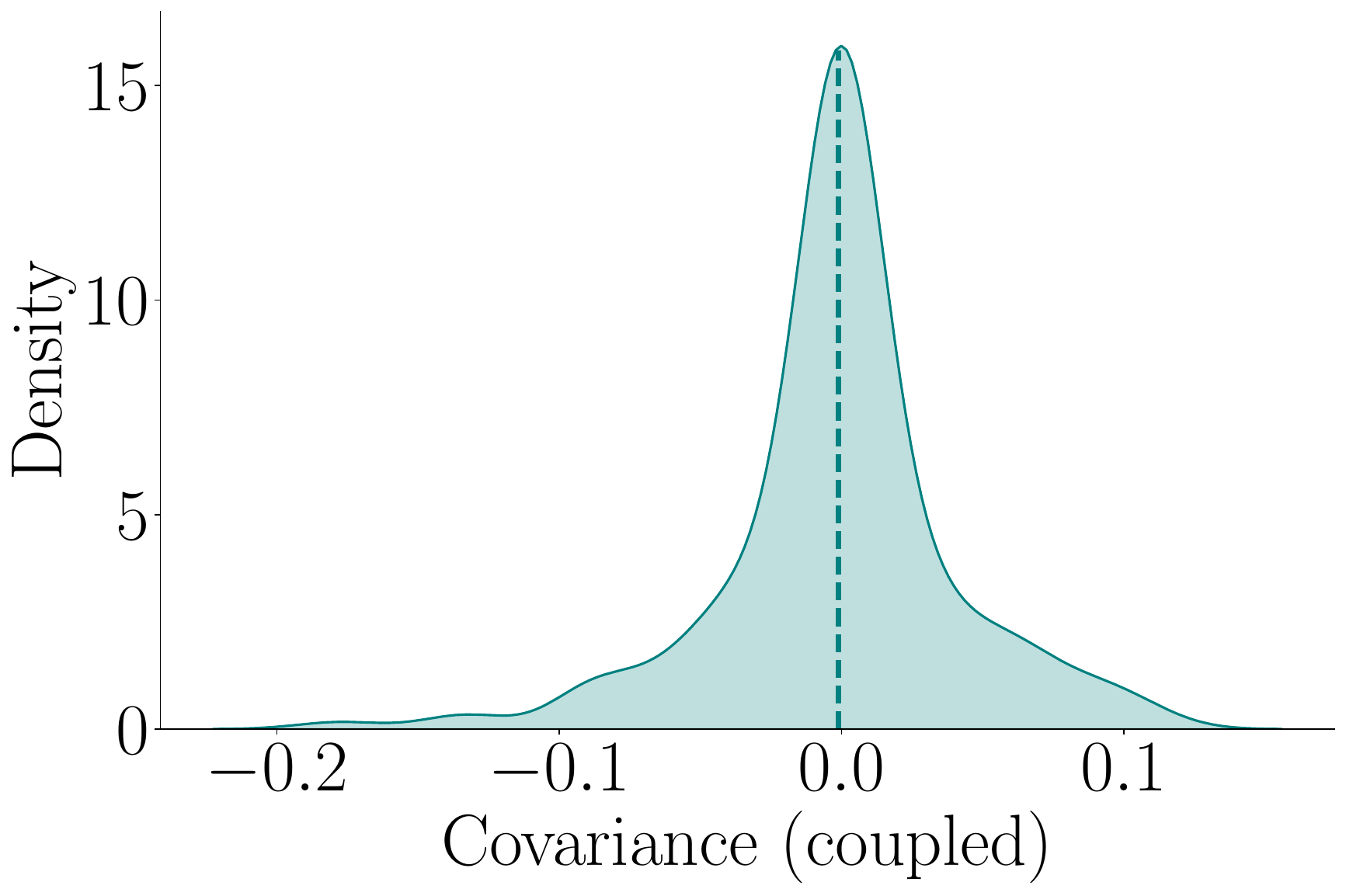} &
    \includegraphics[width=0.23\linewidth]{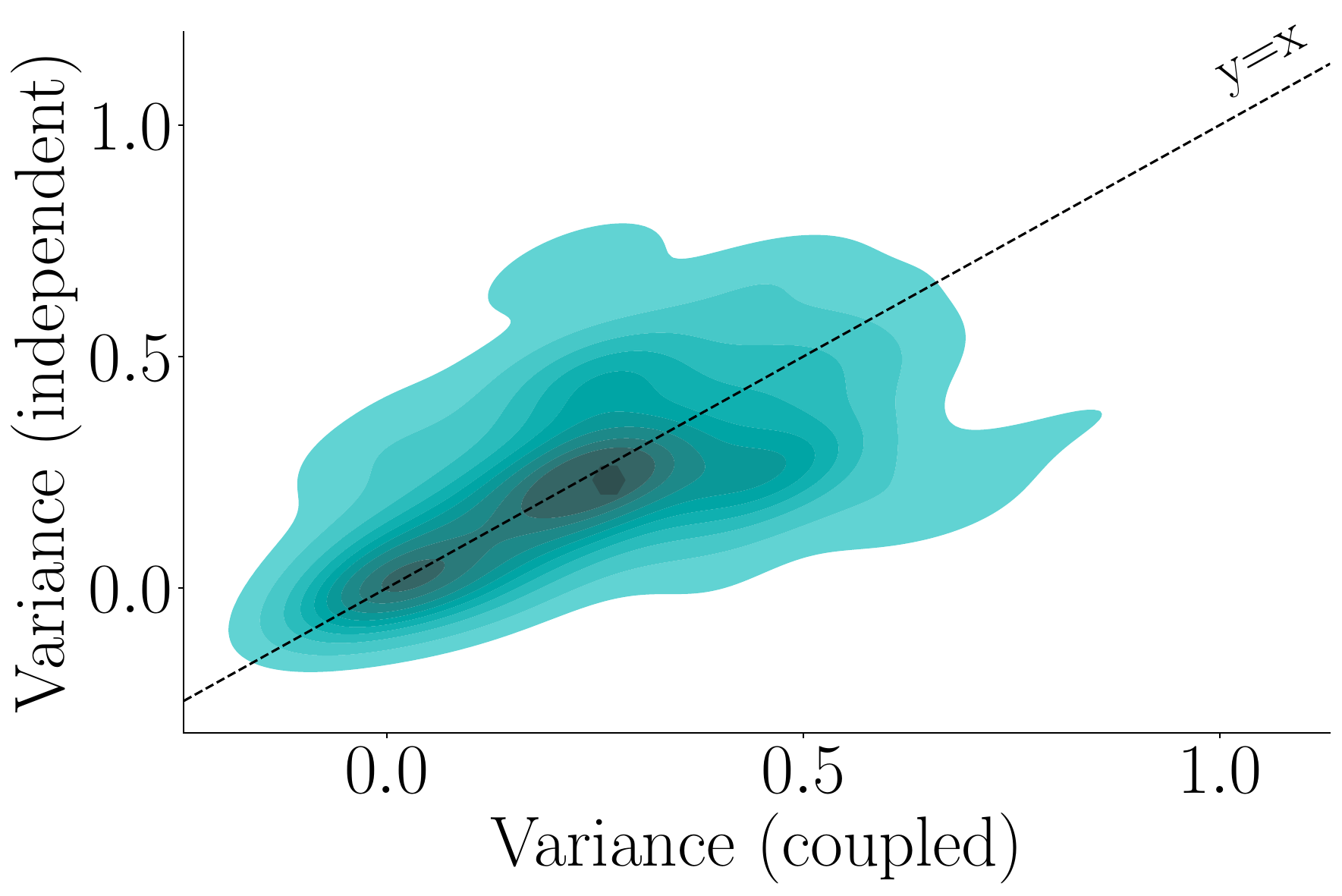} &
    \includegraphics[width=0.23\linewidth]{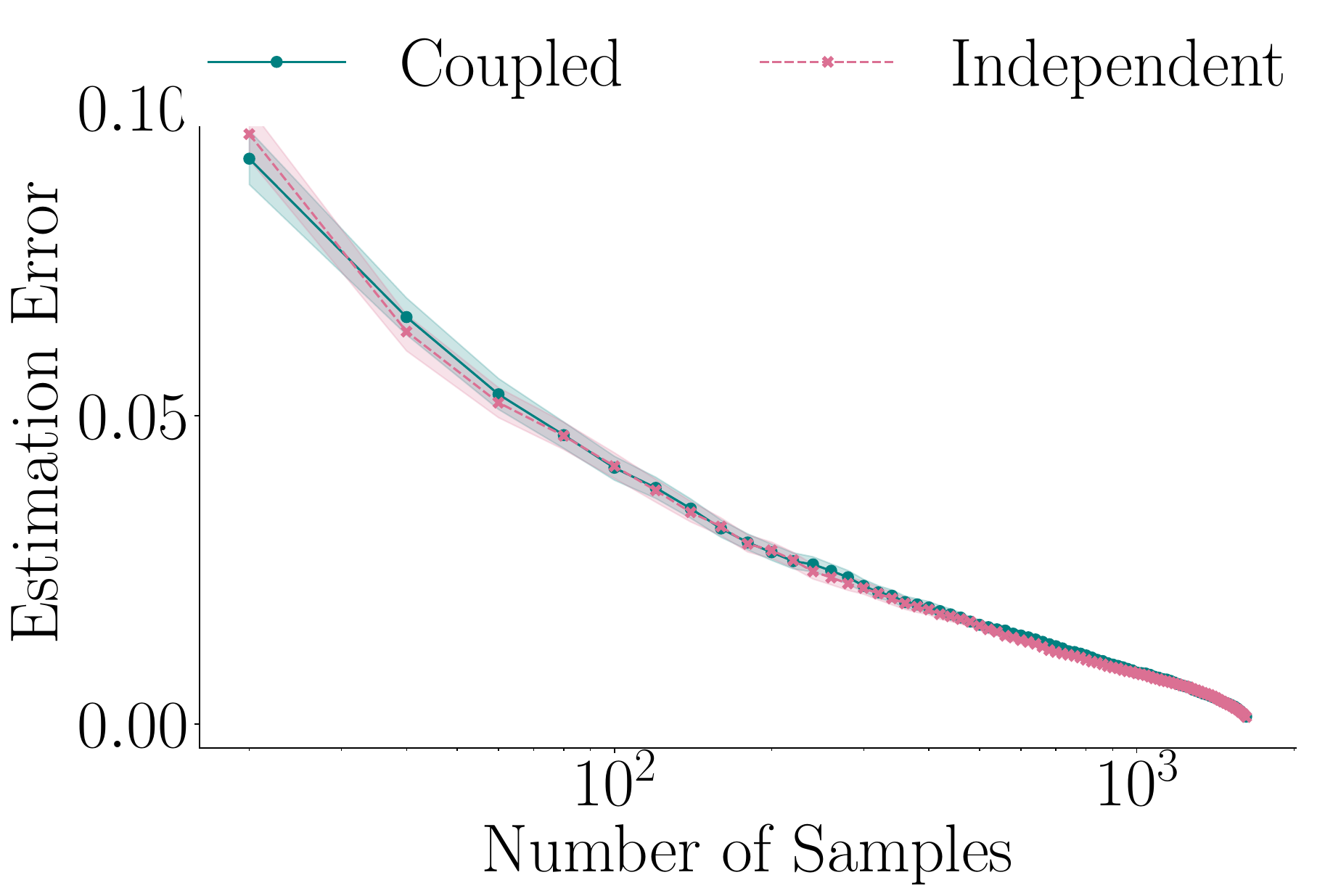} \\ \\
    \multicolumn{3}{c}{\texttt{1B} vs. \texttt{8B}}\\
    \includegraphics[width=0.23\linewidth]{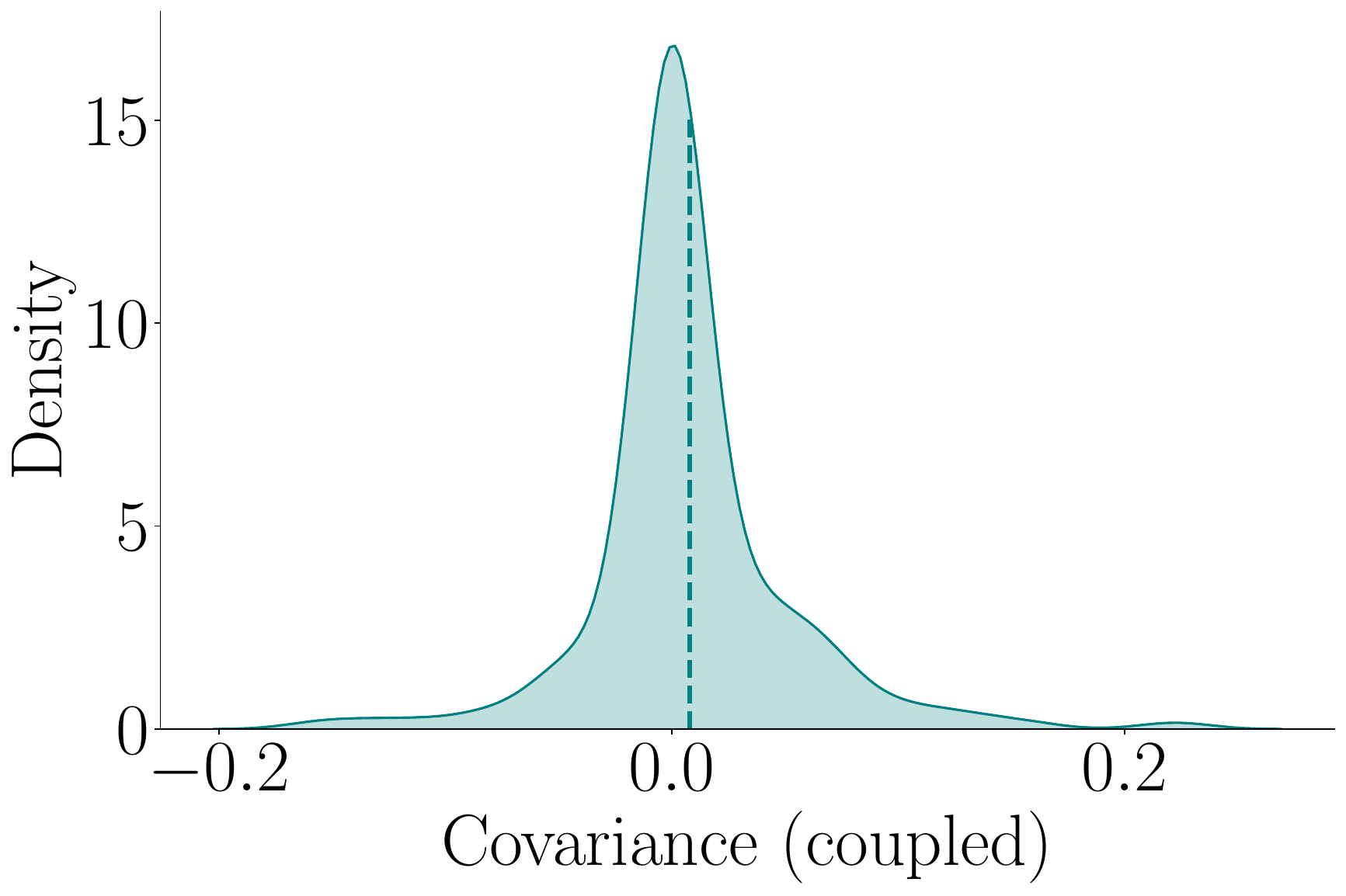} &
    \includegraphics[width=0.23\linewidth]{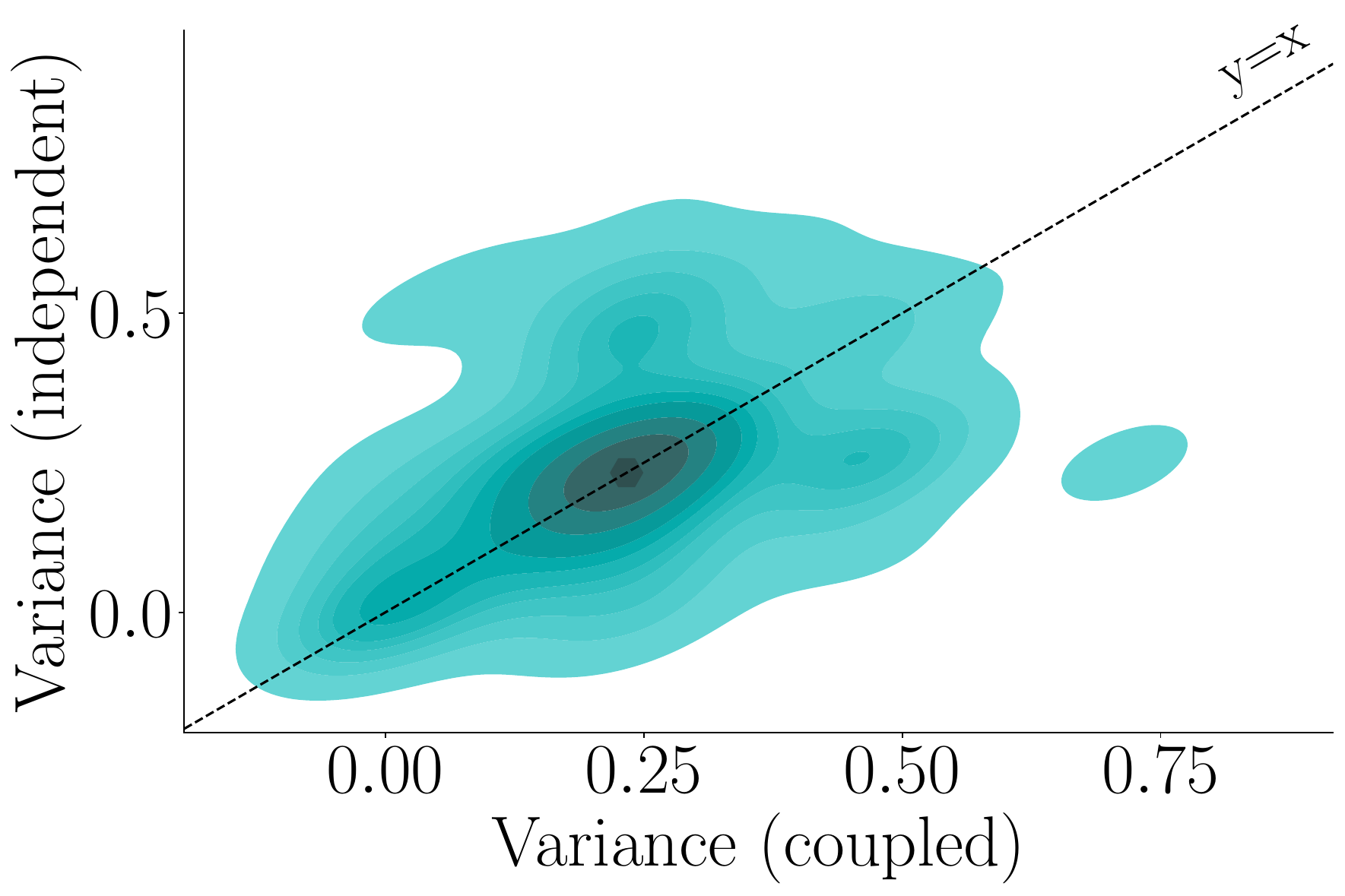} &
    \includegraphics[width=0.23\linewidth]{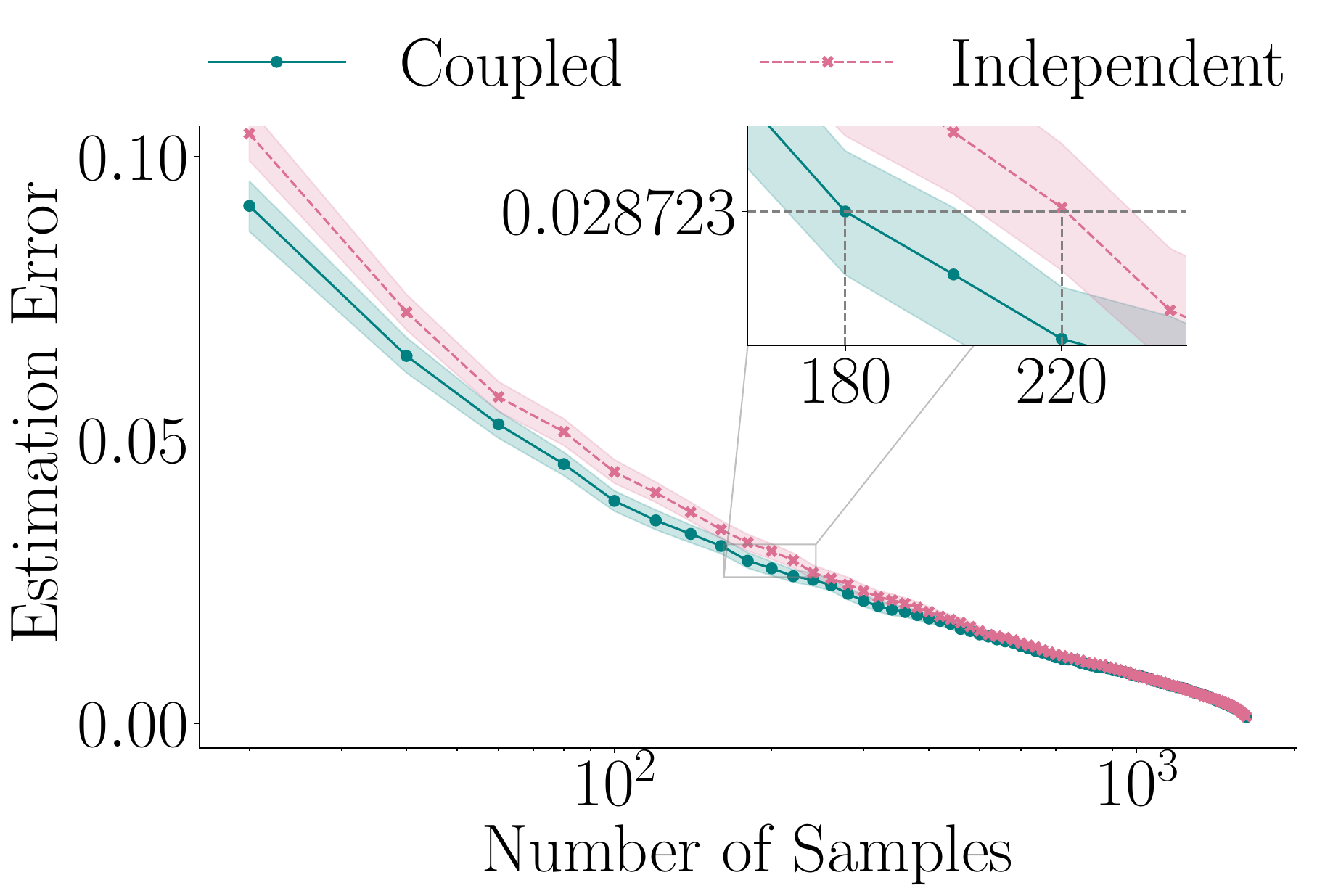} \\ \\
    \multicolumn{3}{c}{\texttt{3B} vs. \texttt{8B}}\\
    \includegraphics[width=0.23\linewidth]{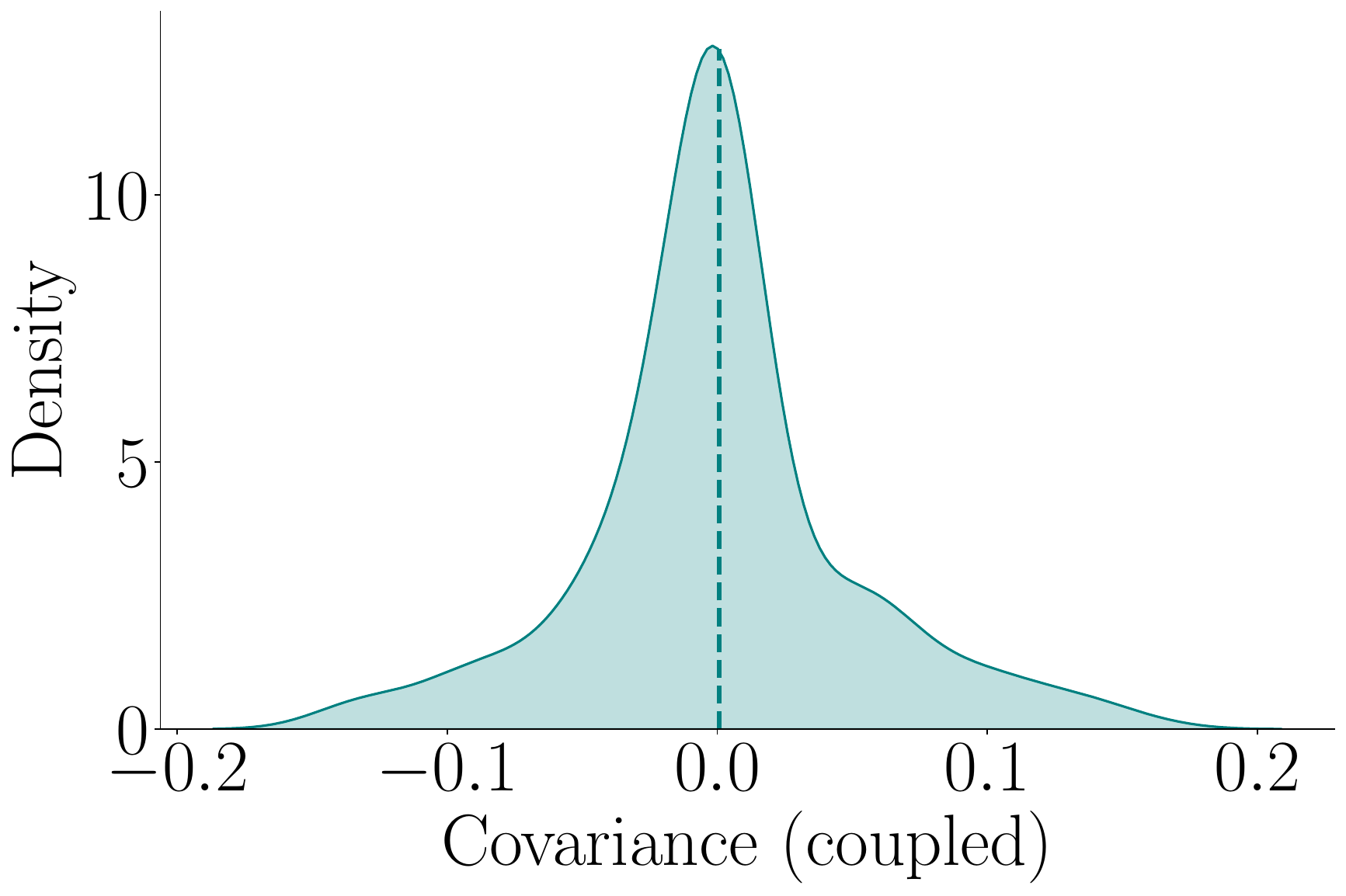} &
    \includegraphics[width=0.23\linewidth]{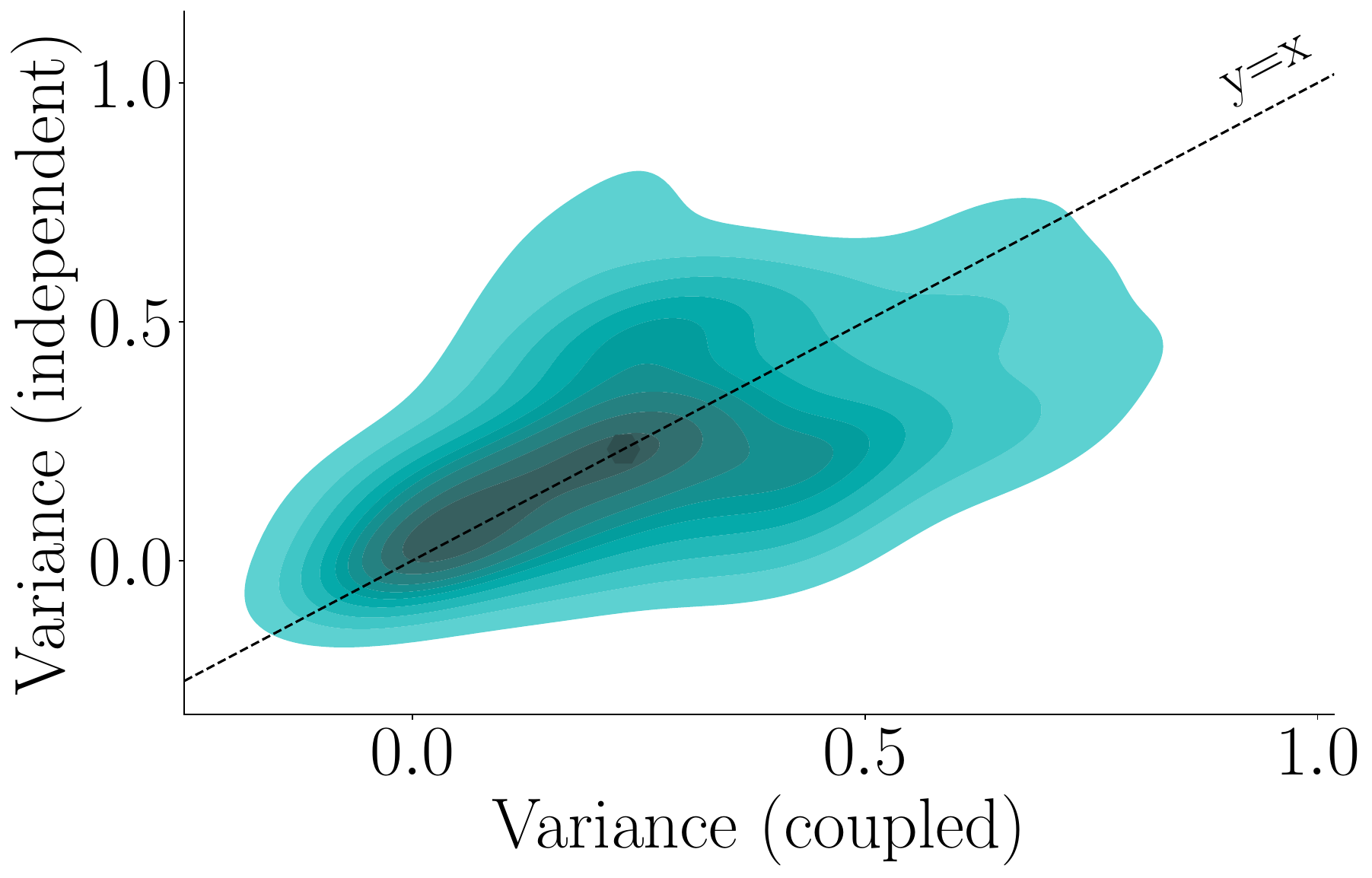} &
    \includegraphics[width=0.23\linewidth]{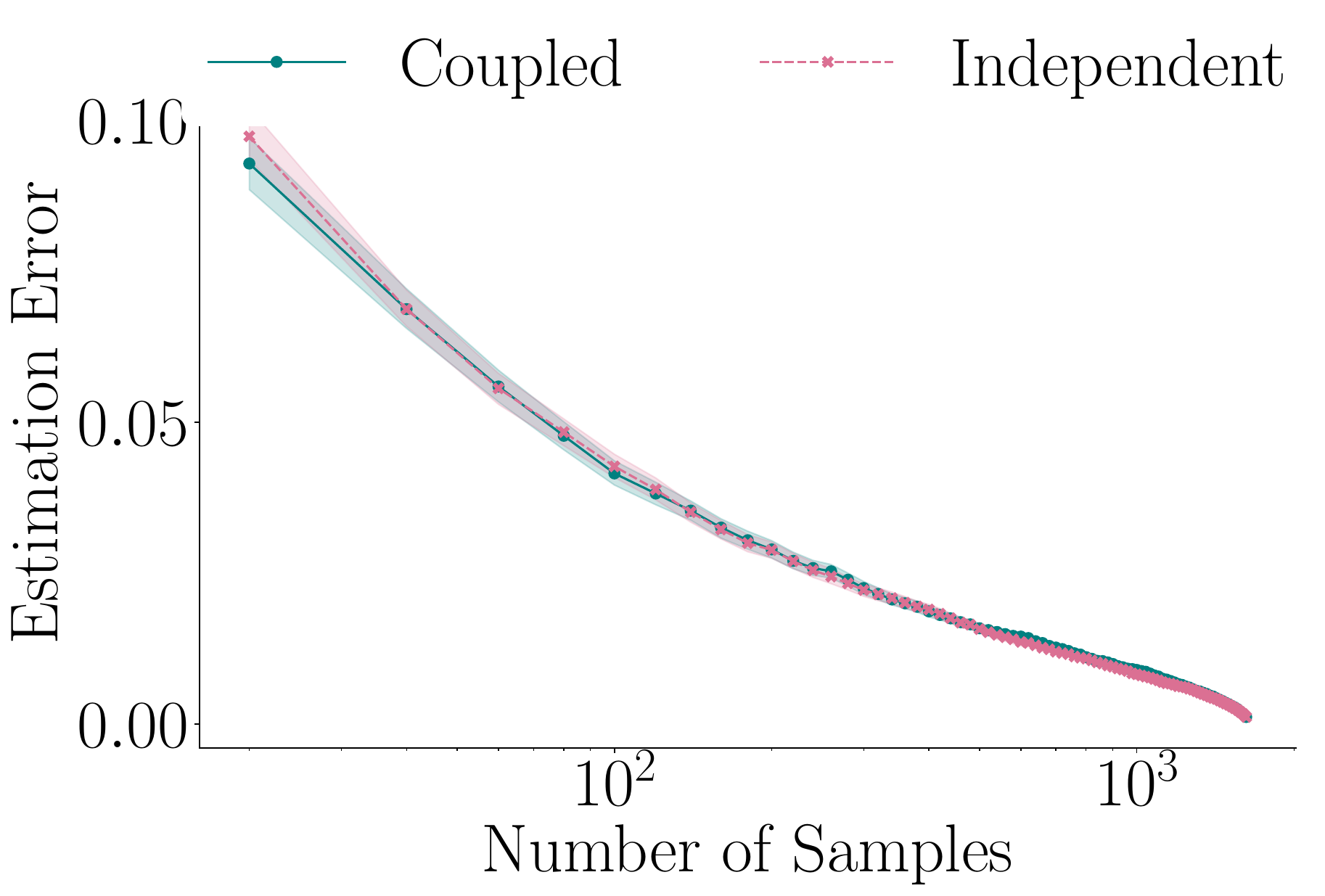} \\ \\
    \multicolumn{3}{c}{\texttt{8B} vs. \texttt{bnb-8bit}}\\
    \includegraphics[width=0.23\linewidth]{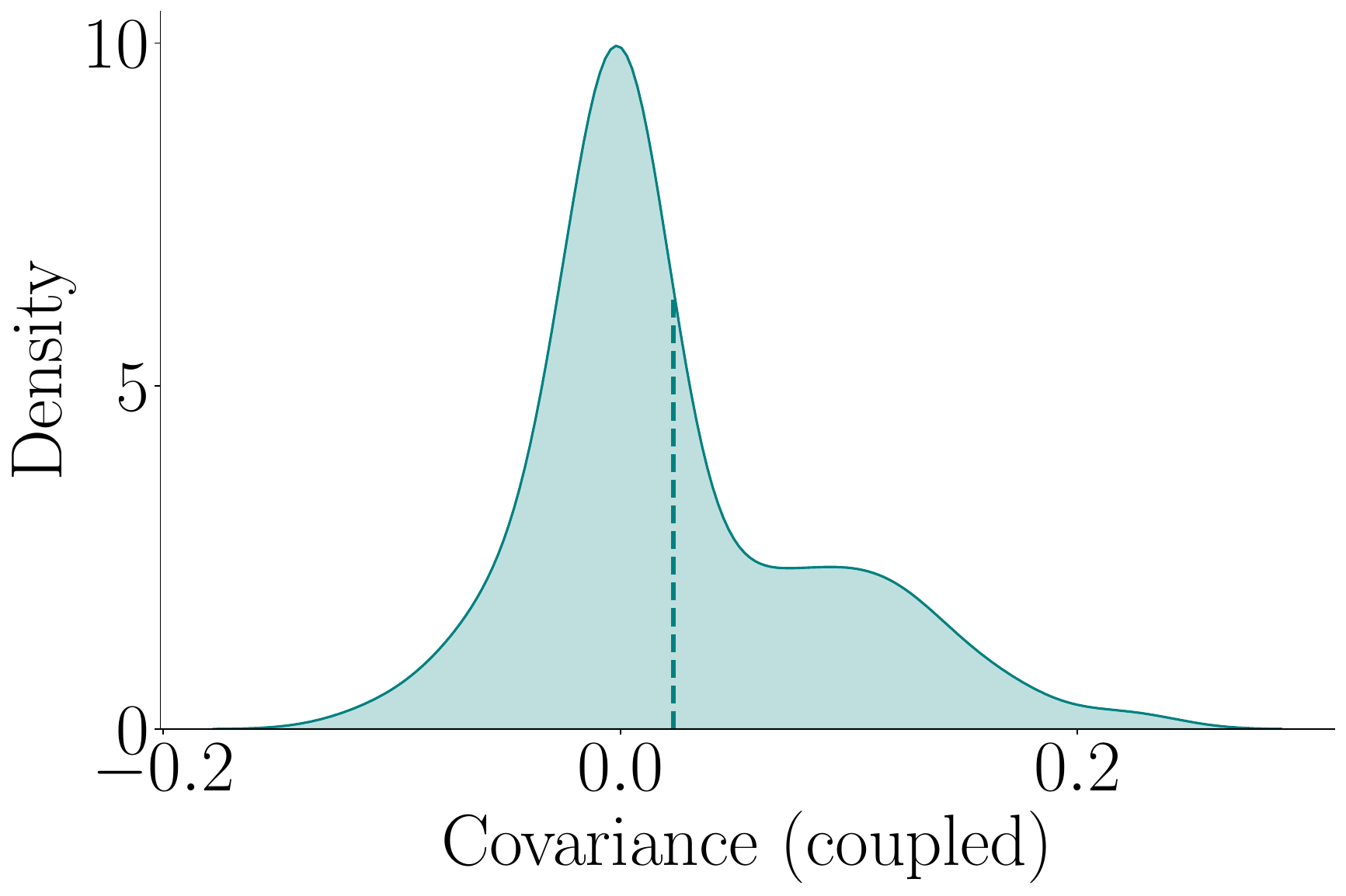} &
    \includegraphics[width=0.23\linewidth]{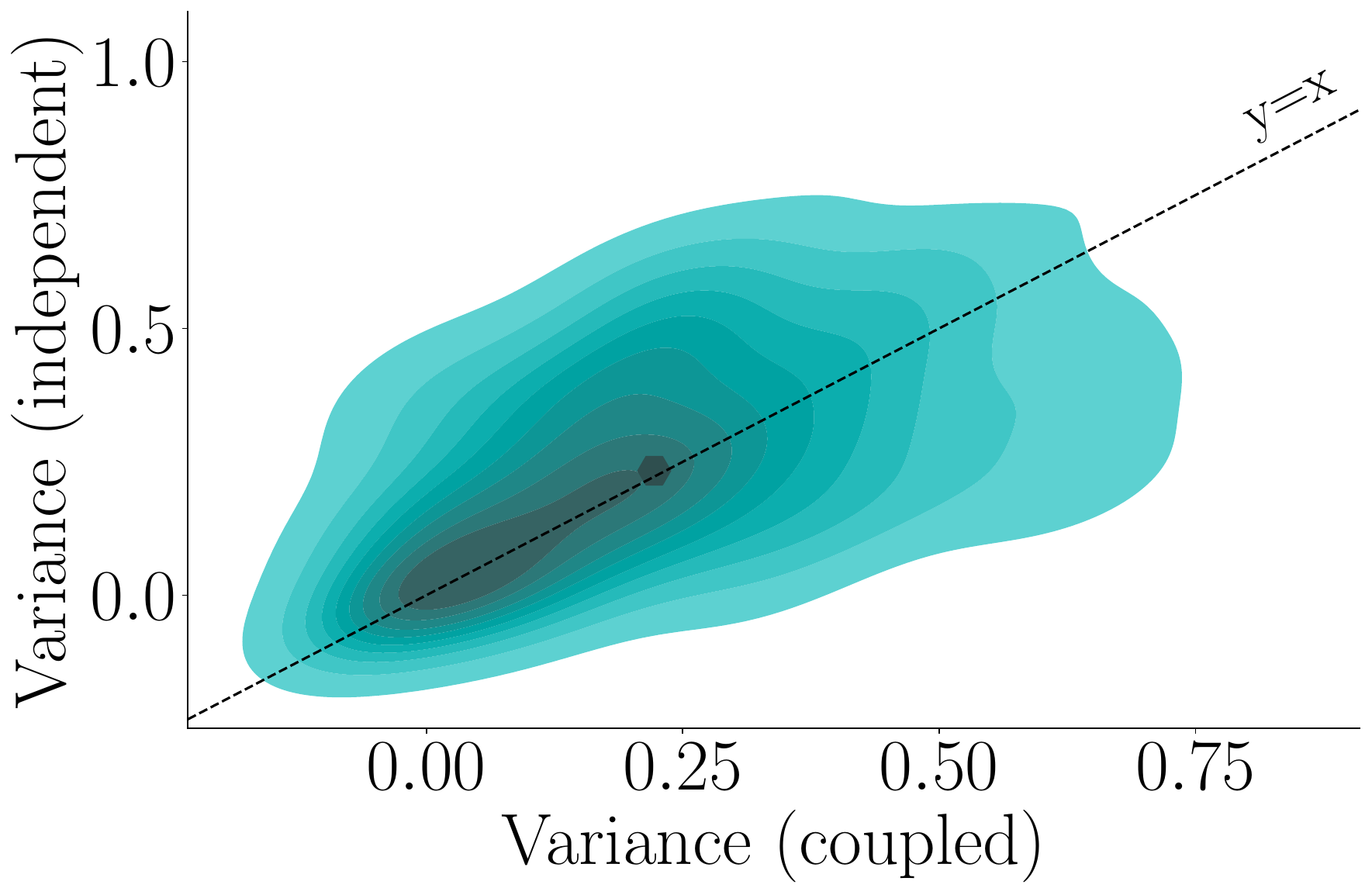} &
    \includegraphics[width=0.23\linewidth]{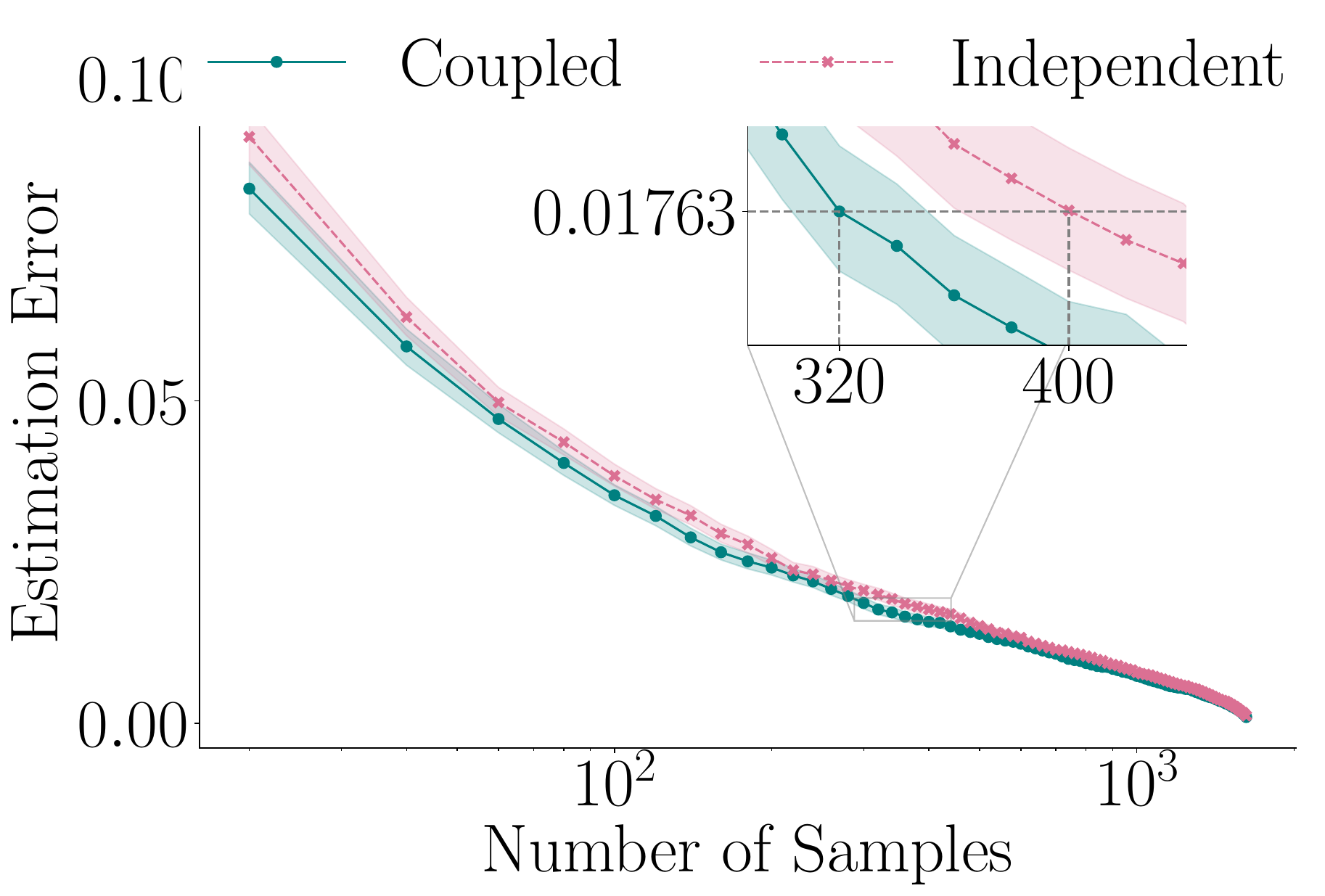} \\ \\

    \multicolumn{3}{c}{\texttt{8B} vs. \texttt{bnb-4bit}}\\
    \includegraphics[width=0.23\linewidth]{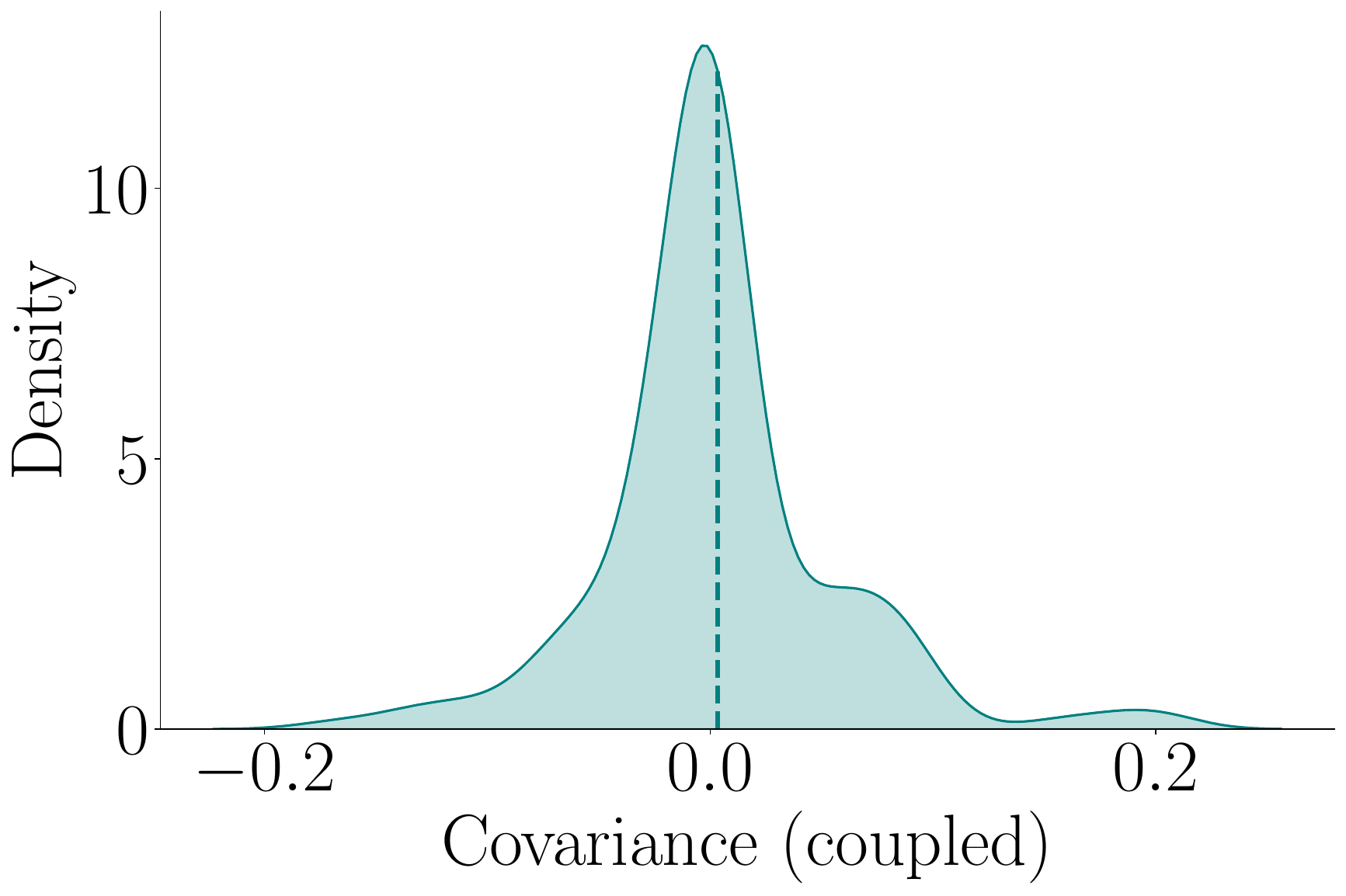} &
    \includegraphics[width=0.23\linewidth]{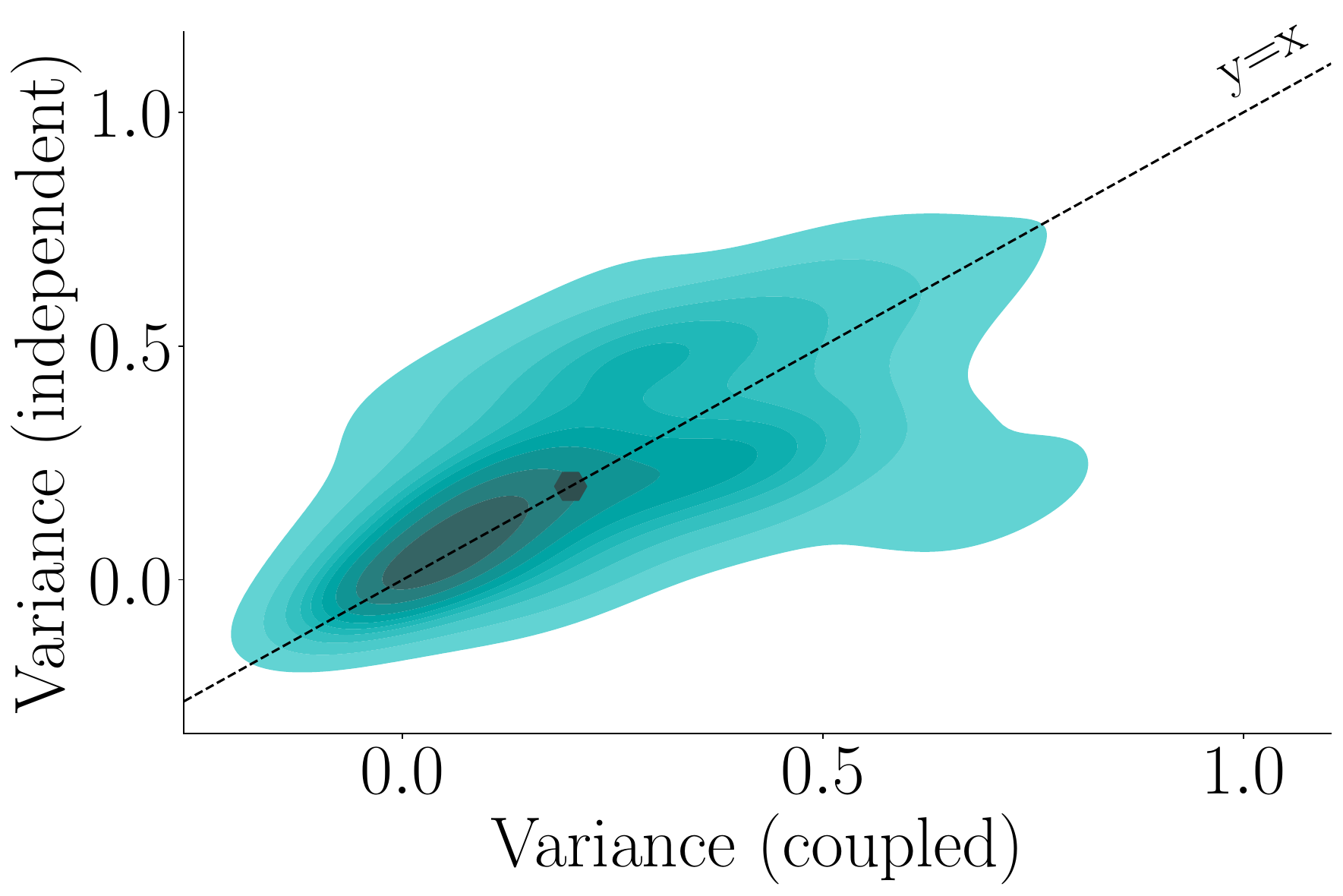} &
    \includegraphics[width=0.23\linewidth]{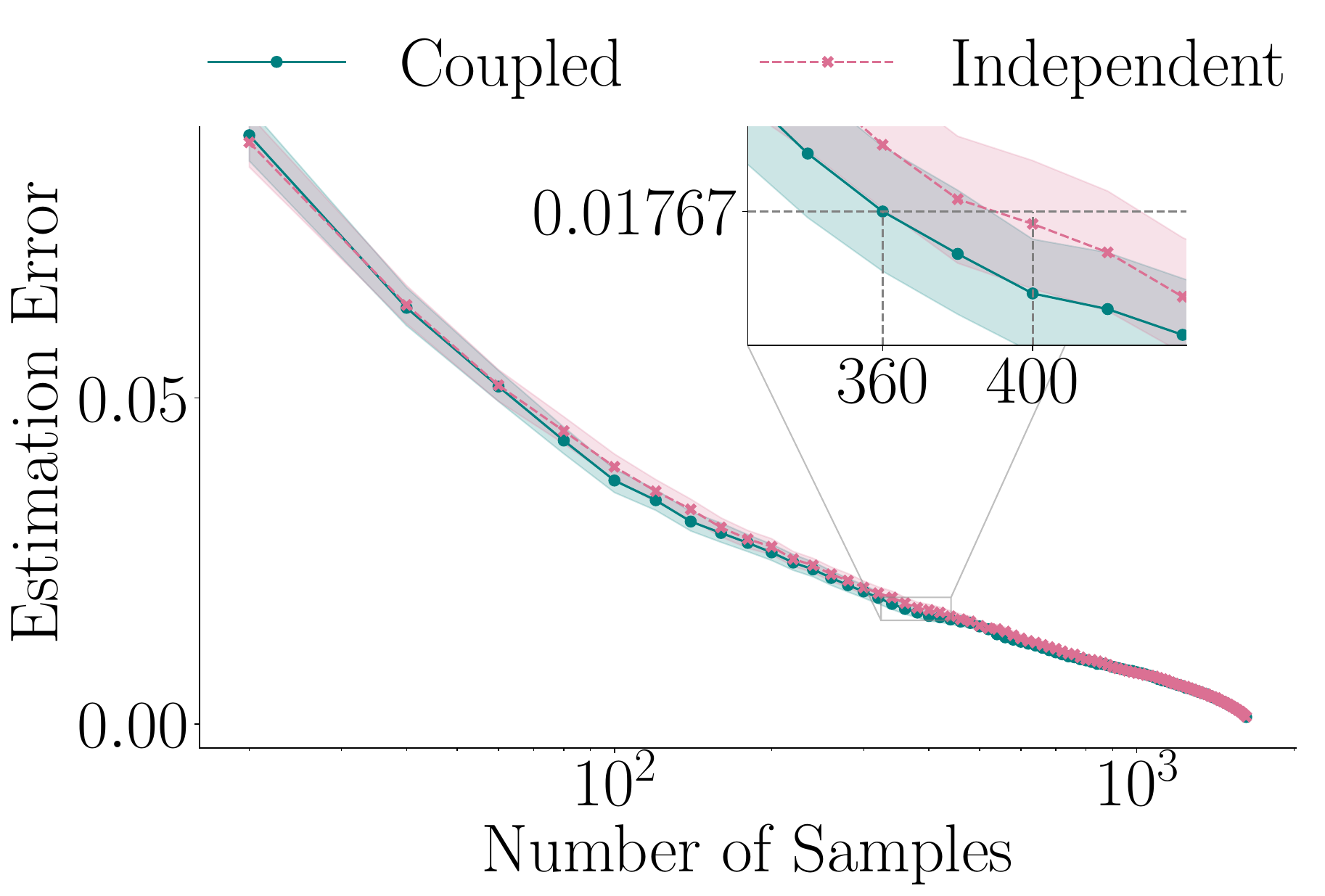} \\ \\
    (a) Score covariance & (b) Variance of the score difference & (c) Estimation error vs. \# samples \\

\end{tabular}
    \caption{\textbf{Comparison between several pairs of LLMs in the \texttt{Llama} family on programming problems from the HumanEval dataset.}
    Panels in column (a) show the kernel density estimate (KDE) of the covariance between the scores of the two LLMs on each problem under coupled generation; the dashed lines correspond to average values. Panels in column (b) show the KDE of the variance of the difference between the scores of the LLMs on each question under coupled and independent generation; the highlighted points correspond to median values. Panels in column (c) show the absolute error in the estimation of the expected difference between the scores of the LLMs against the number of samples; for each point on the x-axis, we perform $1{,}000$ sub-samplings and shaded areas correspond to $95\%$ confidence intervals.}
    \label{fig:human-eval-first-5}
\end{figure}

\begin{figure}[ht]
\centering
\begin{tabular}{c c c}
    \multicolumn{3}{c}{\texttt{8B} vs. \texttt{AWQ-INT4}}\\
    \includegraphics[width=0.23\linewidth]{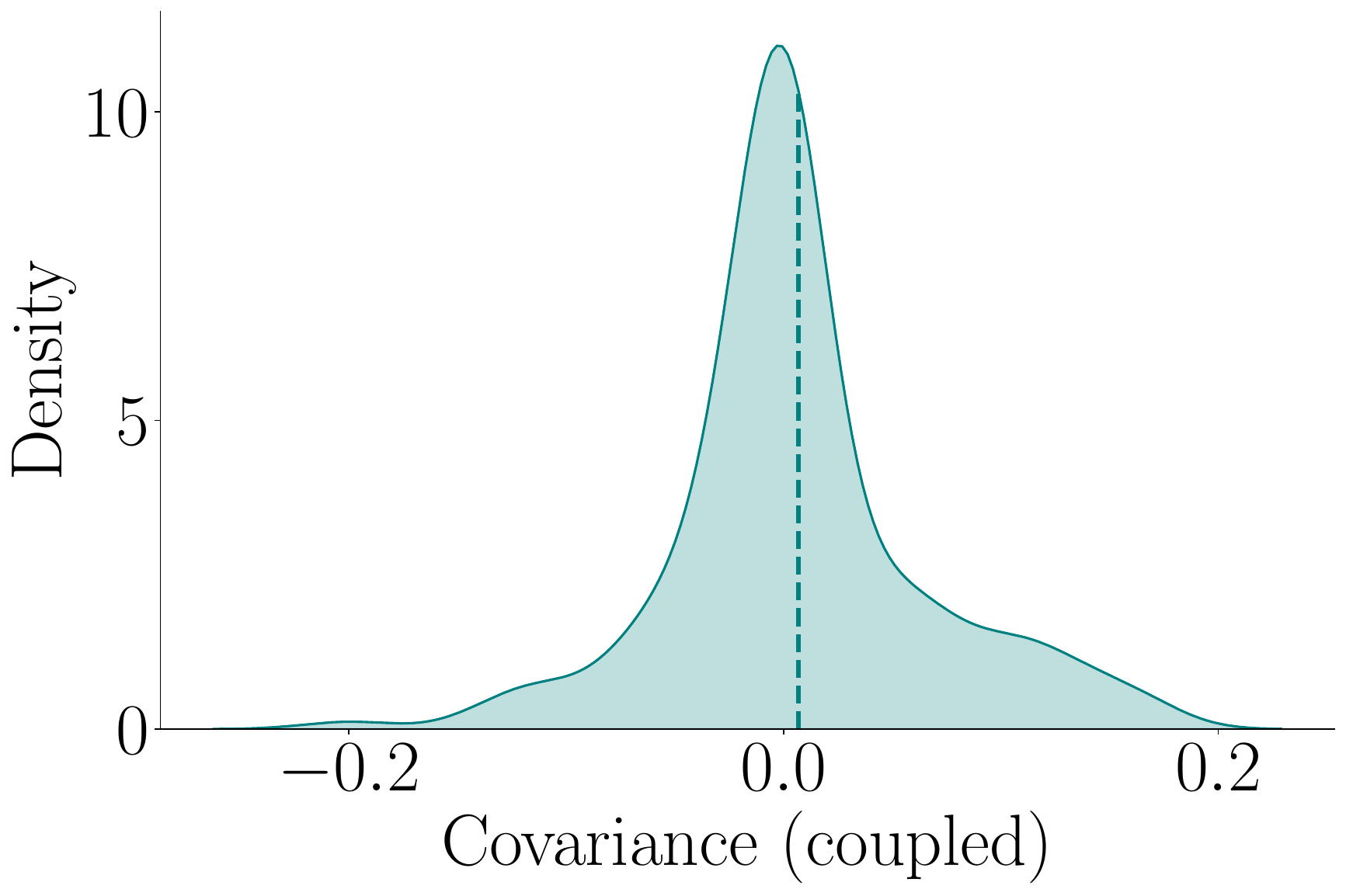} &
    \includegraphics[width=0.23\linewidth]{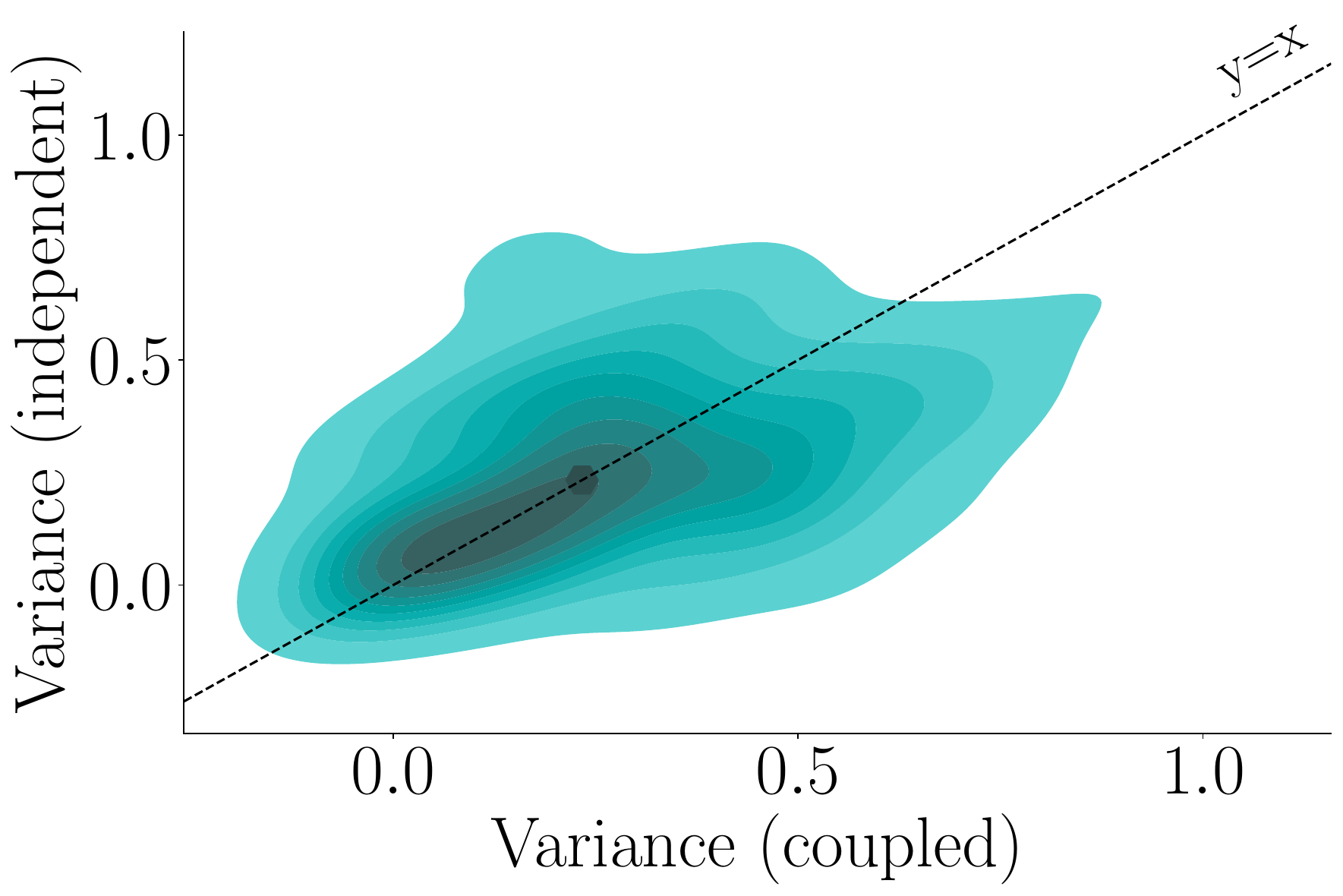} &
    \includegraphics[width=0.23\linewidth]{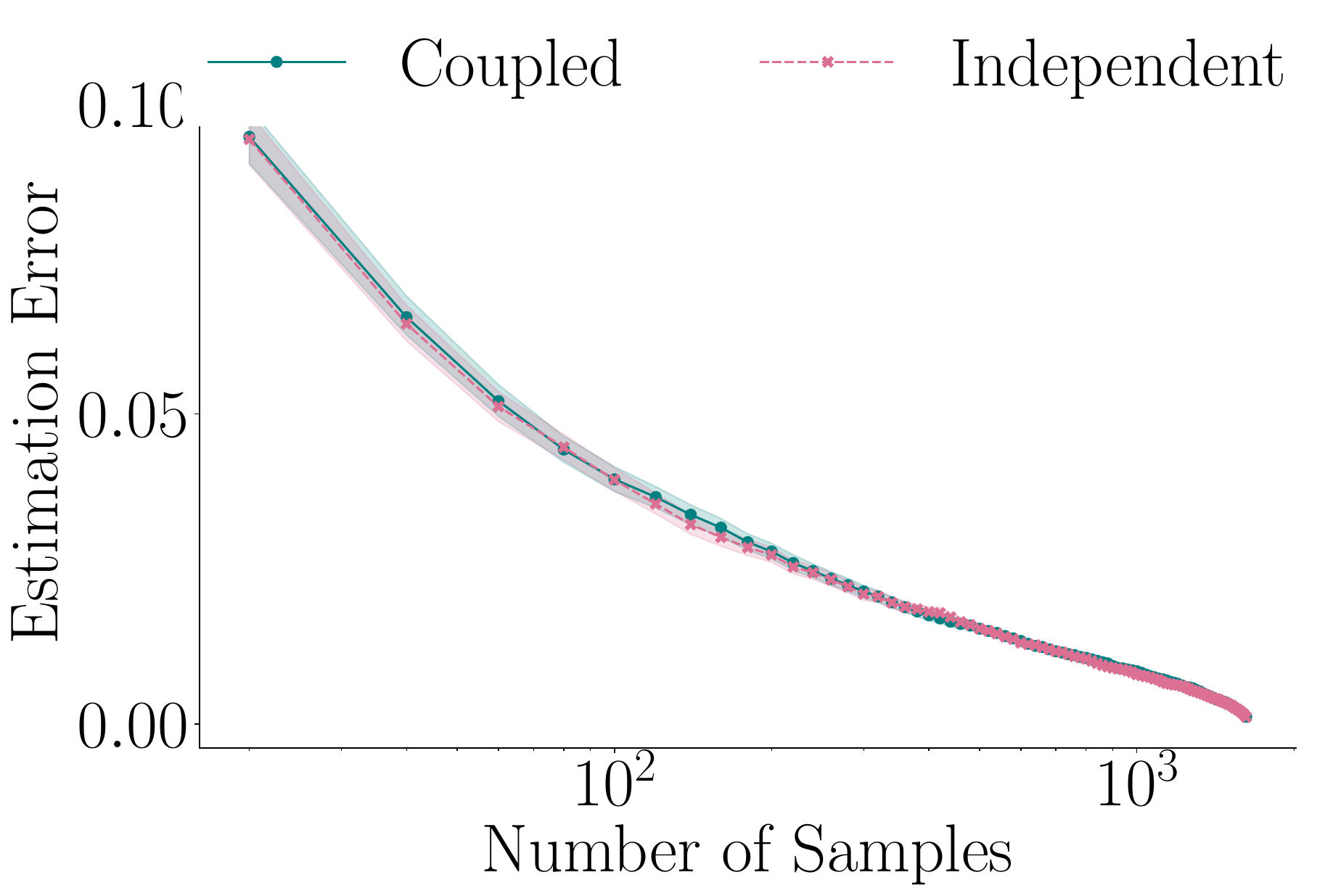} \\ \\
    \multicolumn{3}{c}{\texttt{3B} vs. \texttt{bnb-8bit}}\\
    \includegraphics[width=0.23\linewidth]{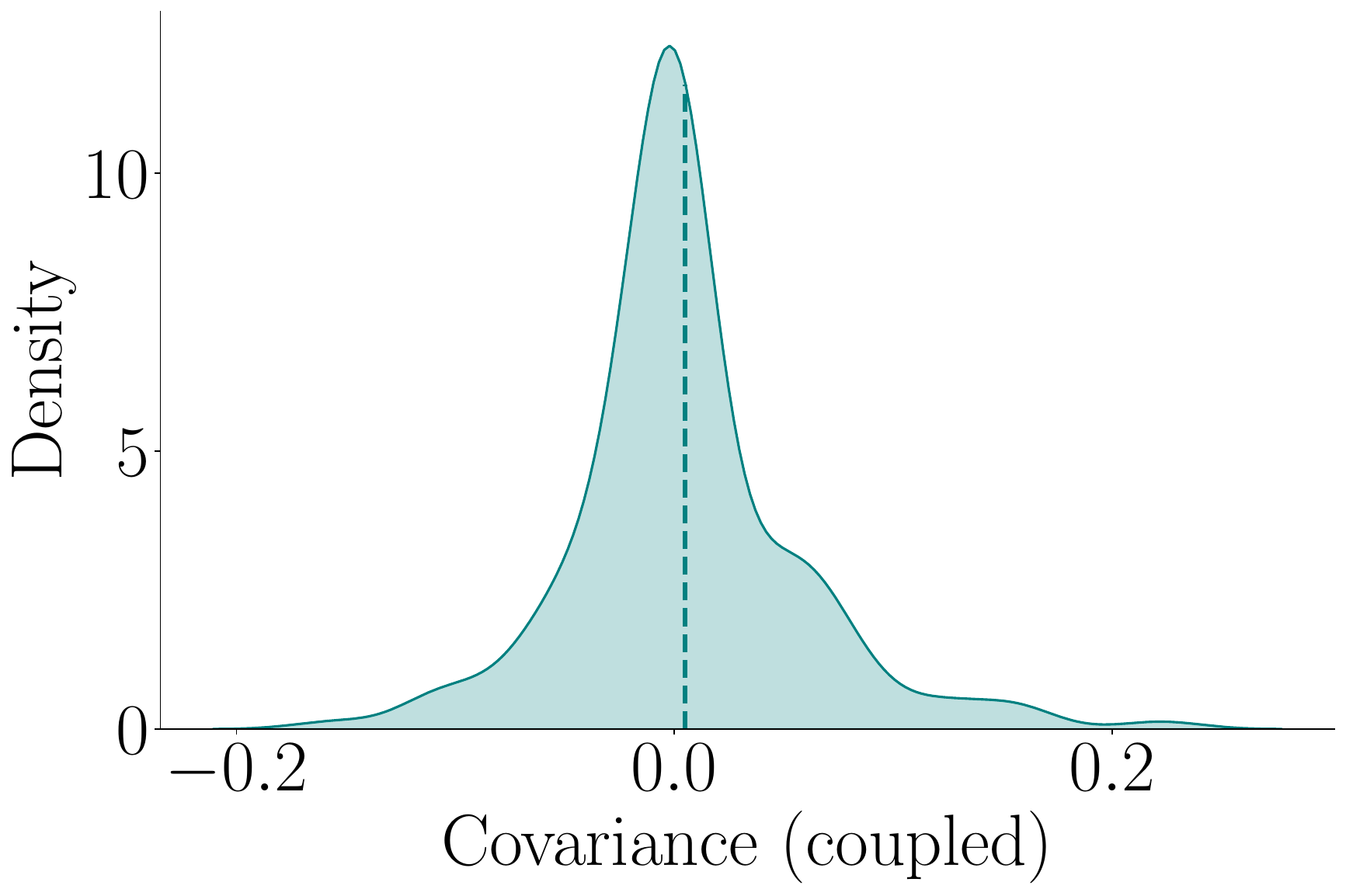} &
    \includegraphics[width=0.23\linewidth]{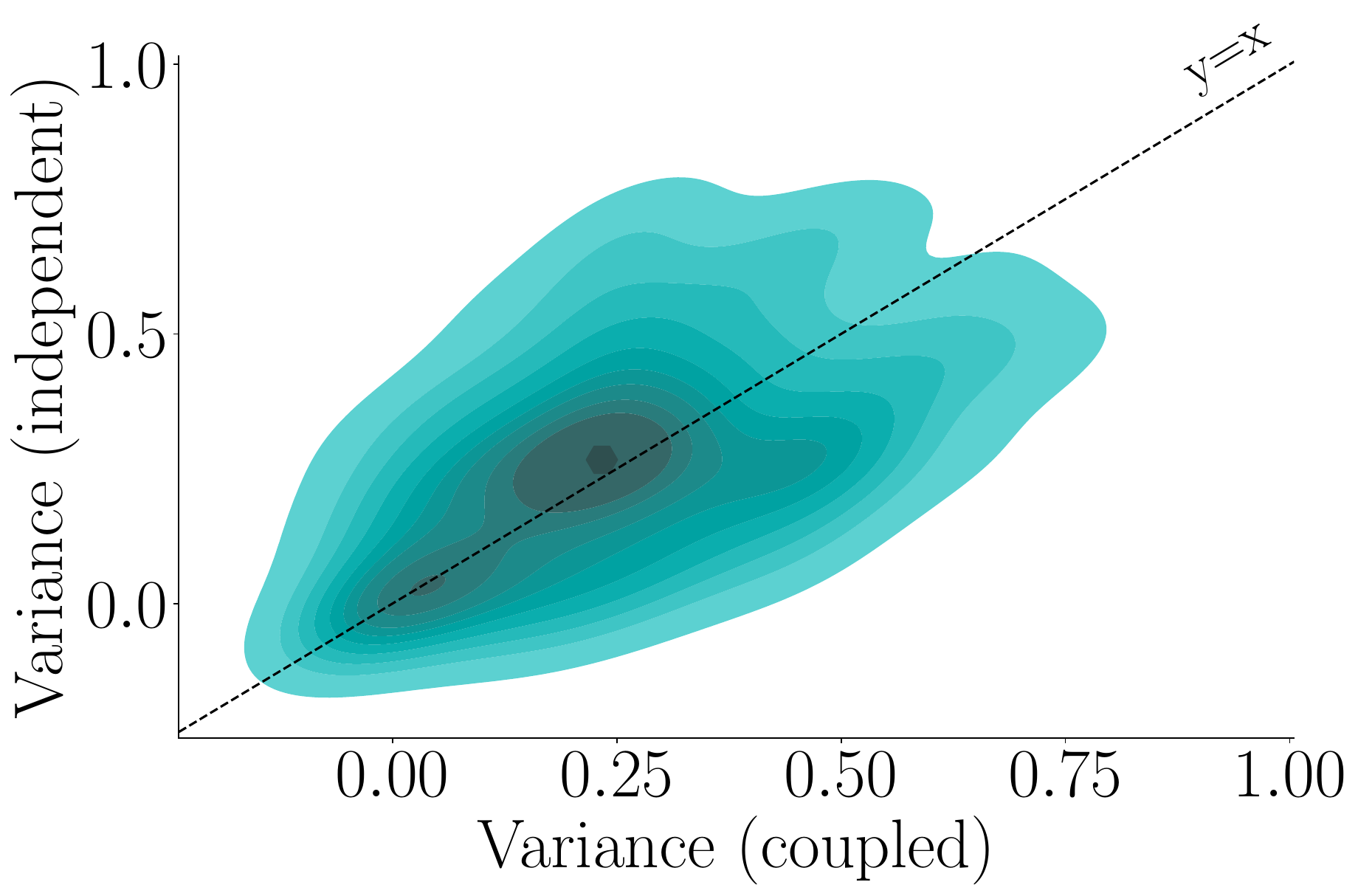} &
    \includegraphics[width=0.23\linewidth]{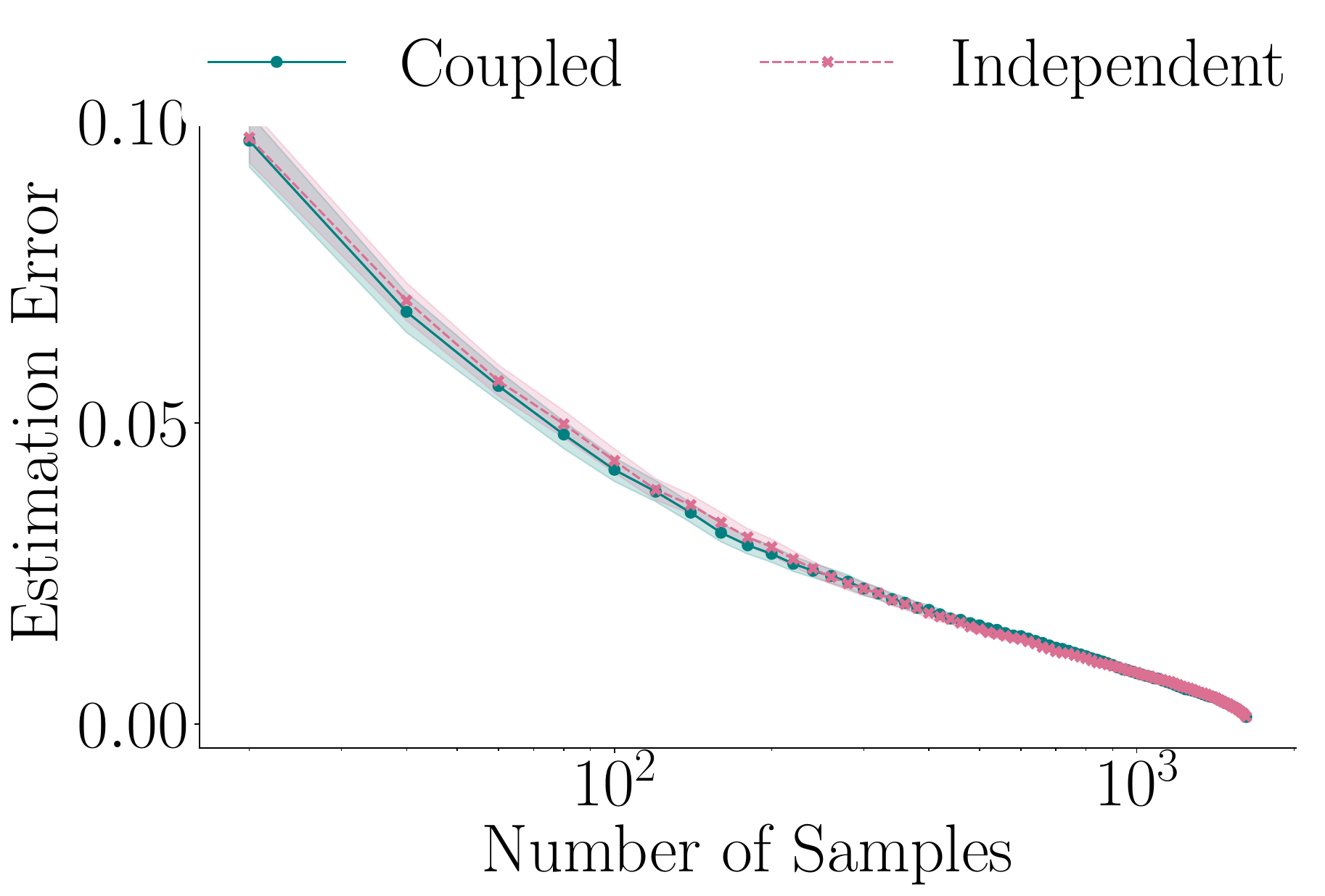} \\ \\
    \multicolumn{3}{c}{\texttt{3B} vs. \texttt{bnb-4bit}}\\
    \includegraphics[width=0.23\linewidth]{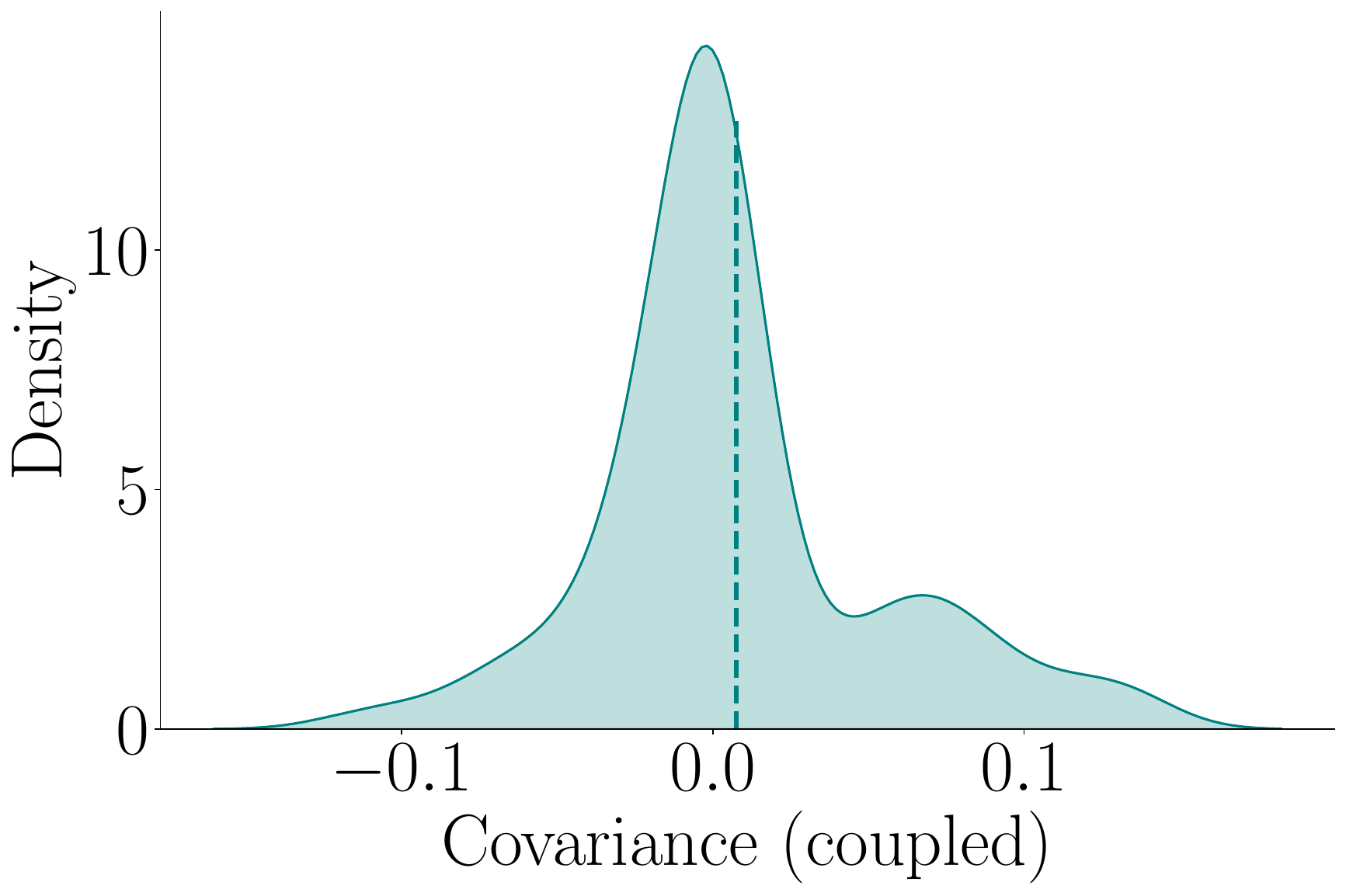} &
    \includegraphics[width=0.23\linewidth]{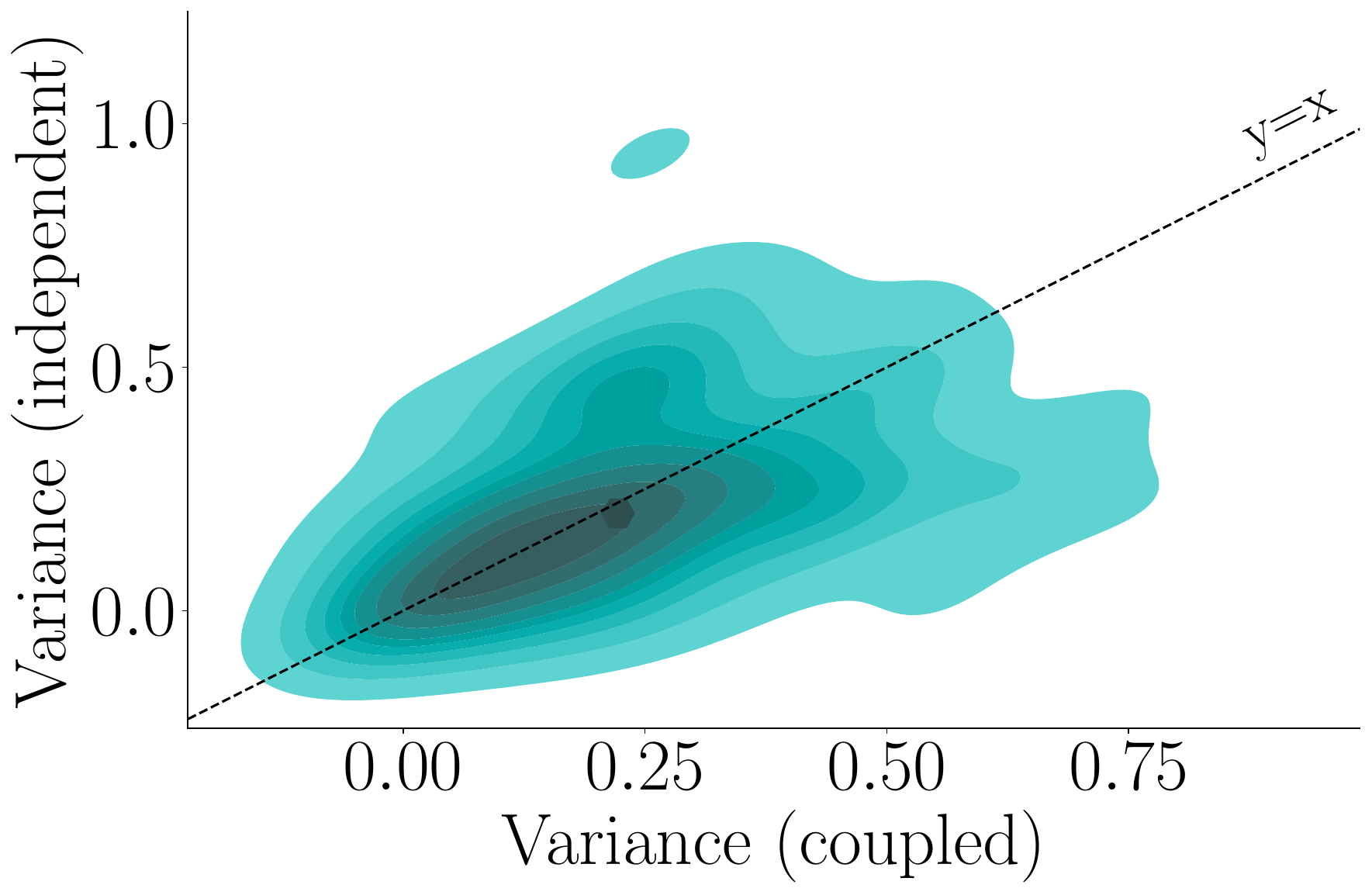} &
    \includegraphics[width=0.23\linewidth]{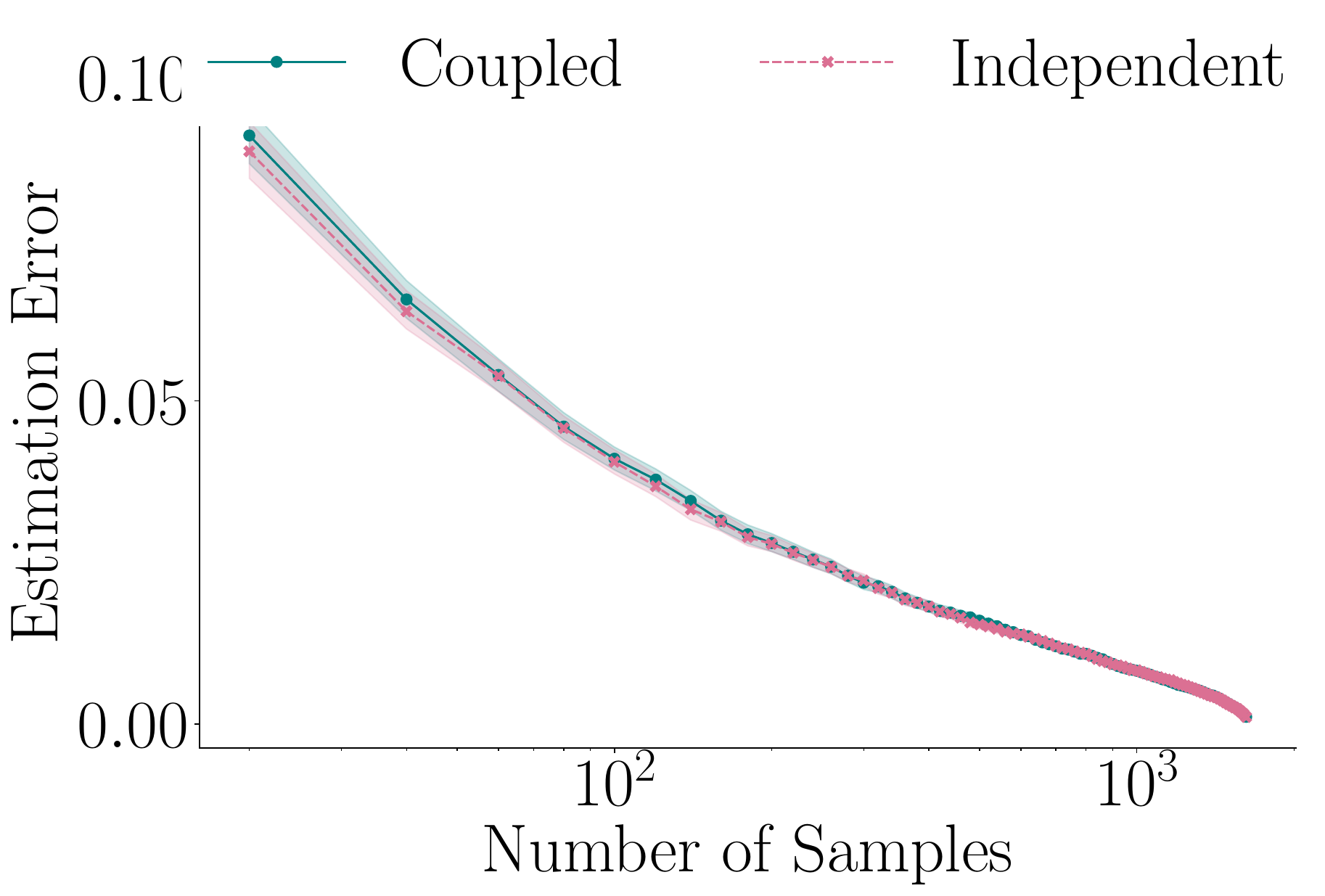} \\ \\
    \multicolumn{3}{c}{\texttt{3B} vs. \texttt{AWQ-INT4}}\\
    \includegraphics[width=0.23\linewidth]{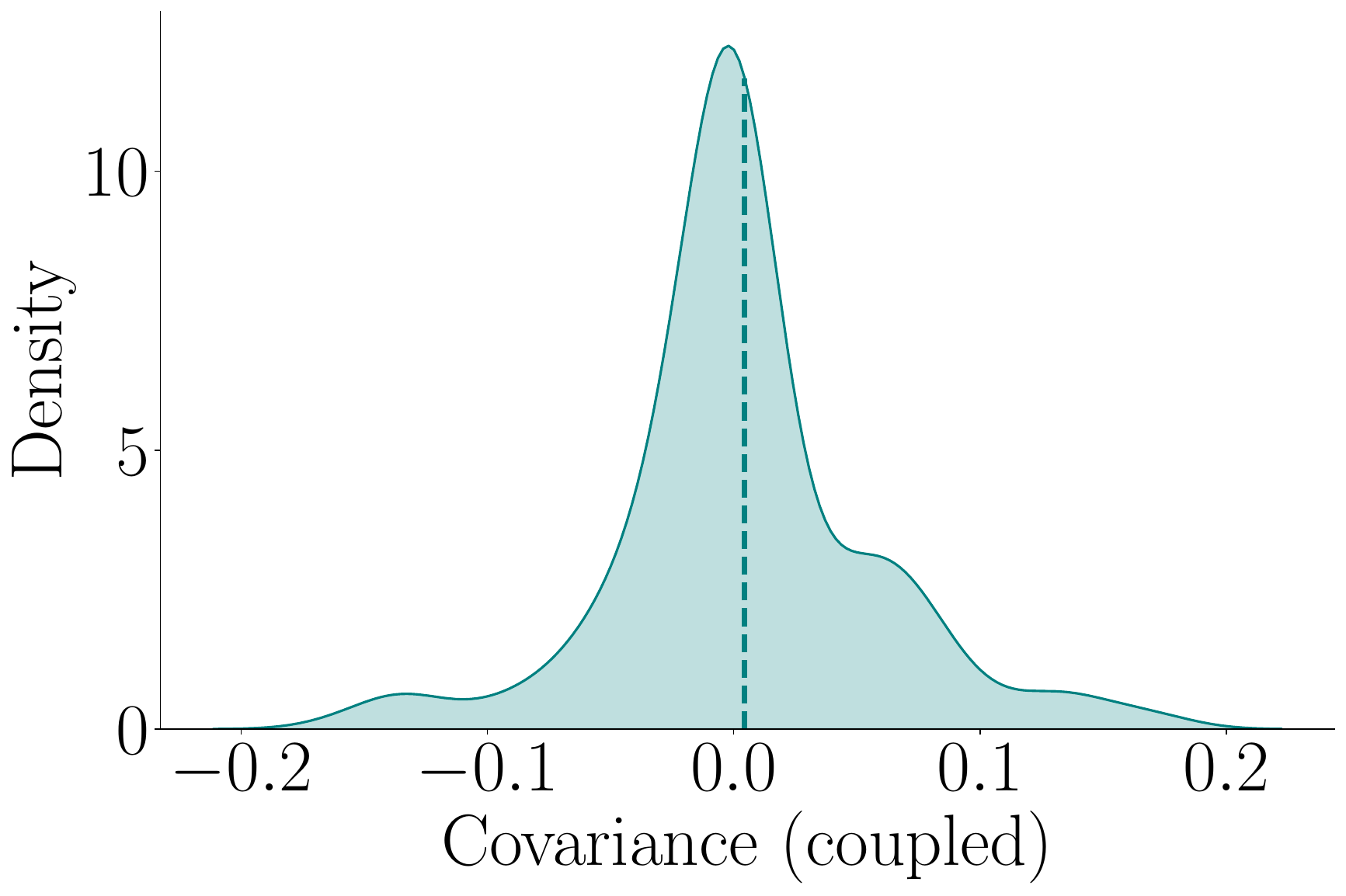} &
    \includegraphics[width=0.23\linewidth]{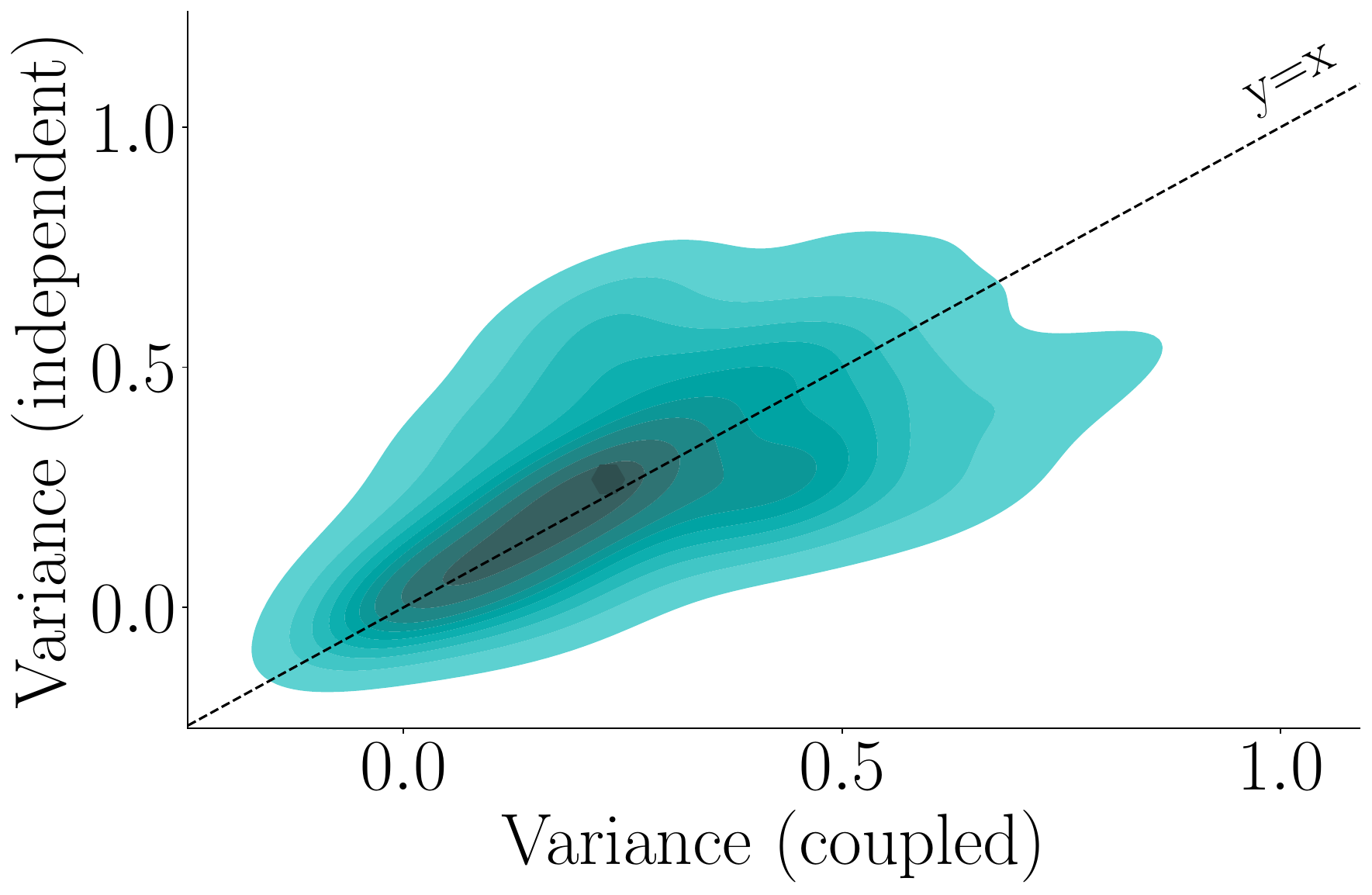} &
    \includegraphics[width=0.23\linewidth]{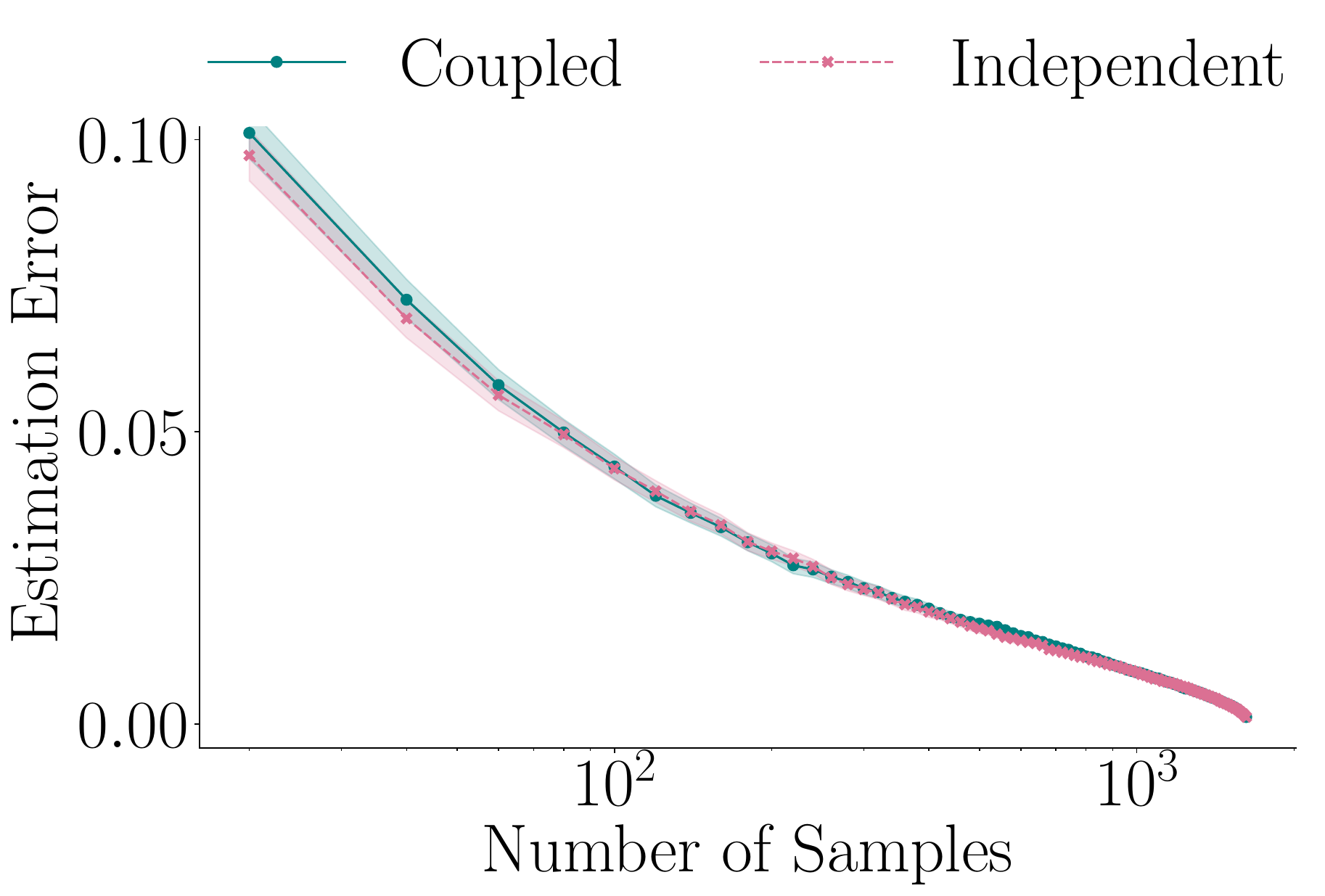} \\ \\

    \multicolumn{3}{c}{\texttt{1B} vs. \texttt{bnb-8bit}}\\
    \includegraphics[width=0.23\linewidth]{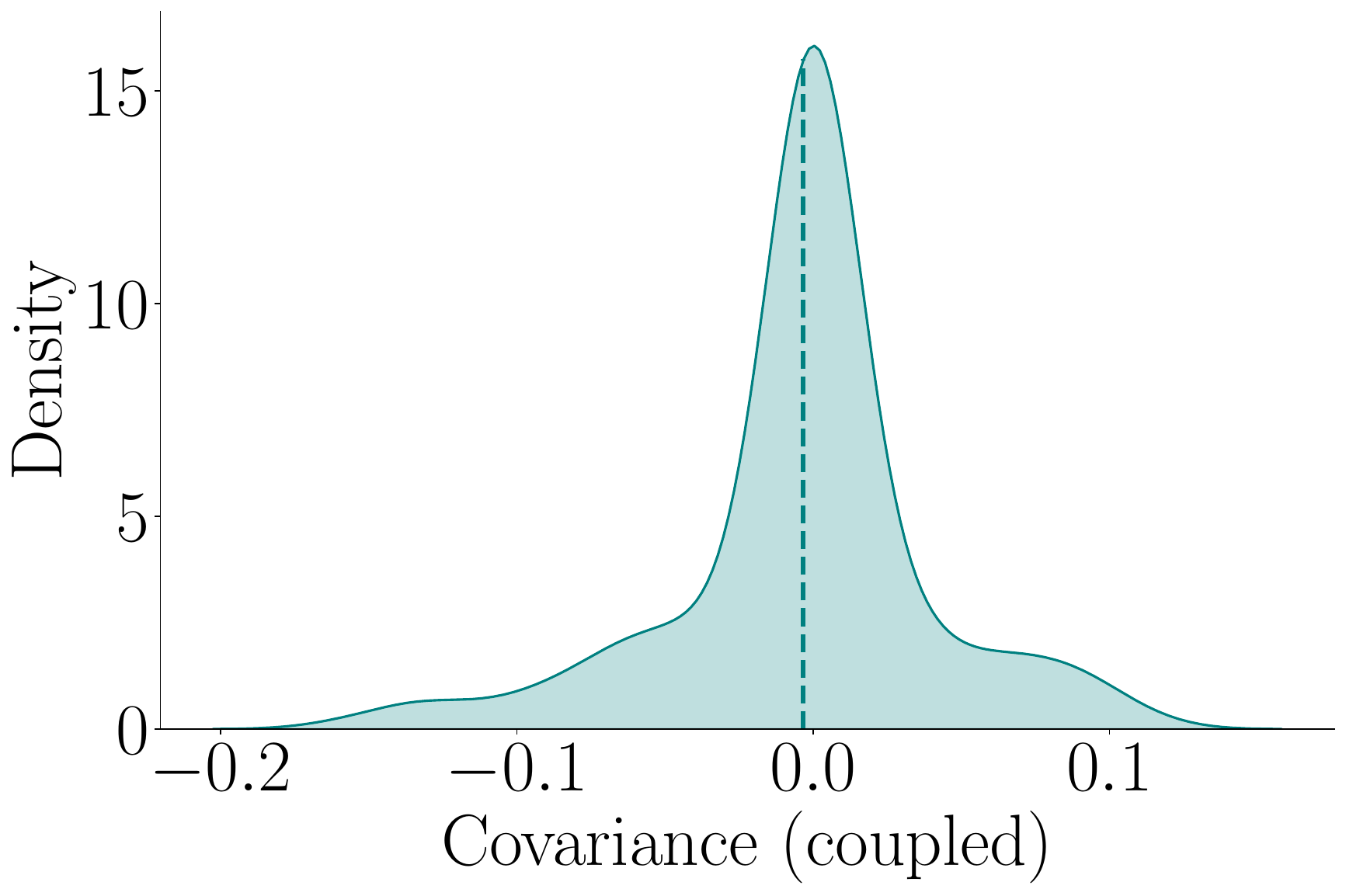} &
    \includegraphics[width=0.23\linewidth]{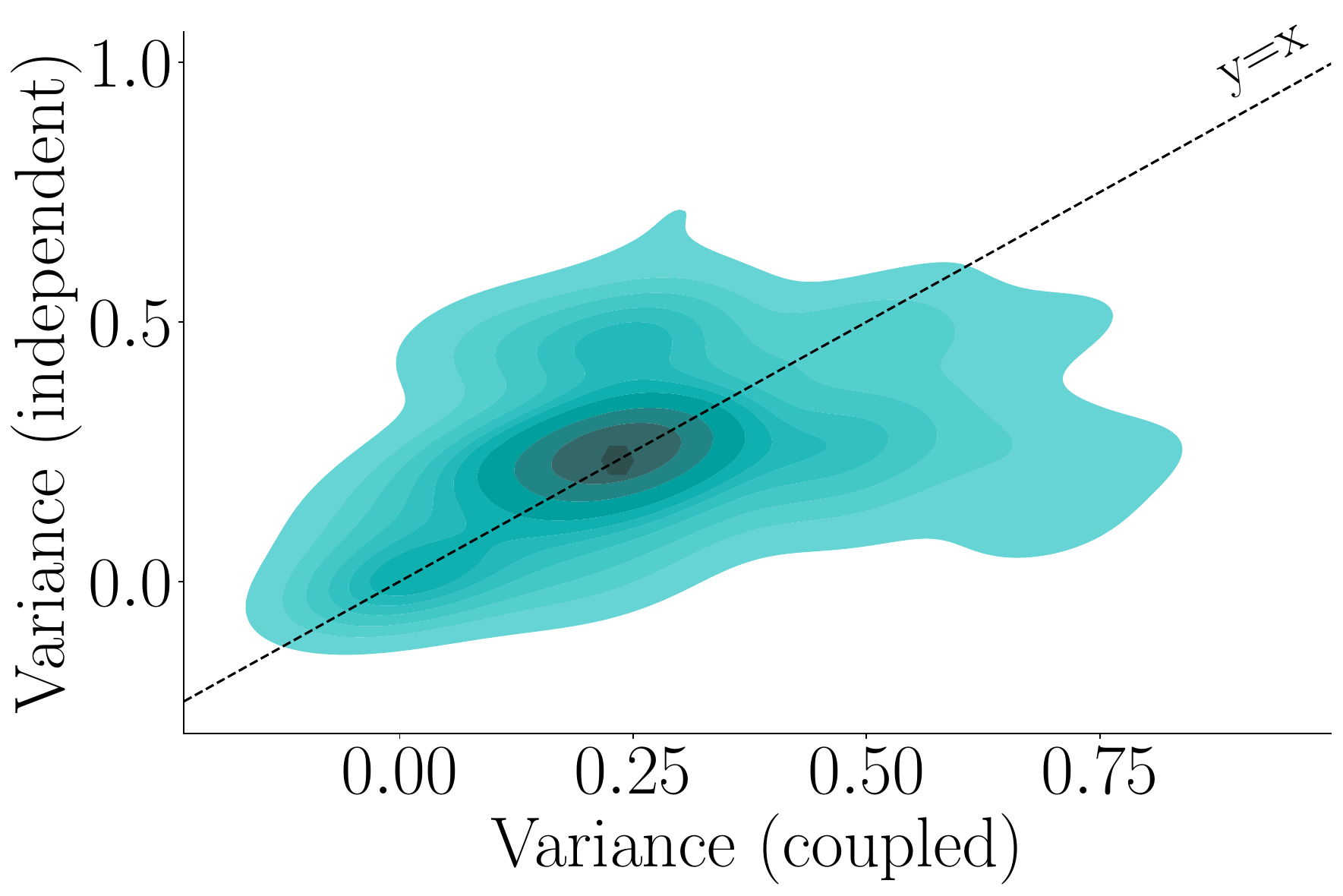} &
    \includegraphics[width=0.23\linewidth]{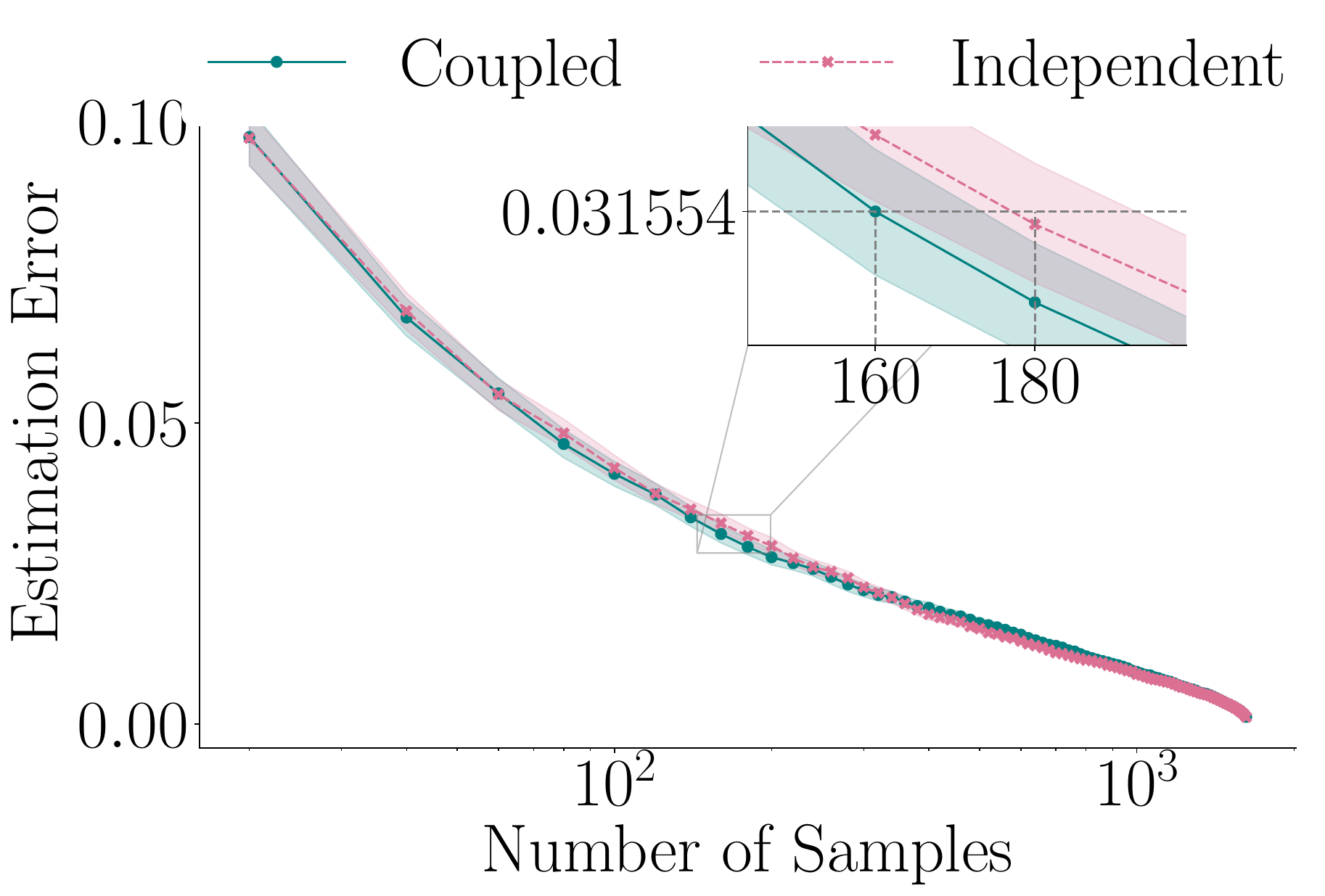} \\ \\
    (a) Score covariance & (b) Variance of the score difference & (c) Estimation error vs. \# samples \\

\end{tabular}
    \caption{\textbf{Comparison between several pairs of LLMs in the \texttt{Llama} family on programming problems from the HumanEval dataset.}
    Panels in column (a) show the kernel density estimate (KDE) of the covariance between the scores of the two LLMs on each problem under coupled generation; the dashed lines correspond to average values. Panels in column (b) show the KDE of the variance of the difference between the scores of the LLMs on each question under coupled and independent generation; the highlighted points correspond to median values. Panels in column (c) show the absolute error in the estimation of the expected difference between the scores of the LLMs against the number of samples; for each point on the x-axis, we perform $1{,}000$ sub-samplings and shaded areas correspond to $95\%$ confidence intervals.}
    \label{fig:human-eval-second-5}
\end{figure}

\begin{figure}[ht]
\centering
\begin{tabular}{c c c}
    \multicolumn{3}{c}{\texttt{1B} vs. \texttt{bnb-4bit}}\\
    \includegraphics[width=0.23\linewidth]{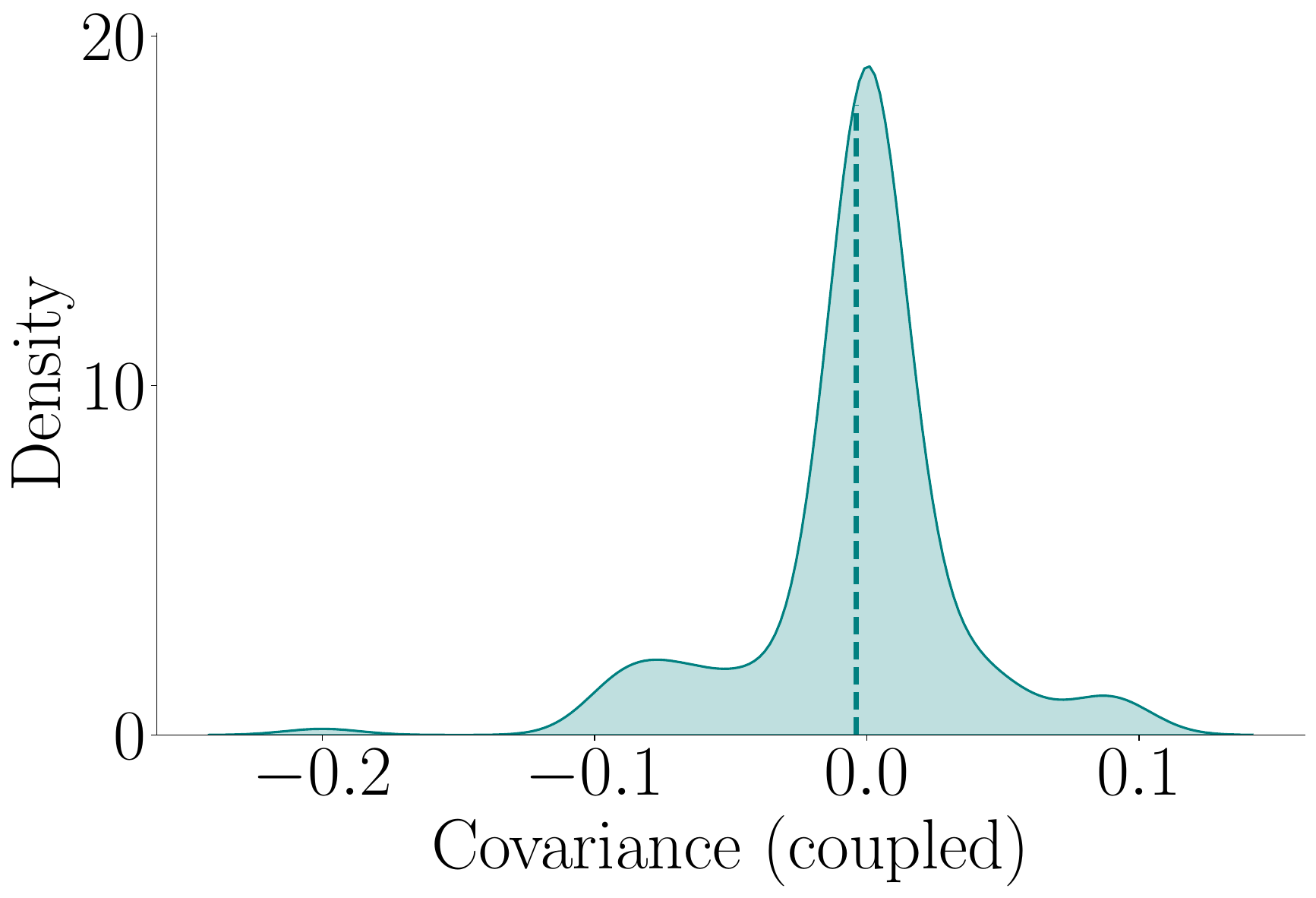} &
    \includegraphics[width=0.23\linewidth]{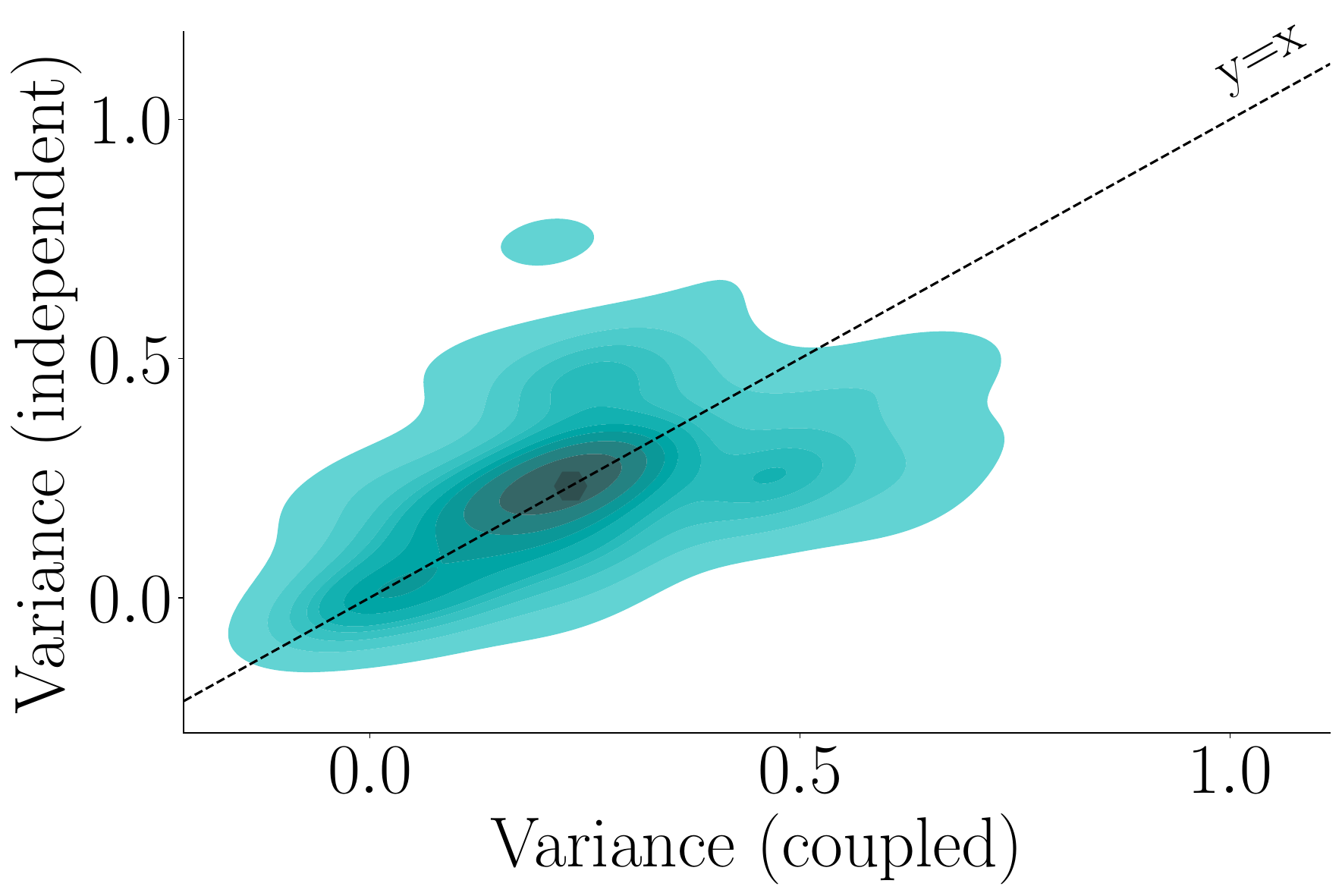} &
    \includegraphics[width=0.23\linewidth]{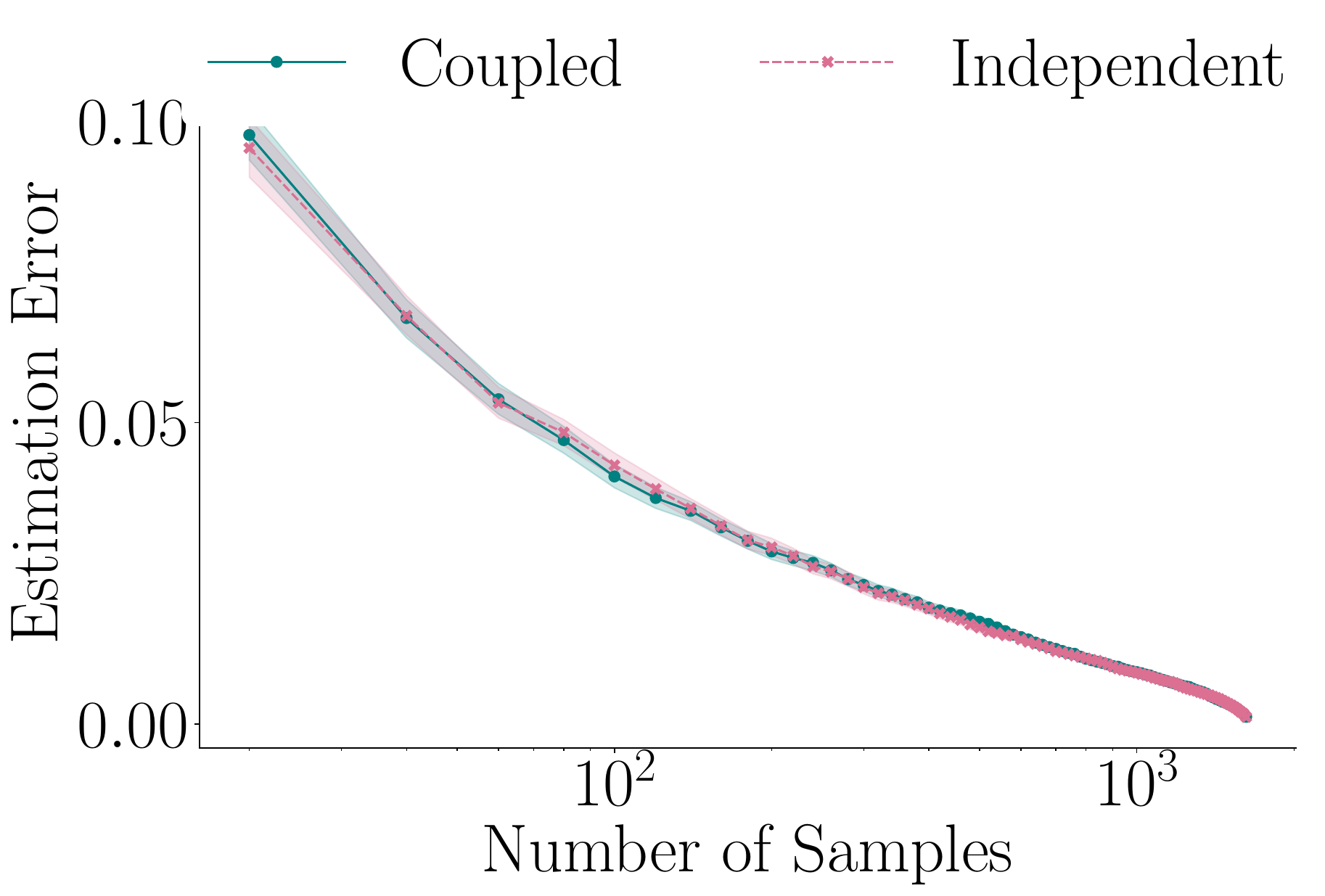} \\ \\
    \multicolumn{3}{c}{\texttt{1B} vs. \texttt{AWQ-INT4}}\\
    \includegraphics[width=0.23\linewidth]{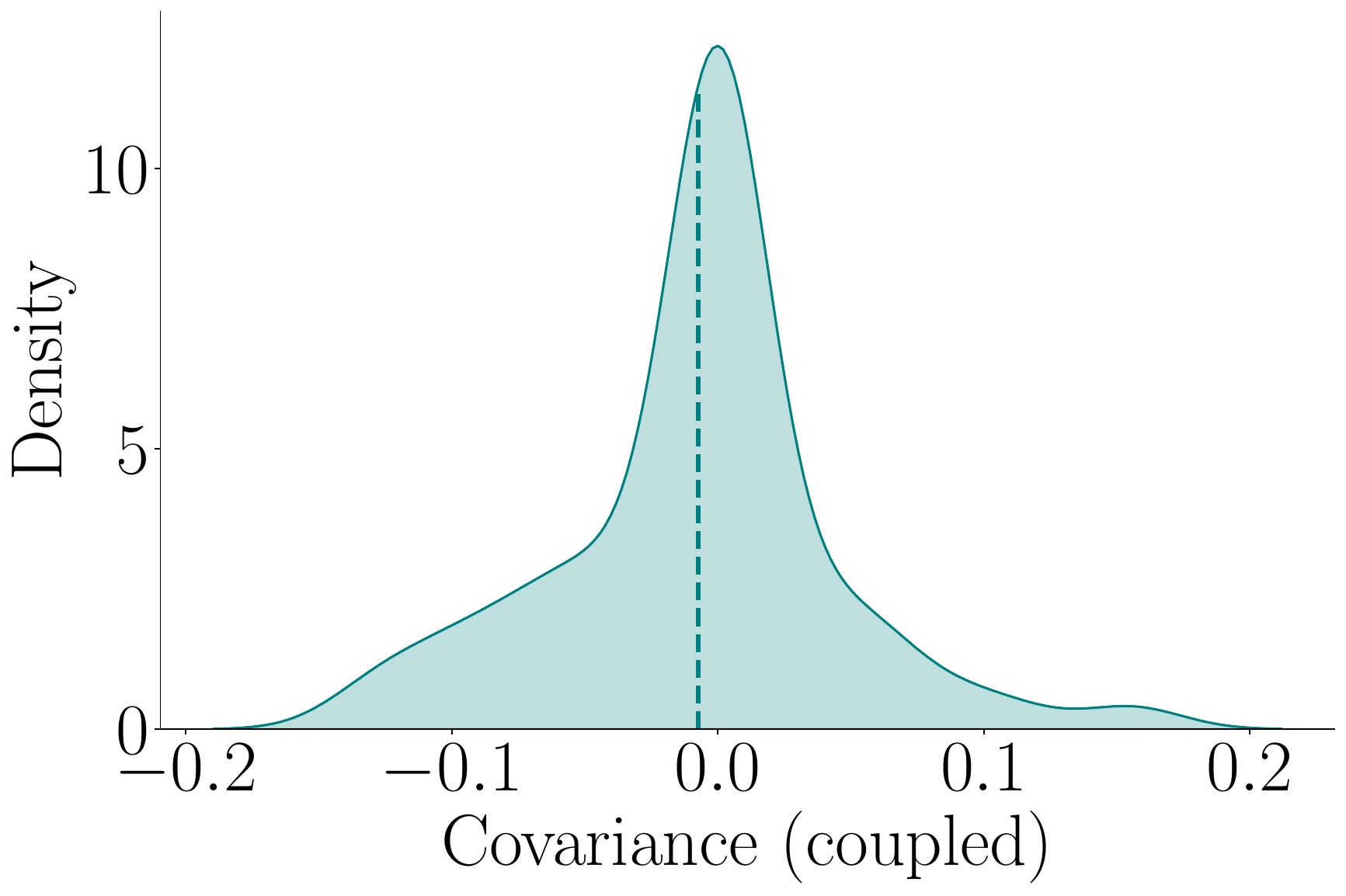} &
    \includegraphics[width=0.23\linewidth]{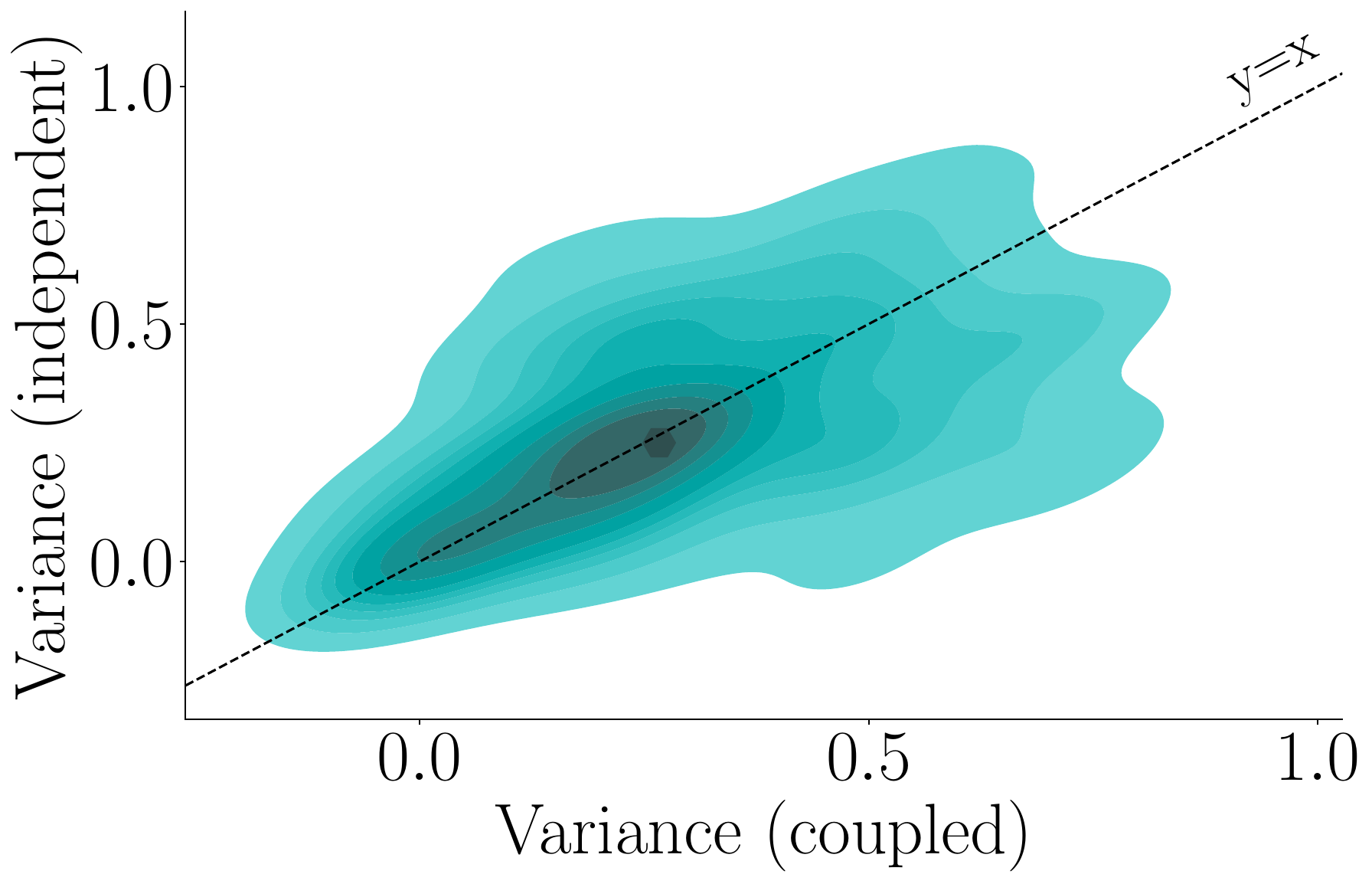} &
    \includegraphics[width=0.23\linewidth]{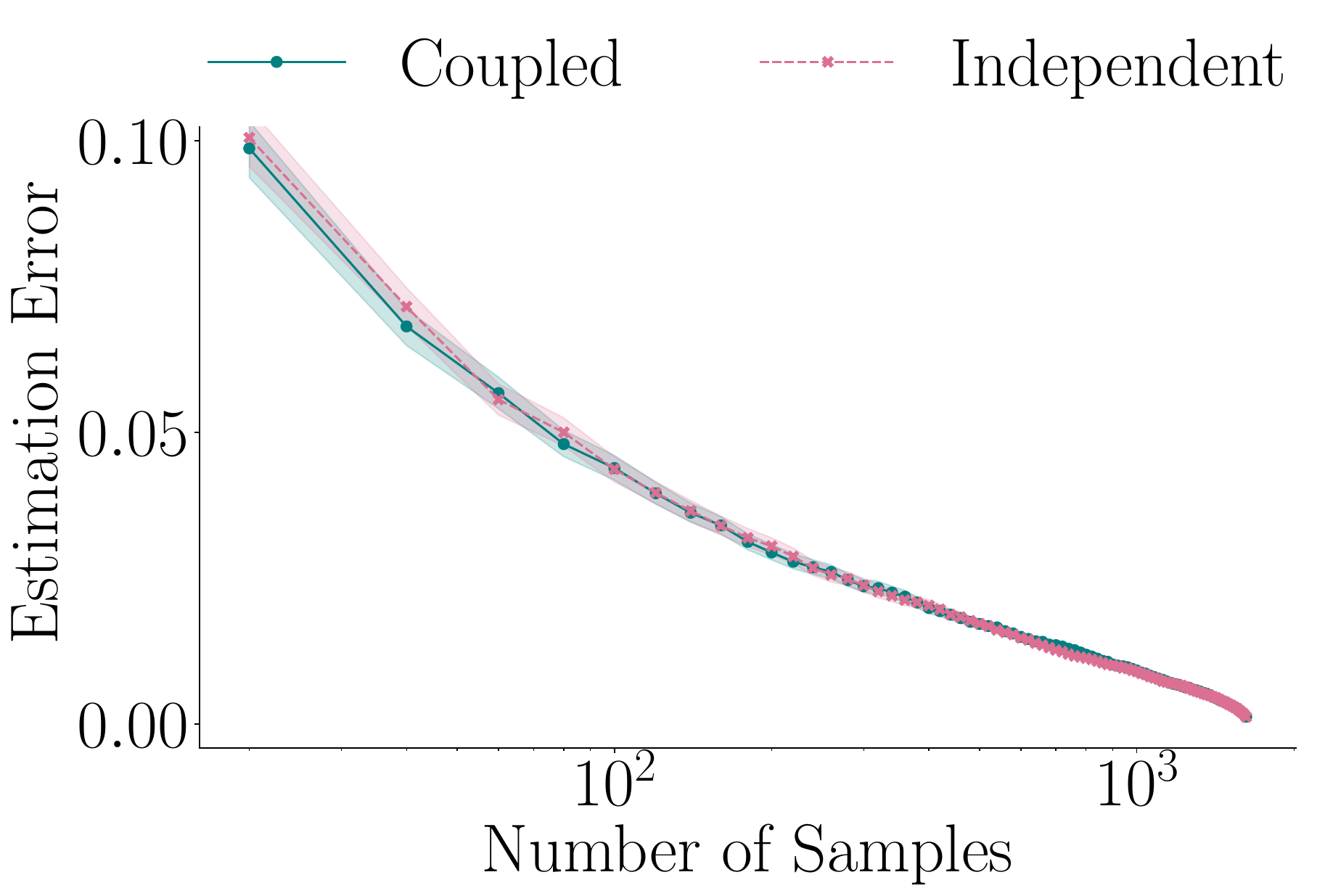} \\ \\
    \multicolumn{3}{c}{\texttt{bnb-8bit} vs. \texttt{bnb-4bit}}\\
    \includegraphics[width=0.23\linewidth]{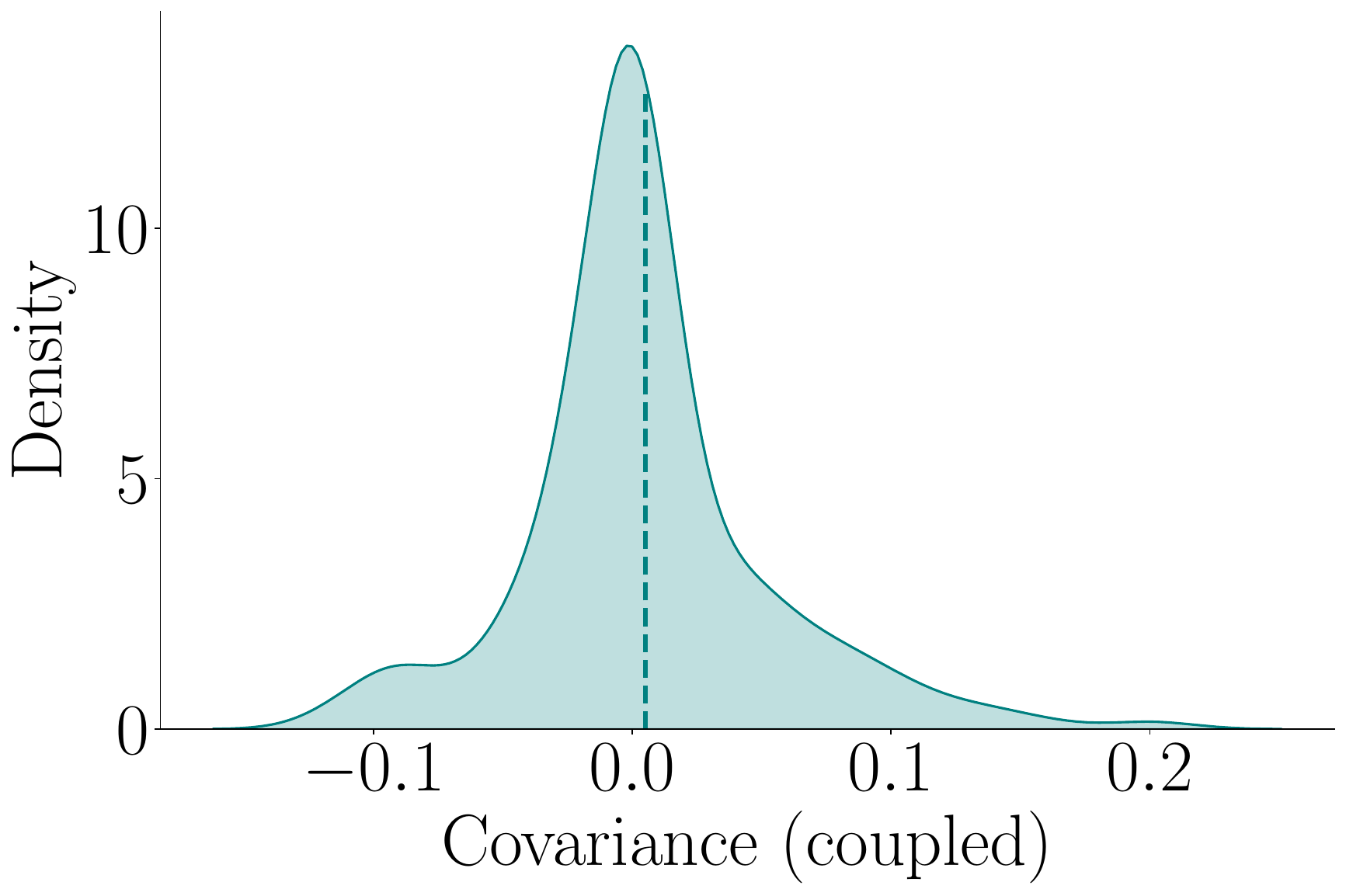} &
    \includegraphics[width=0.23\linewidth]{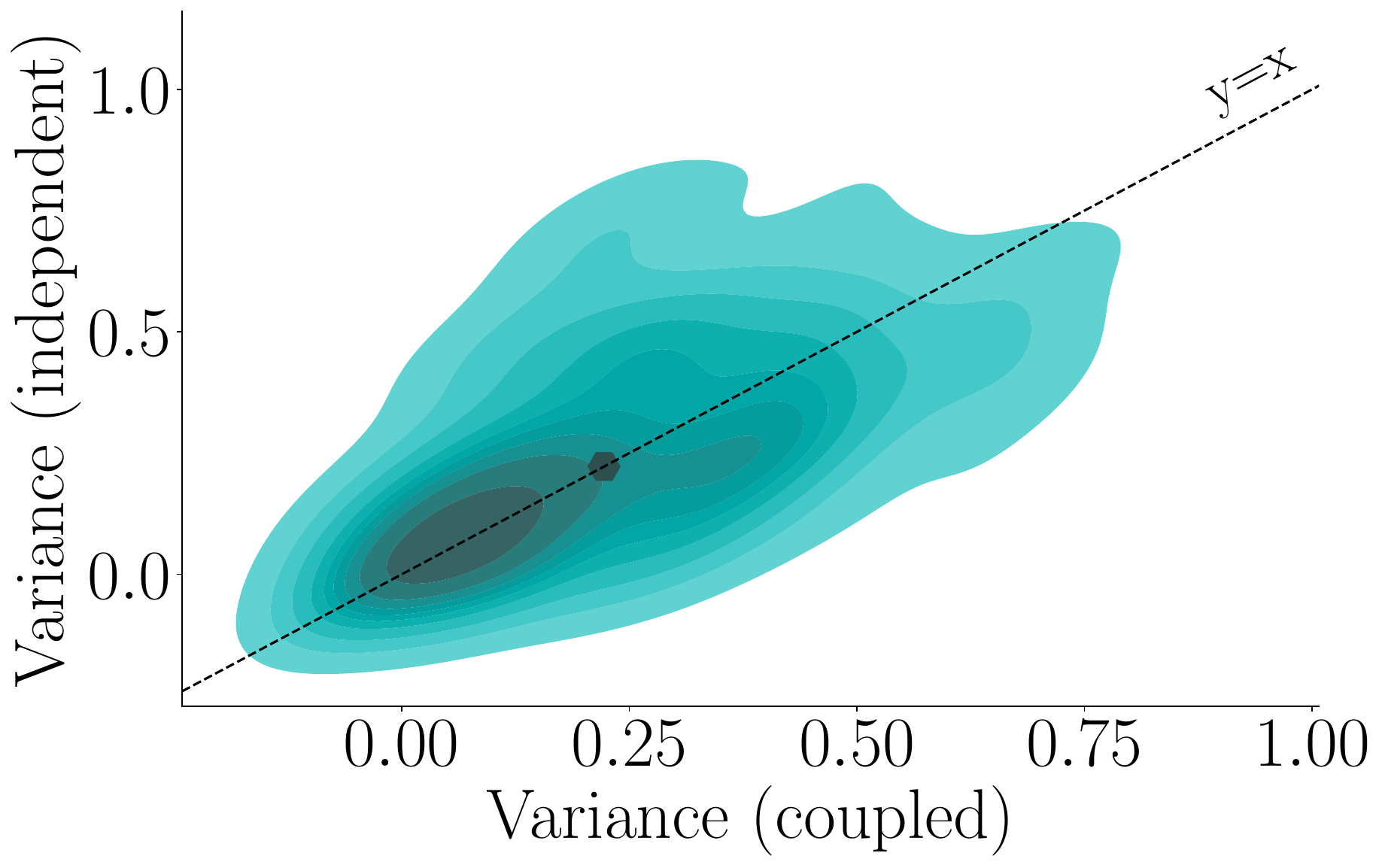} &
    \includegraphics[width=0.23\linewidth]{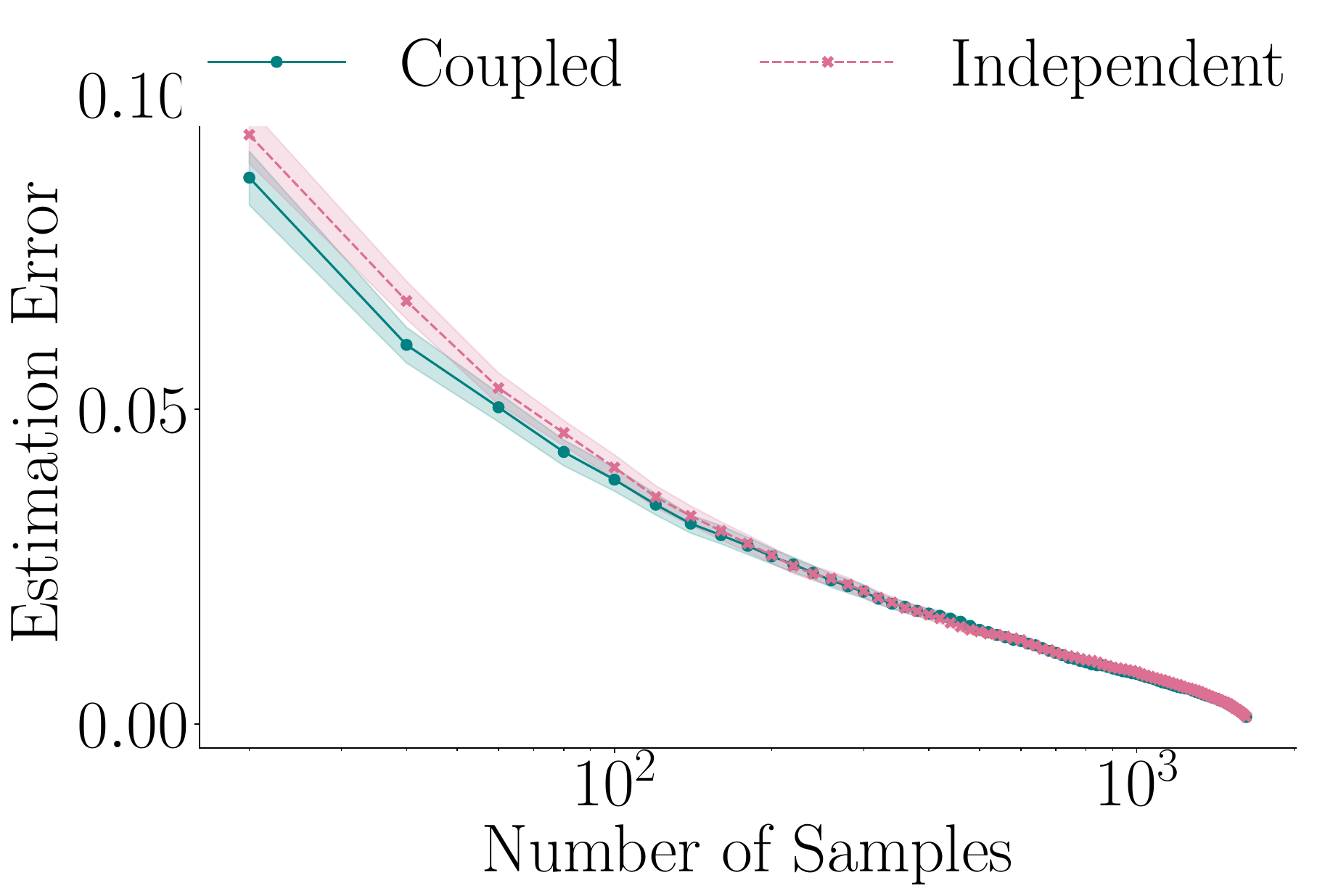} \\ \\
    \multicolumn{3}{c}{\texttt{bnb-8bit} vs. \texttt{AWQ-INT4}}\\
    \includegraphics[width=0.23\linewidth]{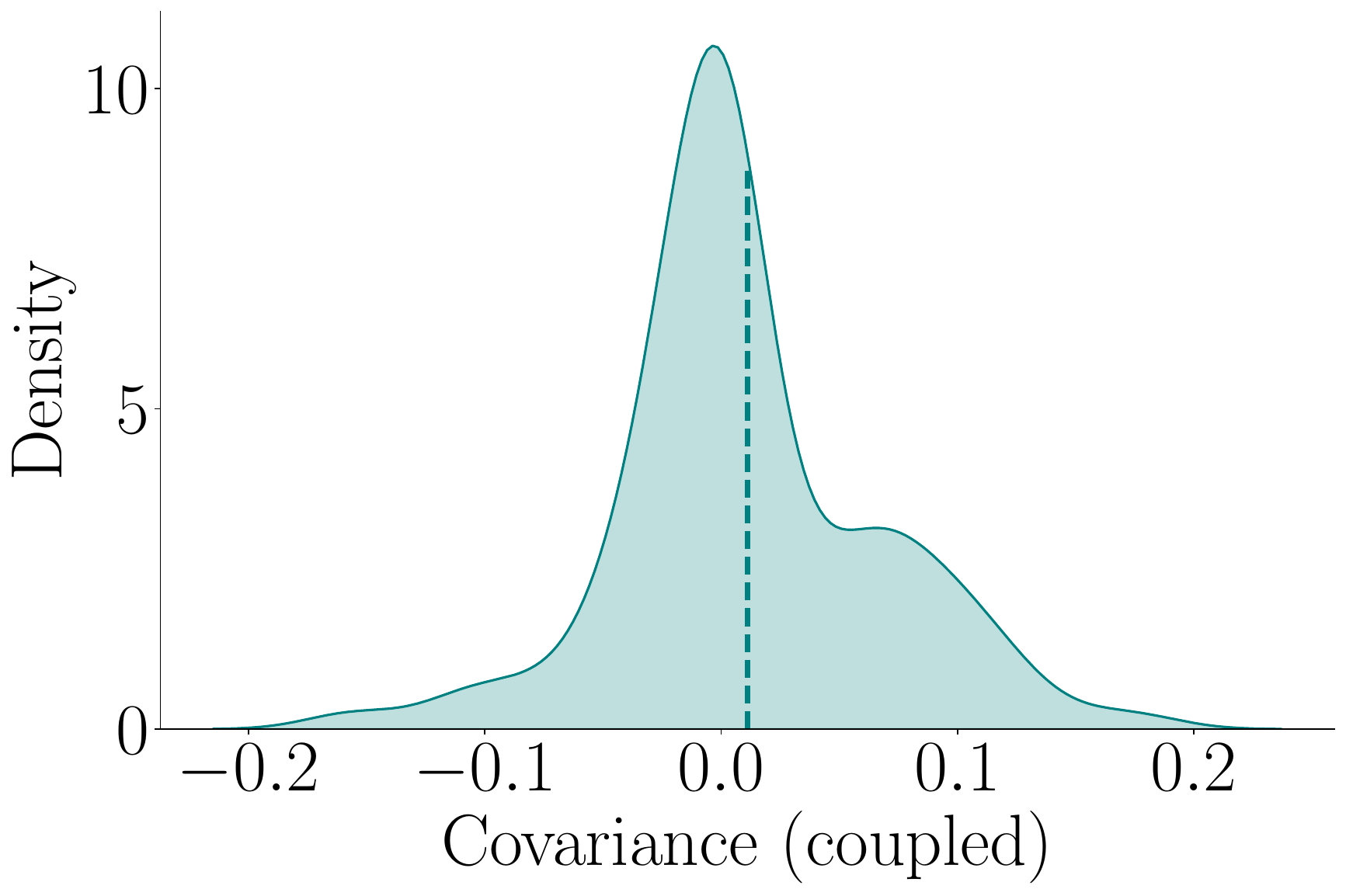} &
    \includegraphics[width=0.23\linewidth]{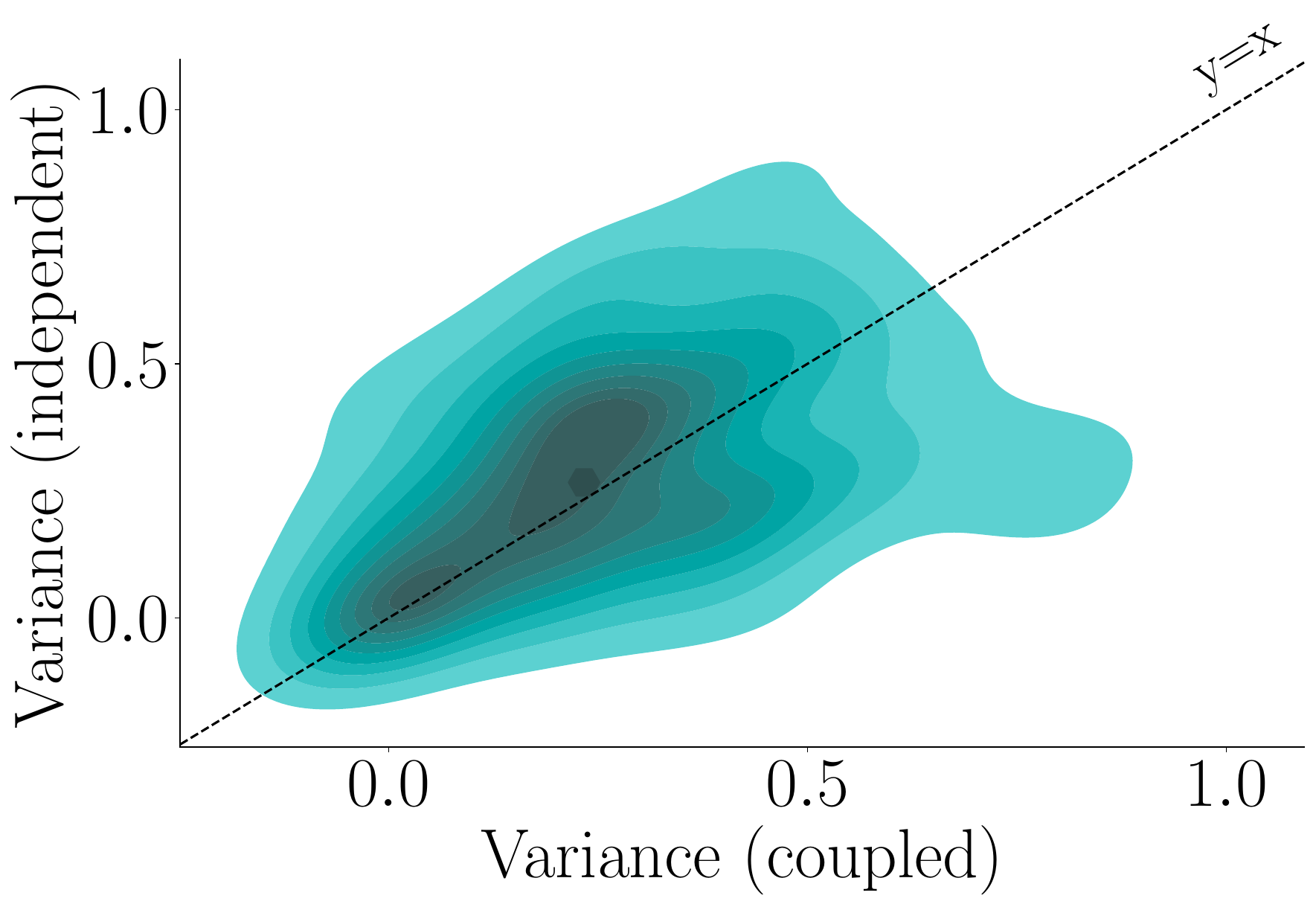} &
    \includegraphics[width=0.23\linewidth]{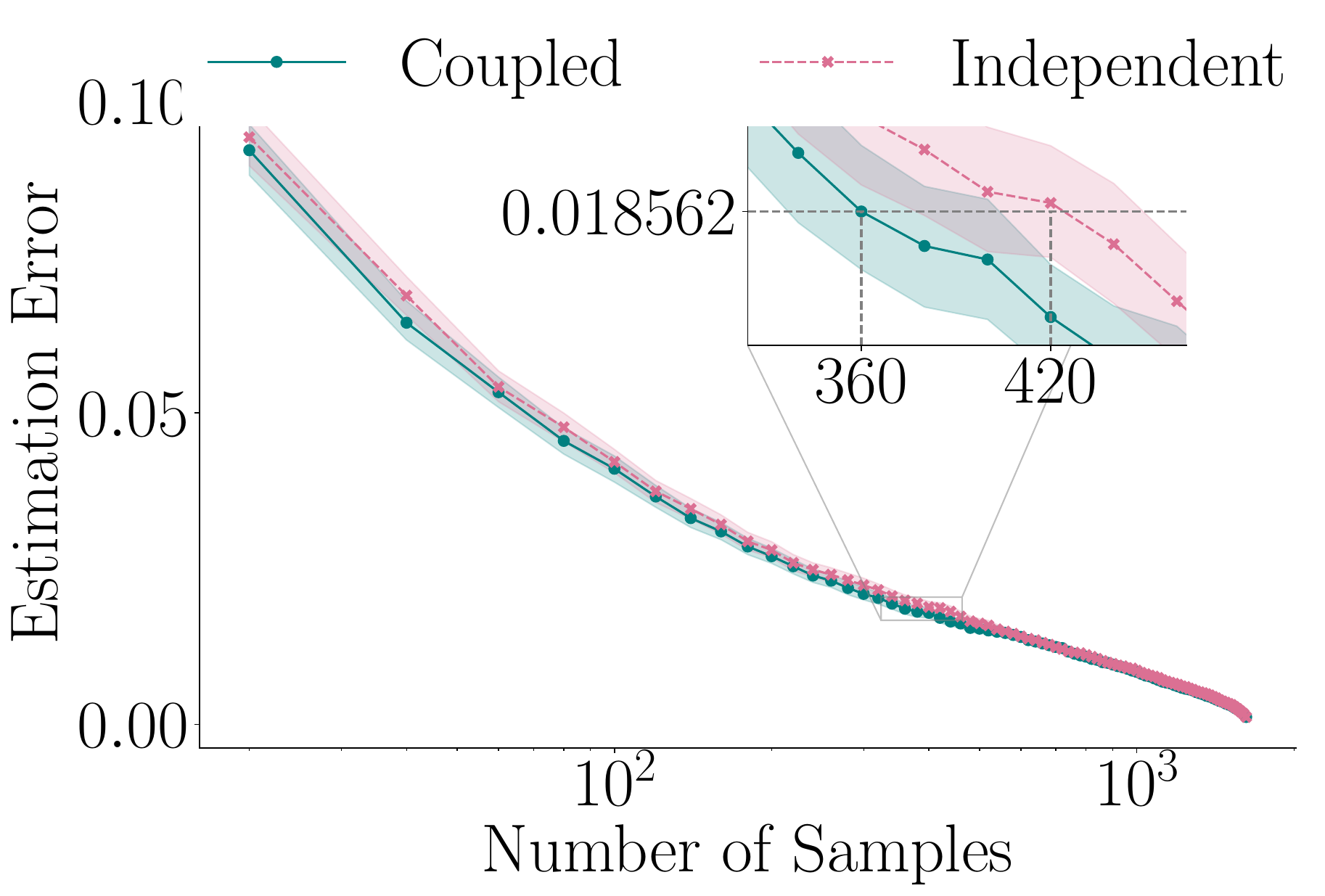} \\ \\

    \multicolumn{3}{c}{\texttt{bnb-4bit} vs. \texttt{AWQ-INT4}}\\
    \includegraphics[width=0.23\linewidth]{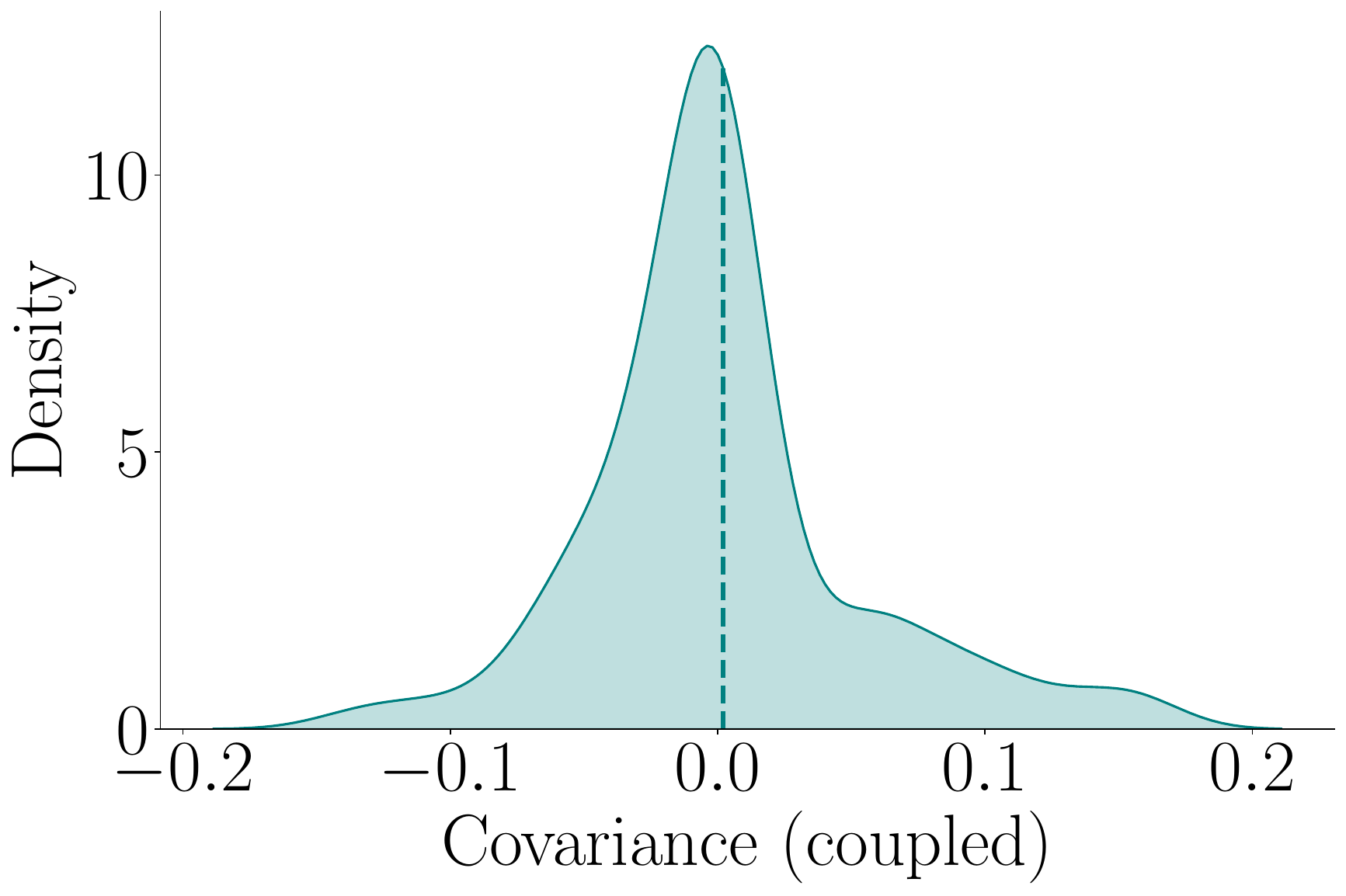} &
    \includegraphics[width=0.23\linewidth]{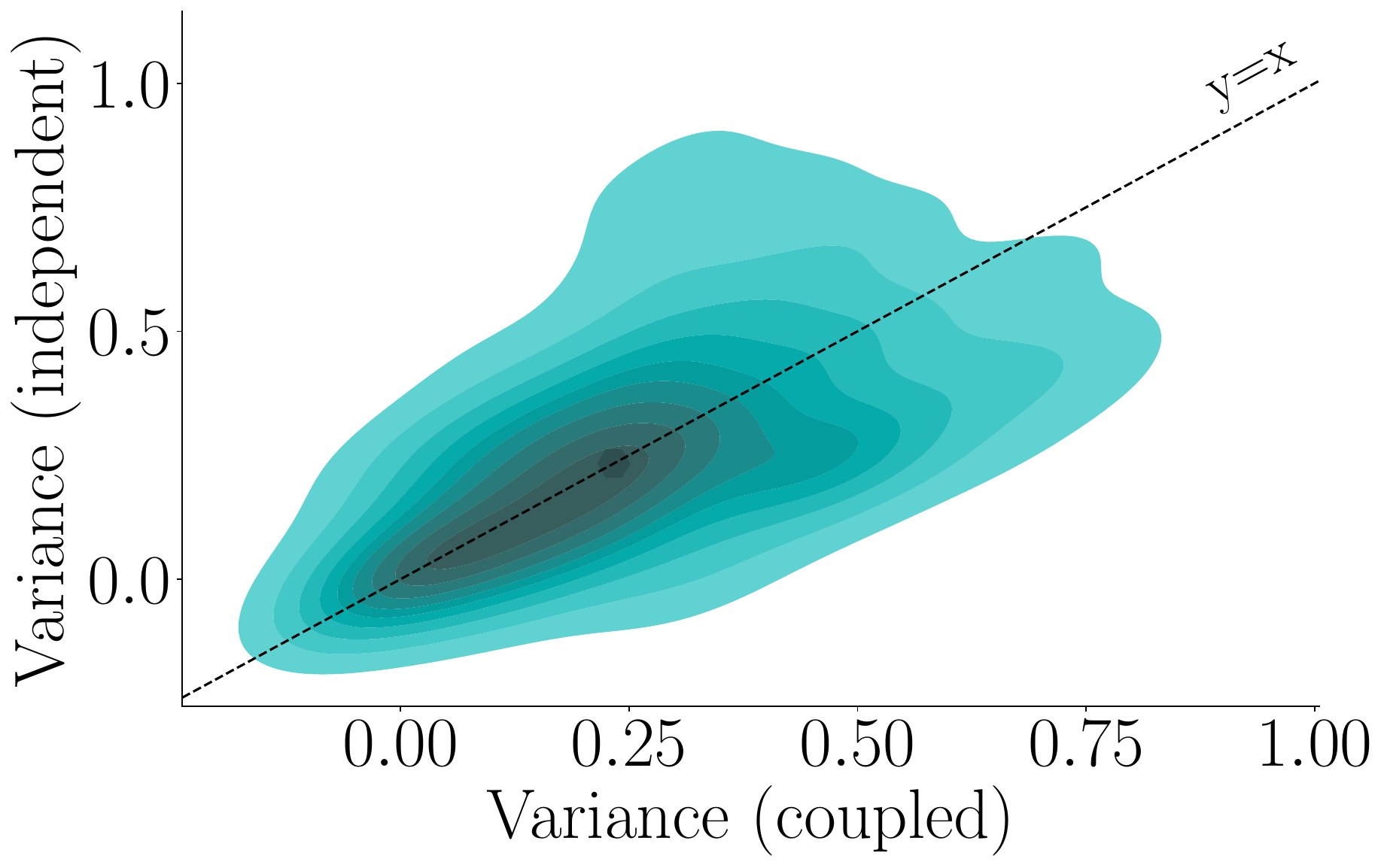} &
    \includegraphics[width=0.23\linewidth]{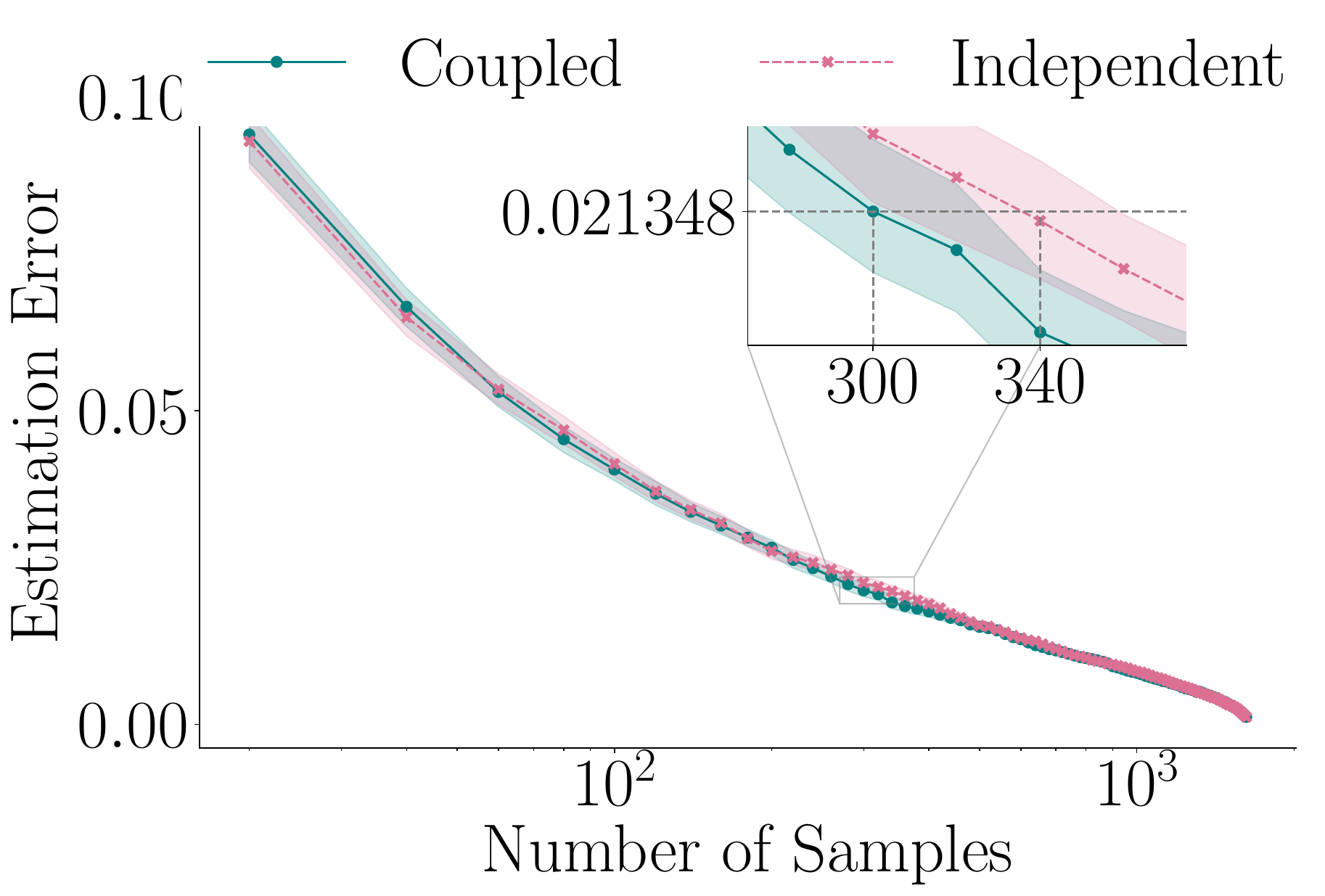} \\ \\
    (a) Score covariance & (b) Variance of the score difference & (c) Estimation error vs. \# samples \\

\end{tabular}
    \caption{\textbf{Comparison between several pairs of LLMs in the \texttt{Llama} family on programming problems from the HumanEval dataset.}
    Panels in column (a) show the kernel density estimate (KDE) of the covariance between the scores of the two LLMs on each problem under coupled generation; the dashed lines correspond to average values. Panels in column (b) show the KDE of the variance of the difference between the scores of the LLMs on each question under coupled and independent generation; the highlighted points correspond to median values. Panels in column (c) show the absolute error in the estimation of the expected difference between the scores of the LLMs against the number of samples; for each point on the x-axis, we perform $1{,}000$ sub-samplings and shaded areas correspond to $95\%$ confidence intervals.}
    \label{fig:human-eval-third-5}
\end{figure}

\xhdr{Results}
Figures~\ref{fig:gsm8k-first-5}-~\ref{fig:human-eval-third-5} show the results for models in the \texttt{Llama} family on the GSM8K and HumanEval datasets. On both datasets, we find that   
 sufficiently similar LLMs, for example, \texttt{8B} and \texttt{bnb-8bit},
have positively correlated scores under coupled generation and thus the variance of the difference in scores is lower under coupled generation than under independent, in agreement with Proposition~\ref{prop:variance} and~\ref{prop:var_gumbel}. These pairs of LLMs also require fewer samples under coupled generation than under independent to achieve equivalent error in the estimation of the expected difference between the scores of the LLMs.\footnote{For each dataset, we compute the error in the estimation of the expected difference in scores as a function of the available sample size as in Section~\ref{sec:mmlu}; here, we use $13{,}190$ samples from  GSM8K and $1{,}680$ samples from HumanEval to estimate (proxies of) the ground truth expected score difference.}
Other pairs of LLMs have  non-positively correlated scores, resulting in similar variance under coupled and independent generation, and requiring the same number of samples to achieve equivalent error in the estimation of the expected difference between the scores of the LLMs. 
%
%
%

\clearpage
\newpage

\clearpage
\newpage

\subsection{Pairwise Comparisons} \label{app:lmsys}
\vspace{-2mm}
%
\begin{figure}[h]
\centering
\includegraphics[width=0.71\linewidth]{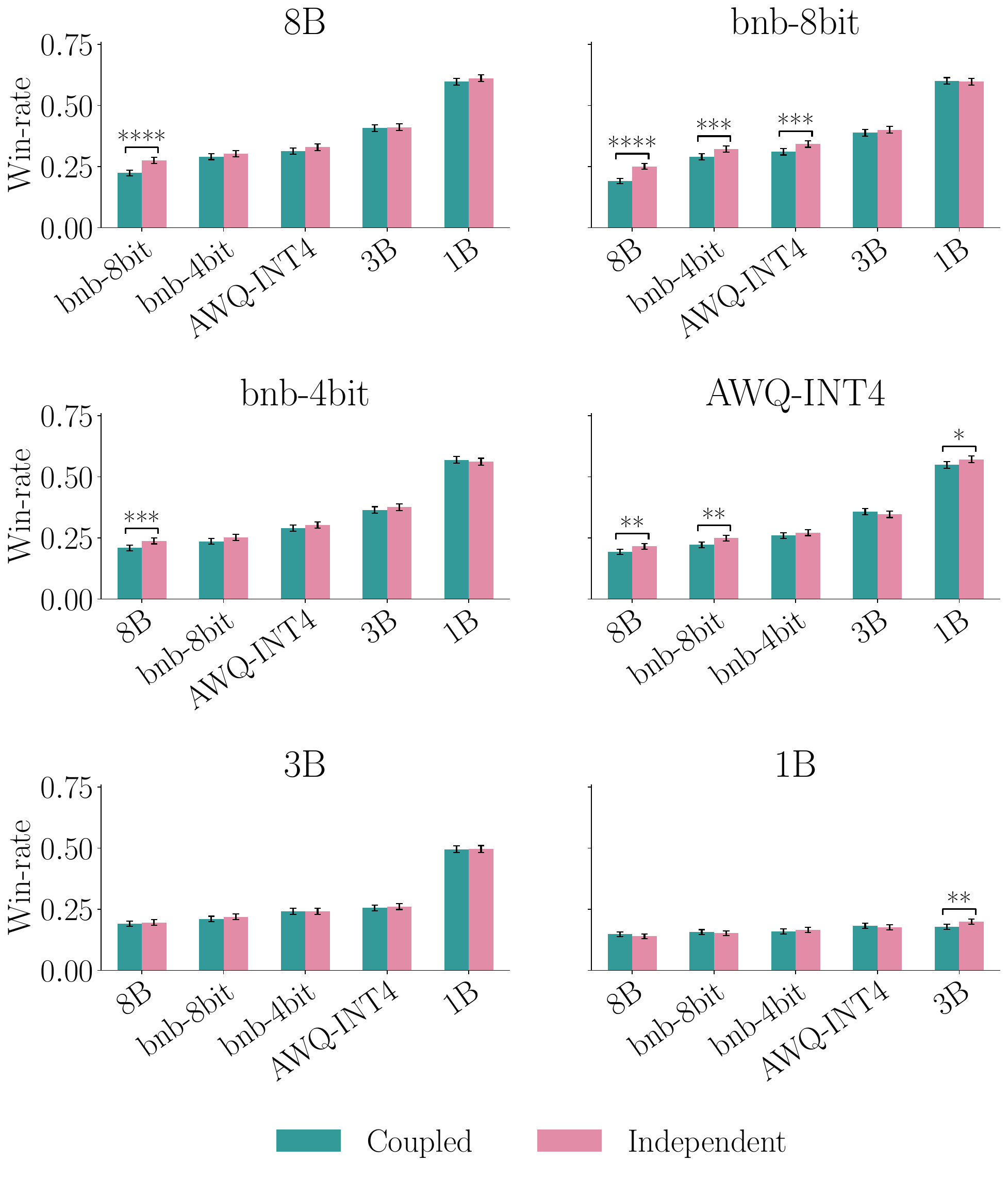}
\caption{
\textbf{Empirical win-rate of each LLM against other LLMs in the \texttt{Llama} family 
on questions from the LMSYS-Chat-1M dataset.} 
Empirical estimate of the win-rate under coupled autoregressive generation as given by Eq.~\ref{eq:coupled-generation-win-rates} and under independent generation generation as given by  Eq.~\ref{eq:independent-generation-win-rates}. 
Each empirical win-rate is computed using pairwise comparisons between the outputs of each LLM and any other LLM over $500$ questions with $10$ (different) random seeds.
The error bars correspond to $95\%$ confidence intervals.
For each pair of empirical win-rates, we conduct a two-tailed test, to test the hypothesis that the empirical win-rates are the same; (\fourstars, \threestars, \twostars, \onestar) indicate $p$-values ($<0.0001$, $<0.001$, $<0.01$, $<0.05$), respectively.
}
\label{fig:lmsys-all-llms}
\end{figure}

\clearpage
\newpage

\section{EXPERIMENTS WITH LLMS IN THE \texttt{Qwen} FAMILY}
\label{app:qwen}
In this section, we experiment with LLMs from the \texttt{Qwen} family.
For brevity, we shorten the names of the LLMs, as listed in Table~\ref{tab:qwen-names}.

\begin{table}[h]
    \centering
    \begin{tabular}{ll} \toprule
         Full name & Shortened name \\ \midrule
         \texttt{Qwen2.5-7B-Instruct} & \texttt{2.5-7B}\\
         \texttt{Qwen2.5-7B-Instruct-AWQ-INT4} & \texttt{2.5-7B-AWQ-INT4}\\
         \texttt{Qwen2.5-7B-Instruct-bnb-4bit} & \texttt{2.5-7B-bnb-4bit} \\ 
         \texttt{Qwen2.5-7B-Instruct-bnb-8bit} &  \texttt{2.5-7B-bnb-8bit}\\ 
         \texttt{Qwen2.5-3B-Instruct} & \texttt{2.5-3B}\\
         \texttt{DistilQwen2.5-3B-Instruct}& \texttt{2.5-3B-distil} \\ 
         \texttt{Qwen2.5-1.5B-Instruct} & \texttt{2.5-1.5B}\\
         \texttt{Qwen3-8B} & \texttt{3-8B}\\
         \bottomrule
    \end{tabular}
    \caption{Full and shortened names of LLMs in the \texttt{Qwen} family.}
    \label{tab:qwen-names}
\end{table}

\subsection{MMLU Dataset}\label{app:qwen-mmlu} 
\begin{figure}[!!h]
\centering
\begin{tabular}{c c c}
     \multicolumn{3}{c}{\texttt{2.5-7B} vs. \texttt{2.5-7B-bnb-8bit}}\\
    \includegraphics[width=0.23\linewidth]{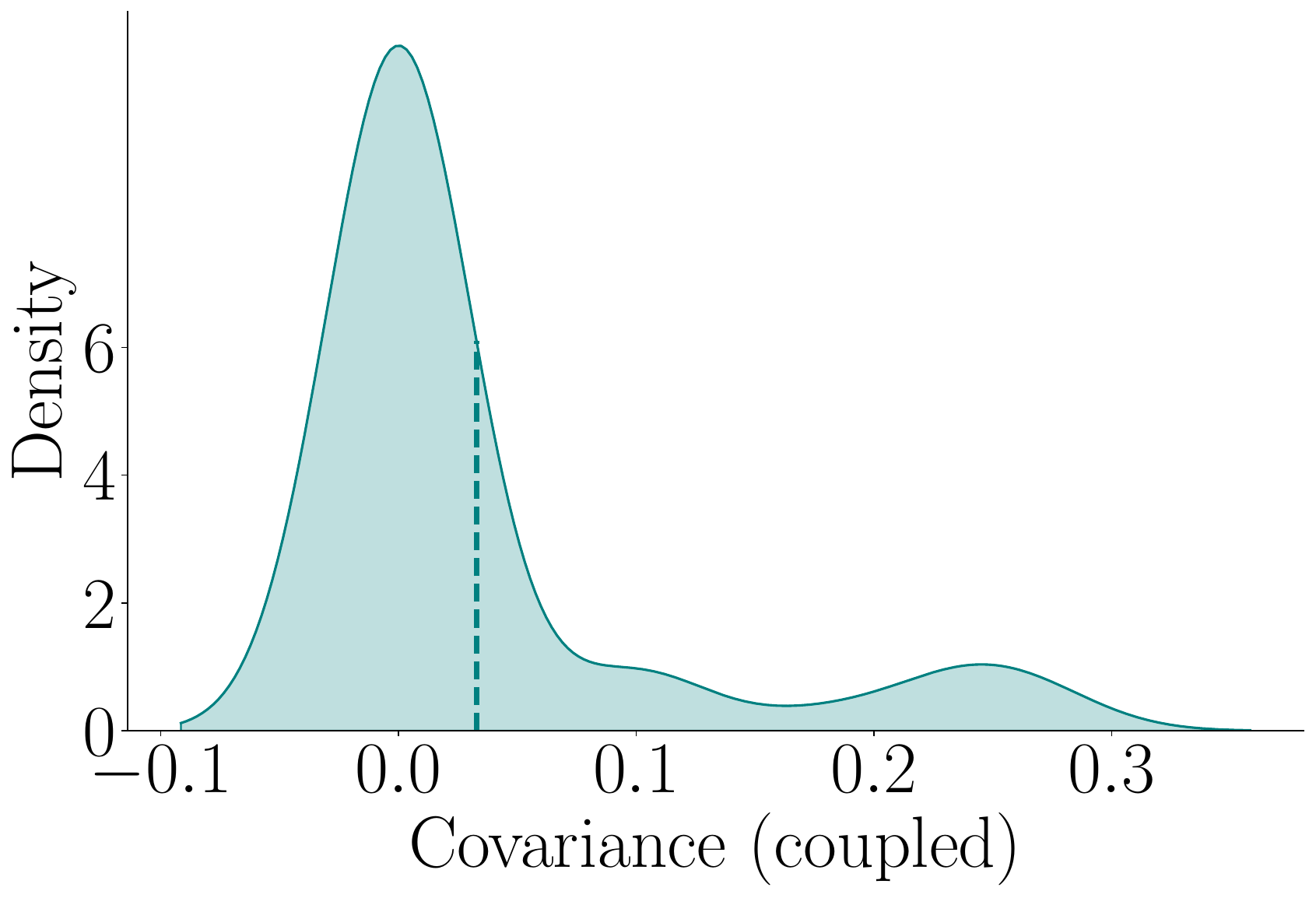} &
    \includegraphics[width=0.23\linewidth]{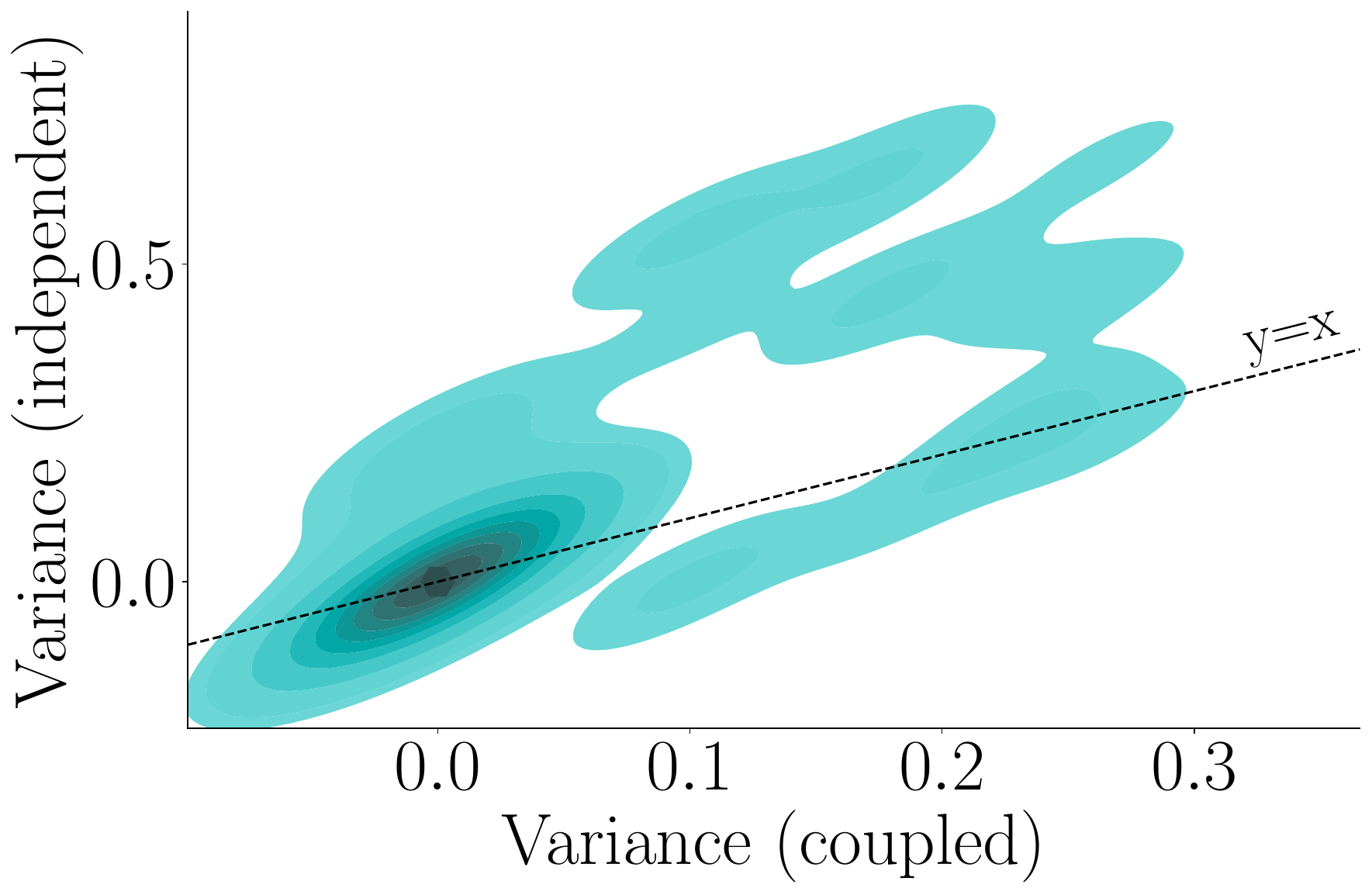} &
    \includegraphics[width=0.23\linewidth]{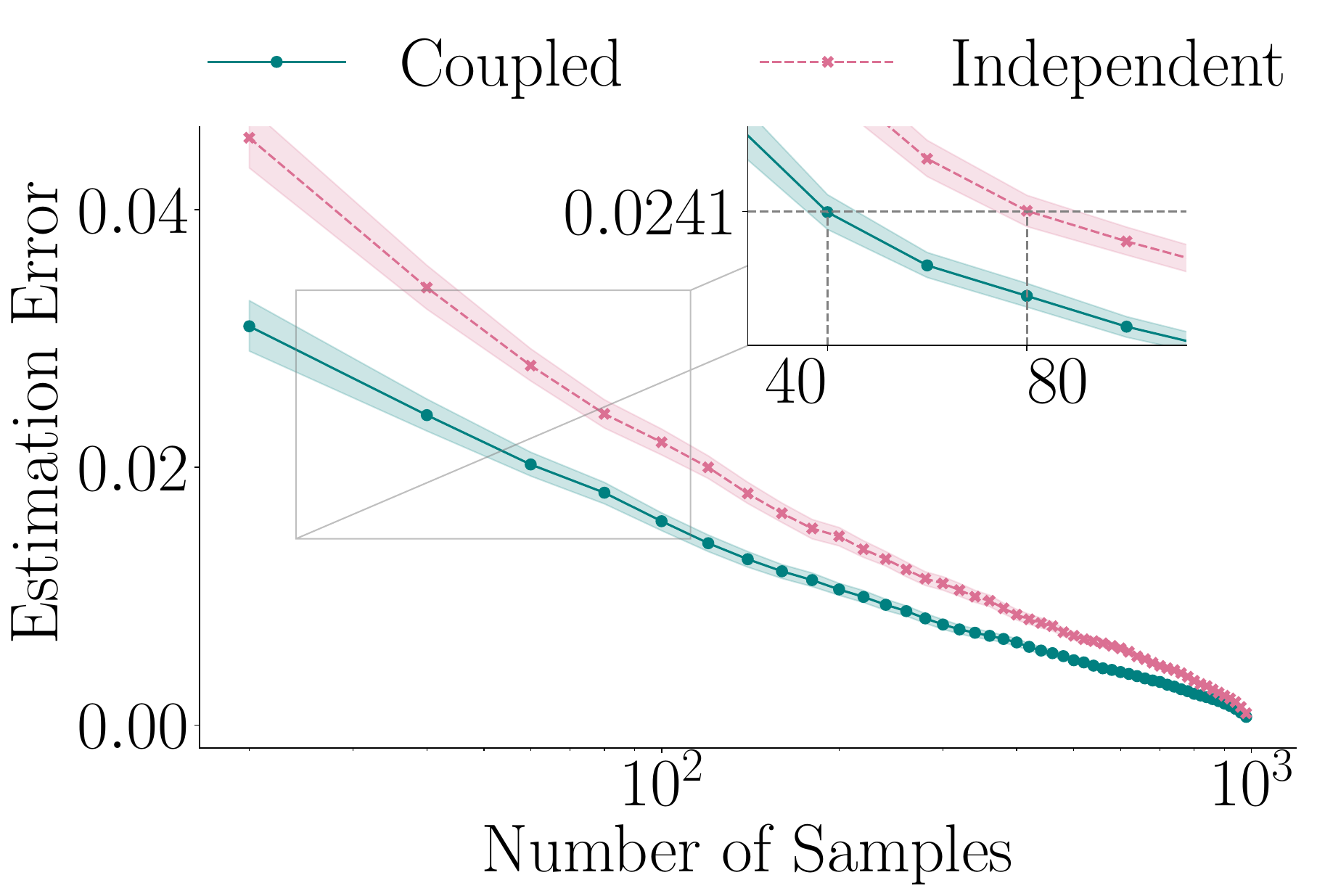} \\ \\
%
    \multicolumn{3}{c}{\texttt{2.5-7B} vs. \texttt{2.5-7B-bnb-4bit}}\\
    \includegraphics[width=0.23\linewidth]{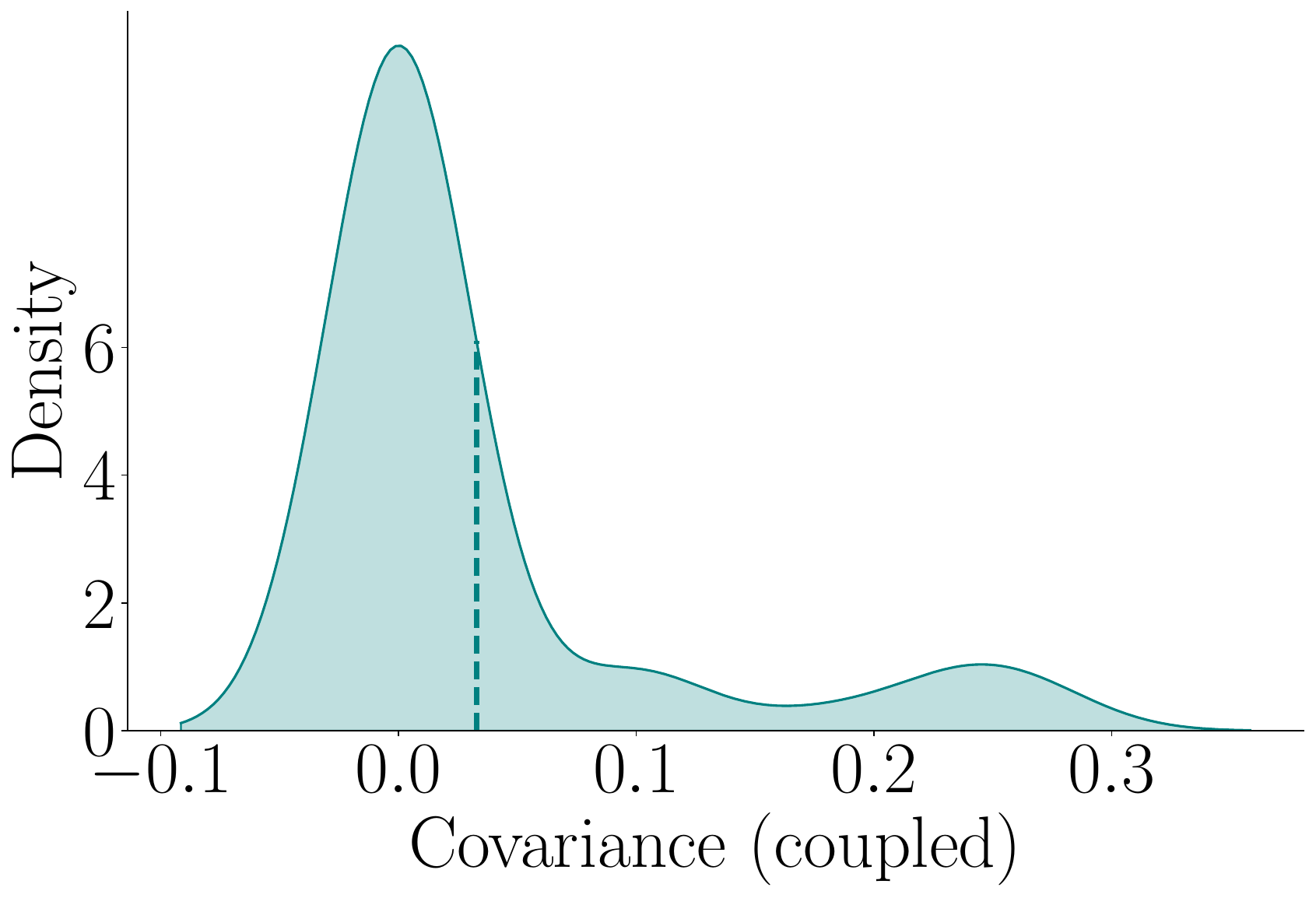} &
    \includegraphics[width=0.23\linewidth]{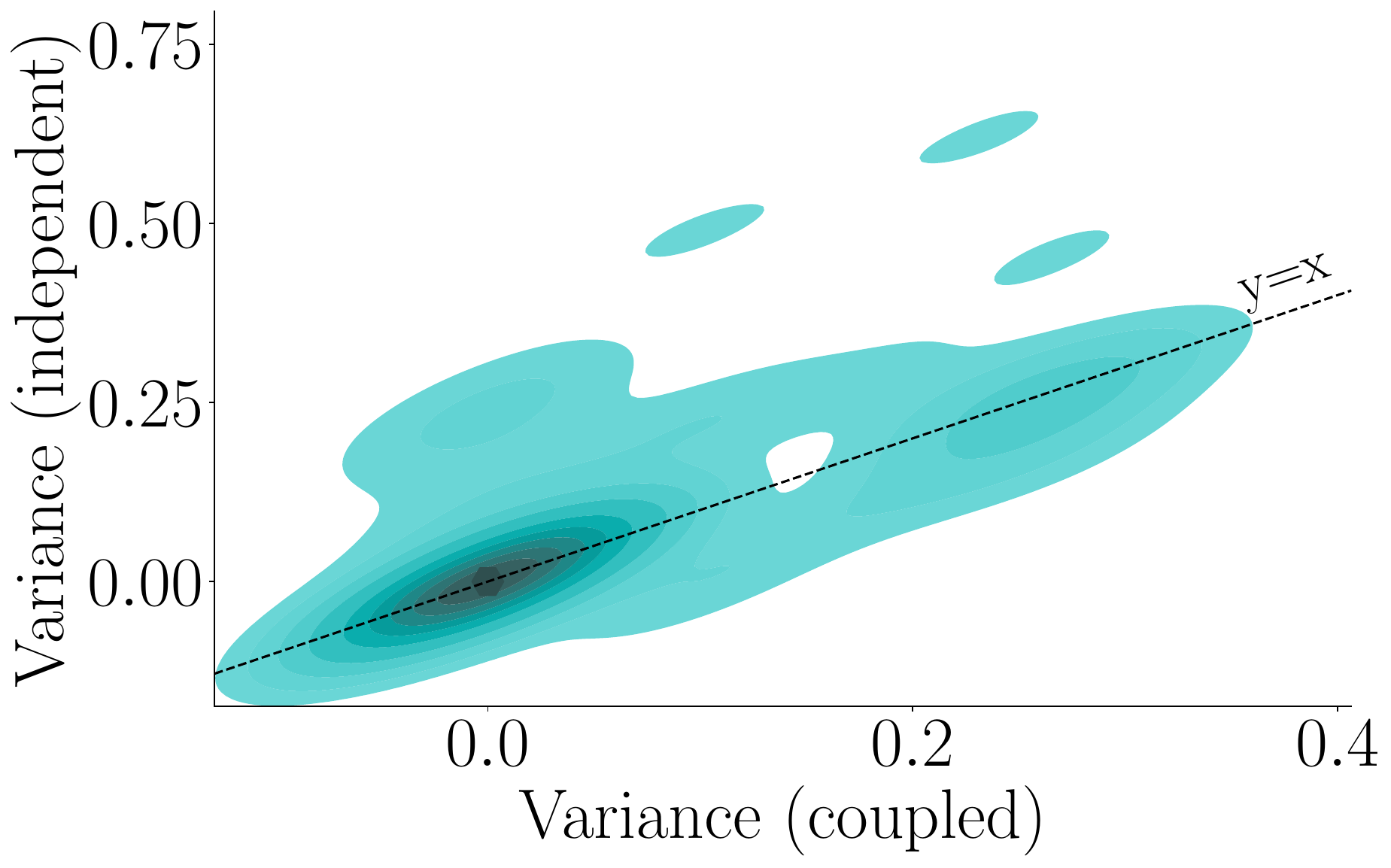} &
    \includegraphics[width=0.23\linewidth]{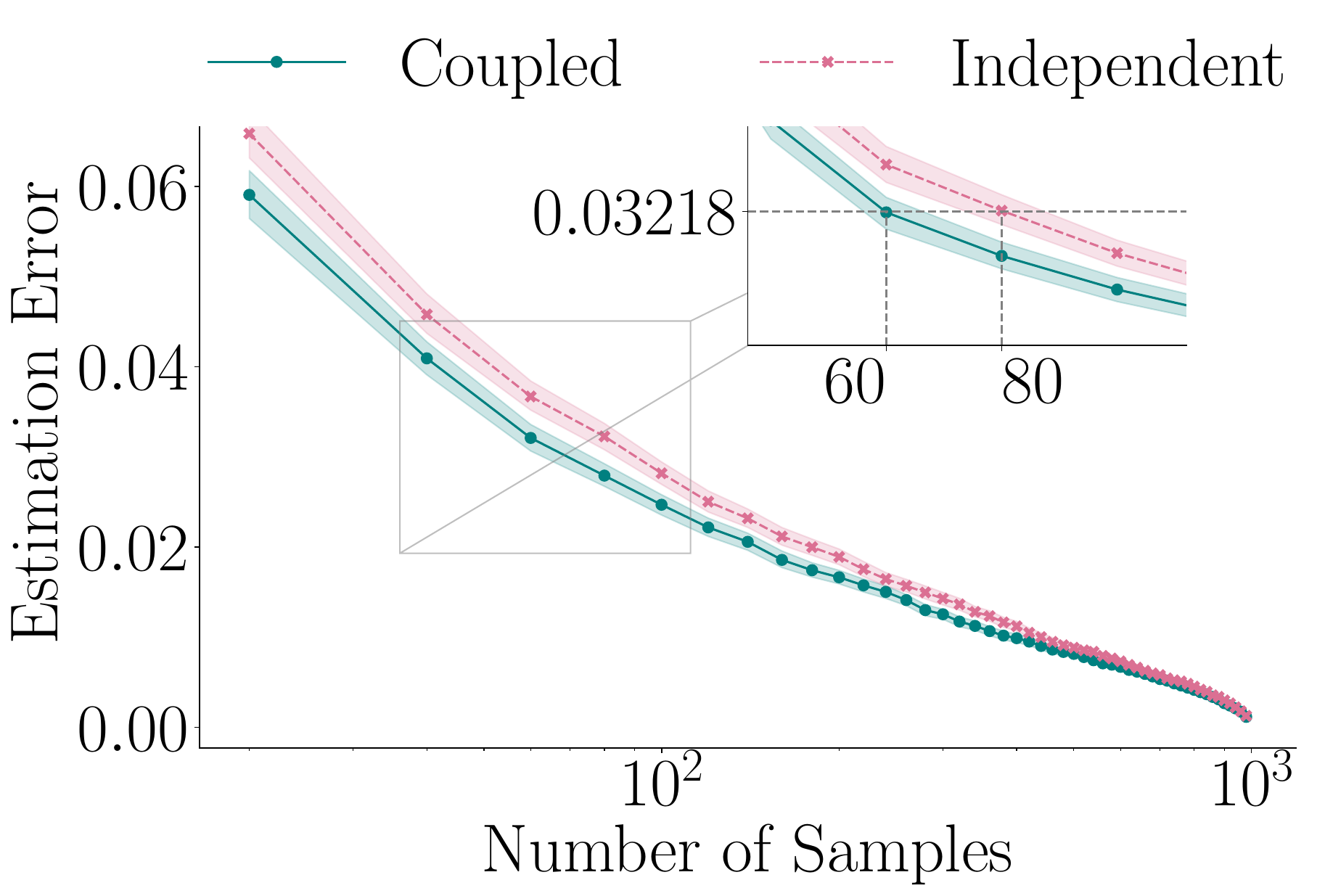} \\ \\
%
     \multicolumn{3}{c}{\texttt{3-8B} vs. \texttt{2.5-7B}}\\
    \includegraphics[width=0.23\linewidth]{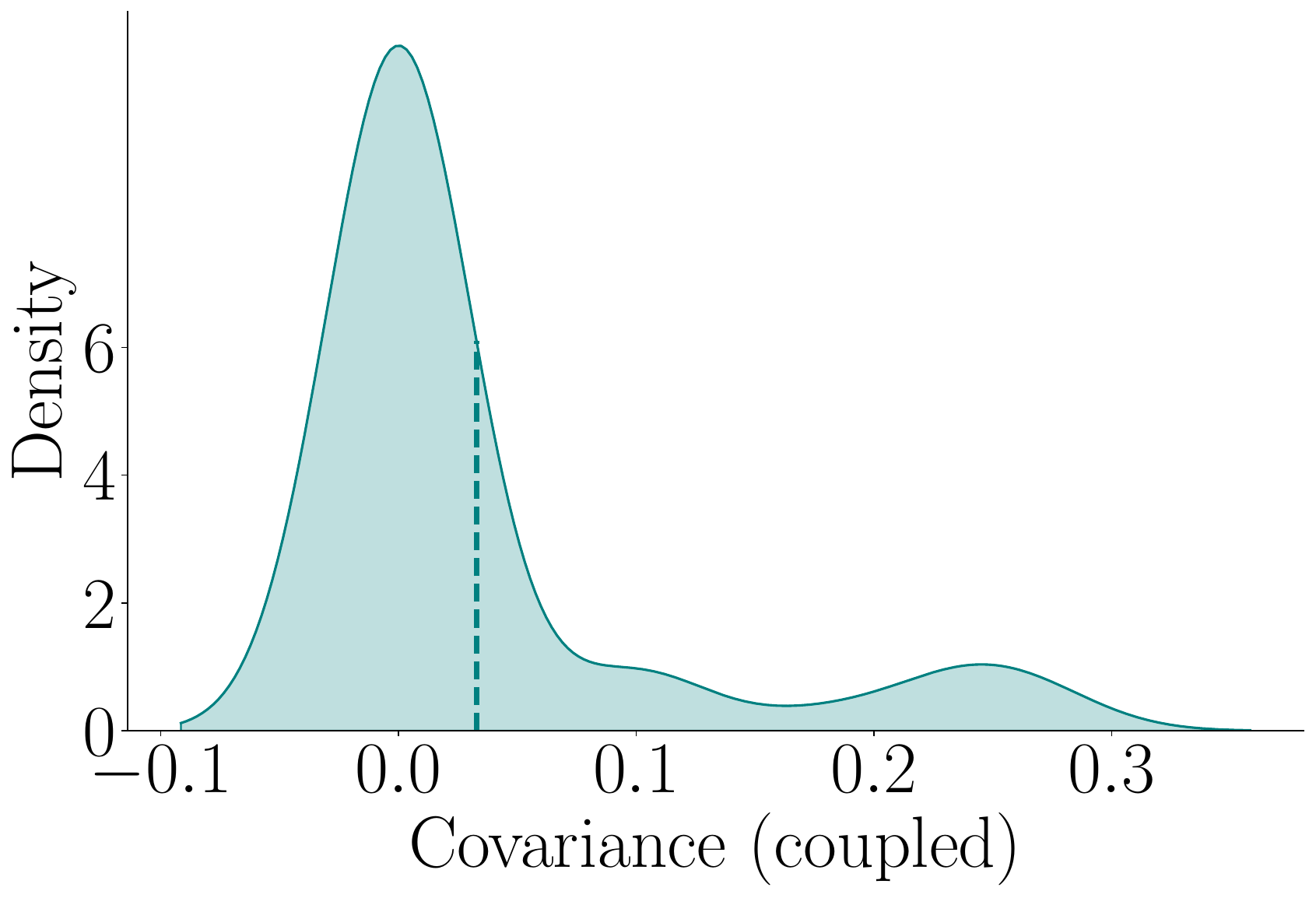} &
    \includegraphics[width=0.23\linewidth]{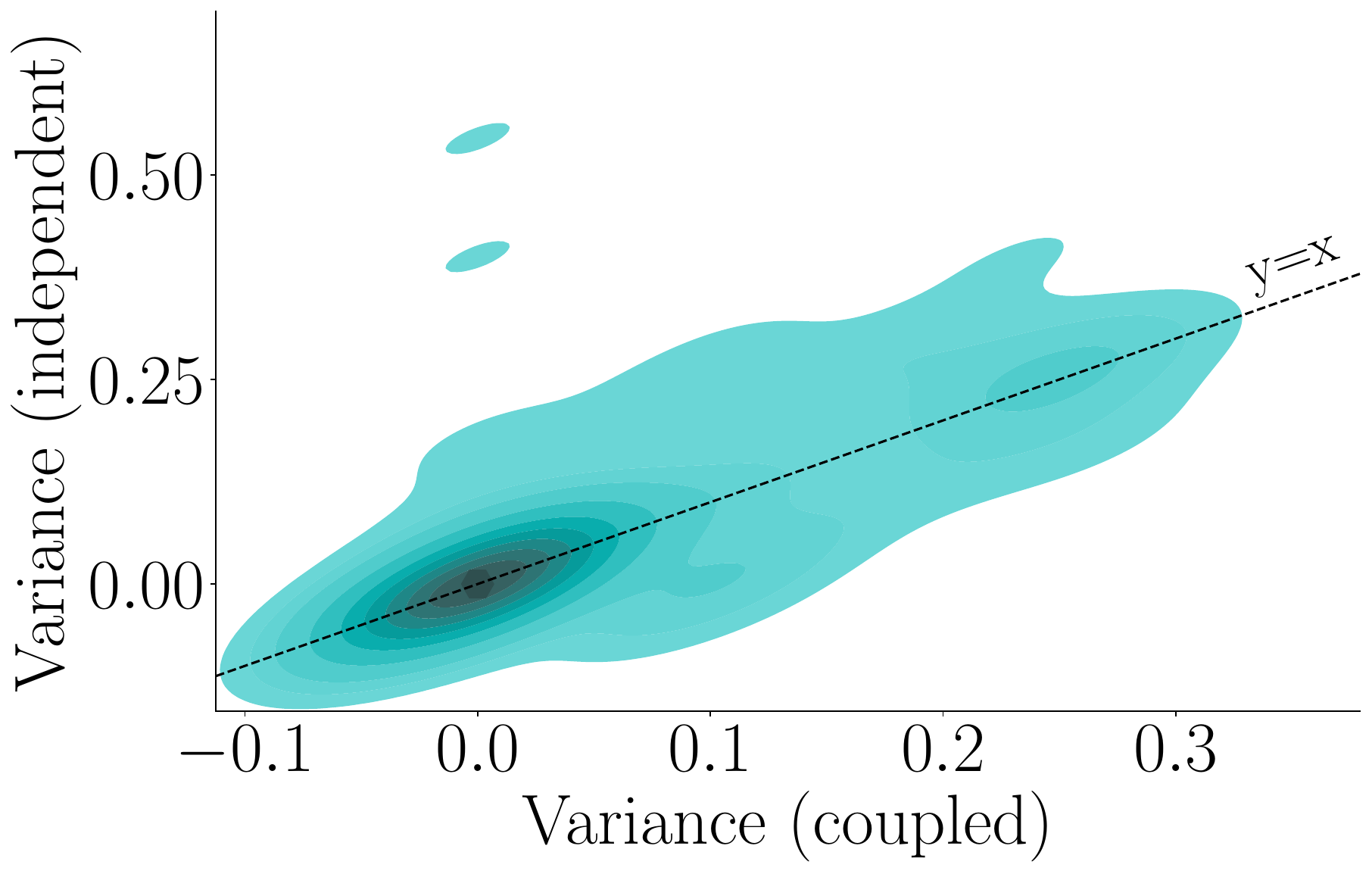} &
    \includegraphics[width=0.23\linewidth]{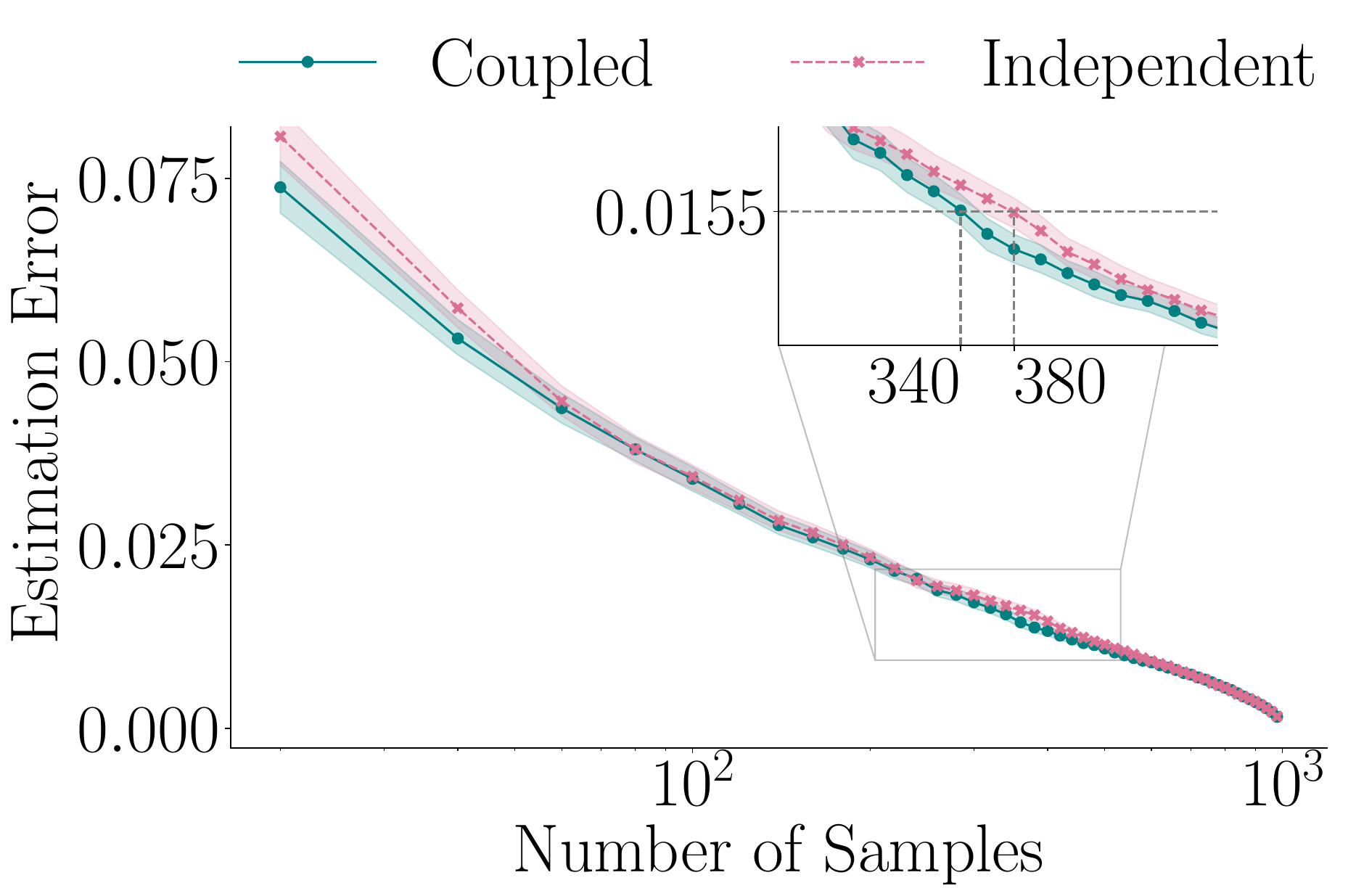} \\ \\
%
  \multicolumn{3}{c}{\texttt{2.5-3B} vs. \texttt{2.5-7B}}\\
    \includegraphics[width=0.23\linewidth]{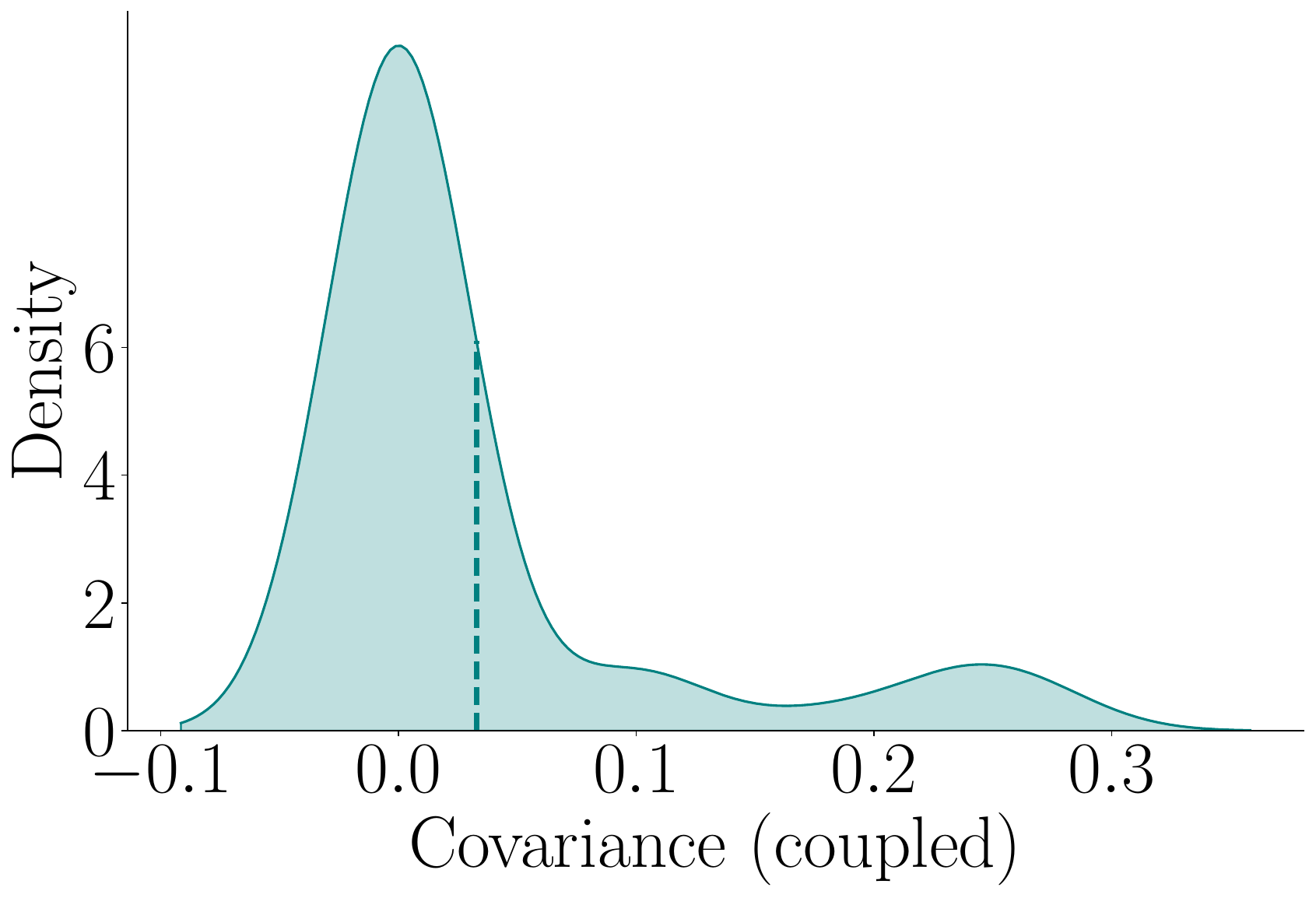} &
    \includegraphics[width=0.23\linewidth]{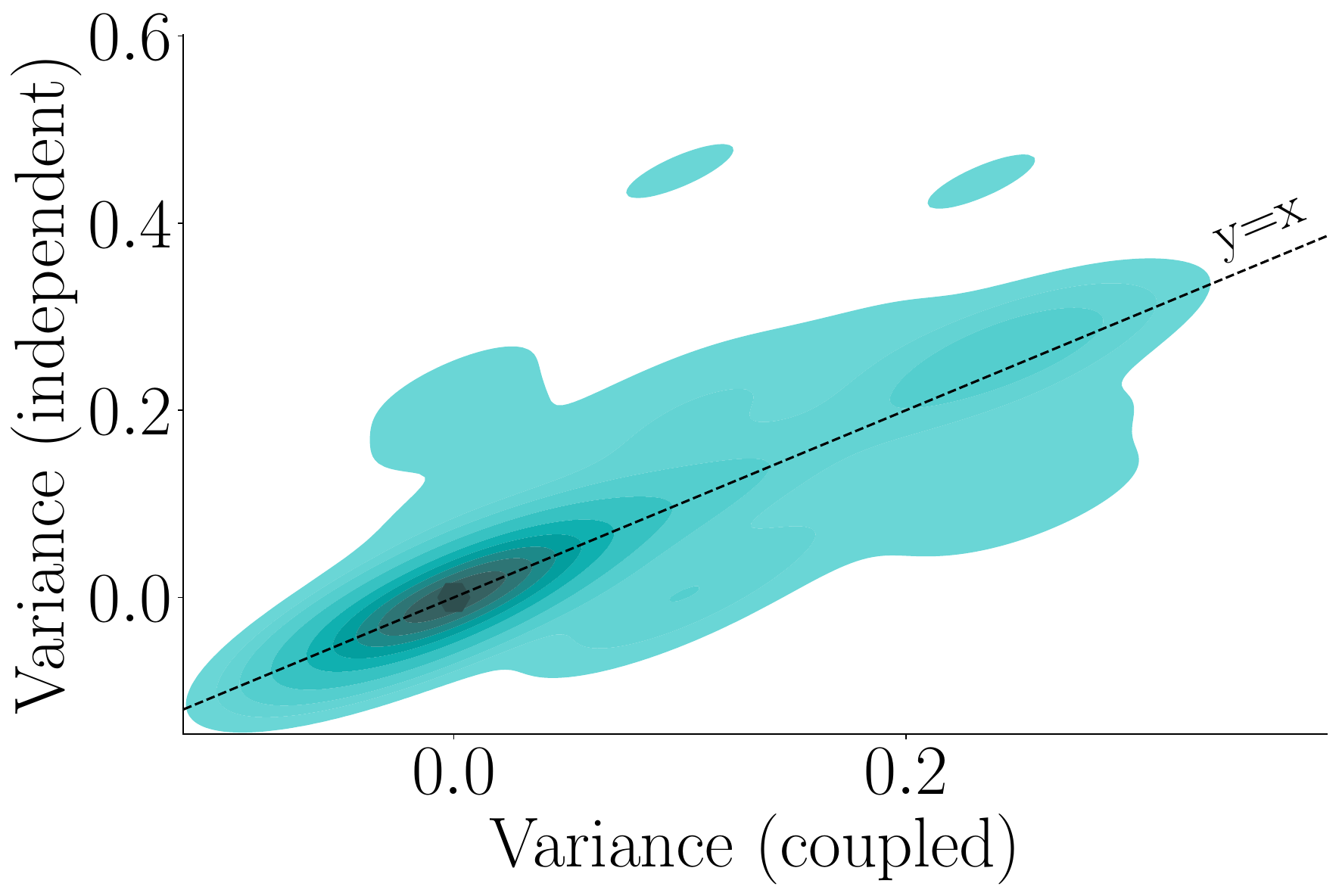} &
    \includegraphics[width=0.23\linewidth]{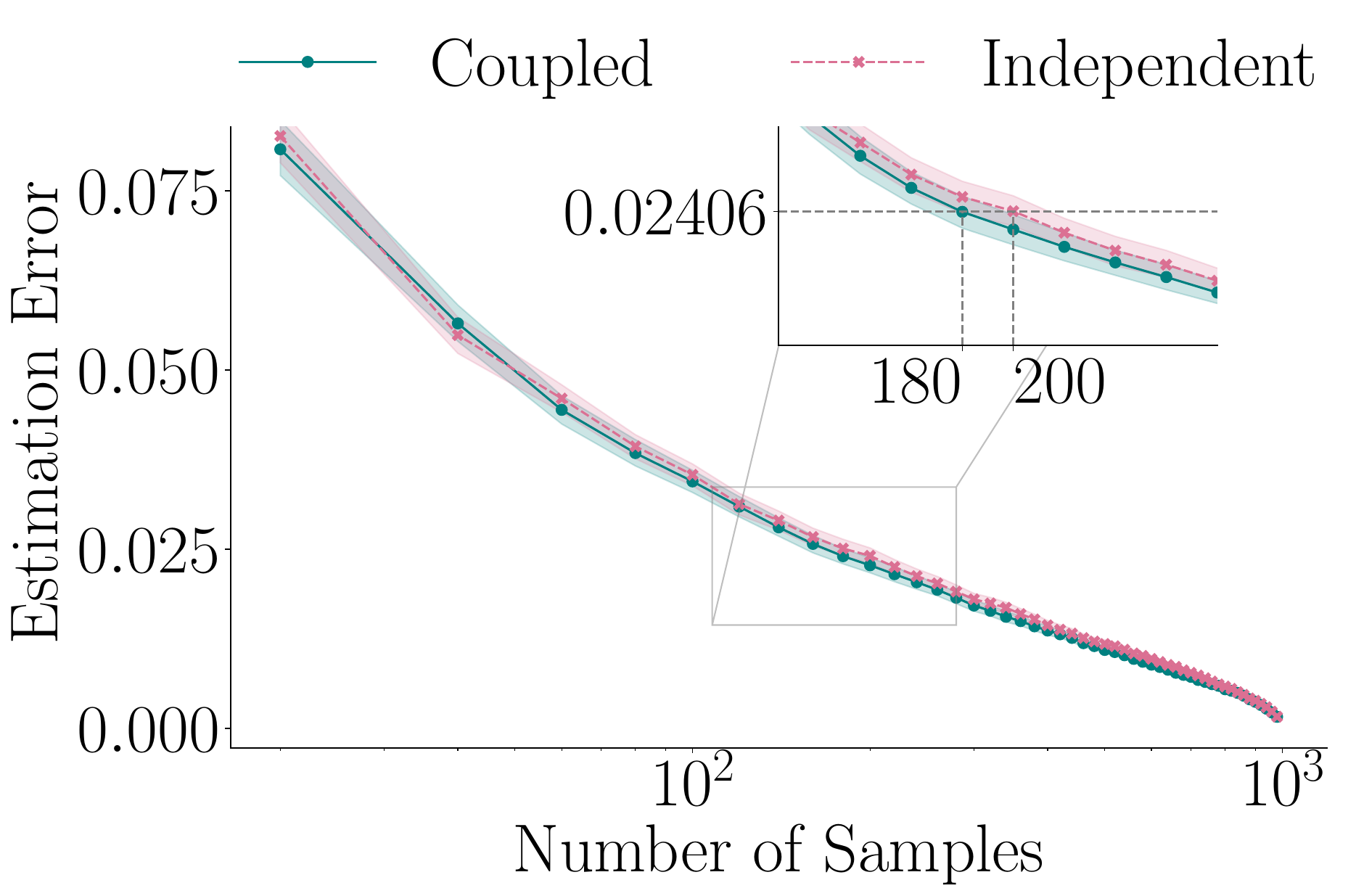} \\ \\

    (a) Score covariance & (b) Variance of the score difference & (c) Estimation error vs. \# samples \\ 
\end{tabular}
    \caption{\textbf{Comparison between four pairs of LLMs in the \texttt{Qwen} family on multiple-choice questions from the ``college computer science'' knowledge area of the MMLU dataset.}
    Panels in column (a) show the kernel density estimate (KDE) of the covariance between the scores of the two LLMs on each question under coupled generation; the dashed lines correspond to average values. Panels in column (b) show the KDE of the variance of the difference between the scores of the LLMs on each question under coupled and independent generation; the highlighted points correspond to median values. Panels in column (c) show the absolute error in the estimation of the expected difference between the scores of the LLMs against the number of samples; for each point on the x-axis, we perform $1{,}000$ sub-samplings and shaded areas correspond to $95\%$ confidence intervals.}
    \label{fig:mmlu-qwen-first-4}
\end{figure}

\begin{figure}[h]
\centering
\begin{tabular}{c c c}
    \multicolumn{3}{c}{\texttt{2.5-1.5B} vs. \texttt{2.5-7B}}\\
    \includegraphics[width=0.23\linewidth]{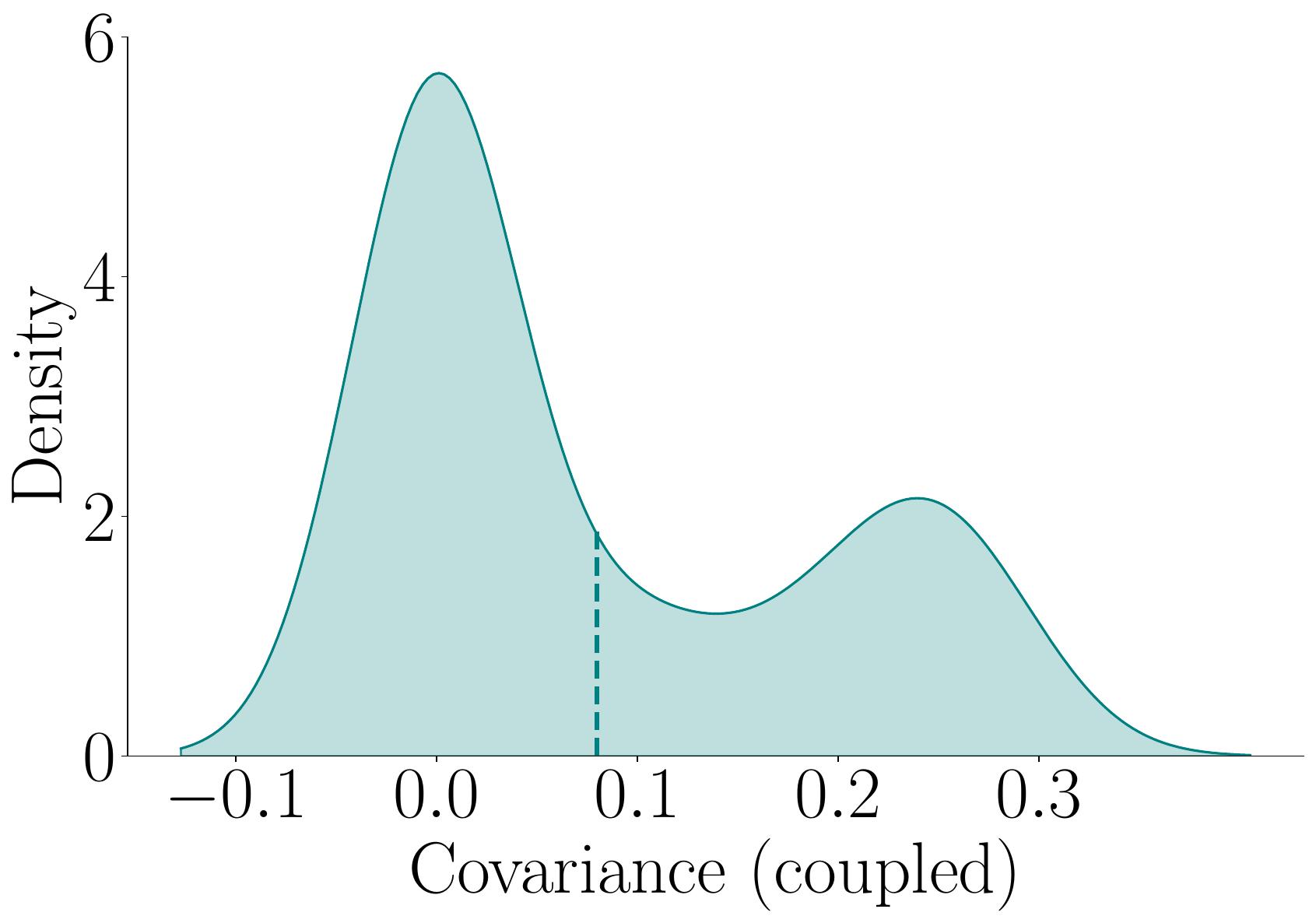} &
    \includegraphics[width=0.23\linewidth]{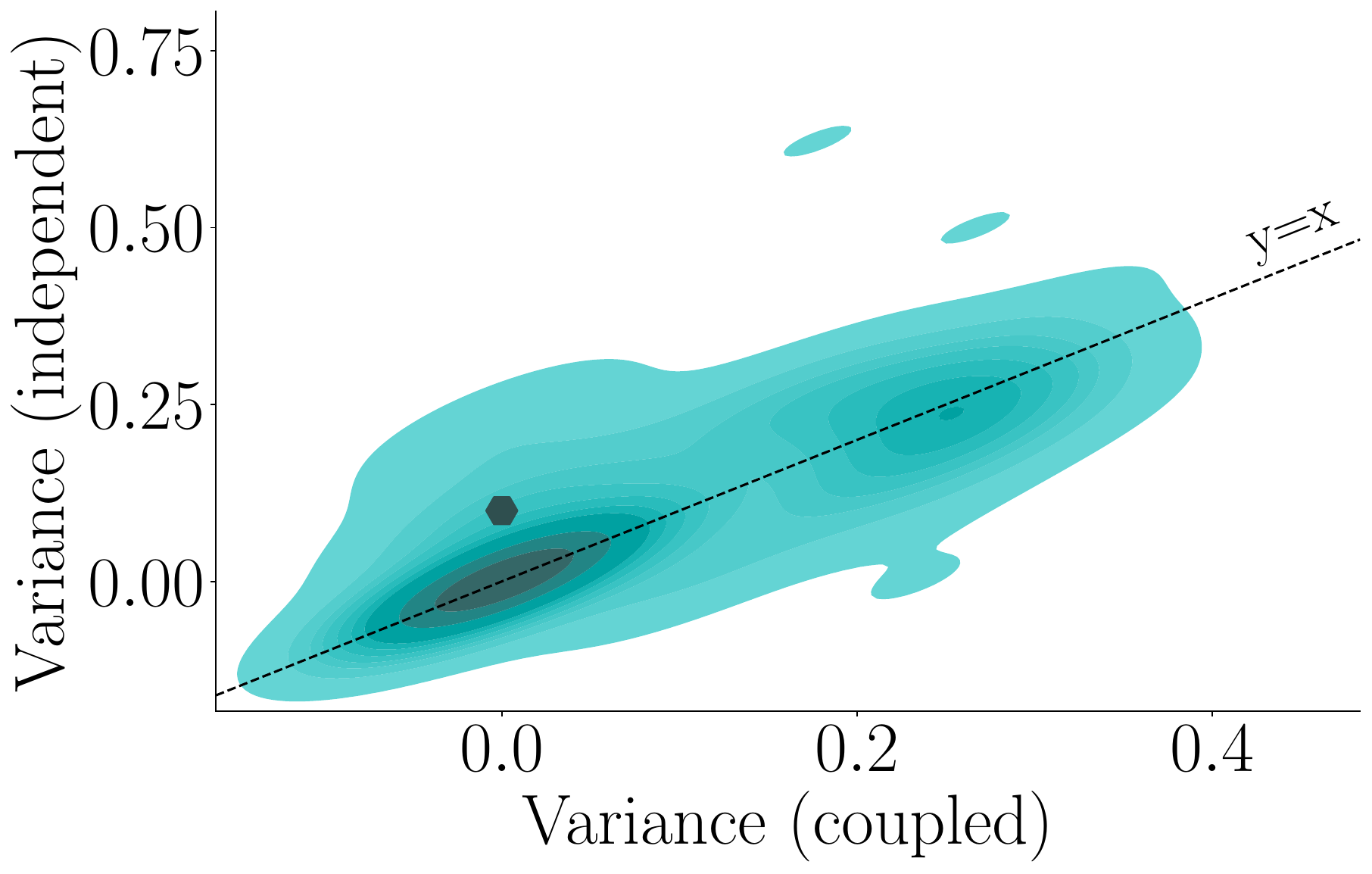} &
    \includegraphics[width=0.23\linewidth]{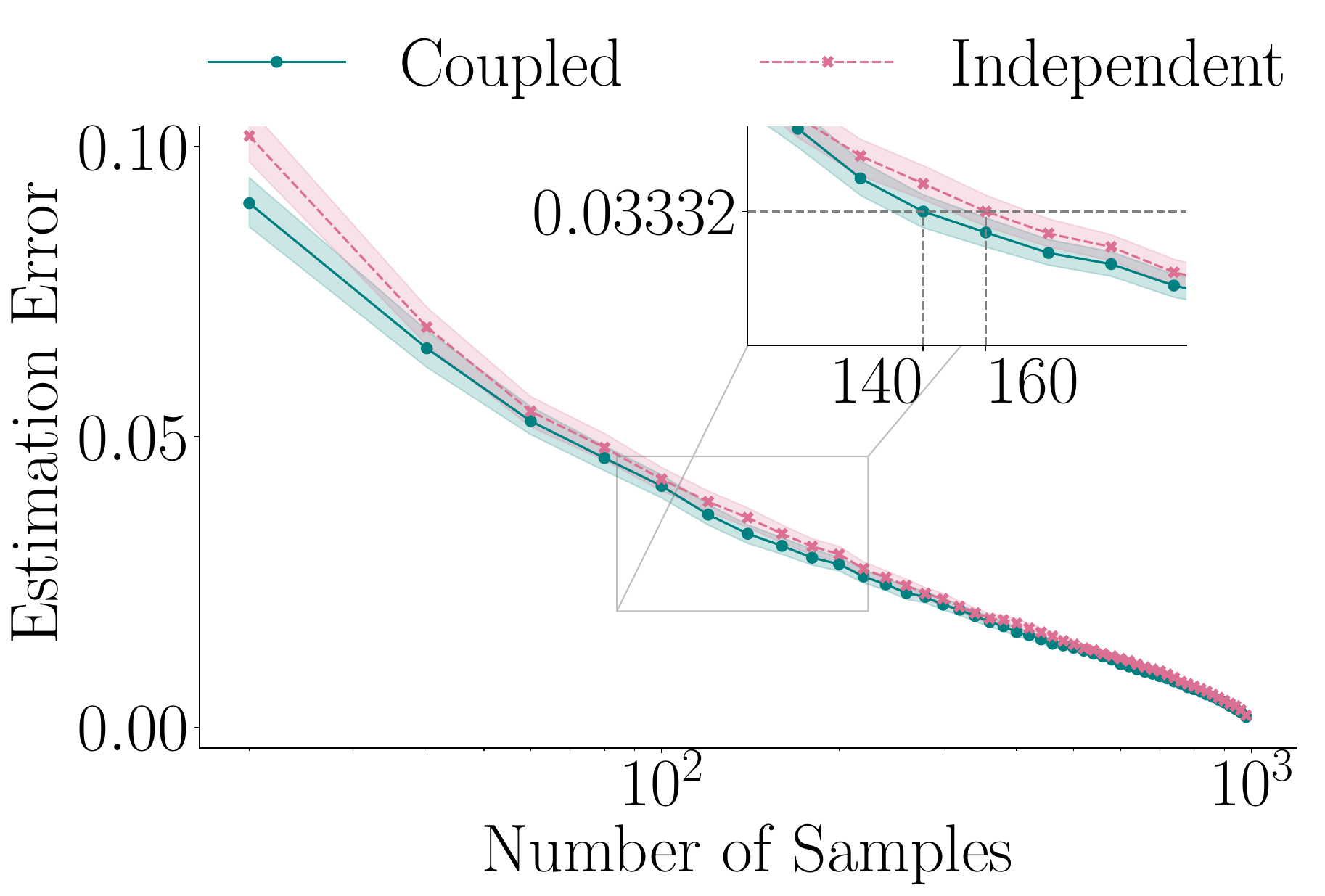} \\ \\
    \multicolumn{3}{c}{\texttt{2.5-3B-distil} vs. \texttt{2.5-3B}}\\
    \includegraphics[width=0.23\linewidth]{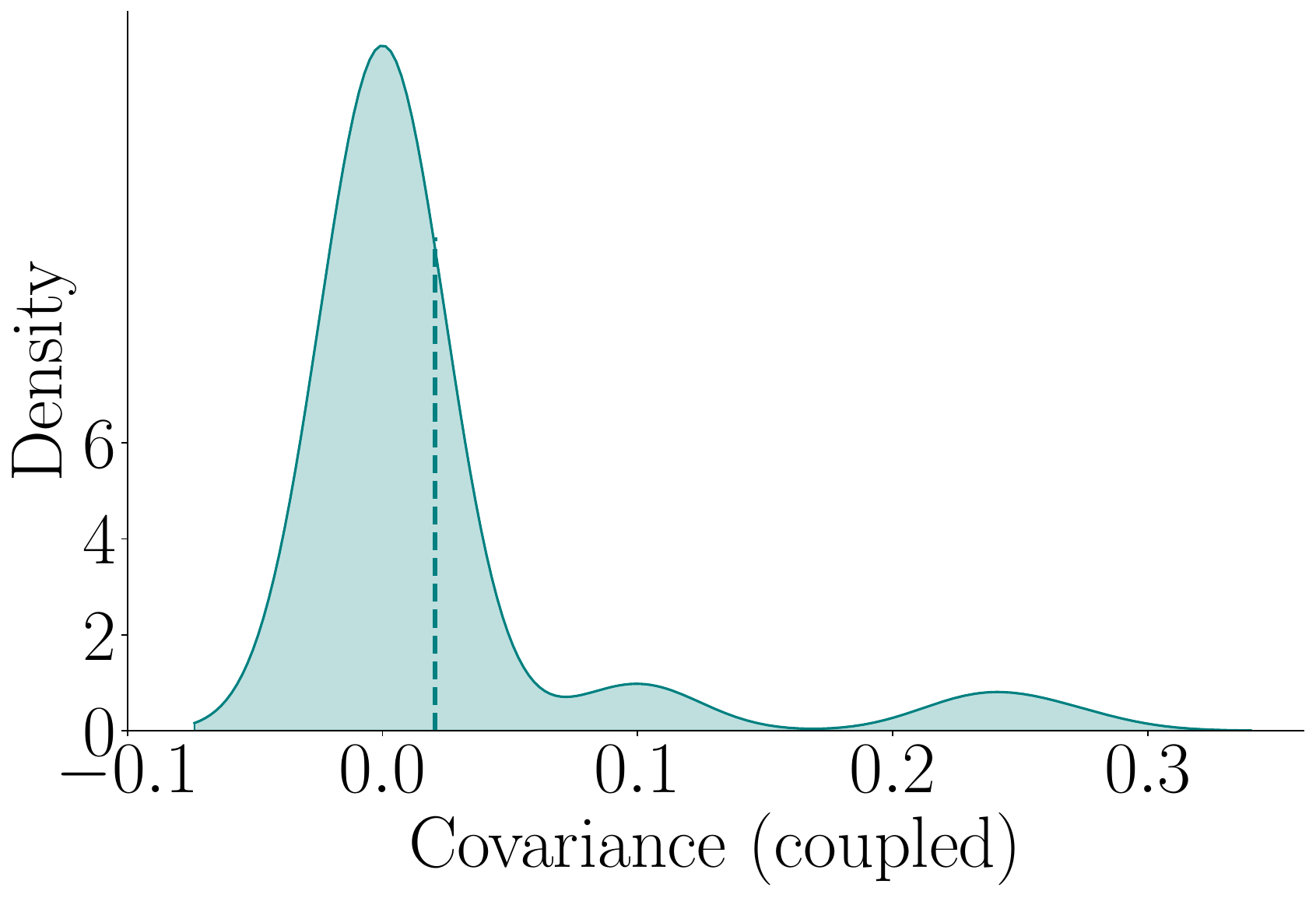} &
    \includegraphics[width=0.23\linewidth]{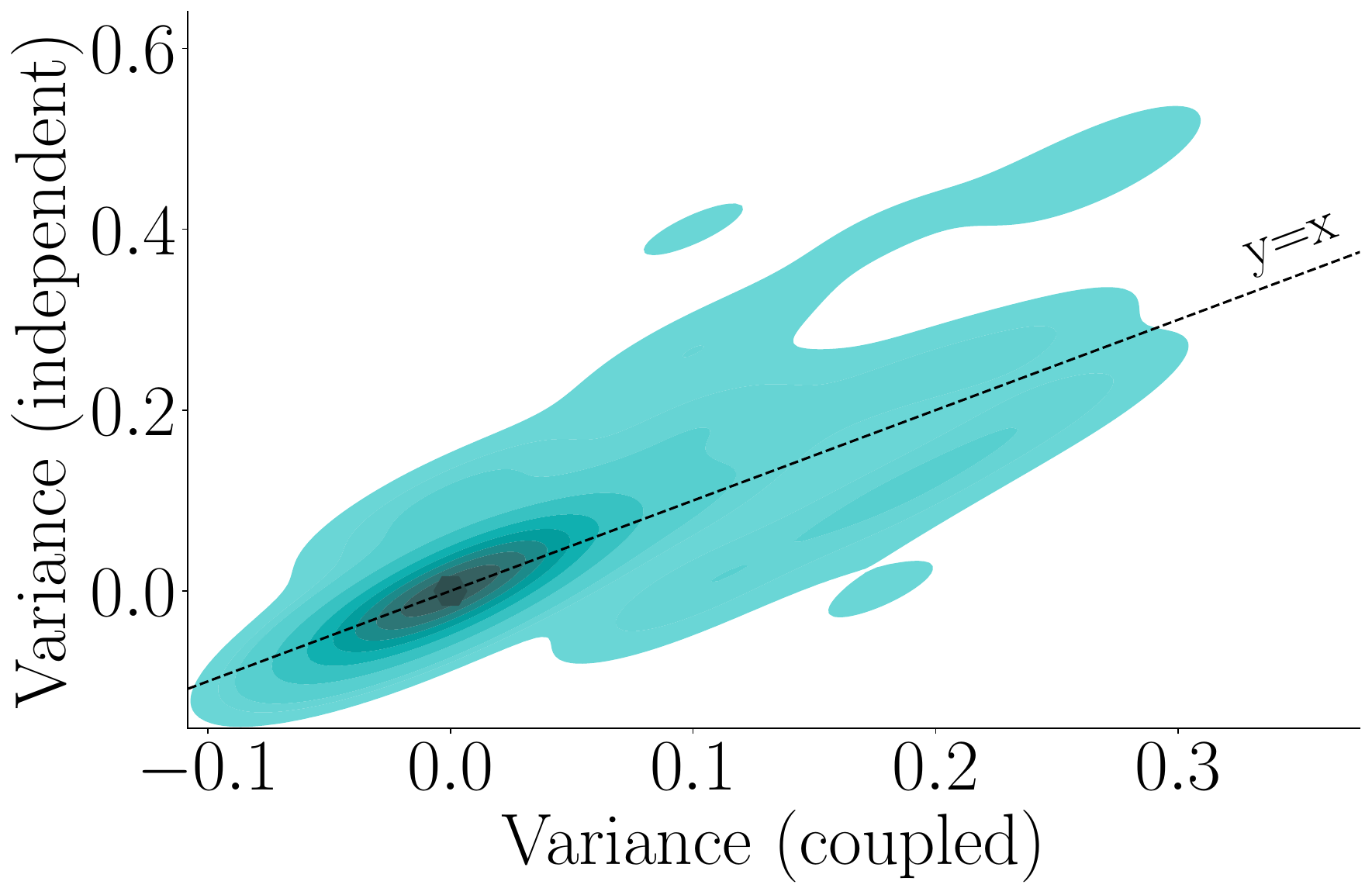} &
    \includegraphics[width=0.23\linewidth]{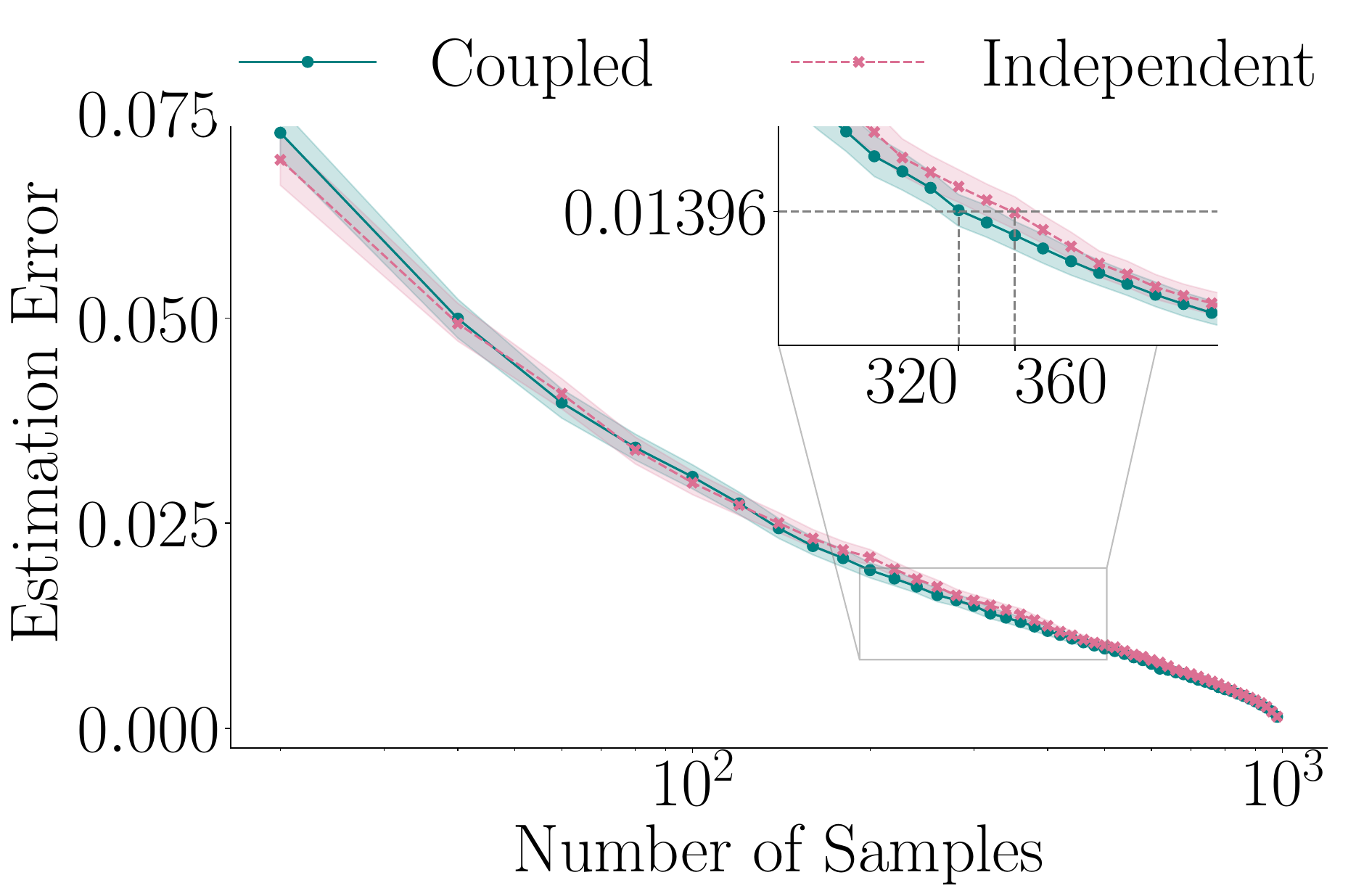} \\ \\
     \multicolumn{3}{c}{\texttt{2.5-3B} vs. \texttt{2.5-7B-bnb-4bit}}\\
    \includegraphics[width=0.23\linewidth]{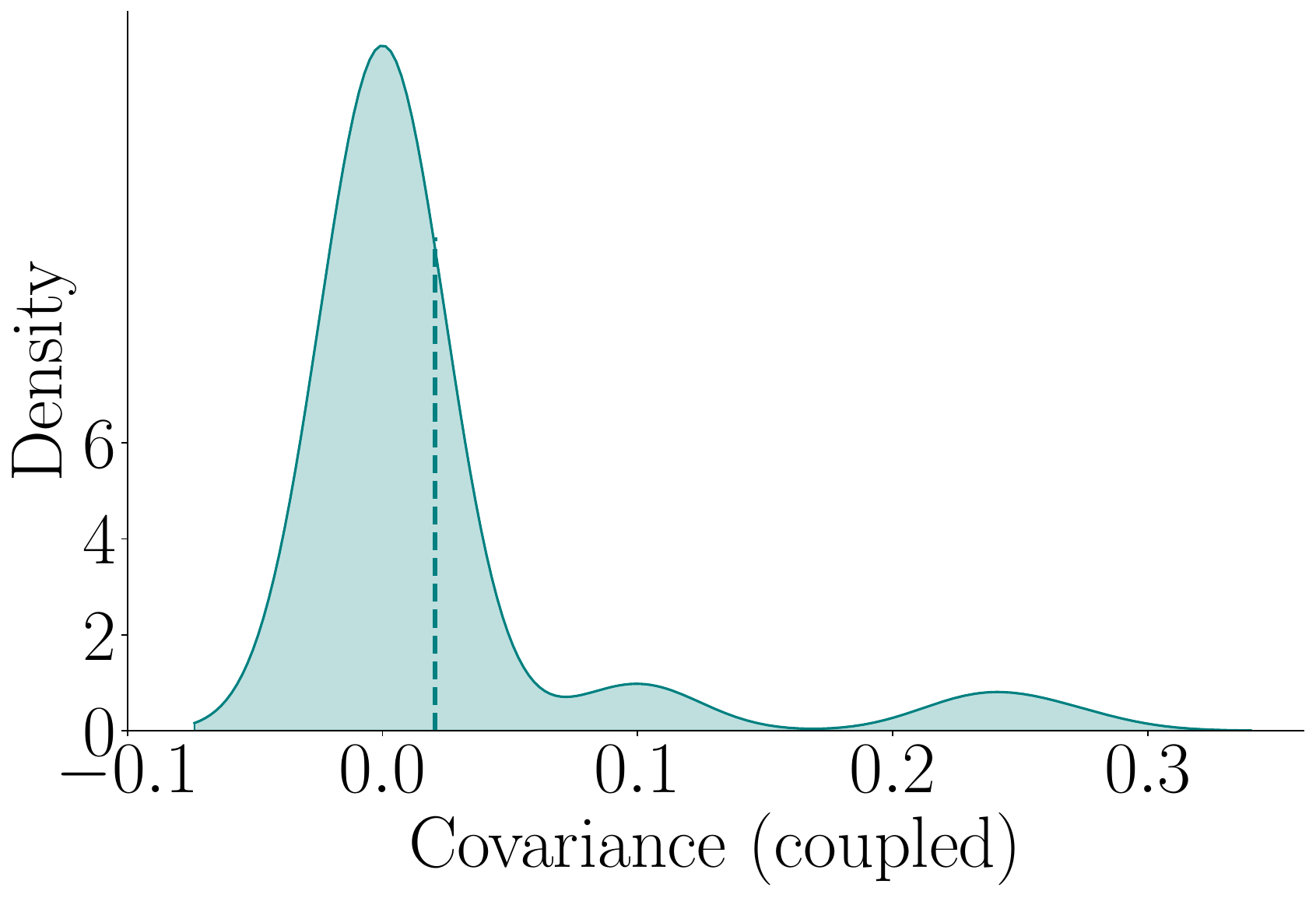} &
    \includegraphics[width=0.23\linewidth]{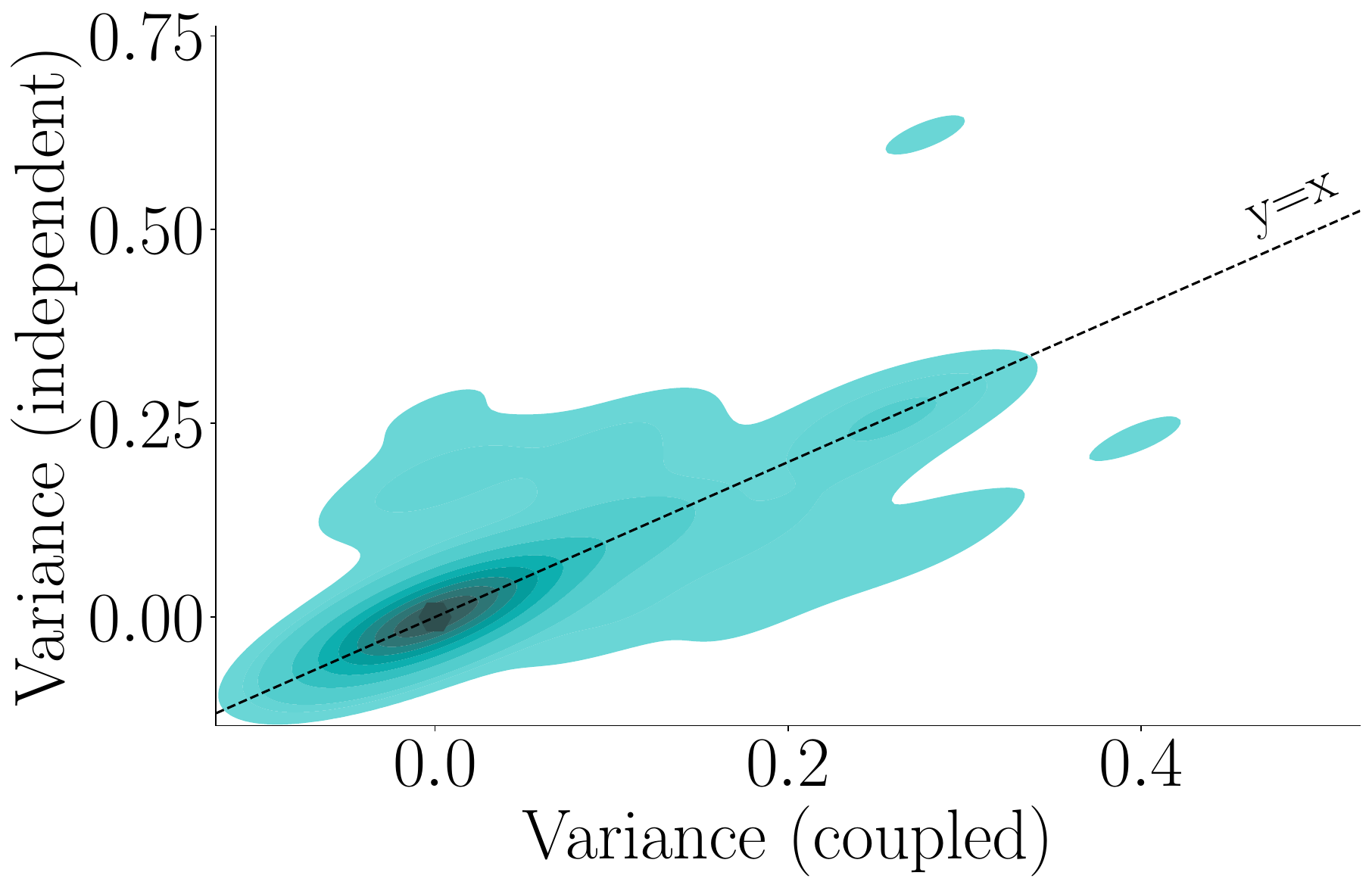} &
    \includegraphics[width=0.23\linewidth]{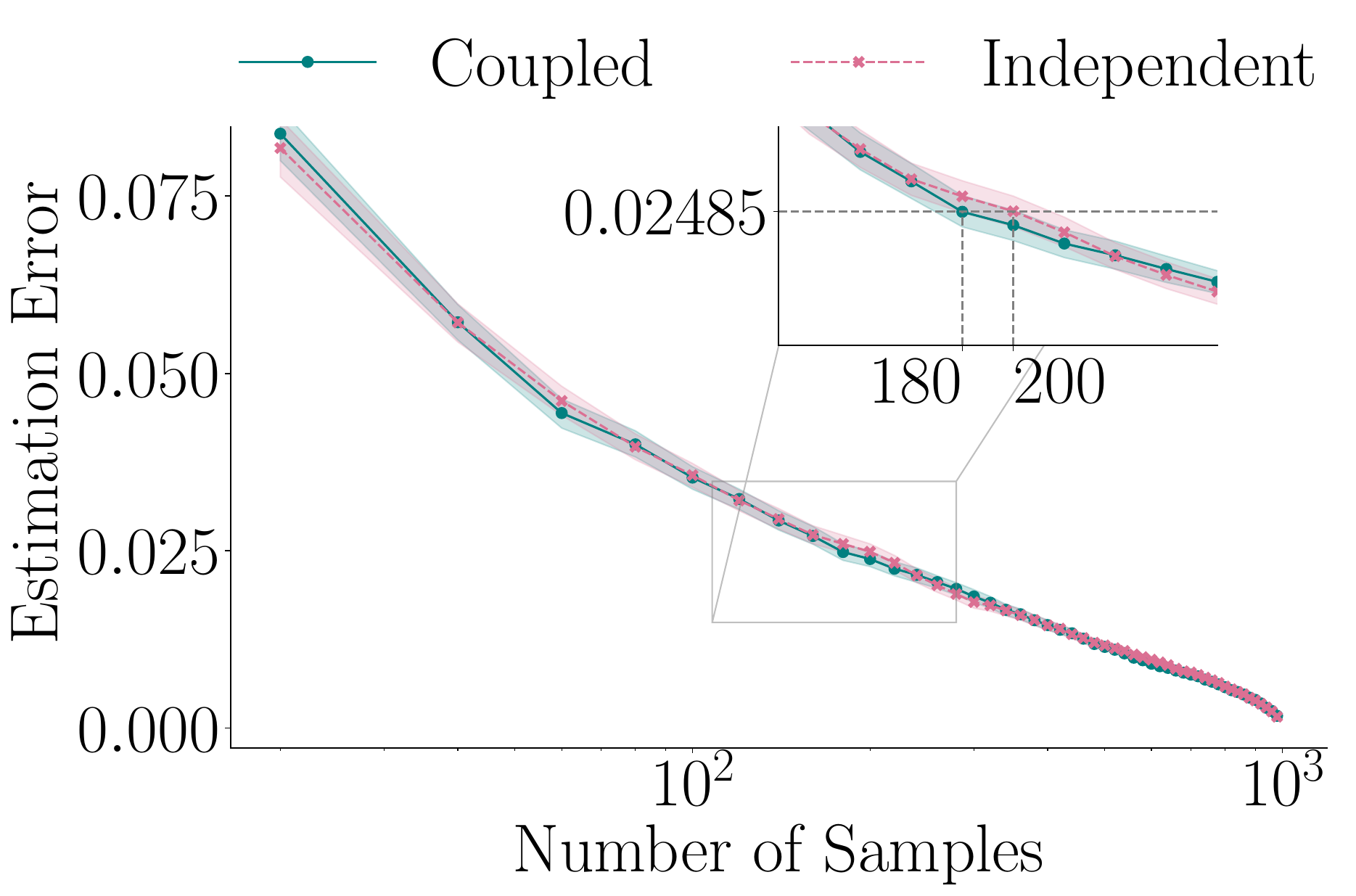} \\ \\
 \multicolumn{3}{c}{\texttt{3-8B} vs. \texttt{2.5-3B}}\\
    \includegraphics[width=0.23\linewidth]{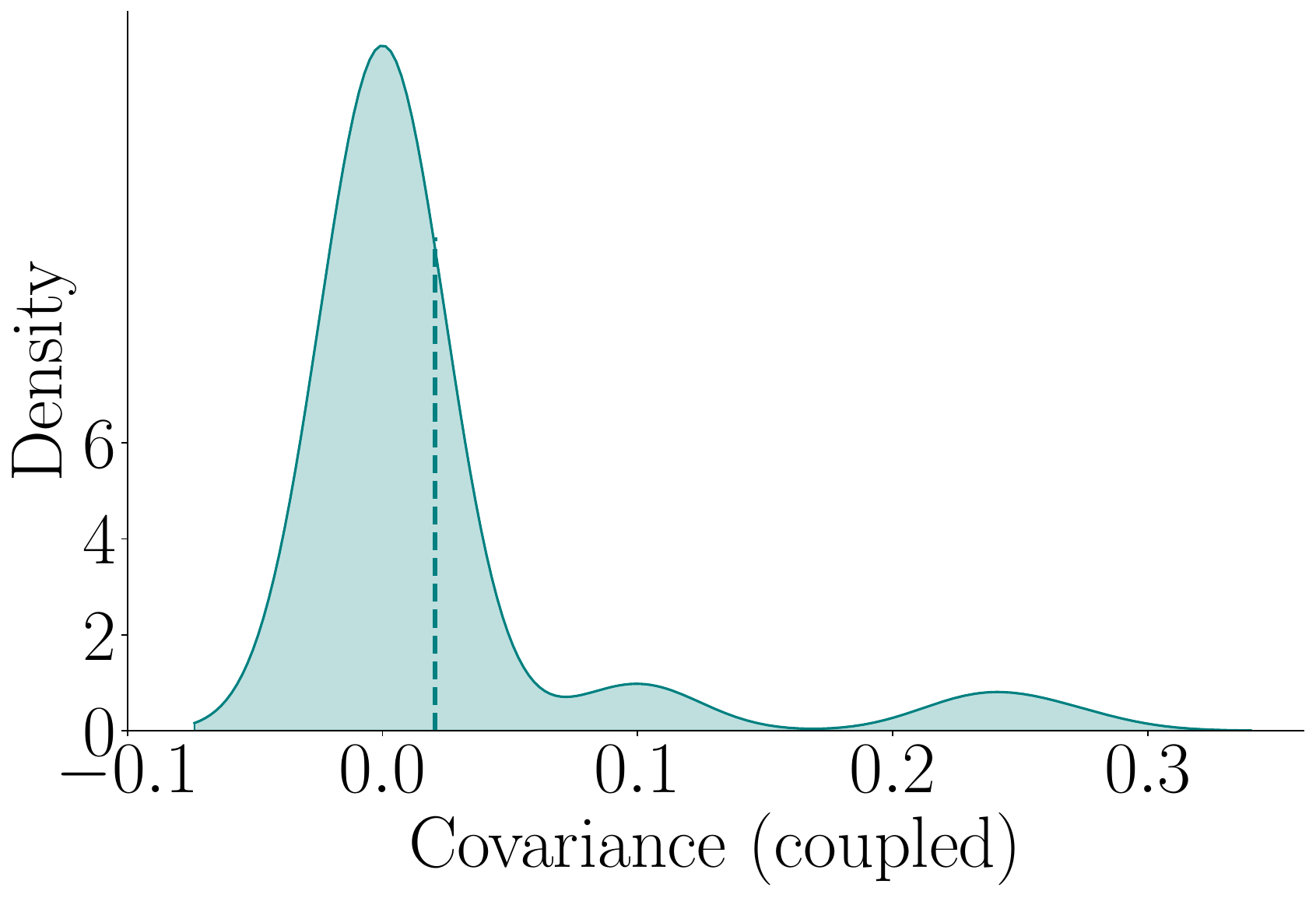} &
    \includegraphics[width=0.23\linewidth]{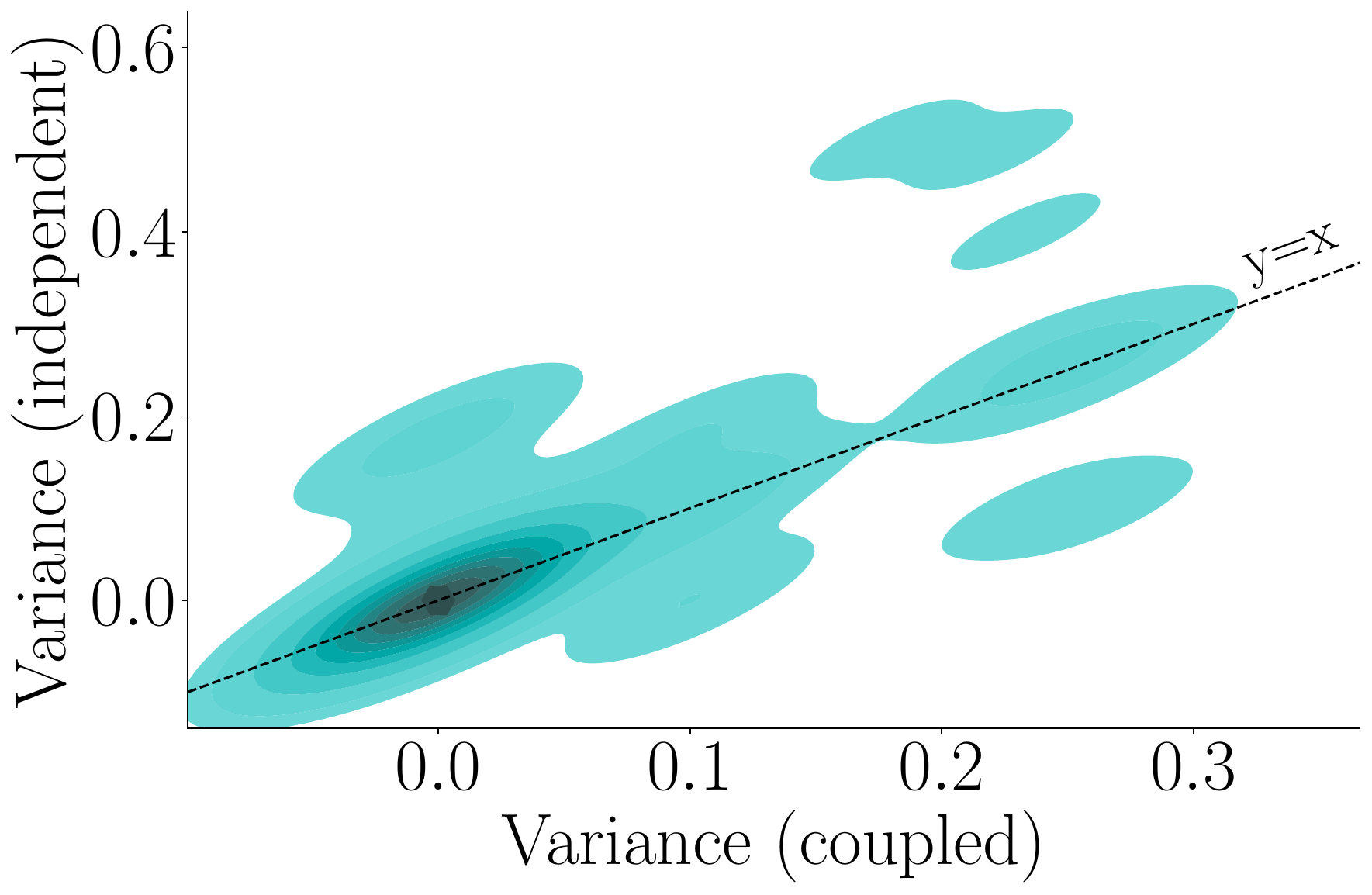} &
    \includegraphics[width=0.23\linewidth]{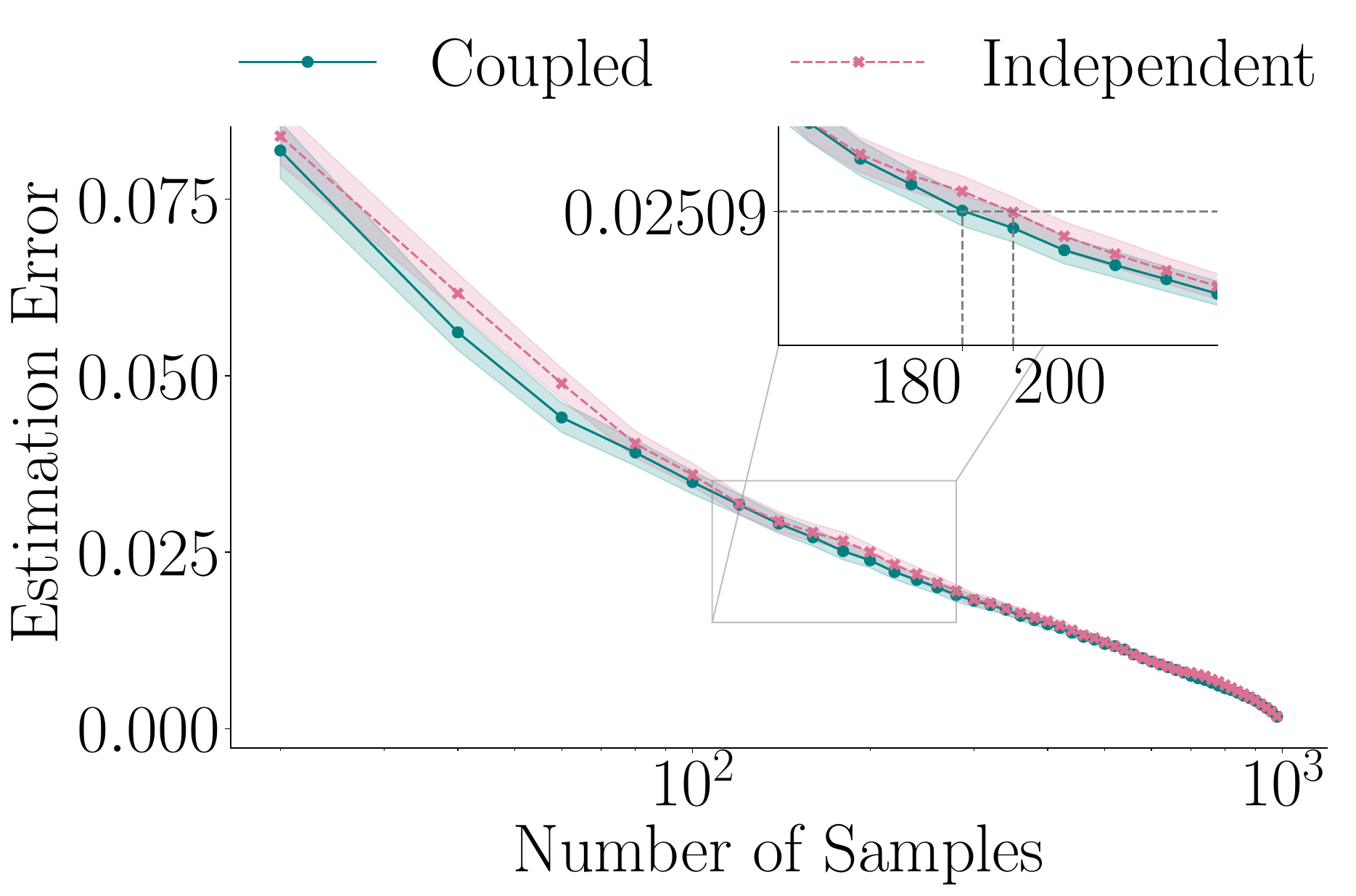} \\ \\

    (a) Score covariance & (b) Variance of the score difference & (c) Estimation error vs. \# samples \\ 
    
\end{tabular}
    \caption{\textbf{Comparison between four pairs of LLMs in the \texttt{Qwen} family on multiple-choice questions from the ``college computer science'' knowledge area of the MMLU dataset.}
    Panels in column (a) show the kernel density estimate (KDE) of the covariance between the scores of the two LLMs on each question under coupled generation; the dashed lines correspond to average values. Panels in column (b) show the KDE of the variance of the difference between the scores of the LLMs on each question under coupled and independent generation; the highlighted points correspond to median values. Panels in column (c) show the absolute error in the estimation of the expected difference between the scores of the LLMs against the number of samples; for each point on the x-axis, we perform $1{,}000$ sub-samplings and shaded areas correspond to $95\%$ confidence intervals.}
    \label{fig:mmlu-qwen-second-4}
\end{figure}

\begin{figure}[h]
\centering
\begin{tabular}{c c c}
     \multicolumn{3}{c}{\texttt{2.5-1.5B} vs. \texttt{2.5-7B-bnb-8bit}}\\
    \includegraphics[width=0.23\linewidth]{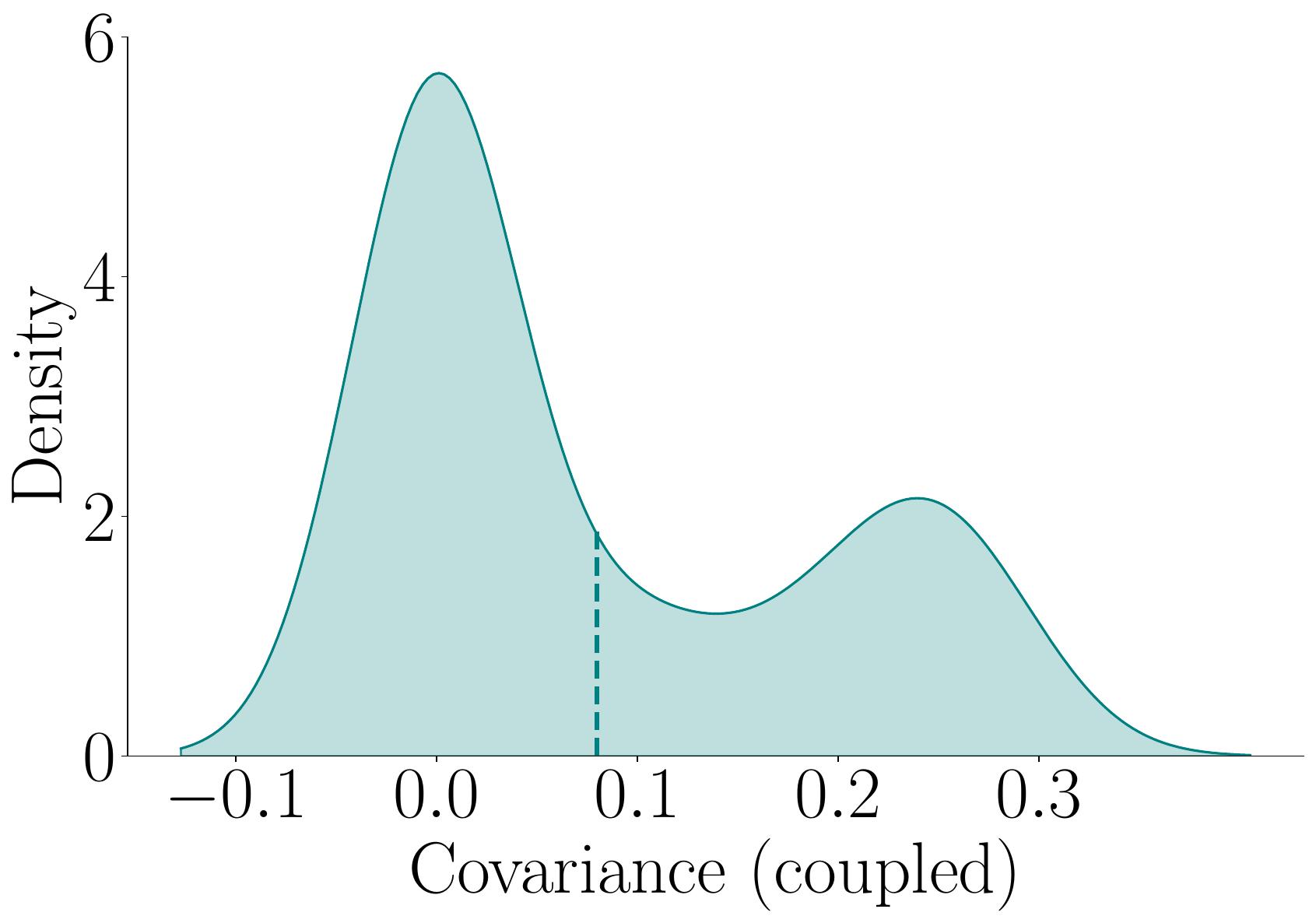} &
    \includegraphics[width=0.23\linewidth]{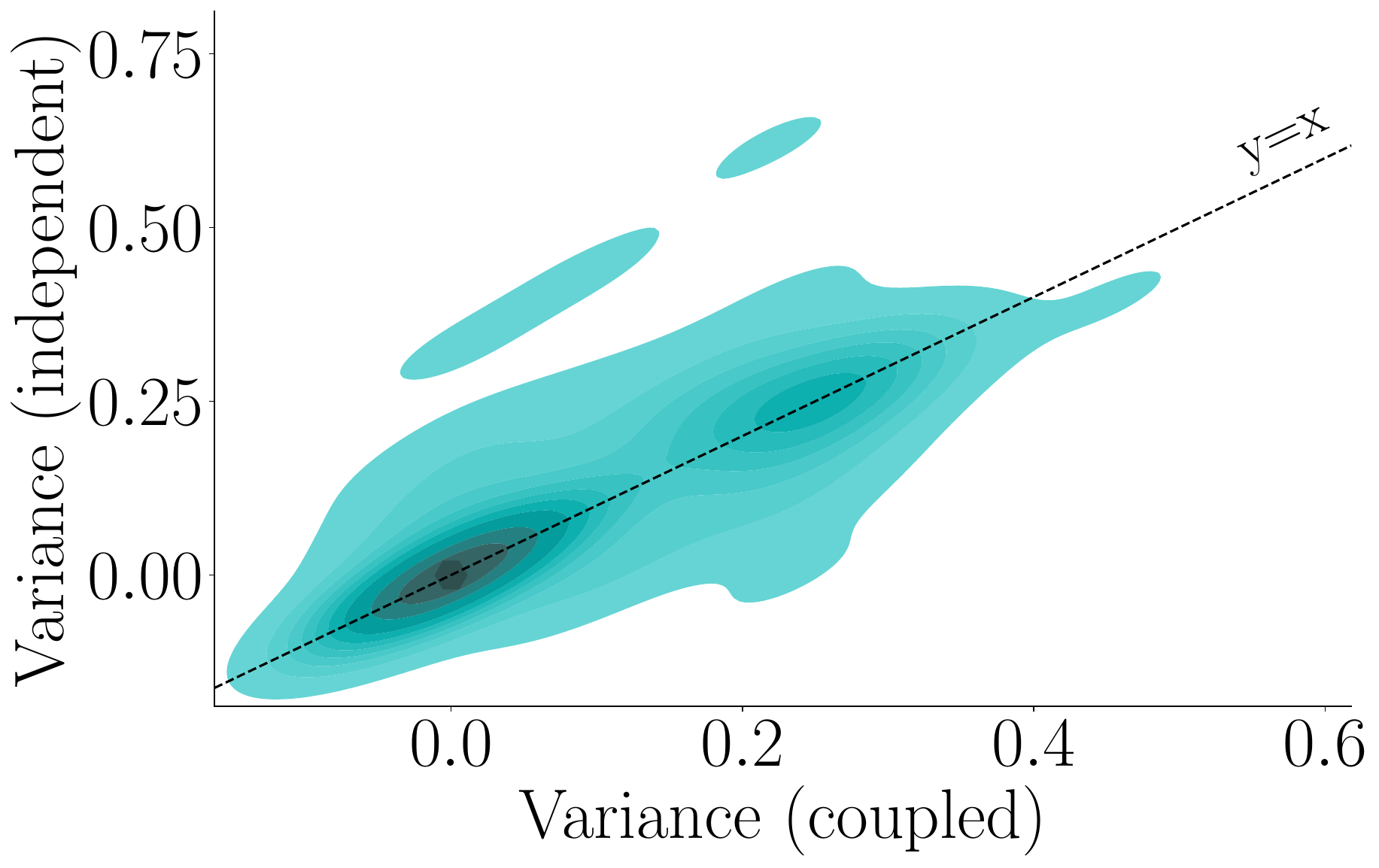} &
    \includegraphics[width=0.23\linewidth]{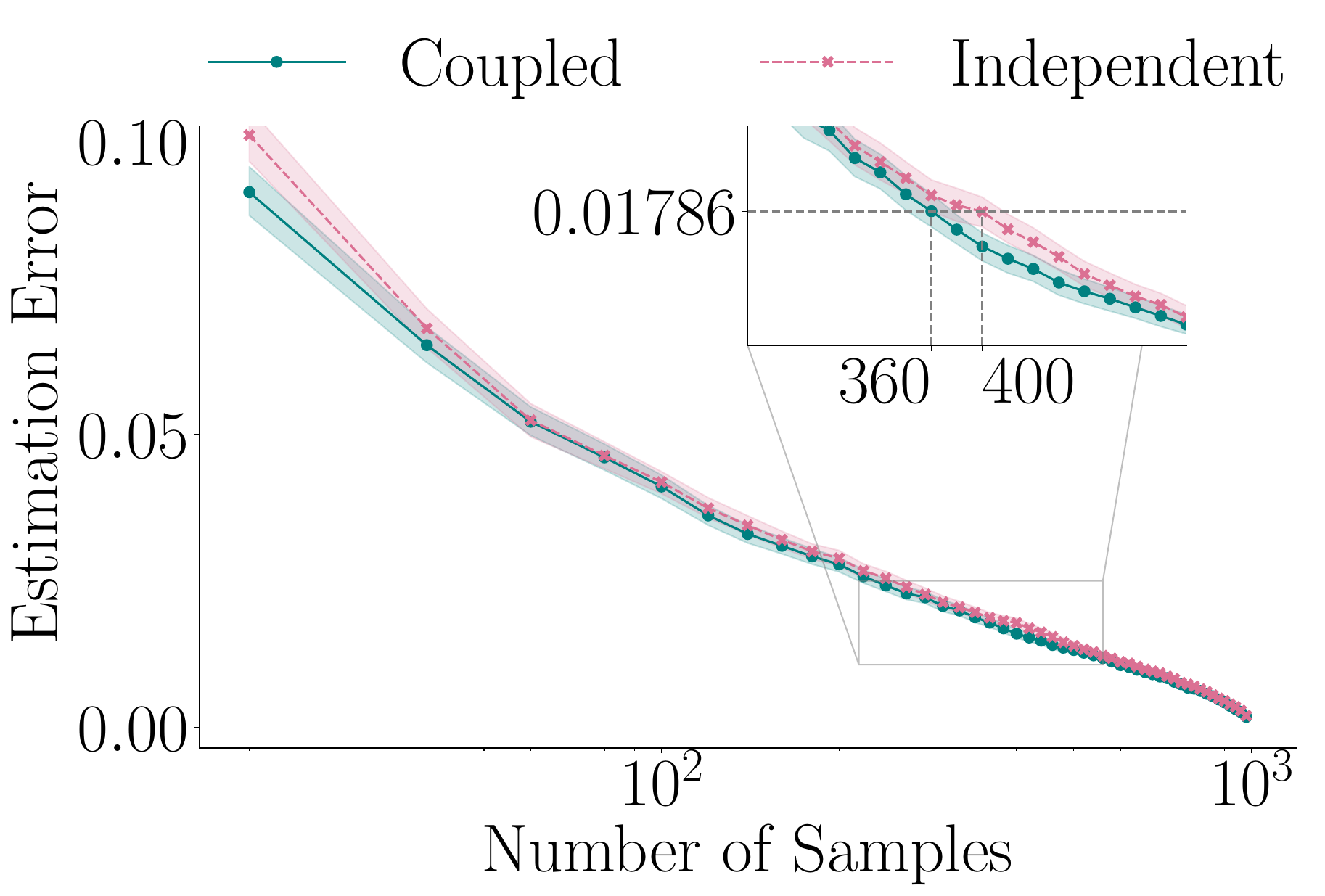} \\ \\
    \multicolumn{3}{c}{\texttt{2.5-7B-bnb-4bit} vs. \texttt{2.5-7B-AWQ-INT4}}\\
    \includegraphics[width=0.23\linewidth]{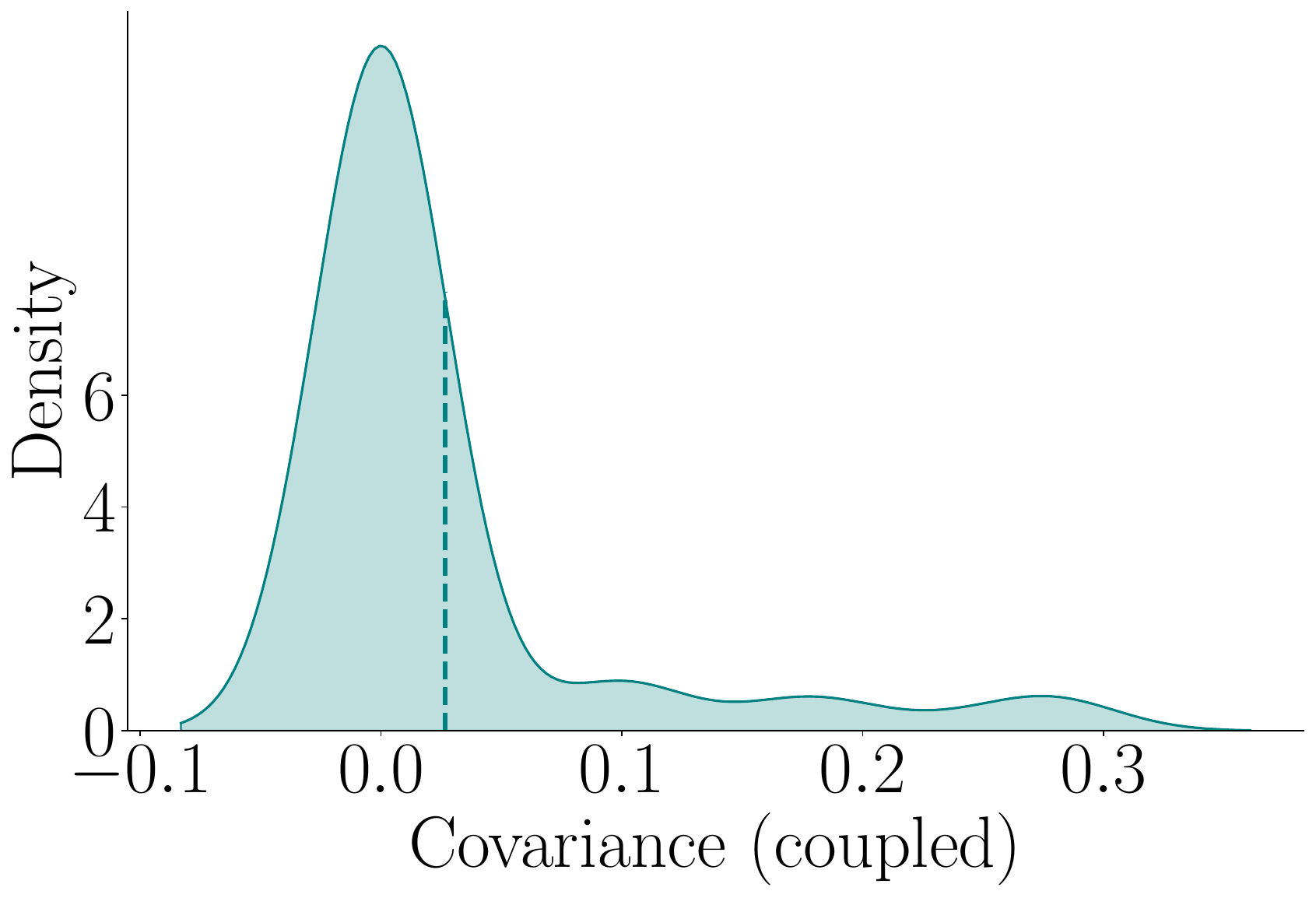} &
    \includegraphics[width=0.23\linewidth]{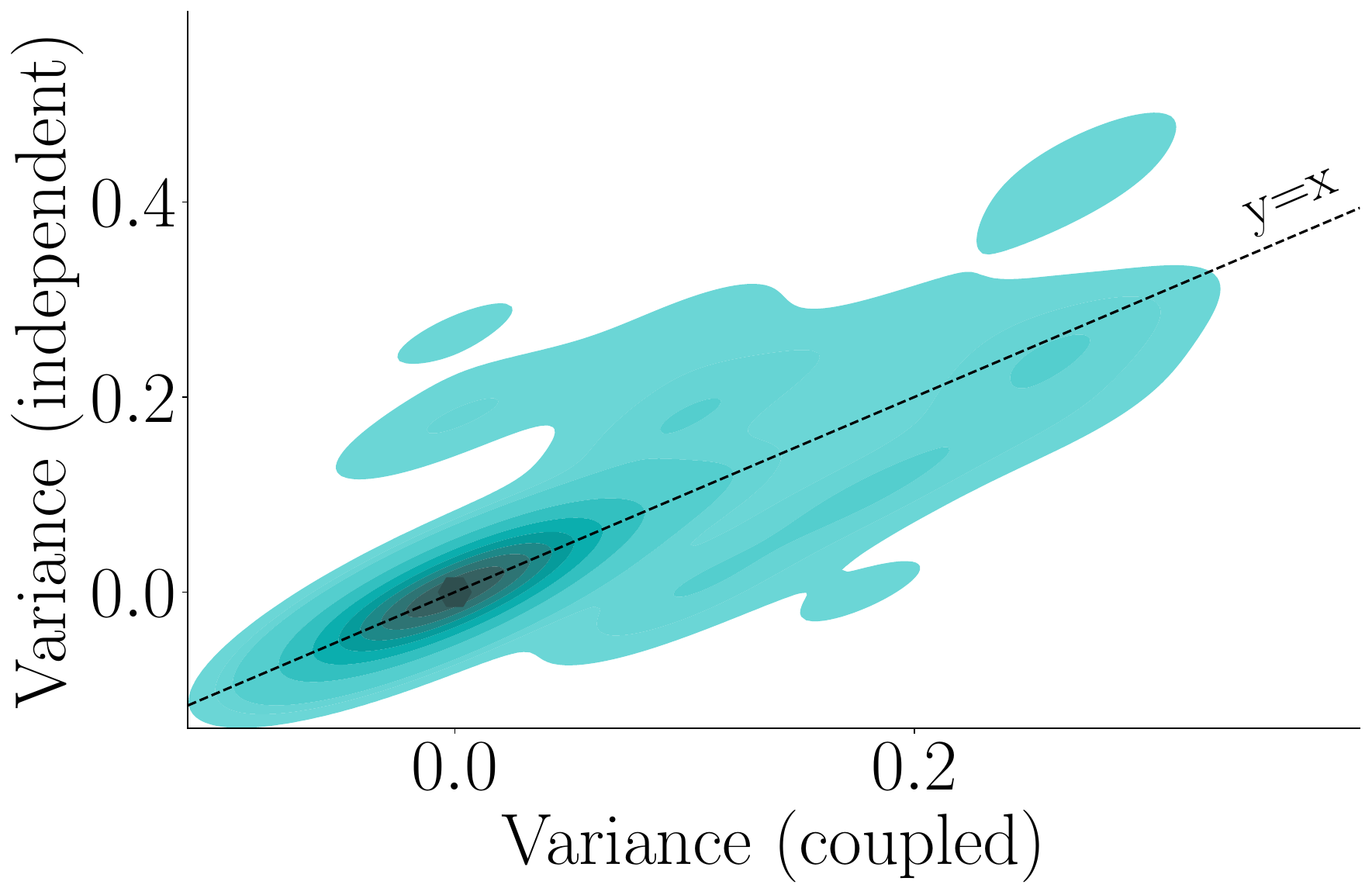} &
    \includegraphics[width=0.23\linewidth]{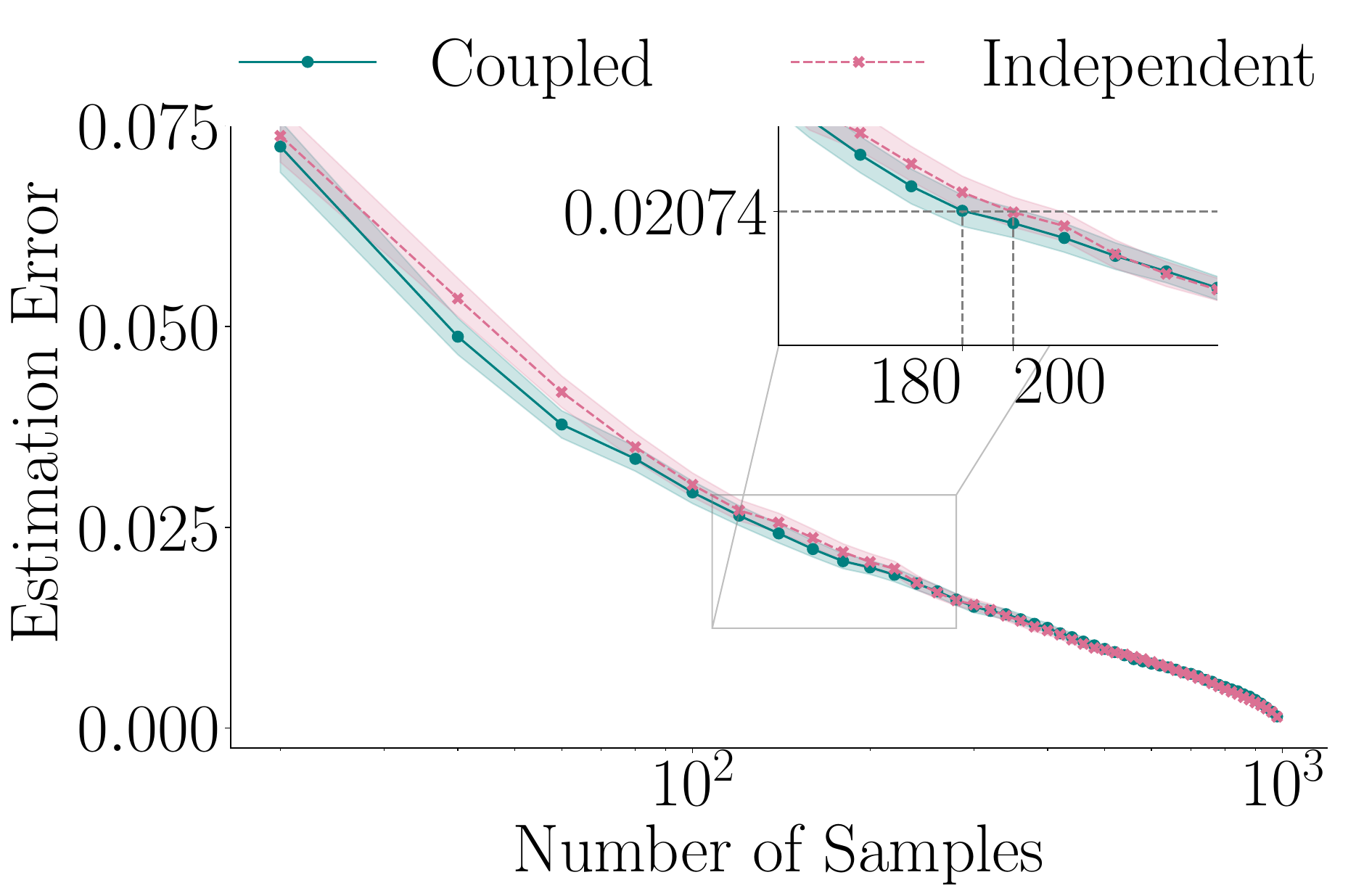} \\ \\
     \multicolumn{3}{c}{\texttt{3-8B} vs. \texttt{2.5-7B-AWQ-INT4}}\\
    \includegraphics[width=0.23\linewidth]{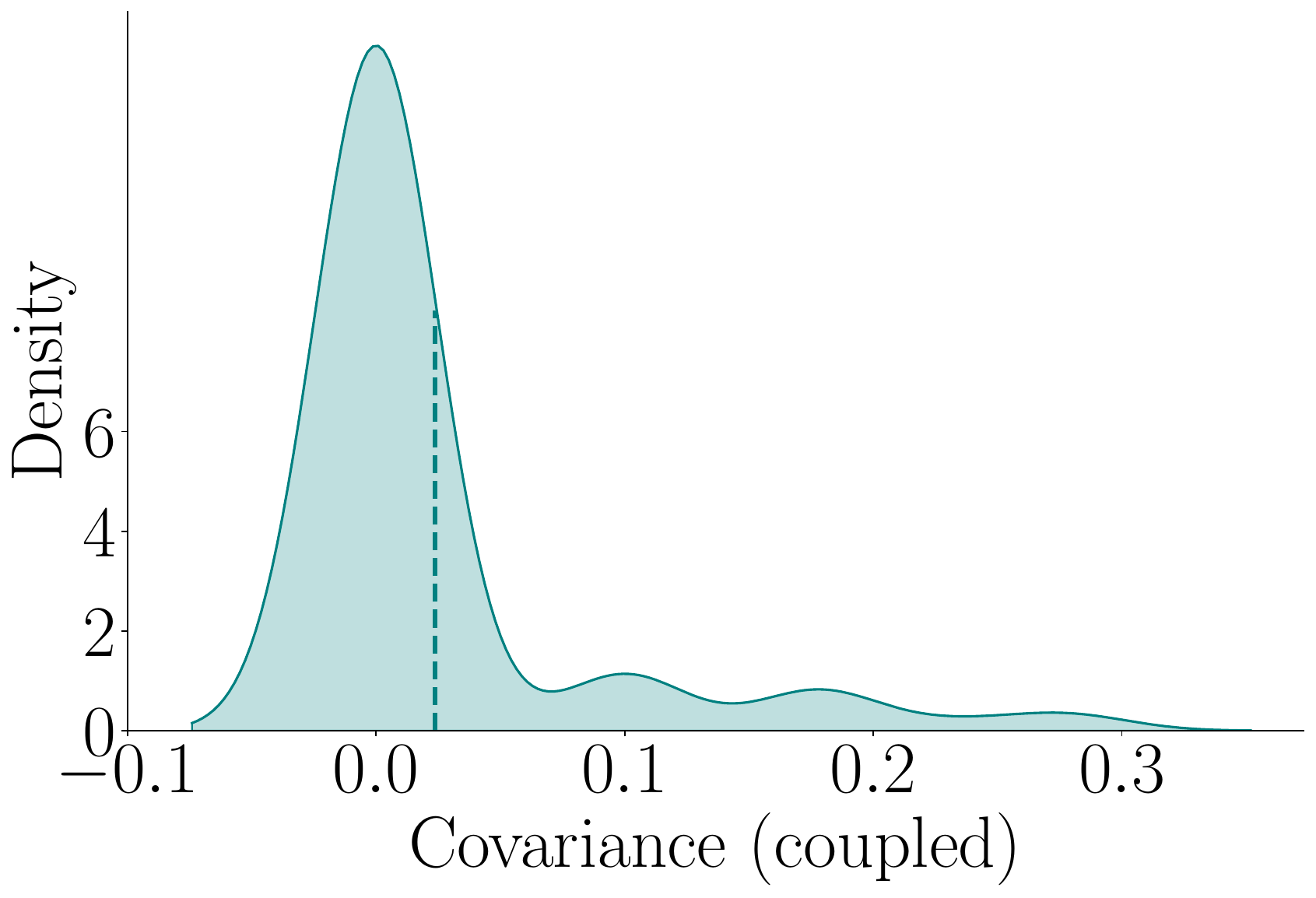} &
    \includegraphics[width=0.23\linewidth]{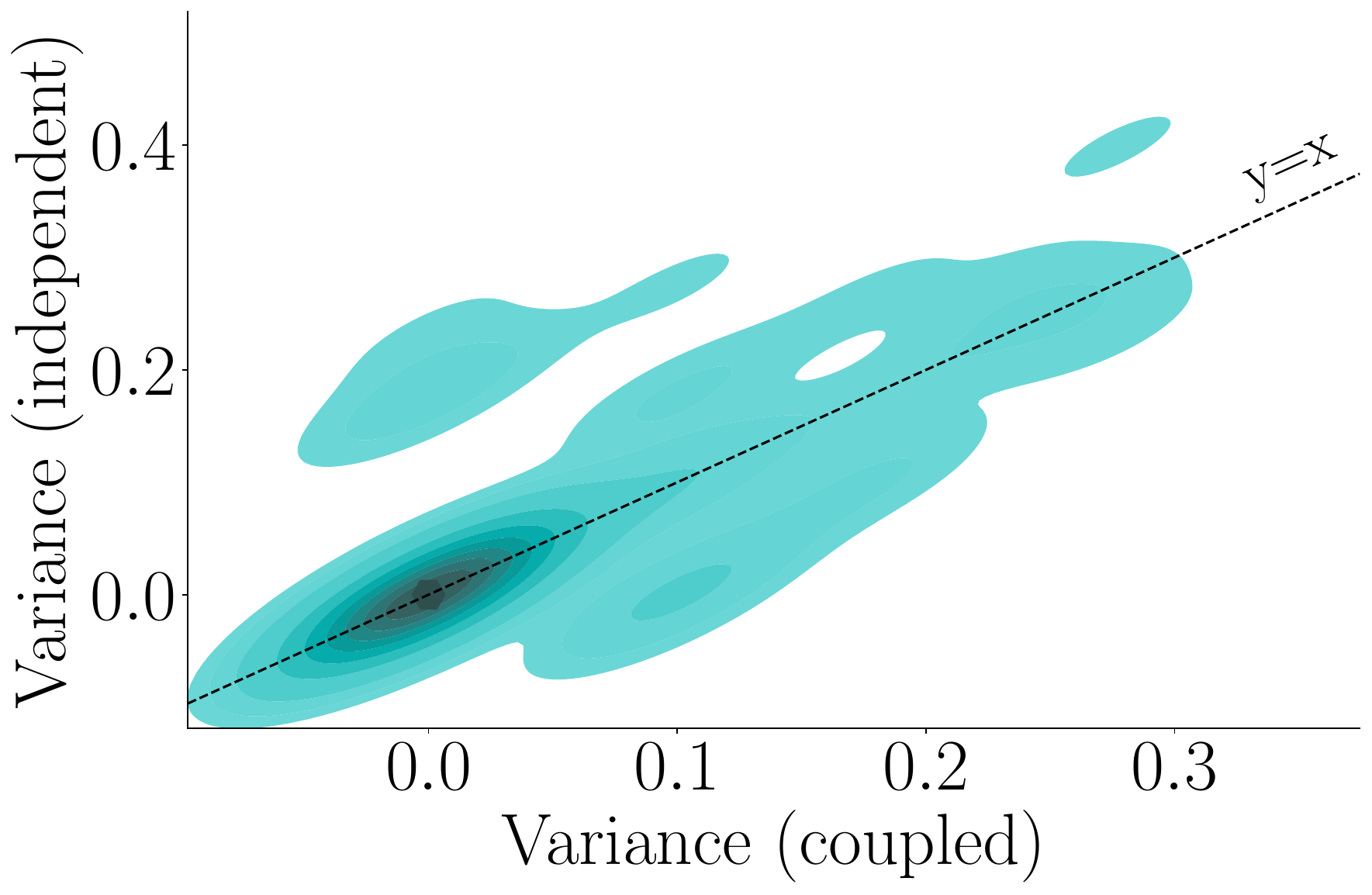} &
    \includegraphics[width=0.23\linewidth]{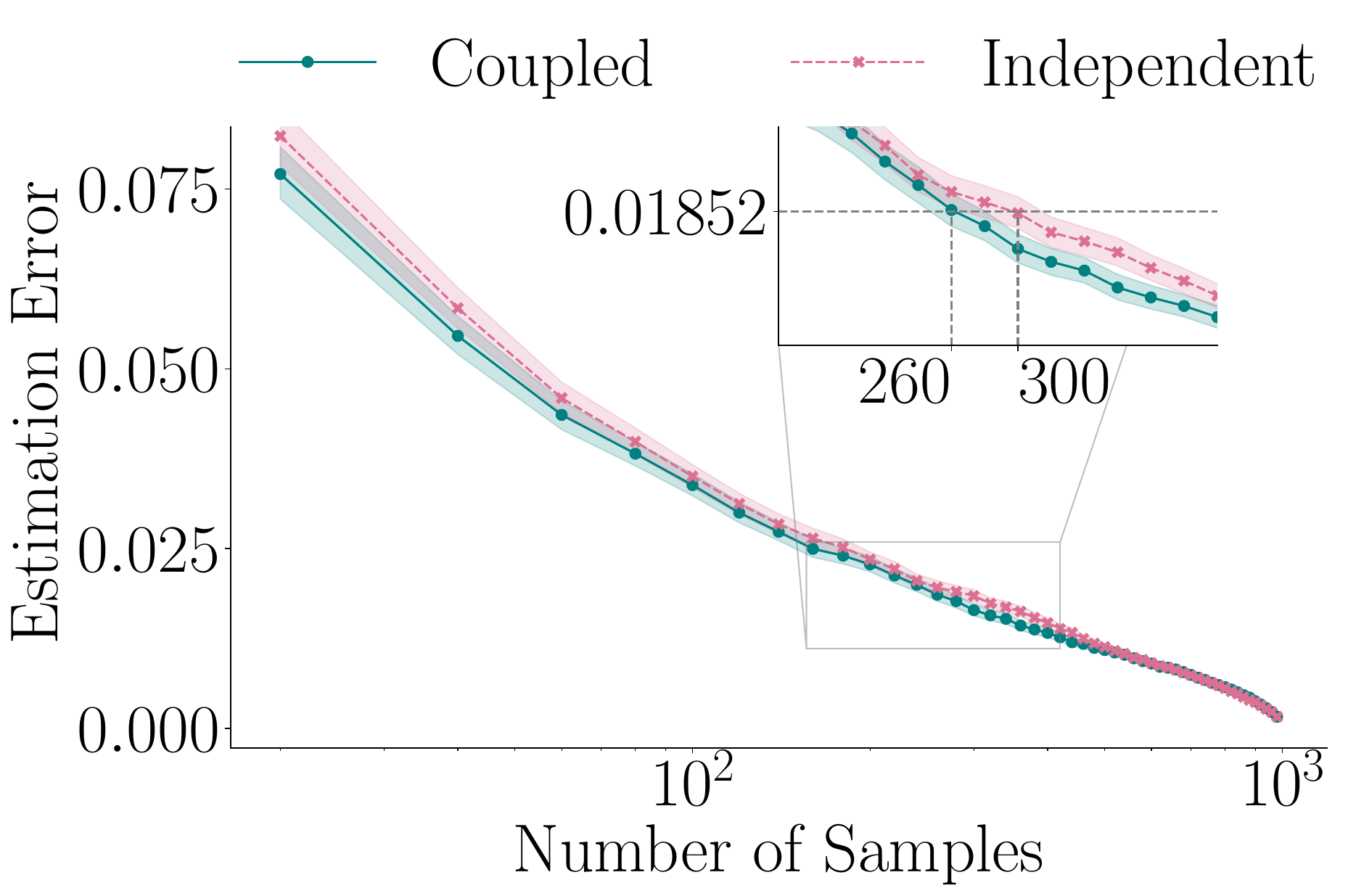} \\ \\

    (a) Score covariance & (b) Variance of the score difference & (c) Estimation error vs. \# samples \\ 
    
\end{tabular}
    \caption{\textbf{Comparison between three pairs of LLMs in the \texttt{Qwen} family on multiple-choice questions from the ``college computer science'' knowledge area of the MMLU dataset.}
    Panels in column (a) show the kernel density estimate (KDE) of the covariance between the scores of the two LLMs on each question under coupled generation; the dashed lines correspond to average values. Panels in column (b) show the KDE of the variance of the difference between the scores of the LLMs on each question under coupled and independent generation; the highlighted points correspond to median values. Panels in column (c) show the absolute error in the estimation of the expected difference between the scores of the LLMs against the number of samples; for each point on the x-axis, we perform $1{,}000$ sub-samplings and shaded areas correspond to $95\%$ confidence intervals.}
    \label{fig:mmlu-qwen-last-3}
\end{figure}

\begin{figure}[h]
\centering
\begin{tabular}{c c c}
    \multicolumn{3}{c}{\textbf{College chemistry}} \\
    \includegraphics[width=0.23\linewidth]{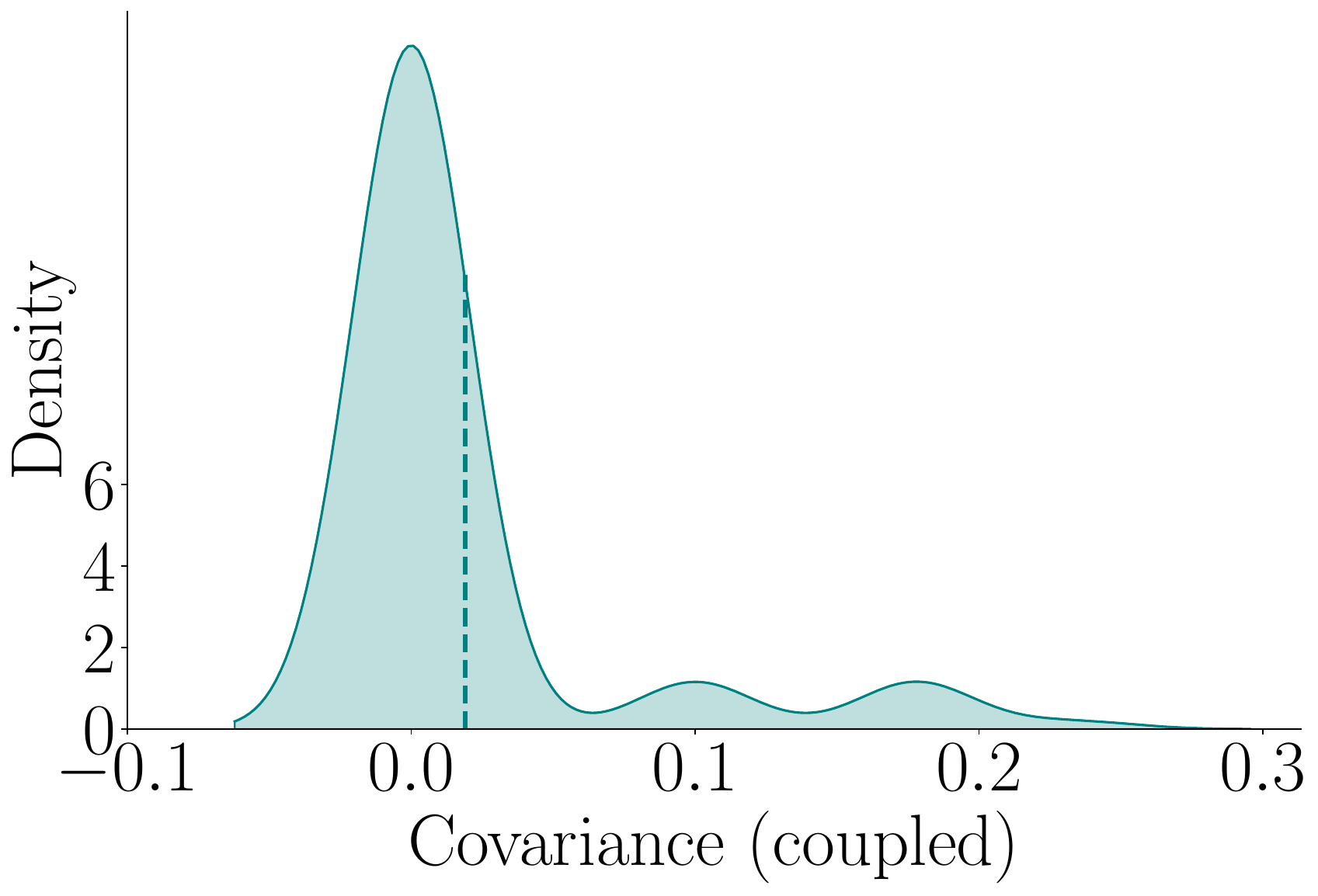} &
    \includegraphics[width=0.23\linewidth]{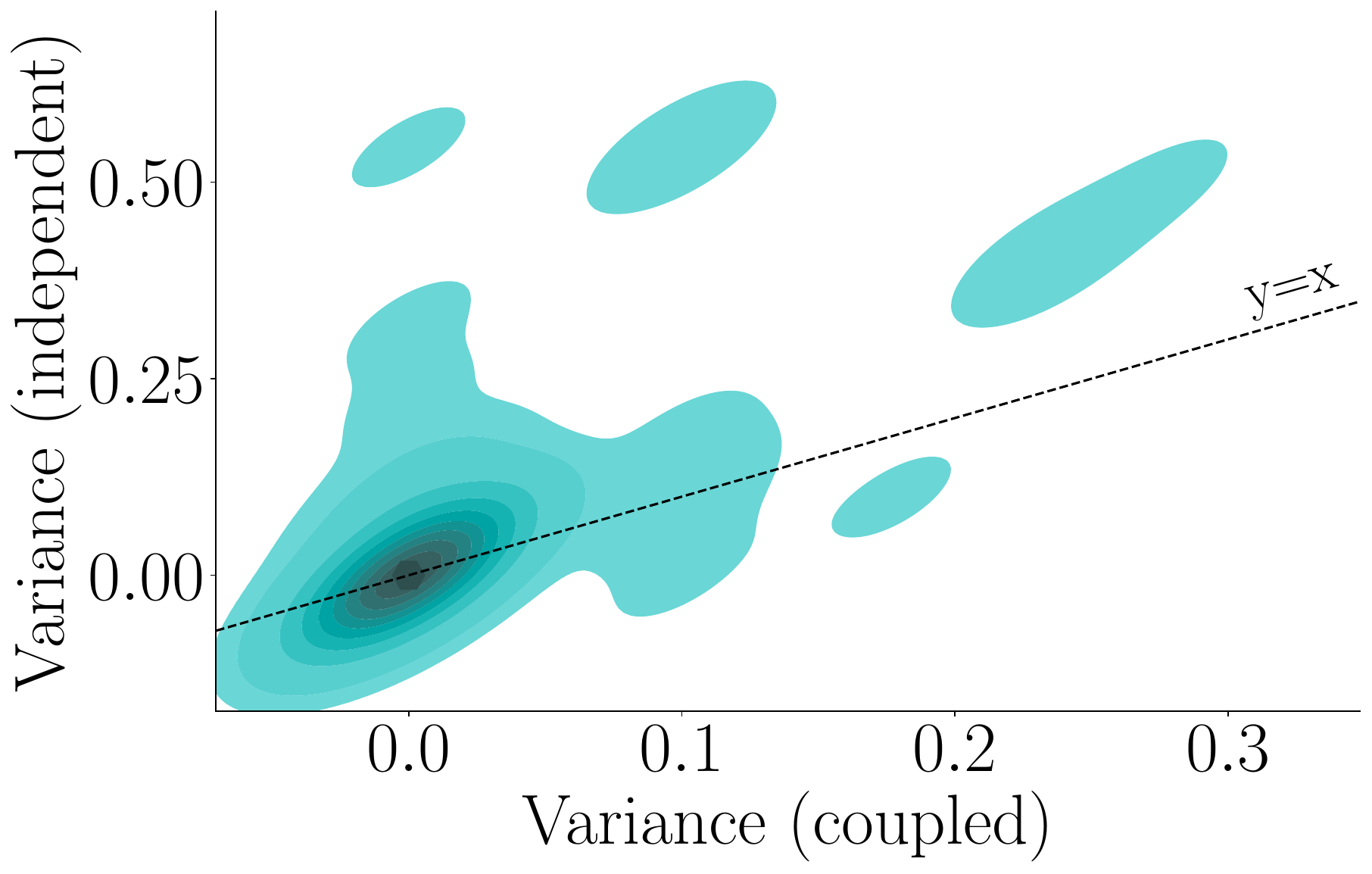} &
    \includegraphics[width=0.23\linewidth]{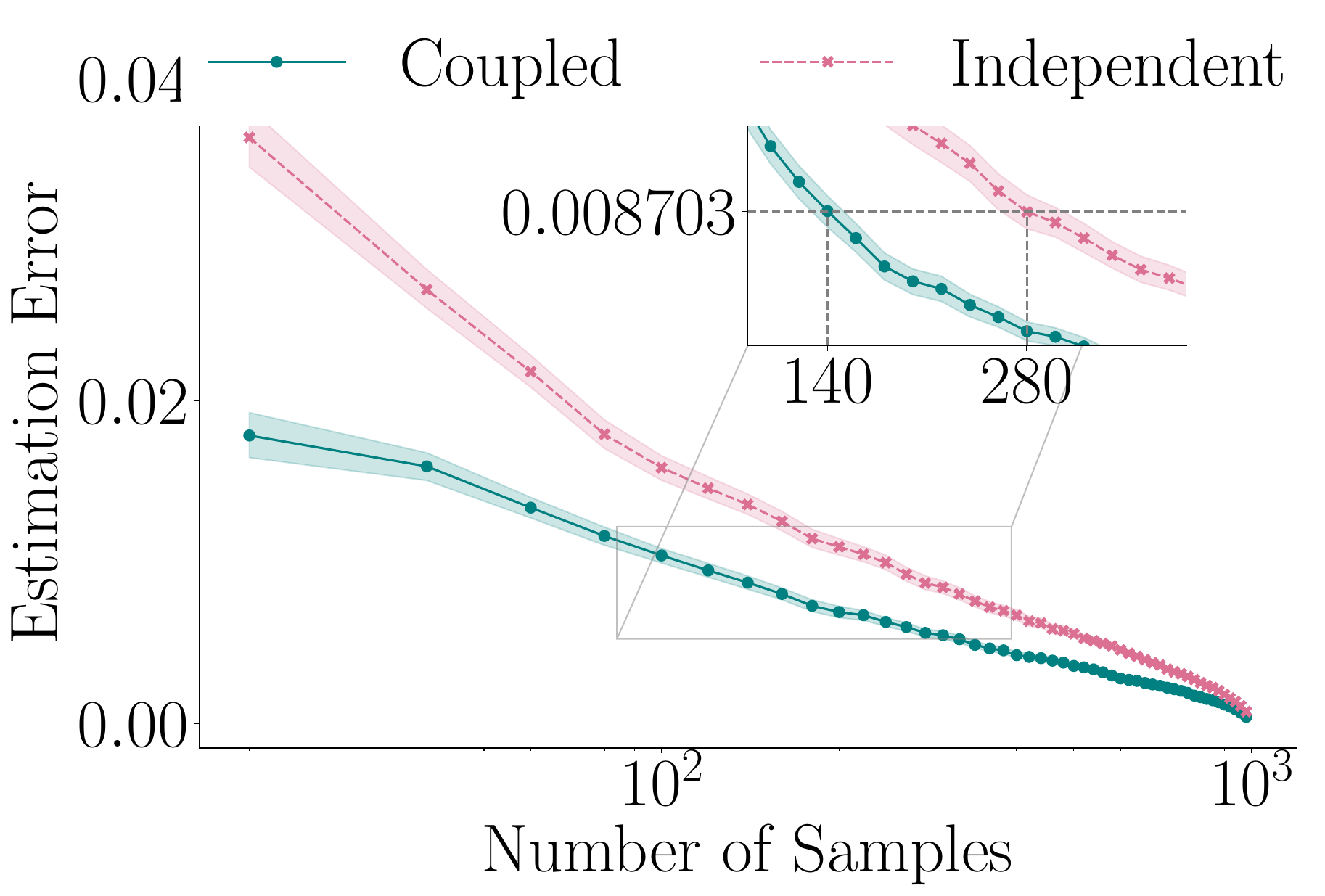} \\
    \multicolumn{3}{c}{\textbf{Professional accounting}} \\
    \includegraphics[width=0.23\linewidth]{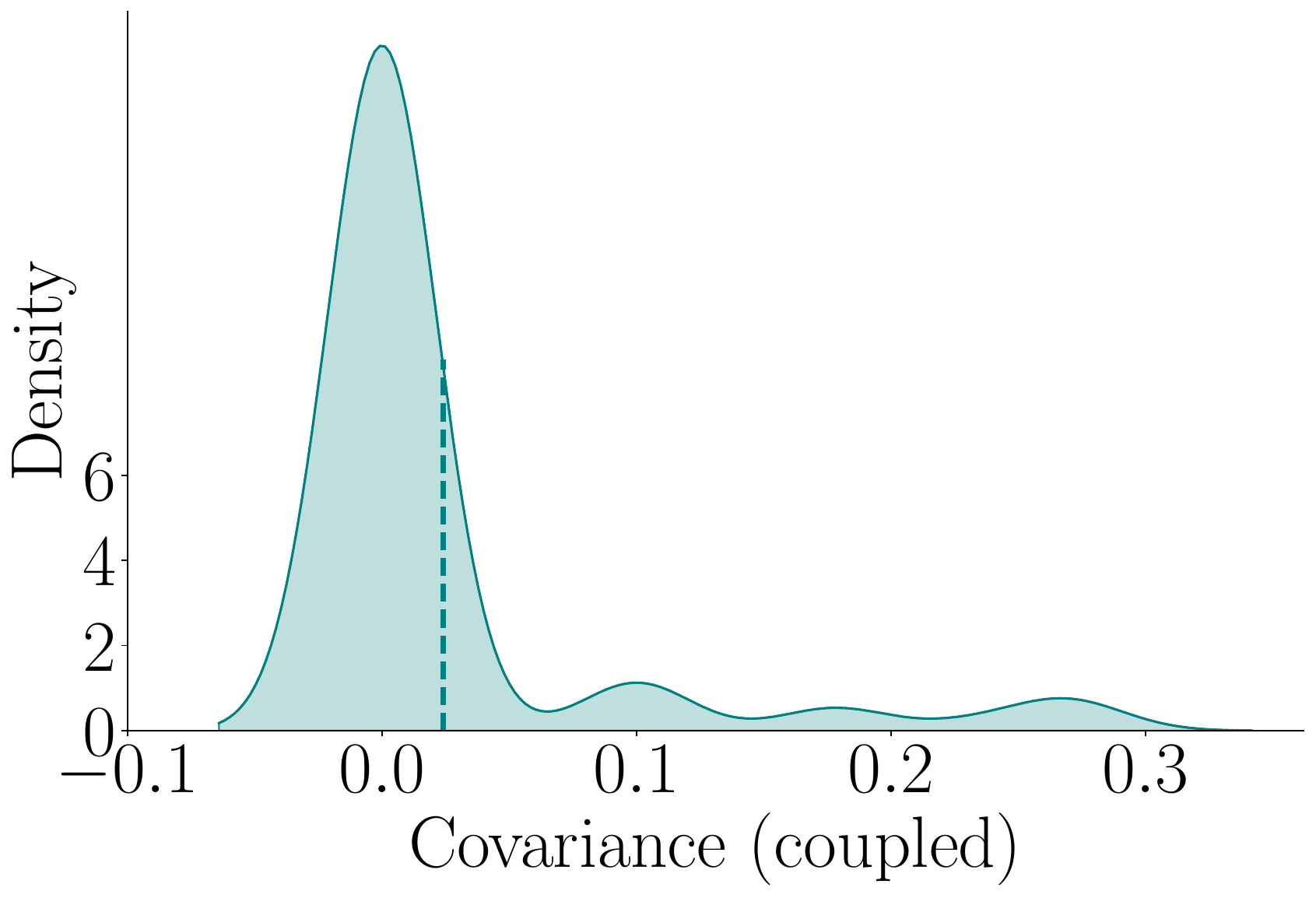} &
    \includegraphics[width=0.23\linewidth]{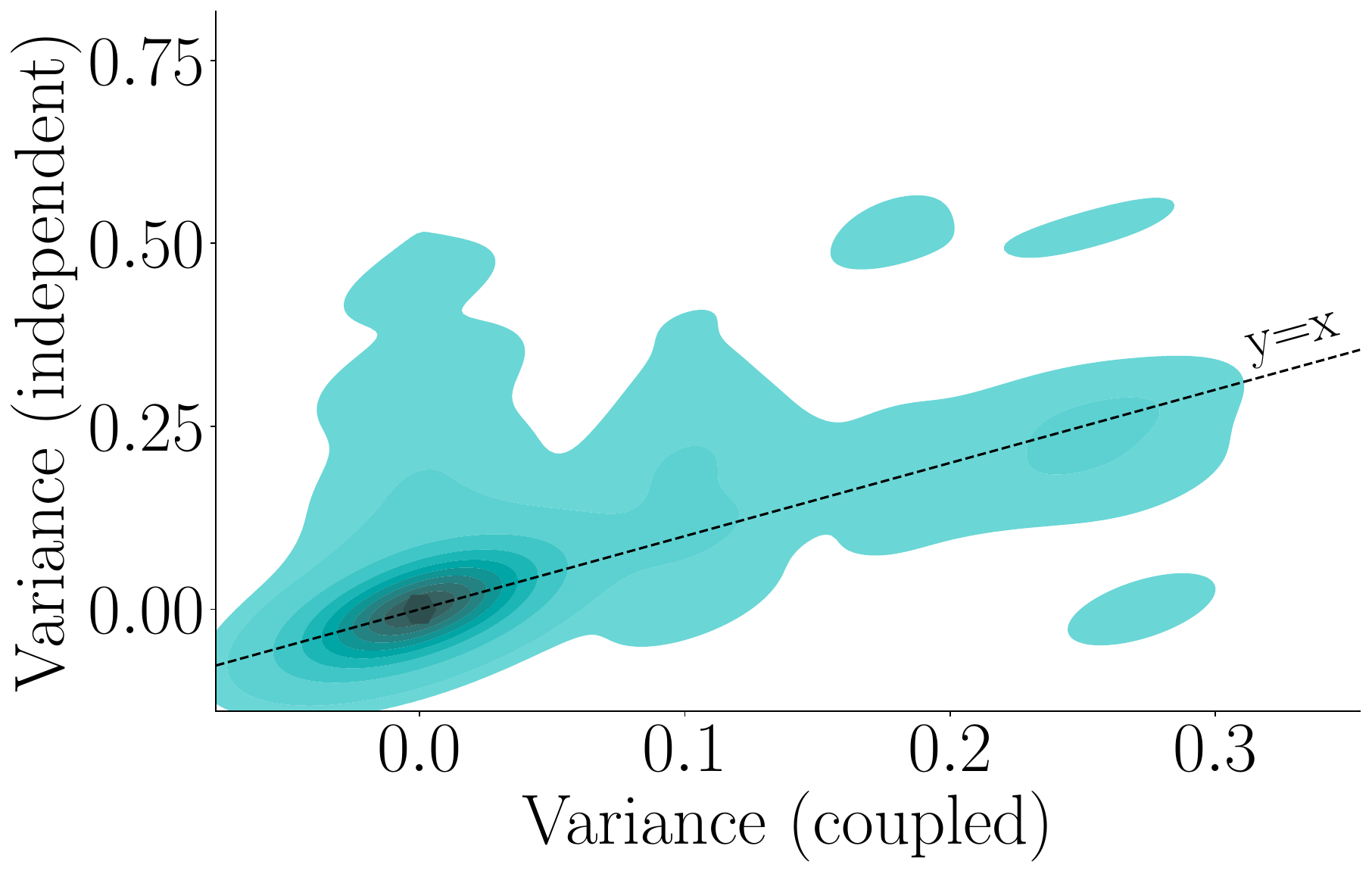} &
    \includegraphics[width=0.23\linewidth]{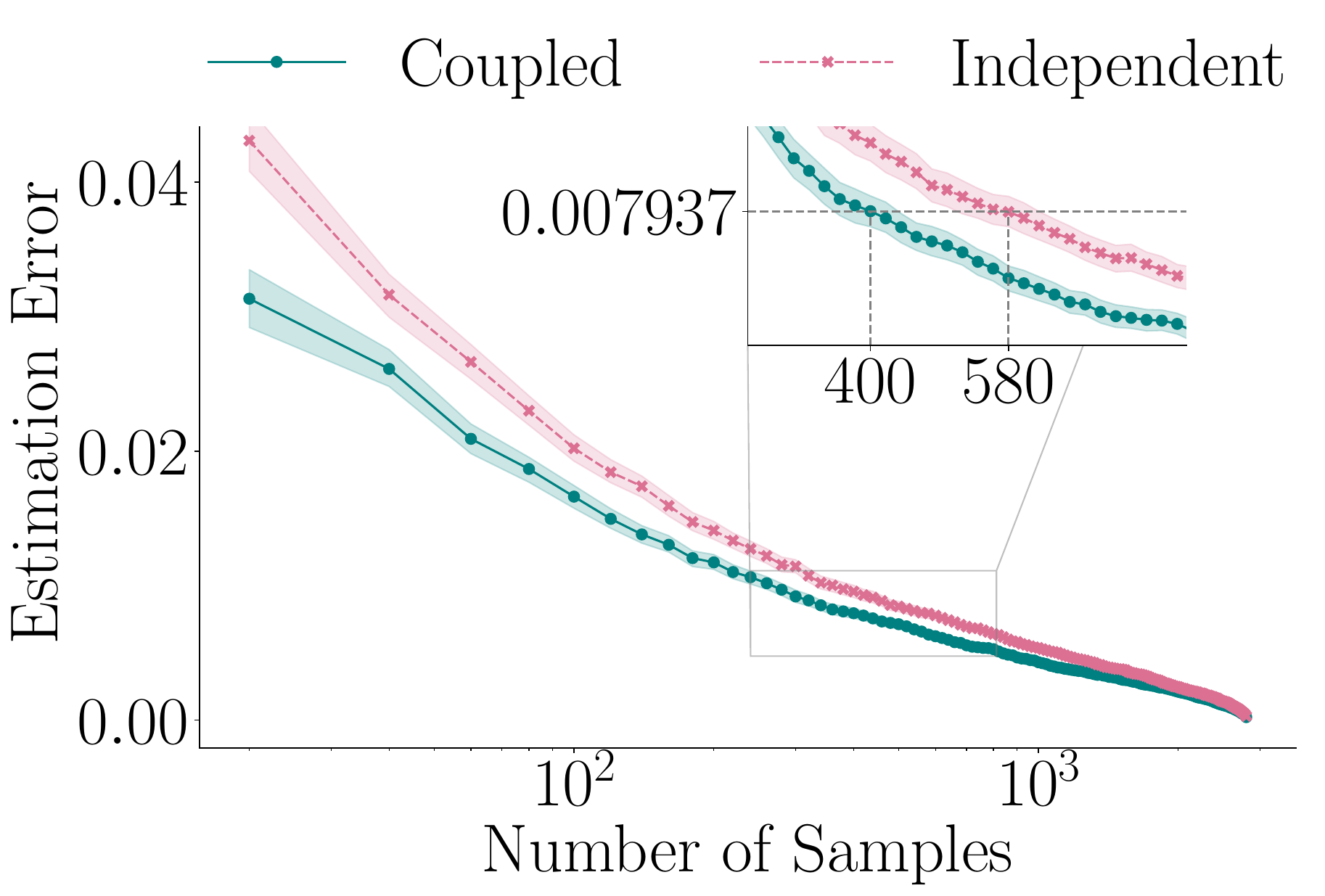} \\
    \multicolumn{3}{c}{\textbf{Professional law}}\\
    \includegraphics[width=0.23\linewidth]{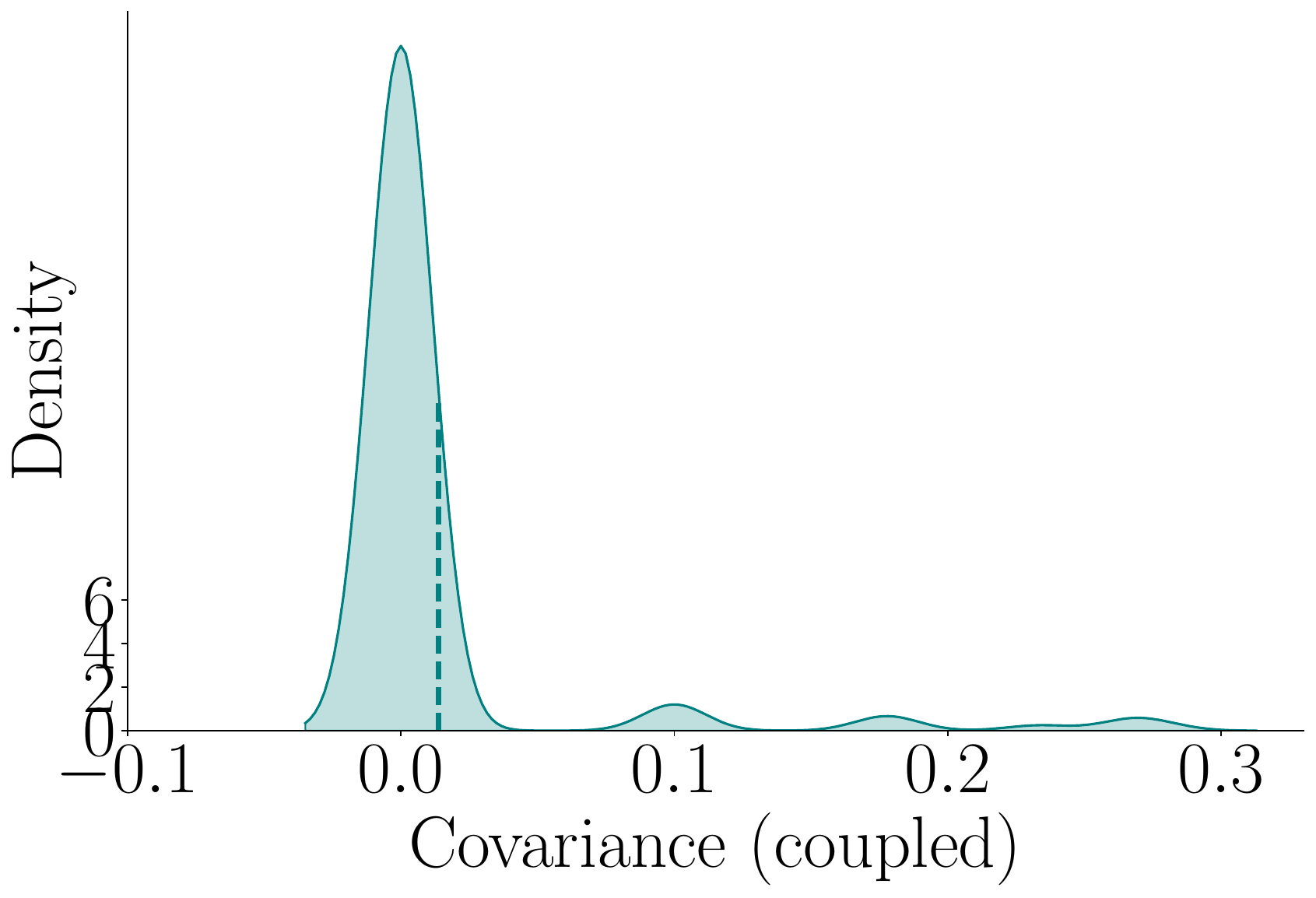} &
    \includegraphics[width=0.23\linewidth]{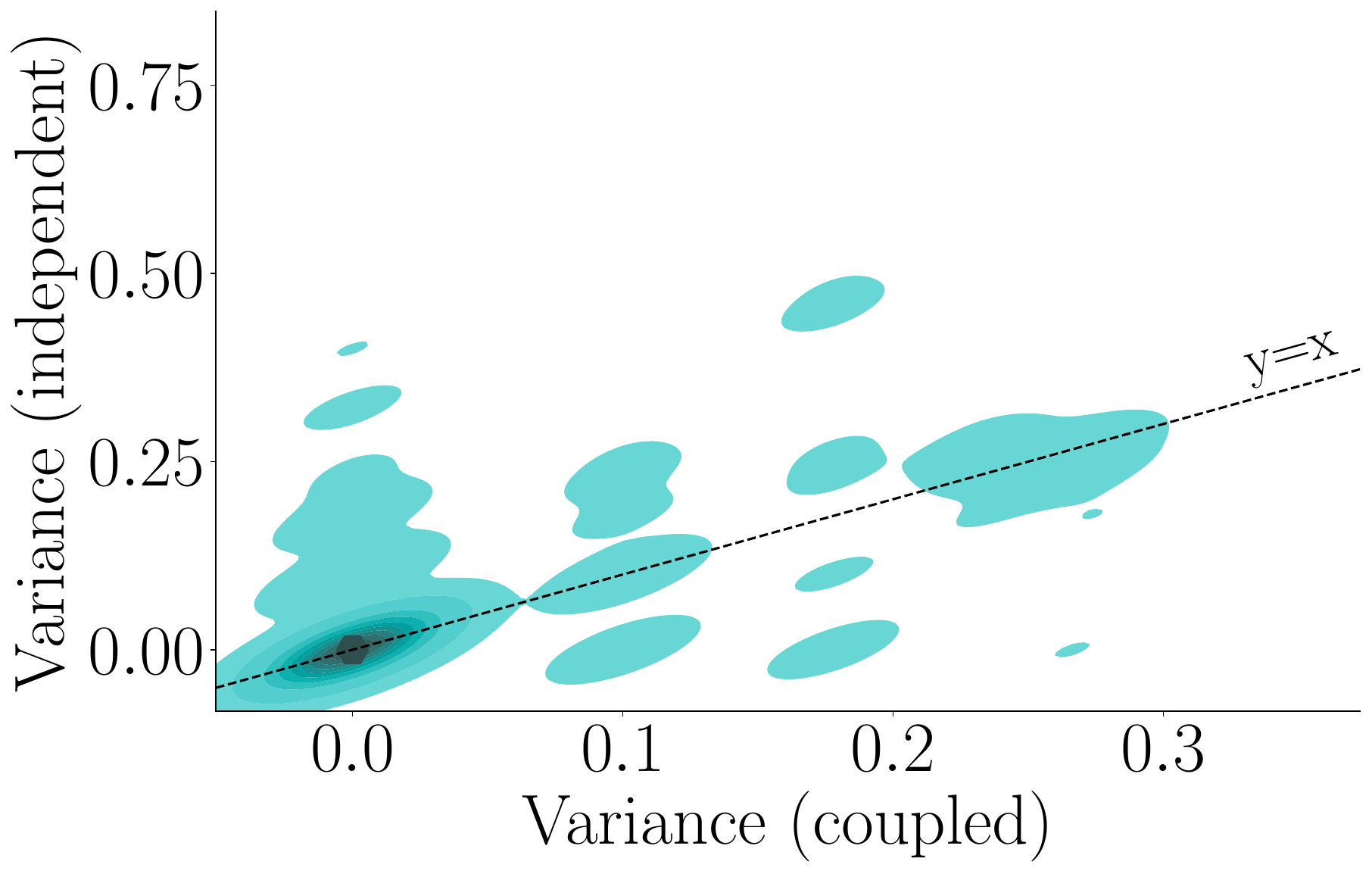} &
    \includegraphics[width=0.23\linewidth]{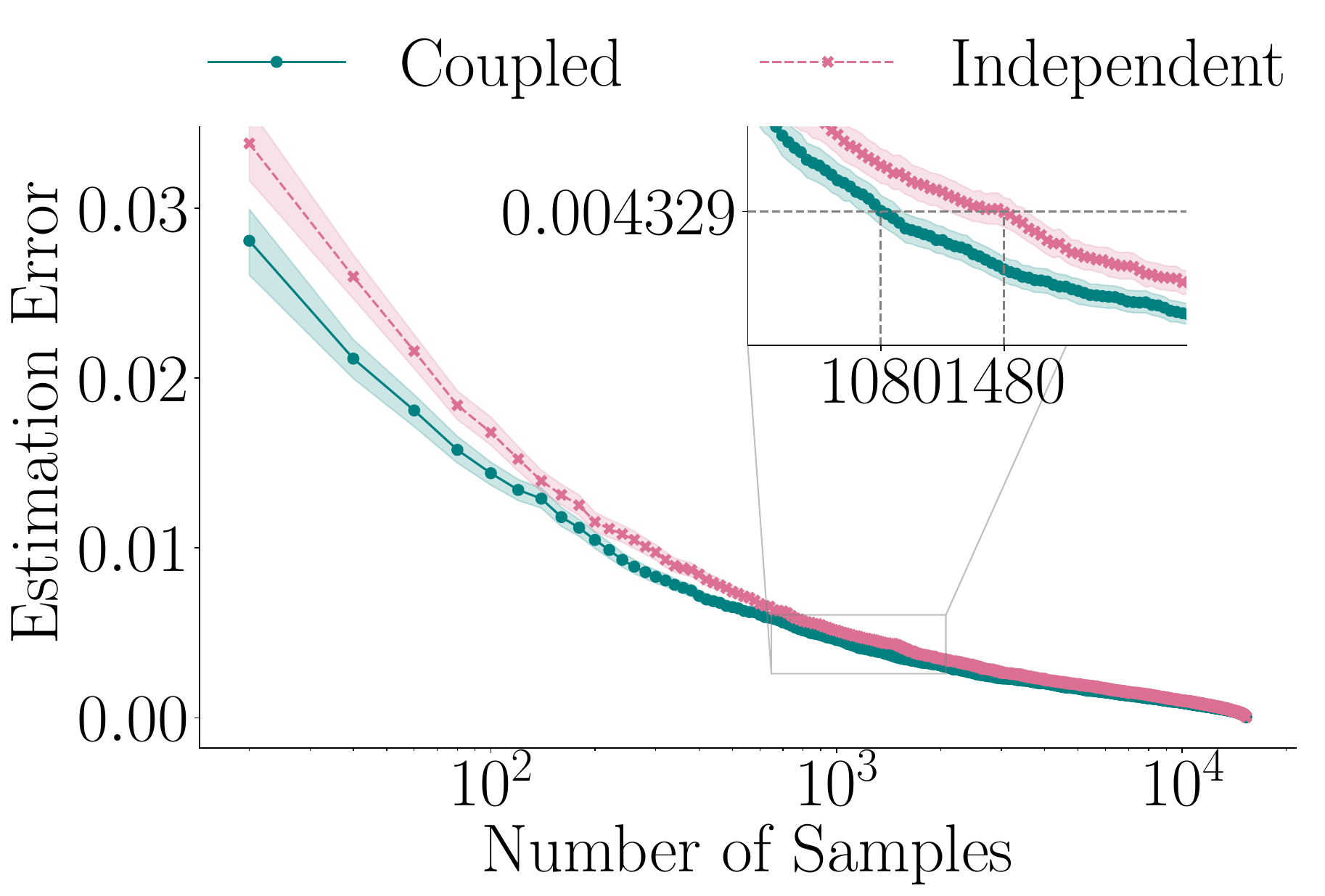} \\ \\
    \multicolumn{3}{c}{\textbf{Professional medicine}} \\   
    \includegraphics[width=0.23\linewidth]{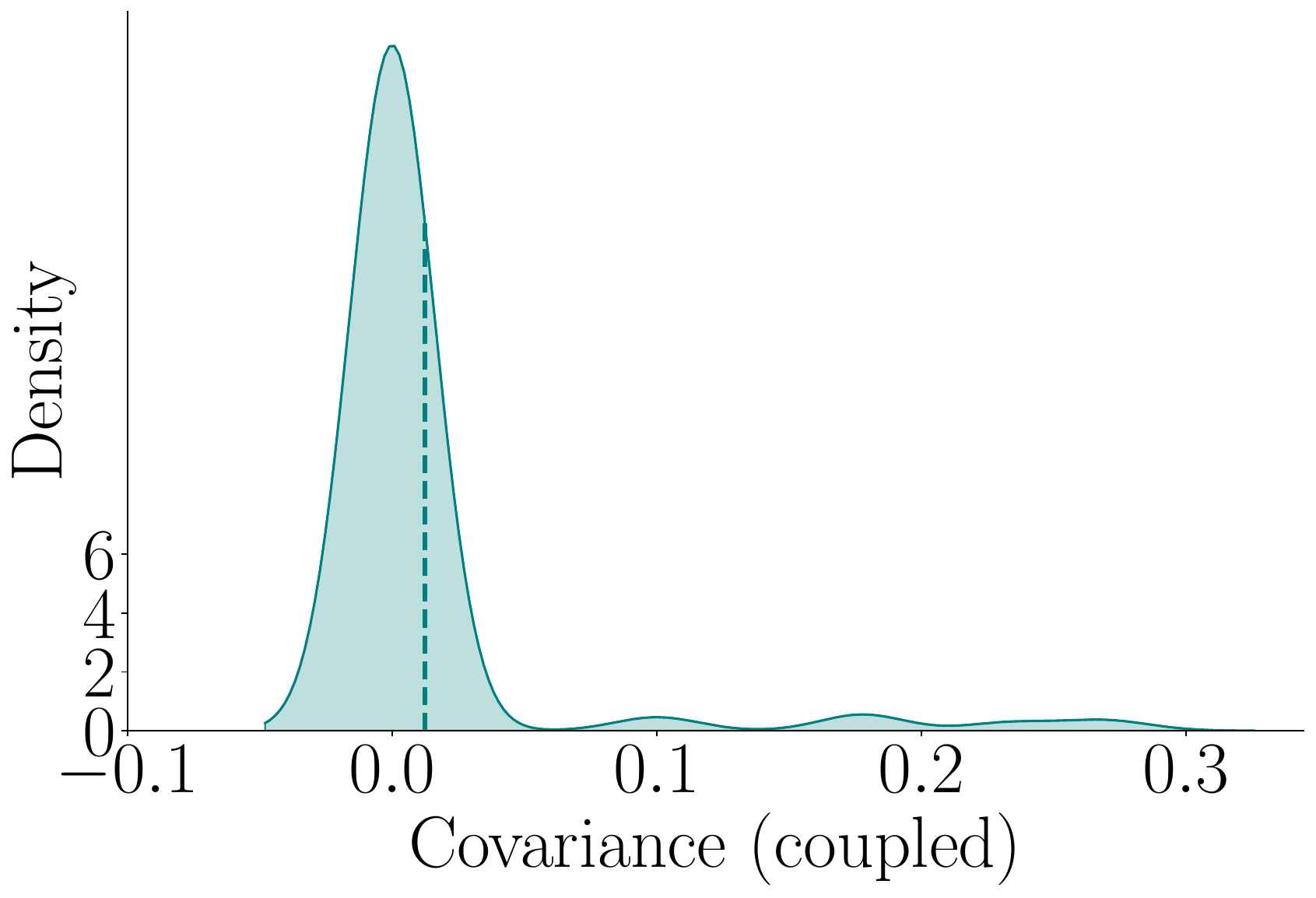} &
    \includegraphics[width=0.23\linewidth]{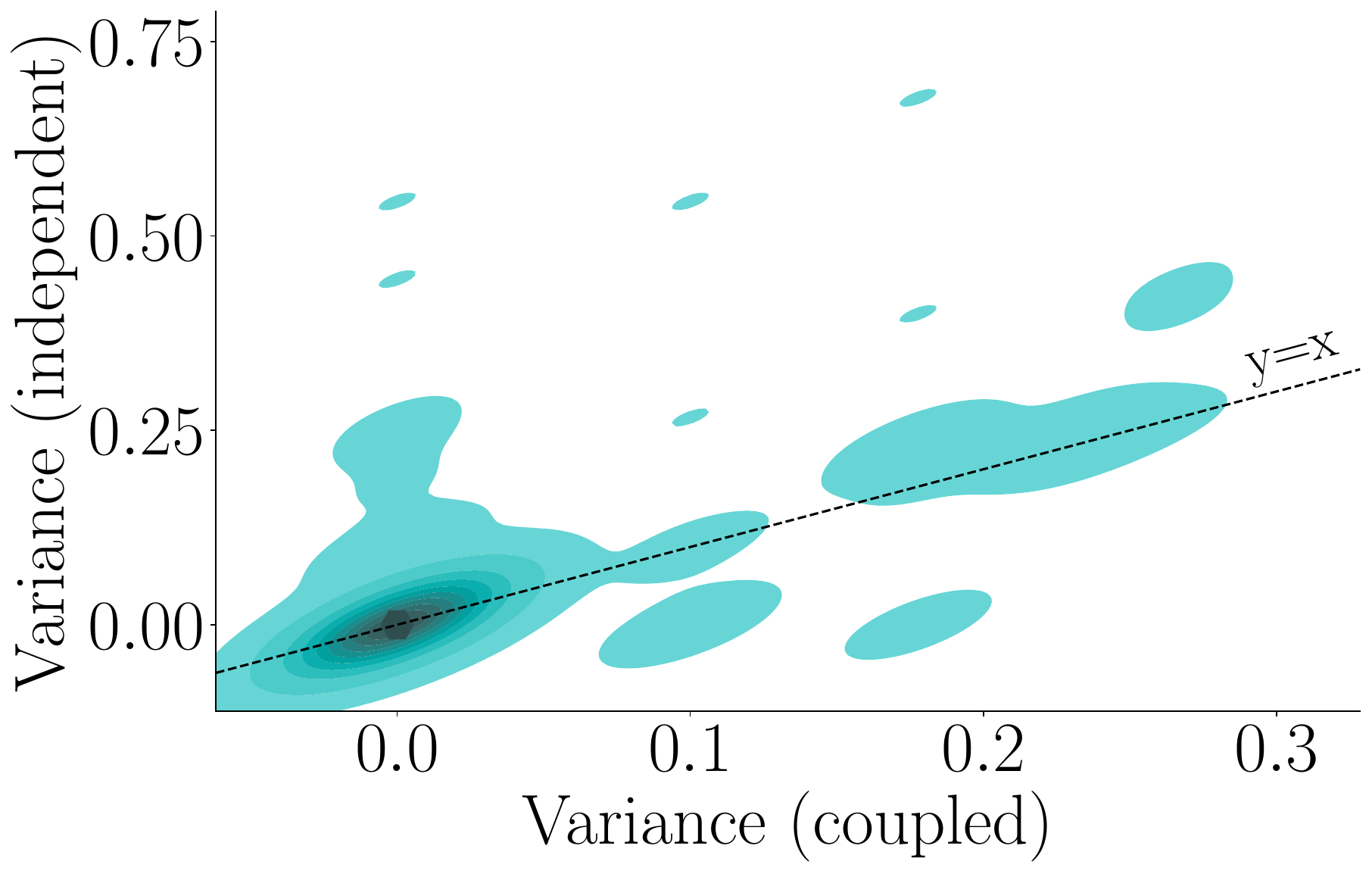} &
    \includegraphics[width=0.23\linewidth]{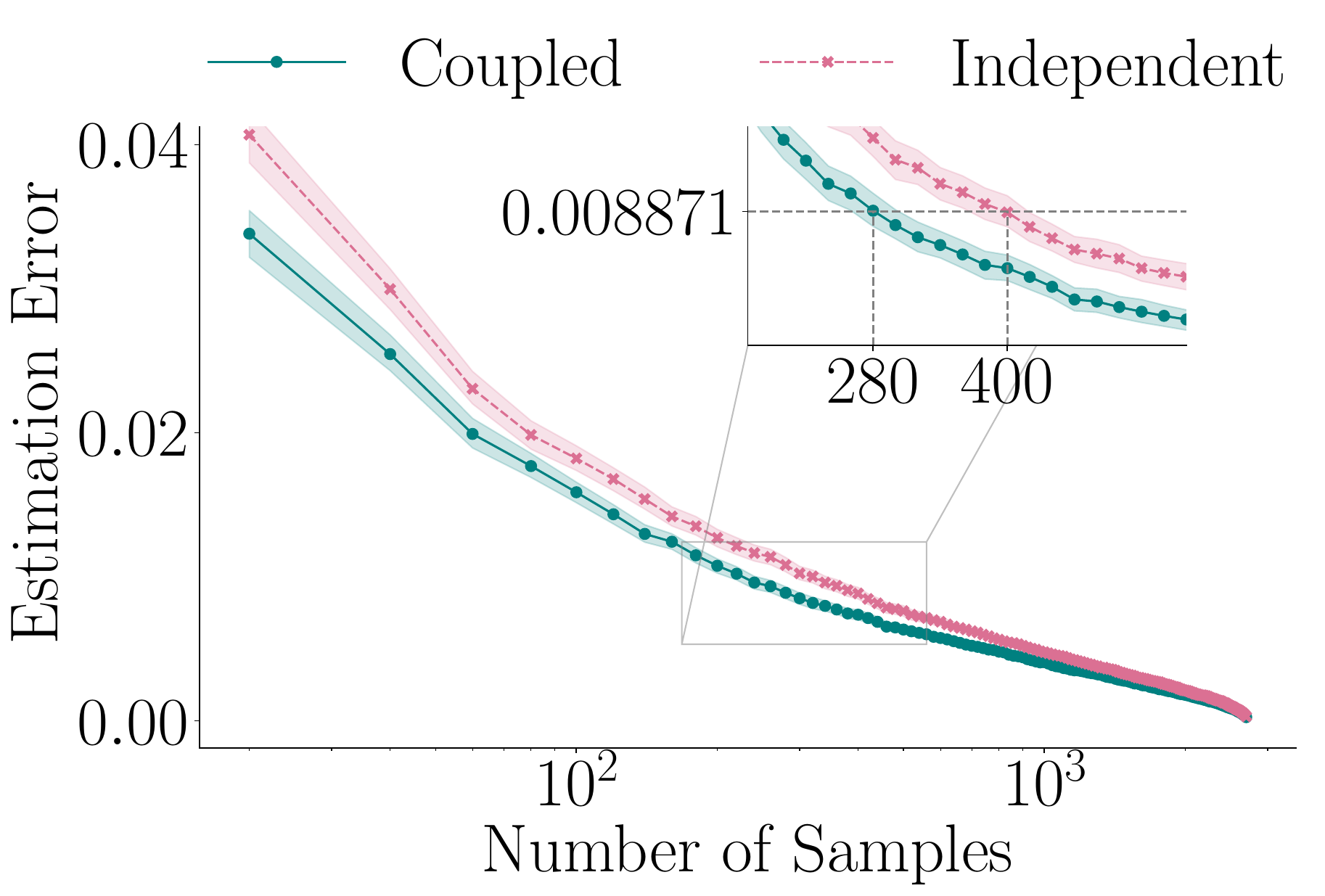} \\
    (a) Score covariance & (b) Variance of the score difference & (c) Estimation error vs. \# samples \\ 
\end{tabular}
    \caption{\textbf{Comparison between \texttt{2.5-7B} and \texttt{2.5-7B-bnb-8bit} from the \texttt{Qwen} family on multiple-choice questions from four knowledge areas of the MMLU dataset.}
    Panels in column (a) show the kernel density estimate (KDE) of the covariance between the scores of the two LLMs on each question under coupled generation; the dashed lines correspond to average values. Panels in column (b) show the KDE of the variance of the difference between the scores of the LLMs on each question under coupled and independent generation; the highlighted points correspond to median values. Panels in column (c) show the absolute error in the estimation of the expected difference between the scores of the LLMs against the number of samples; for each point on the x-axis, we perform $1{,}000$ sub-samplings and shaded areas correspond to $95\%$ confidence intervals. We observe qualitatively similar results for other knowledge areas.}
    \label{fig:mmlu-qwen-7B-vs-7B-8bit-areas}
\end{figure}

Here, we experiment with models from the \texttt{Qwen} family on the MMLU dataset following the setup described in Section~\ref{sec:mmlu}. 
We find that, for $\sim $$40$\% of the pairs of models shown in Table~\ref{tab:qwen-names}, coupled generation leads to at least a $10\%$ reduction in the number of samples required to achieve equivalent error in the estimation of the expected difference between the scores of the LLMs on the knowledge area \texttt{college computer science}.
For brevity, we only show the results for the above mentioned pairs in Figures~\ref{fig:mmlu-qwen-first-4}--\ref{fig:mmlu-qwen-last-3}.
Further, Figure~\ref{fig:mmlu-qwen-7B-vs-7B-8bit-areas} shows the results for \texttt{2.5-7B} against \texttt{2.5-7B-bnb-8bit} on different knowledge areas.
Overall, the results are qualitatively similar to those in Section~\ref{sec:mmlu}.

\vspace{-0.2cm}

\subsection{GSM8K and HumanEval Datasets}\label{app:qwen-gsm8k-human-eval}
\begin{figure}[!!h]
\centering
\begin{tabular}{c c c}
     \multicolumn{3}{c}{\texttt{2.5-7B} vs. \texttt{2.5-7B-bnb-8bit}}\\
    \includegraphics[width=0.23\linewidth]{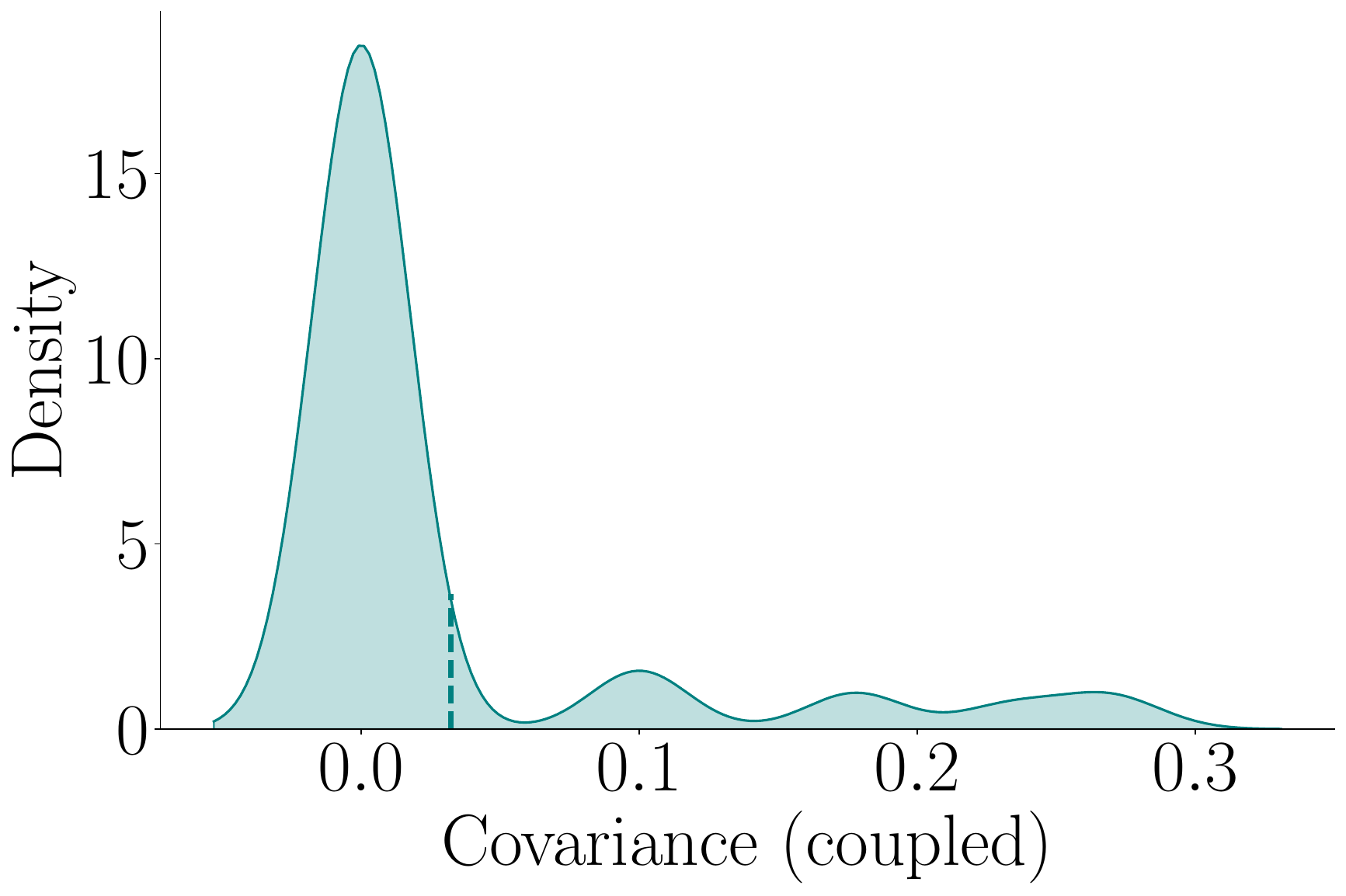} &
    \includegraphics[width=0.23\linewidth]{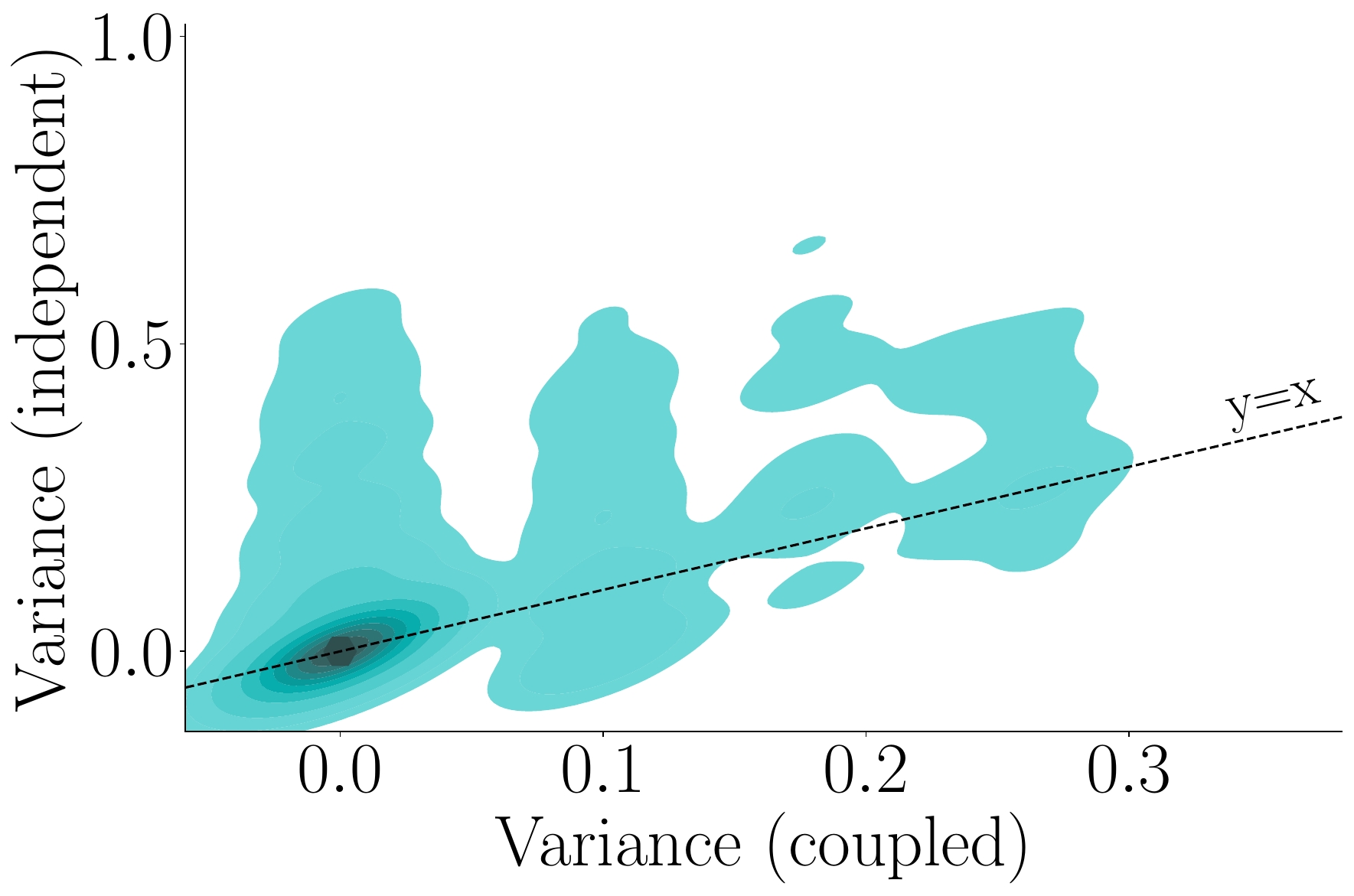} &
    \includegraphics[width=0.23\linewidth]{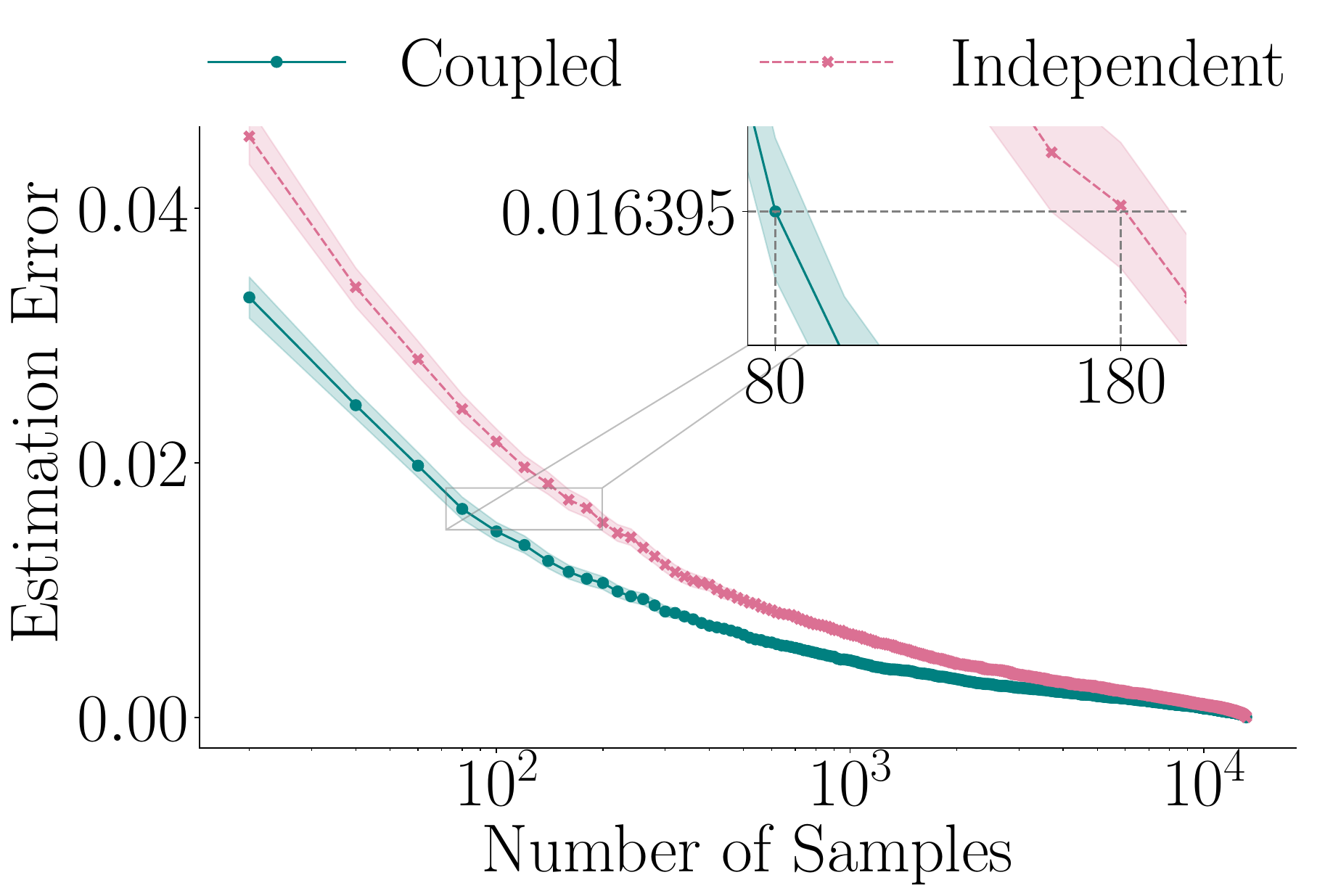} \\ \\
%
    \multicolumn{3}{c}{\texttt{2.5-7B} vs. \texttt{2.5-7B-bnb-4bit}}\\
    \includegraphics[width=0.23\linewidth]{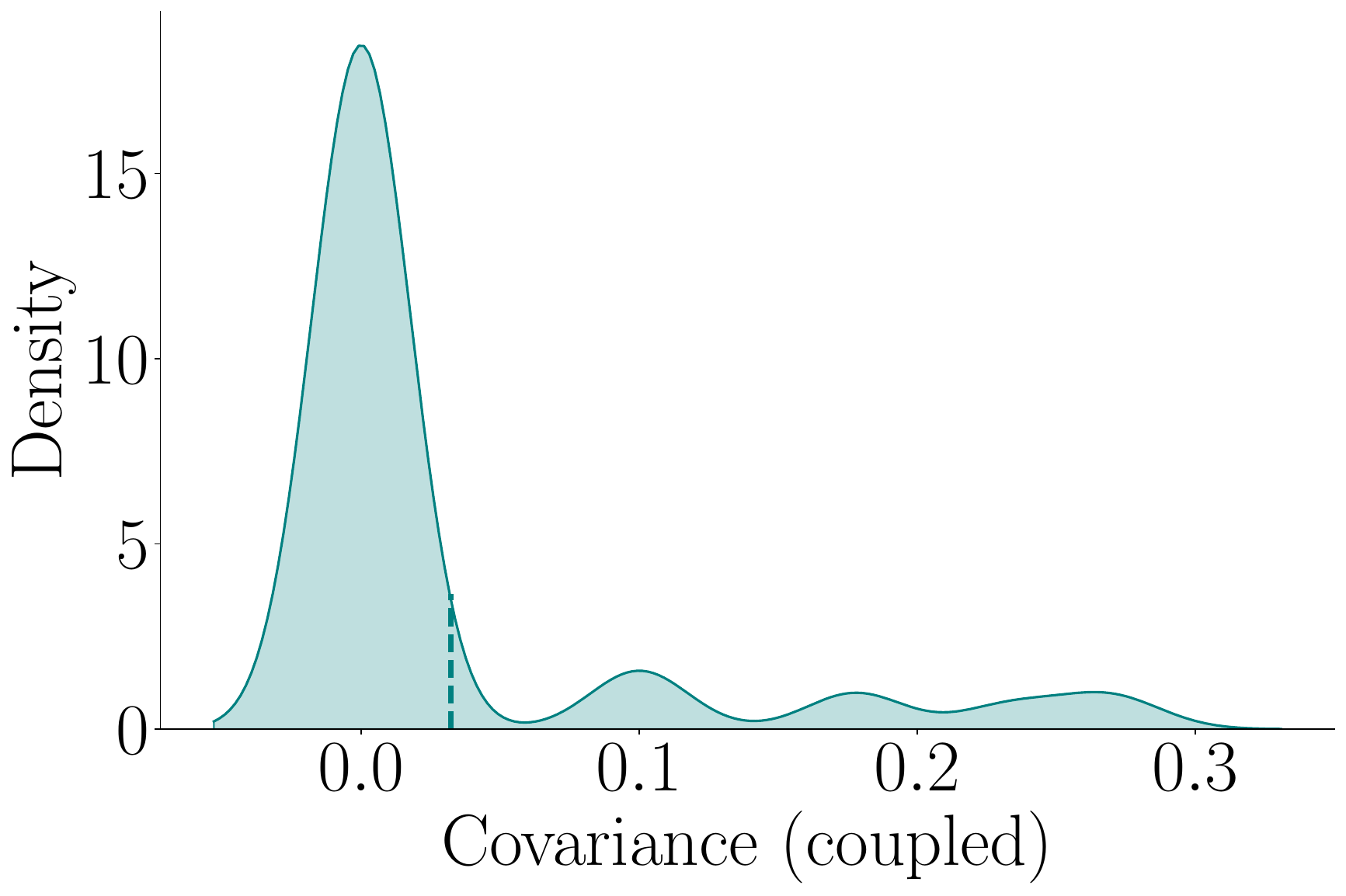} &
    \includegraphics[width=0.23\linewidth]{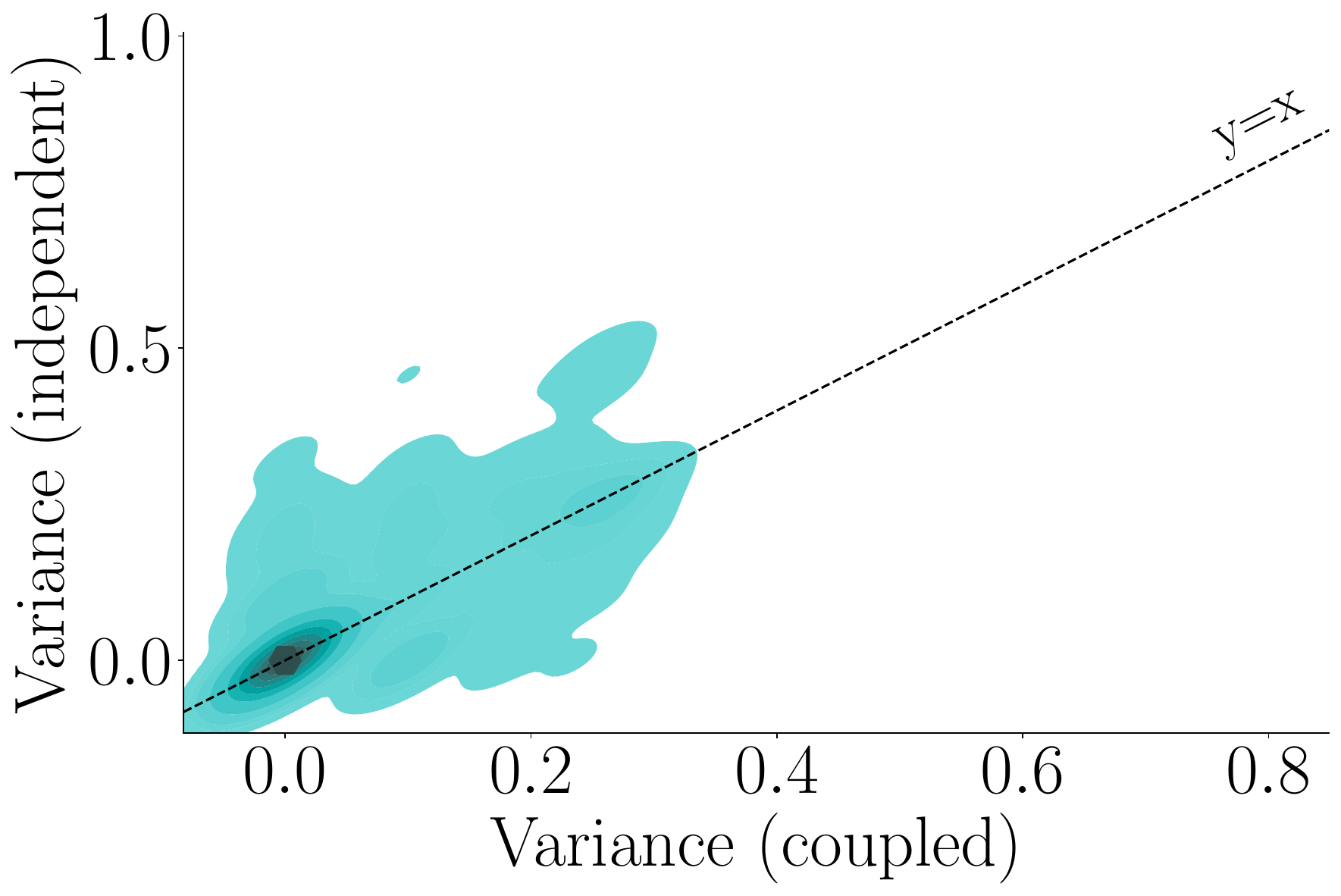} &
    \includegraphics[width=0.23\linewidth]{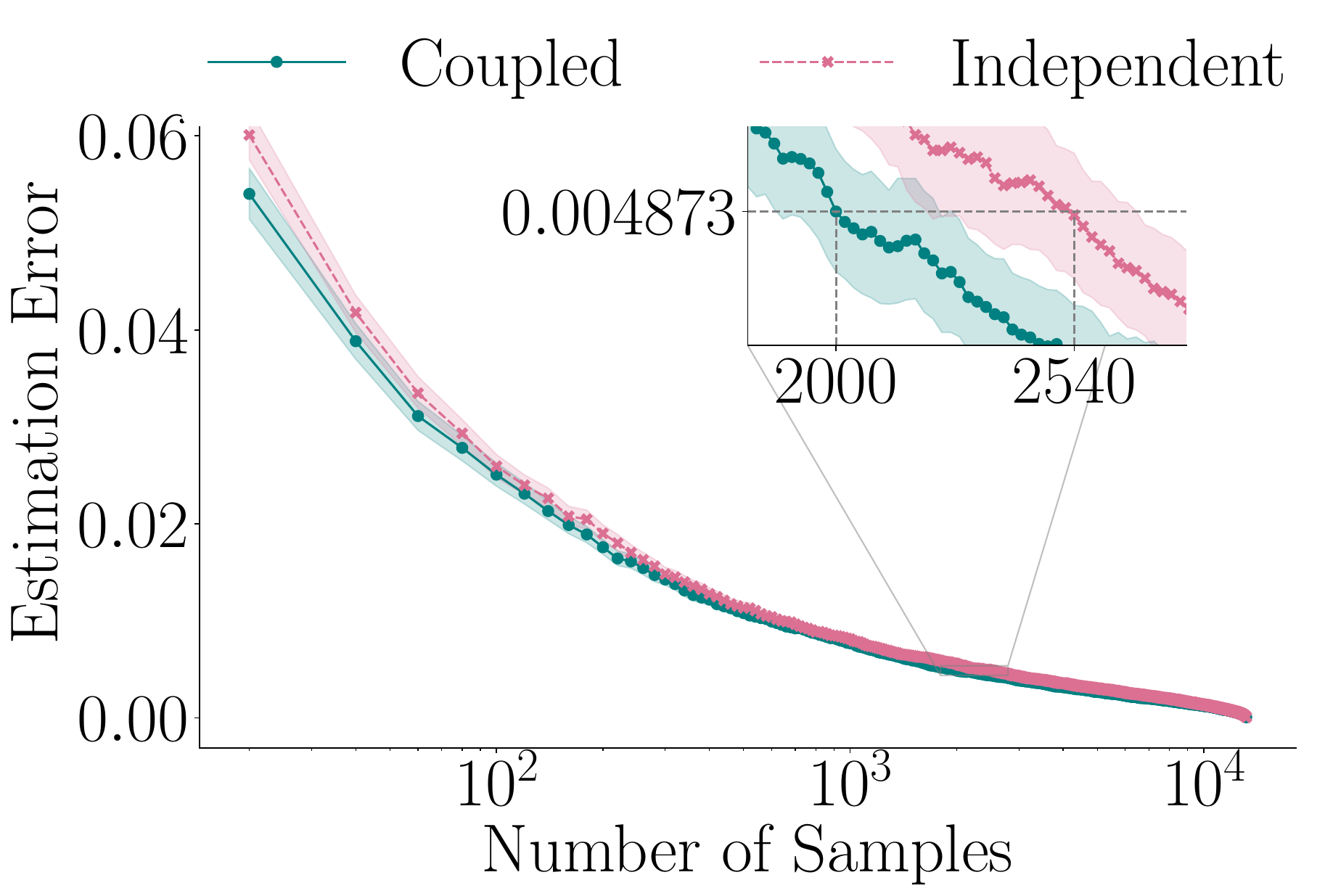} \\ \\
%
     \multicolumn{3}{c}{\texttt{3-8B} vs. \texttt{2.5-7B}}\\
    \includegraphics[width=0.23\linewidth]{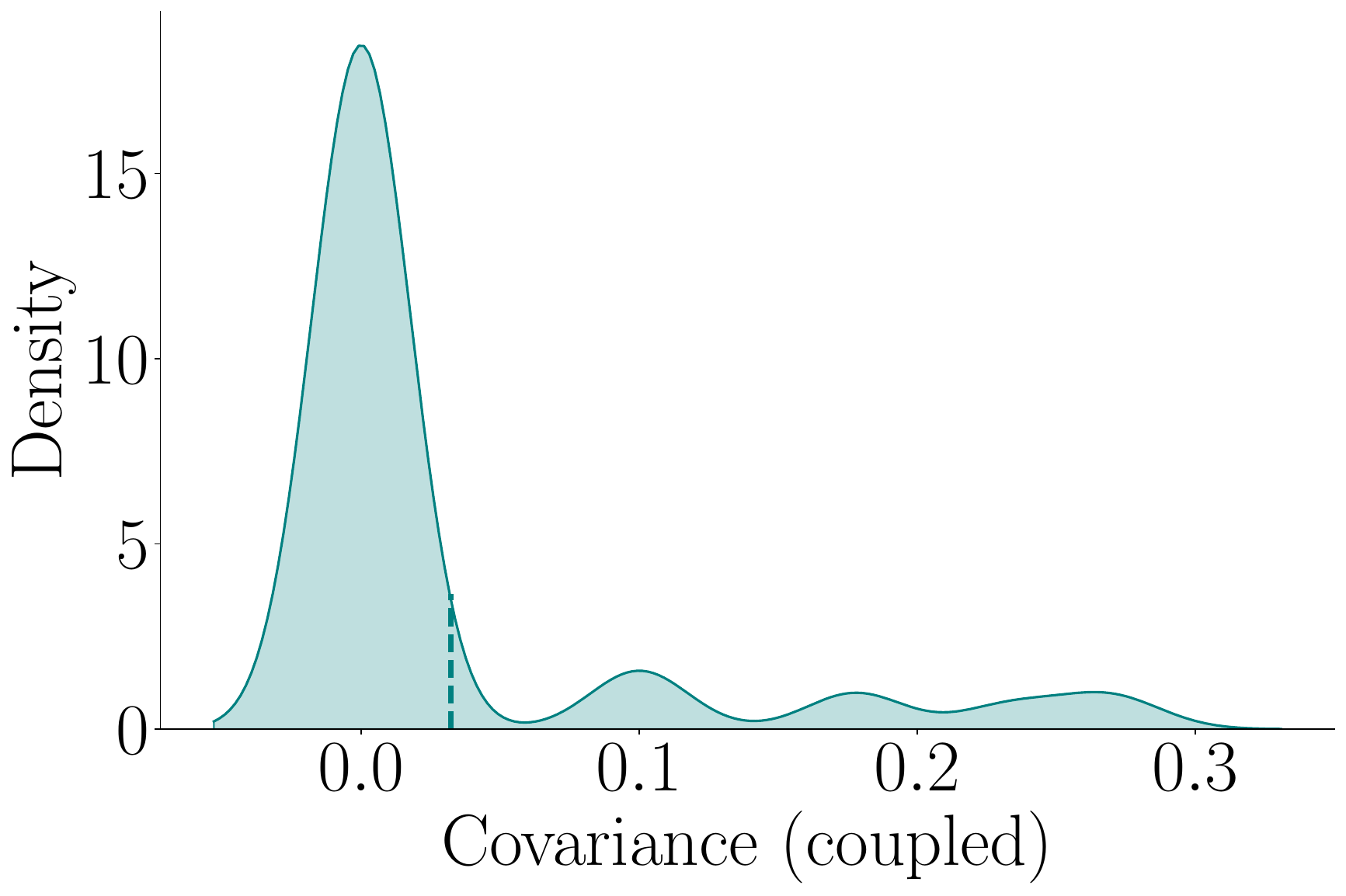} &
    \includegraphics[width=0.23\linewidth]{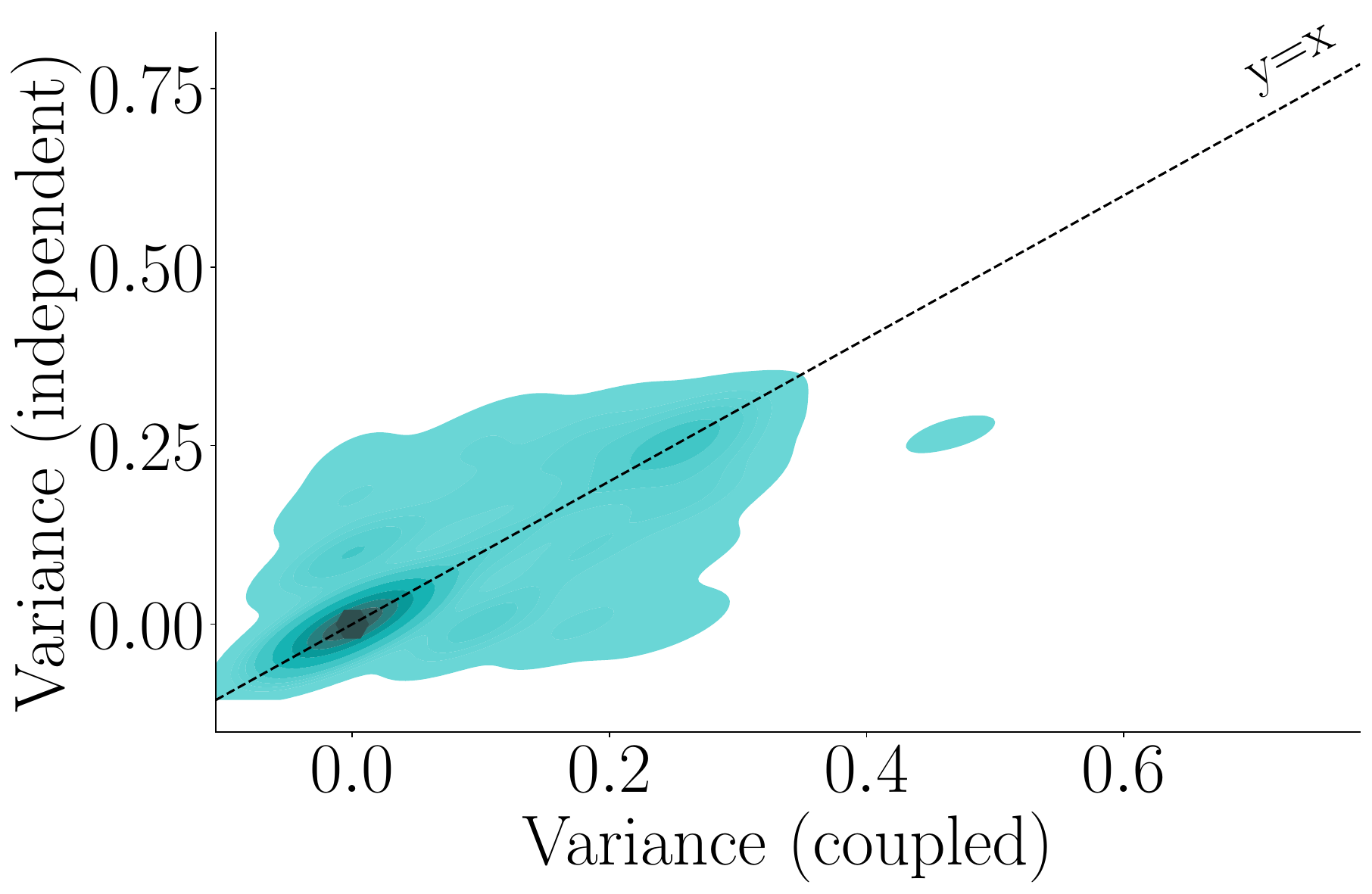} &
    \includegraphics[width=0.23\linewidth]{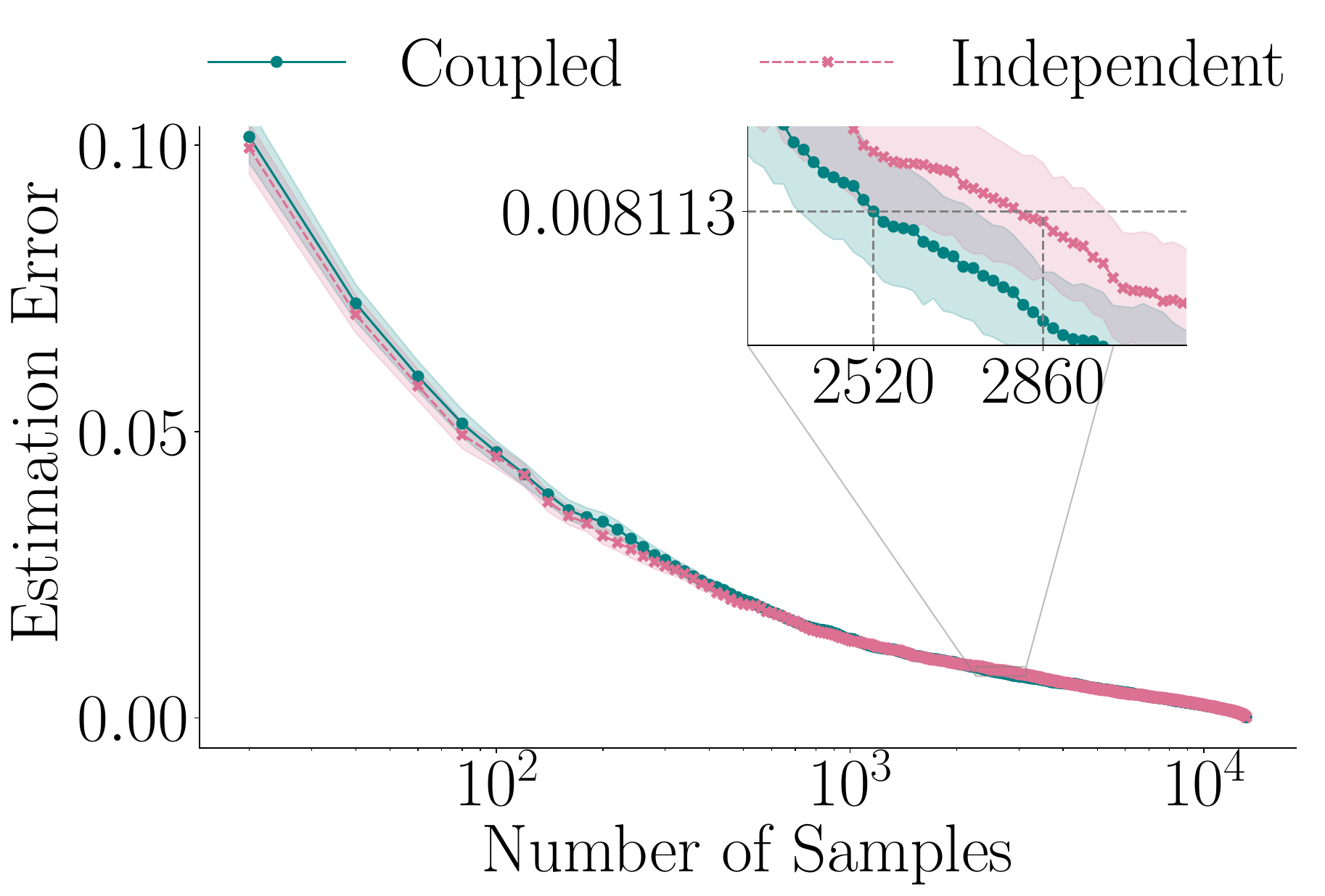} \\ \\
%
  \multicolumn{3}{c}{\texttt{2.5-3B} vs. \texttt{2.5-7B-bnb-4bit}}\\
    \includegraphics[width=0.23\linewidth]{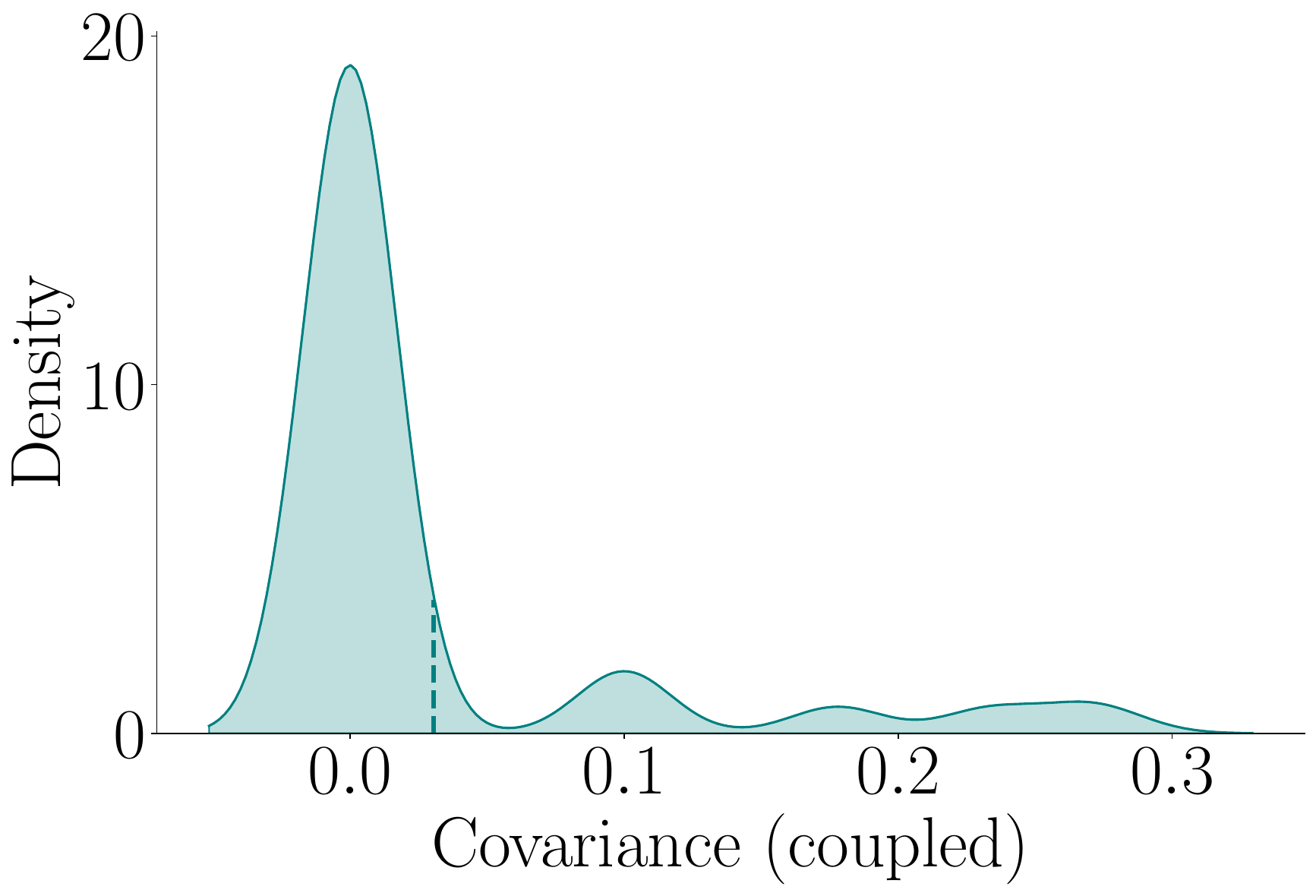} &
    \includegraphics[width=0.23\linewidth]{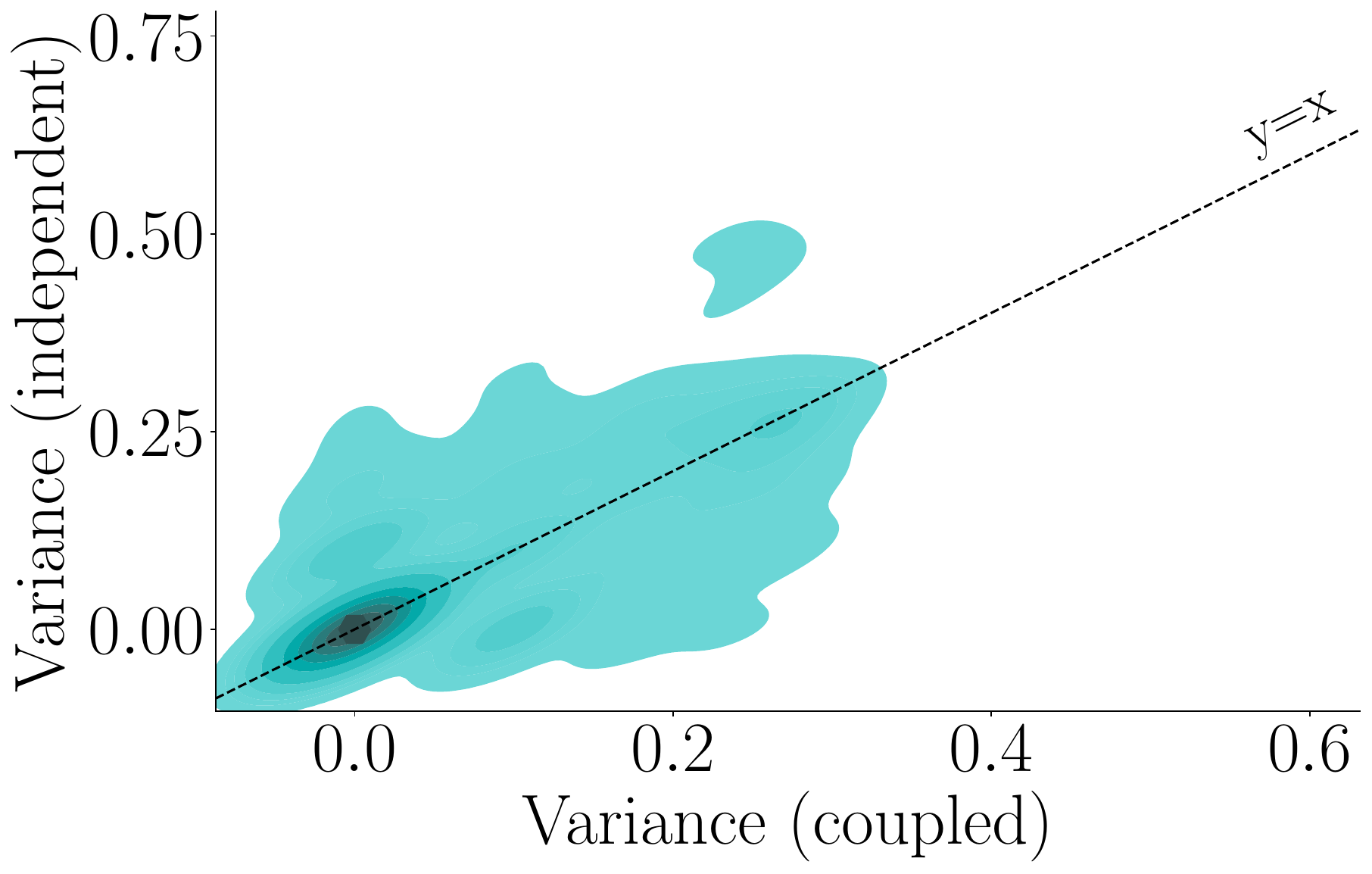} &
    \includegraphics[width=0.23\linewidth]{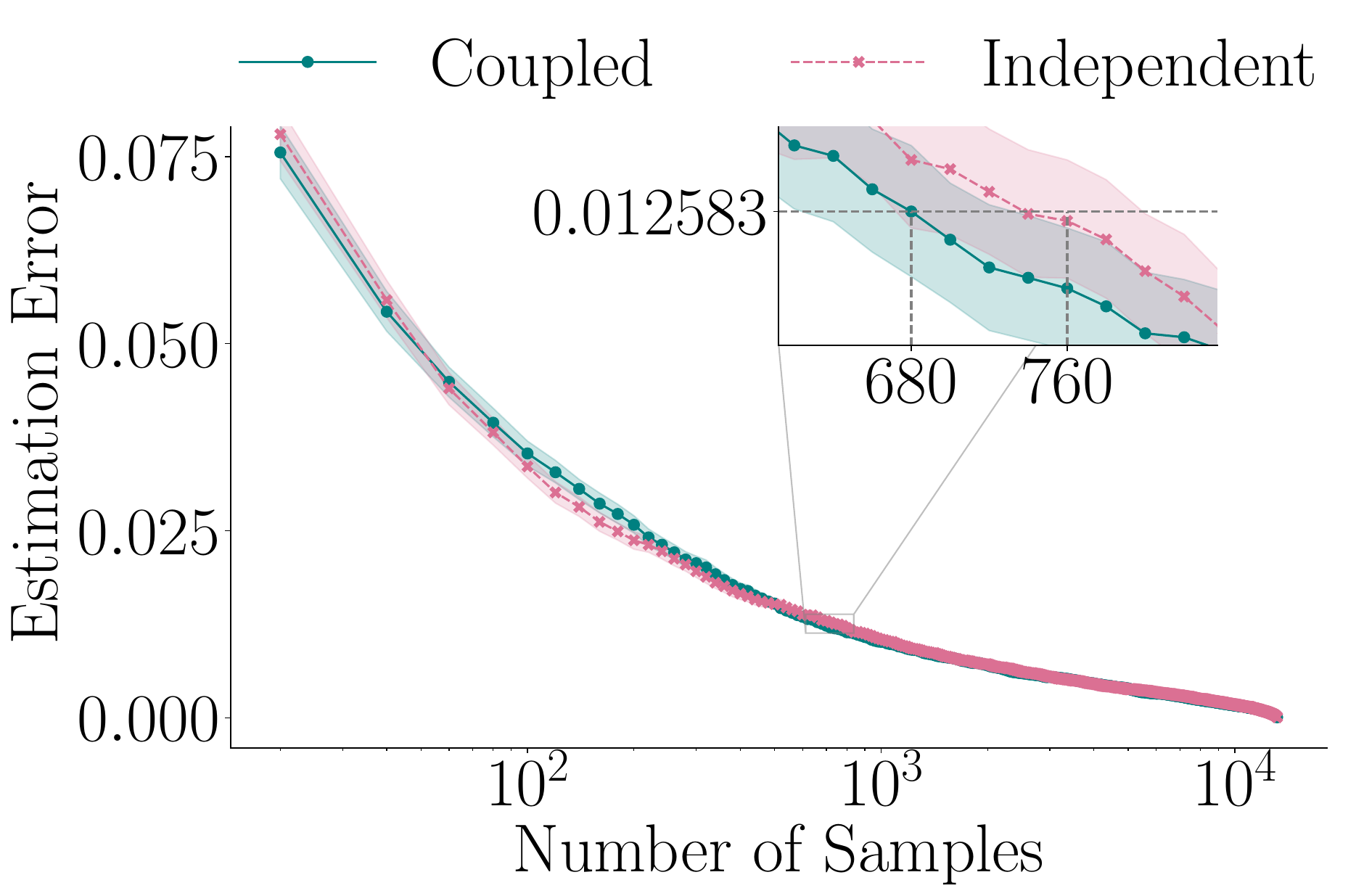} \\ \\

\multicolumn{3}{c}{\texttt{2.5-3B} vs. \texttt{2.5-3B-distil}}\\
    \includegraphics[width=0.23\linewidth]{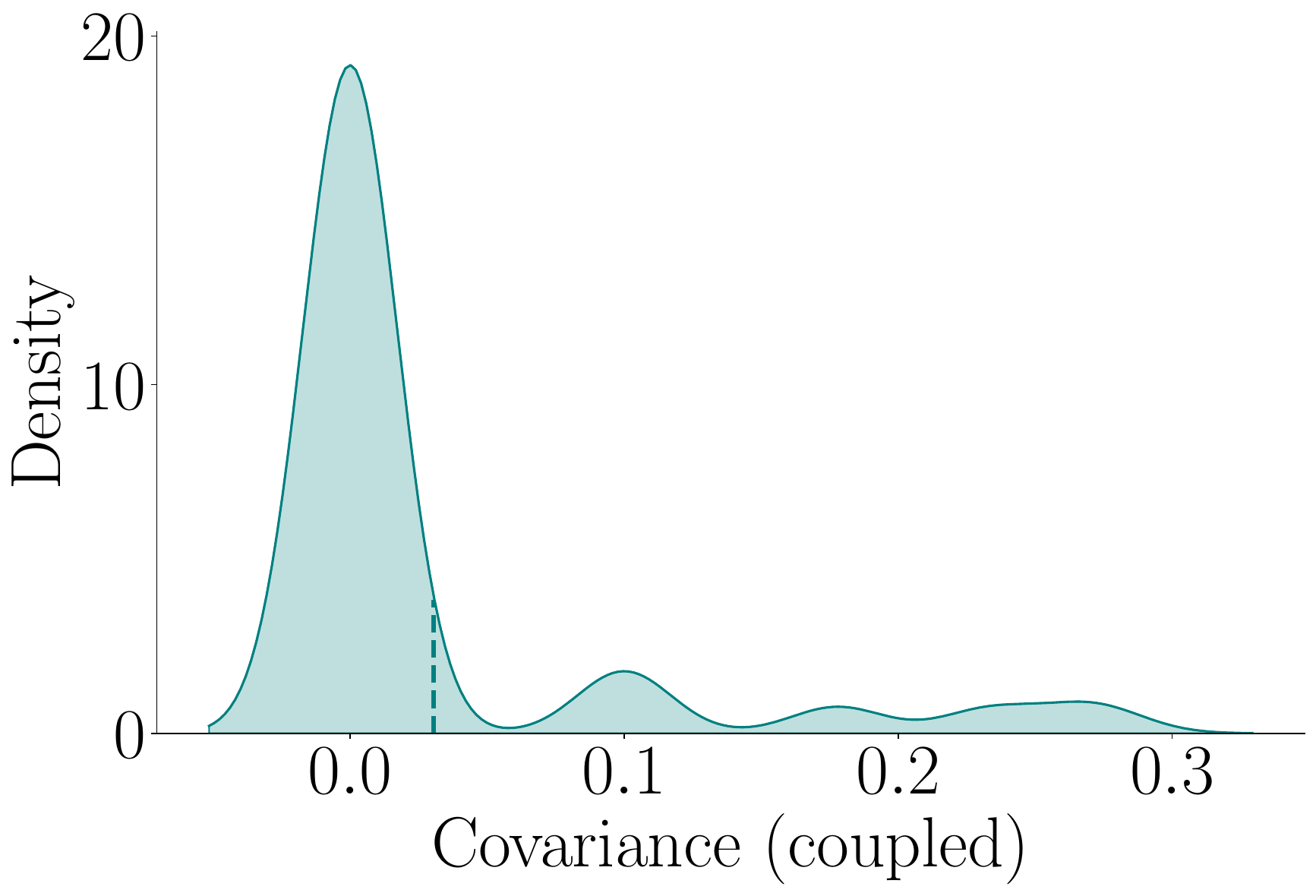} &
    \includegraphics[width=0.23\linewidth]{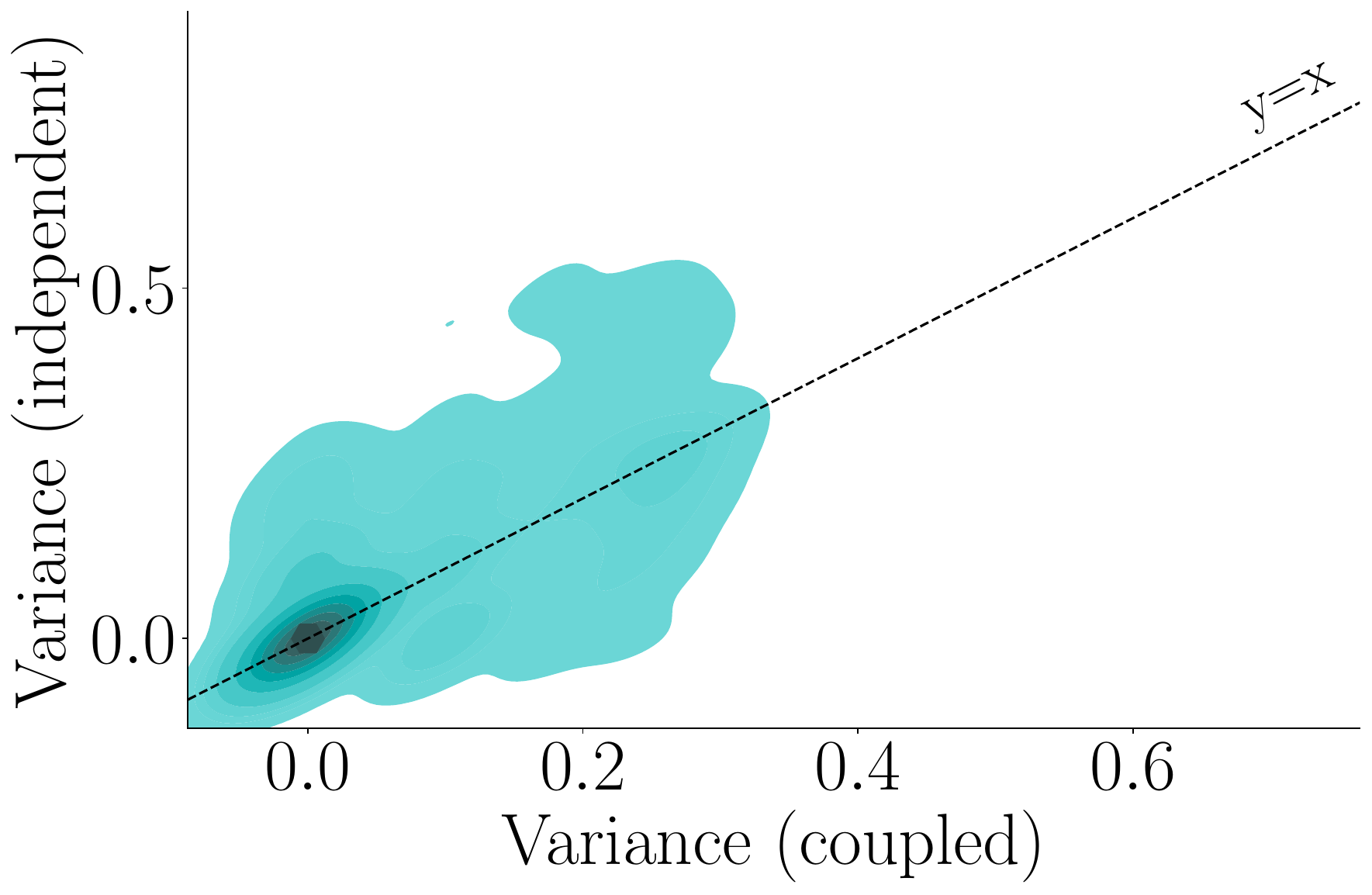} &
    \includegraphics[width=0.23\linewidth]{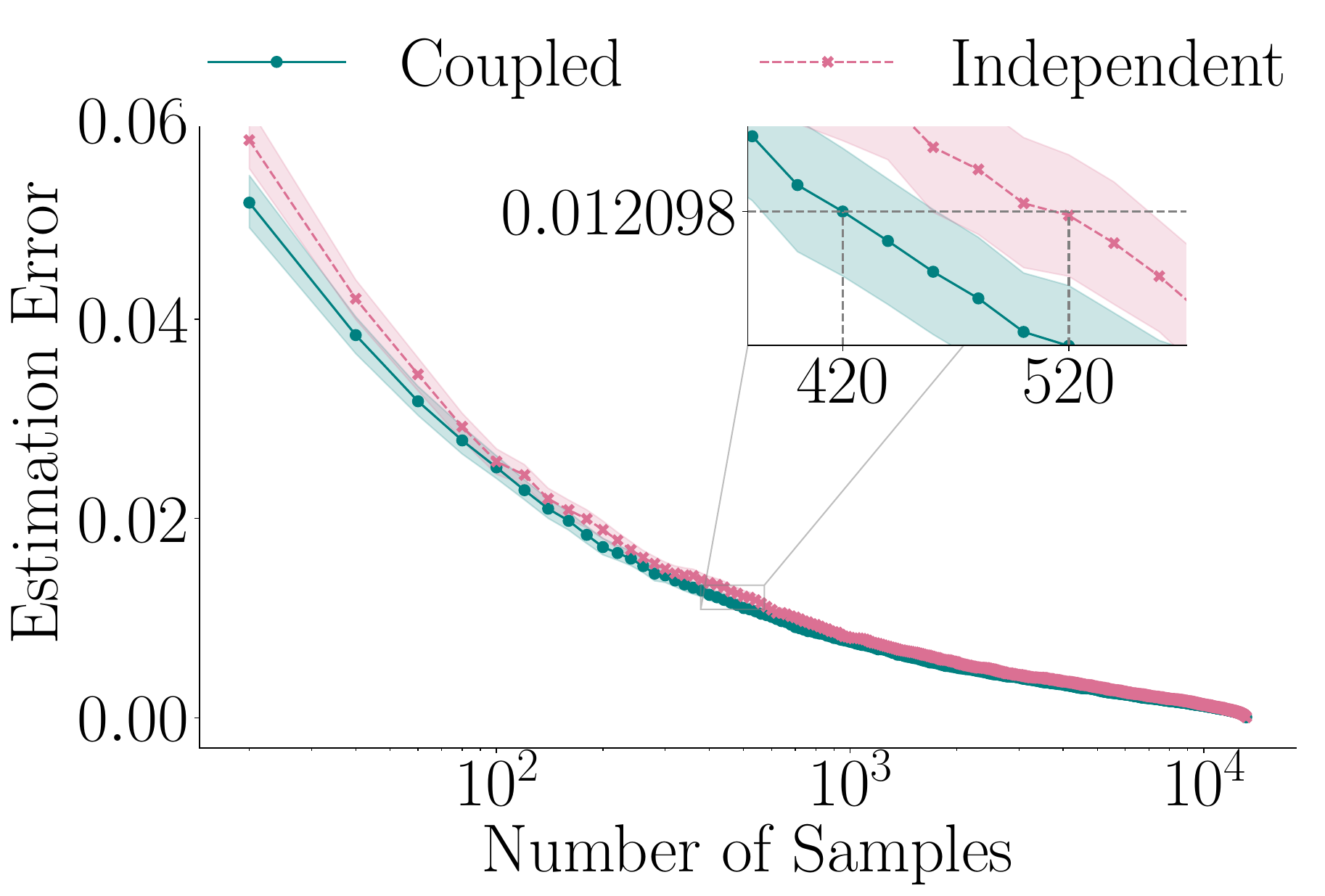} \\ \\
    
    (a) Score covariance & (b) Variance of the score difference & (c) Estimation error vs. \# samples \\ 
\end{tabular}
    \caption{\textbf{Comparison between several pairs of LLMs in the \texttt{Qwen} family on math questions from the GSM8K dataset.}
    Panels in column (a) show the kernel density estimate (KDE) of the covariance between the scores of the two LLMs on each problem under coupled generation; the dashed lines correspond to average values. Panels in column (b) show the KDE of the variance of the difference between the scores of the LLMs on each question under coupled and independent generation; the highlighted points correspond to median values. Panels in column (c) show the absolute error in the estimation of the expected difference between the scores of the LLMs against the number of samples; for each point on the x-axis, we perform $1{,}000$ sub-samplings and shaded areas correspond to $95\%$ confidence intervals.}
    \label{fig:gsm8k-qwen-first-5}
\end{figure}

Here, we experiment with models in the \texttt{Qwen} family on the GSM8K and HumanEval datasets following the setup described in Appendix~\ref{app:llama-gsm8k-humaneval}. 
We find that, for $\sim$61\% and $\sim$$32\%$ of the pairs of models shown in Table~\ref{tab:qwen-names}, coupled generation leads to at least a $10\%$ reduction in the number of samples required to achieve equivalent error in the estimation of the expected difference between the scores of the LLMs on the GSM8K dataset and the HumanEval dataset, respectively.
For brevity, we only show the results for the above mentioned pairs in Figures~\ref{fig:gsm8k-qwen-first-5}--\ref{fig:human-eval-qwen-last-4}. 
Overall, the results are qualitatively similar to those in Appendix~\ref{app:llama-gsm8k-humaneval}.

\vspace{-0.2cm}

\begin{figure}[h]
\centering
\begin{tabular}{c c c}
    \multicolumn{3}{c}{\texttt{2.5-1.5B} vs. \texttt{2.5-7B-bnb-4bit}}\\
    \includegraphics[width=0.23\linewidth]{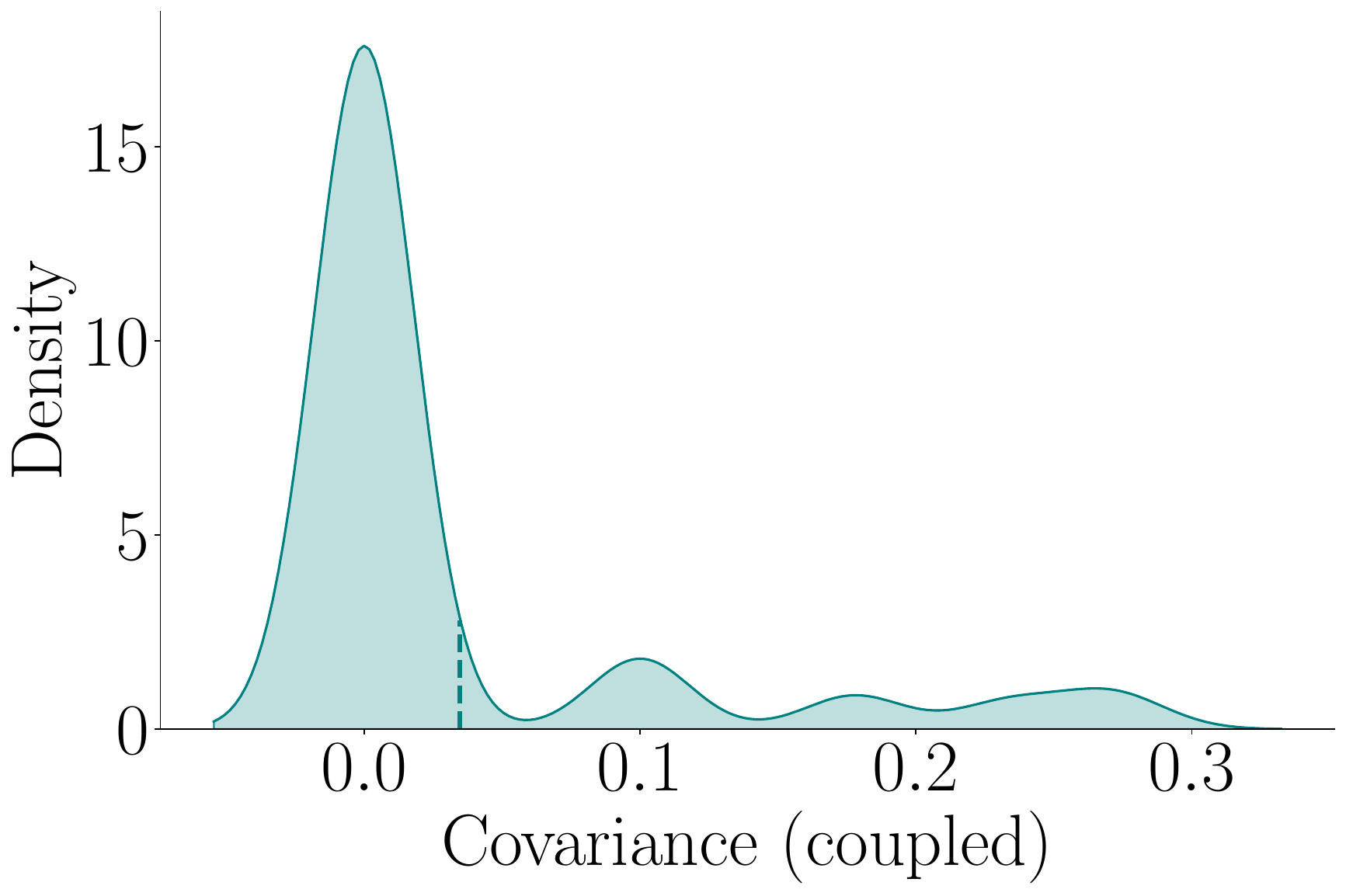} &
    \includegraphics[width=0.23\linewidth]{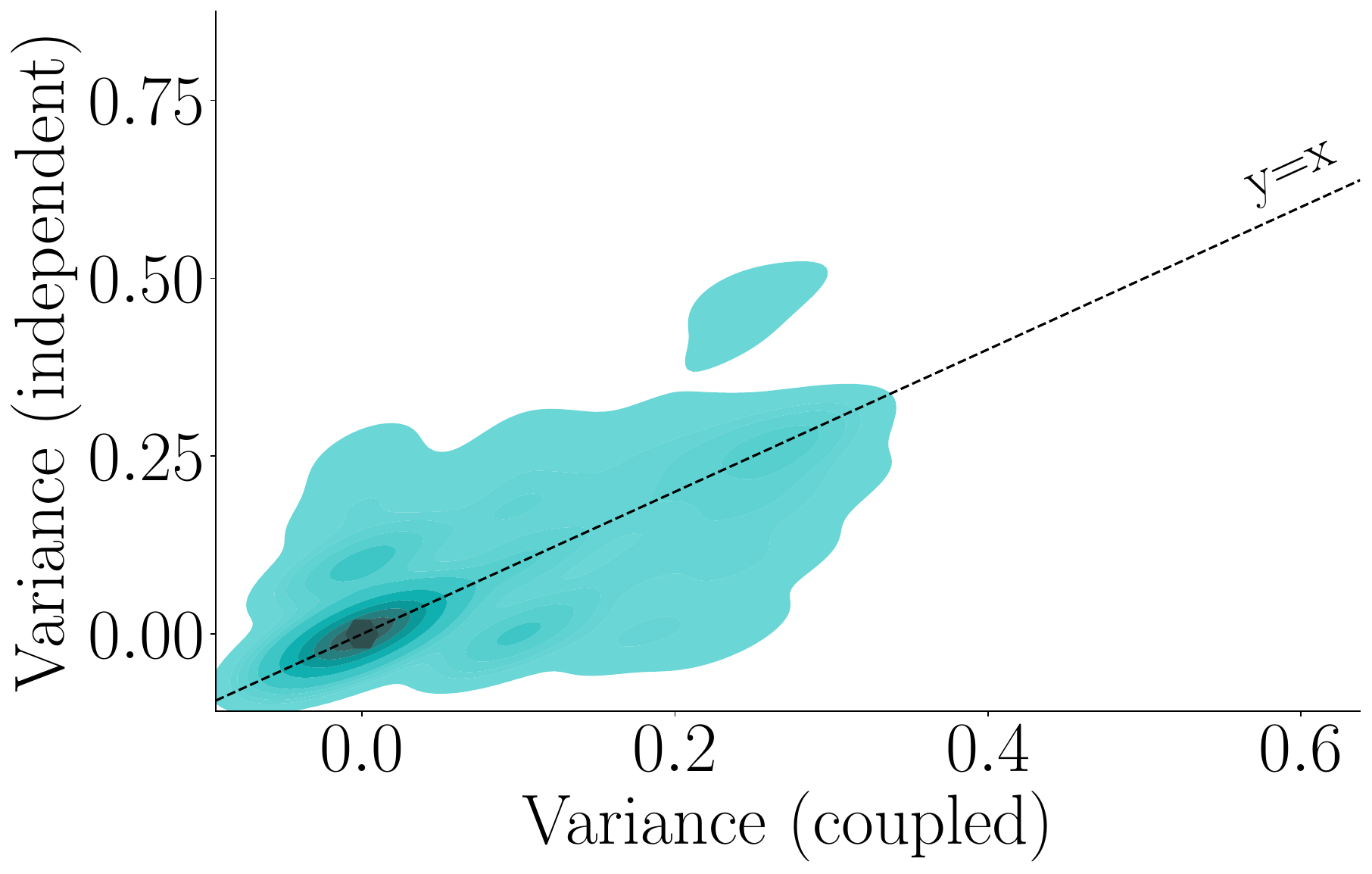} &
    \includegraphics[width=0.23\linewidth]{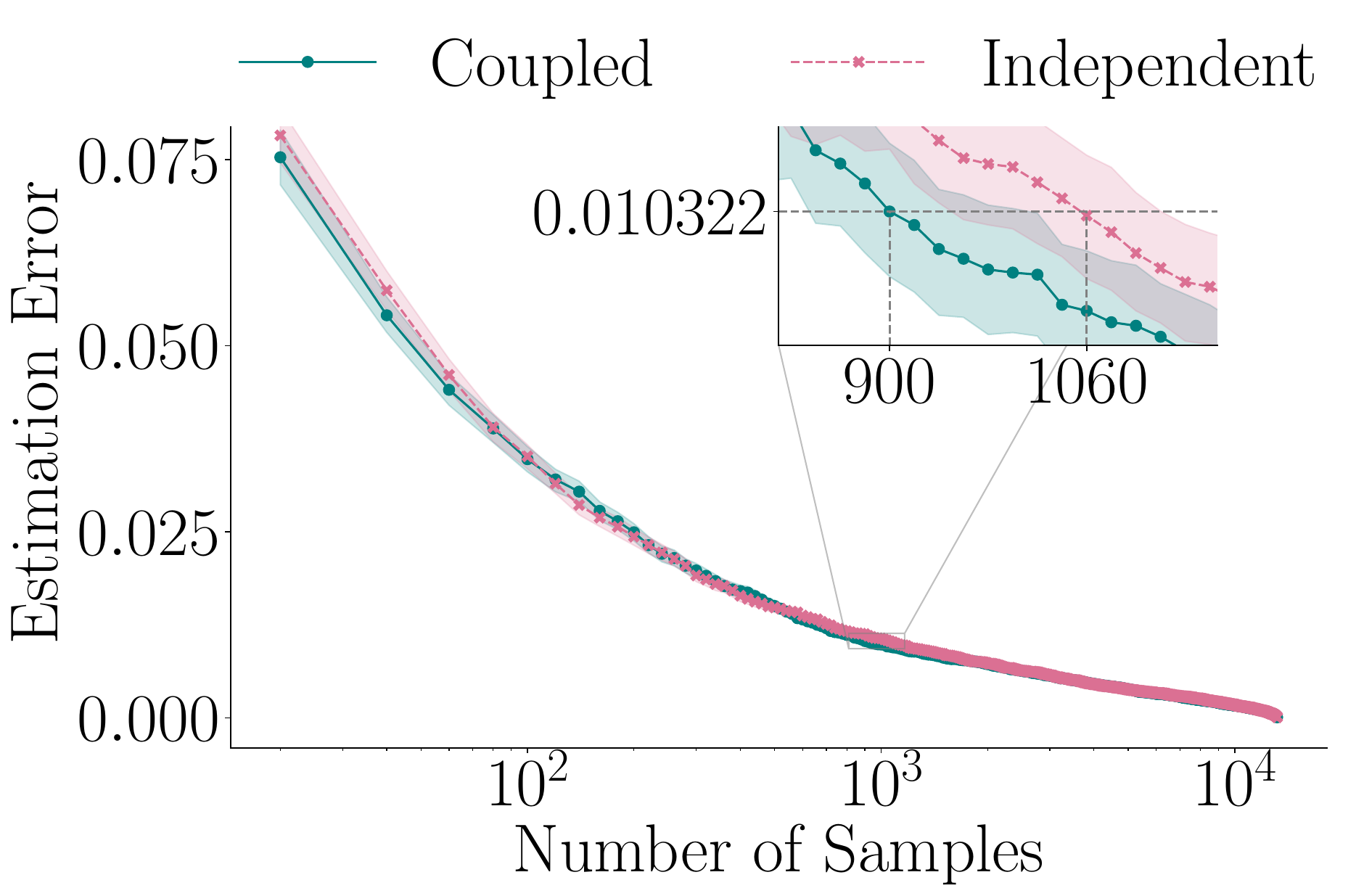} \\ \\
    \multicolumn{3}{c}{\texttt{2.5-1.5B} vs. \texttt{2.5-3B}}\\
    \includegraphics[width=0.23\linewidth]{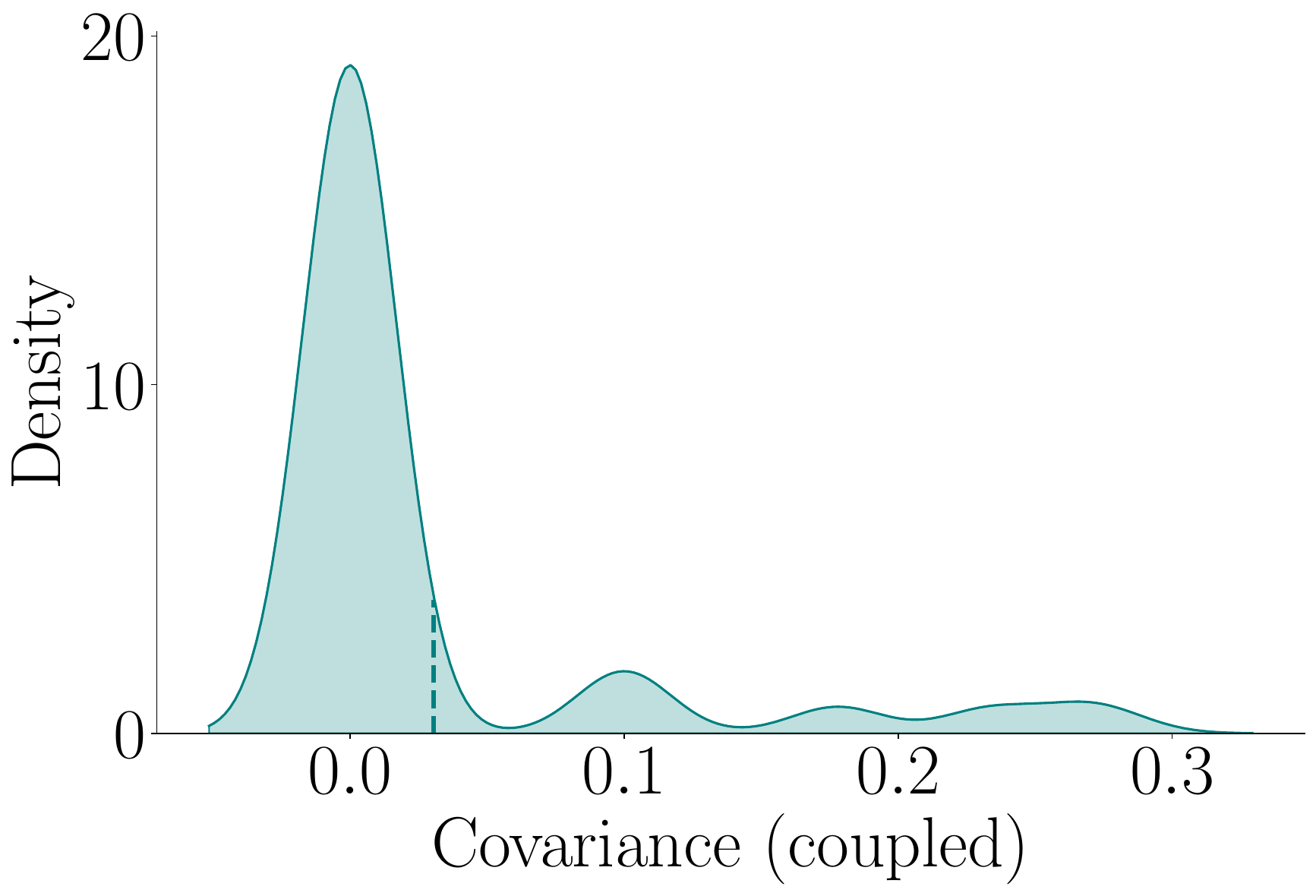} &
    \includegraphics[width=0.23\linewidth]{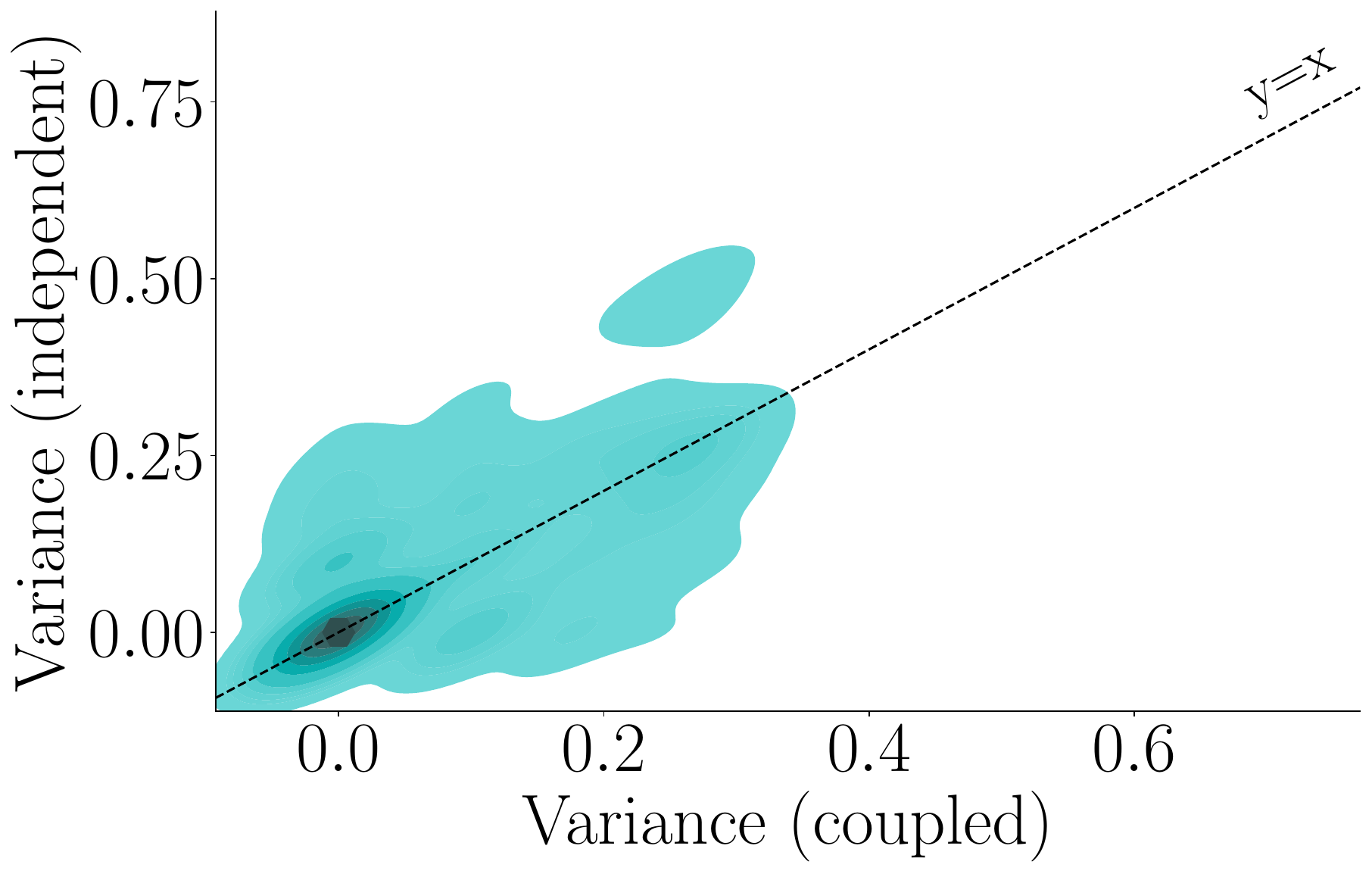} &
    \includegraphics[width=0.23\linewidth]{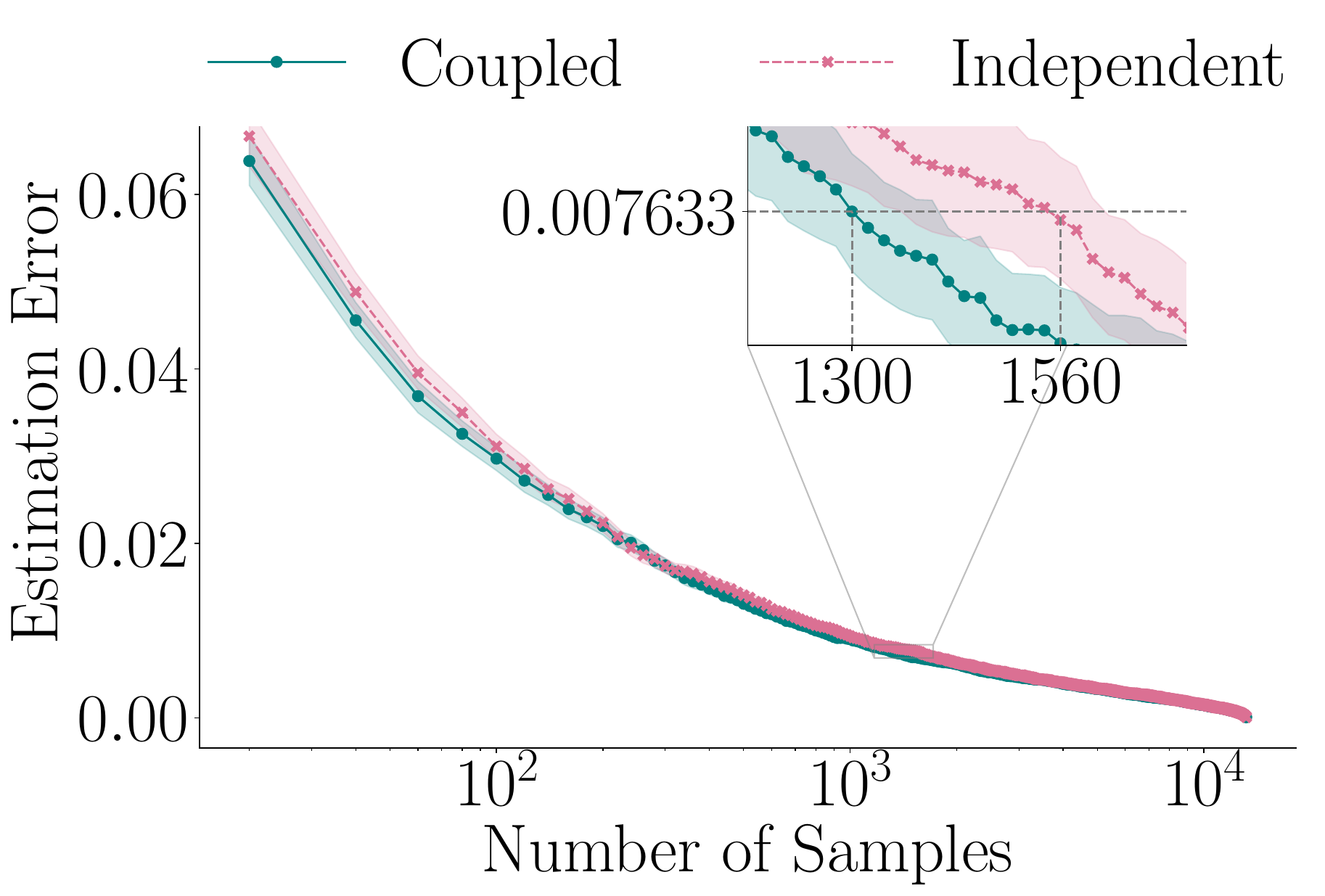} \\ \\
     \multicolumn{3}{c}{\texttt{2.5-1.5B} vs. \texttt{2.5-3B-distil}}\\
    \includegraphics[width=0.23\linewidth]{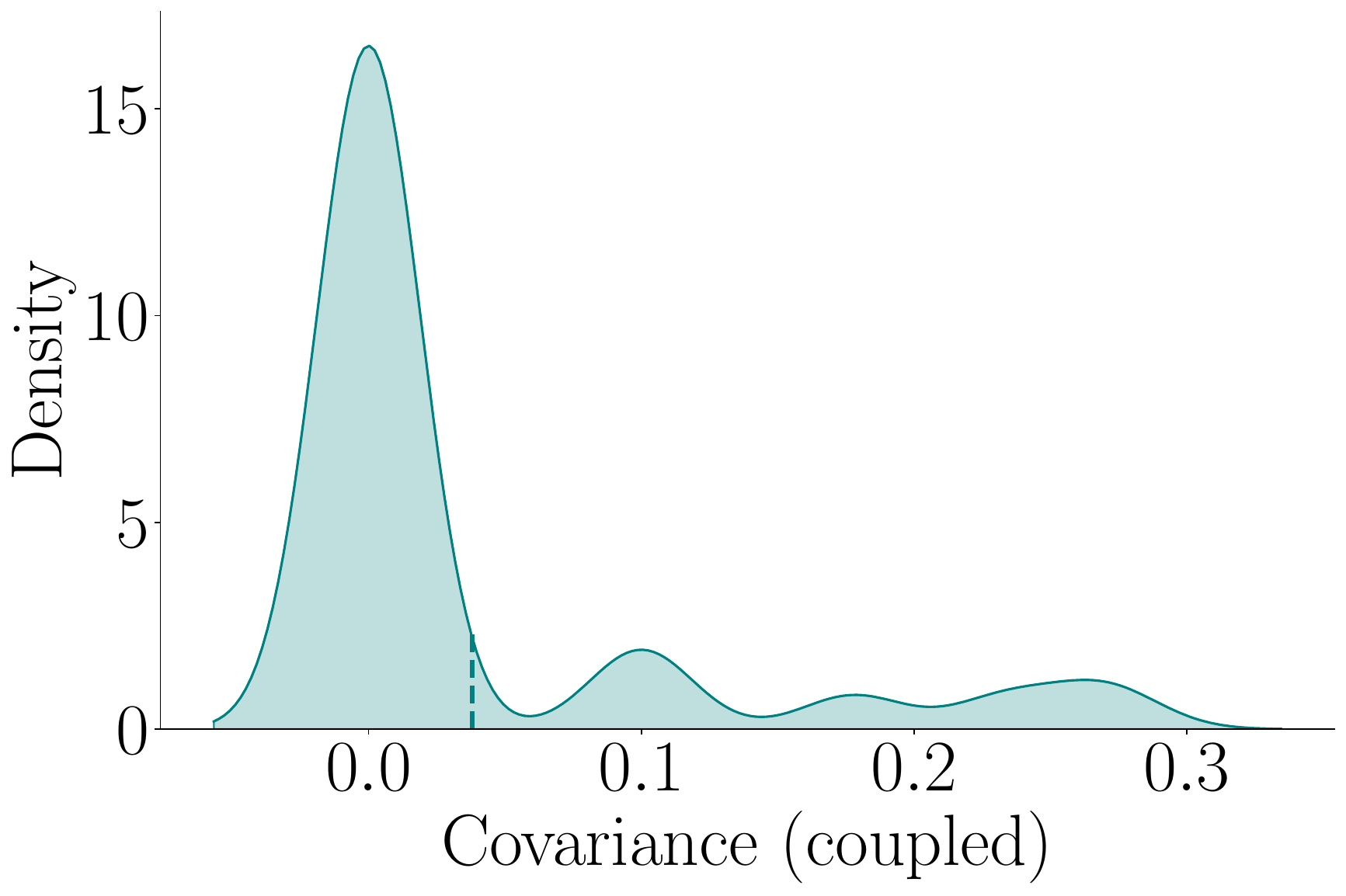} &
    \includegraphics[width=0.23\linewidth]{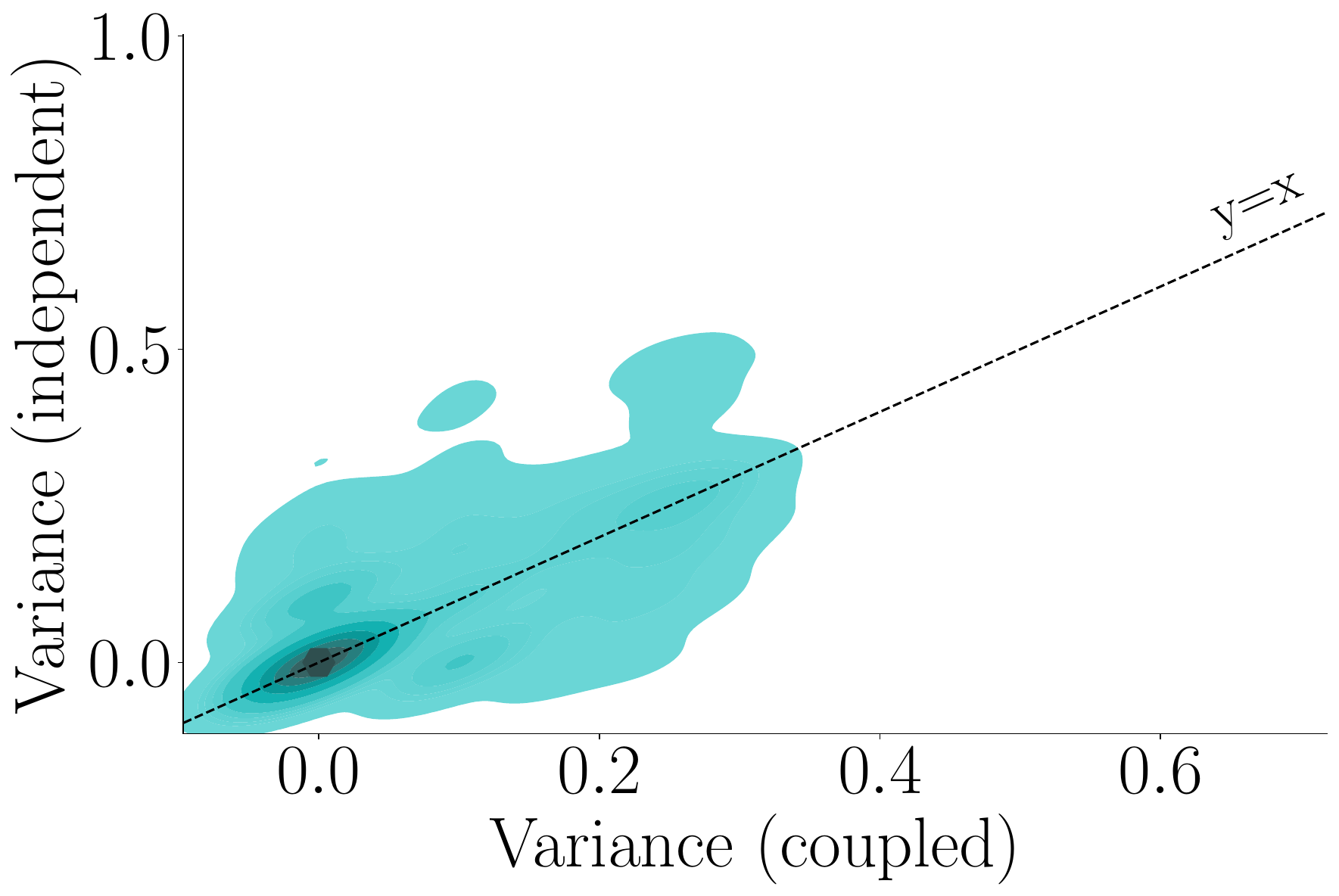} &
    \includegraphics[width=0.23\linewidth]{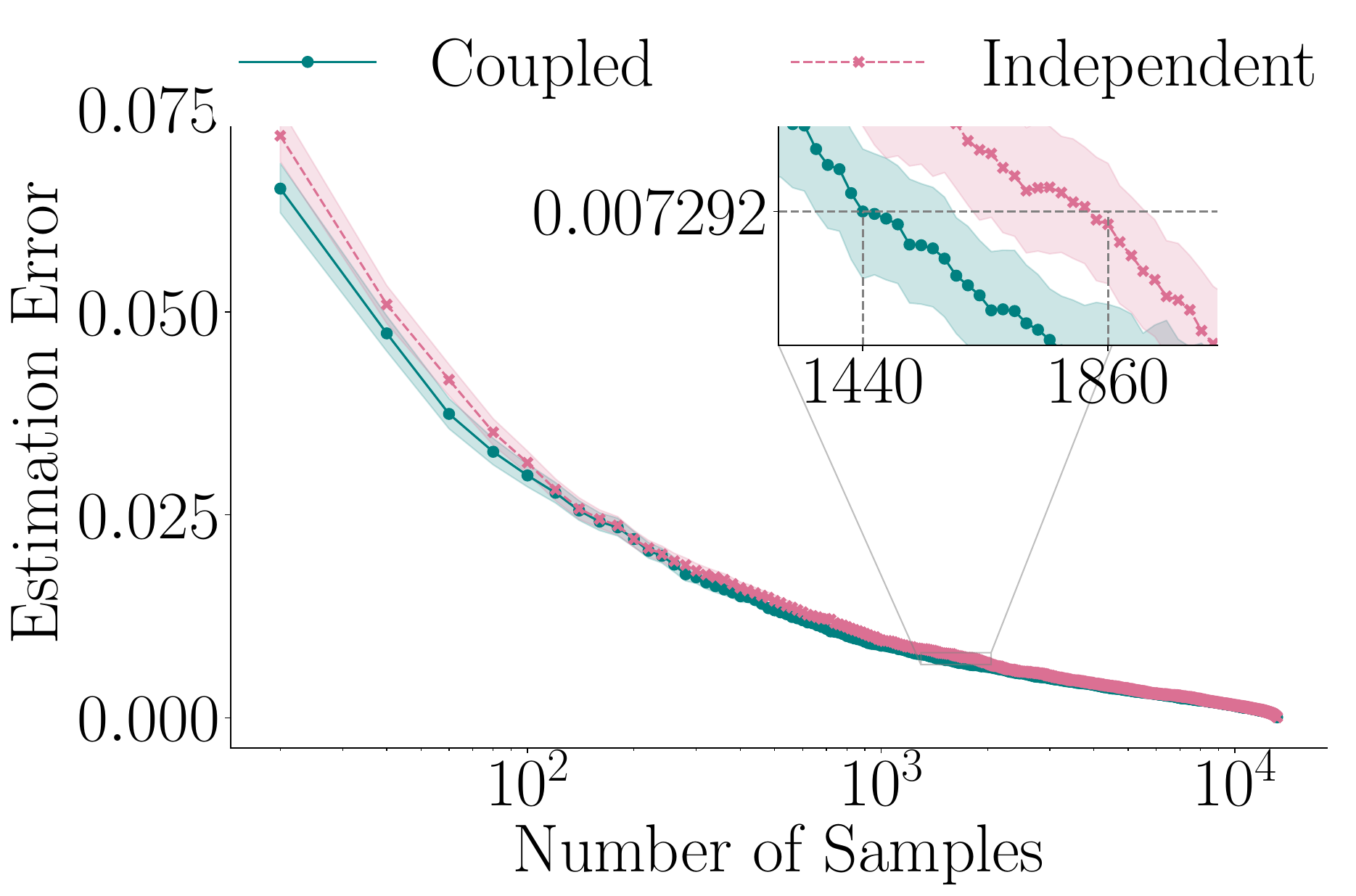} \\ \\
 \multicolumn{3}{c}{\texttt{3-8B} vs. \texttt{2.5-7B-bnb-4bit}}\\
    \includegraphics[width=0.23\linewidth]{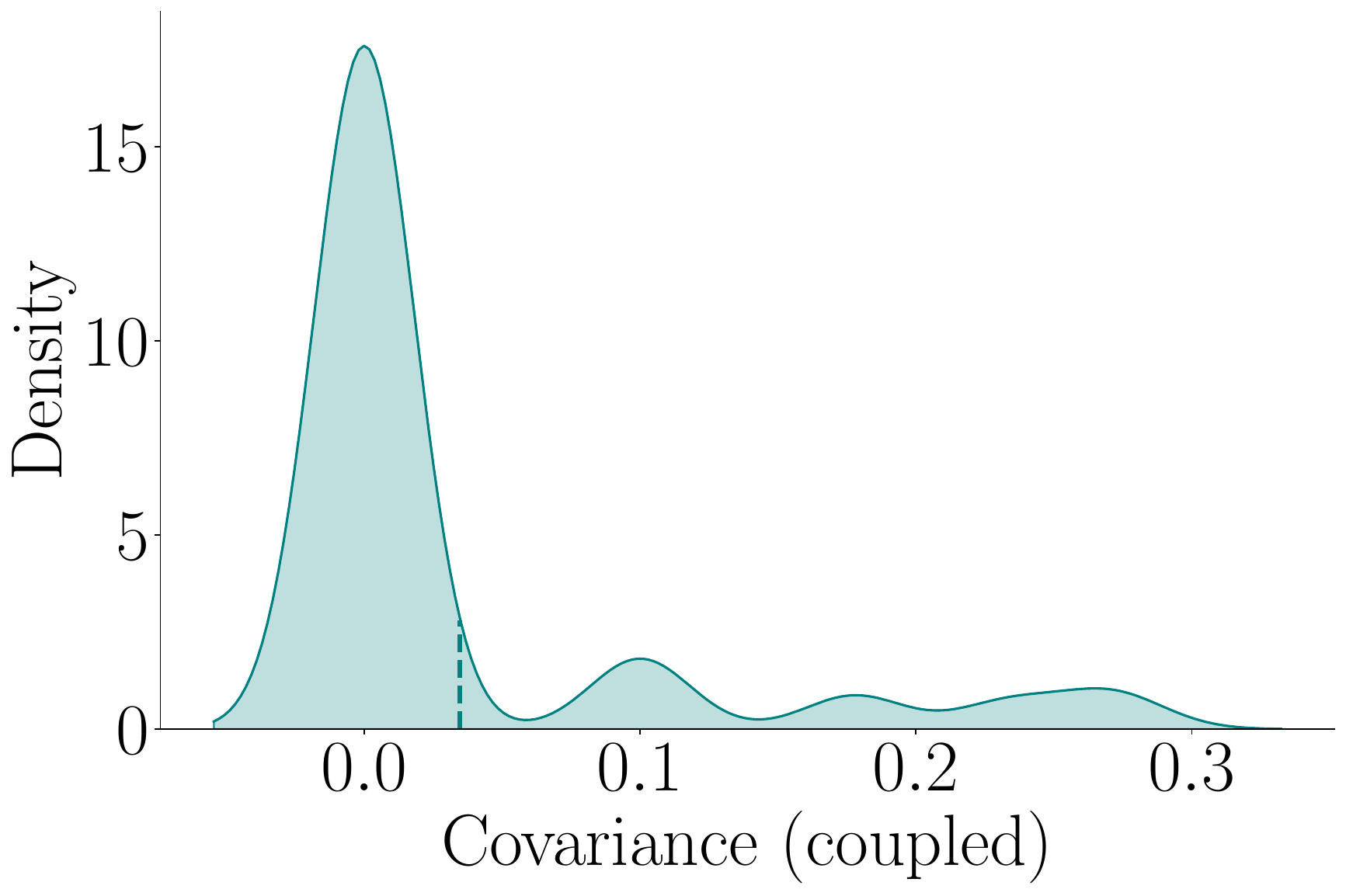} &
    \includegraphics[width=0.23\linewidth]{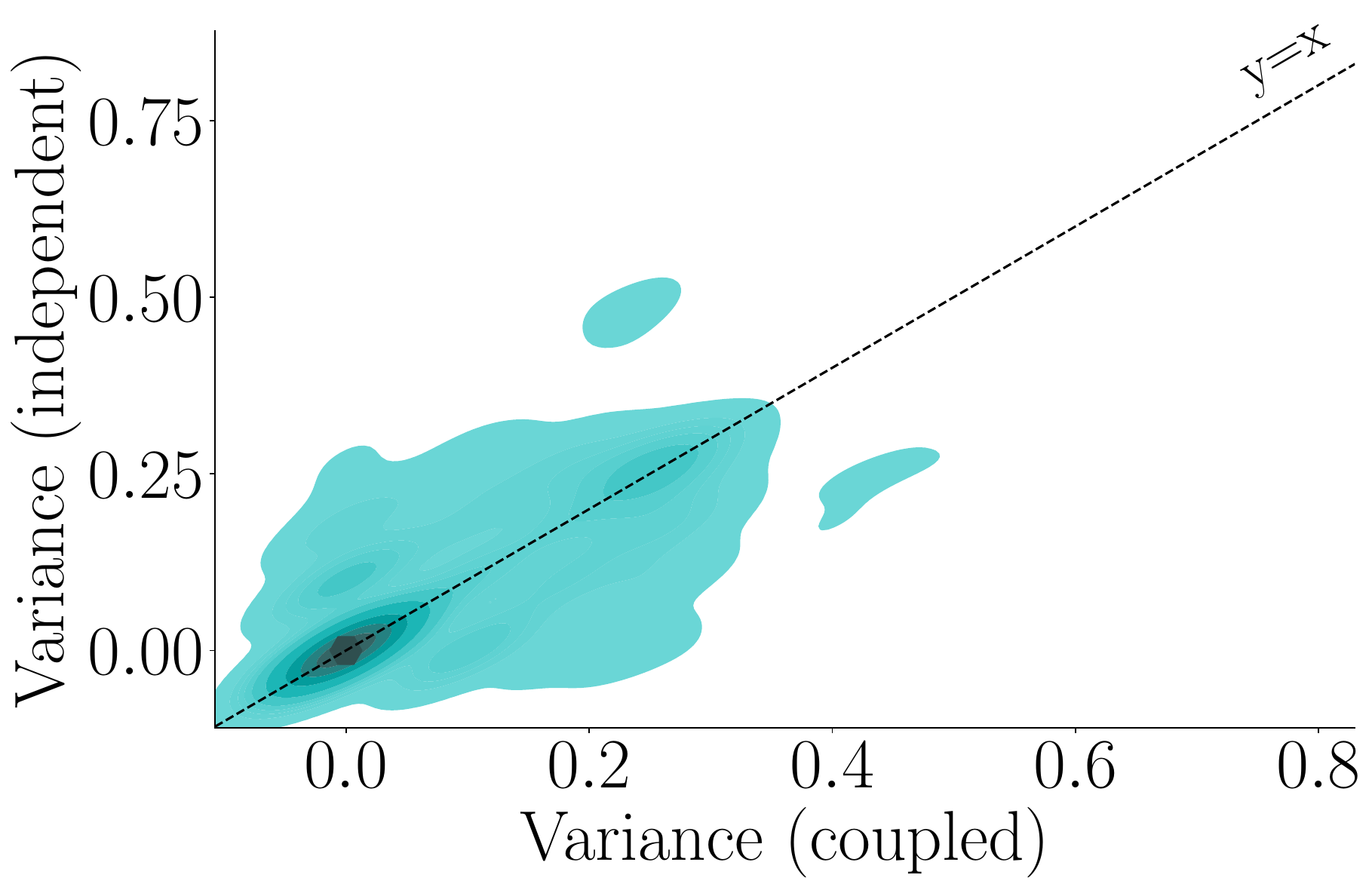} &
    \includegraphics[width=0.23\linewidth]{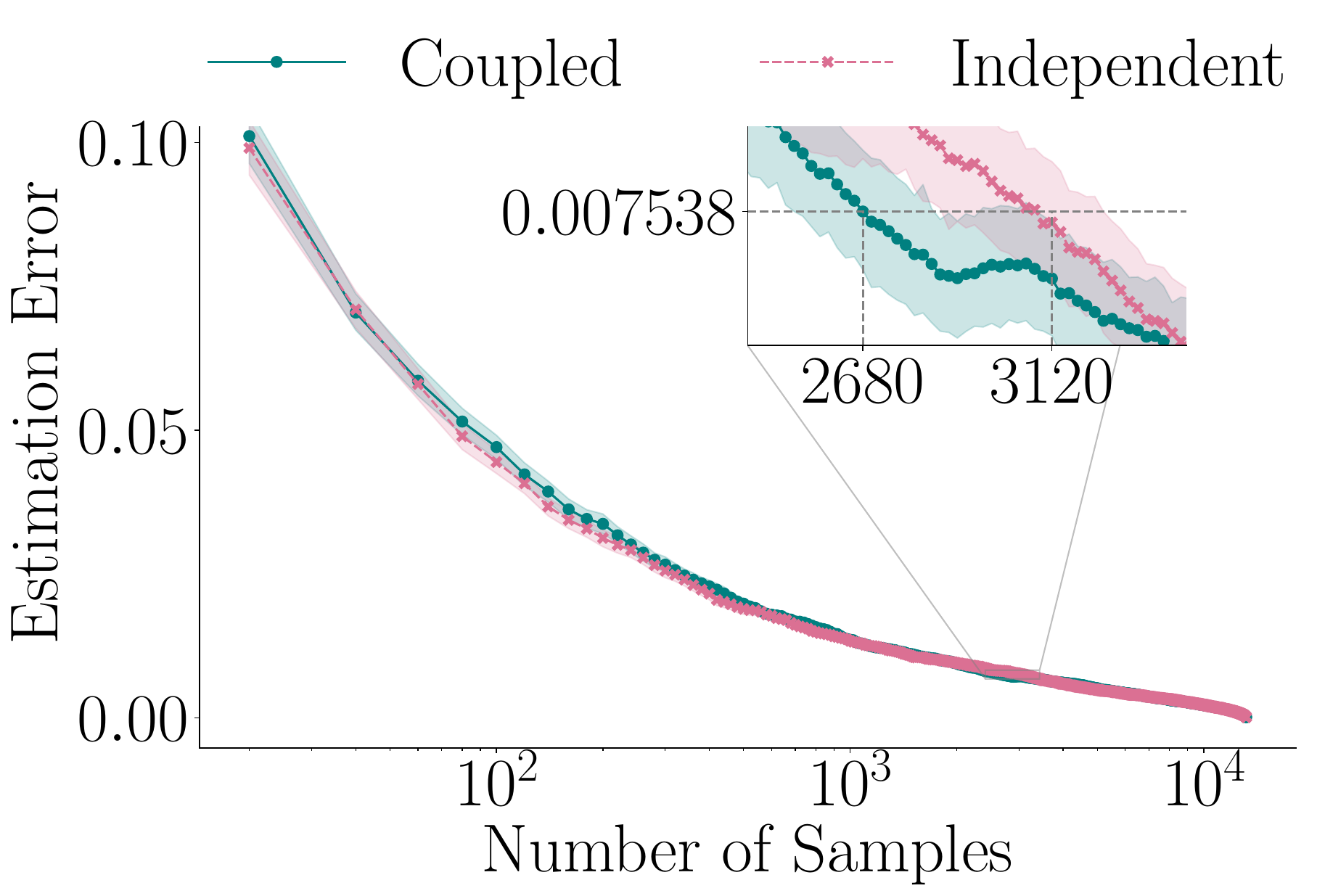} \\ \\
 
 \multicolumn{3}{c}{\texttt{3-8B} vs. \texttt{2.5-7B-bnb-8bit}}\\
    \includegraphics[width=0.23\linewidth]{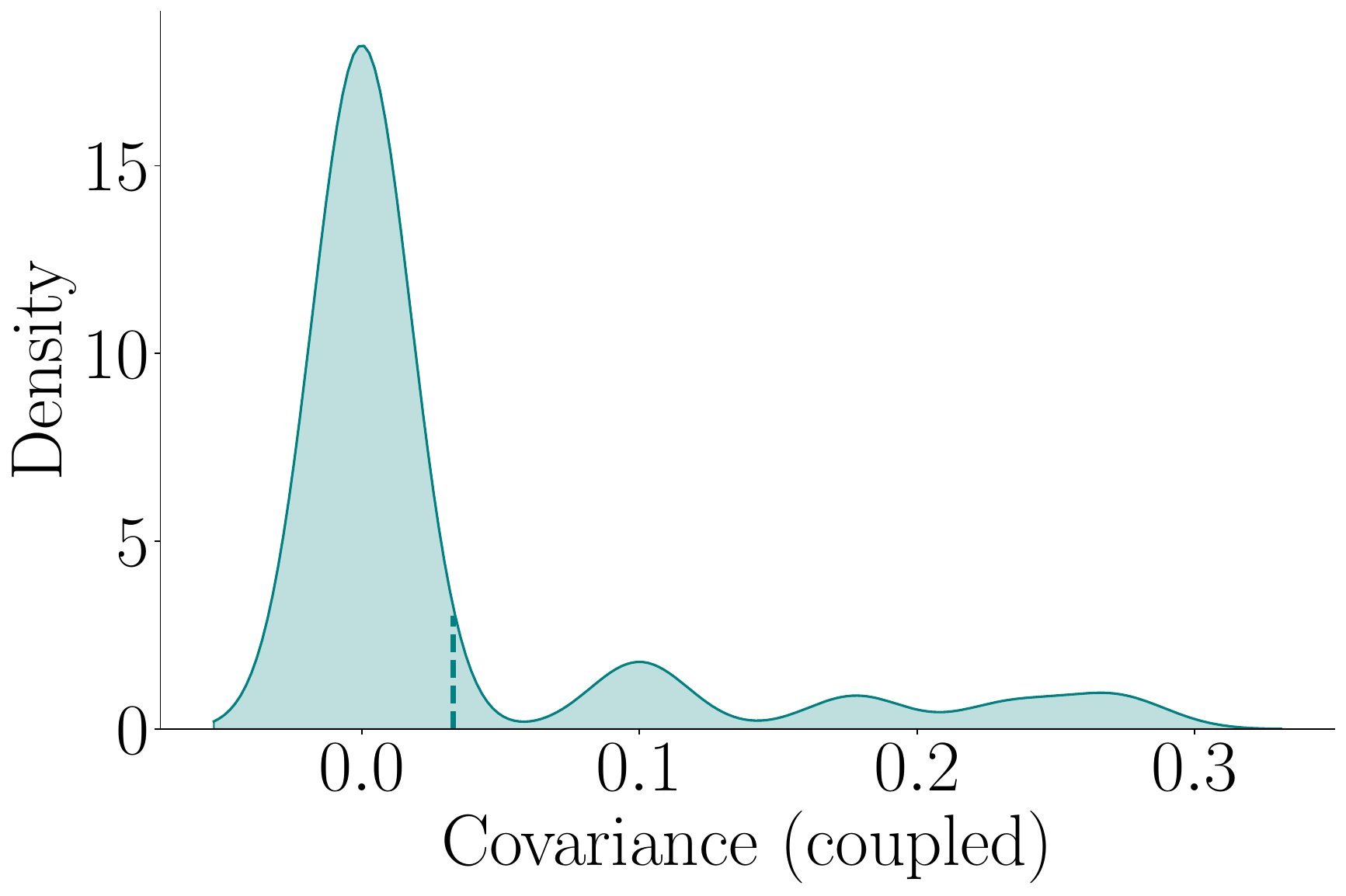} &
    \includegraphics[width=0.23\linewidth]{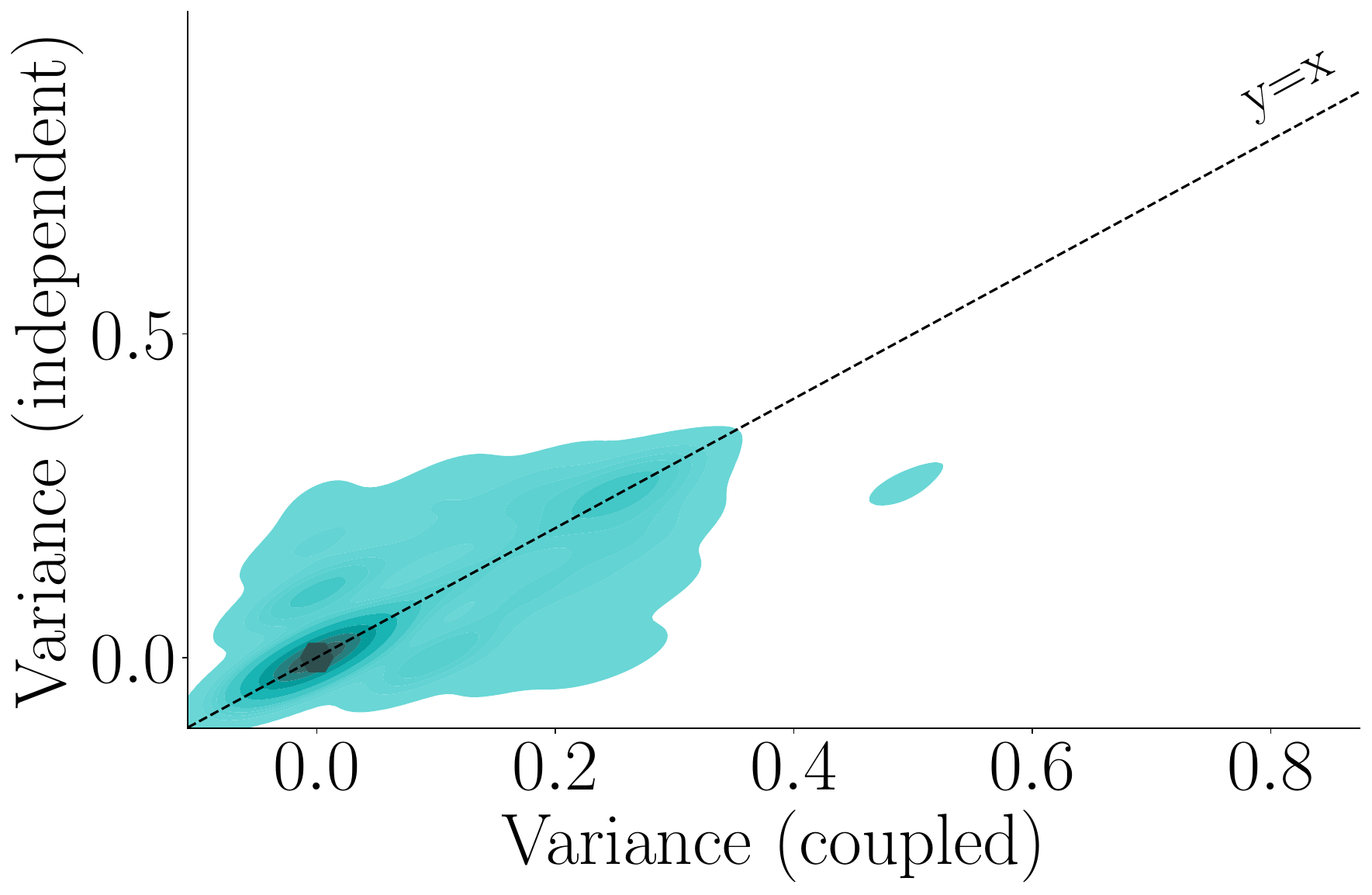} &
    \includegraphics[width=0.23\linewidth]{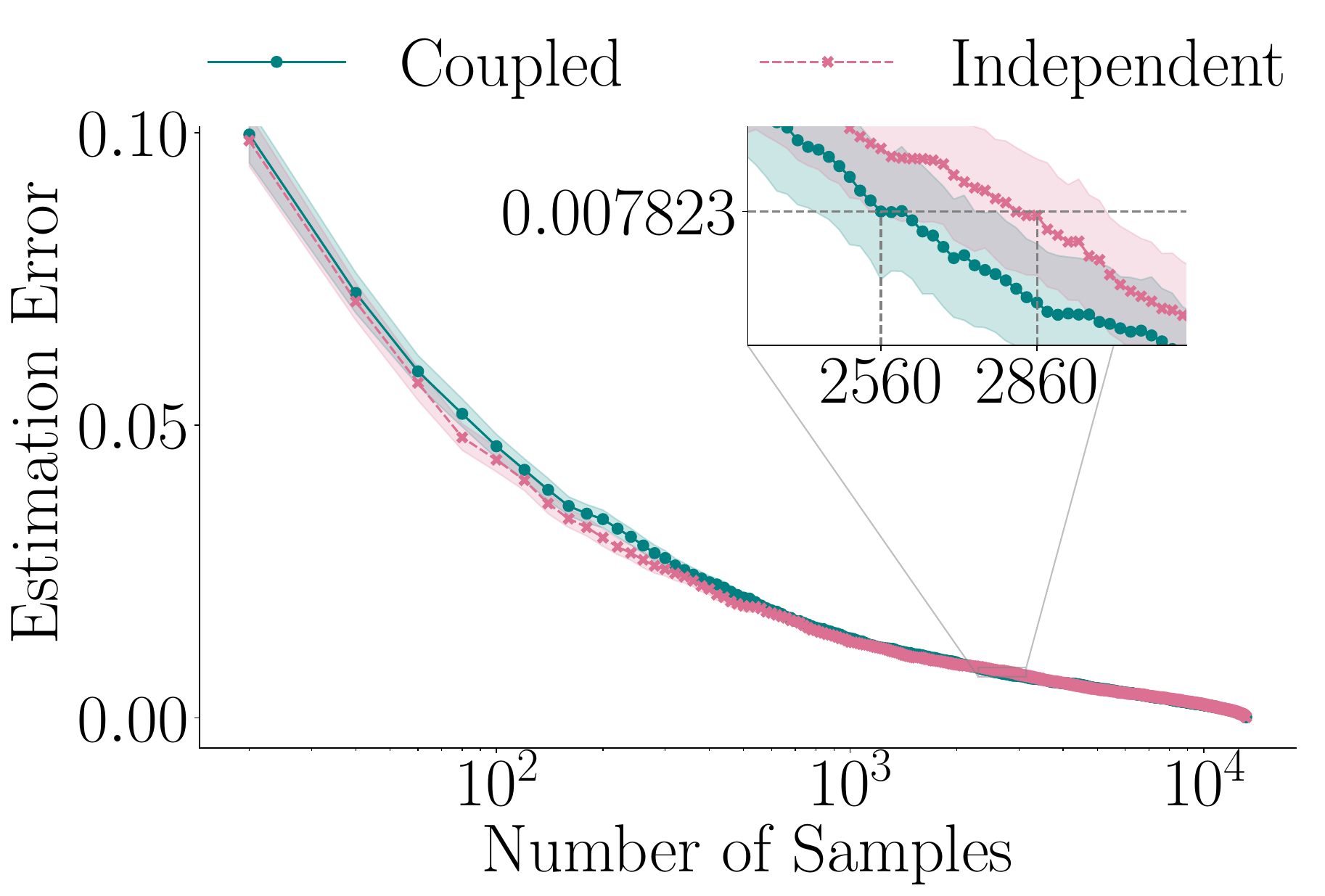} \\ \\
    
    (a) Score covariance & (b) Variance of the score difference & (c) Estimation error vs. \# samples \\ 
    
\end{tabular}
    \caption{\textbf{Comparison between several pairs of LLMs in the \texttt{Qwen} family on math questions from the GSM8K dataset.}
    Panels in column (a) show the kernel density estimate (KDE) of the covariance between the scores of the two LLMs on each problem under coupled generation; the dashed lines correspond to average values. Panels in column (b) show the KDE of the variance of the difference between the scores of the LLMs on each question under coupled and independent generation; the highlighted points correspond to median values. Panels in column (c) show the absolute error in the estimation of the expected difference between the scores of the LLMs against the number of samples; for each point on the x-axis, we perform $1{,}000$ sub-samplings and shaded areas correspond to $95\%$ confidence intervals.}
    \label{fig:gsm8k-qwen-second-5}
\end{figure}

\begin{figure}[h]
\centering
\begin{tabular}{c c c}
     \multicolumn{3}{c}{\texttt{2.5-7B-AWQ-INT4} vs. \texttt{2.5-7B-bnb-8bit}}\\
    \includegraphics[width=0.23\linewidth]{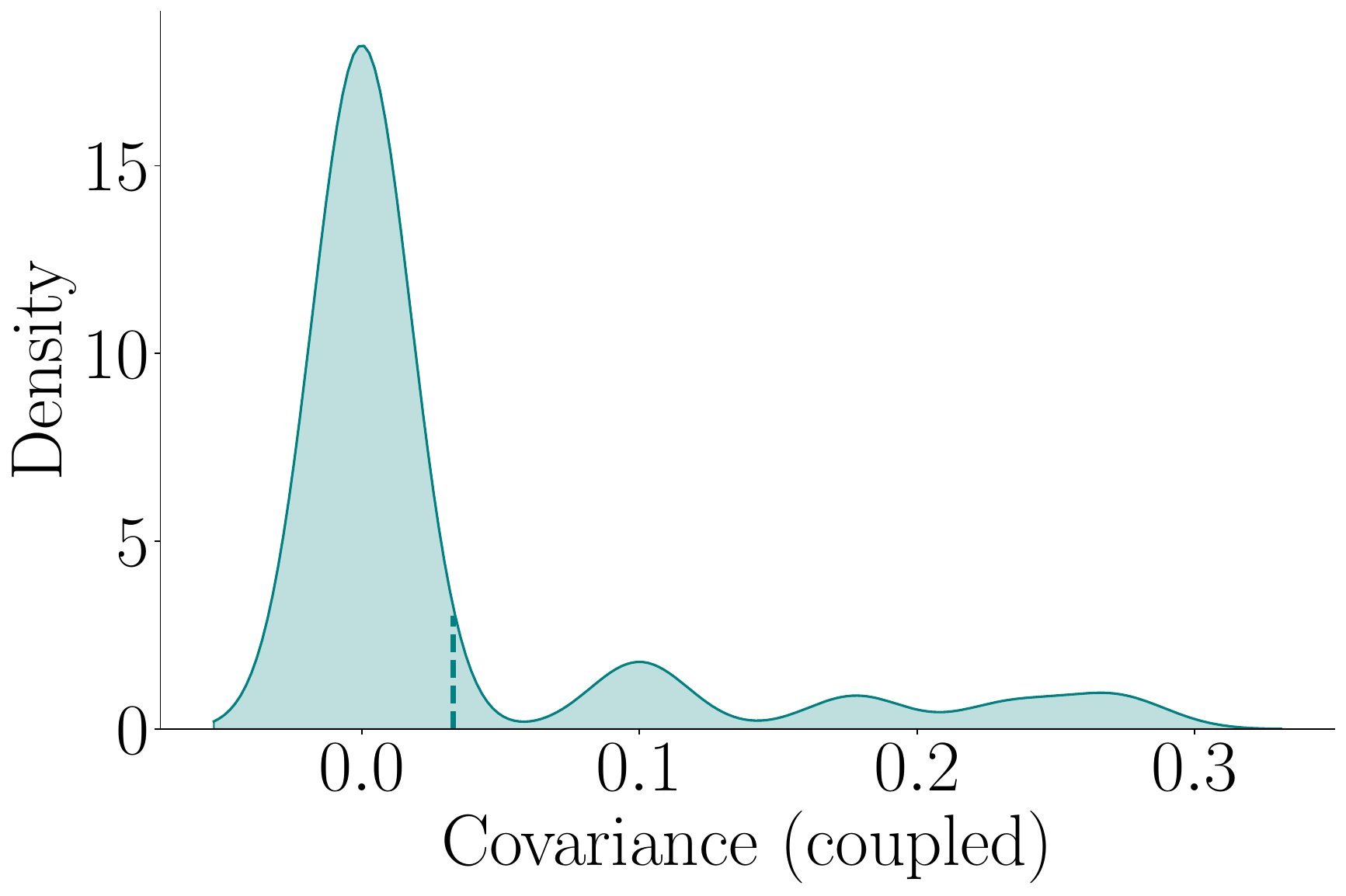} &
    \includegraphics[width=0.23\linewidth]{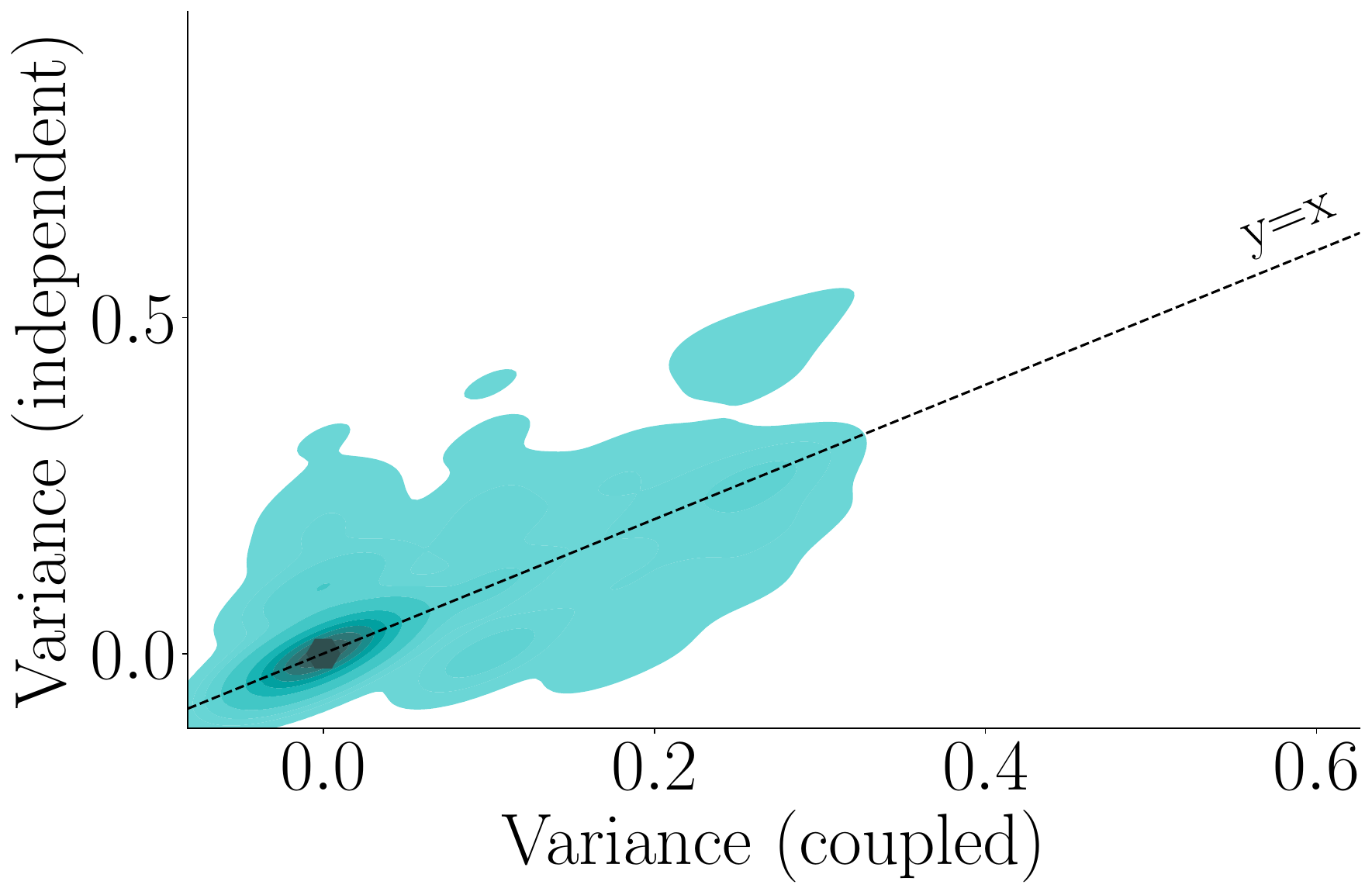} &
    \includegraphics[width=0.23\linewidth]{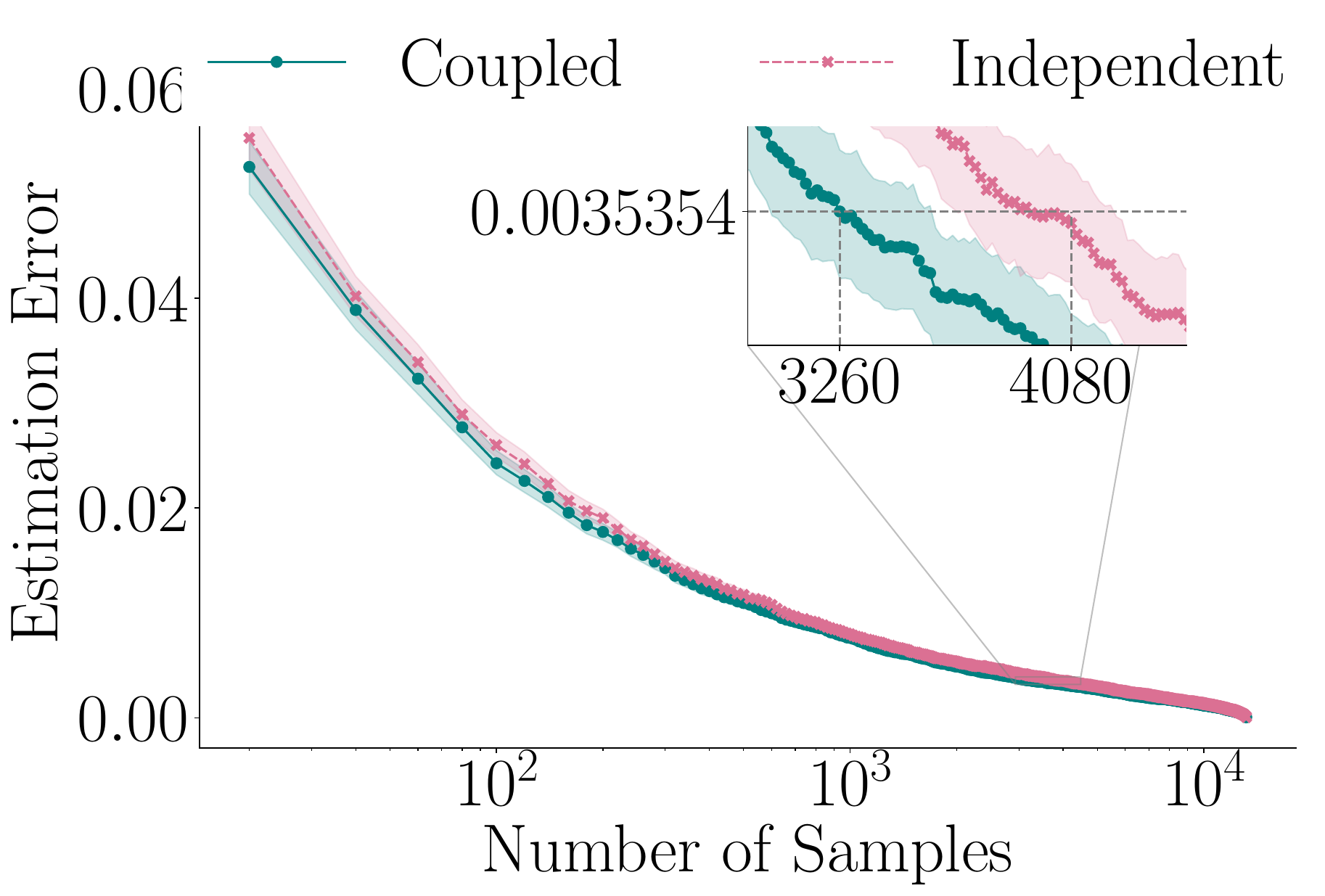} \\ \\
    \multicolumn{3}{c}{\texttt{2.5-7B} vs. \texttt{2.5-7B-AWQ-INT4}}\\
    \includegraphics[width=0.23\linewidth]{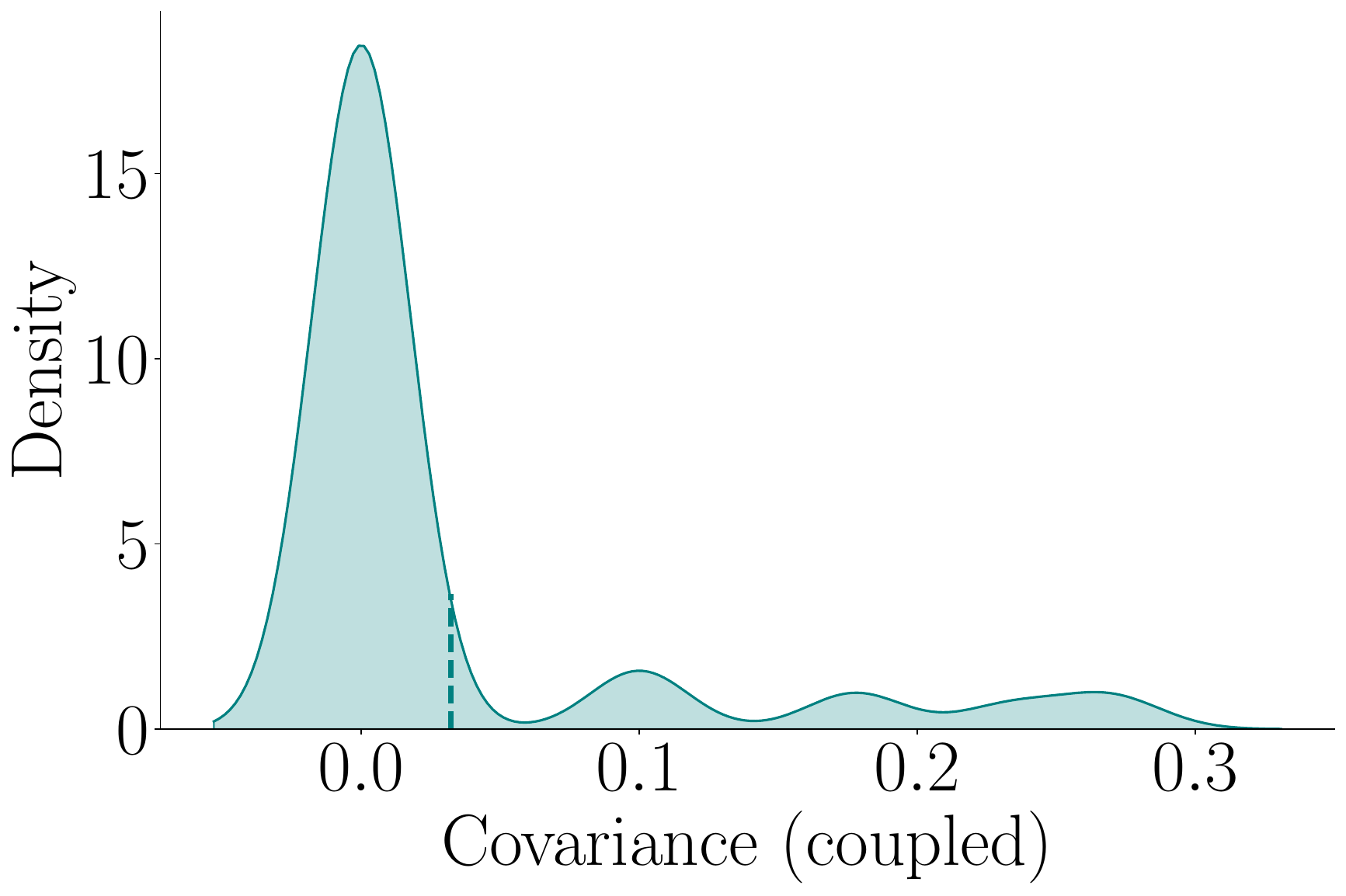} &
    \includegraphics[width=0.23\linewidth]{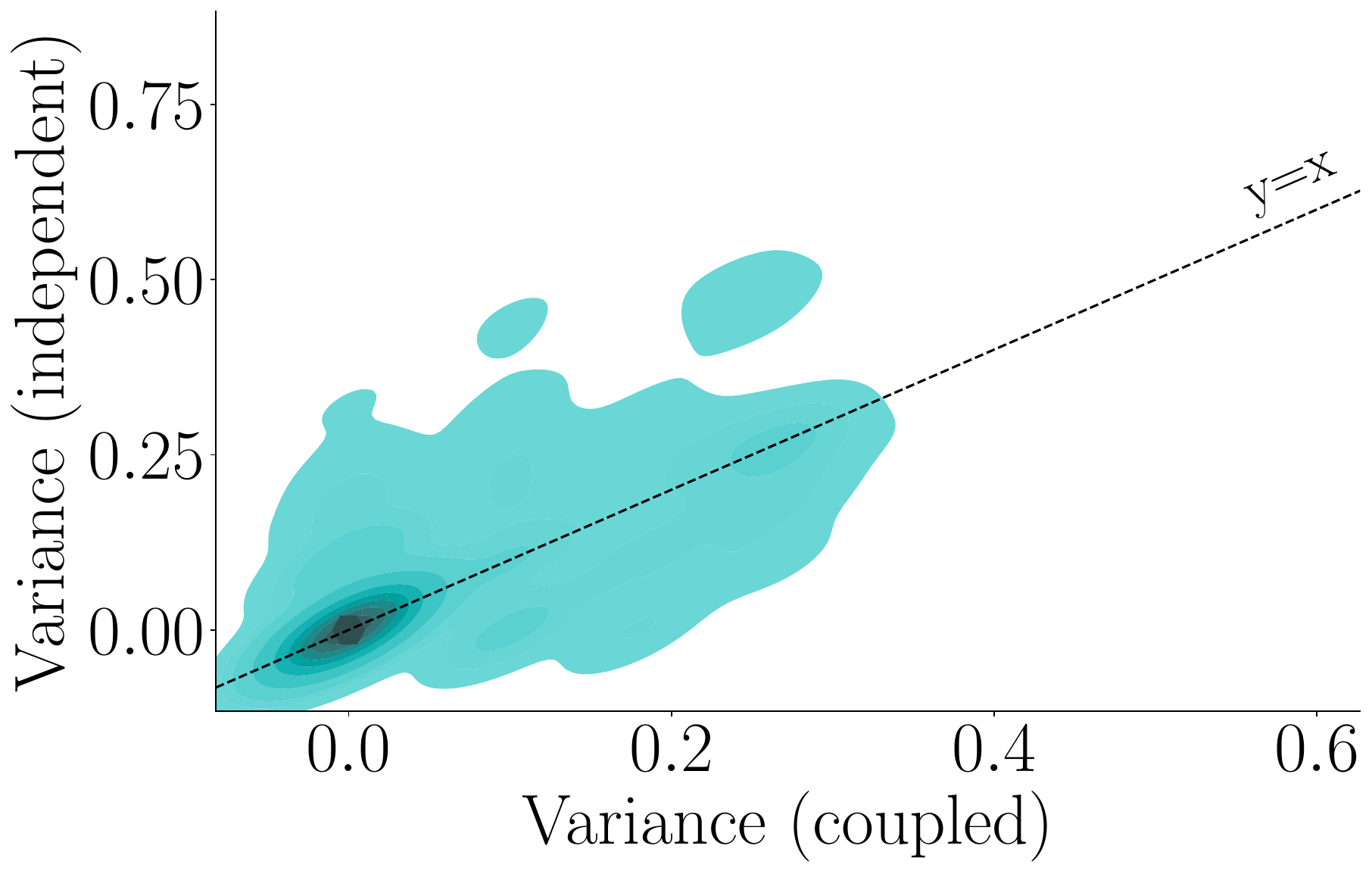} &
    \includegraphics[width=0.23\linewidth]{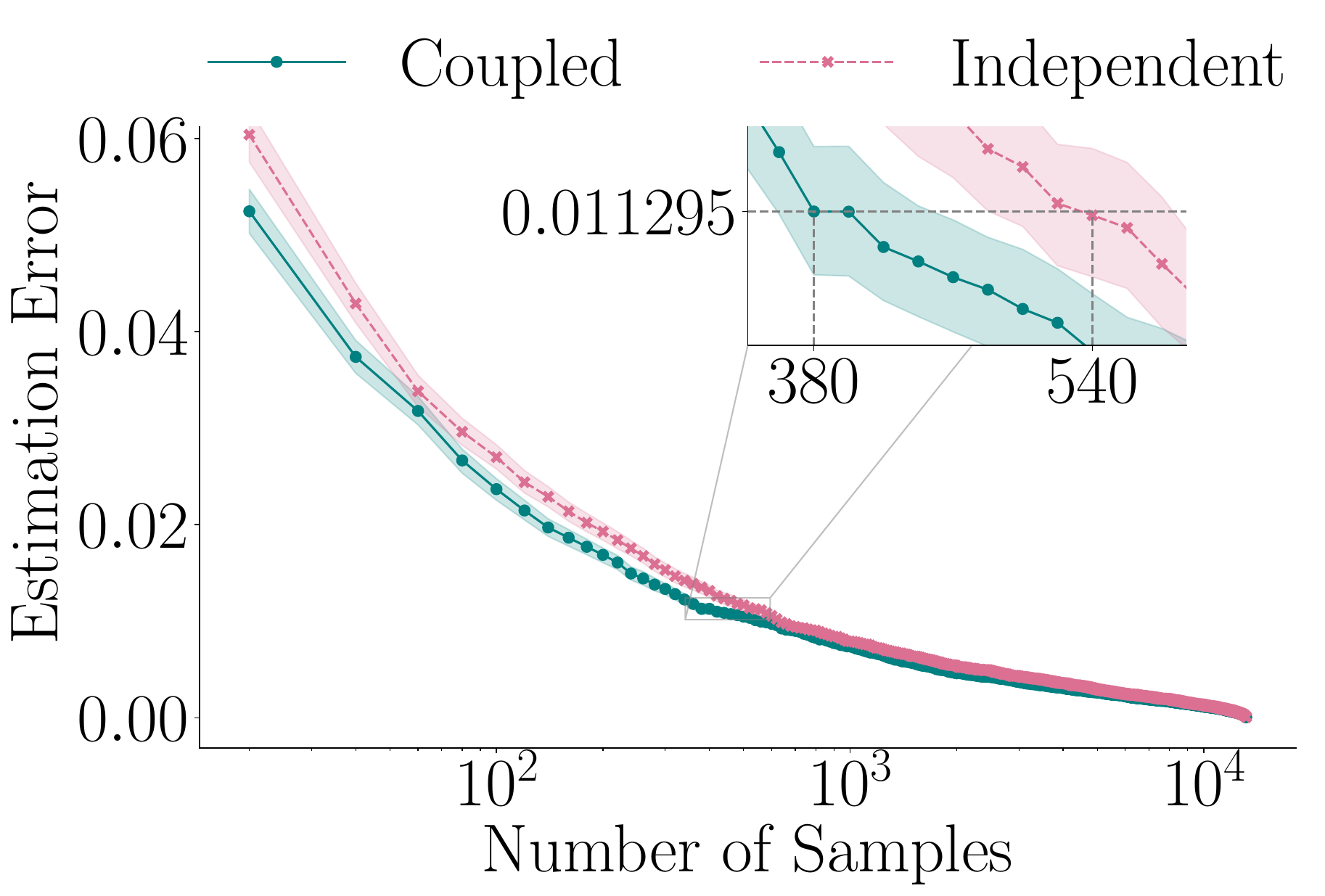} \\ \\
     \multicolumn{3}{c}{\texttt{2.5-7B-bnb-4bit} vs. \texttt{2.5-3B-distil}}\\
    \includegraphics[width=0.23\linewidth]{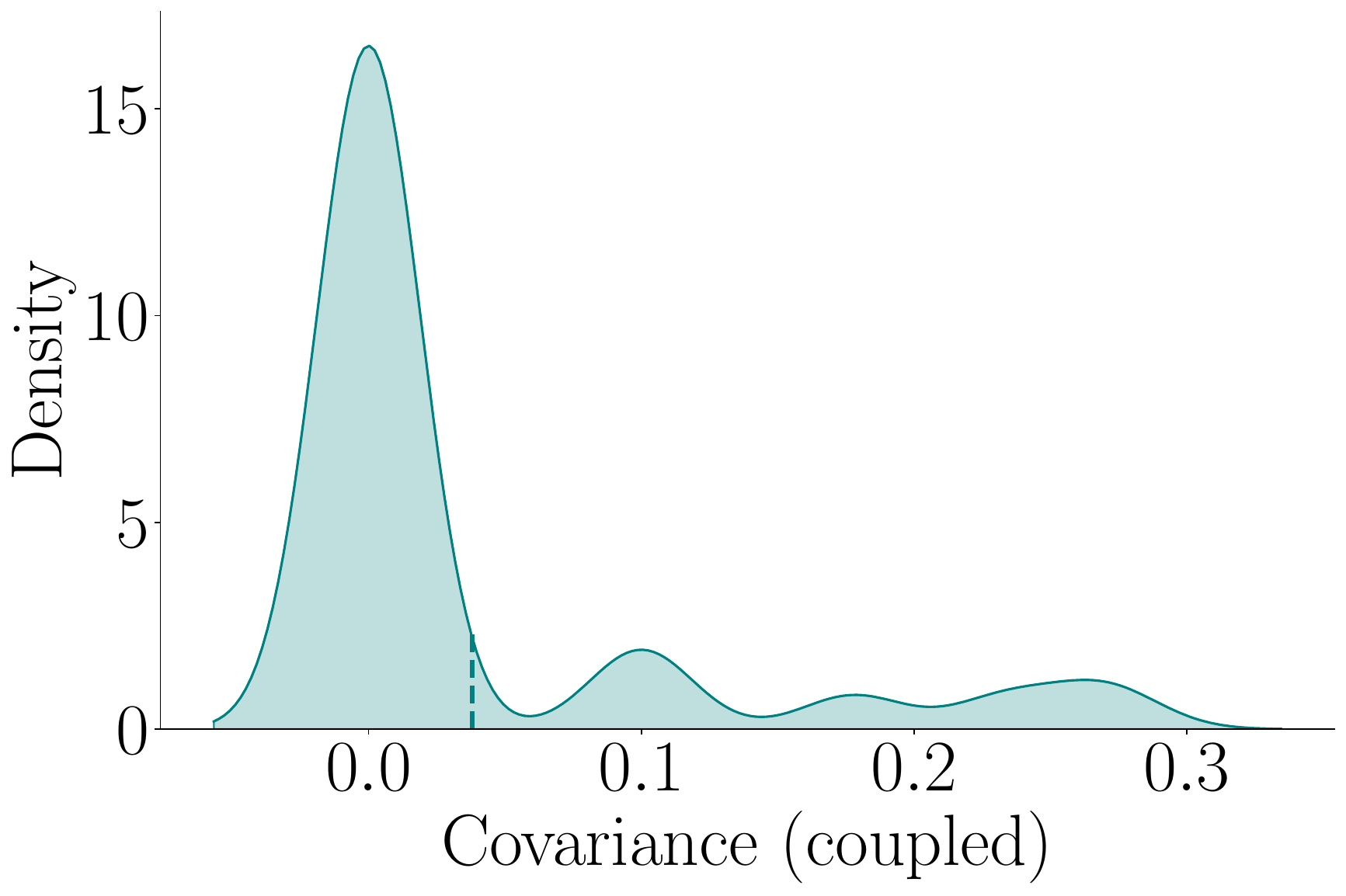} &
    \includegraphics[width=0.23\linewidth]{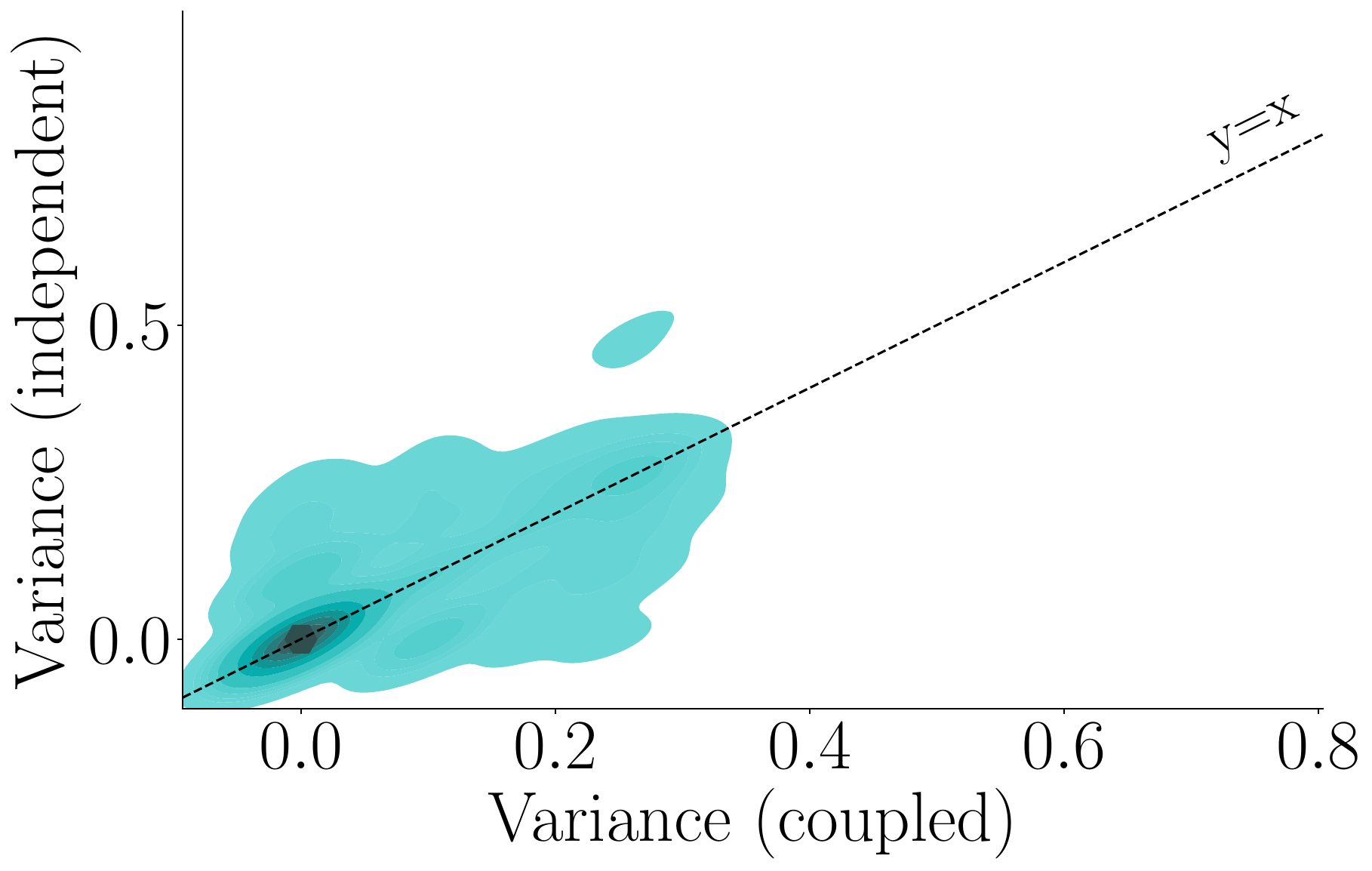} &
    \includegraphics[width=0.23\linewidth]{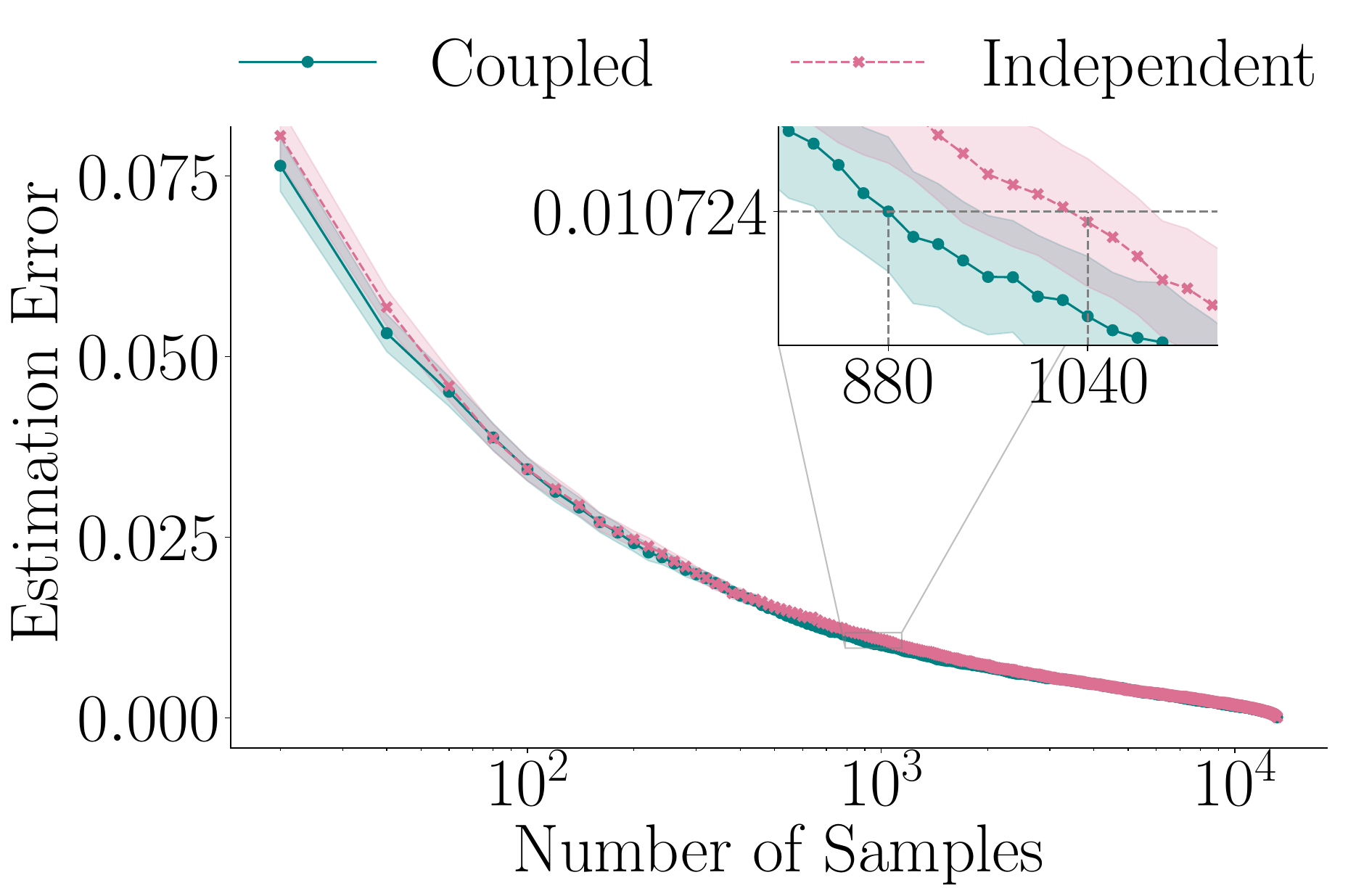} \\ \\
    \multicolumn{3}{c}{\texttt{2.5-7B-bnb-4bit} vs. \texttt{2.5-7B-AWQ-INT4}}\\
    \includegraphics[width=0.23\linewidth]{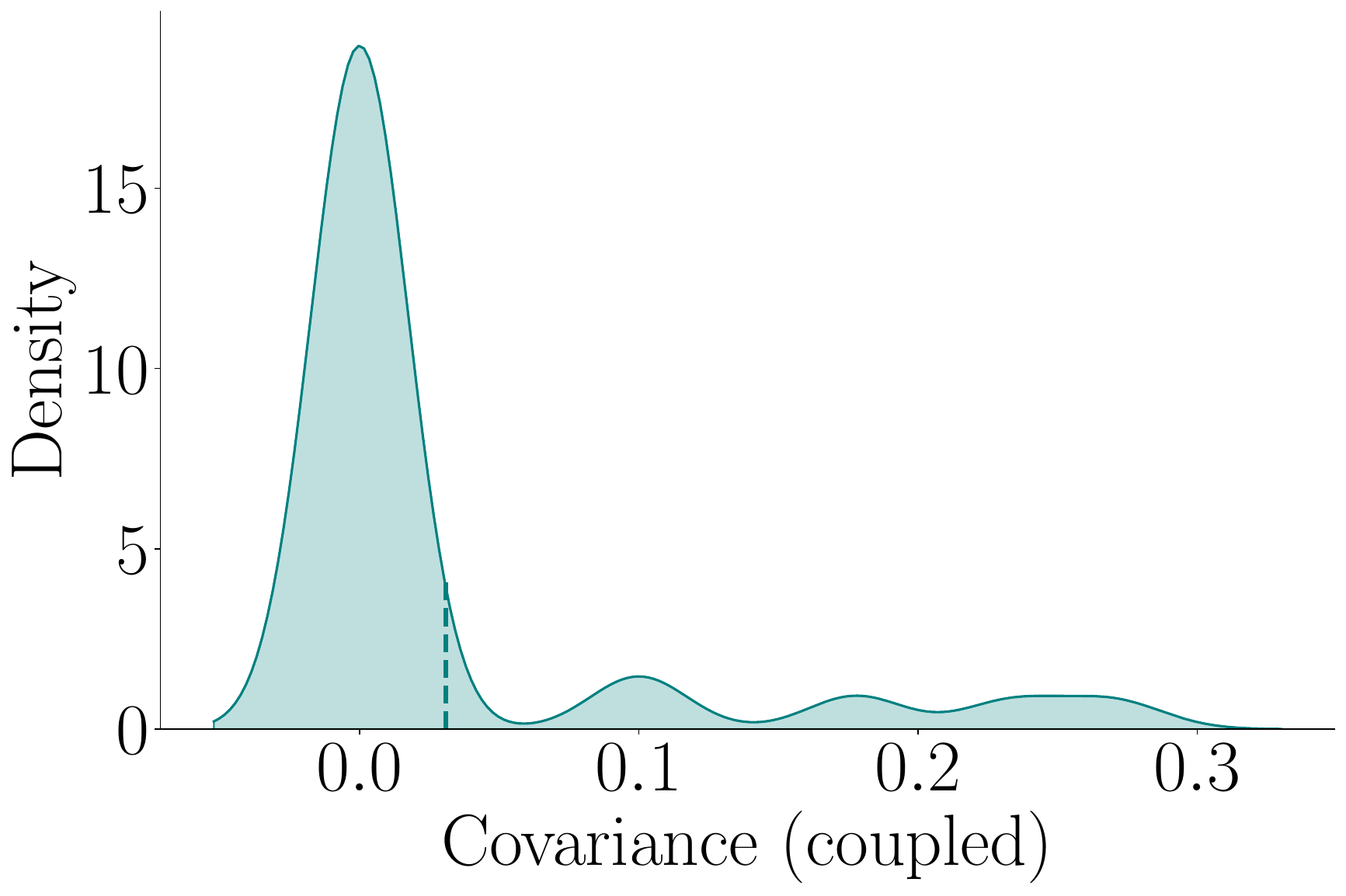} &
    \includegraphics[width=0.23\linewidth]{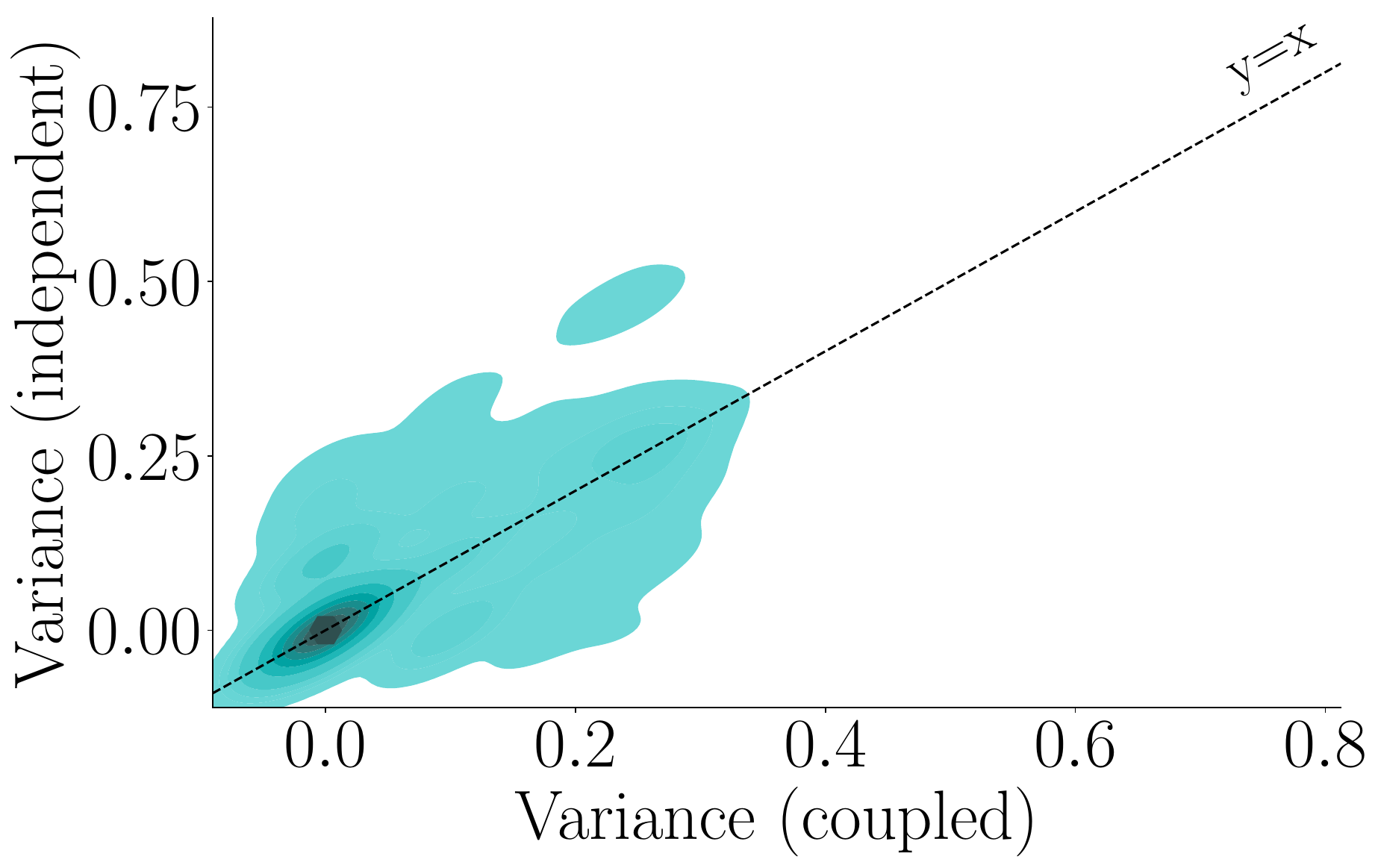} &
    \includegraphics[width=0.23\linewidth]{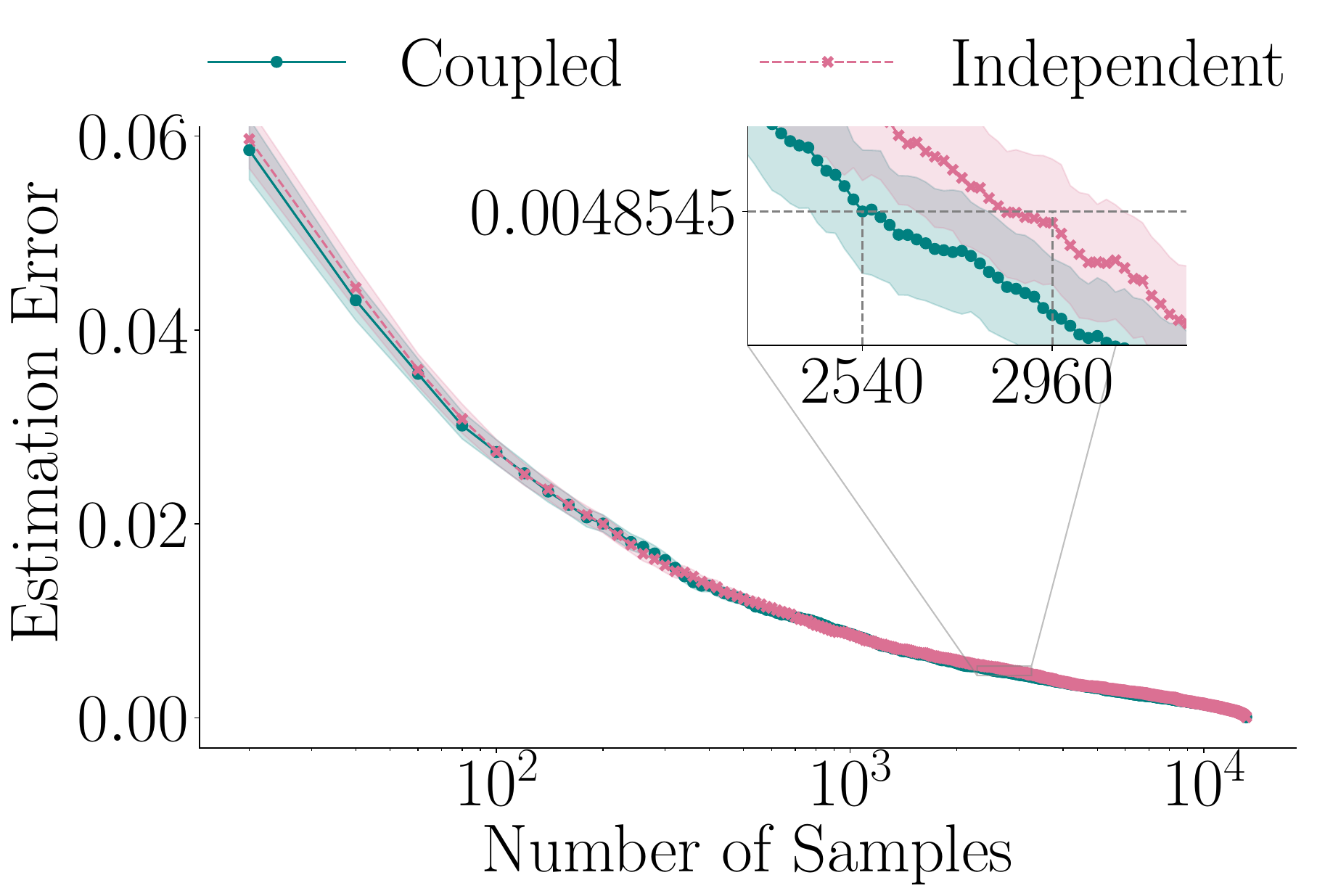} \\ \\  

        \multicolumn{3}{c}{\texttt{2.5-7B-bnb-4bit} vs. \texttt{2.5-7B-bnb-8bit}}\\
    \includegraphics[width=0.23\linewidth]{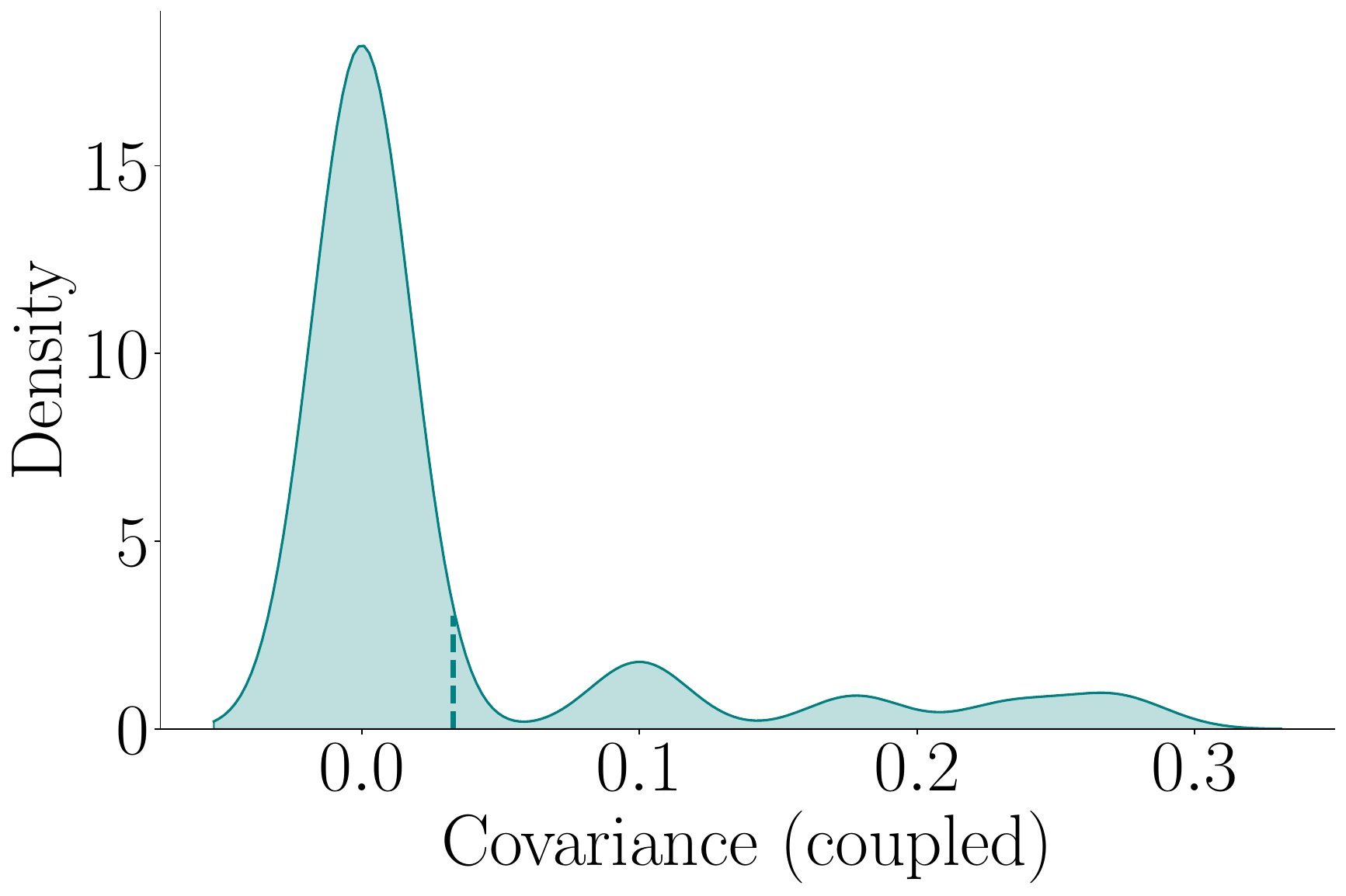} &
    \includegraphics[width=0.23\linewidth]{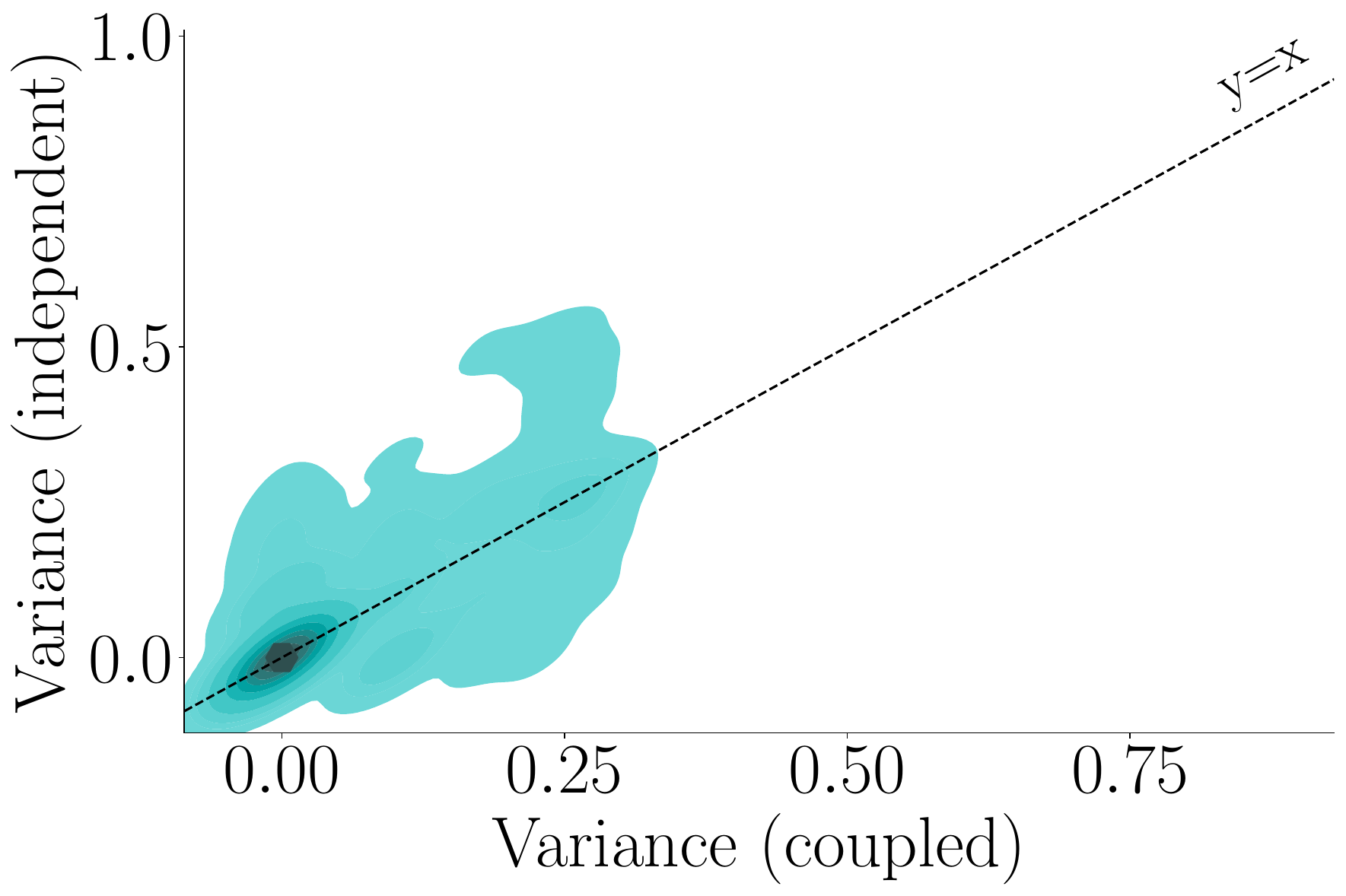} &
    \includegraphics[width=0.23\linewidth]{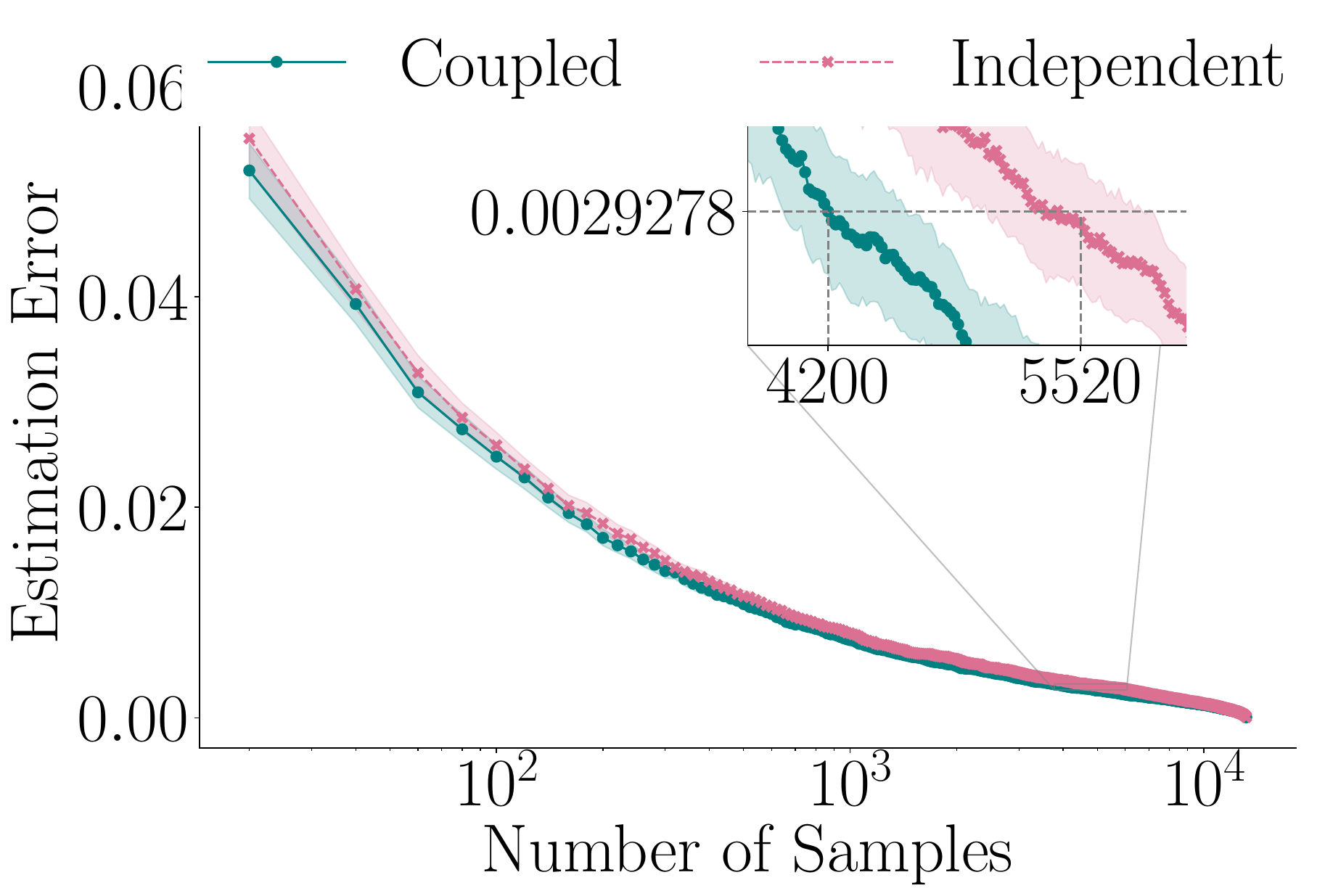} \\ \\ 
    
    (a) Score covariance & (b) Variance of the score difference & (c) Estimation error vs. \# samples \\ 
    
\end{tabular}
    \caption{\textbf{Comparison between several pairs of LLMs in the \texttt{Qwen} family on math questions from the GSM8K dataset.}
    Panels in column (a) show the kernel density estimate (KDE) of the covariance between the scores of the two LLMs on each problem under coupled generation; the dashed lines correspond to average values. Panels in column (b) show the KDE of the variance of the difference between the scores of the LLMs on each question under coupled and independent generation; the highlighted points correspond to median values. Panels in column (c) show the absolute error in the estimation of the expected difference between the scores of the LLMs against the number of samples; for each point on the x-axis, we perform $1{,}000$ sub-samplings and shaded areas correspond to $95\%$ confidence intervals.}
    \label{fig:gsm8k-qwen-third-5}
\end{figure}

\begin{figure}[h]
\centering
\begin{tabular}{c c c}
     \multicolumn{3}{c}{\texttt{2.5-3B-distil} vs. \texttt{2.5-7B-bnb-8bit}}\\
    \includegraphics[width=0.23\linewidth]{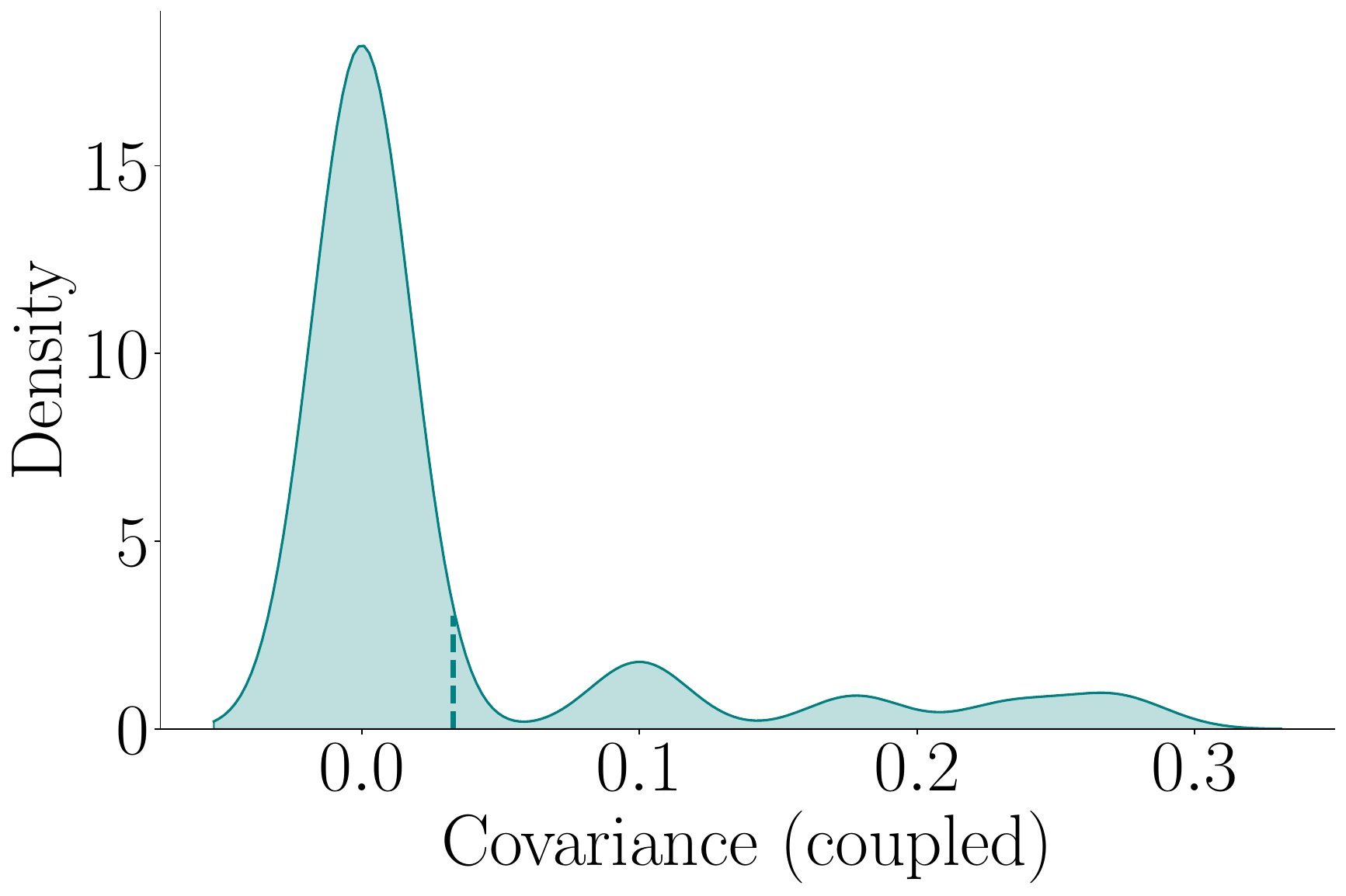} &
    \includegraphics[width=0.23\linewidth]{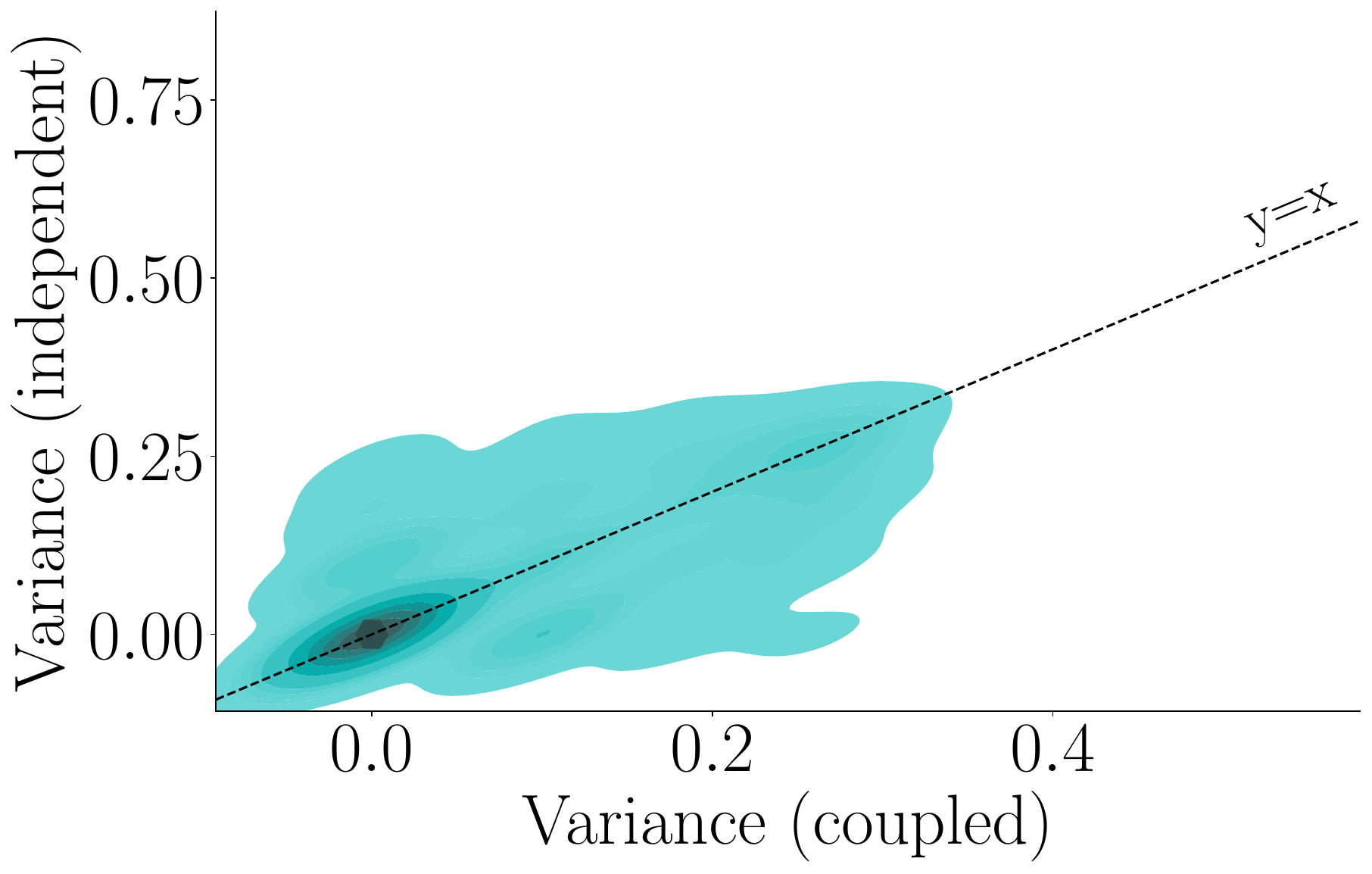} &
    \includegraphics[width=0.23\linewidth]{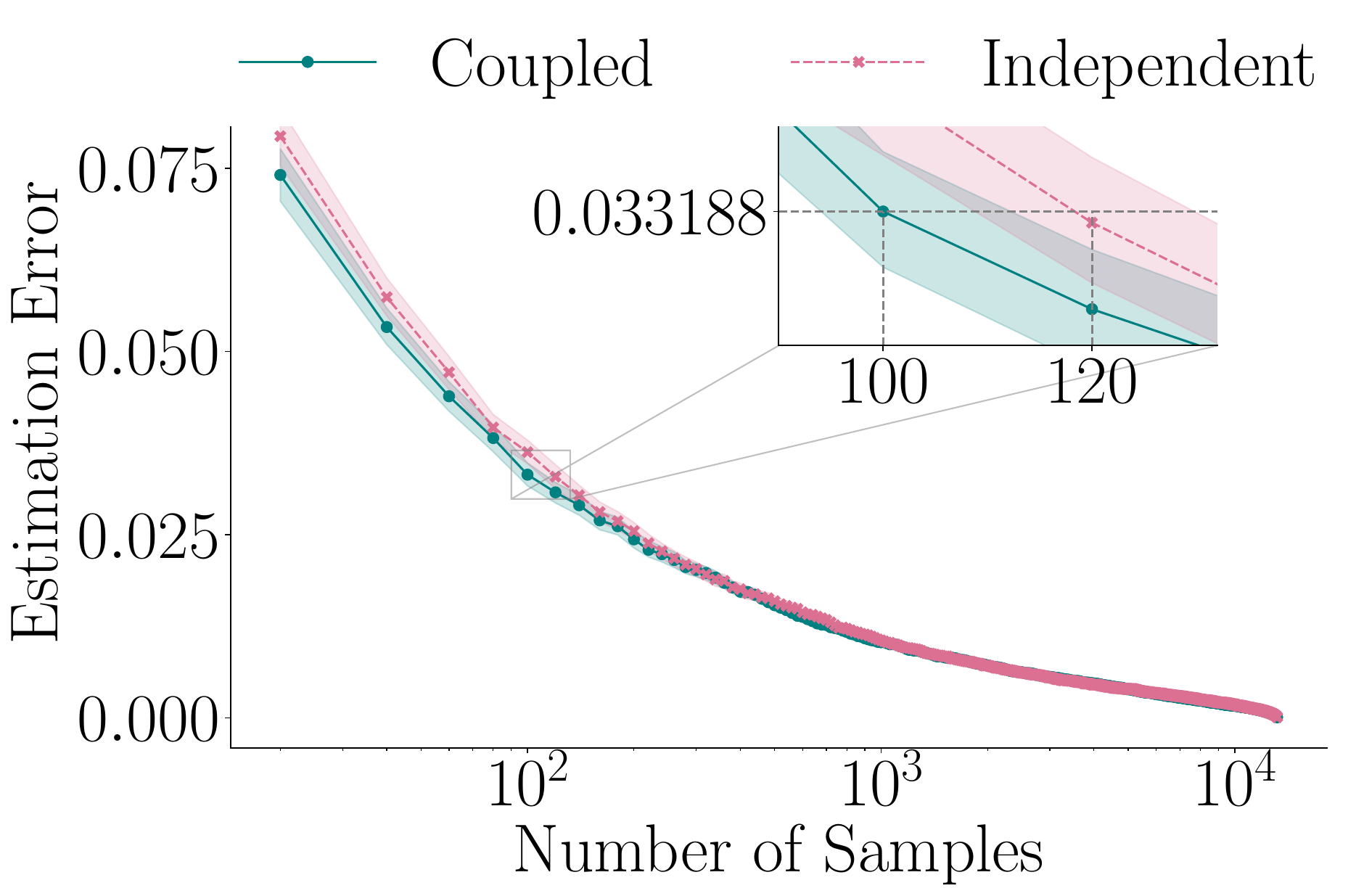} \\ \\
    \multicolumn{3}{c}{\texttt{2.5-3B-distil} vs. \texttt{2.5-7B}}\\
    \includegraphics[width=0.23\linewidth]{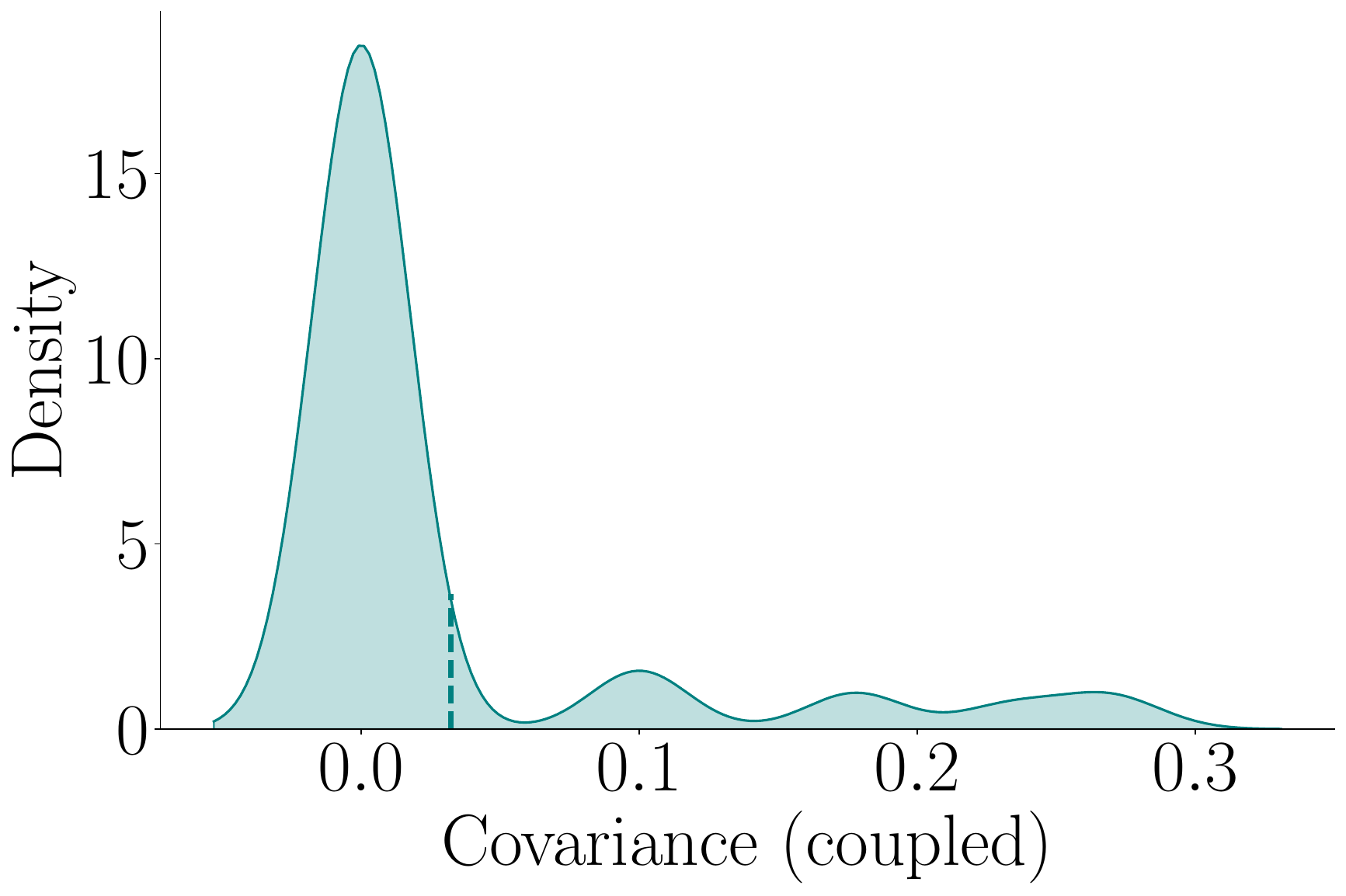} &
    \includegraphics[width=0.23\linewidth]{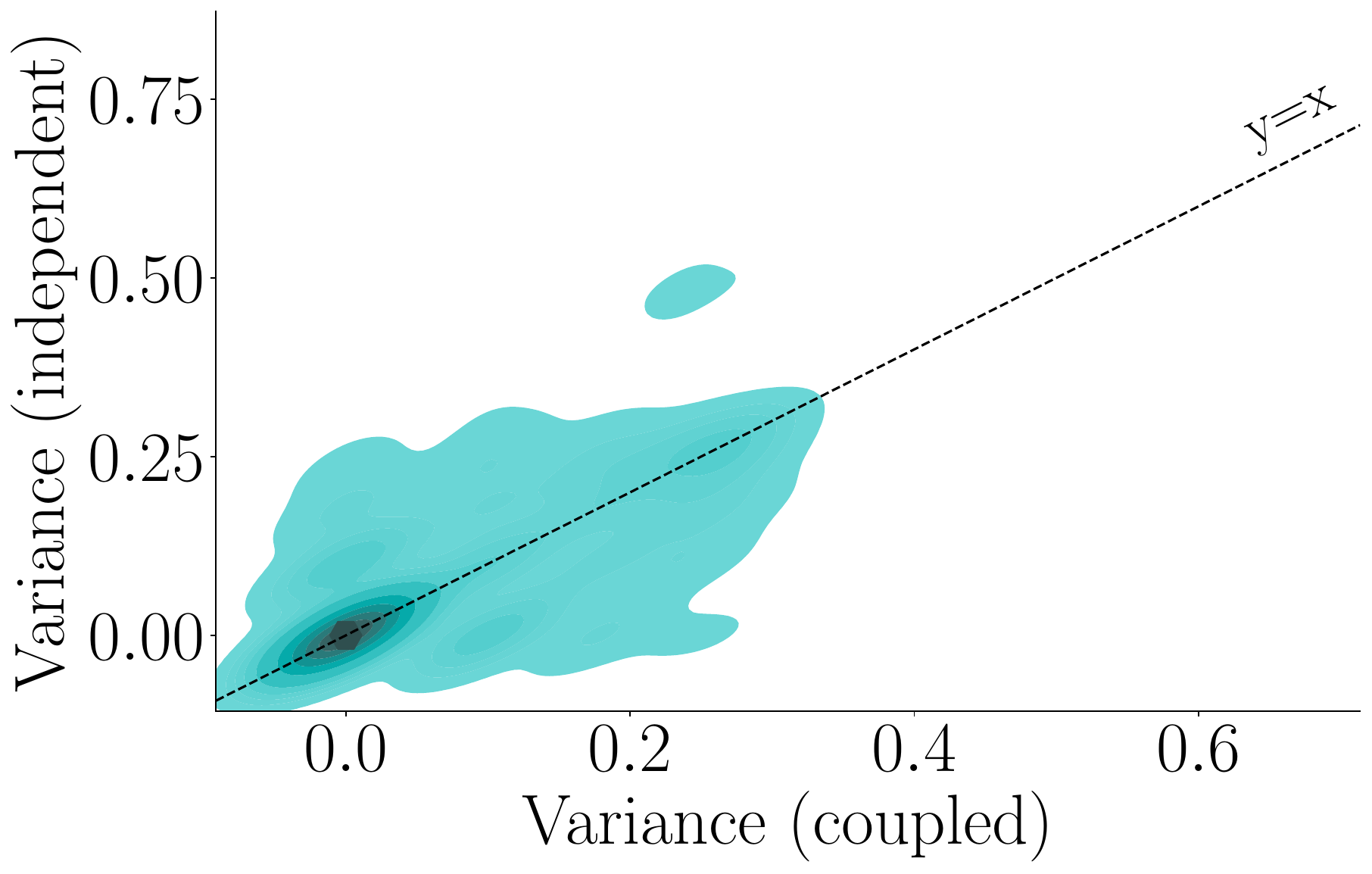} &
    \includegraphics[width=0.23\linewidth]{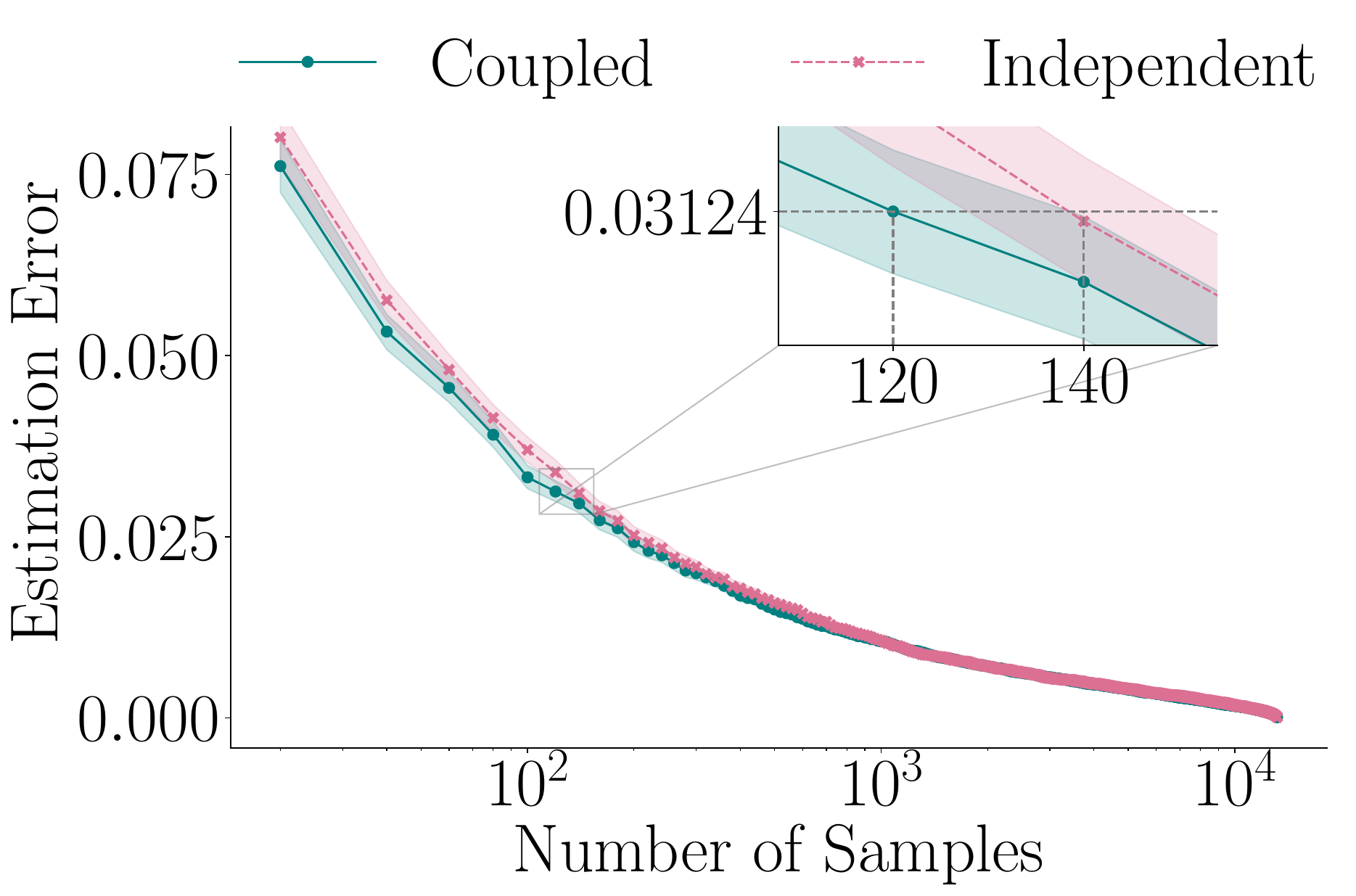} \\ \\

    (a) Score covariance & (b) Variance of the score difference & (c) Estimation error vs. \# samples \\ 
    
\end{tabular}
    \caption{\textbf{Comparison between several pairs of LLMs in the \texttt{Qwen} family on math questions from the GSM8K dataset.}
    Panels in column (a) show the kernel density estimate (KDE) of the covariance between the scores of the two LLMs on each problem under coupled generation; the dashed lines correspond to average values. Panels in column (b) show the KDE of the variance of the difference between the scores of the LLMs on each question under coupled and independent generation; the highlighted points correspond to median values. Panels in column (c) show the absolute error in the estimation of the expected difference between the scores of the LLMs against the number of samples; for each point on the x-axis, we perform $1{,}000$ sub-samplings and shaded areas correspond to $95\%$ confidence intervals.}
    \label{fig:gsm8k-qwen-last-2}
\end{figure}

\begin{figure}[!!h]
\centering
\begin{tabular}{c c c}
     \multicolumn{3}{c}{\texttt{2.5-7B} vs. \texttt{2.5-7B-bnb-8bit}}\\
    \includegraphics[width=0.23\linewidth]{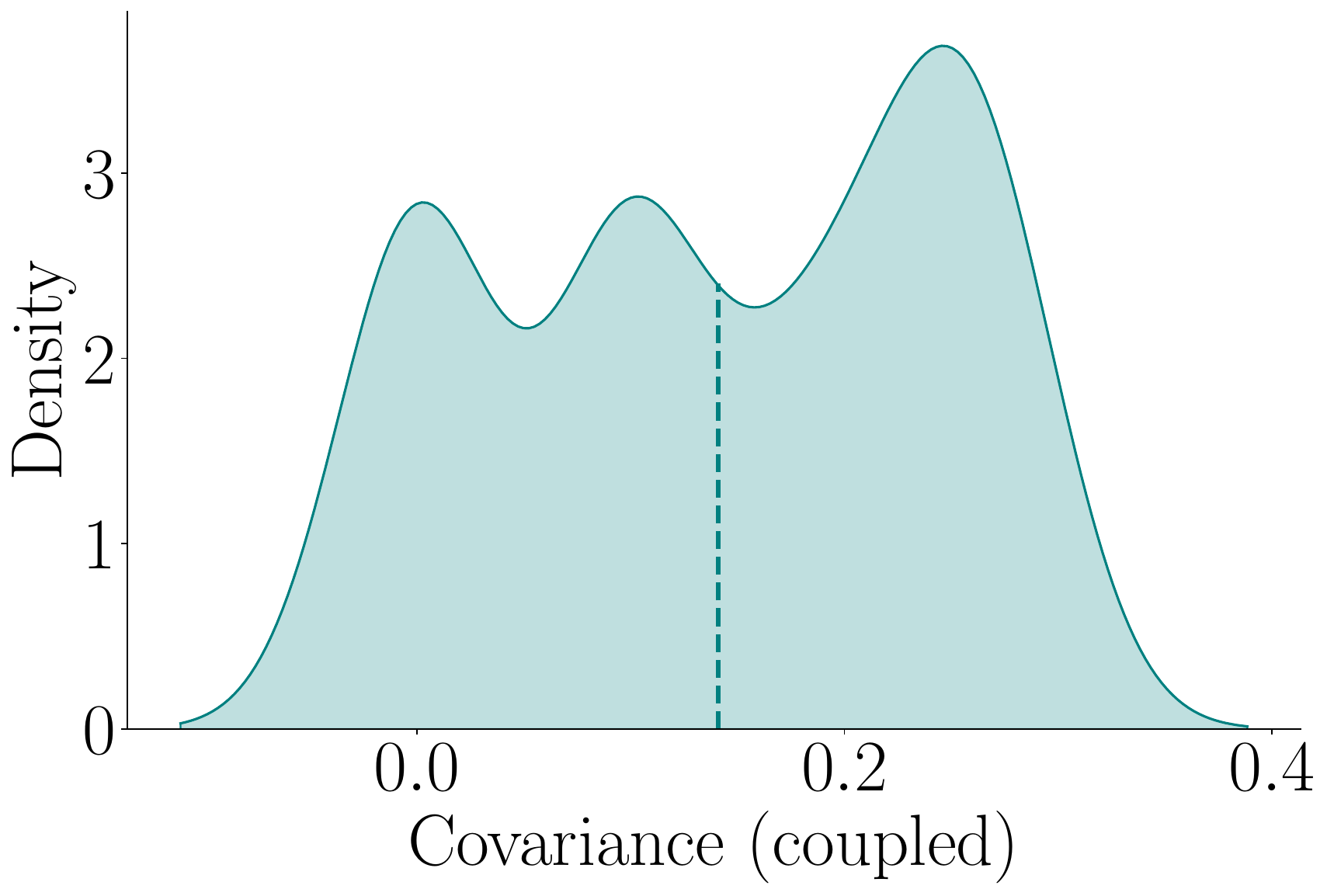} &
    \includegraphics[width=0.23\linewidth]{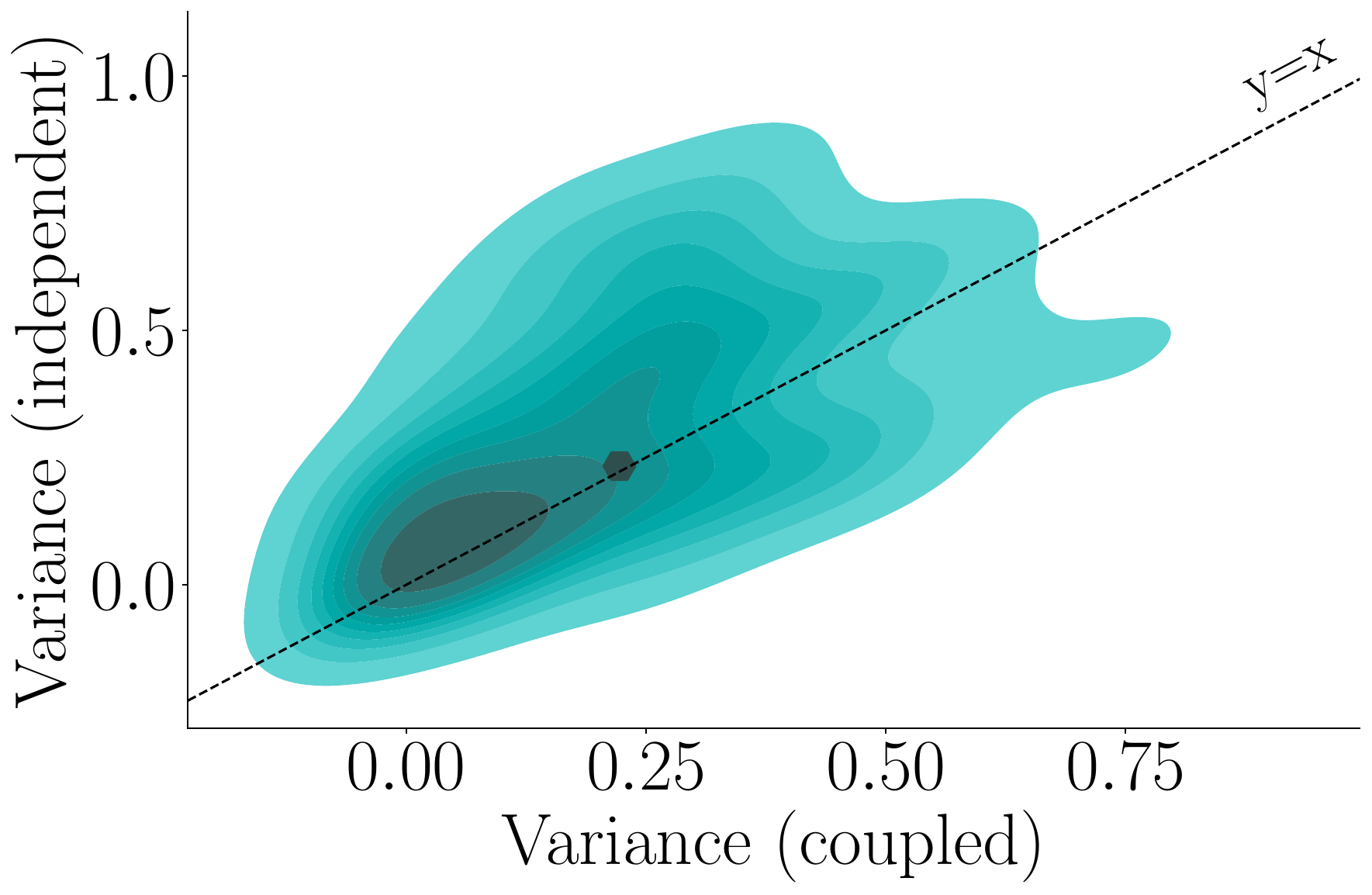} &
    \includegraphics[width=0.23\linewidth]{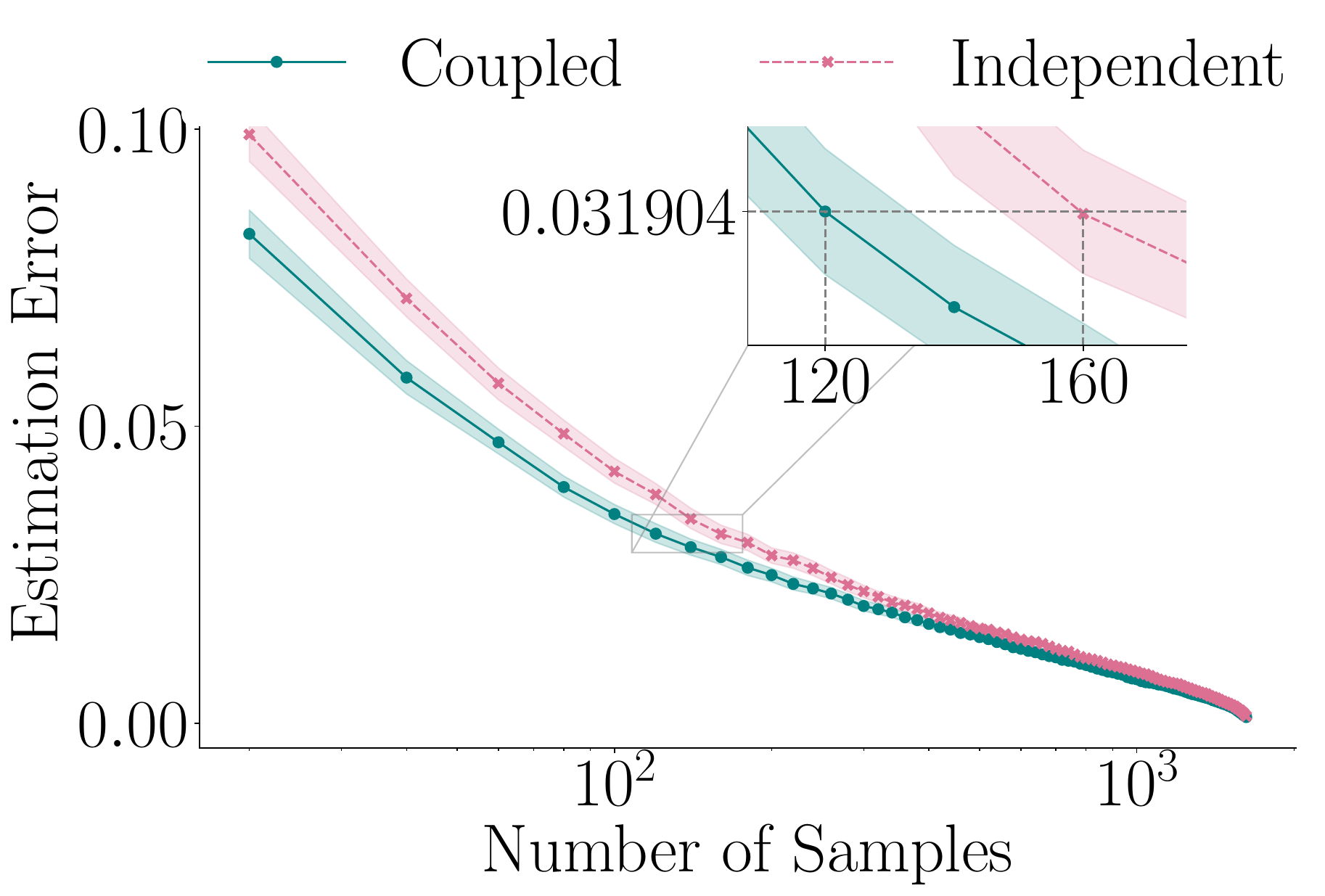} \\ \\
%
    \multicolumn{3}{c}{\texttt{2.5-7B} vs. \texttt{2.5-7B-bnb-4bit}}\\
    \includegraphics[width=0.23\linewidth]{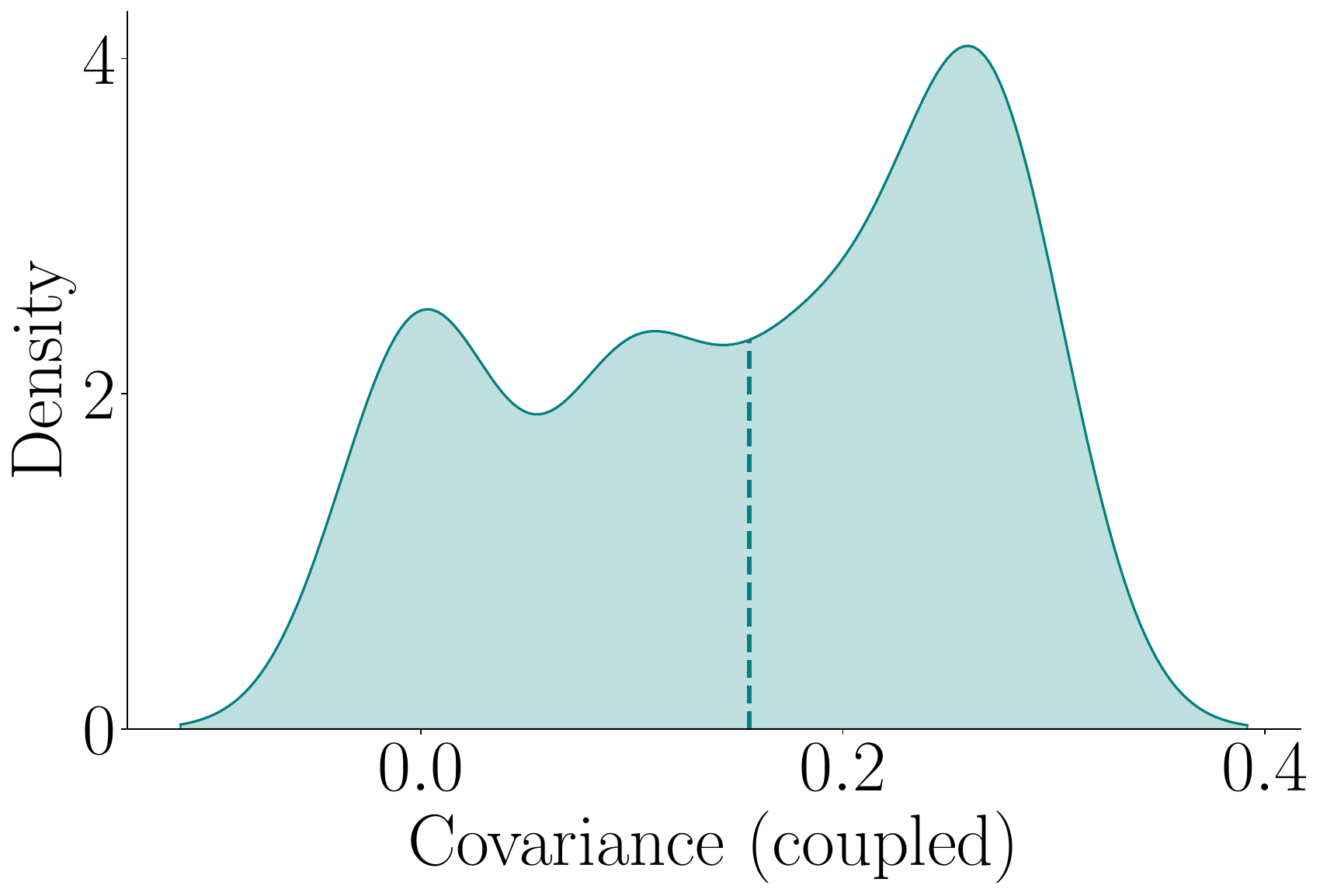} &
    \includegraphics[width=0.23\linewidth]{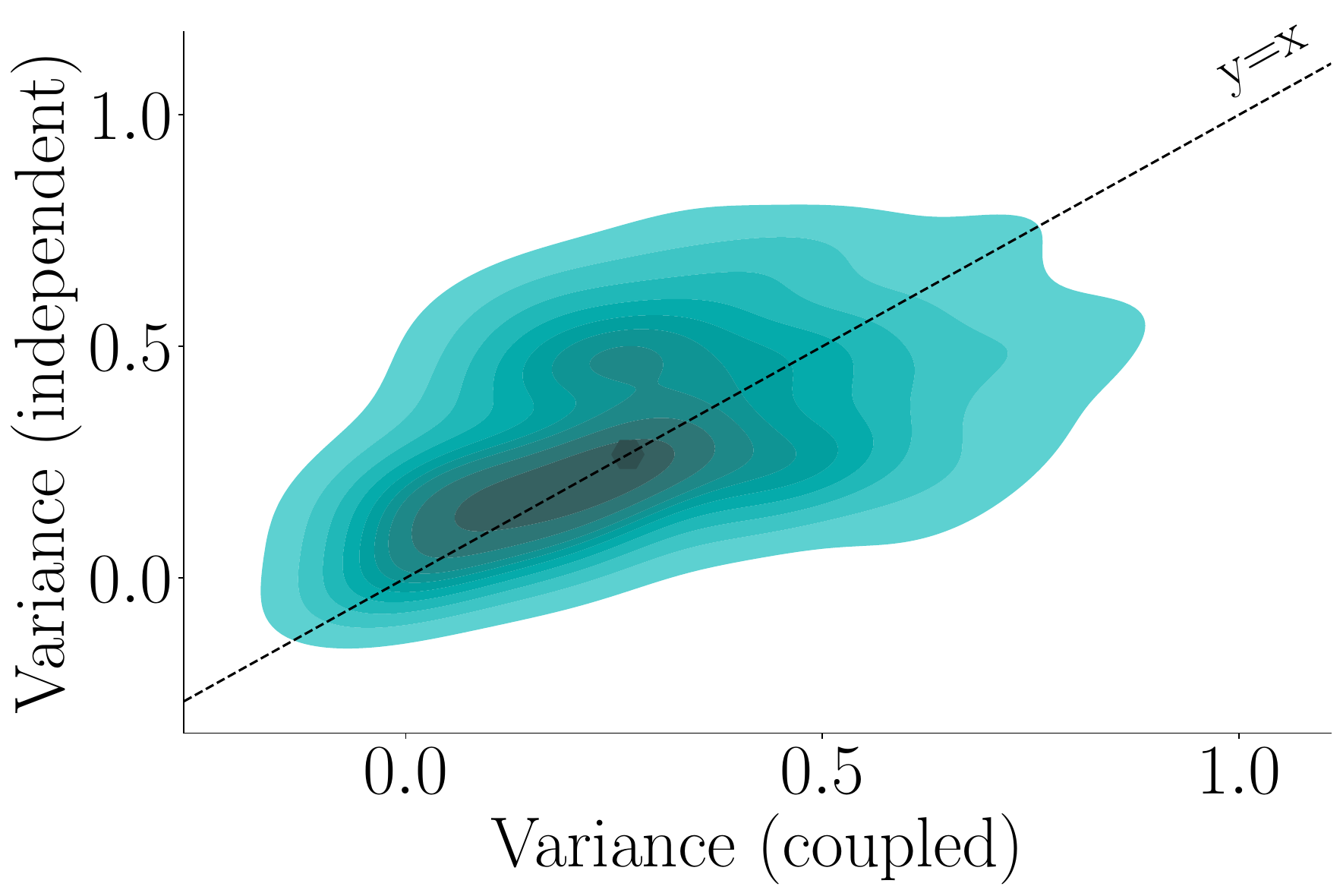} &
    \includegraphics[width=0.23\linewidth]{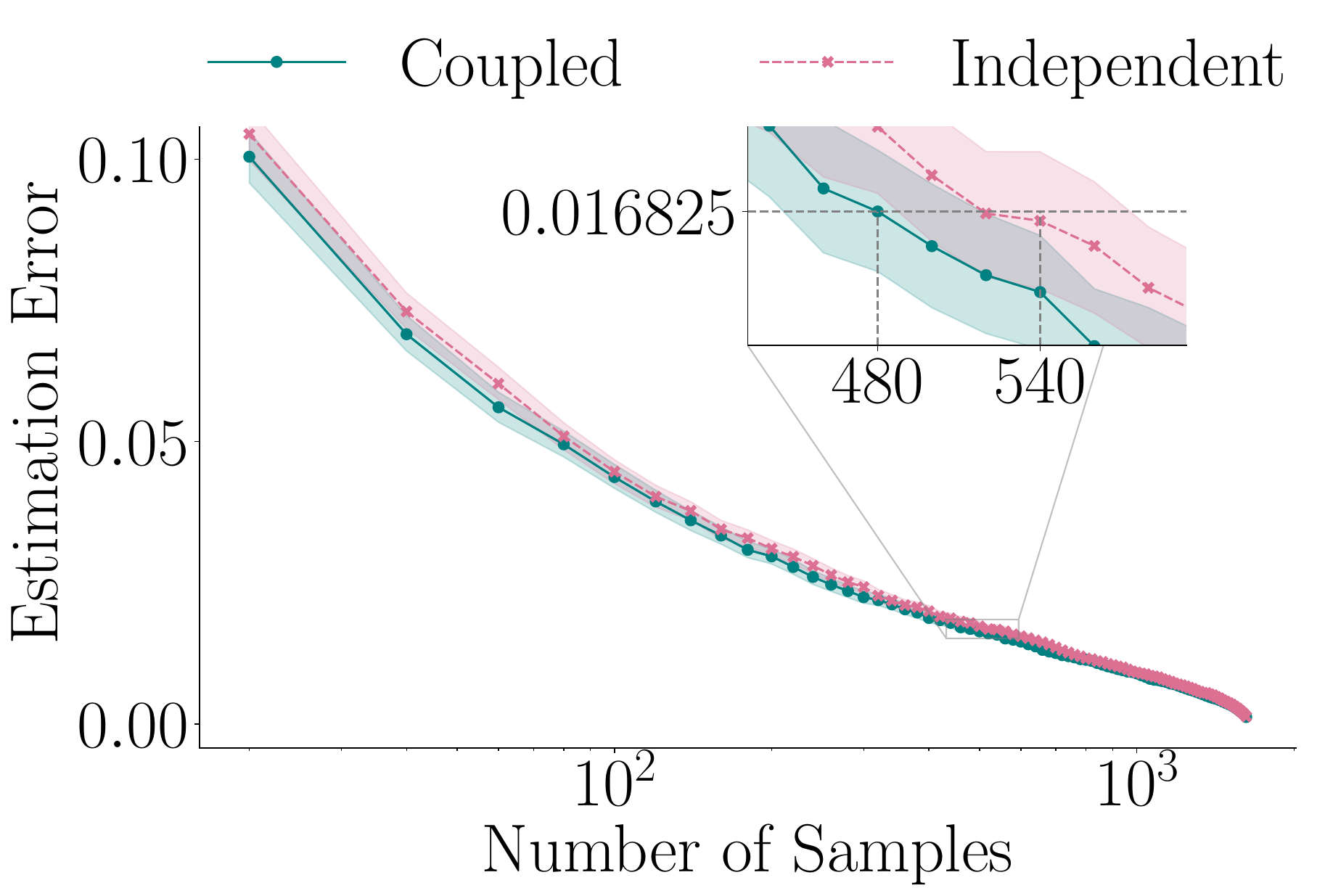} \\ \\
%
     \multicolumn{3}{c}{\texttt{2.5-7B} vs. \texttt{2.5-7B-AWQ-INT4}}\\
    \includegraphics[width=0.23\linewidth]{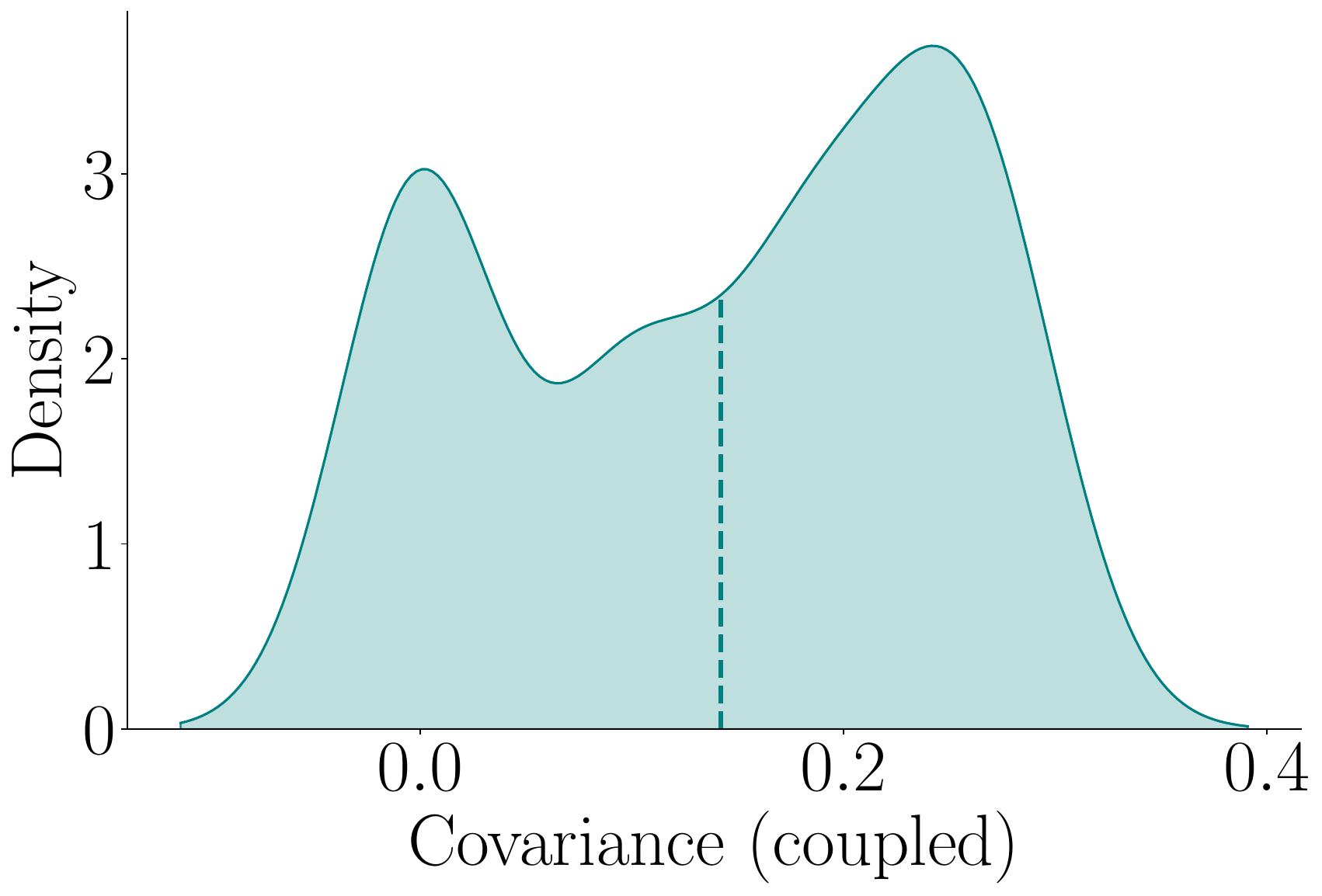} &
    \includegraphics[width=0.23\linewidth]{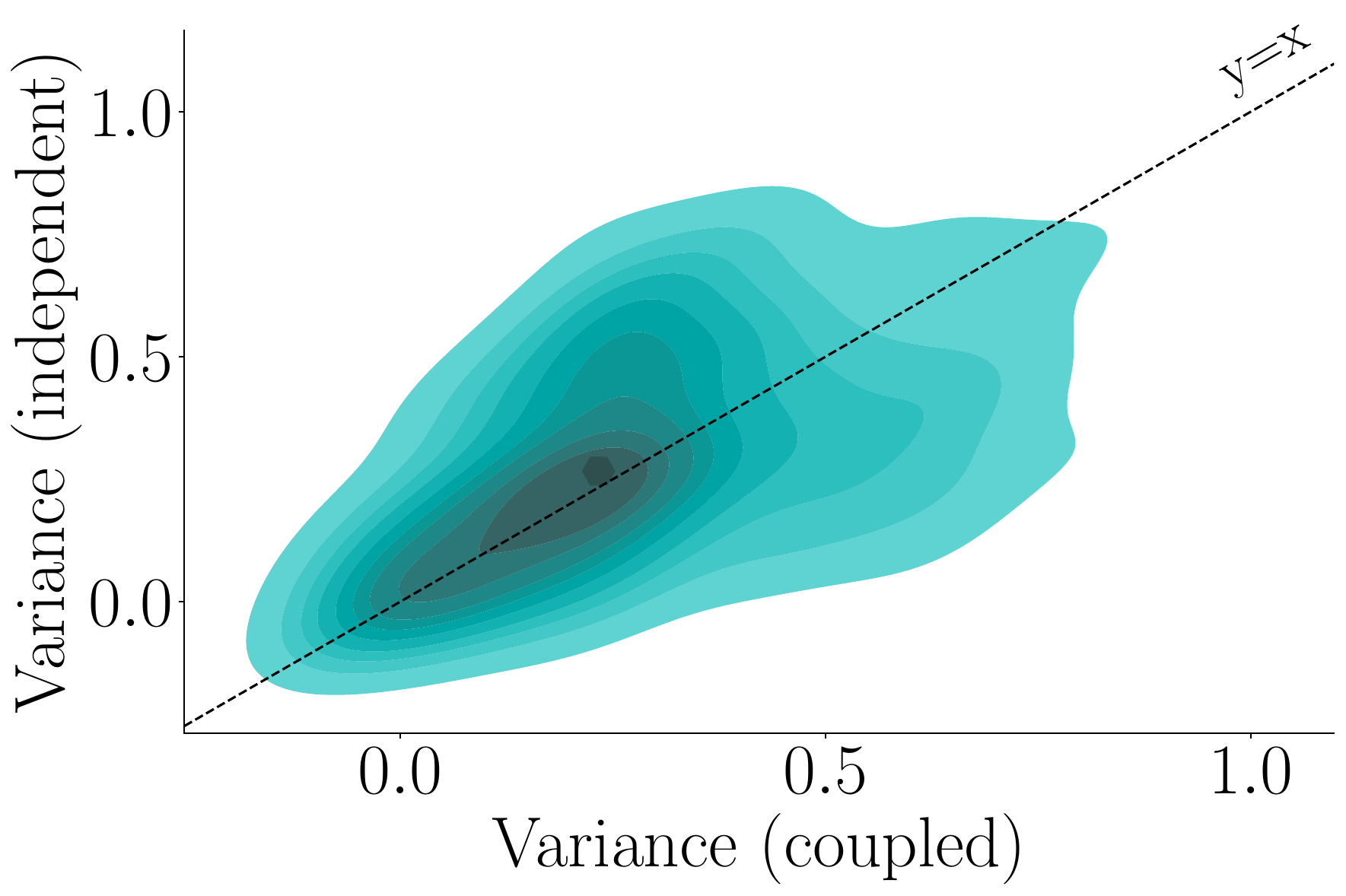} &
    \includegraphics[width=0.23\linewidth]{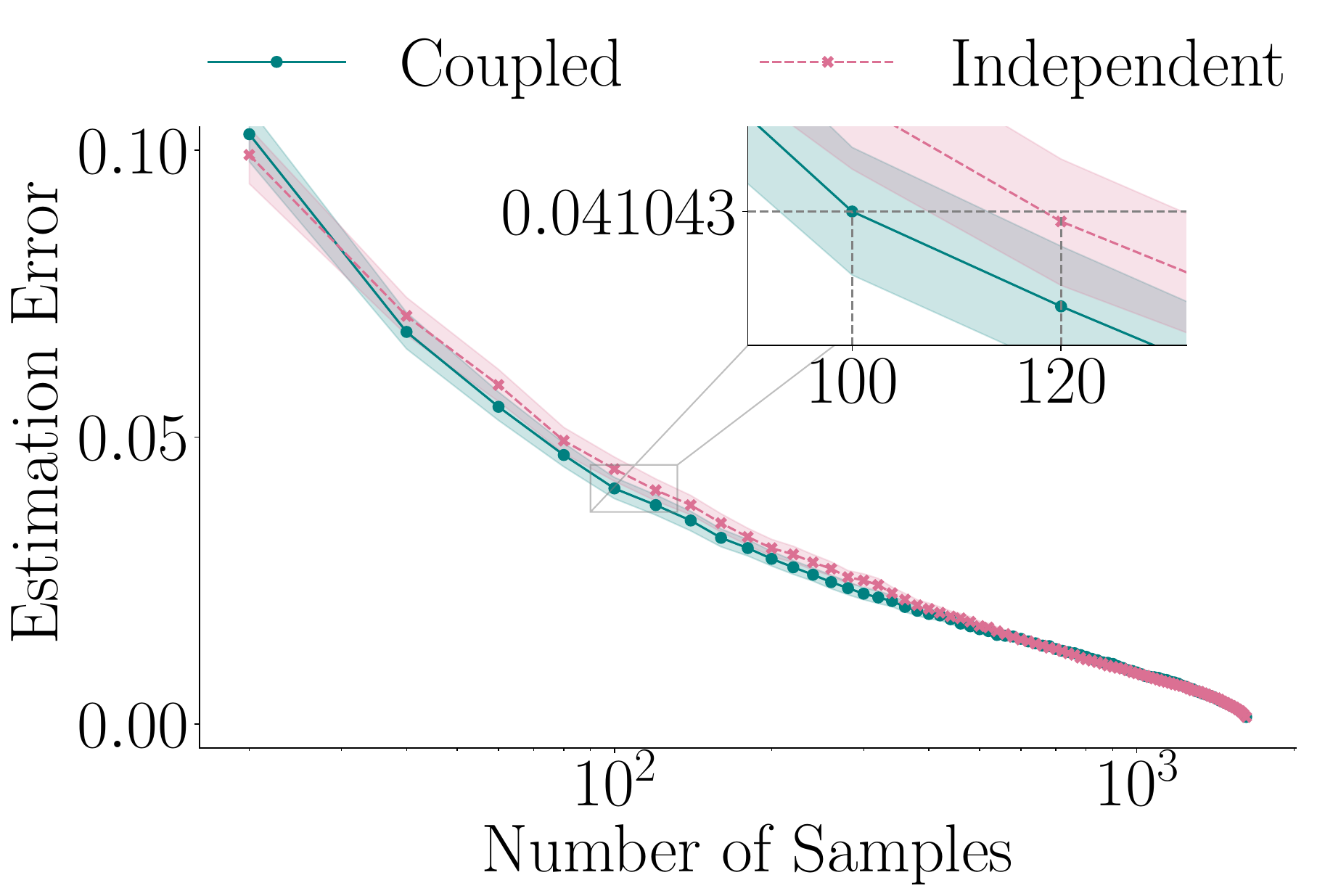} \\ \\
%
  \multicolumn{3}{c}{\texttt{2.5-7B-bnb-4bit} vs. \texttt{2.5-7B-AWQ-INT4}}\\
    \includegraphics[width=0.23\linewidth]{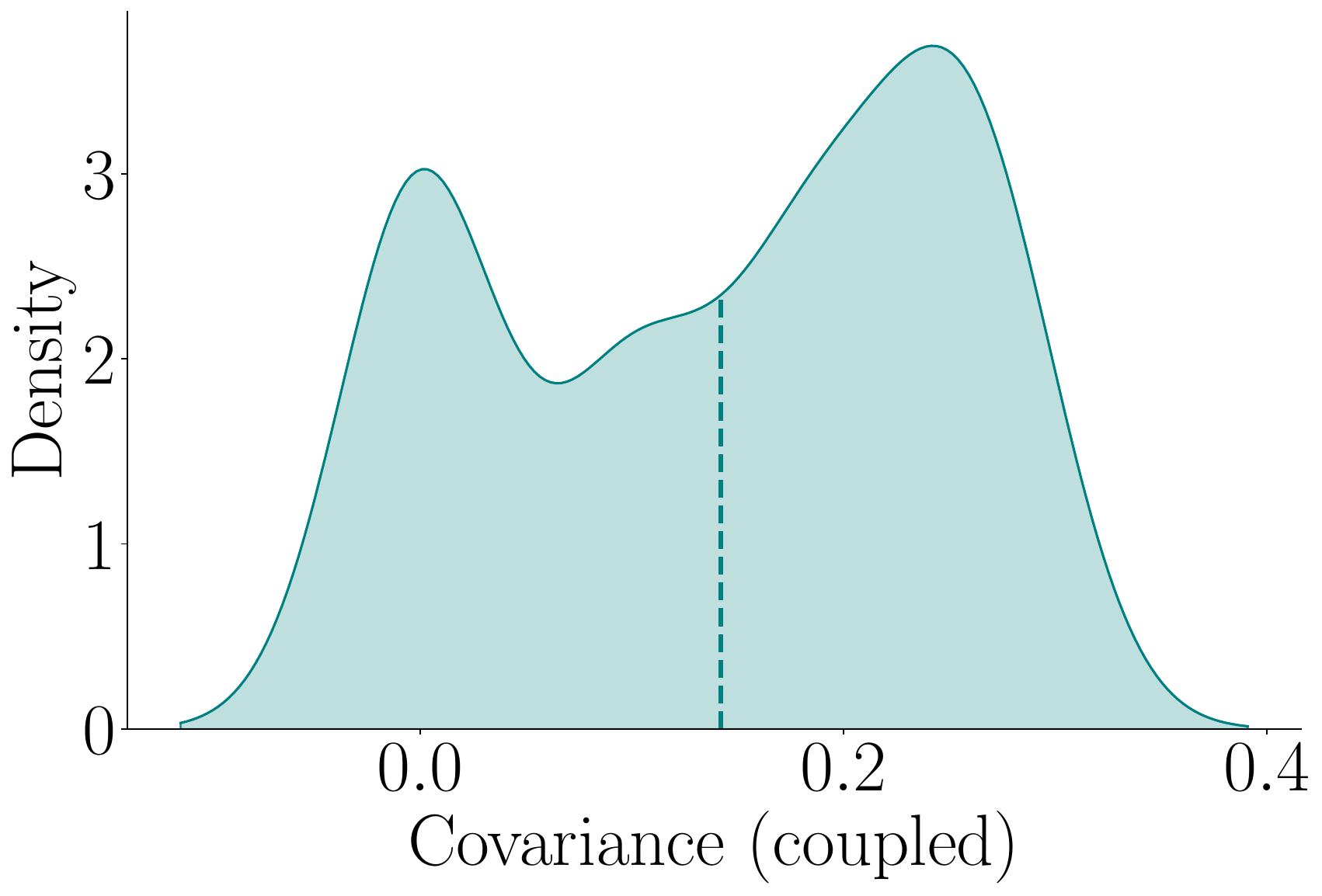} &
    \includegraphics[width=0.23\linewidth]{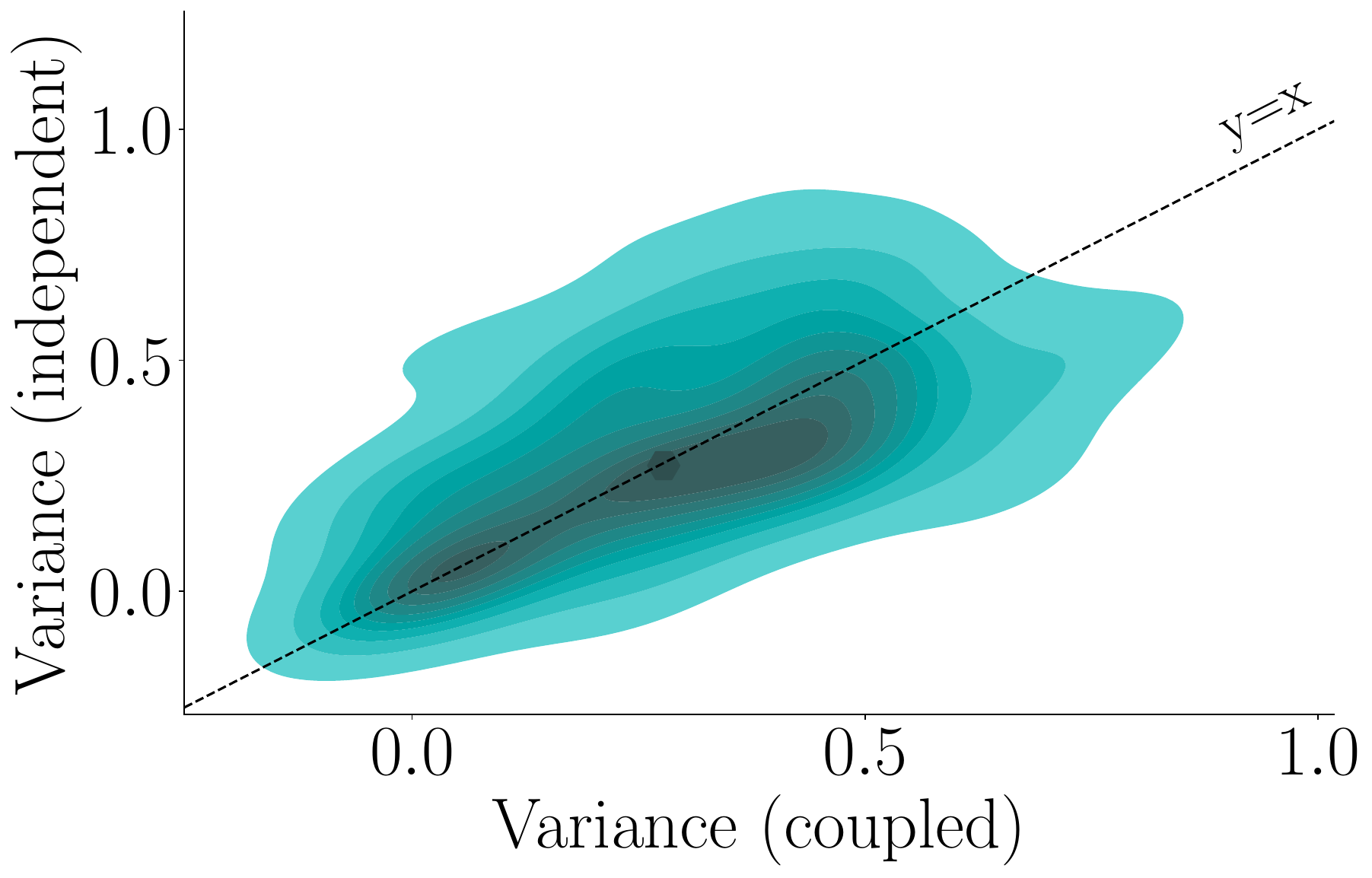} &
    \includegraphics[width=0.23\linewidth]{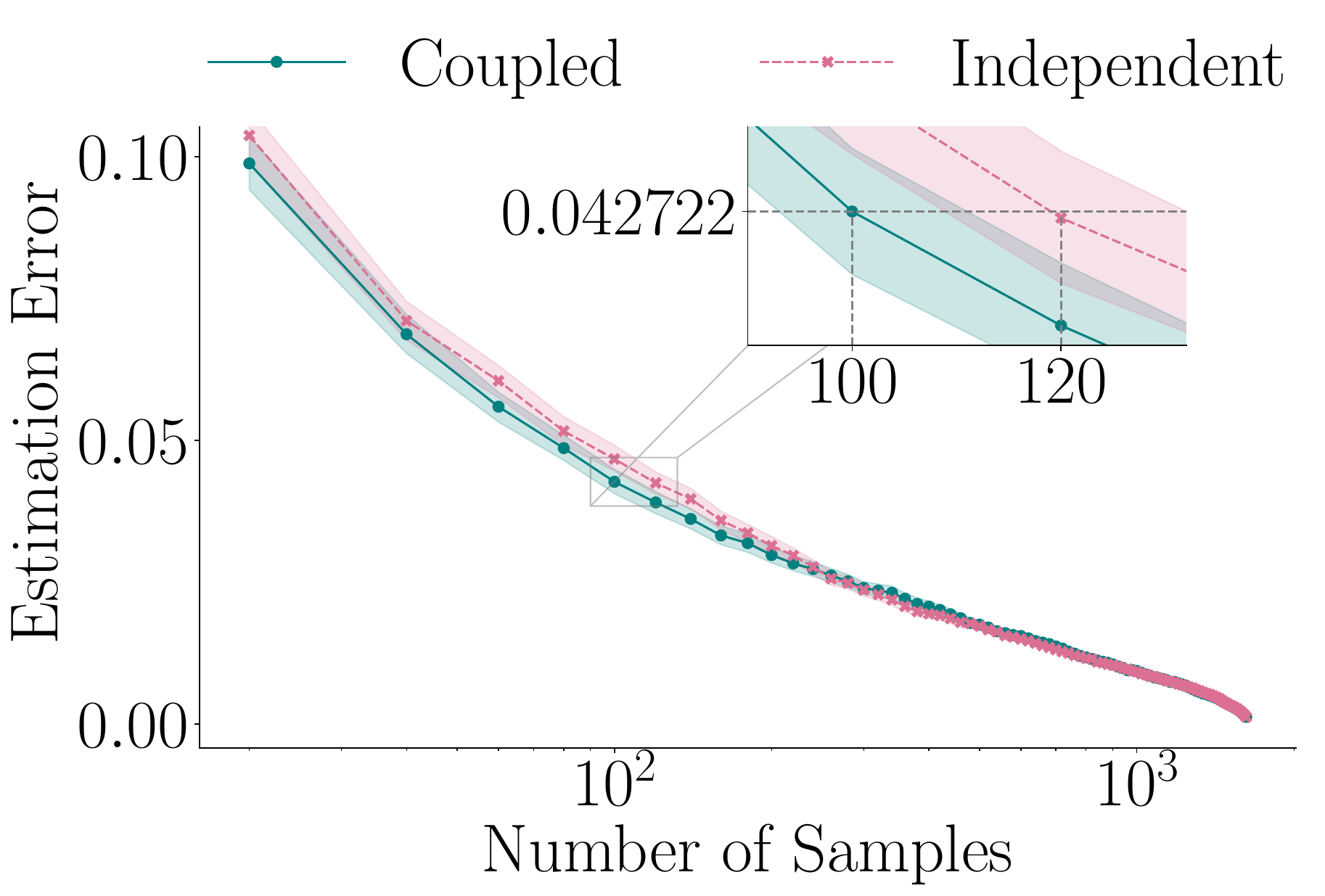} \\ \\
 
%
  \multicolumn{3}{c}{\texttt{2.5-7B} vs. \texttt{2.5-7B-distil}}\\
    \includegraphics[width=0.23\linewidth]{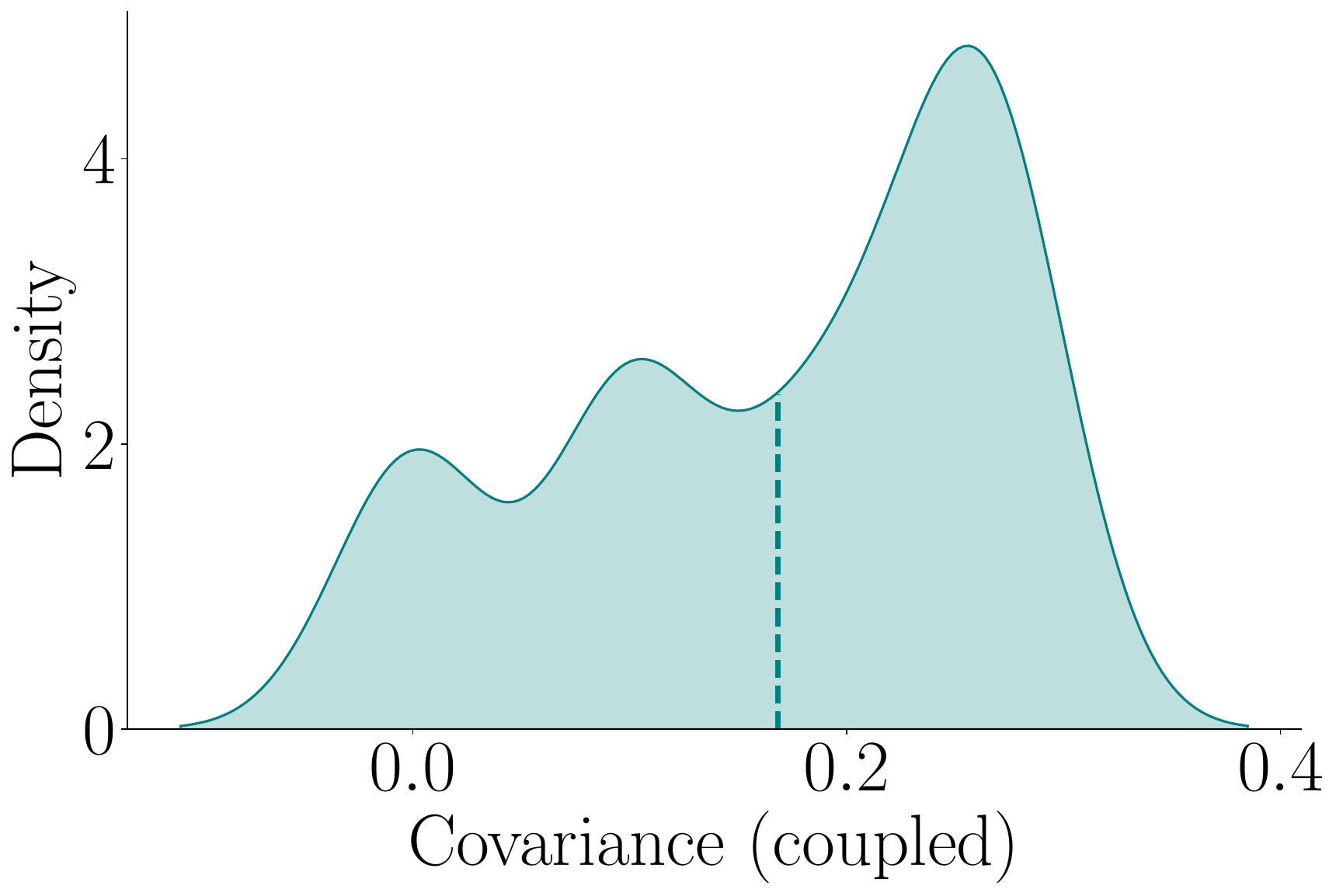} &
    \includegraphics[width=0.23\linewidth]{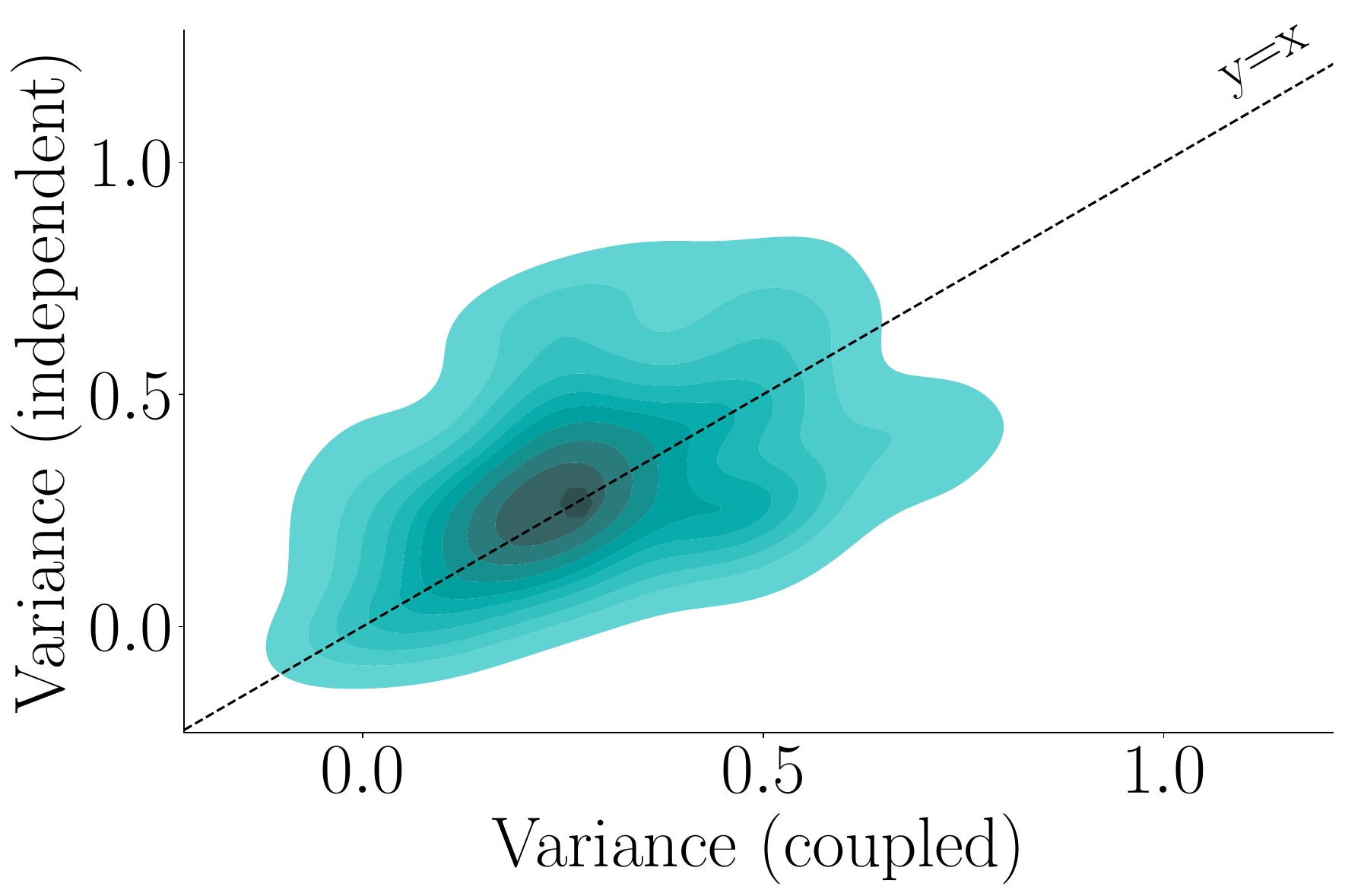} &
    \includegraphics[width=0.23\linewidth]{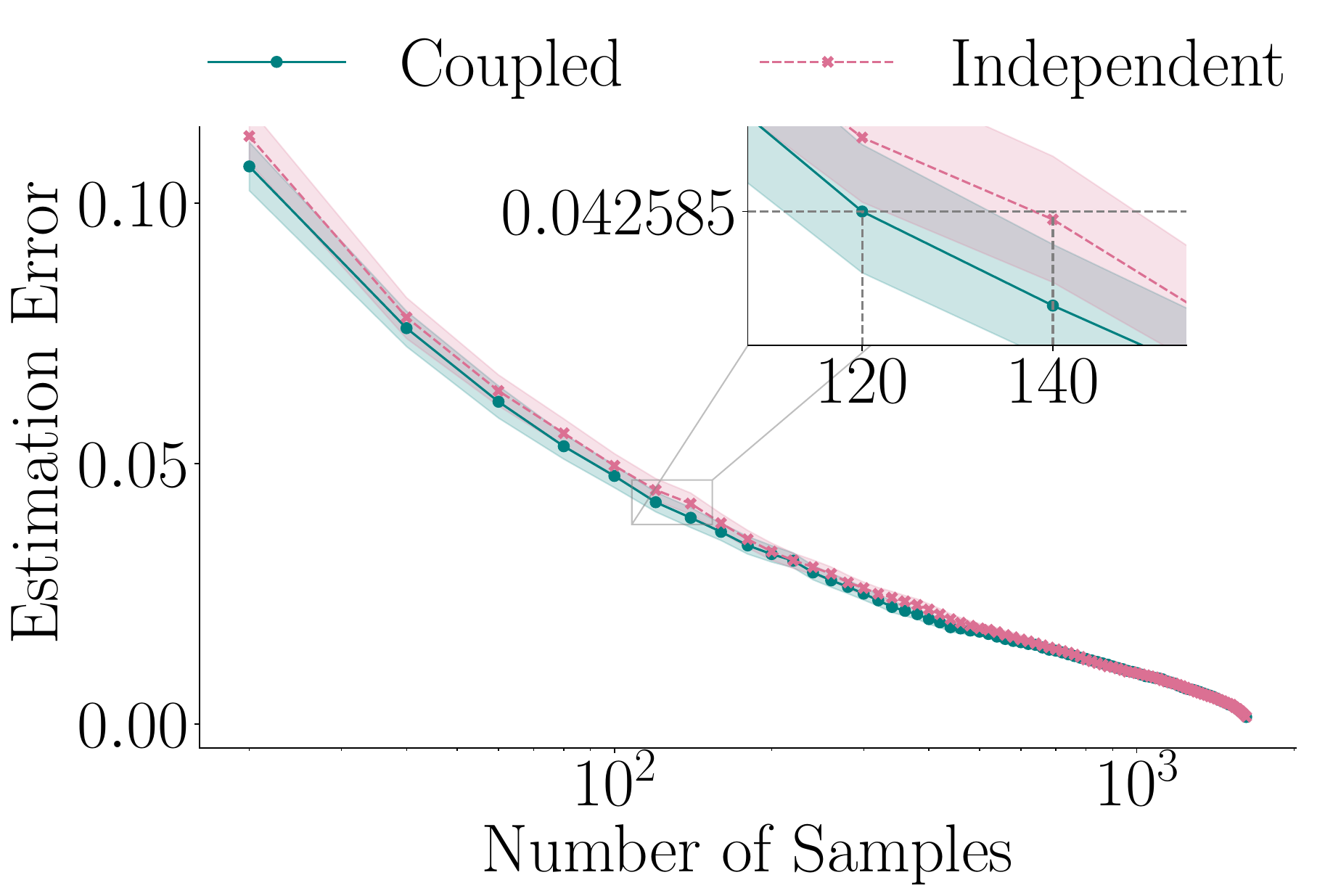} \\ \\

    (a) Score covariance & (b) Variance of the score difference & (c) Estimation error vs. \# samples \\ 
\end{tabular}
     \caption{\textbf{Comparison between several pairs of LLMs in the \texttt{Qwen} family on programming problems from the HumanEval dataset.}
    Panels in column (a) show the kernel density estimate (KDE) of the covariance between the scores of the two LLMs on each problem under coupled generation; the dashed lines correspond to average values. Panels in column (b) show the KDE of the variance of the difference between the scores of the LLMs on each question under coupled and independent generation; the highlighted points correspond to median values. Panels in column (c) show the absolute error in the estimation of the expected difference between the scores of the LLMs against the number of samples; for each point on the x-axis, we perform $1{,}000$ sub-samplings and shaded areas correspond to $95\%$ confidence intervals.}
    \label{fig:human-eval-qwen-first-5}
\end{figure}
\vspace{-0.2cm}

\begin{figure}[!!h]
\centering
\begin{tabular}{c c c}
     \multicolumn{3}{c}{\texttt{2.5-3B-distil} vs. \texttt{2.5-7B-bnb-8bit}}\\
    \includegraphics[width=0.23\linewidth]{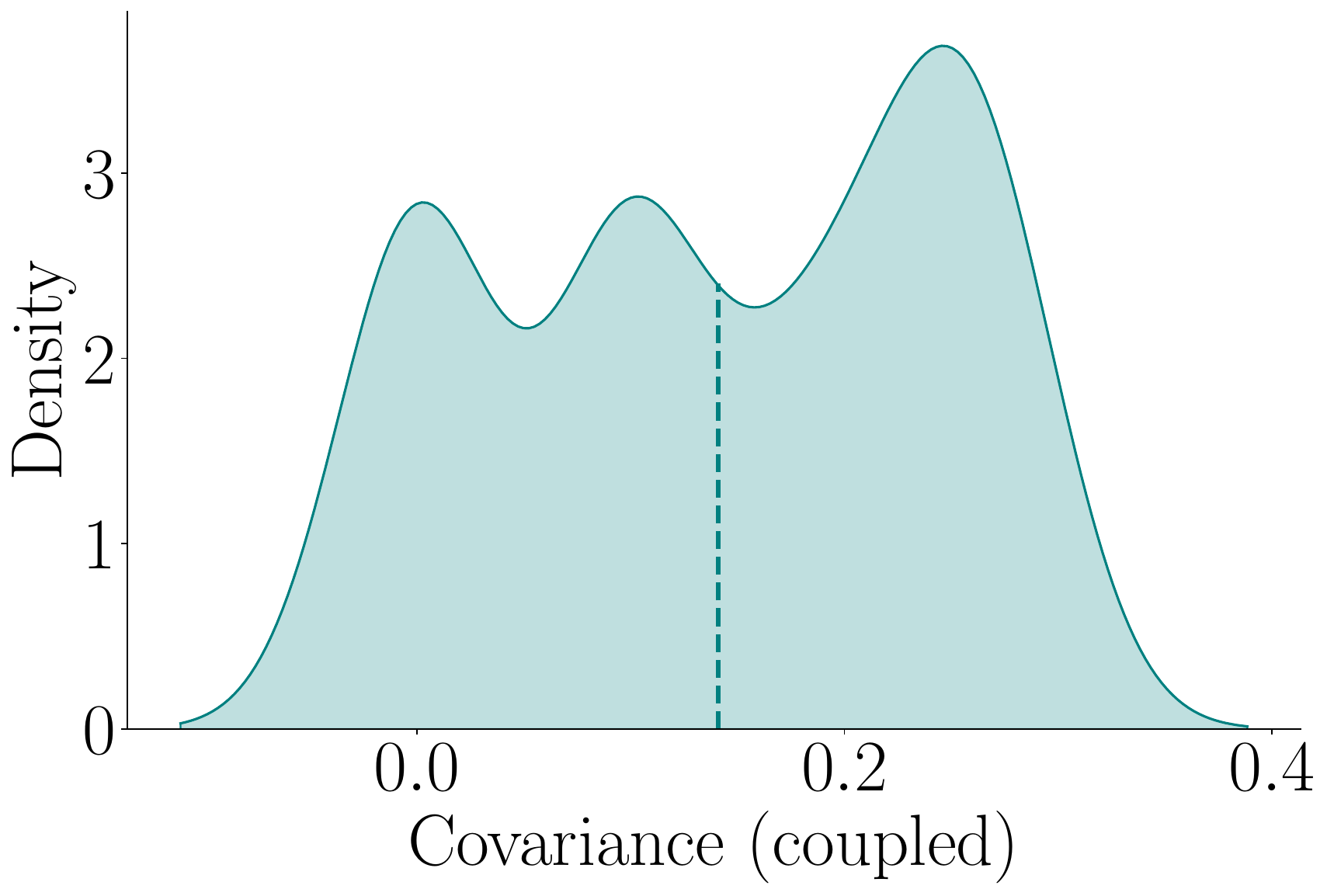} &
    \includegraphics[width=0.23\linewidth]{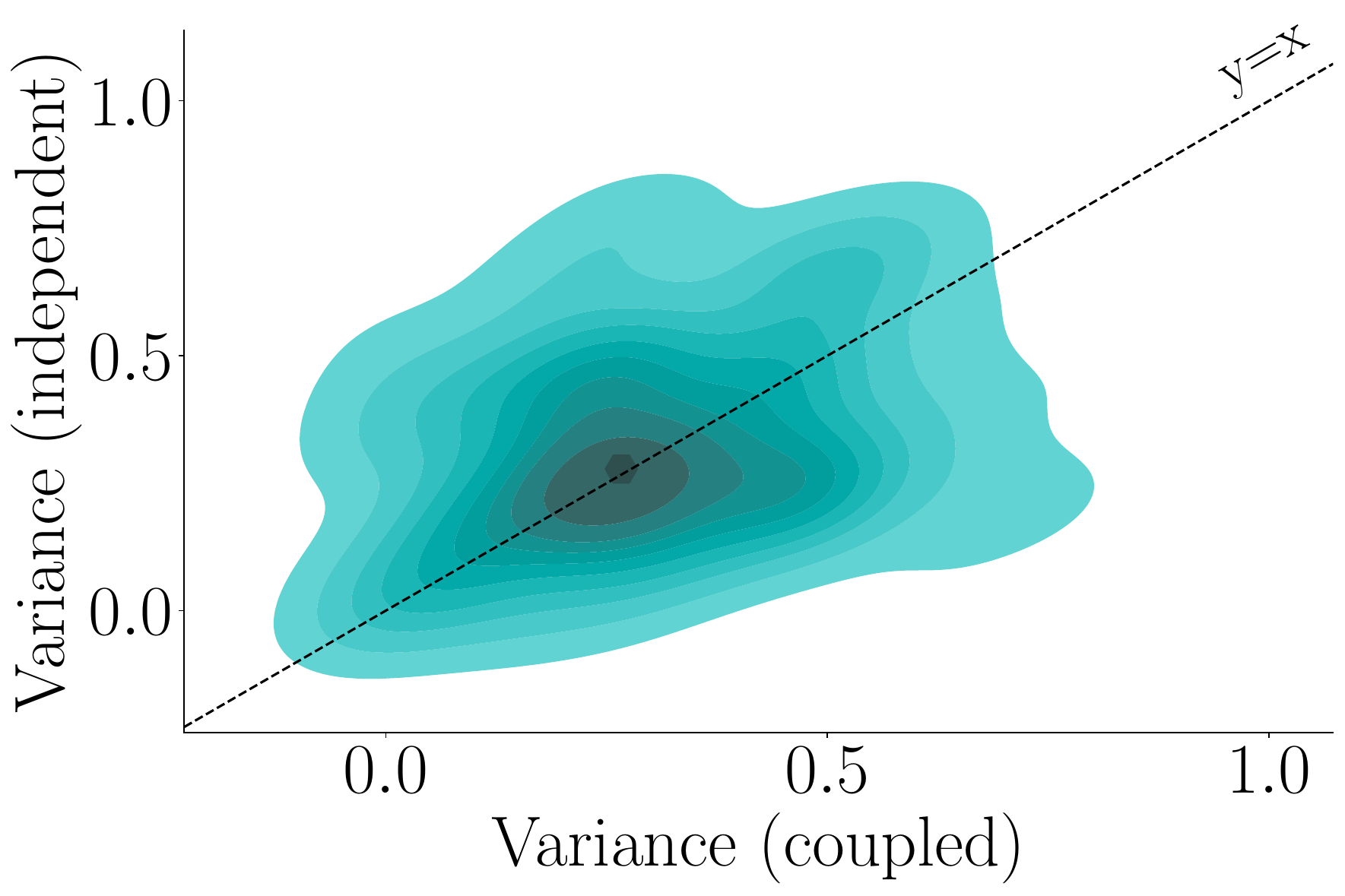} &
    \includegraphics[width=0.23\linewidth]{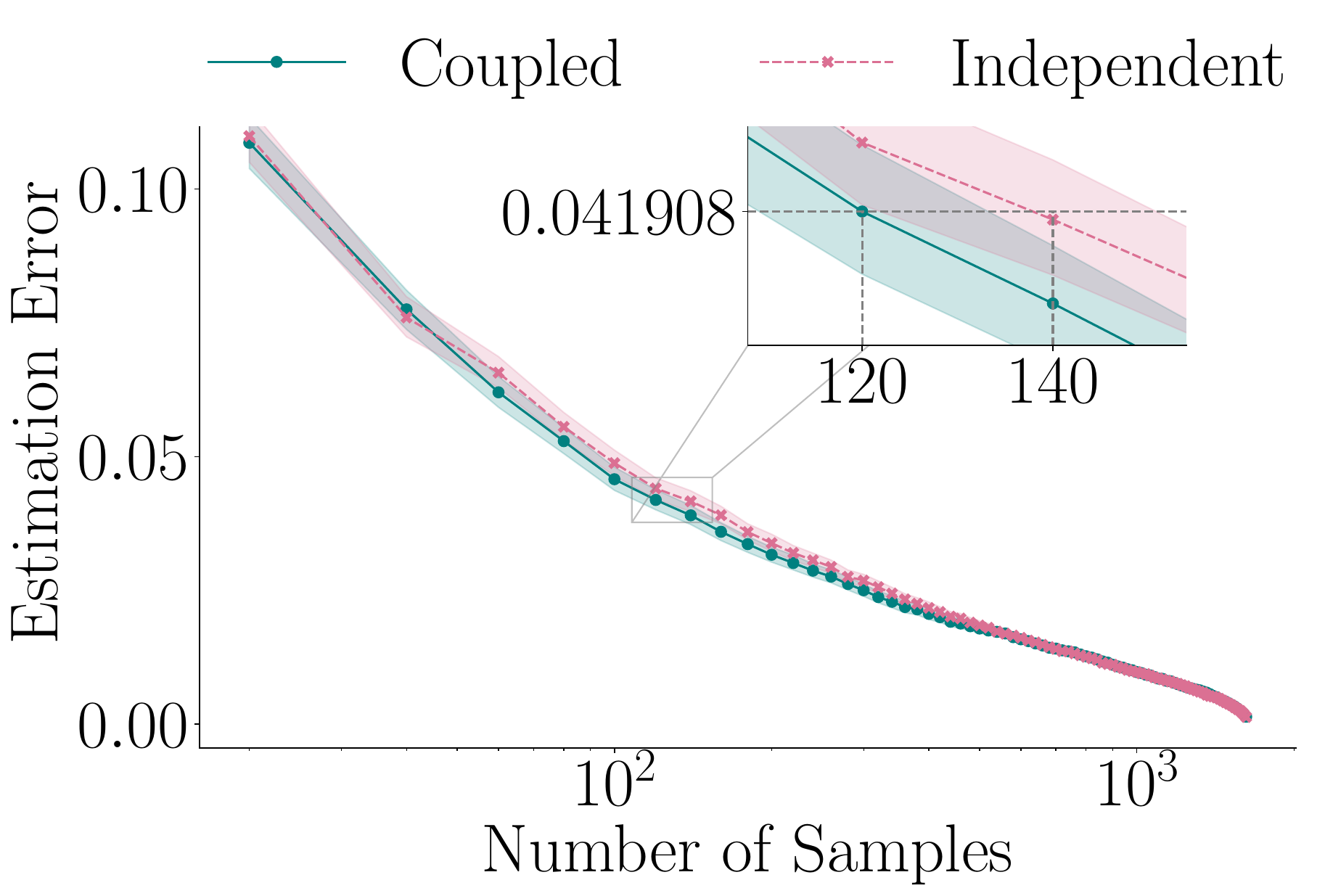} \\ \\
%
    \multicolumn{3}{c}{\texttt{2.5-3B-distil} vs. \texttt{2.5-7B-AWQ-INT4}}\\
    \includegraphics[width=0.23\linewidth]{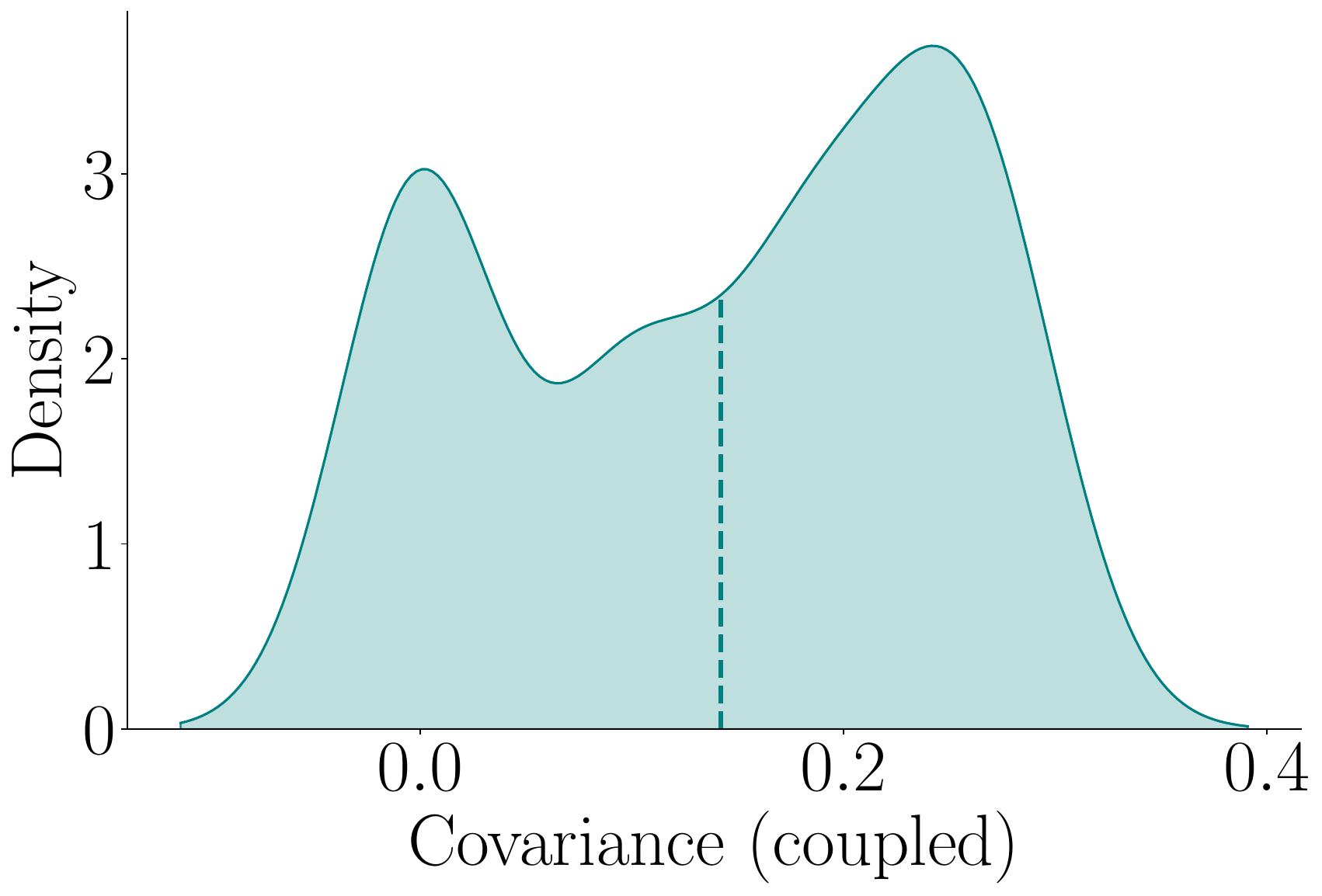} &
    \includegraphics[width=0.23\linewidth]{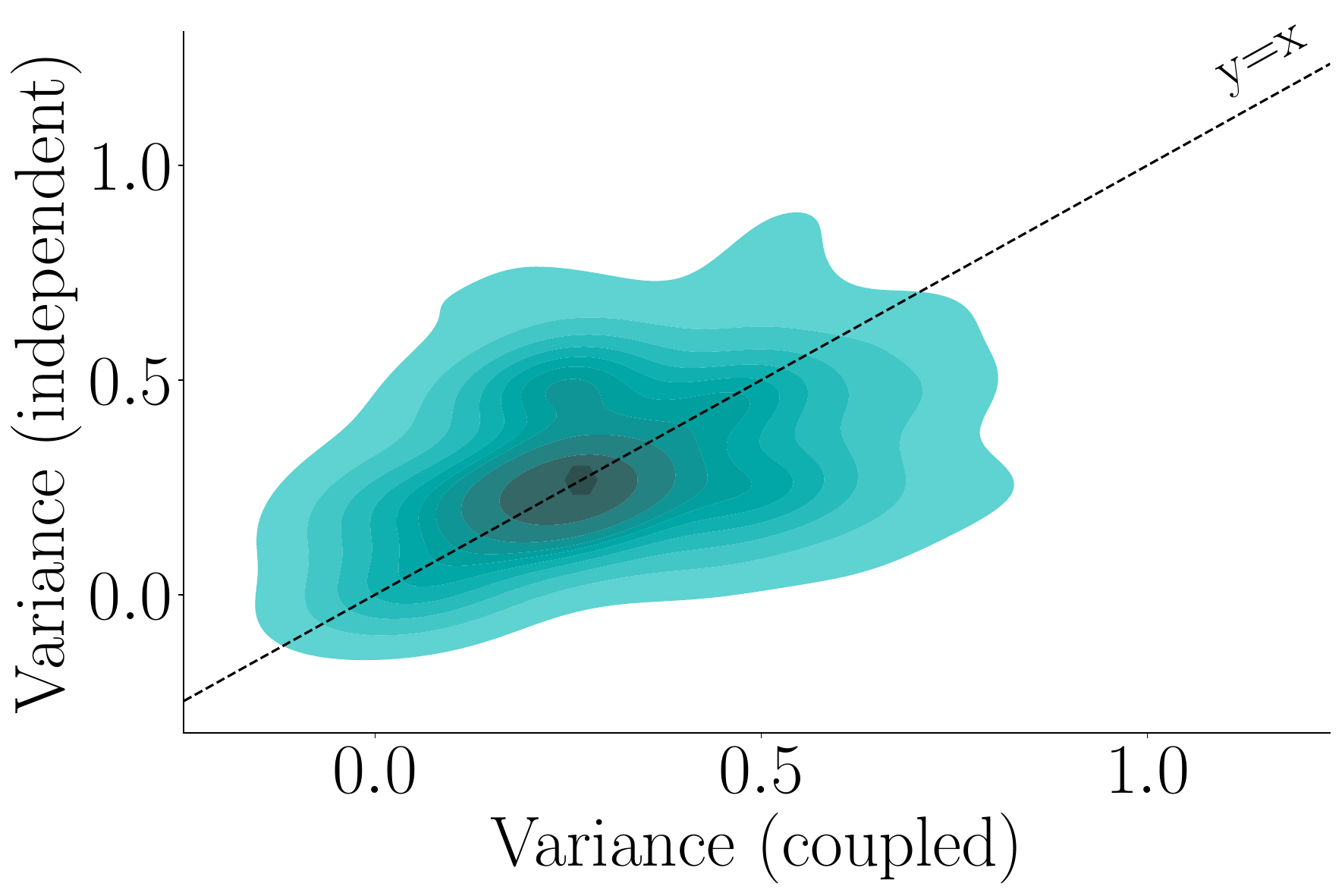} &
    \includegraphics[width=0.23\linewidth]{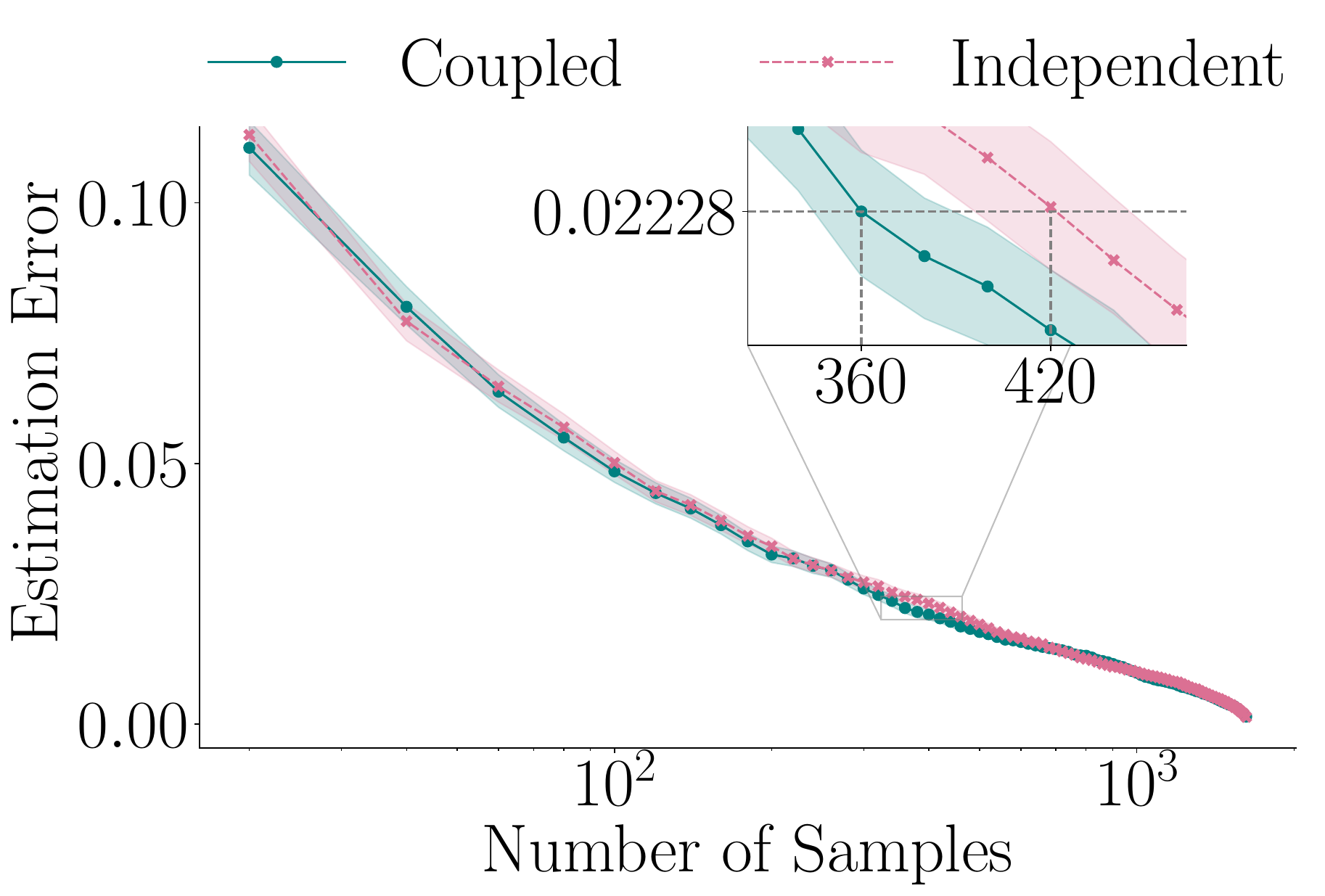} \\ \\

  \multicolumn{3}{c}{\texttt{2.5-3B-distil} vs. \texttt{2.5-1.5B}}\\
    \includegraphics[width=0.23\linewidth]{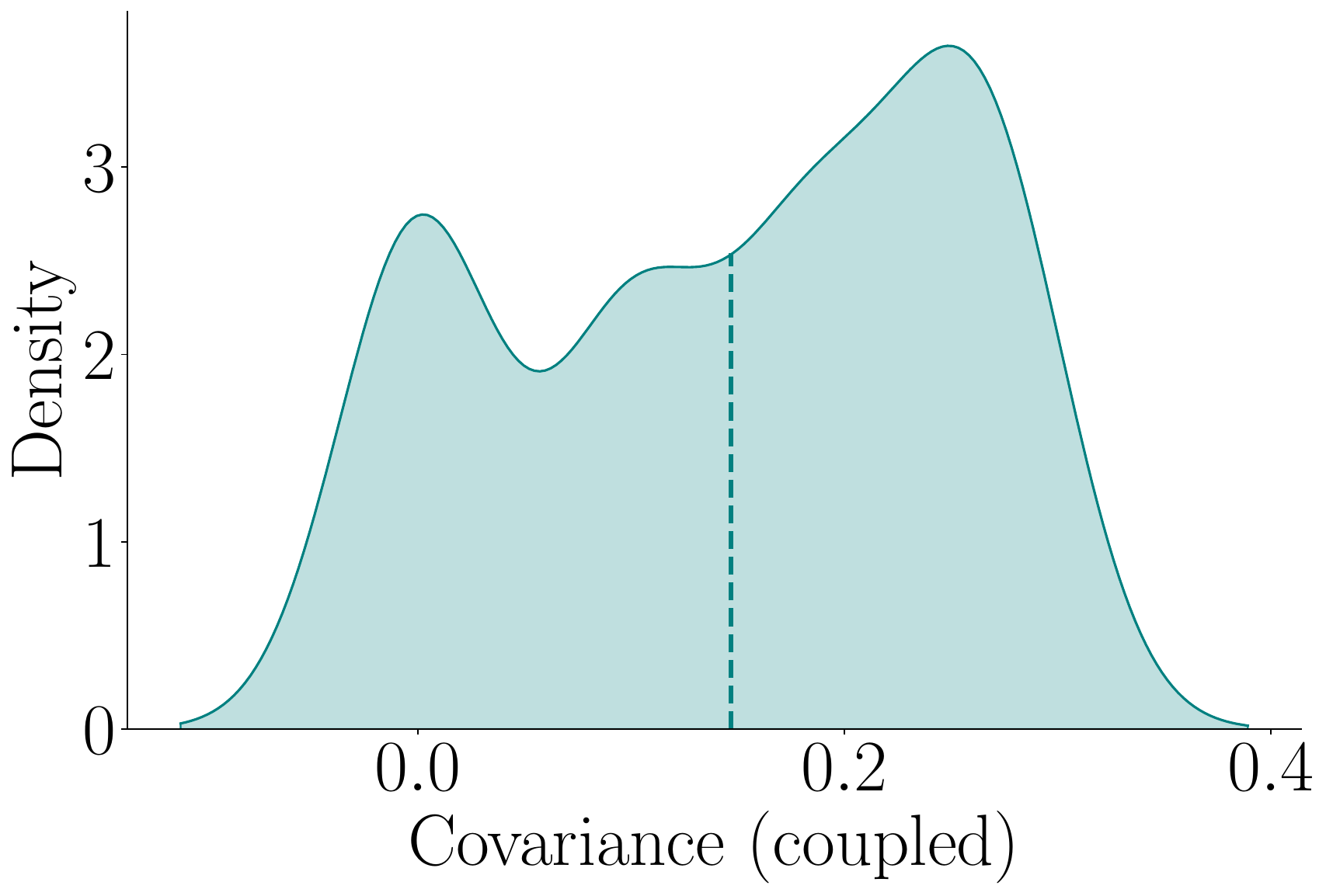} &
    \includegraphics[width=0.23\linewidth]{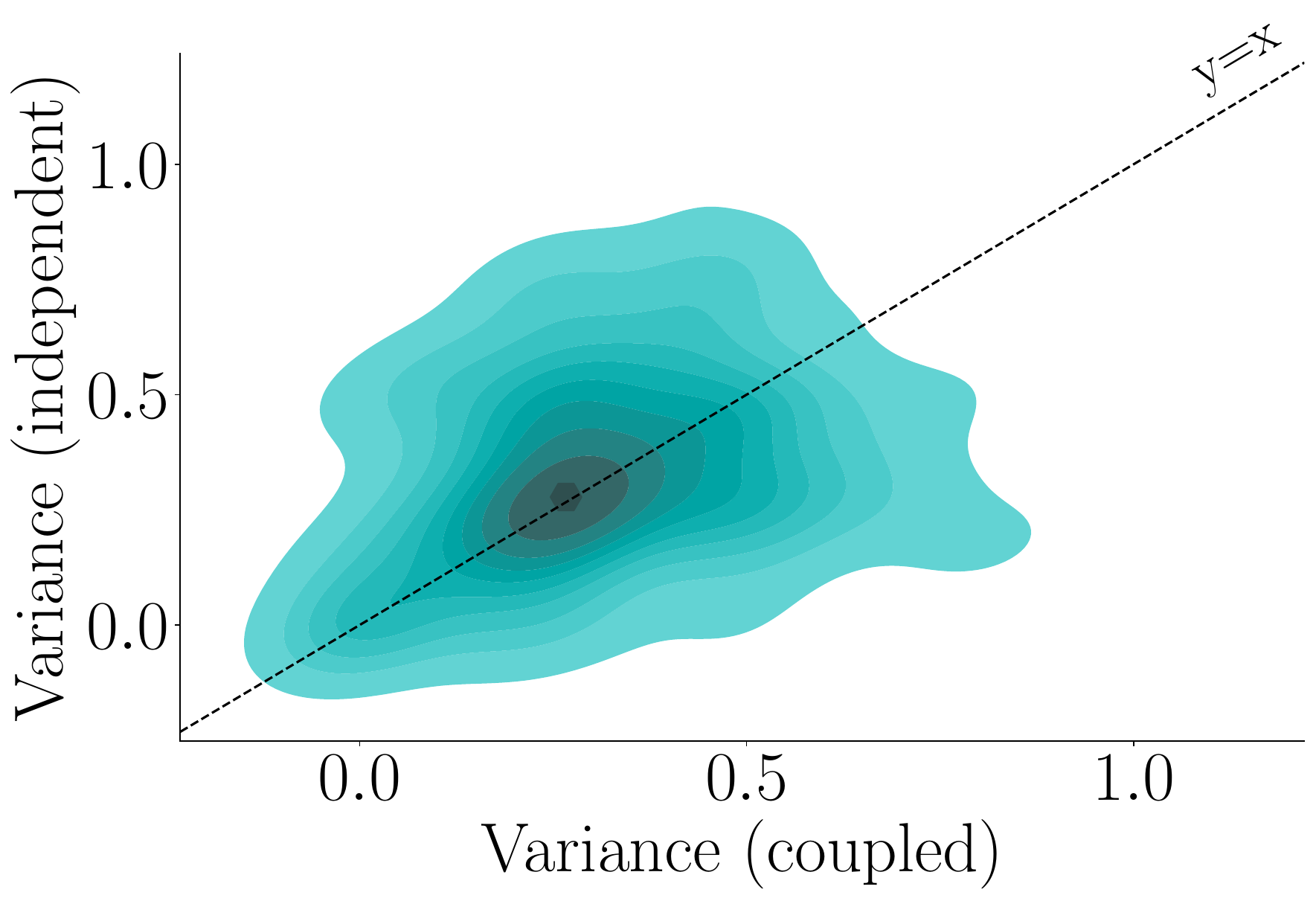} &
    \includegraphics[width=0.23\linewidth]{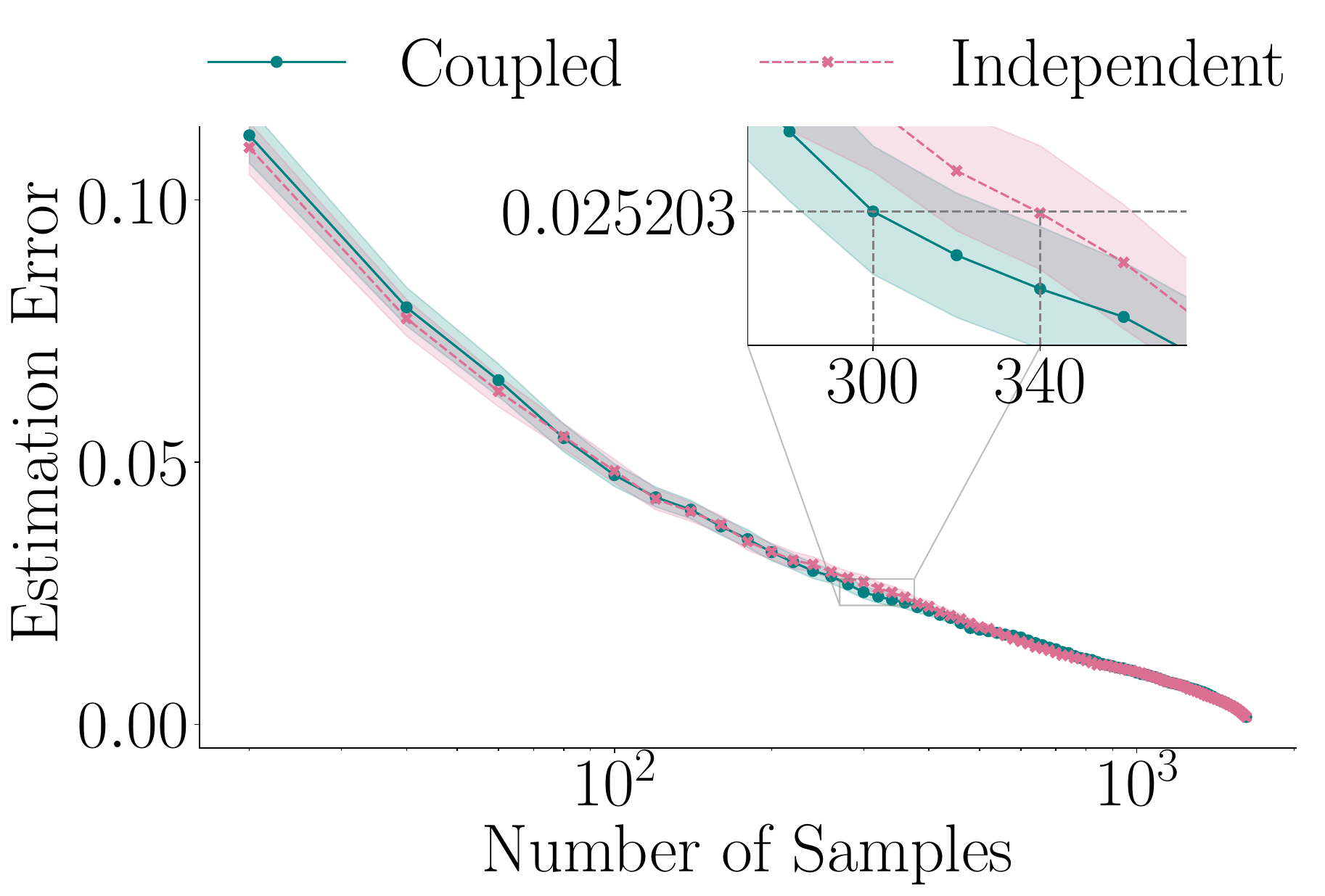} \\ \\

%
  \multicolumn{3}{c}{\texttt{2.5-1.5B} vs. \texttt{3-8B}}\\
    \includegraphics[width=0.23\linewidth]{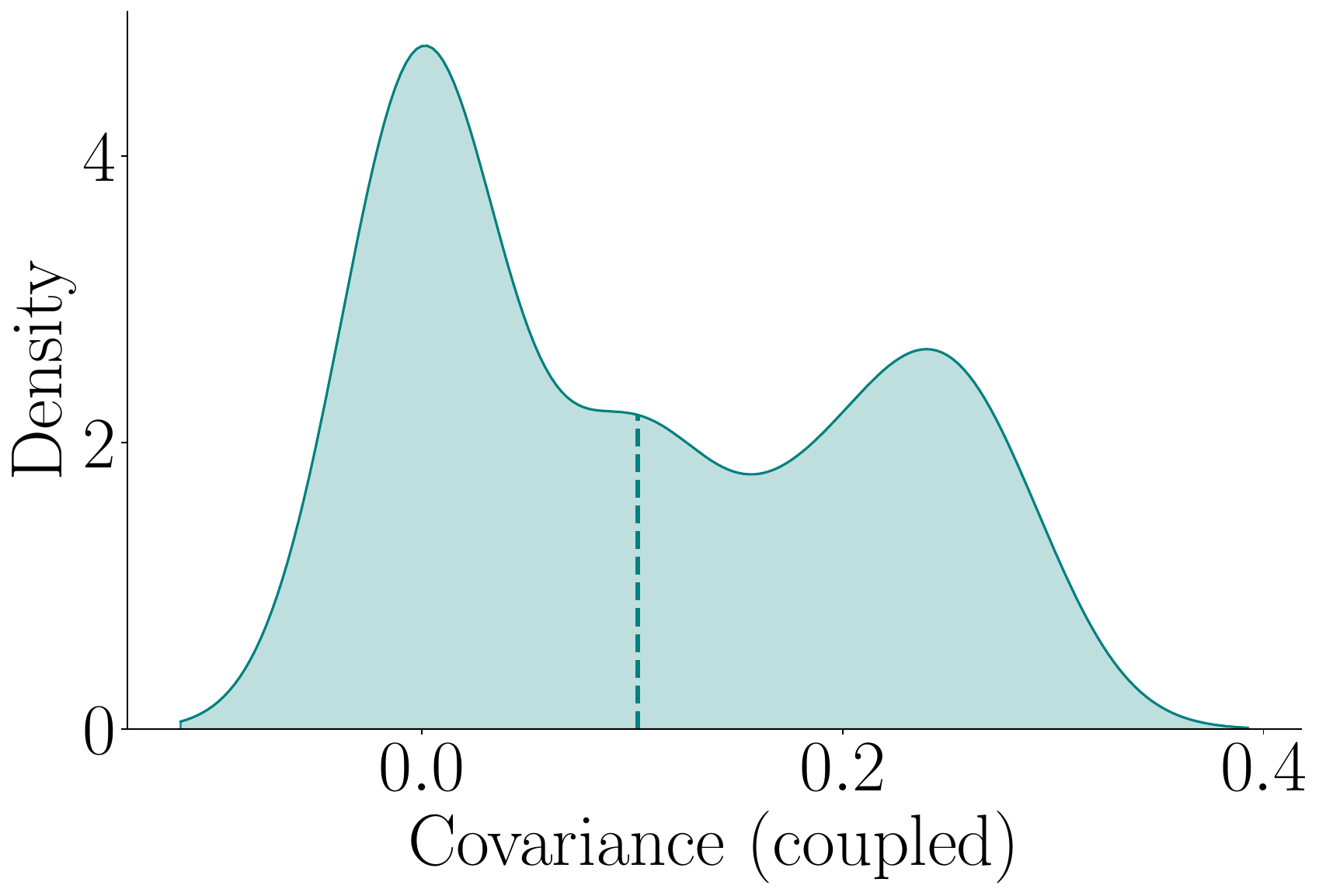} &
    \includegraphics[width=0.23\linewidth]{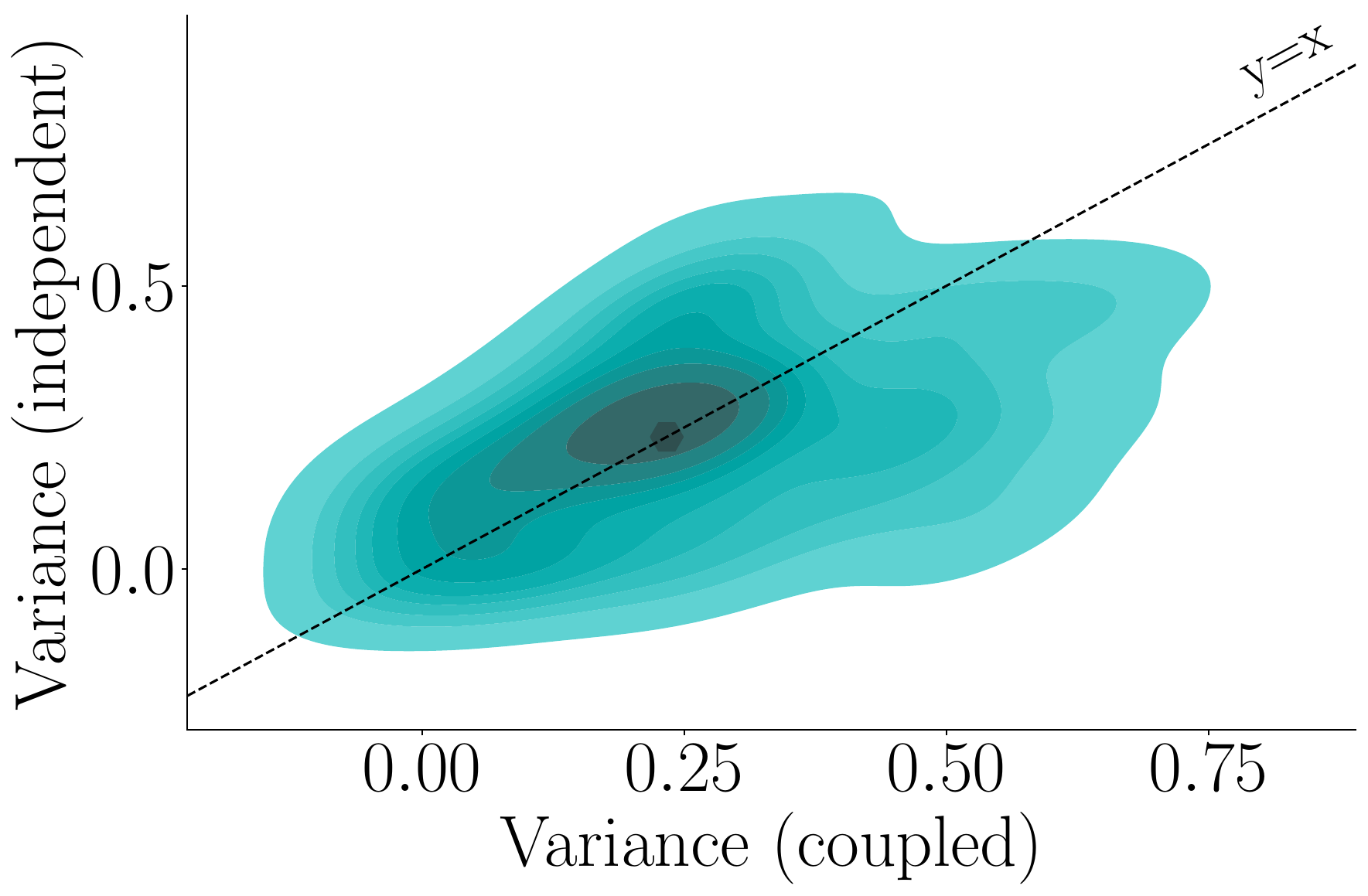} &
    \includegraphics[width=0.23\linewidth]{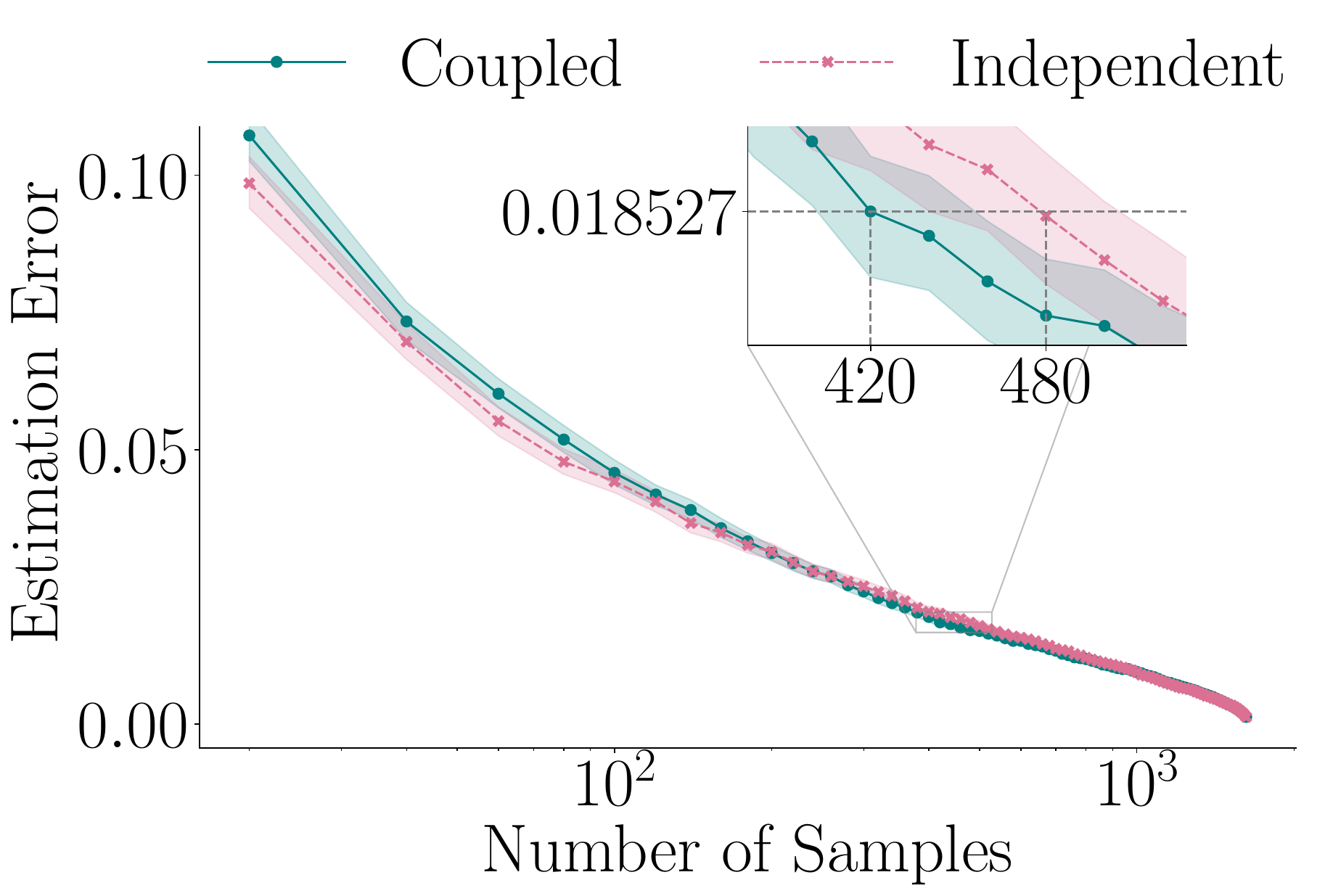} \\ \\

    (a) Score covariance & (b) Variance of the score difference & (c) Estimation error vs. \# samples \\ 
\end{tabular}
     \caption{\textbf{Comparison between several pairs of LLMs in the \texttt{Qwen} family on programming problems from the HumanEval dataset.}
    Panels in column (a) show the kernel density estimate (KDE) of the covariance between the scores of the two LLMs on each problem under coupled generation; the dashed lines correspond to average values. Panels in column (b) show the KDE of the variance of the difference between the scores of the LLMs on each question under coupled and independent generation; the highlighted points correspond to median values. Panels in column (c) show the absolute error in the estimation of the expected difference between the scores of the LLMs against the number of samples; for each point on the x-axis, we perform $1{,}000$ sub-samplings and shaded areas correspond to $95\%$ confidence intervals.}
    \label{fig:human-eval-qwen-last-4}
\end{figure}
\vspace{-0.2cm}

\subsection{Pairwise Comparisons}

\begin{figure}[h]
\centering
\includegraphics[width=0.72\linewidth]{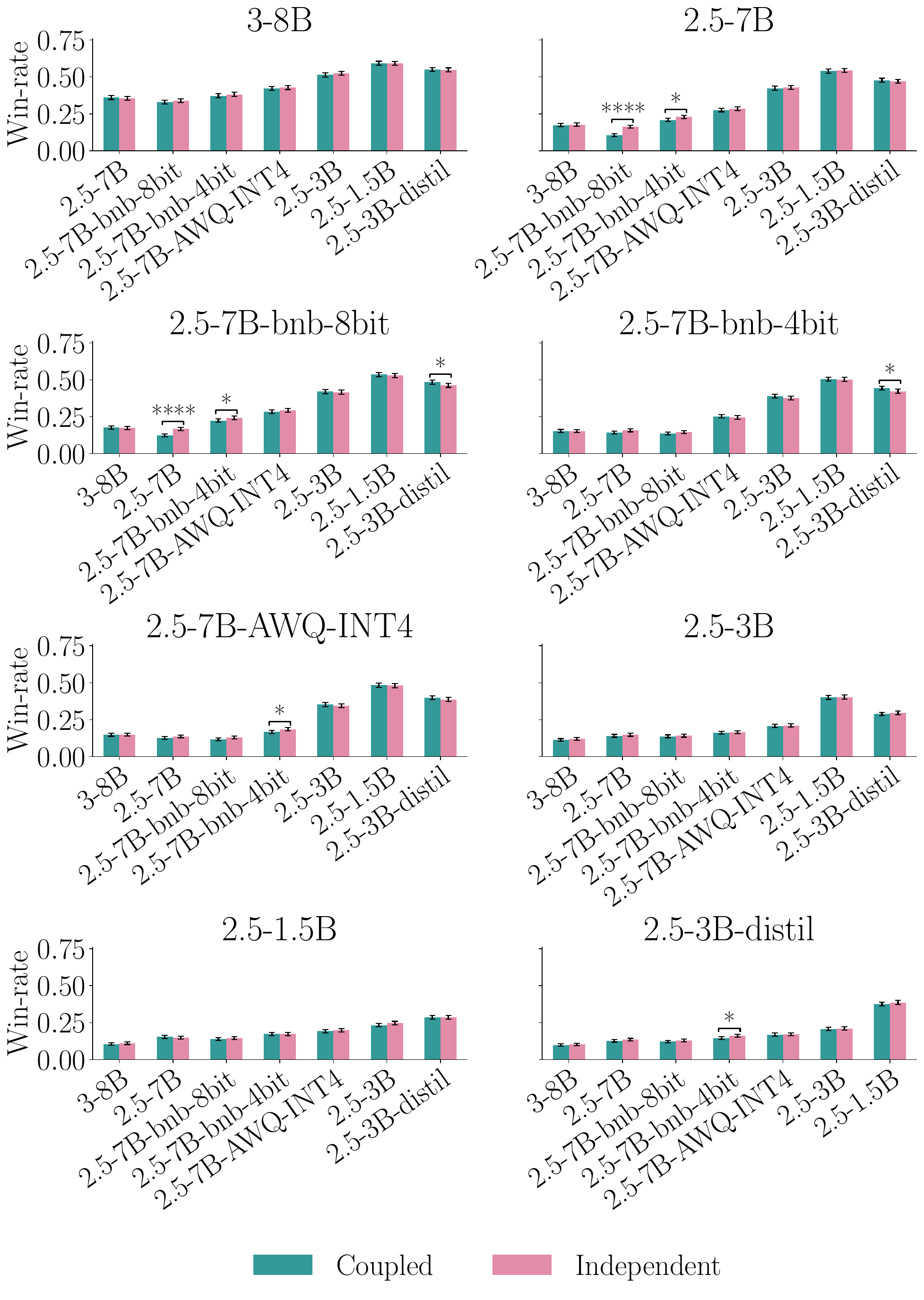}
\caption{
\textbf{Empirical win-rate of each LLM against other LLMs in the \texttt{Qwen} family
on questions from the LMSYS-Chat-1M dataset.} 
Empirical estimate of the win-rate under coupled autoregressive generation as given by Eq.~\ref{eq:coupled-generation-win-rates} and under independent generation generation as given by  Eq.~\ref{eq:independent-generation-win-rates}. 
Each empirical win-rate is computed using pairwise comparisons between the outputs of each LLM and any other LLM over $500$ questions with $10$ (different) random seeds.
The error bars correspond to $95\%$ confidence intervals.
For each pair of empirical win-rates, we conduct a two-tailed test, to test the hypothesis that the empirical win-rates are the same; (\fourstars, \threestars, \twostars, \onestar) indicate $p$-values ($<0.0001$, $<0.001$, $<0.01$, $<0.05$), respectively.
}
\label{fig:lmsys-all-qwen}
\end{figure}

Here, we experiment with models from the \texttt{Qwen} family using pairwise comparisons between their outputs by a strong LLM, when prompted with open-ended questions from the LMSYS Chatbot Arena platform, following the setup described in Section~\ref{sec:pairwise}. 
Table~\ref{tab:lmsys-ranking-qwen} and Figure~\ref{fig:lmsys-all-qwen} summarizes the results, which are qualitatively similar to those in Section~\ref{sec:pairwise}.

\begin{table}[h]
\centering
    \begin{tabular}{l c c c c}
    \toprule & \multicolumn{2}{c}{Coupled} &  \multicolumn{2}{c}{Independent} \\
    \cmidrule(r{1mm}){2-3} \cmidrule(l{1mm}){4-5}
    {LLM} & Rank & Avg. win-rate 
    & Rank & Avg. win-rate \\ 
    \midrule \texttt{3-8B} & 1 & 0.4486 $\pm$ 0.0013 & 1 & 0.4524 $\pm$ 0.0013 \\ \texttt{2.5-7B-bnb-8bit} & 2 & 0.3211 $\pm$ 0.0012 & 2 & 0.3264 $\pm$ 0.0012 \\ \texttt{2.5-7B} & 3 & 0.3148 $\pm$ 0.0012 & 2 & 0.3279 $\pm$ 0.0012 \\ \texttt{2.5-7B-bnb-4bit} & 4 & 0.2892 $\pm$ 0.0012 & 4 & 0.2862 $\pm$ 0.0012 \\ \texttt{2.5-7B-AWQ-INT4} & 5 & 0.2561 $\pm$ 0.0011 & 5 & 0.2591 $\pm$ 0.0012 \\ \texttt{2.5-3B} & 6 & 0.2072 $\pm$ 0.0011 & 6 & 0.2121 $\pm$ 0.0011 \\ \texttt{2.5-1.5B} & 7 & 0.1834 $\pm$ 0.0010 & 7 & 0.1875 $\pm$ 0.0010 \\ \texttt{2.5-3B-distil} & 8 & 0.1780 $\pm$ 0.0010 & 8 & 0.1850 $\pm$ 0.0010 \\ \bottomrule \end{tabular}
    \caption{Average win-rate of each LLM across all other LLMs in the \texttt{Qwen} family ($\pm$ $95\%$ confidence intervals). To derive the rankings, for each LLM, we choose the lowest ranking provided by the method of~\citet{chatzi2024prediction}.}
    \label{tab:lmsys-ranking-qwen}

\end{table}

\clearpage
\newpage

\section{EXPERIMENTS WITH LLMS IN THE \texttt{Mistral} FAMILY}
\label{app:mistral}
In this section, we experiment with LLMs in from the \texttt{Mistral} family.
For brevity, we shorten the names of the LLMs, as listed in Table~\ref{tab:mistral-names}.

\begin{table}[h]
    \centering
    \begin{tabular}{ll} \toprule
         Full name & Shortened name \\ \midrule
         \texttt{Mistral-7B-Instruct-v0.3} & \texttt{v0.3}\\
         \texttt{Mistral-7B-Instruct-v0.3-bnb-4bit} & \texttt{v0.3-bnb-4bit} \\ 
         \texttt{Mistral-7B-Instruct-v0.3-bnb-8bit} &  \texttt{v0.3-bnb-8bit}\\ 
         \texttt{Mistral-7B-Instruct-v0.2} & \texttt{v0.2}\\
         \texttt{Mistral-7B-Instruct-v0.1}& \texttt{v0.1} \\ 
         \bottomrule
    \end{tabular}
    \caption{Full and shortened names of LLMs in the \texttt{Mistral} family.}
    \label{tab:mistral-names}
\end{table}

\subsection{MMLU Dataset}
\begin{figure}[!!h]
\centering
\begin{tabular}{c c c}
     \multicolumn{3}{c}{\texttt{v0.3} vs. \texttt{v0.3-bnb-8bit}}\\
    \includegraphics[width=0.23\linewidth]{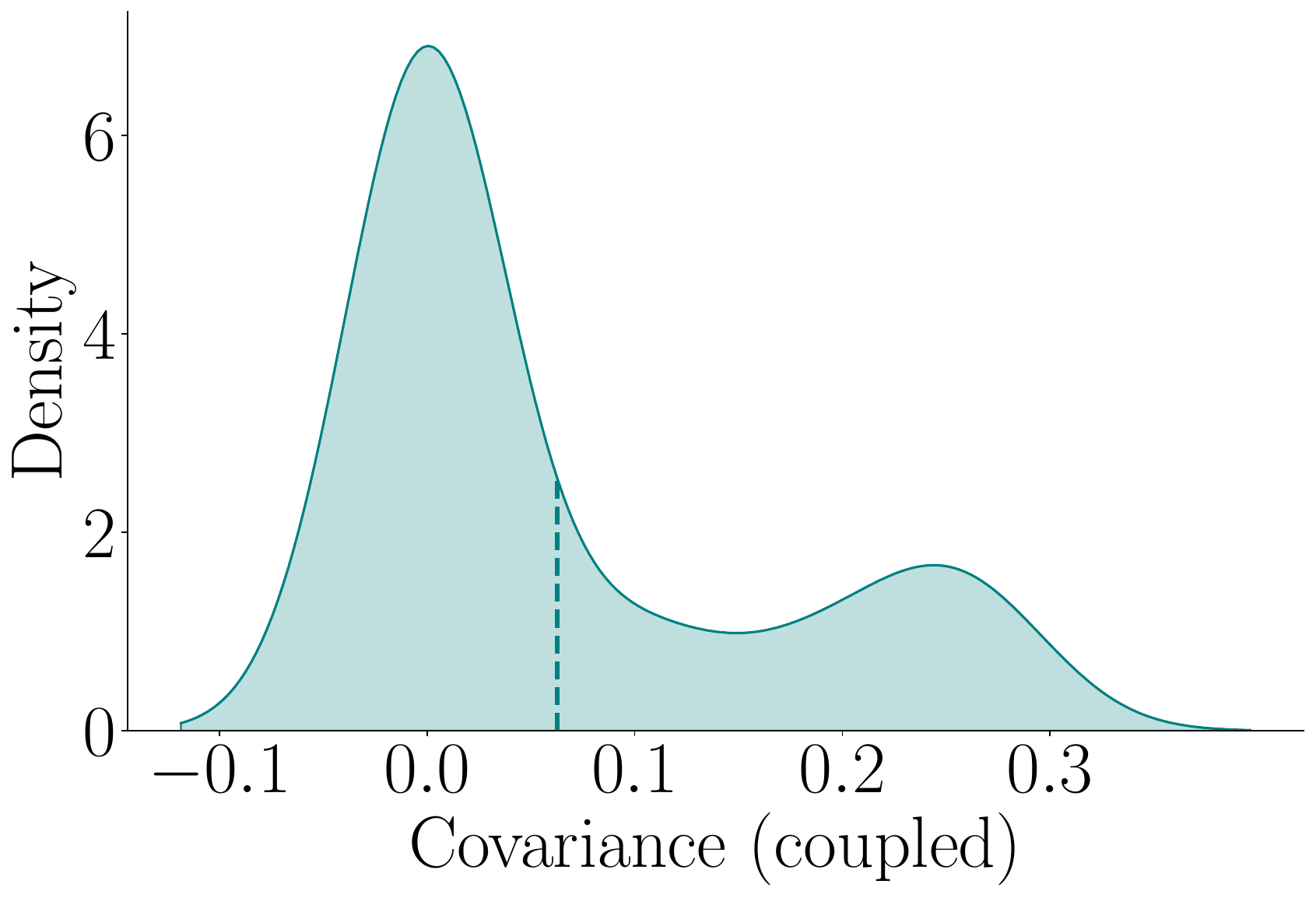} &
    \includegraphics[width=0.23\linewidth]{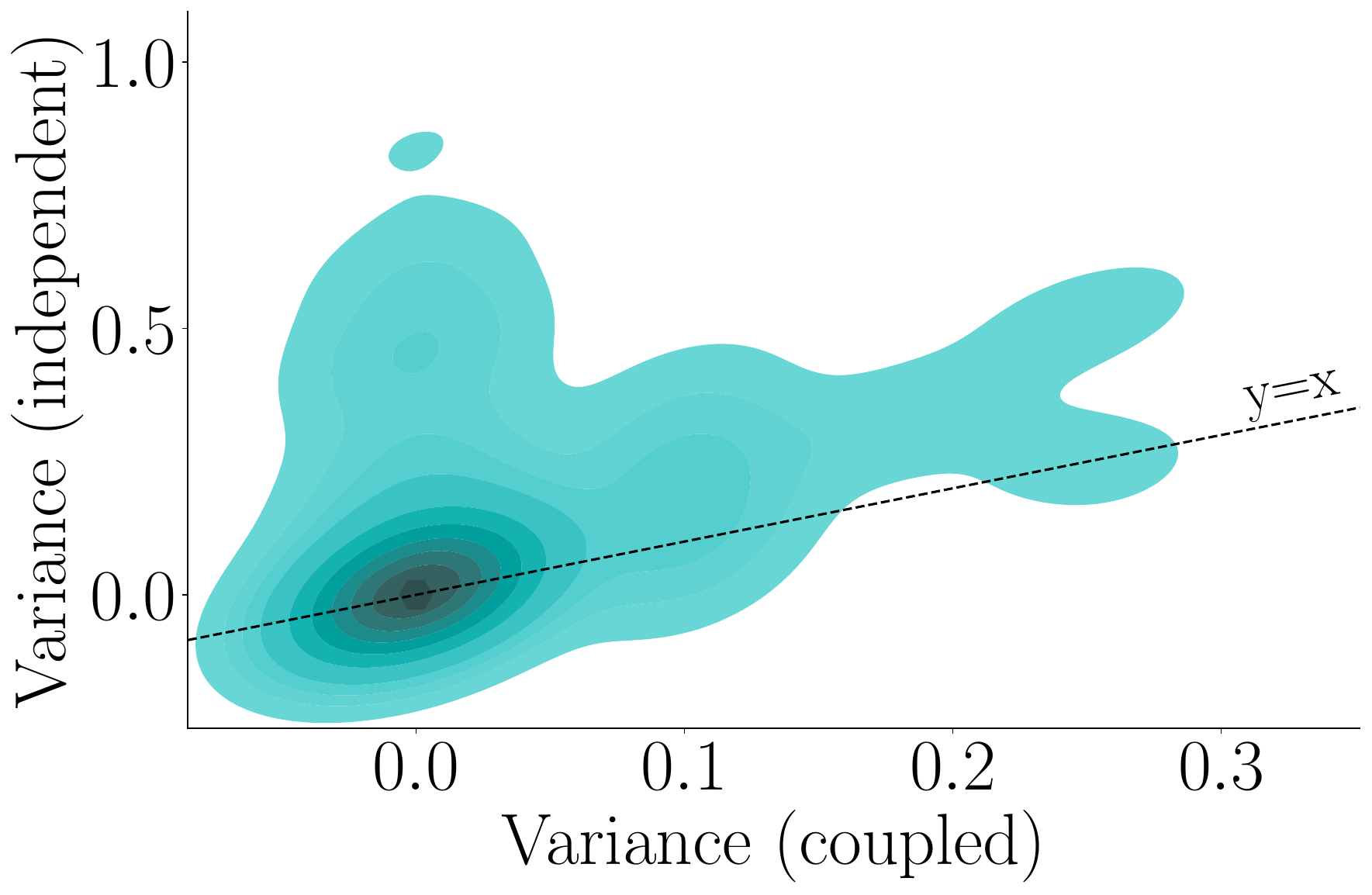} &
    \includegraphics[width=0.23\linewidth]{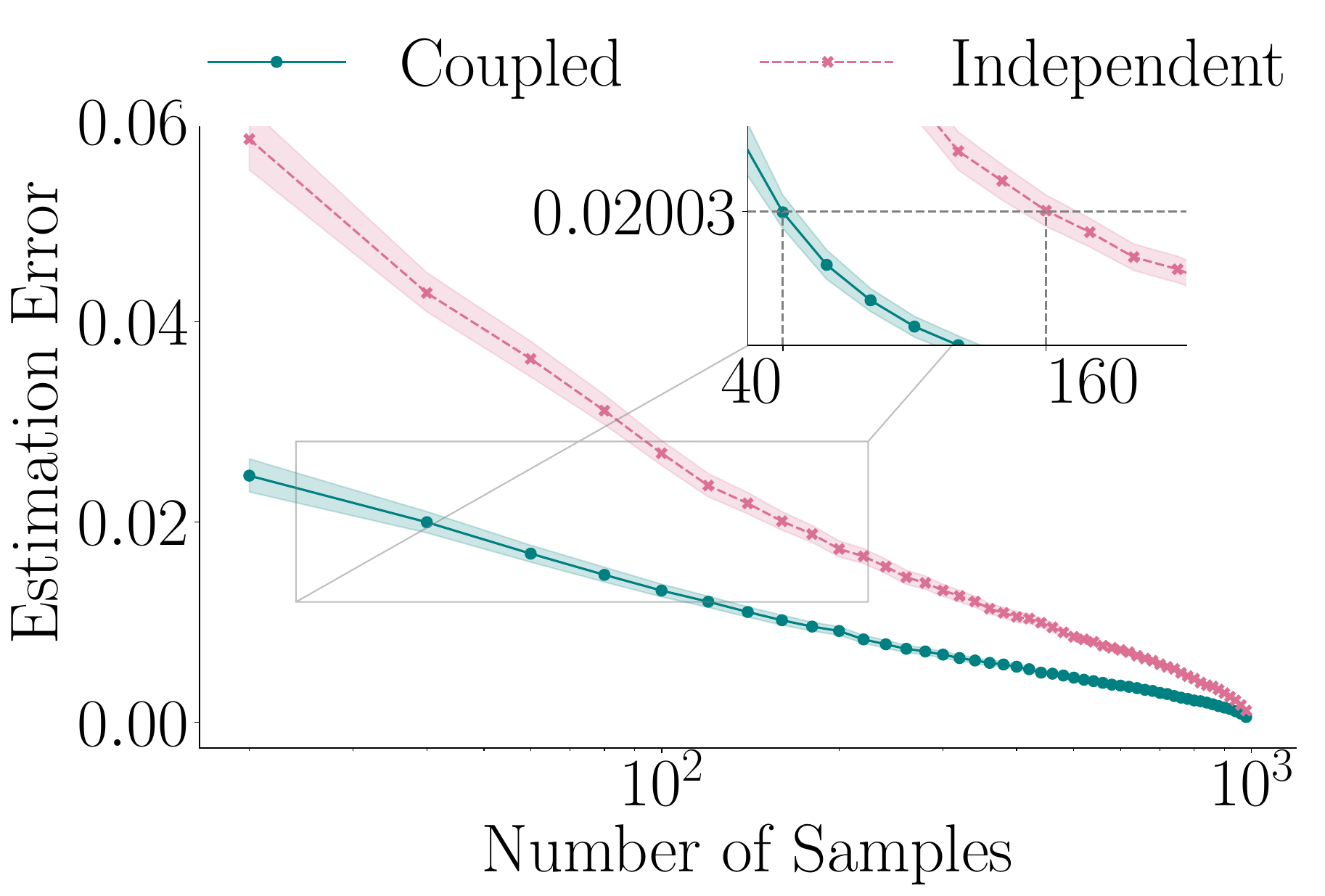} \\ \\
%
    \multicolumn{3}{c}{\texttt{v0.3} vs. \texttt{v0.3-bnb-4bit}}\\
    \includegraphics[width=0.23\linewidth]{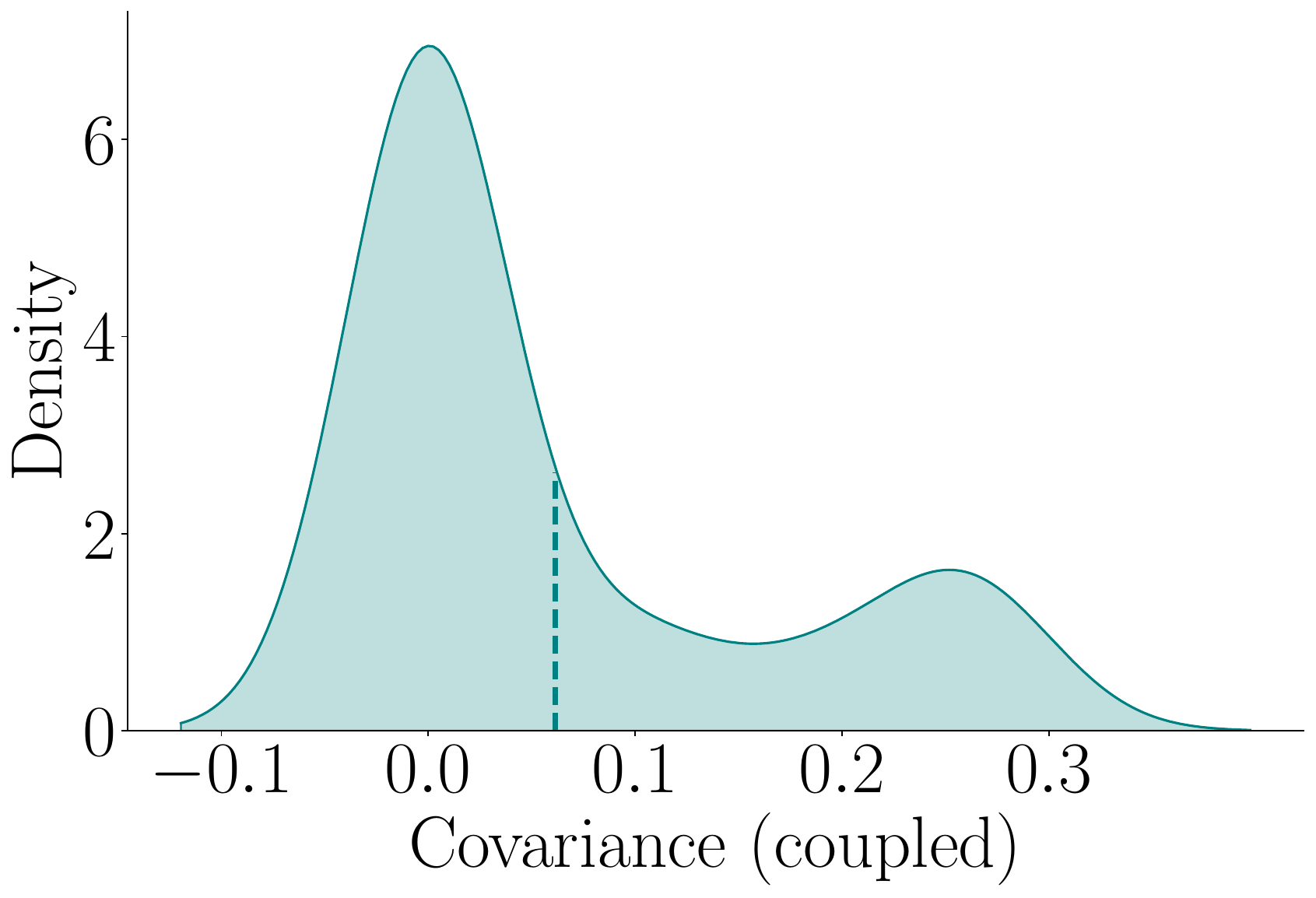} &
    \includegraphics[width=0.23\linewidth]{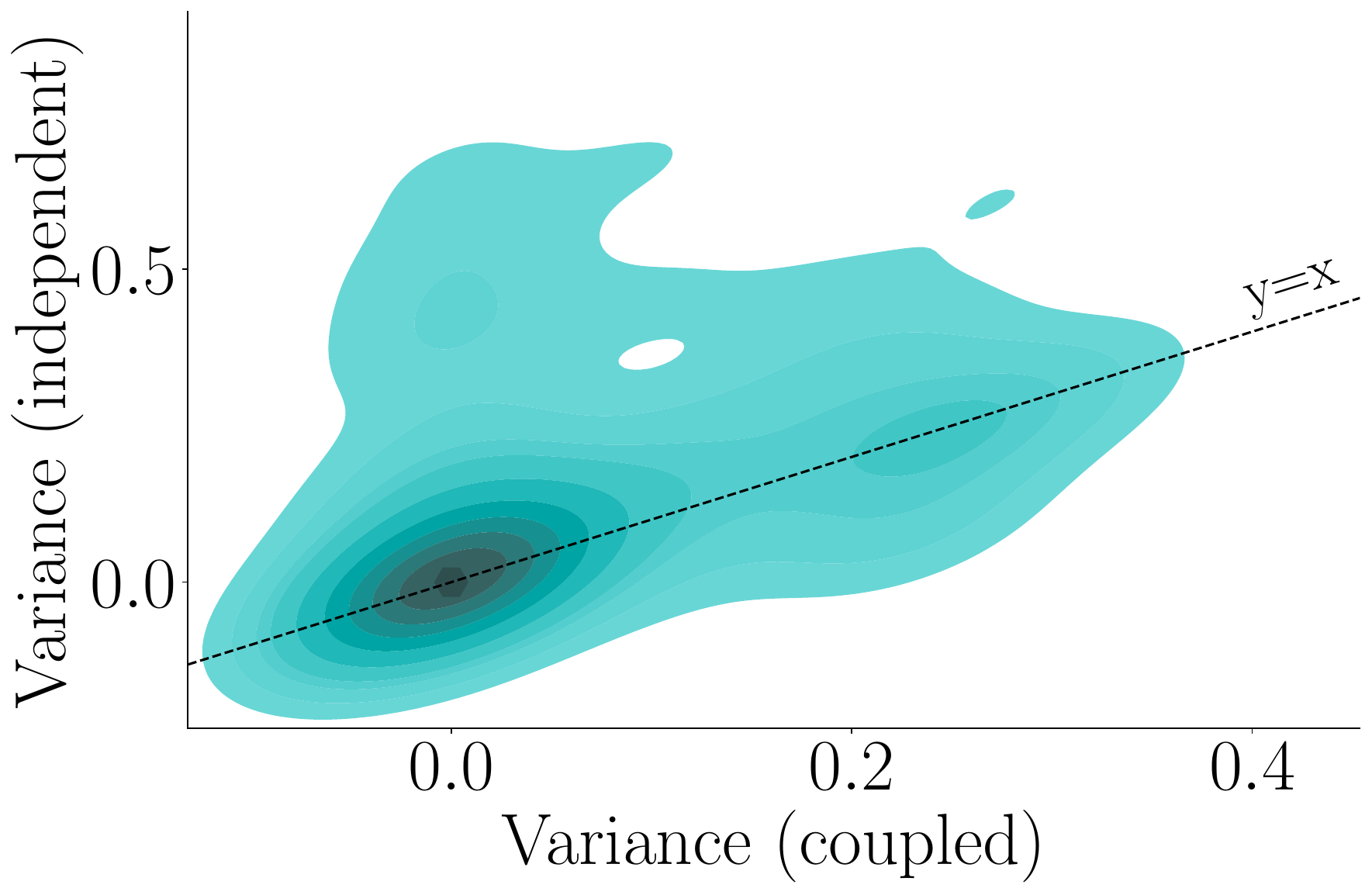} &
    \includegraphics[width=0.23\linewidth]{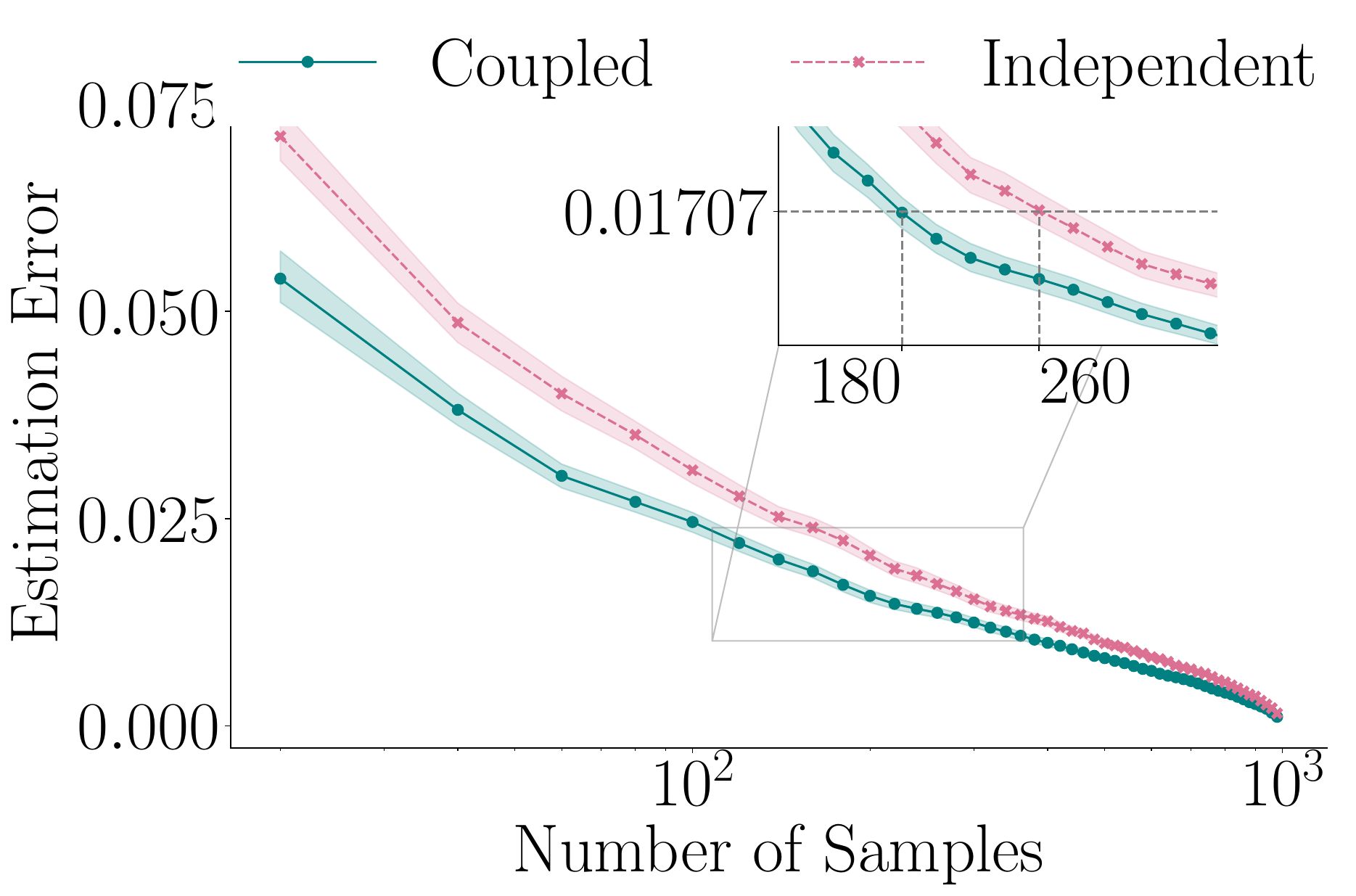} \\ \\
%
     \multicolumn{3}{c}{\texttt{v0.3-bnb-8bit} vs. \texttt{v0.3-bnb-4bit}}\\
    \includegraphics[width=0.23\linewidth]{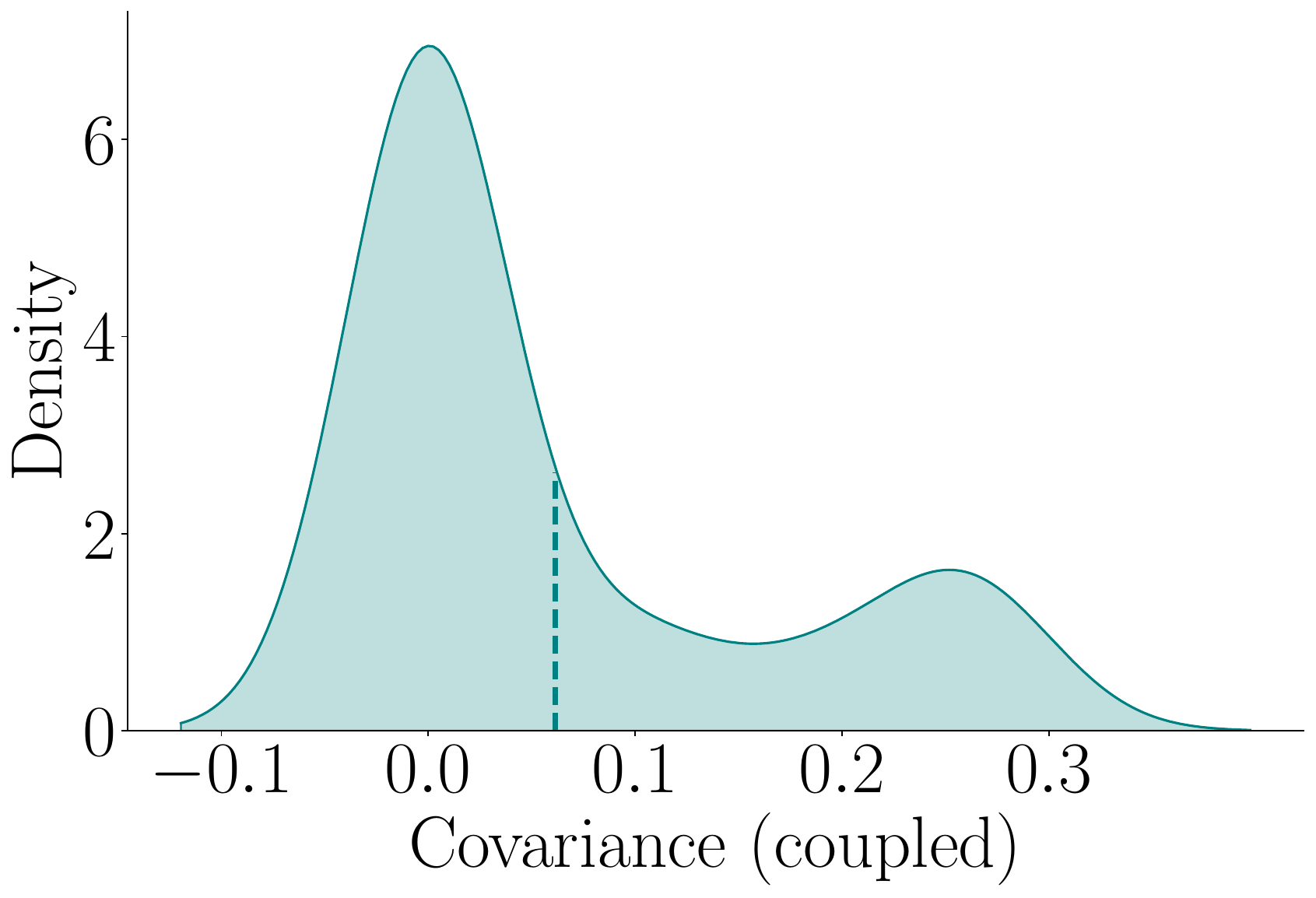} &
    \includegraphics[width=0.23\linewidth]{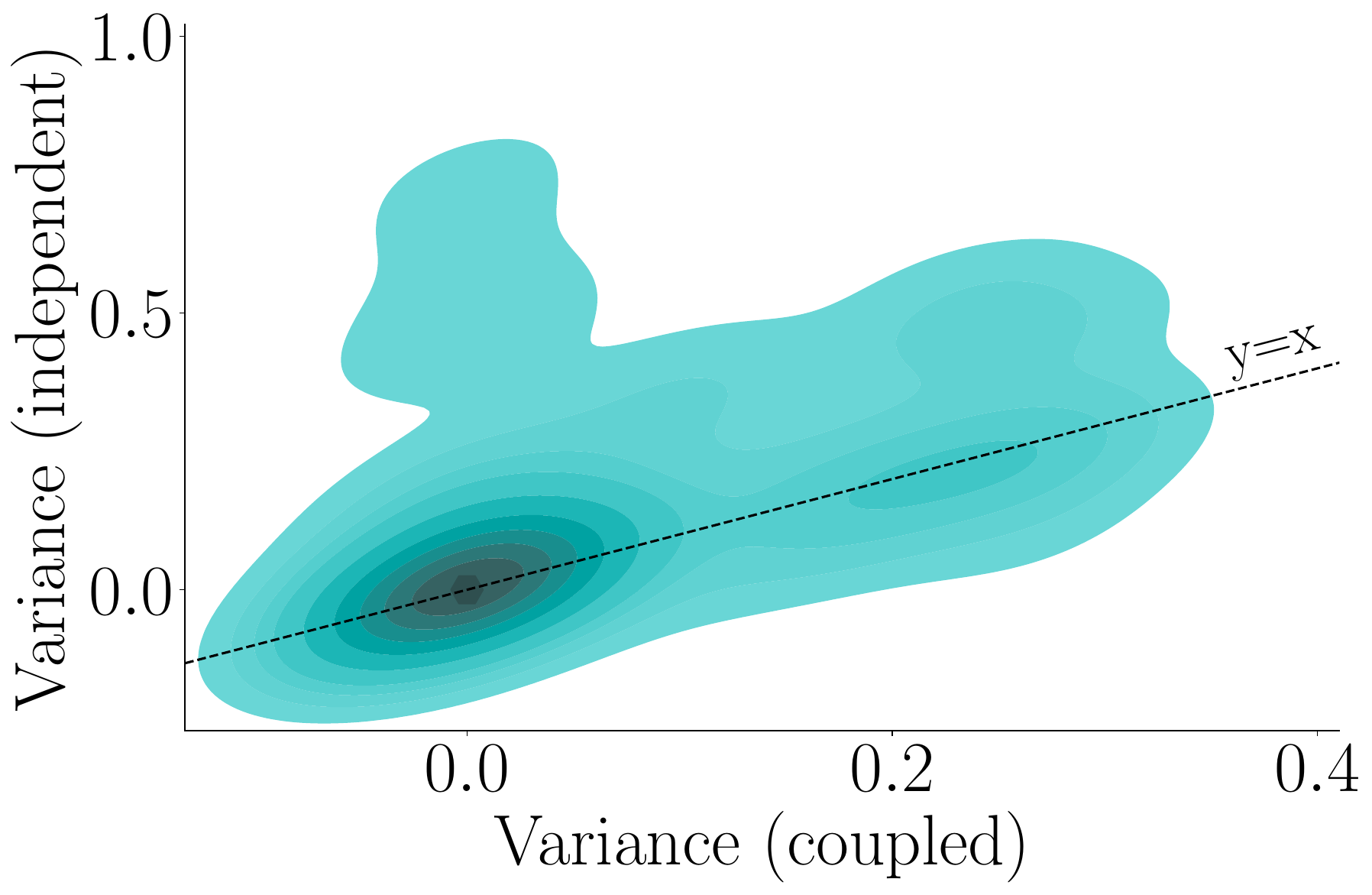} &
    \includegraphics[width=0.23\linewidth]{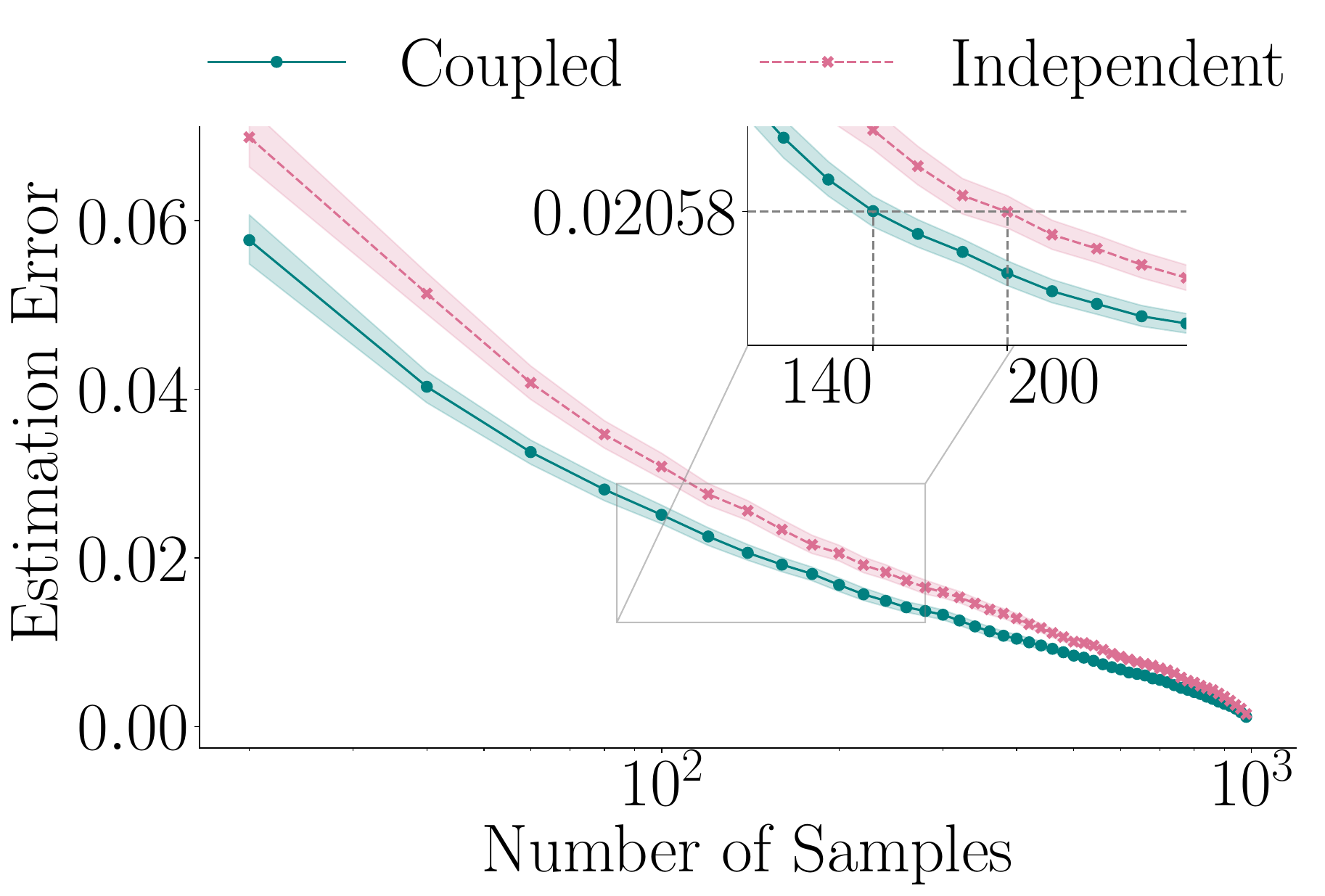} \\ \\
%
  \multicolumn{3}{c}{\texttt{v0.3} vs. \texttt{v0.2}}\\
    \includegraphics[width=0.23\linewidth]{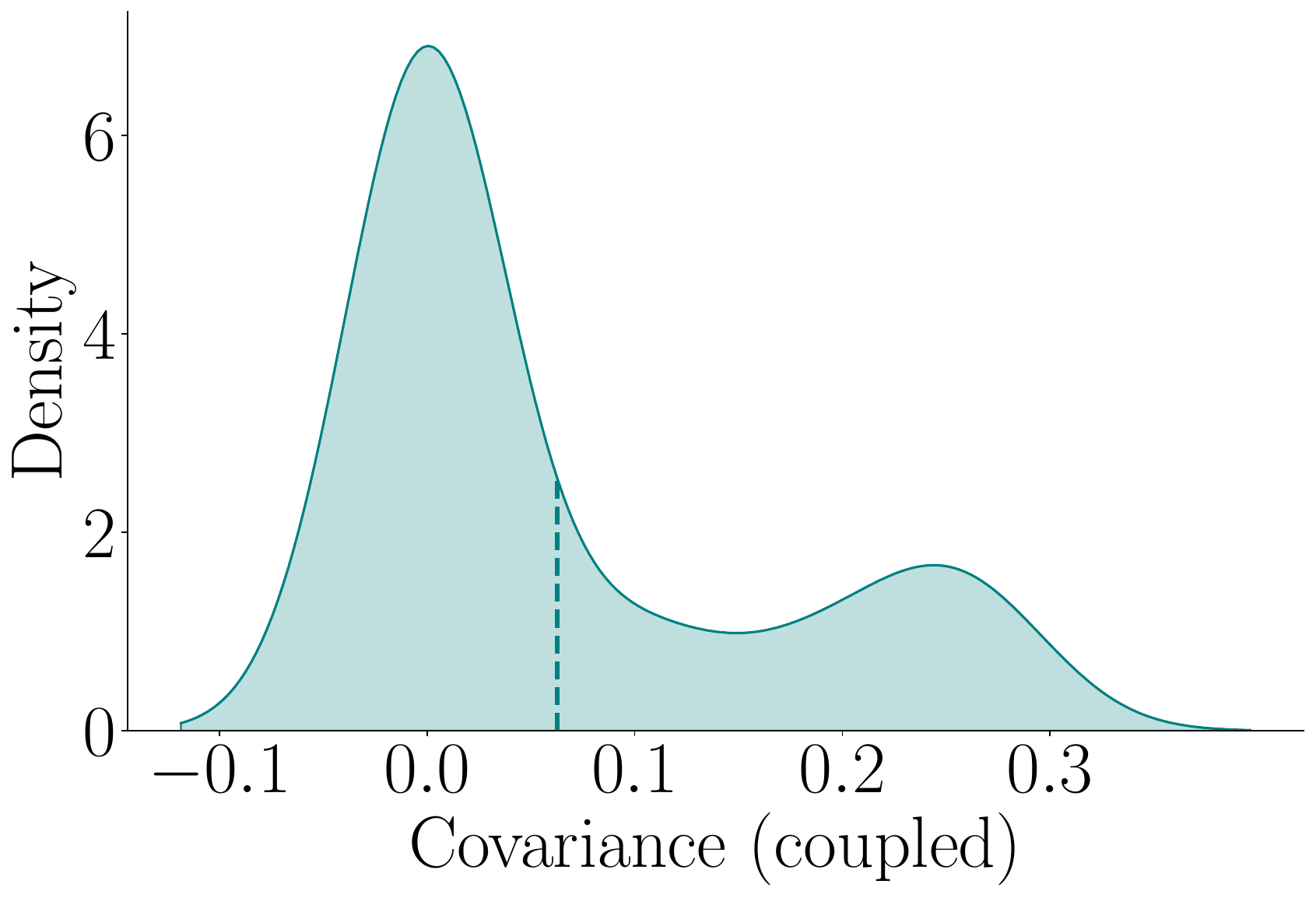} &
    \includegraphics[width=0.23\linewidth]{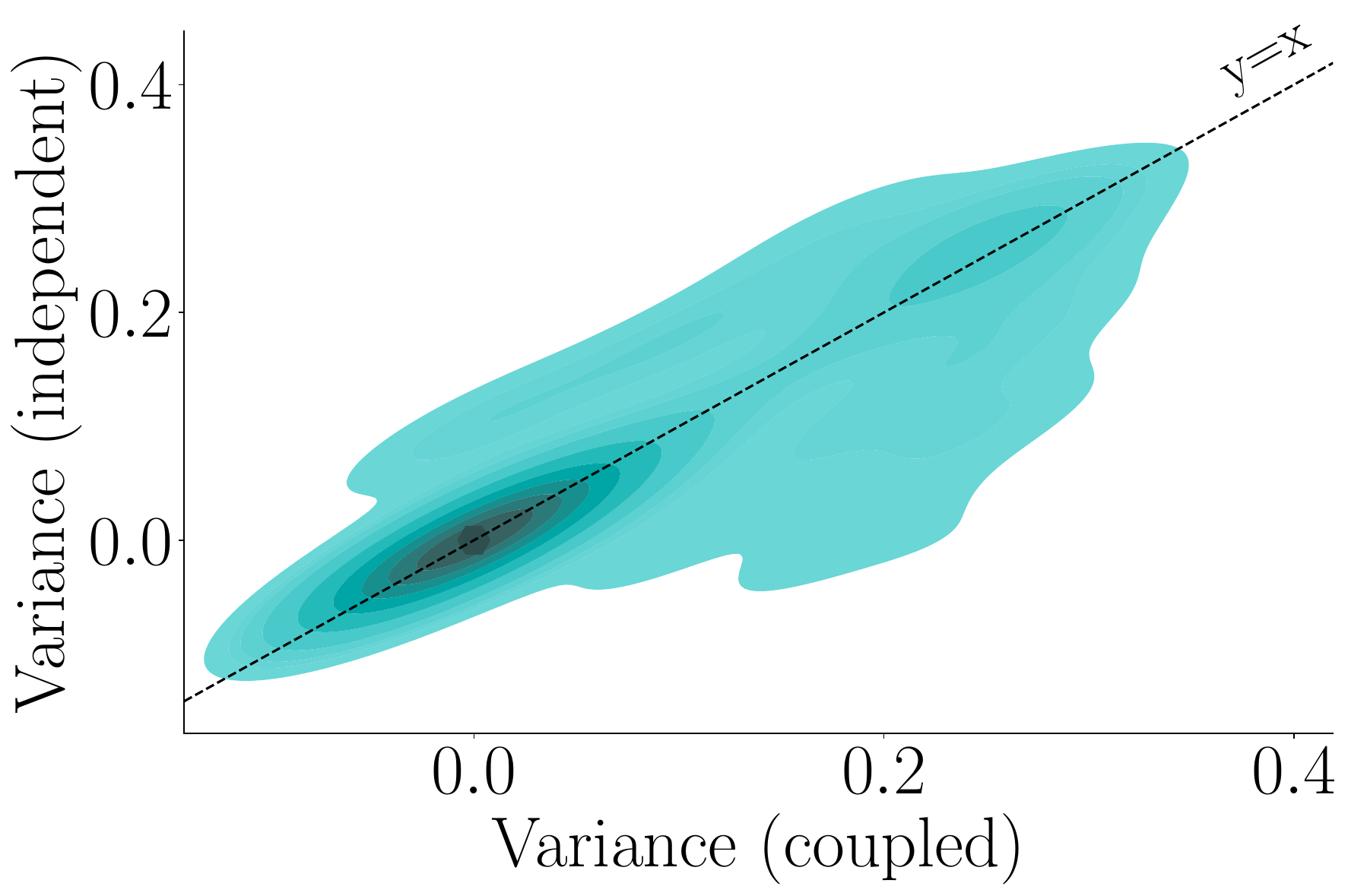} &
    \includegraphics[width=0.23\linewidth]{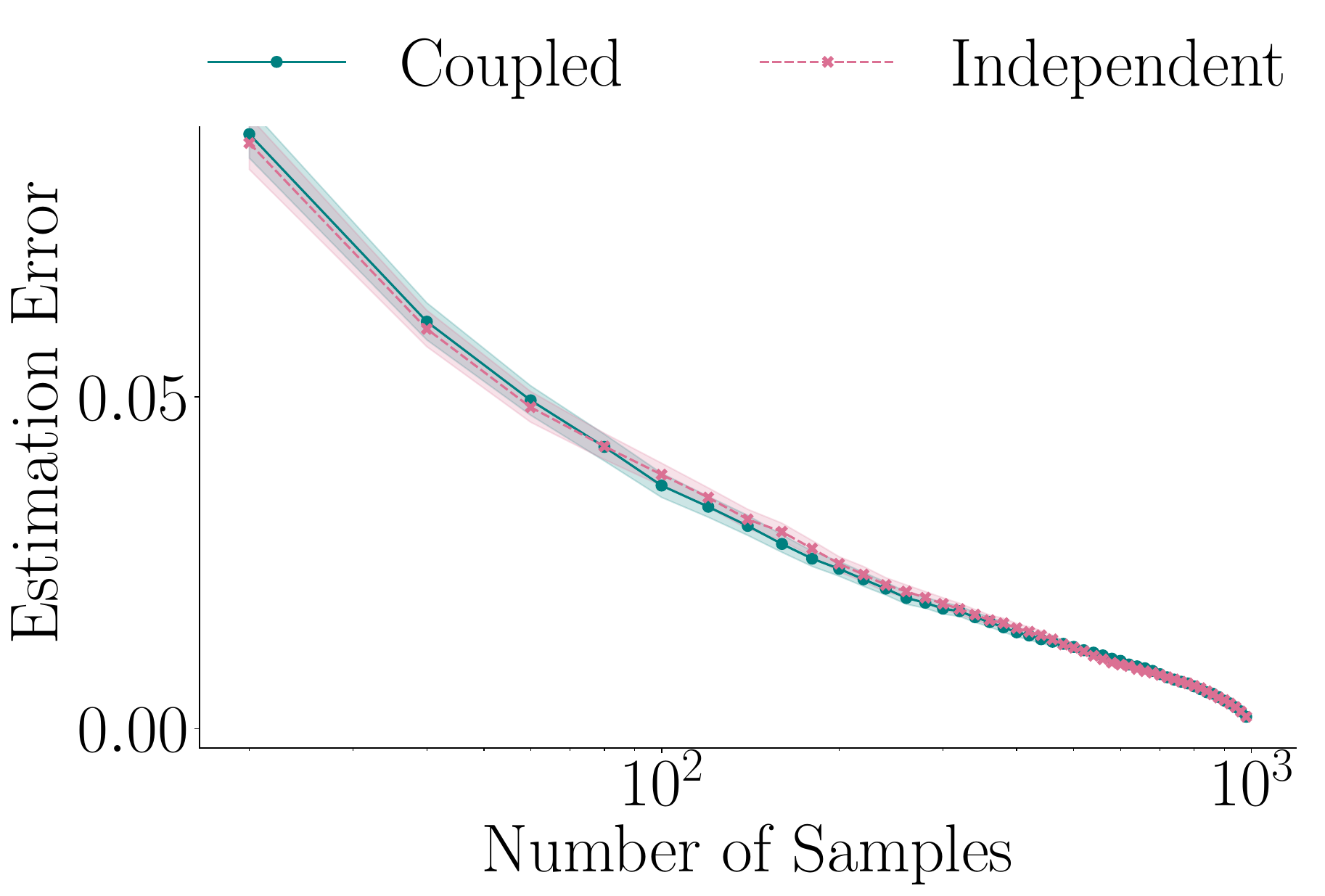} \\ \\

 \multicolumn{3}{c}{\texttt{v0.3} vs. \texttt{v0.1}}\\
    \includegraphics[width=0.23\linewidth]{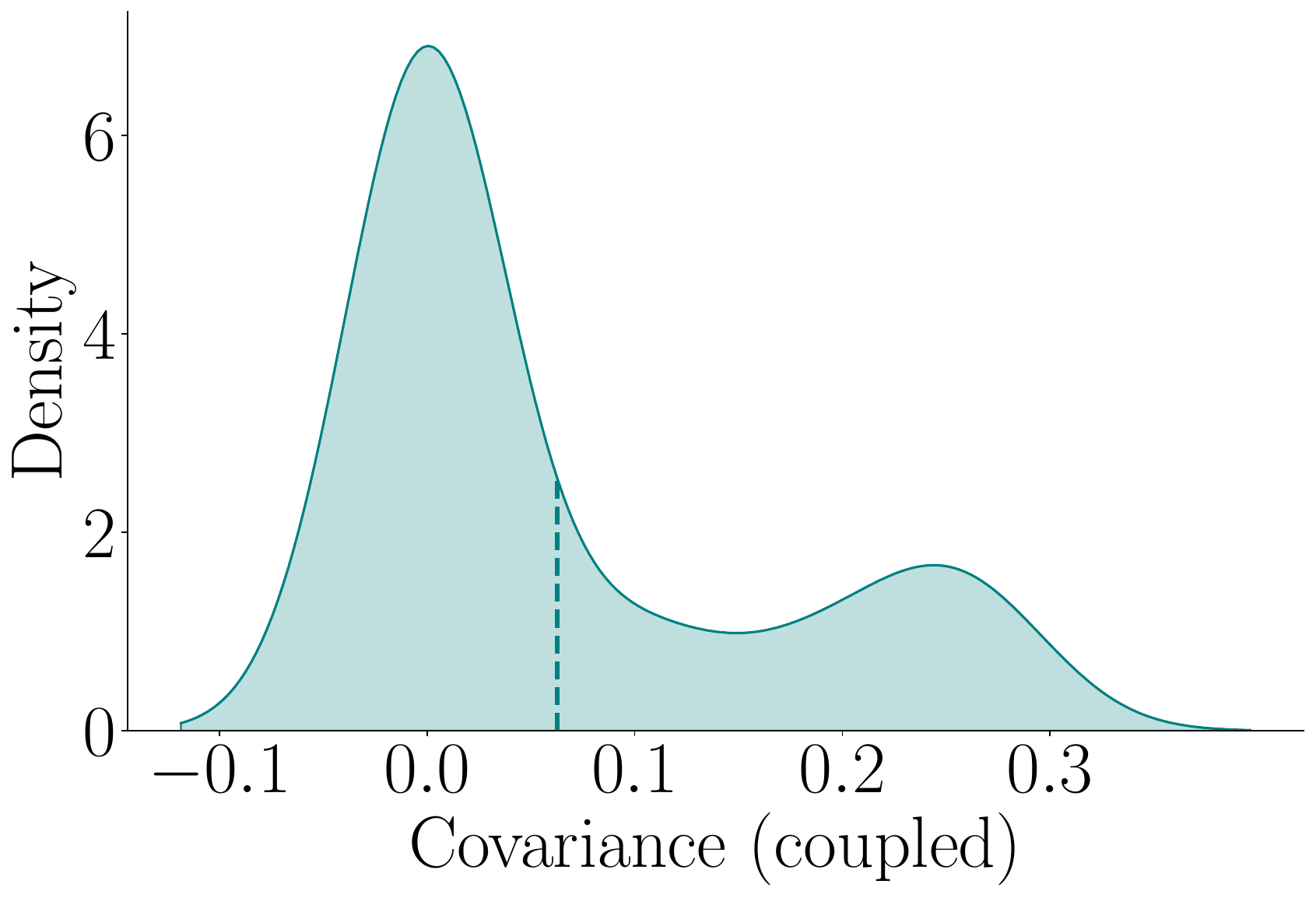} &
    \includegraphics[width=0.23\linewidth]{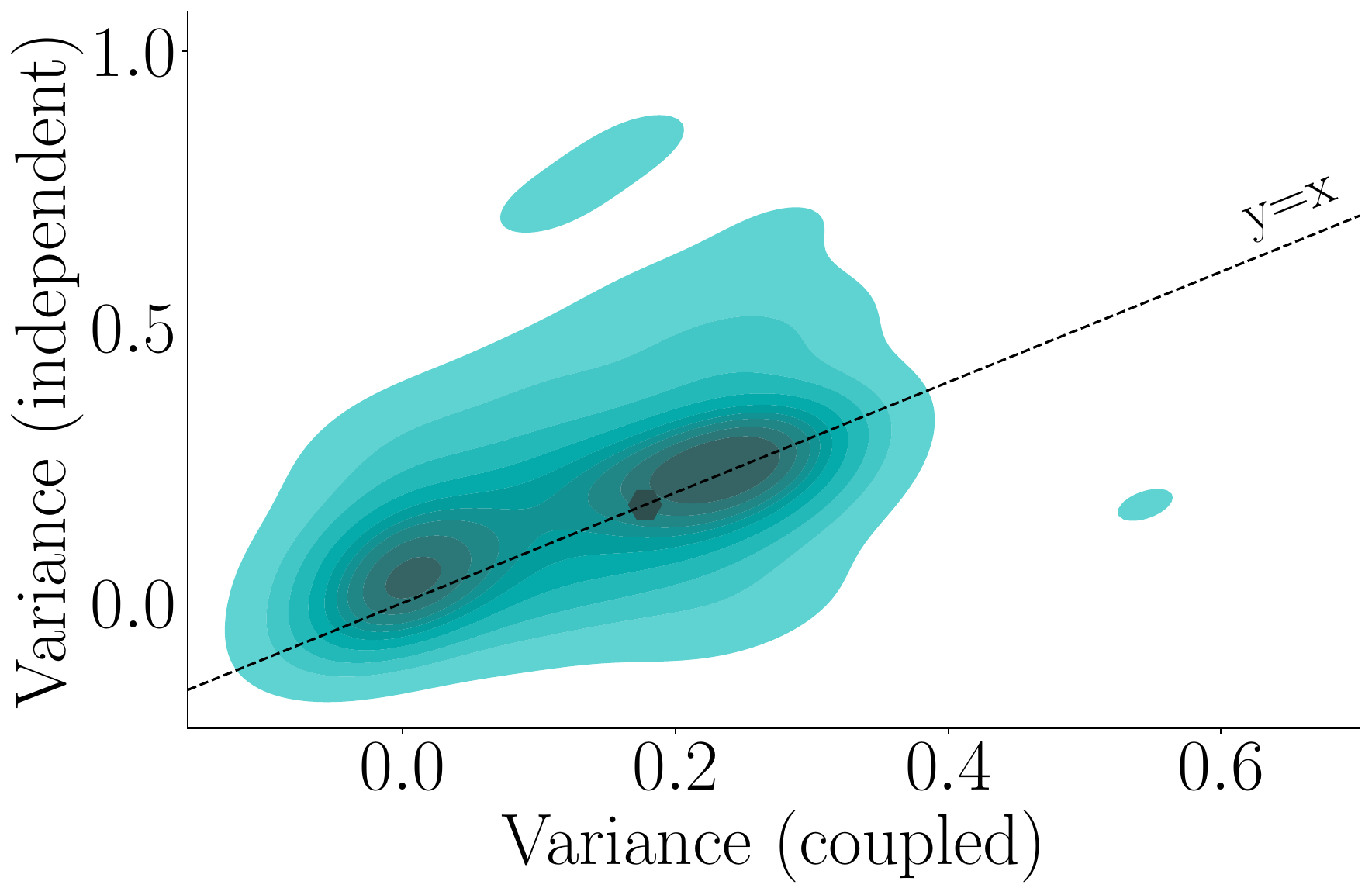} &
    \includegraphics[width=0.23\linewidth]{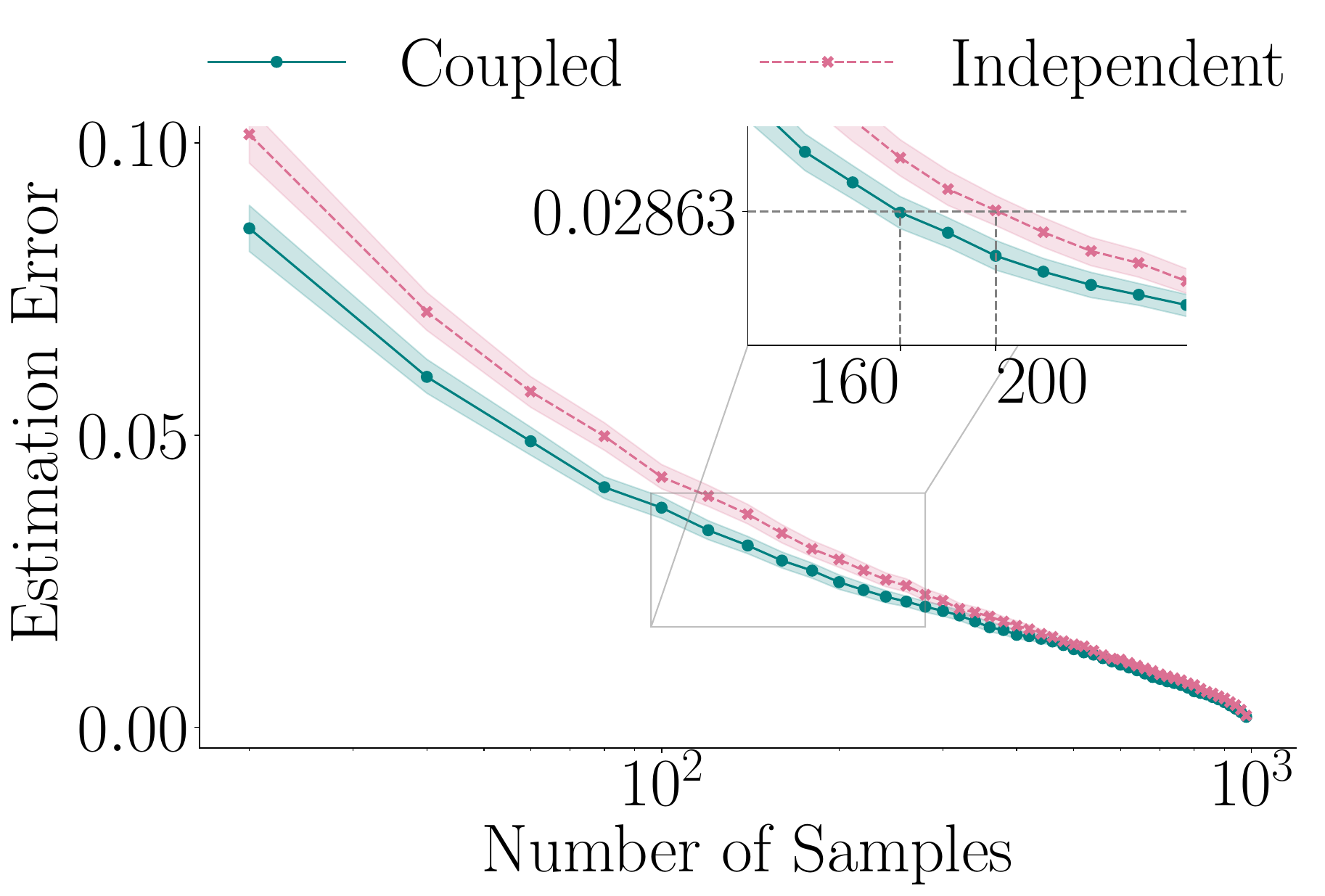} \\ \\
    (a) Score covariance & (b) Variance of the score difference & (c) Estimation error vs. \# samples \\ 
\end{tabular}
    \caption{\textbf{Comparison between five pairs of LLMs in the \texttt{Mistral} family on multiple-choice questions from the ``college computer science'' knowledge area of the MMLU dataset.}
    Panels in column (a) show the kernel density estimate (KDE) of the covariance between the scores of the two LLMs on each question under coupled generation; the dashed lines correspond to average values. Panels in column (b) show the KDE of the variance of the difference between the scores of the LLMs on each question under coupled and independent generation; the highlighted points correspond to median values. Panels in column (c) show the absolute error in the estimation of the expected difference between the scores of the LLMs against the number of samples; for each point on the x-axis, we perform $1{,}000$ sub-samplings and shaded areas correspond to $95\%$ confidence intervals.}
    \label{fig:mmlu-mistral-first}
\end{figure}

\begin{figure}[h]
\centering
\begin{tabular}{c c c}
    \multicolumn{3}{c}{\texttt{v0.3-bnb-8bit} vs. \texttt{v0.2}}\\
    \includegraphics[width=0.23\linewidth]{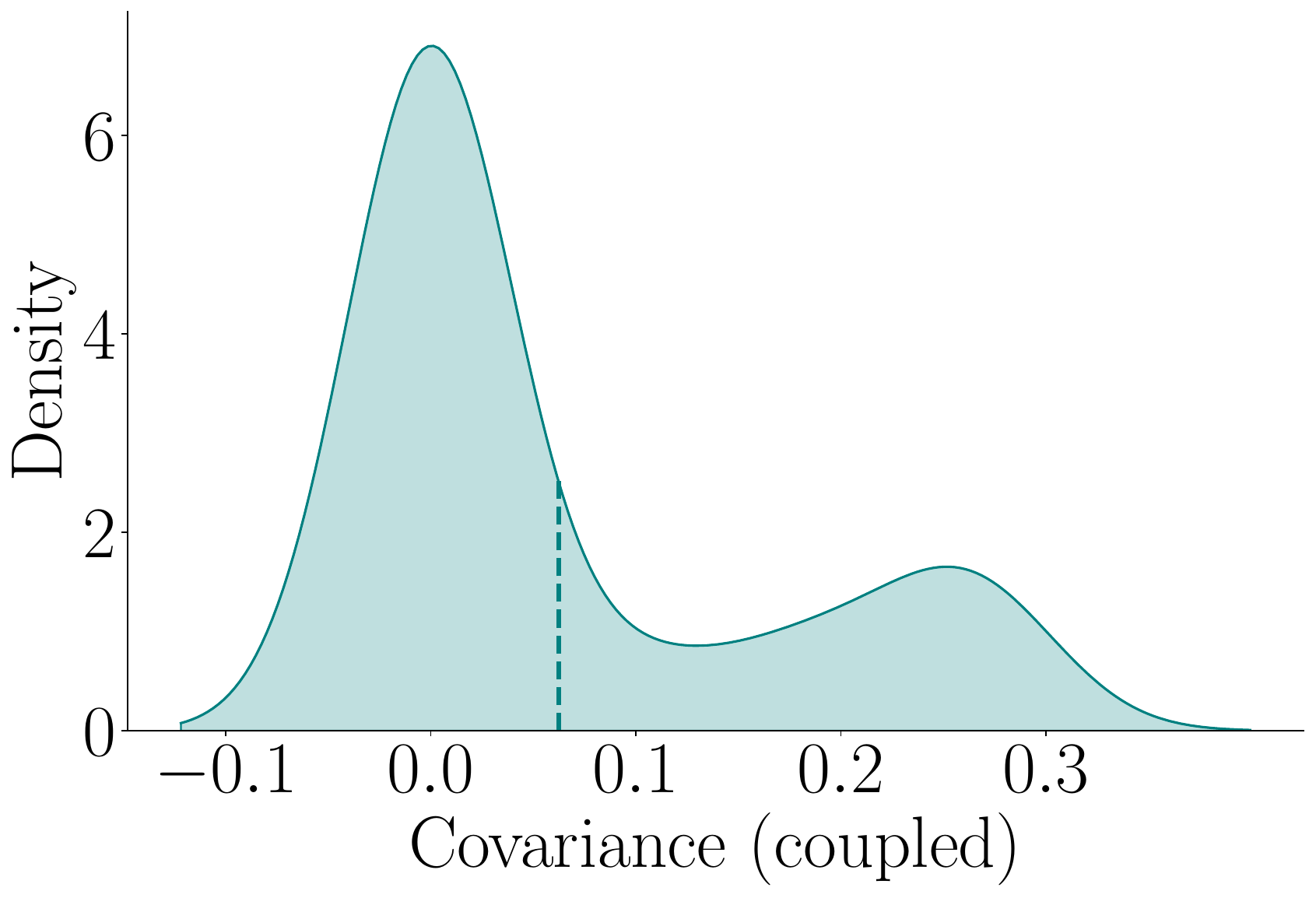} &
    \includegraphics[width=0.23\linewidth]{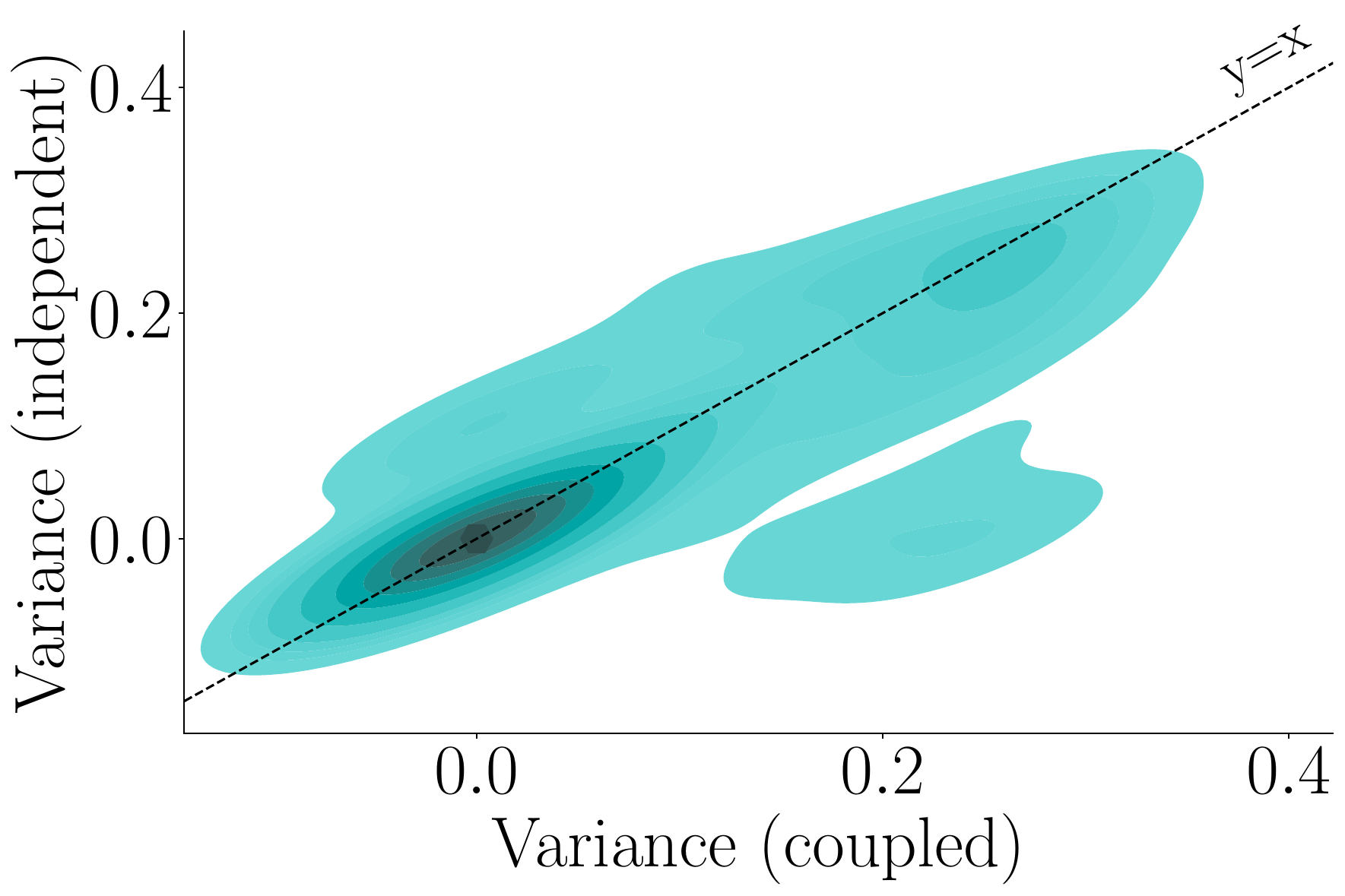} &
    \includegraphics[width=0.23\linewidth]{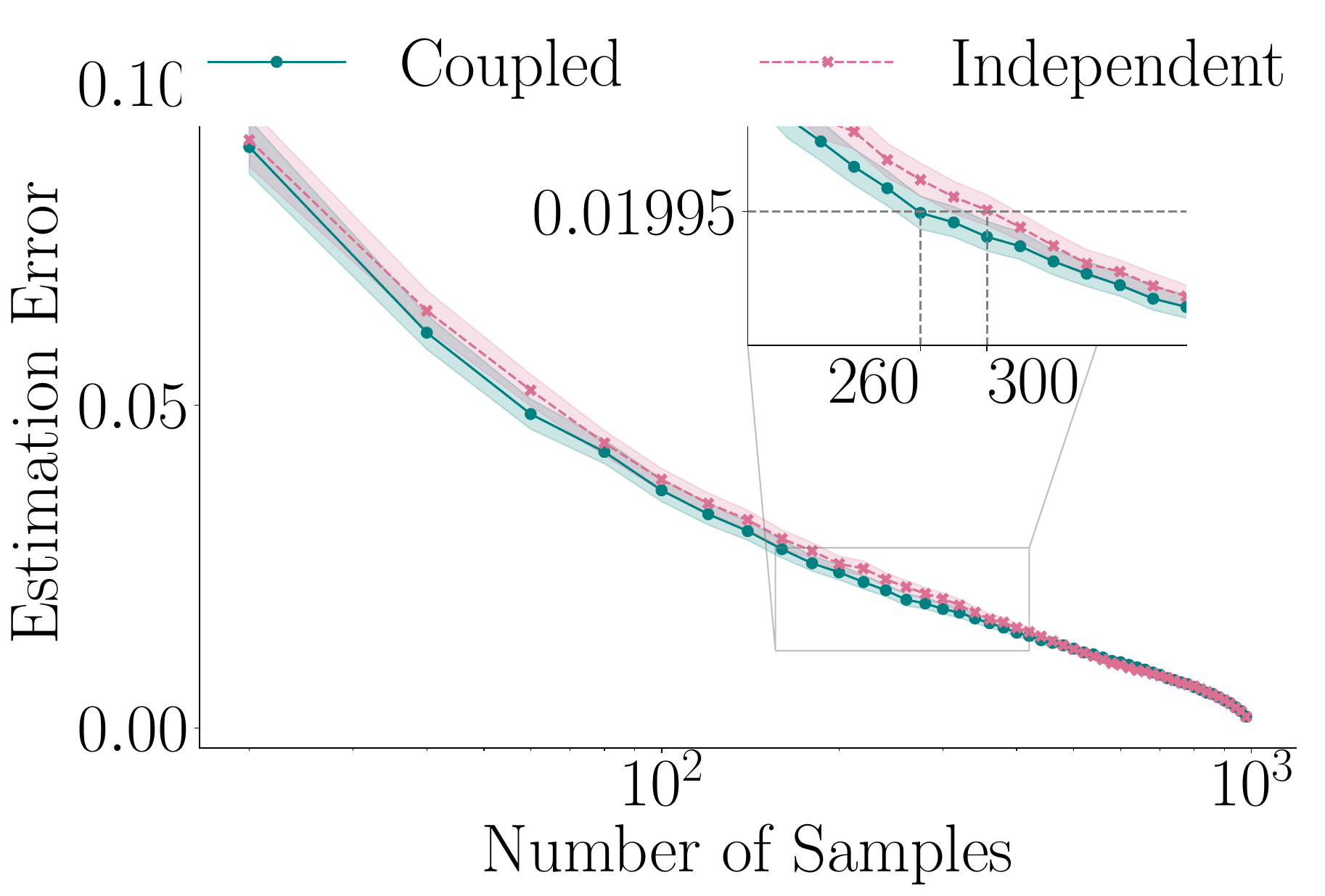} \\ \\
    \multicolumn{3}{c}{\texttt{v0.3-bnb-8bit} vs. \texttt{v0.1}}\\
    \includegraphics[width=0.23\linewidth]{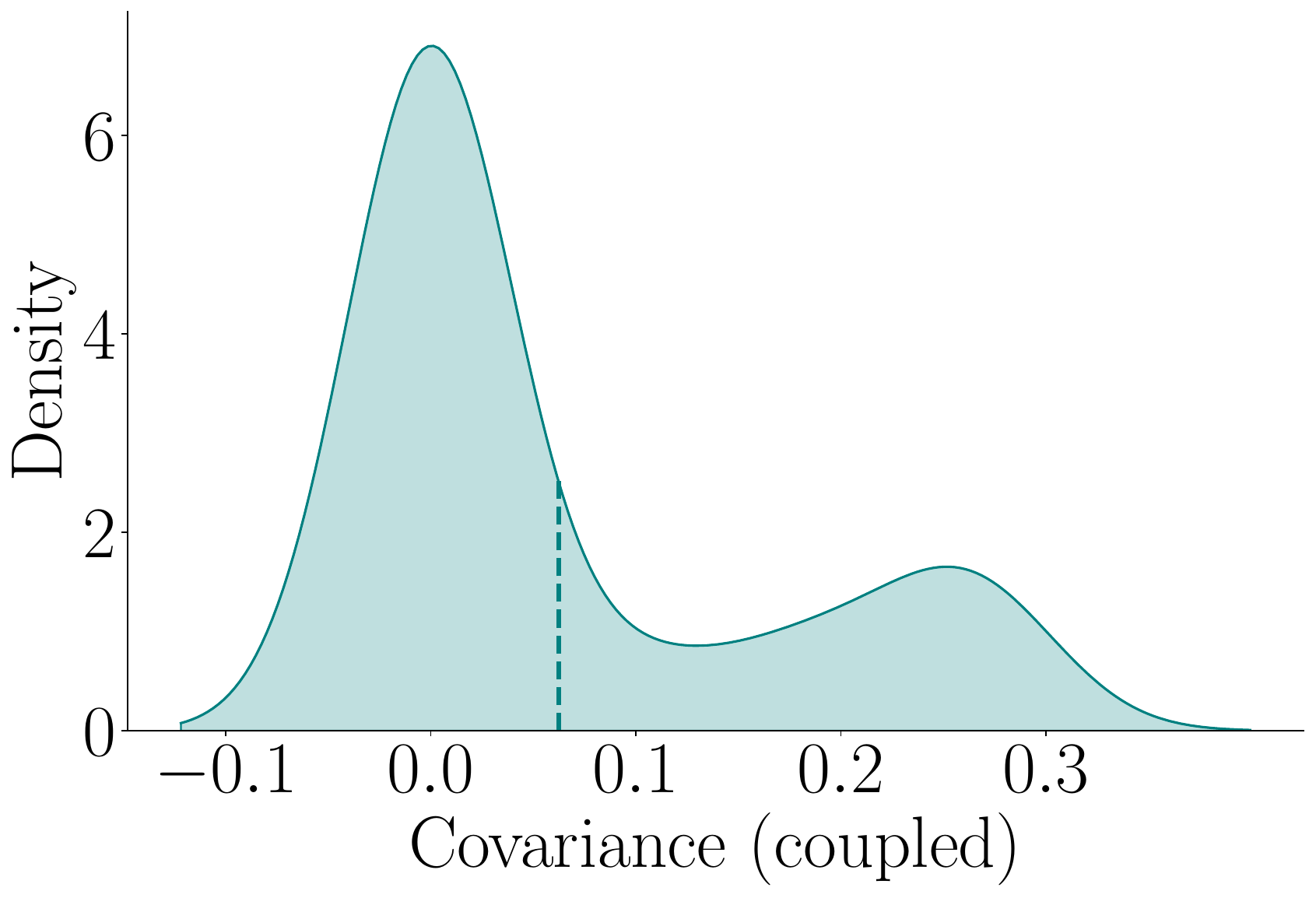} &
    \includegraphics[width=0.23\linewidth]{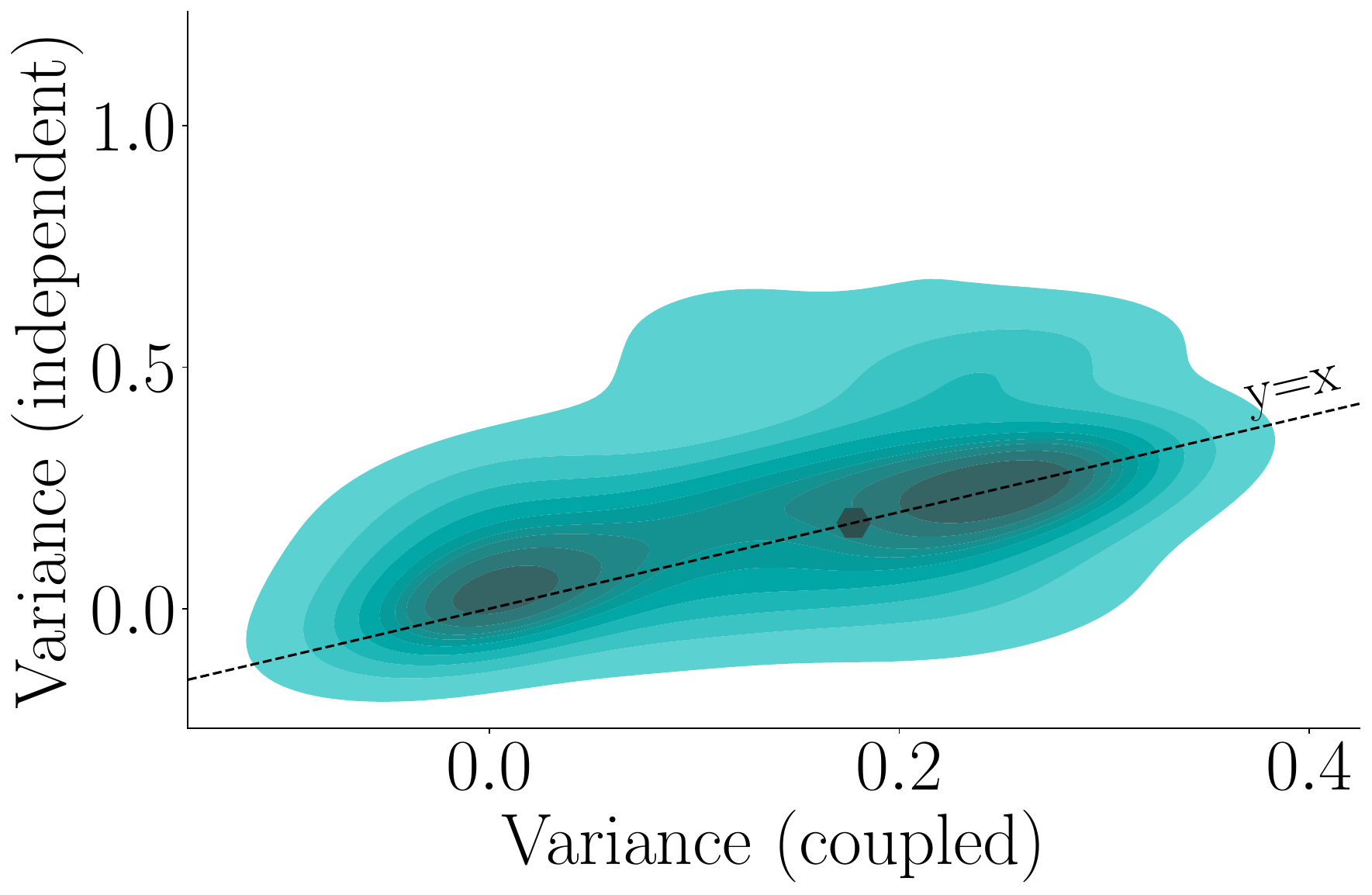} &
    \includegraphics[width=0.23\linewidth]{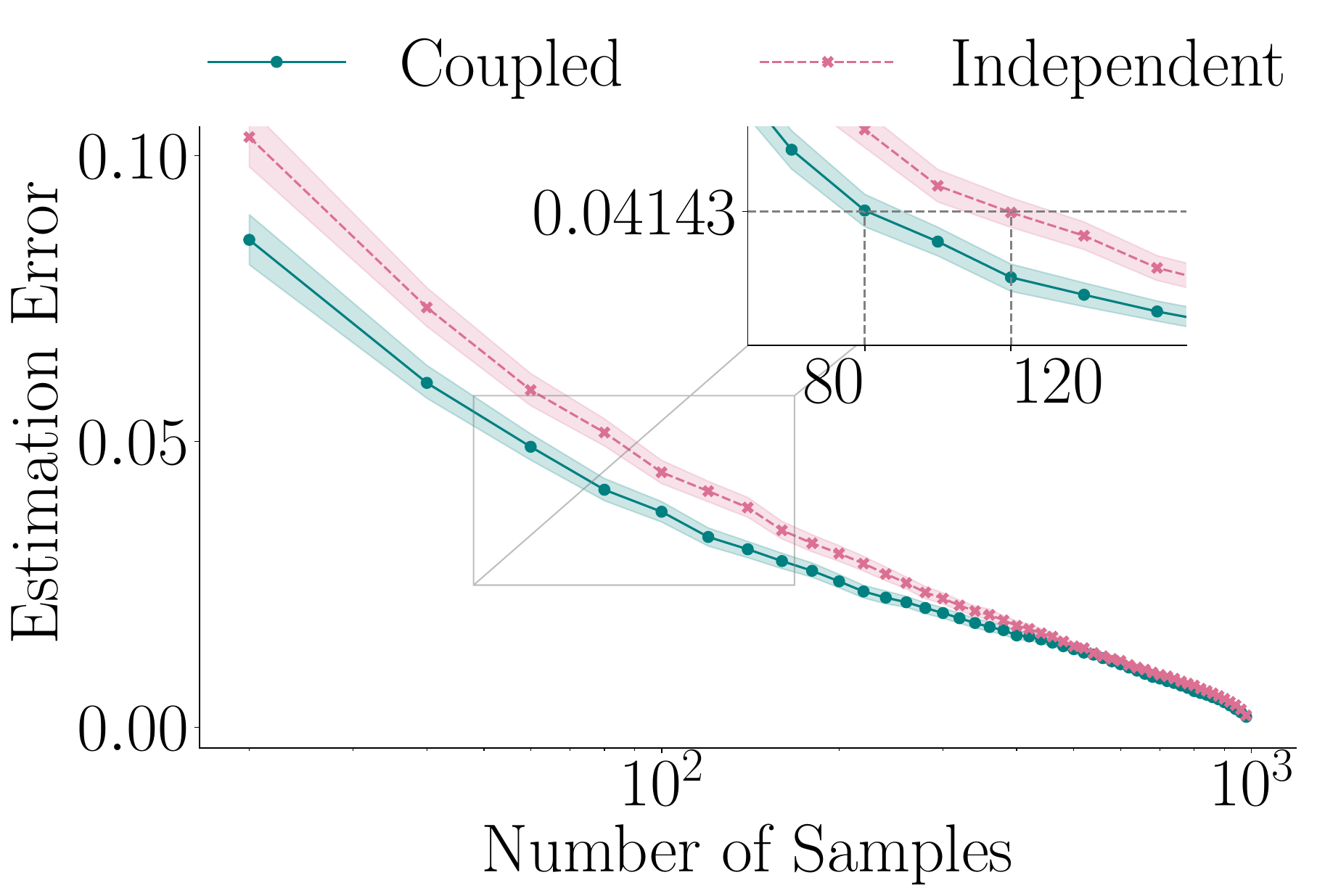} \\ \\
     \multicolumn{3}{c}{\texttt{v0.1} vs. \texttt{v0.3-bnb-4bit}}\\
    \includegraphics[width=0.23\linewidth]{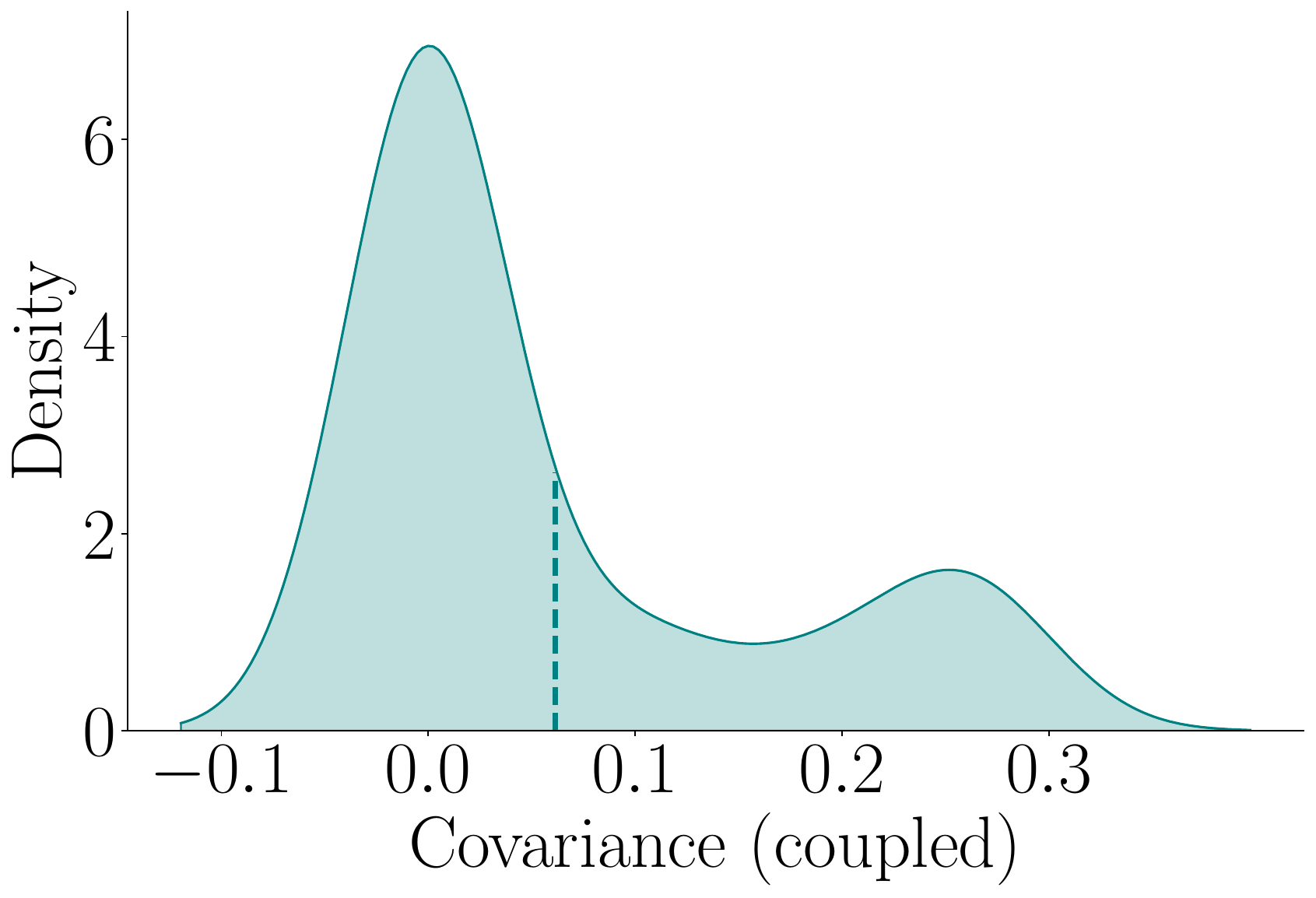} &
    \includegraphics[width=0.23\linewidth]{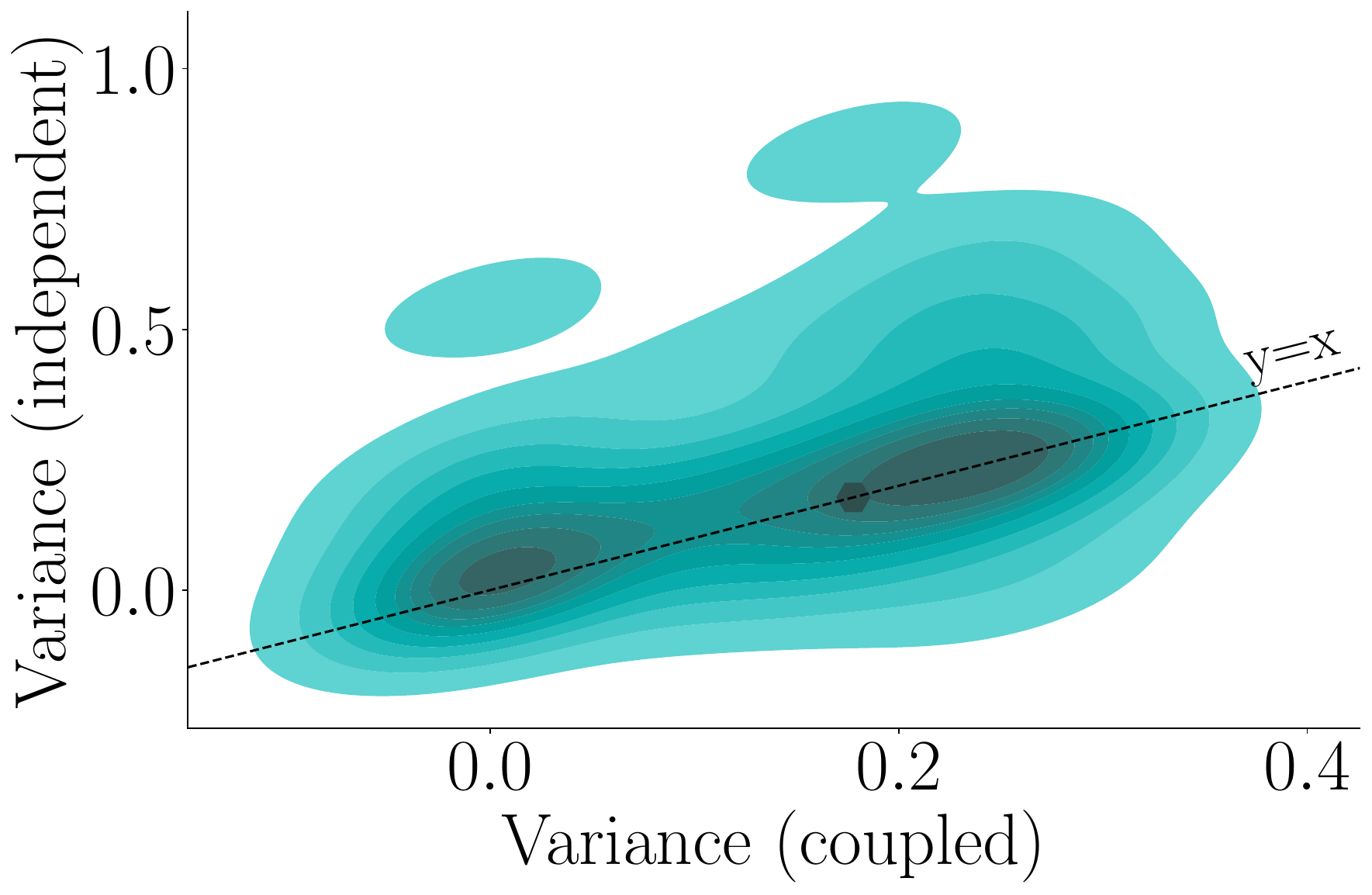} &
    \includegraphics[width=0.23\linewidth]{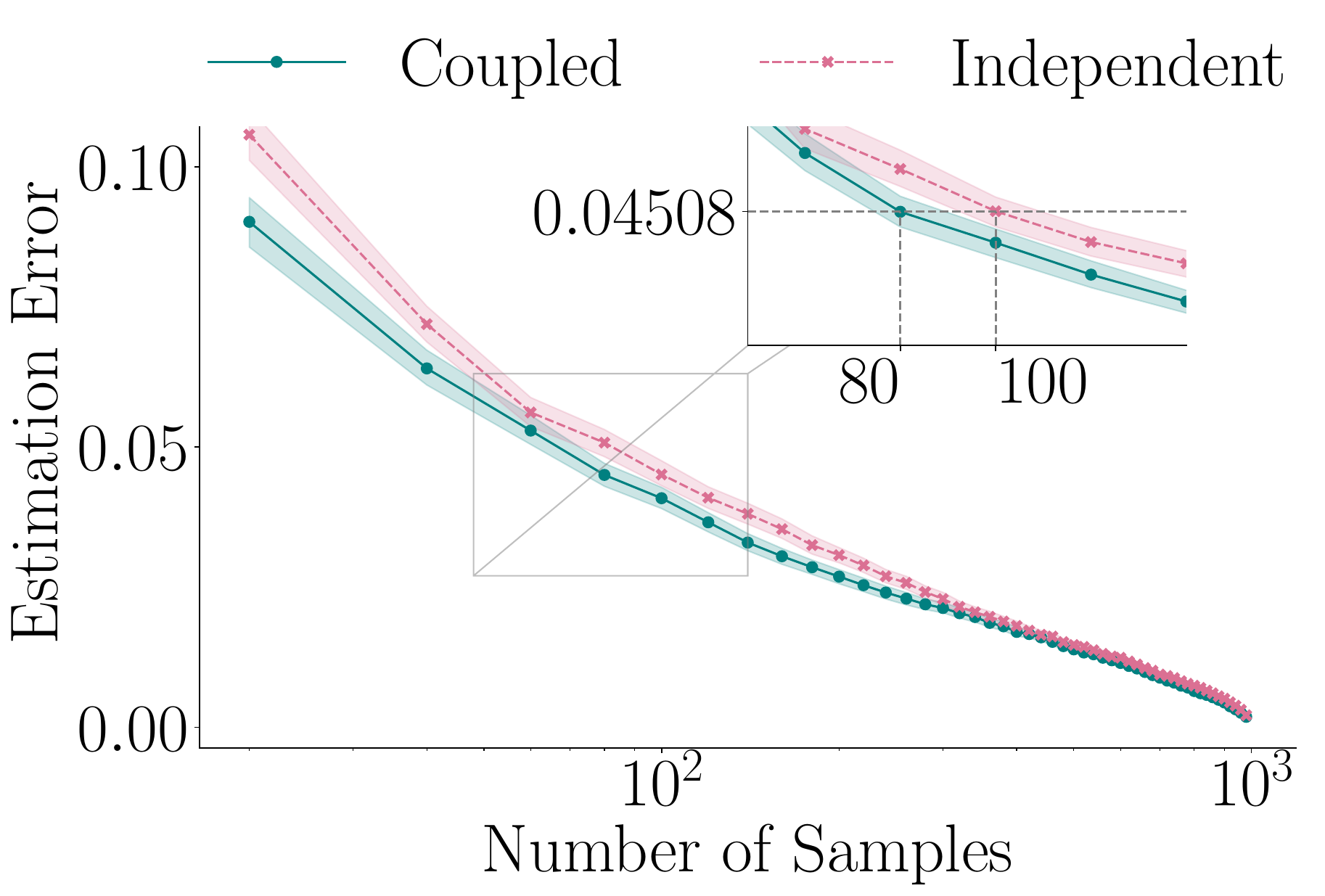} \\ \\
    \multicolumn{3}{c}{\texttt{v0.2} vs. \texttt{v0.3-bnb-4bit}}\\
    \includegraphics[width=0.23\linewidth]{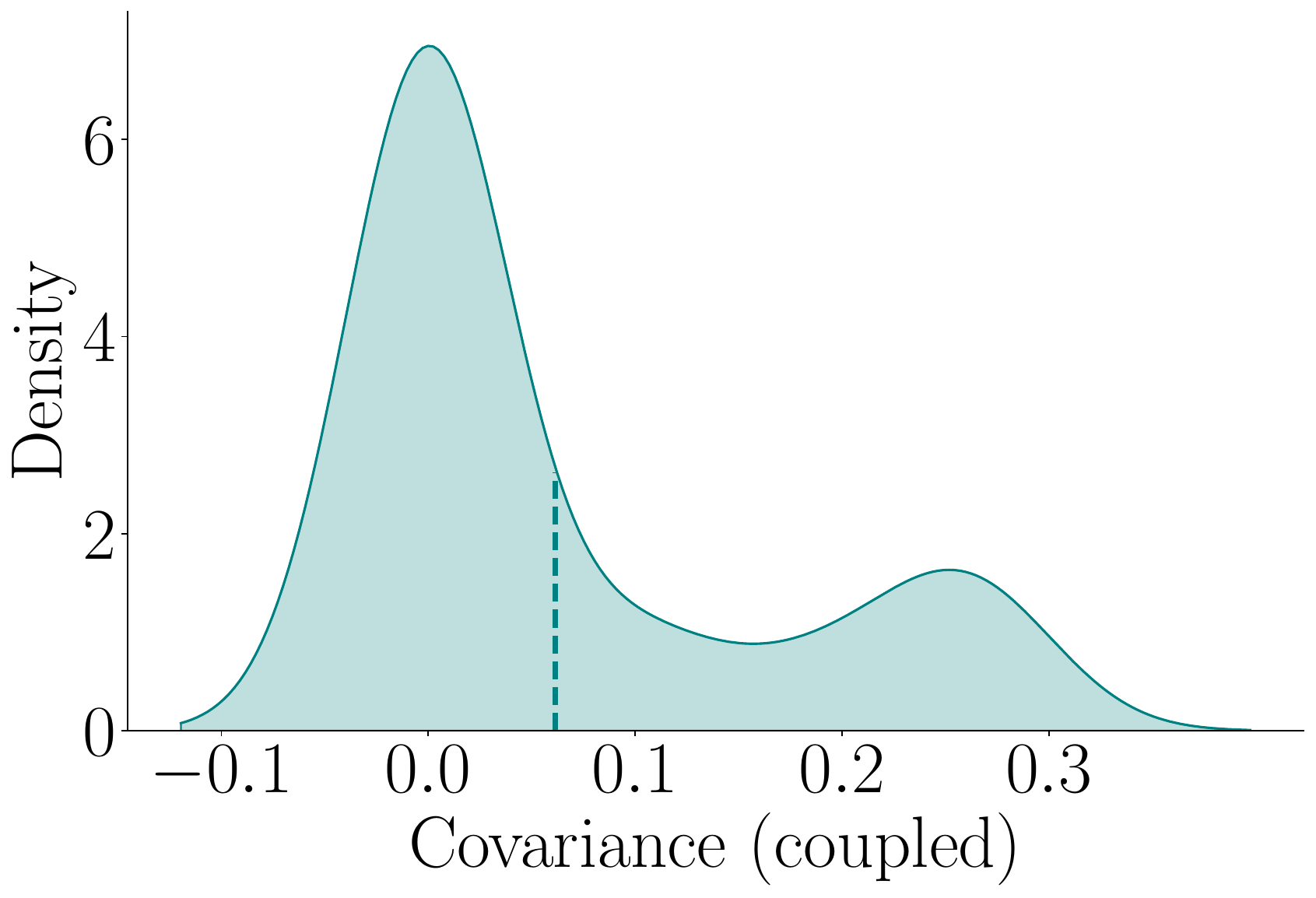} &
    \includegraphics[width=0.23\linewidth]{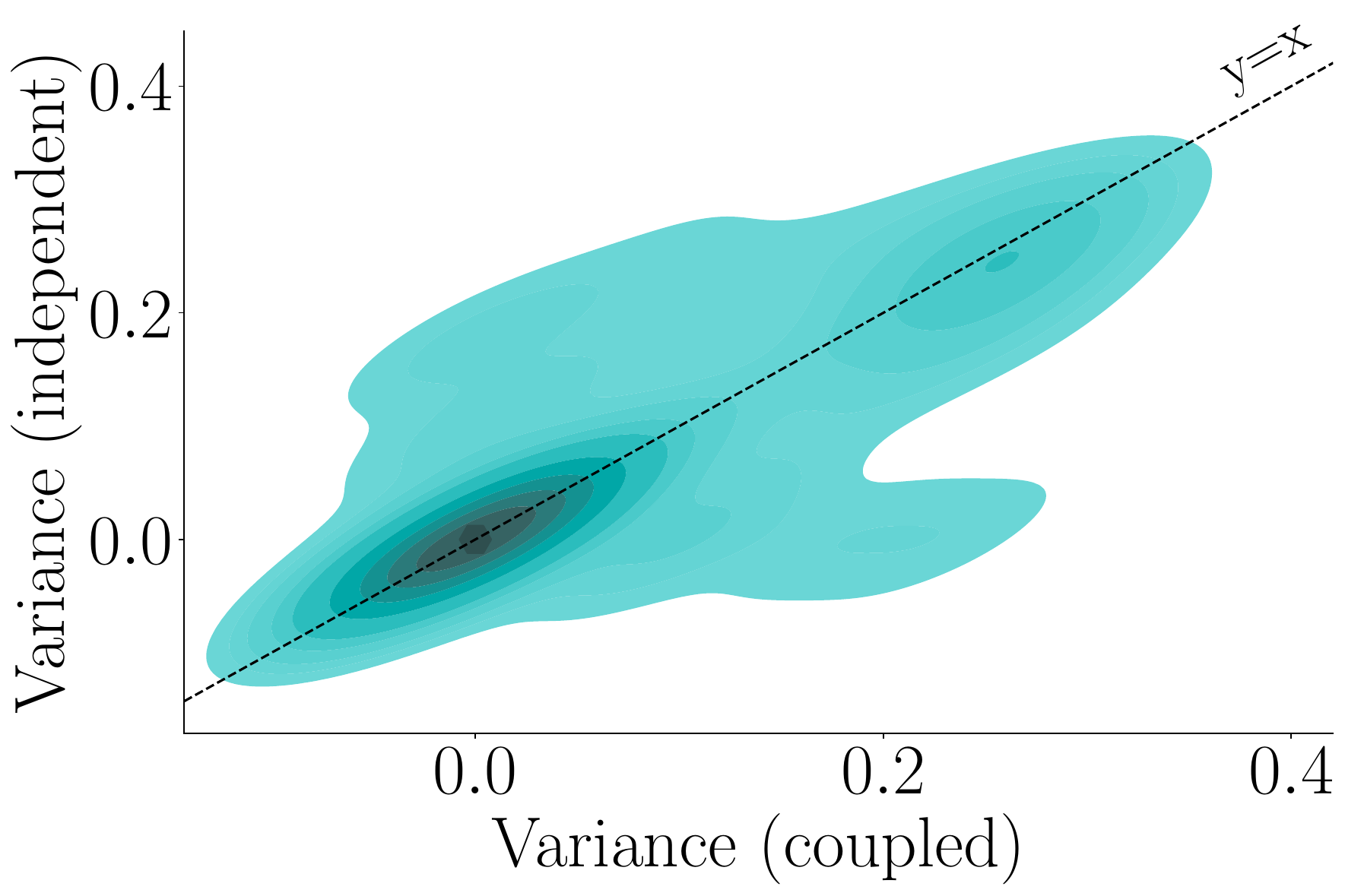} &
    \includegraphics[width=0.23\linewidth]{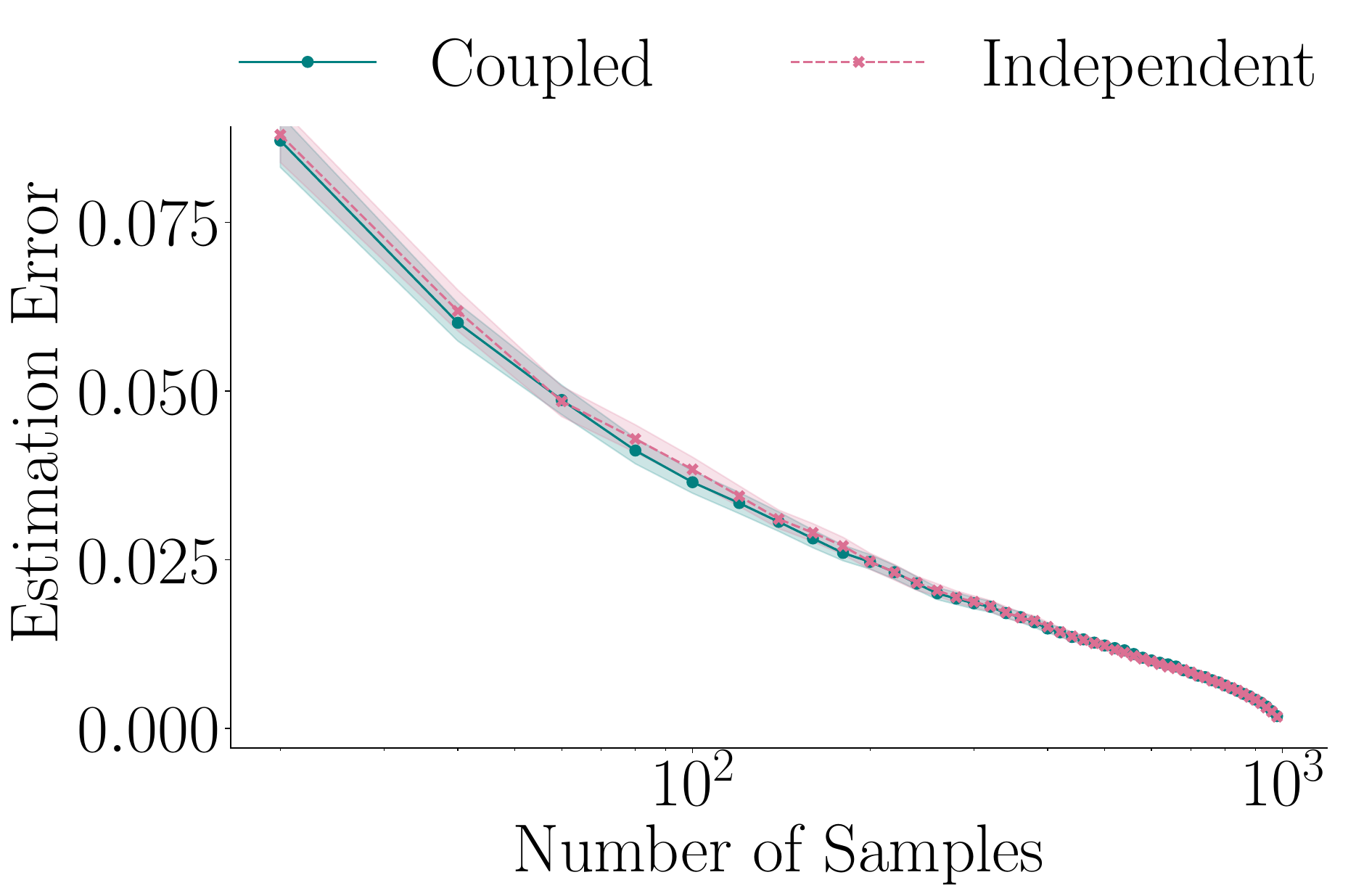} \\ \\
     \multicolumn{3}{c}{\texttt{v0.2} vs. \texttt{v0.1}}\\
    \includegraphics[width=0.23\linewidth]{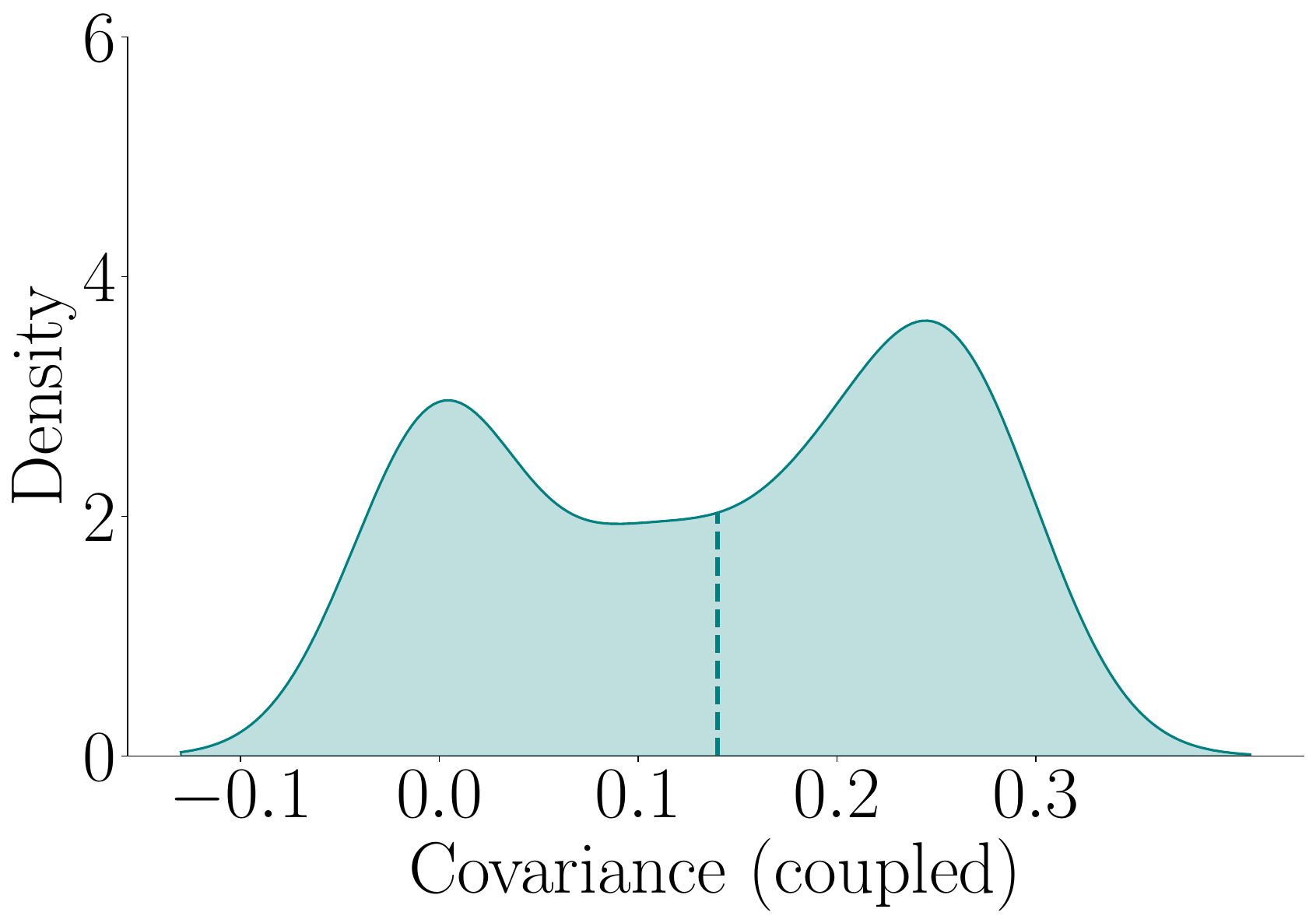} &
    \includegraphics[width=0.23\linewidth]{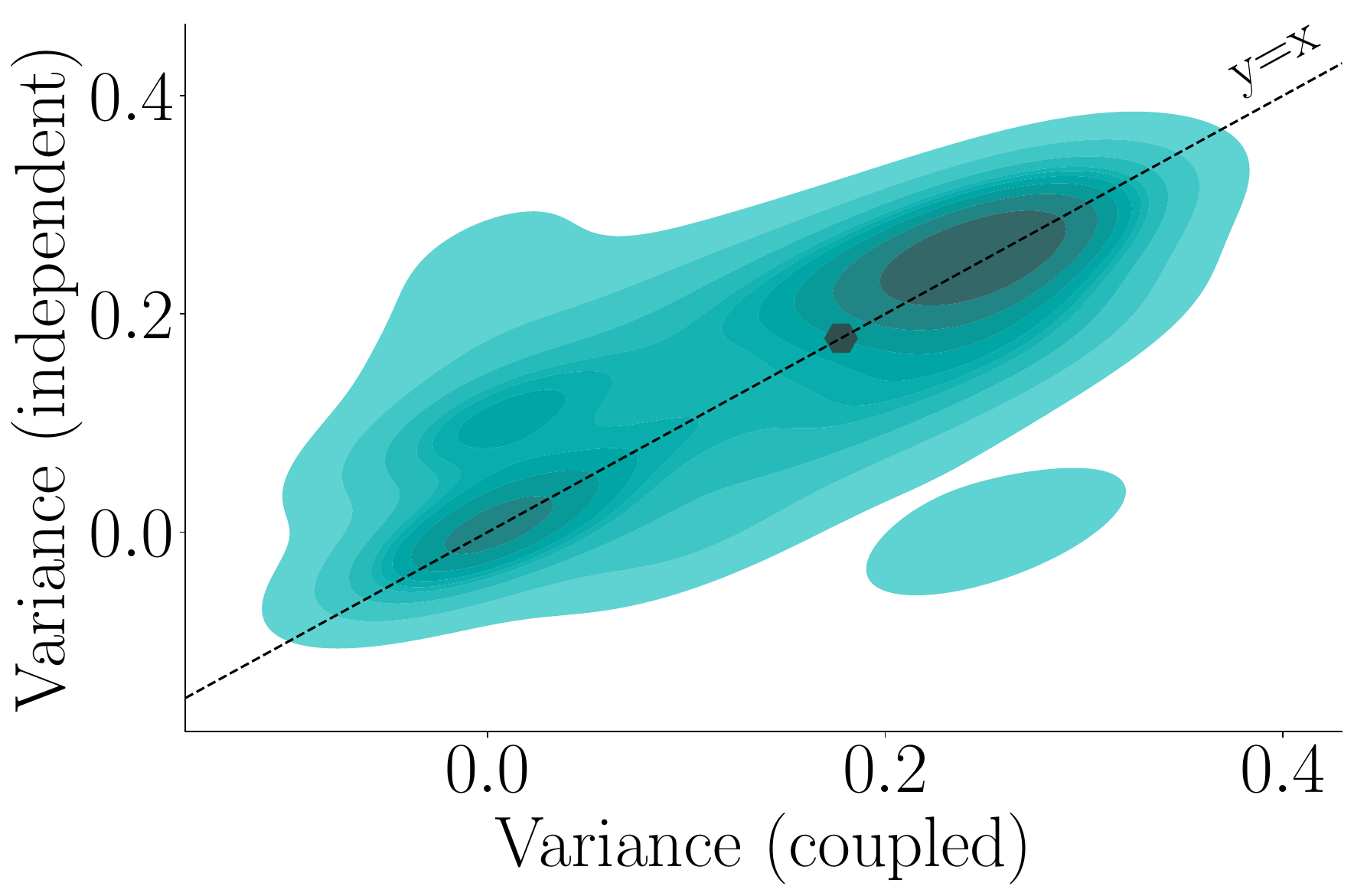} &
    \includegraphics[width=0.23\linewidth]{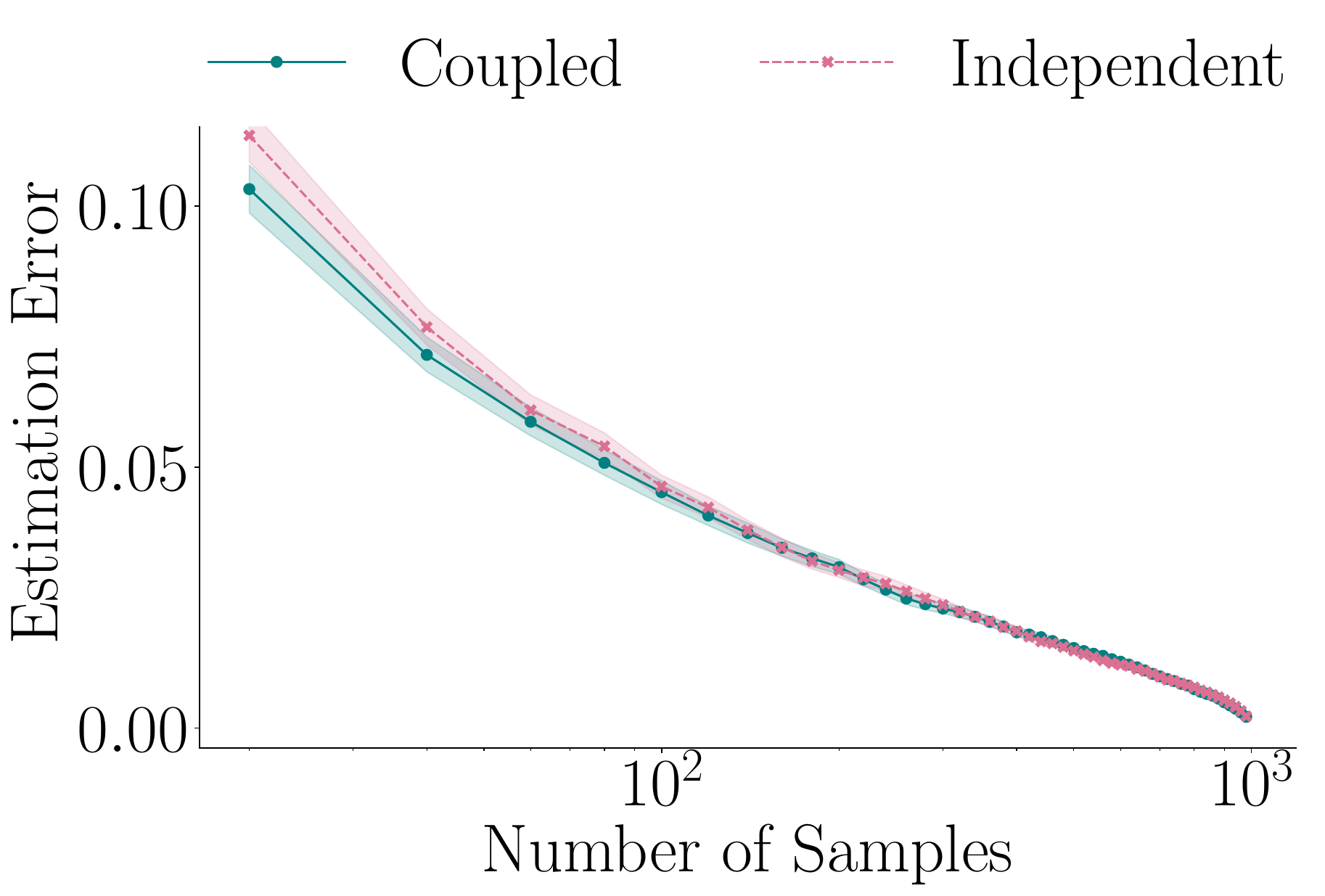} \\ \\
    (a) Score covariance & (b) Variance of the score difference & (c) Estimation error vs. \# samples \\ 
    
\end{tabular}
    \caption{\textbf{Comparison between five pairs of LLMs in the \texttt{Mistral} family on multiple-choice questions from the ``college computer science'' knowledge area of the MMLU dataset.}
    Panels in column (a) show the kernel density estimate (KDE) of the covariance between the scores of the two LLMs on each question under coupled generation; the dashed lines correspond to average values. Panels in column (b) show the KDE of the variance of the difference between the scores of the LLMs on each question under coupled and independent generation; the highlighted points correspond to median values. Panels in column (c) show the absolute error in the estimation of the expected difference between the scores of the LLMs against the number of samples; for each point on the x-axis, we perform $1{,}000$ sub-samplings and shaded areas correspond to $95\%$ confidence intervals.}
    \label{fig:mmlu-mistral-last}
\end{figure}

\begin{figure}[h]
\centering
\begin{tabular}{c c c}
    \multicolumn{3}{c}{\textbf{College chemistry}} \\
    \includegraphics[width=0.23\linewidth]{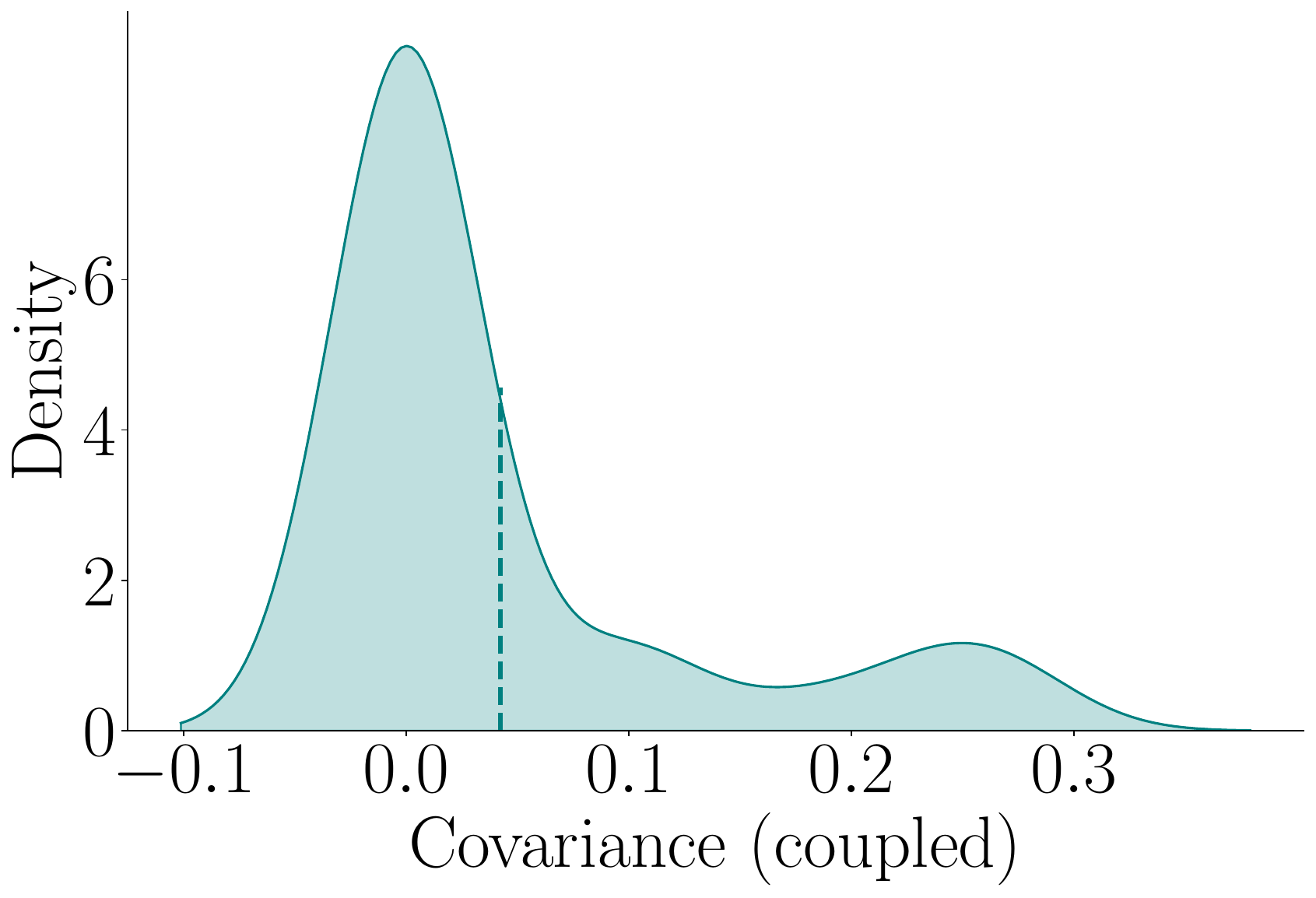} &
    \includegraphics[width=0.23\linewidth]{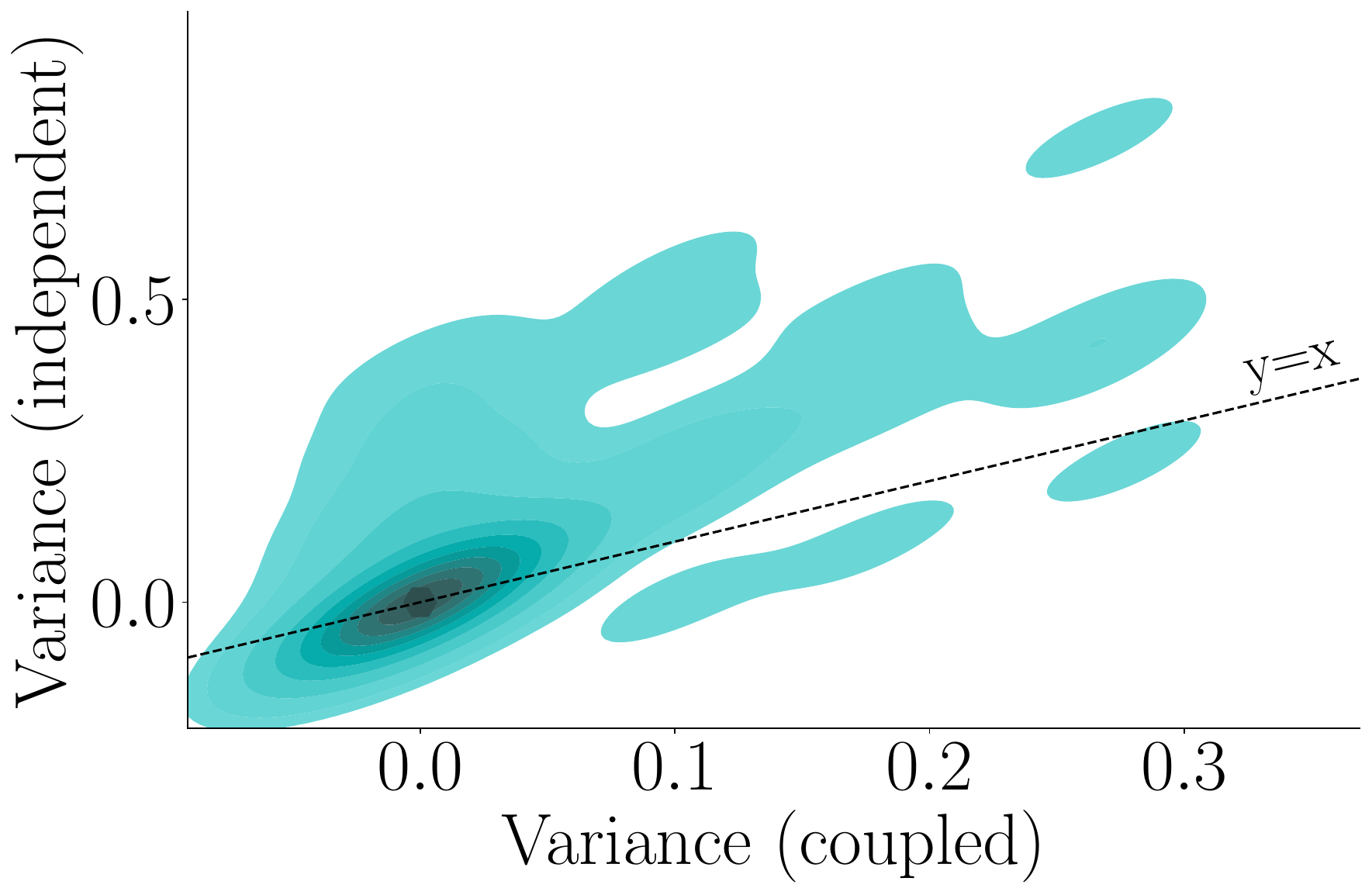} &
    \includegraphics[width=0.23\linewidth]{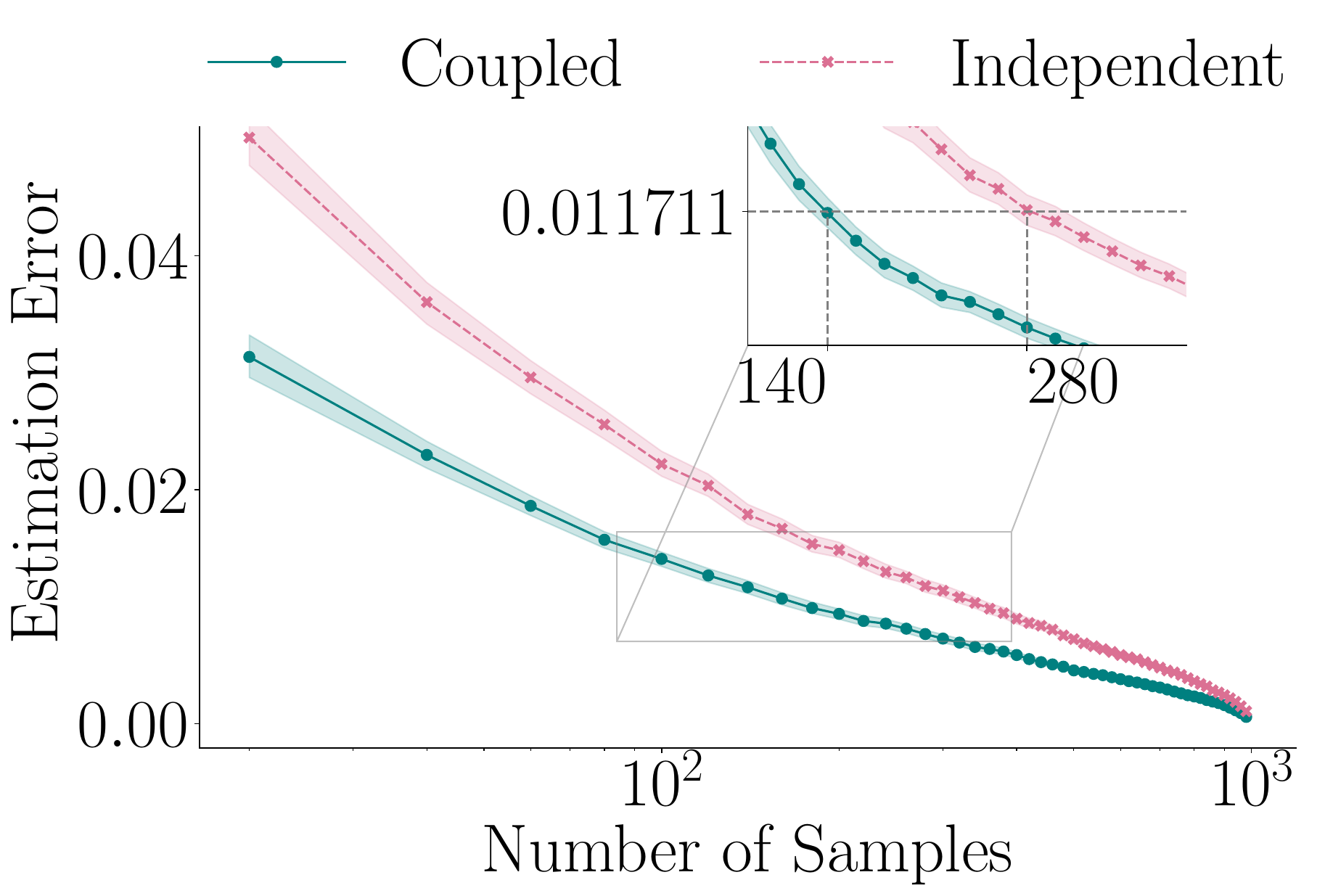} \\
    \multicolumn{3}{c}{\textbf{Professional accounting}} \\
    \includegraphics[width=0.23\linewidth]{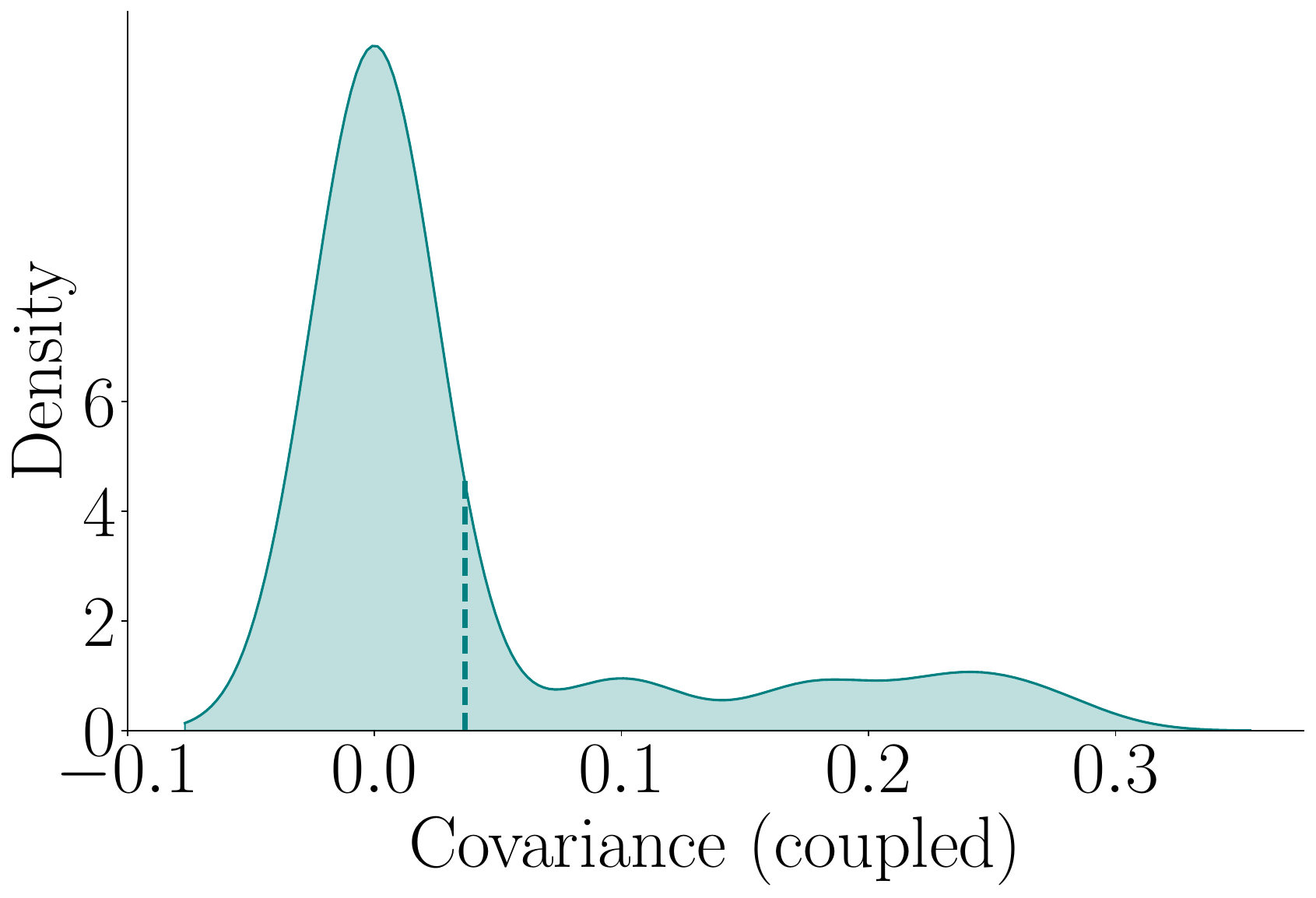} &
    \includegraphics[width=0.23\linewidth]{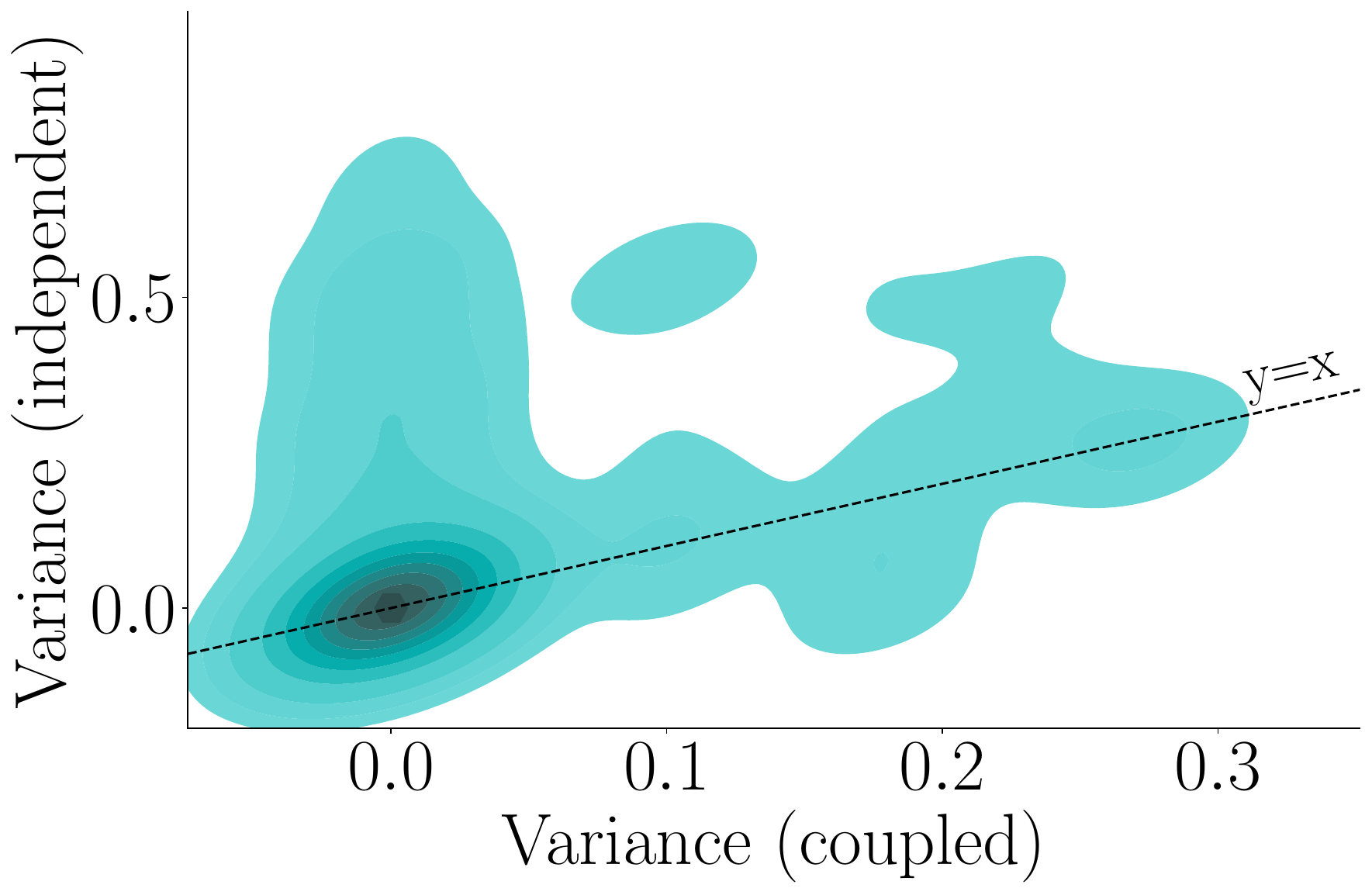} &
    \includegraphics[width=0.23\linewidth]{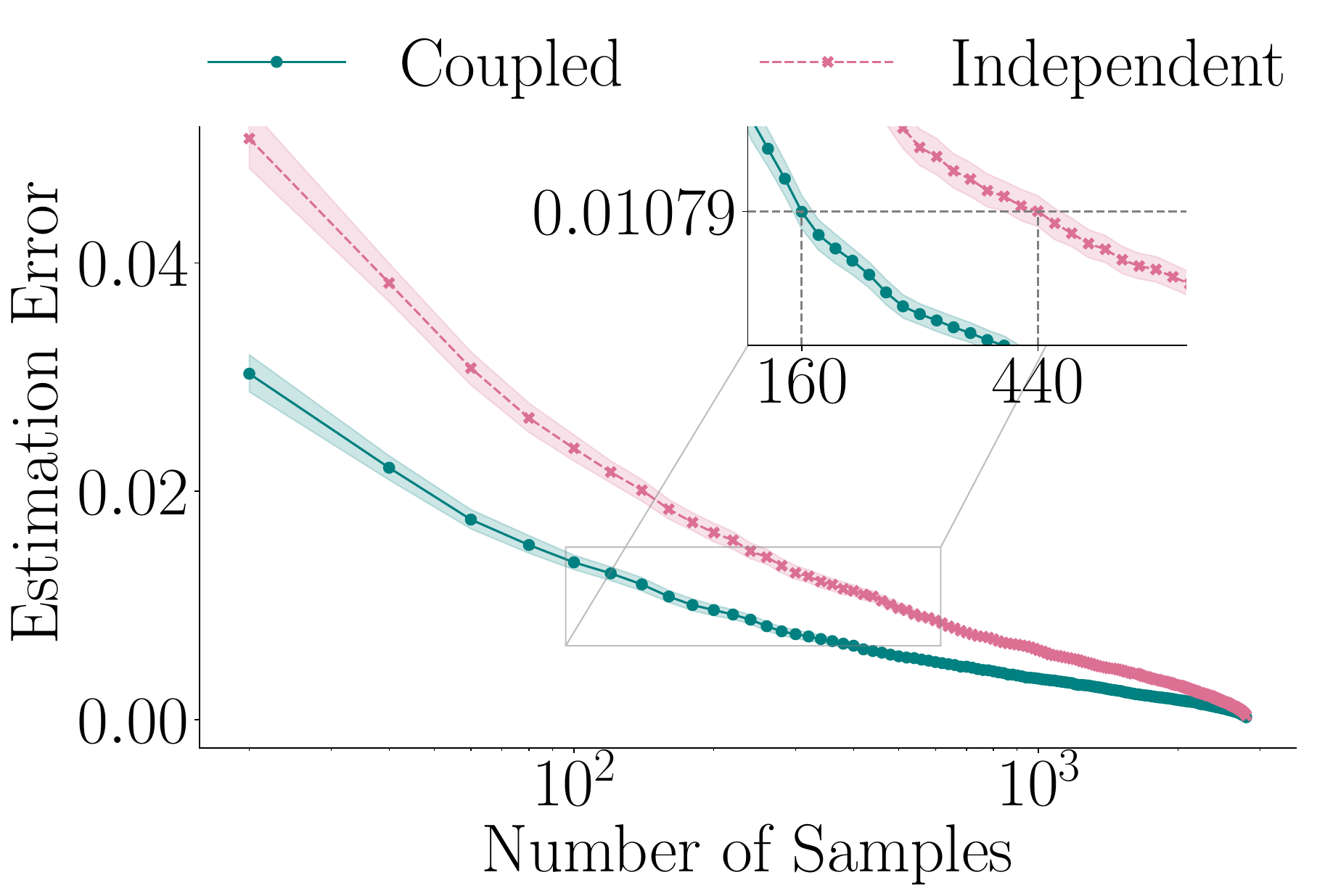} \\
    \multicolumn{3}{c}{\textbf{Professional law}}\\
    \includegraphics[width=0.23\linewidth]{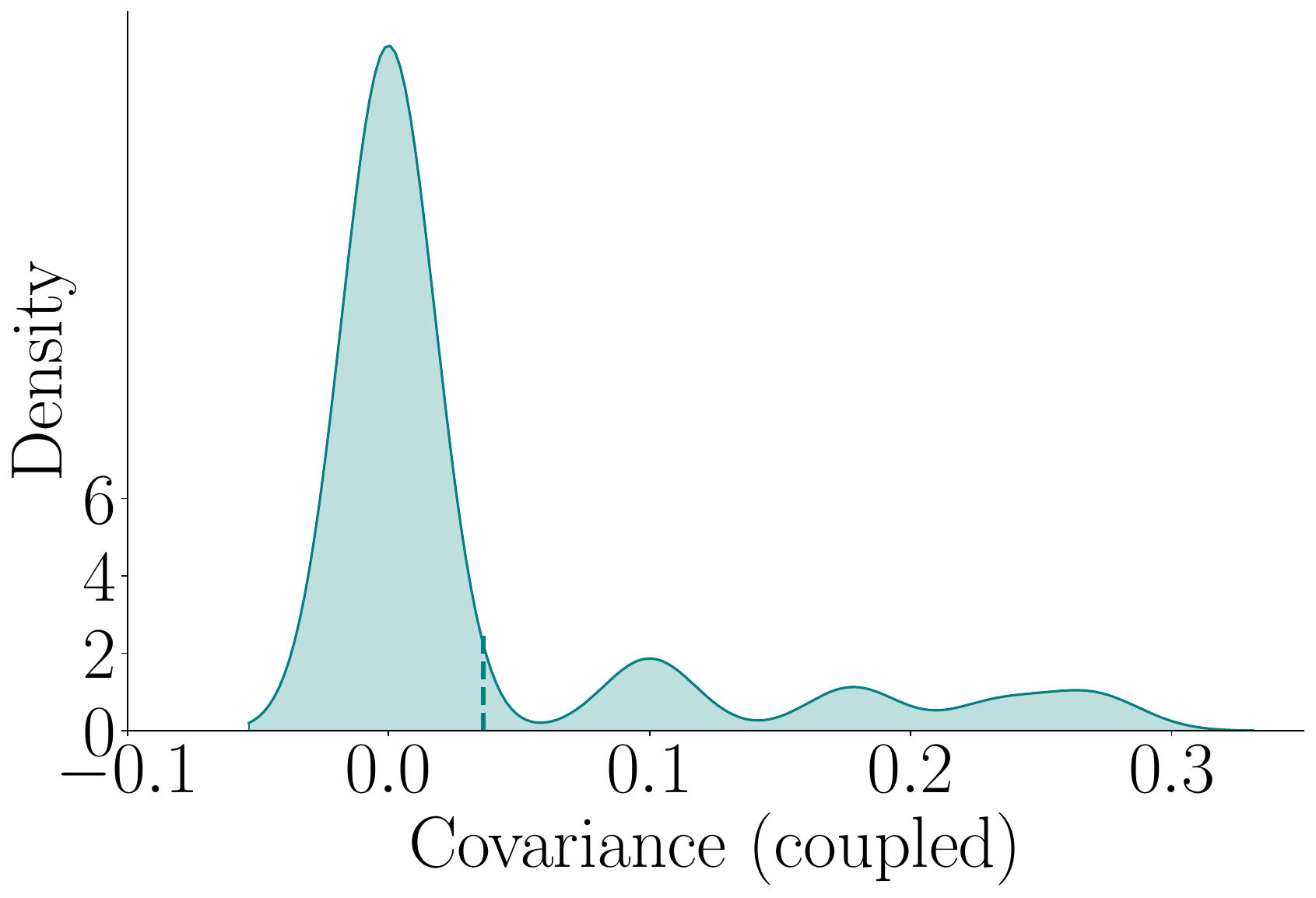} &
    \includegraphics[width=0.23\linewidth]{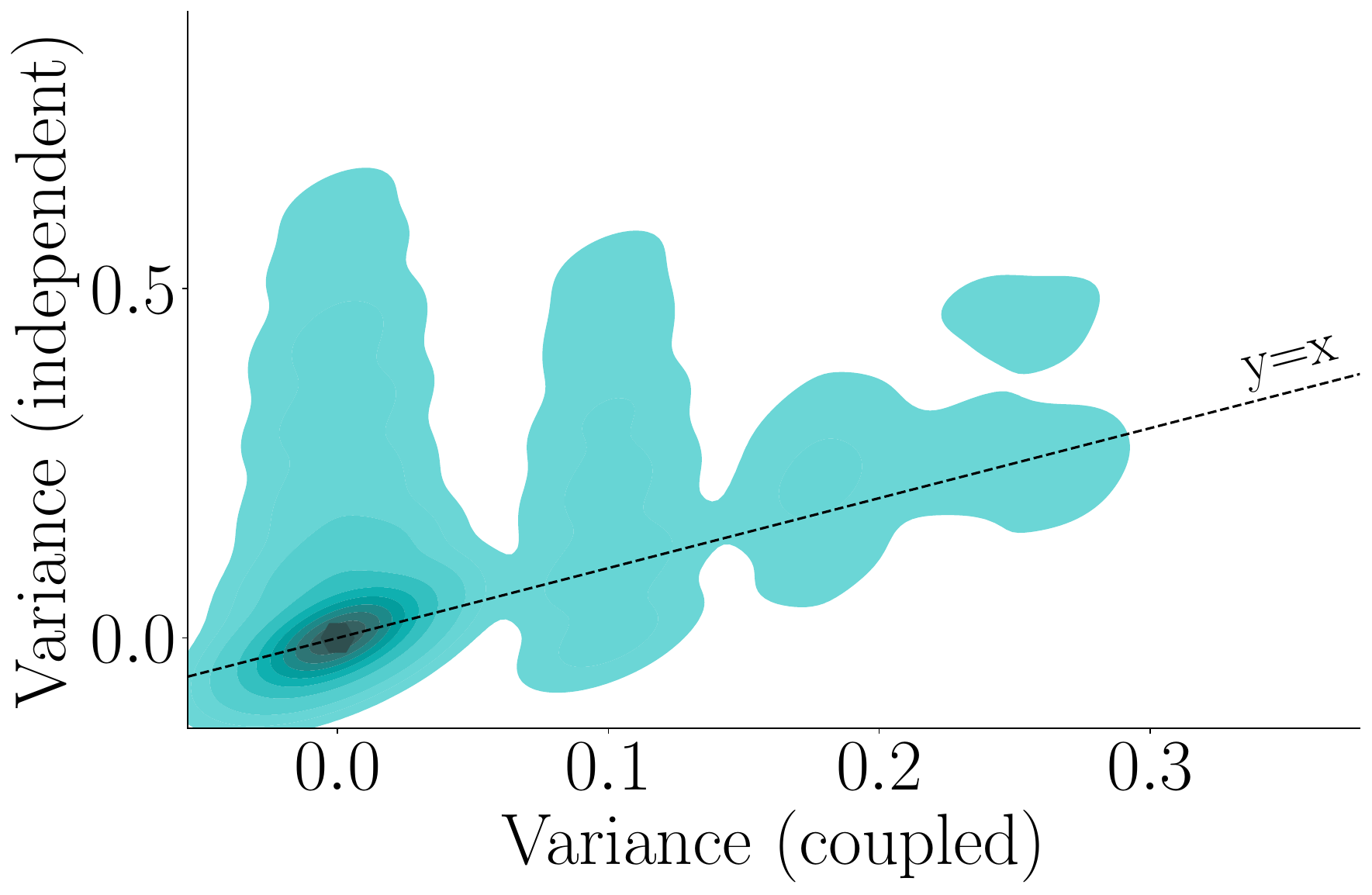} &
    \includegraphics[width=0.23\linewidth]{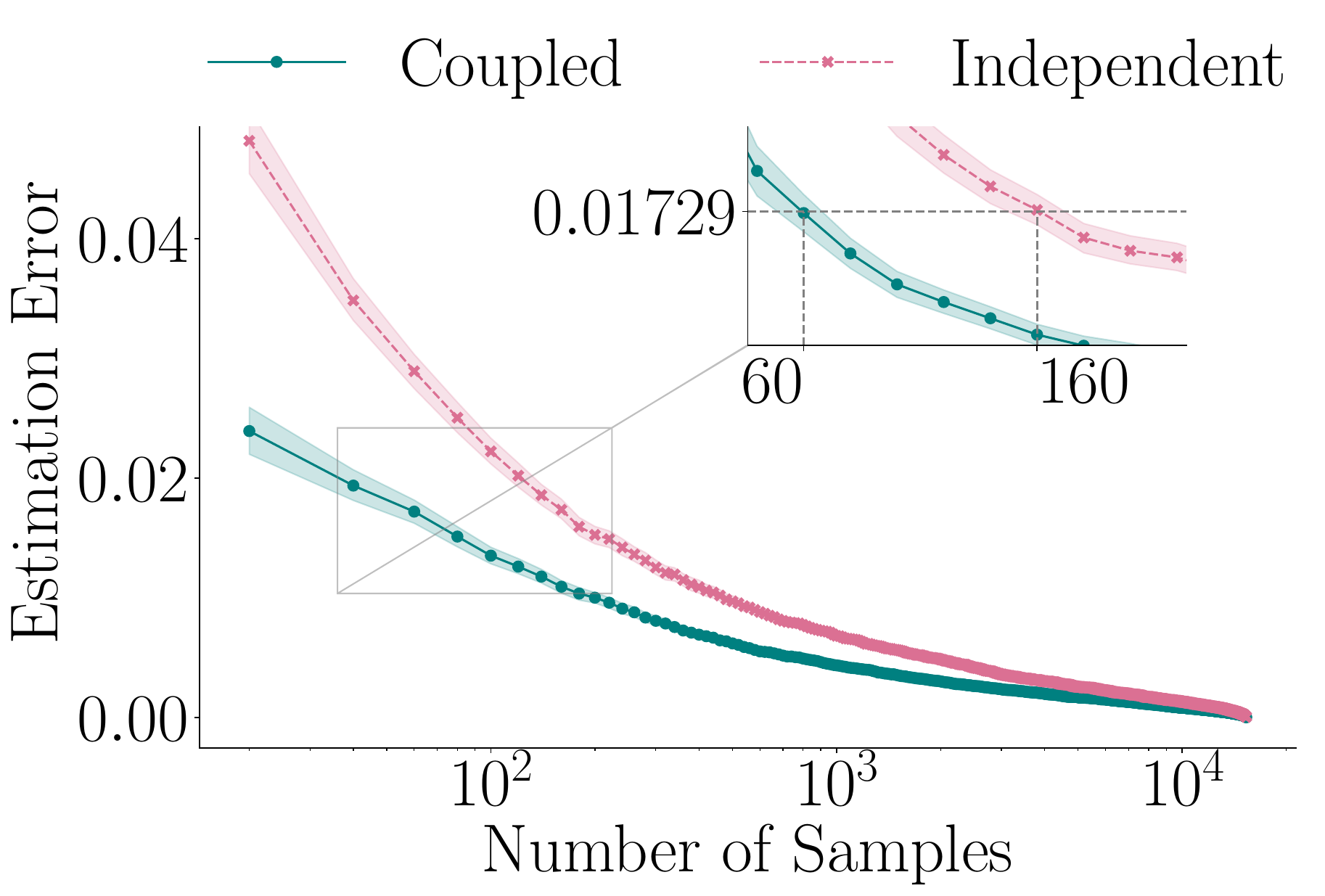} \\ \\
    \multicolumn{3}{c}{\textbf{Professional medicine}} \\   
    \includegraphics[width=0.23\linewidth]{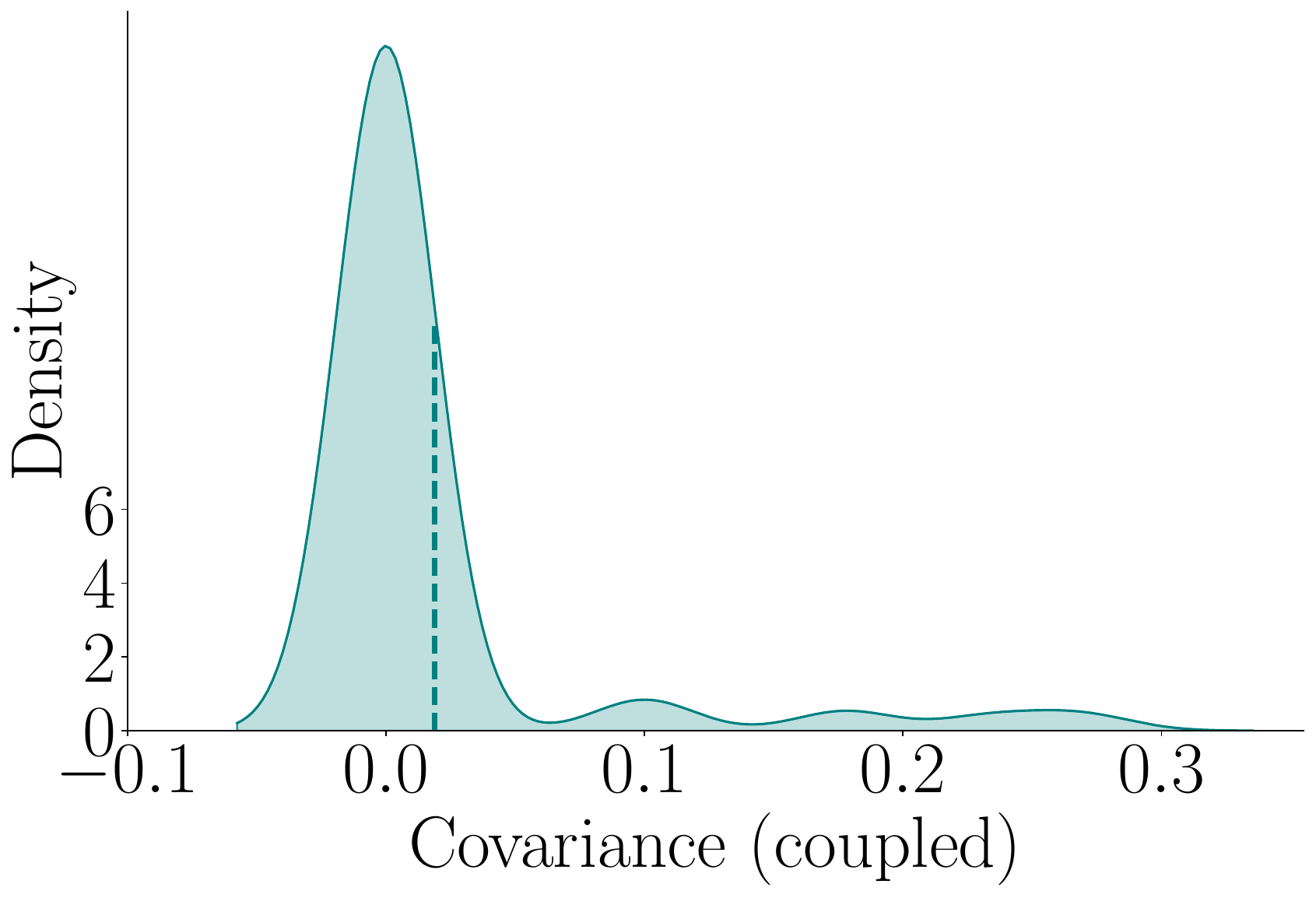} &
    \includegraphics[width=0.23\linewidth]{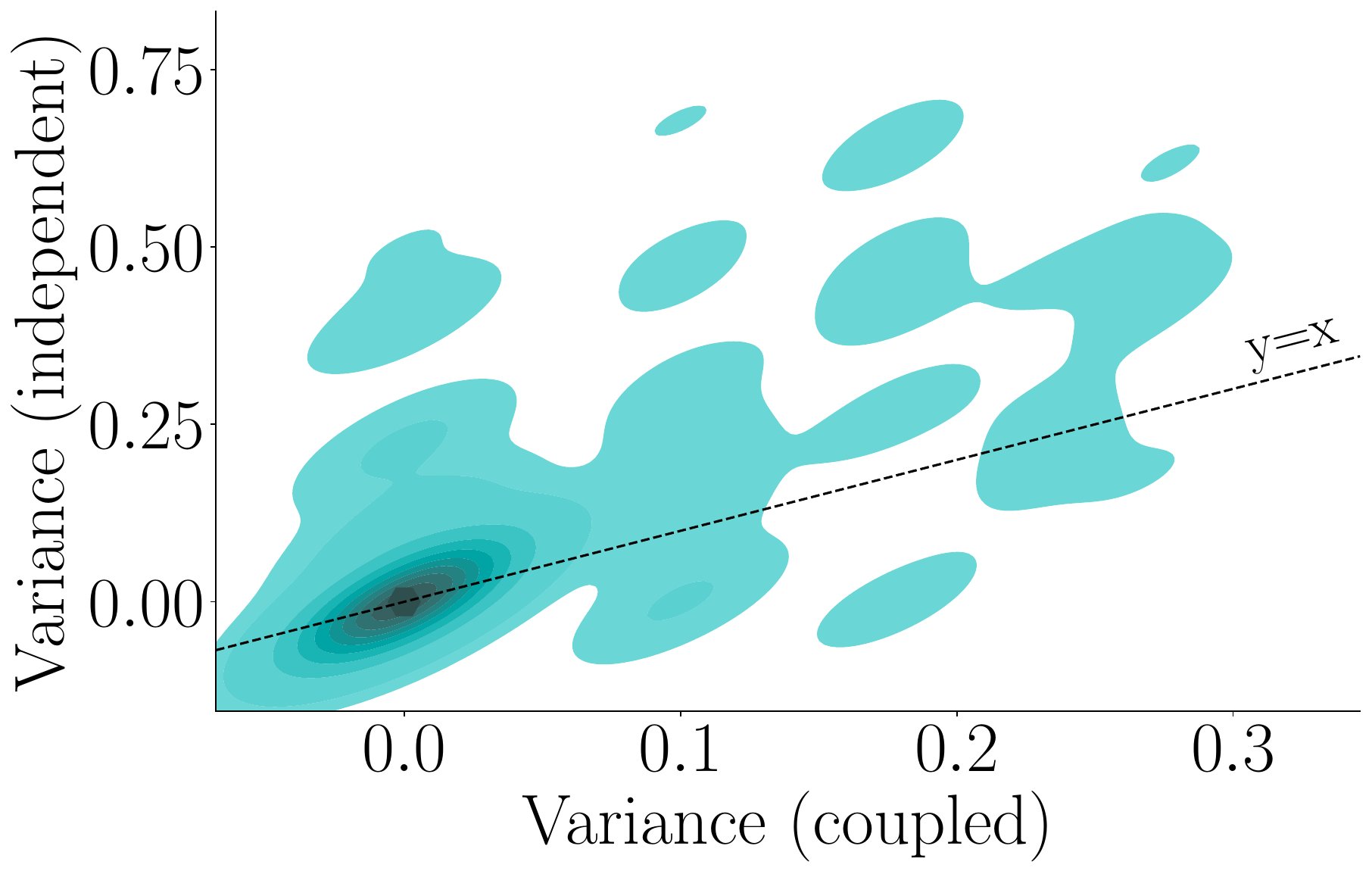} &
    \includegraphics[width=0.23\linewidth]{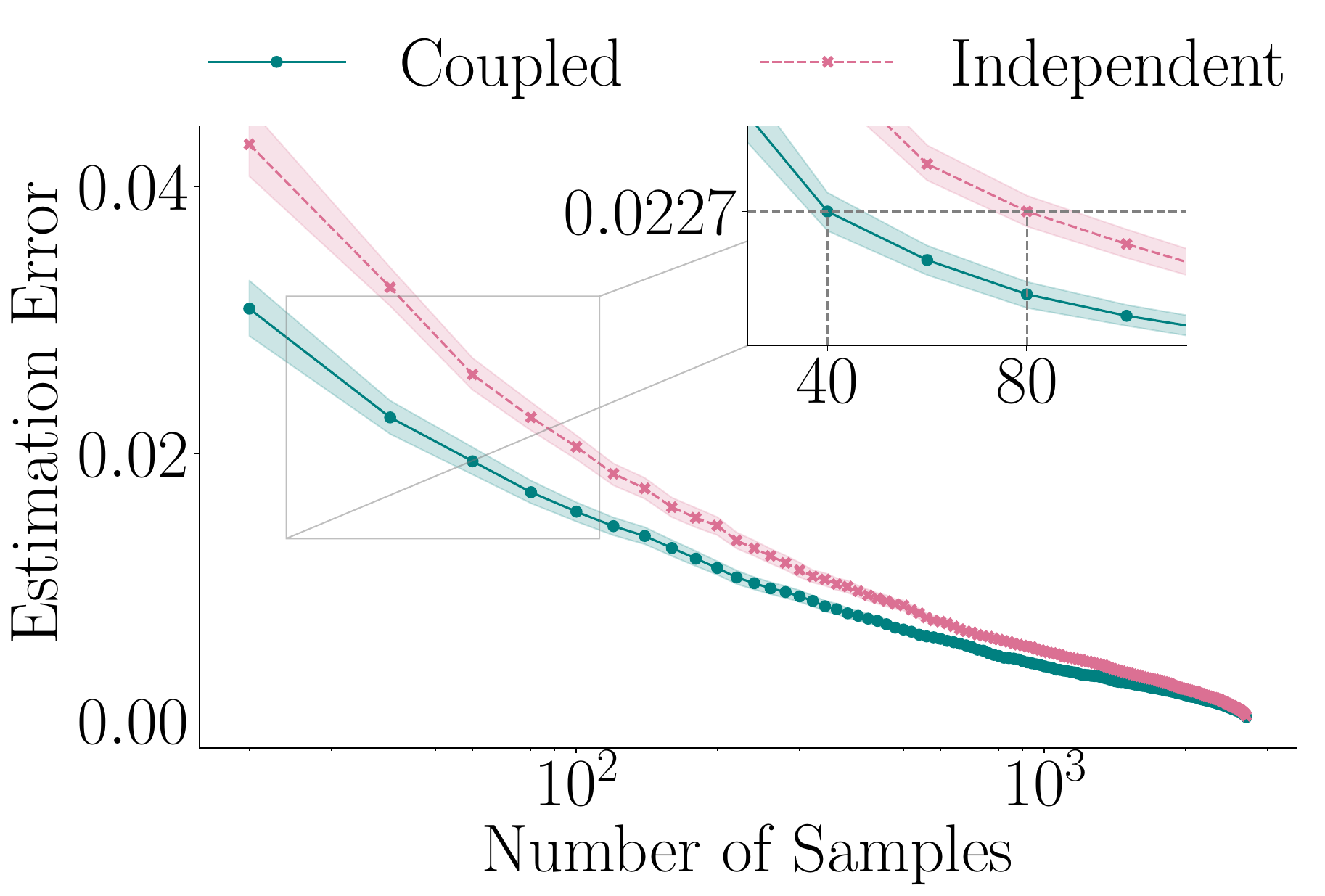} \\
    (a) Score covariance & (b) Variance of the score difference & (c) Estimation error vs. \# samples \\ 
\end{tabular}
    \caption{\textbf{Comparison between \texttt{v0.3} and \texttt{v0.3-bnb-8bit} from the \texttt{Mistral} family on multiple-choice questions from four knowledge areas of the MMLU dataset.}
    Panels in column (a) show the kernel density estimate (KDE) of the covariance between the scores of the two LLMs on each question under coupled generation; the dashed lines correspond to average values. Panels in column (b) show the KDE of the variance of the difference between the scores of the LLMs on each question under coupled and independent generation; the highlighted points correspond to median values. Panels in column (c) show the absolute error in the estimation of the expected difference between the scores of the LLMs against the number of samples; for each point on the x-axis, we perform $1{,}000$ sub-samplings and shaded areas correspond to $95\%$ confidence intervals. We observe qualitatively similar results for other knowledge areas.}
    \label{fig:mmlu-mistral-v03-vs-v03-8bit-areas}
\end{figure}

Here, we experiment with models from the \texttt{Mistral} family on the MMLU dataset following the setup described in Section~\ref{sec:mmlu}. Figures~\ref{fig:mmlu-mistral-first} and~\ref{fig:mmlu-mistral-last} show the results for all pairs of models described in Table~\ref{tab:mistral-names} on the knowledge area \texttt{college computer science}, and Figure~\ref{fig:mmlu-mistral-v03-vs-v03-8bit-areas} shows the results for \texttt{v0.3} against \texttt{v0.3-bnb-8bit} on different knowledge areas. 
Overall, the results are qualitatively similar to those in Section~\ref{sec:mmlu}.

\vspace{-0.2cm}

\subsection{GSM8K and HumanEval Datasets}
\begin{figure}[h]
\centering
\begin{tabular}{c c c}
    \multicolumn{3}{c}{\texttt{v0.3-bnb-8bit} vs. \texttt{v0.2}}\\
    \includegraphics[width=0.23\linewidth]{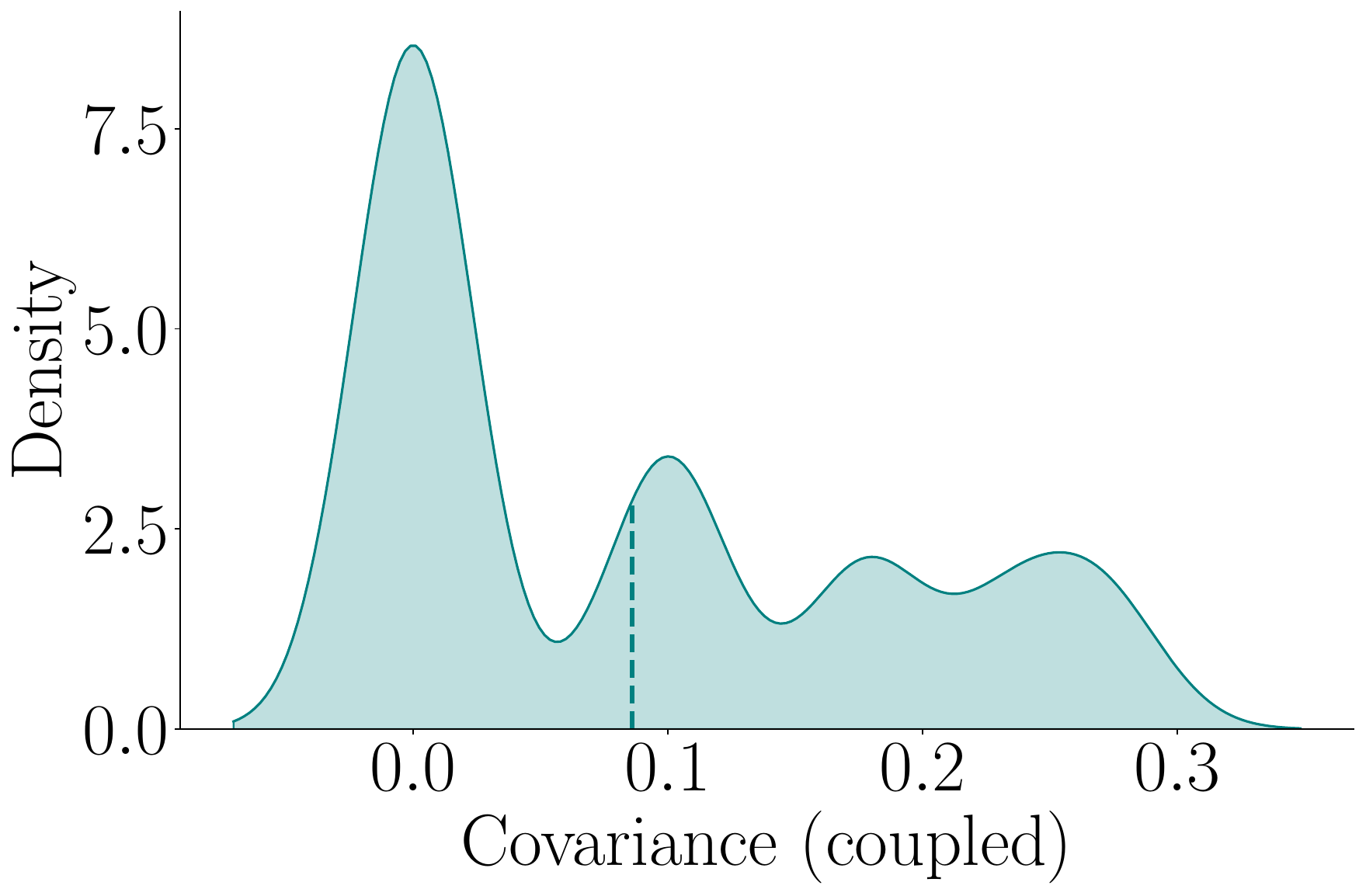} &
    \includegraphics[width=0.23\linewidth]{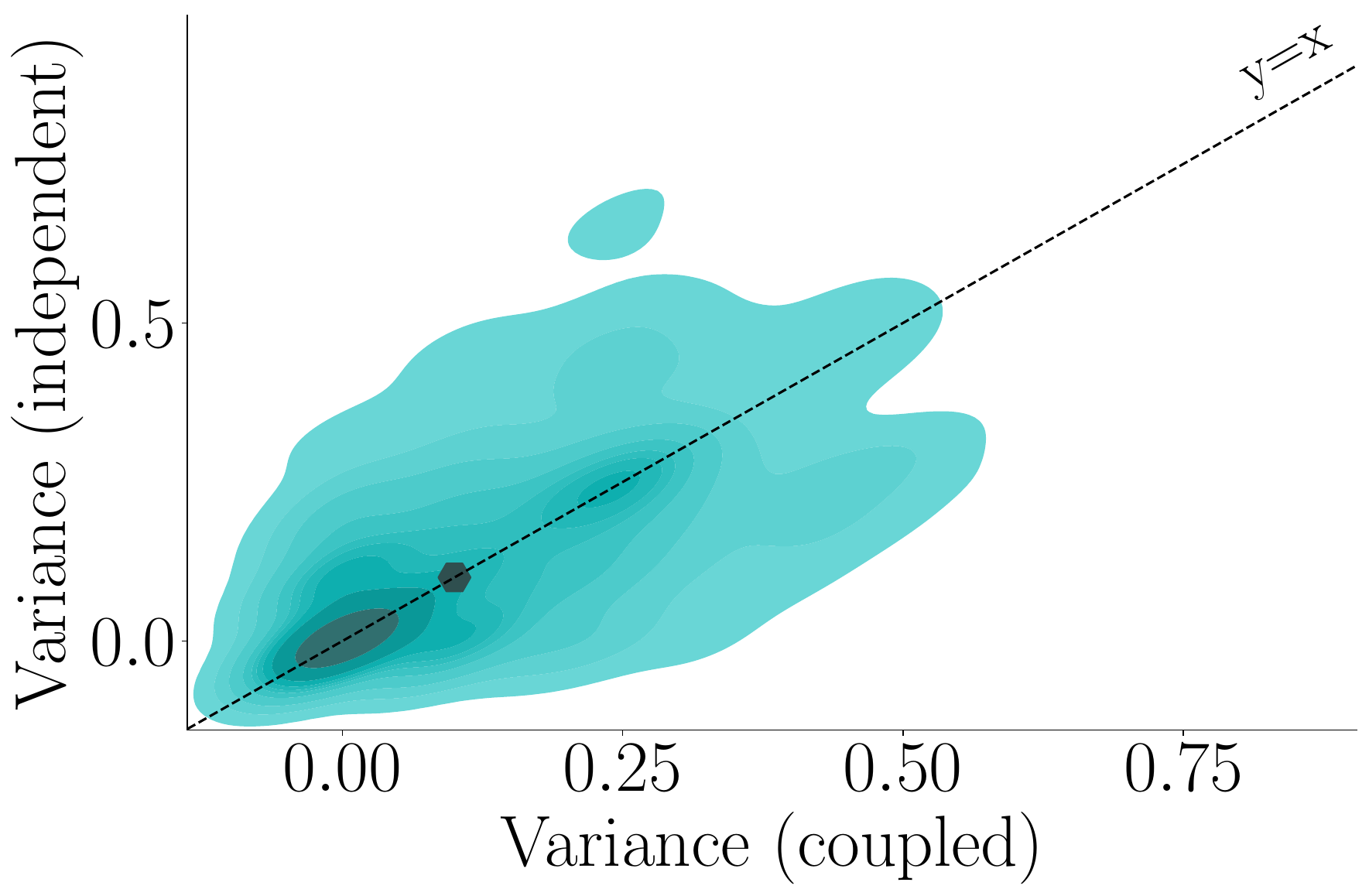} &
    \includegraphics[width=0.23\linewidth]{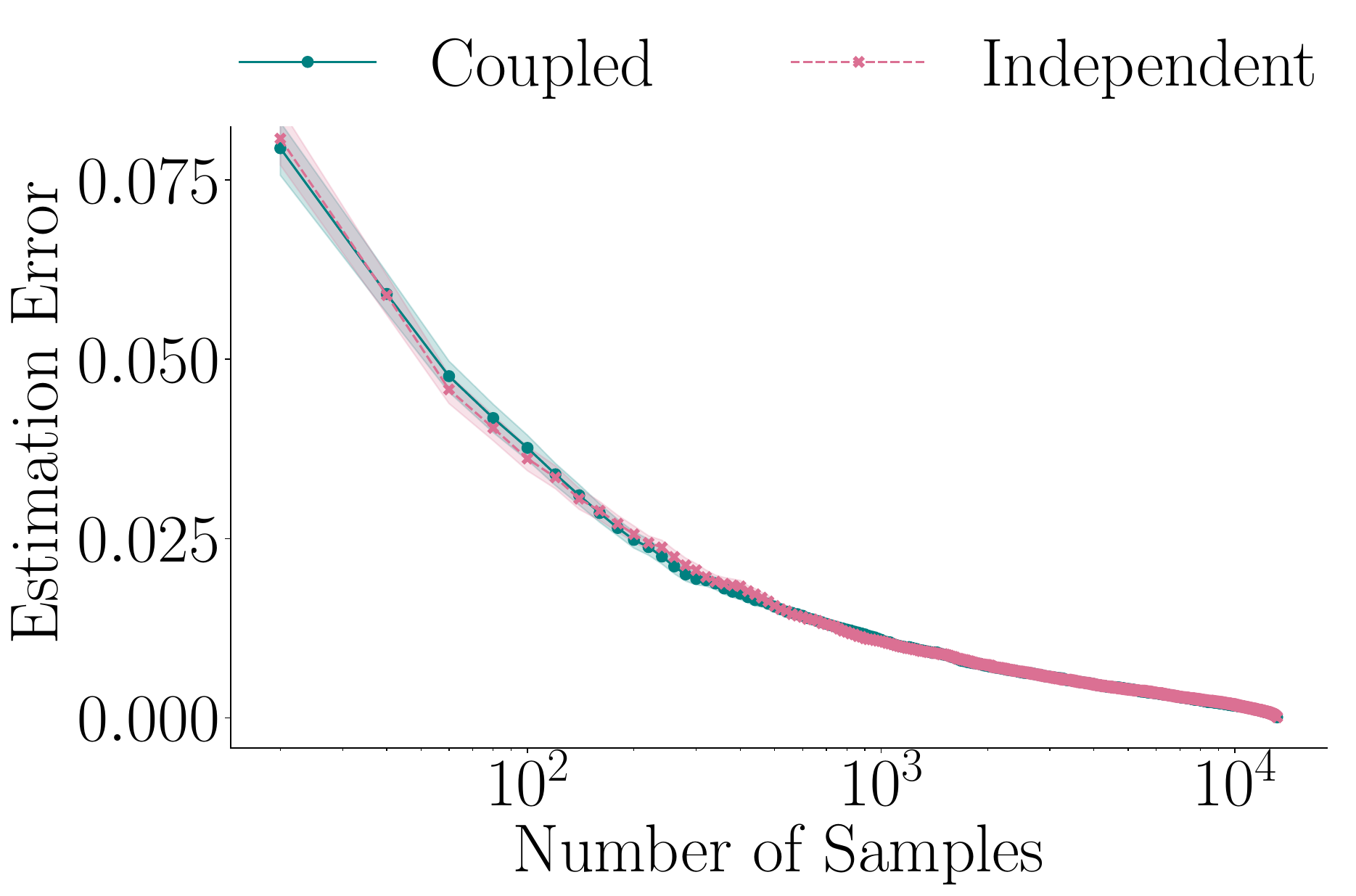} \\ \\
    \multicolumn{3}{c}{\texttt{v0.3-bnb-8bit} vs. \texttt{v0.1}}\\
    \includegraphics[width=0.23\linewidth]{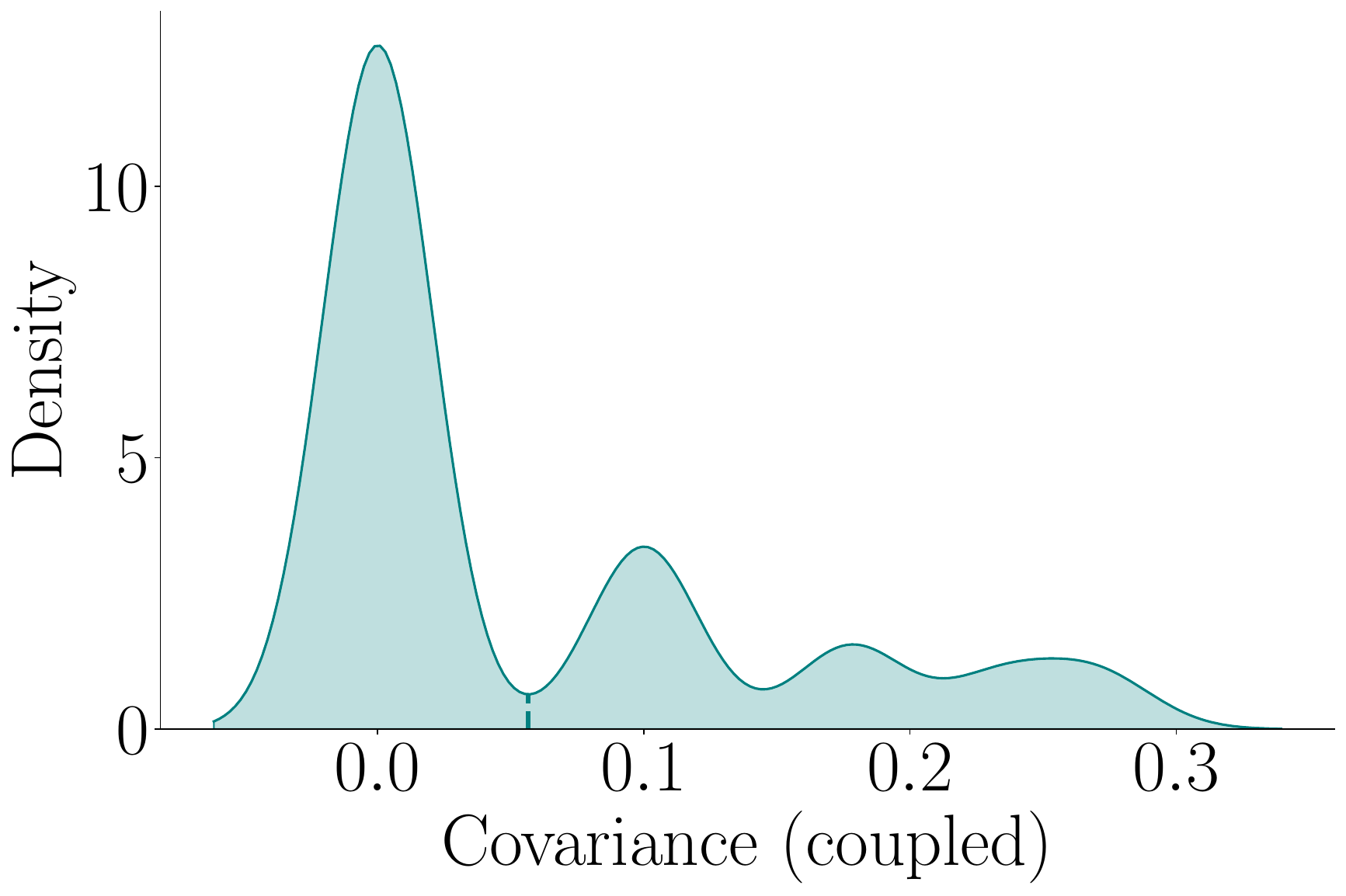} &
    \includegraphics[width=0.23\linewidth]{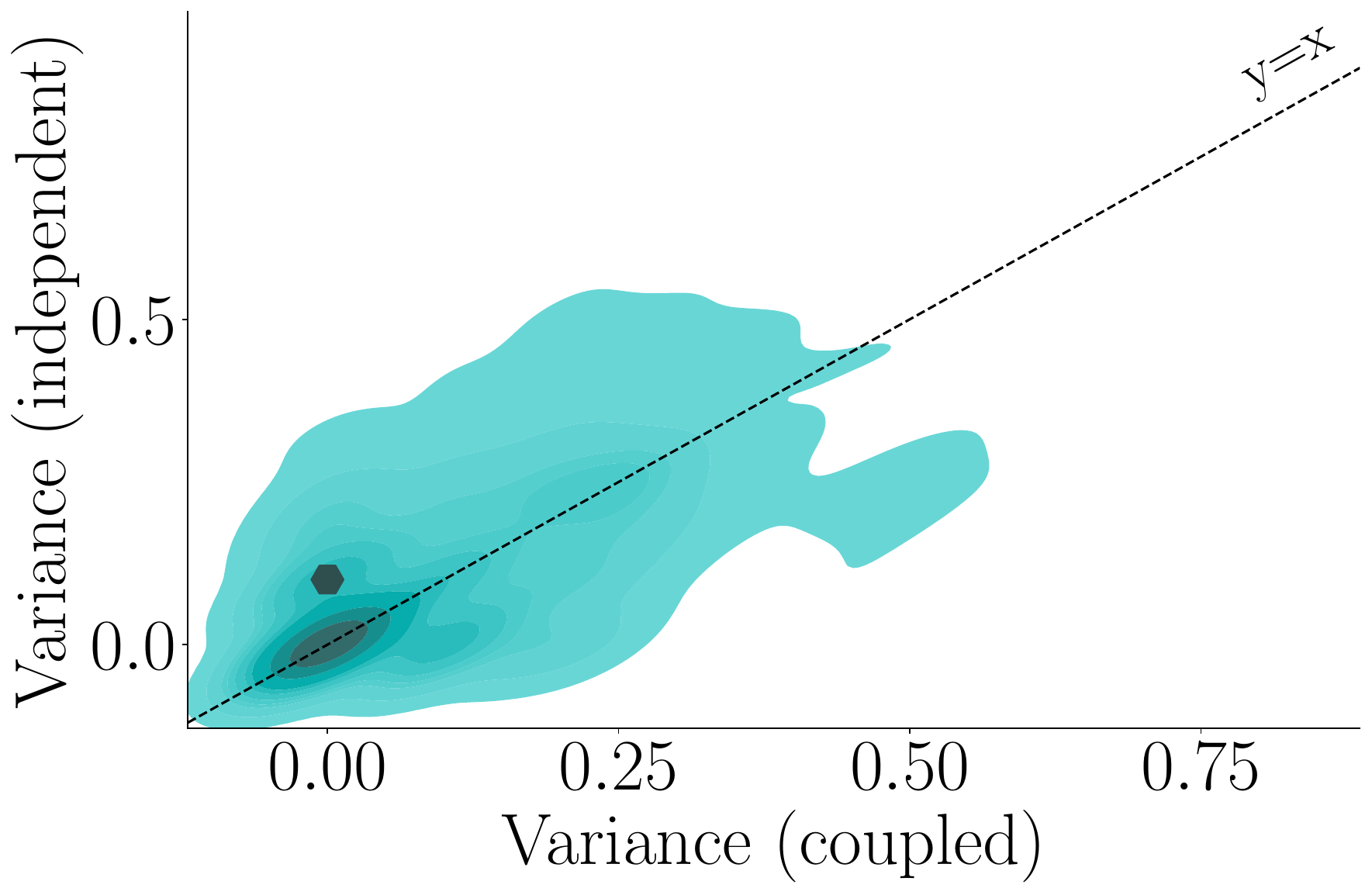} &
    \includegraphics[width=0.23\linewidth]{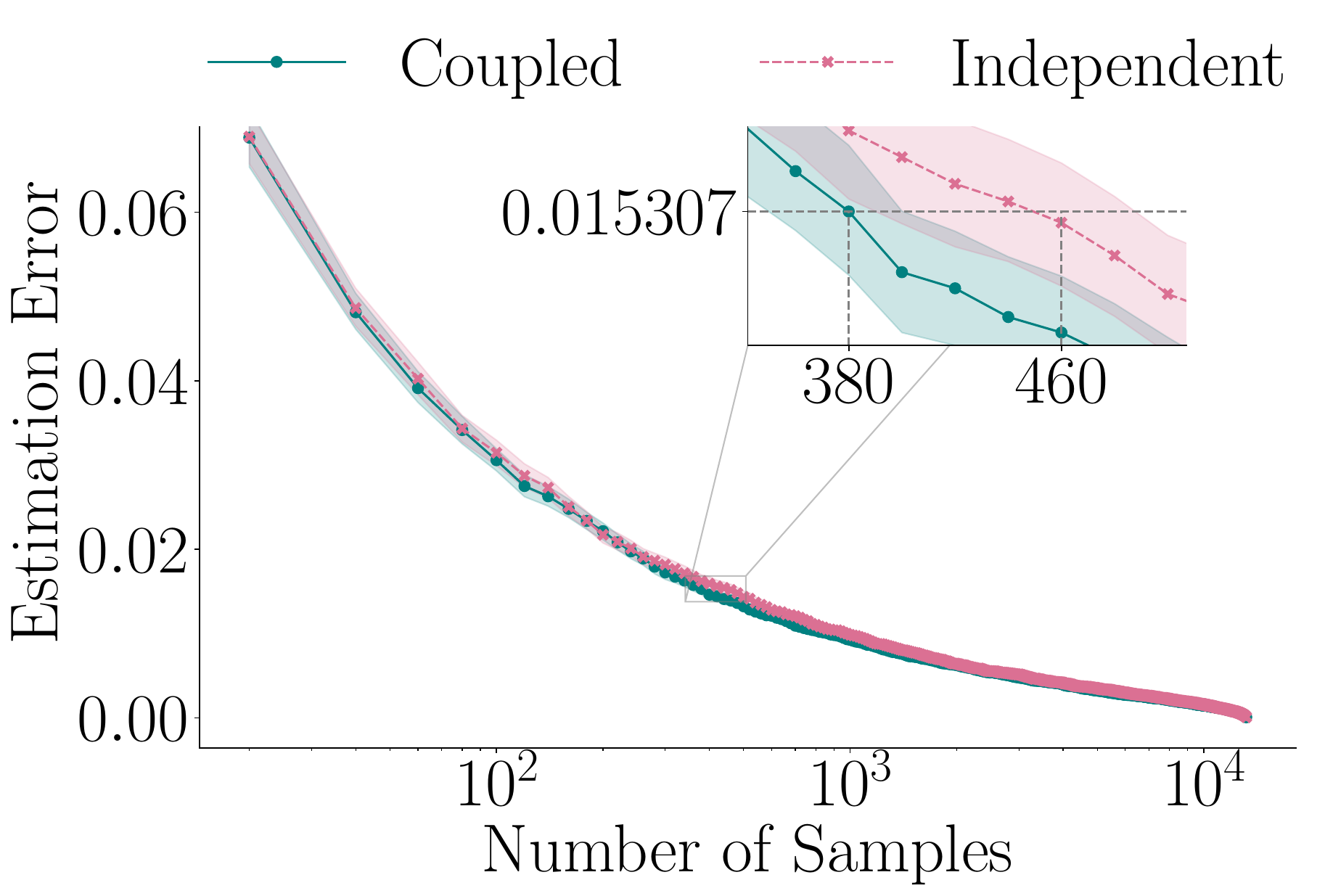} \\ \\
     \multicolumn{3}{c}{\texttt{v0.1} vs. \texttt{v0.3-bnb-4bit}}\\
    \includegraphics[width=0.23\linewidth]{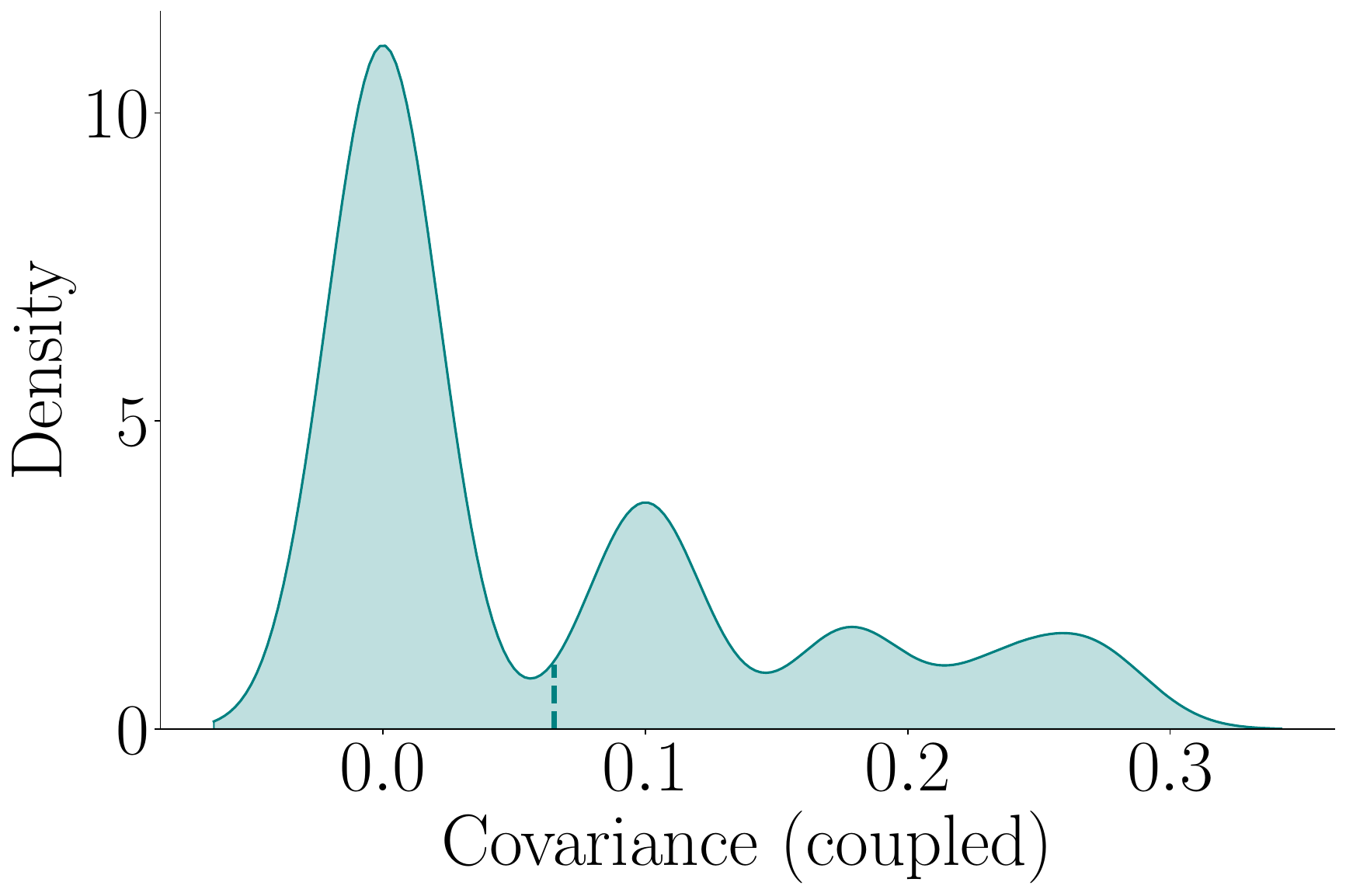} &
    \includegraphics[width=0.23\linewidth]{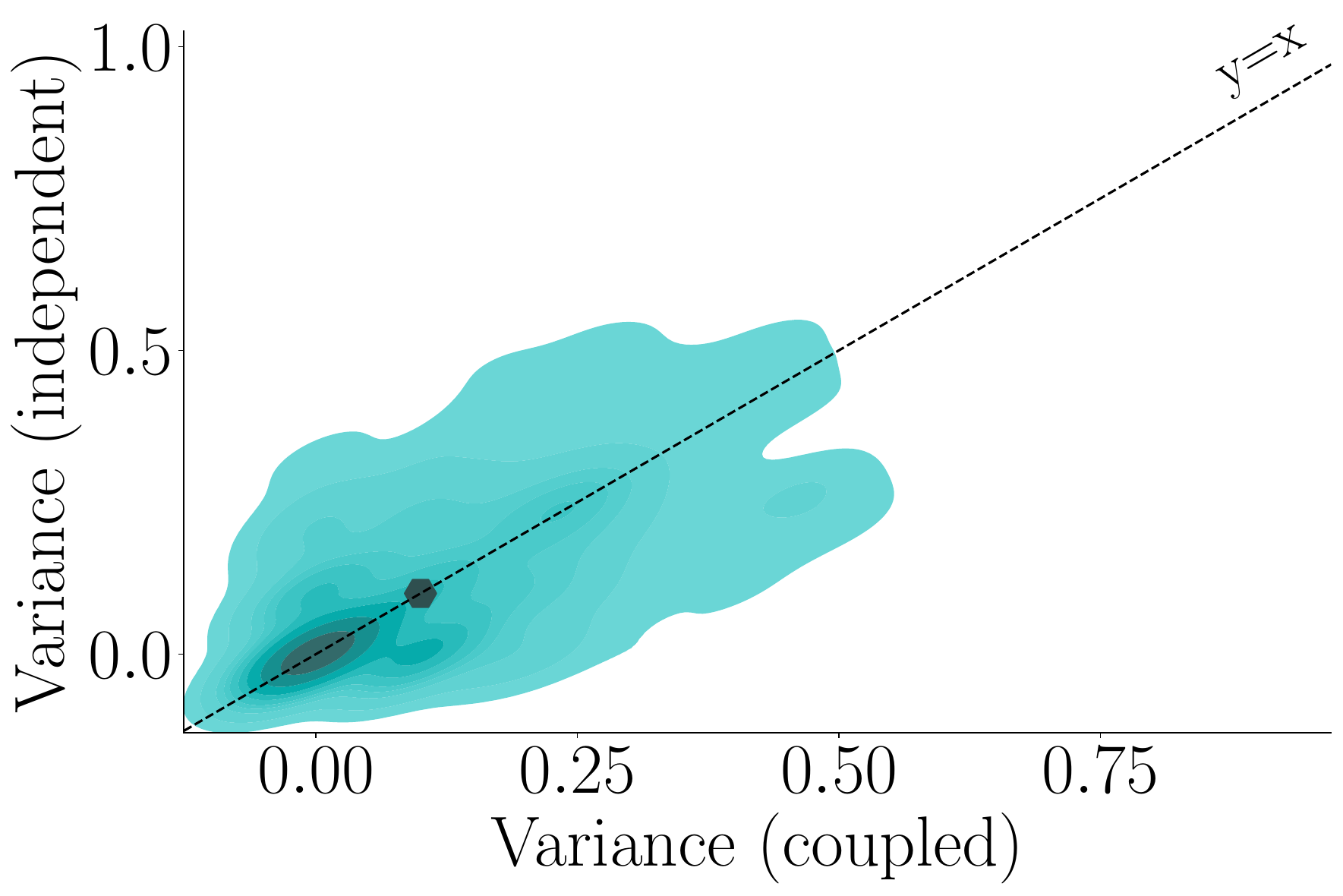} &
    \includegraphics[width=0.23\linewidth]{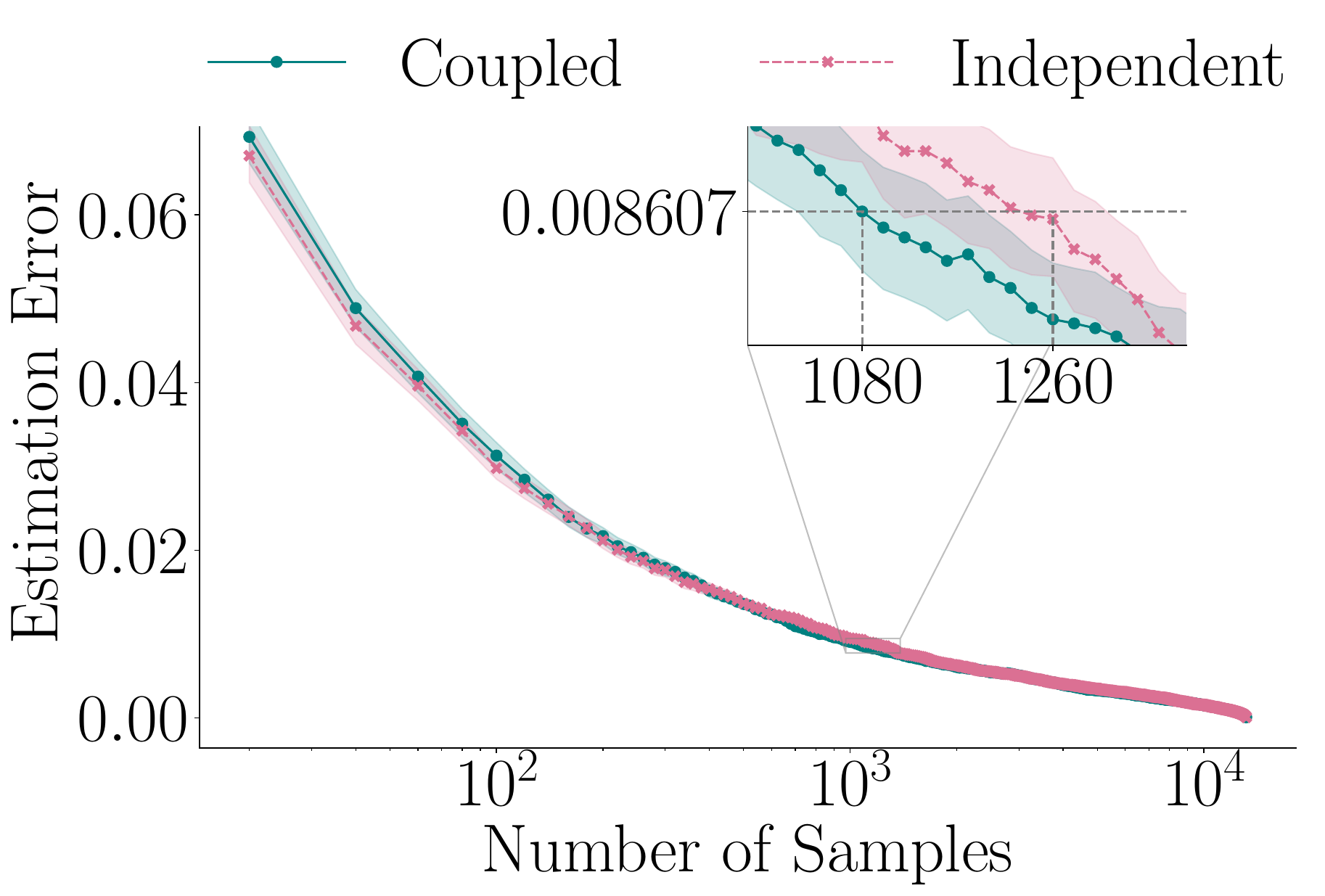} \\ \\
    \multicolumn{3}{c}{\texttt{v0.2} vs. \texttt{v0.3-bnb-4bit}}\\
    \includegraphics[width=0.23\linewidth]{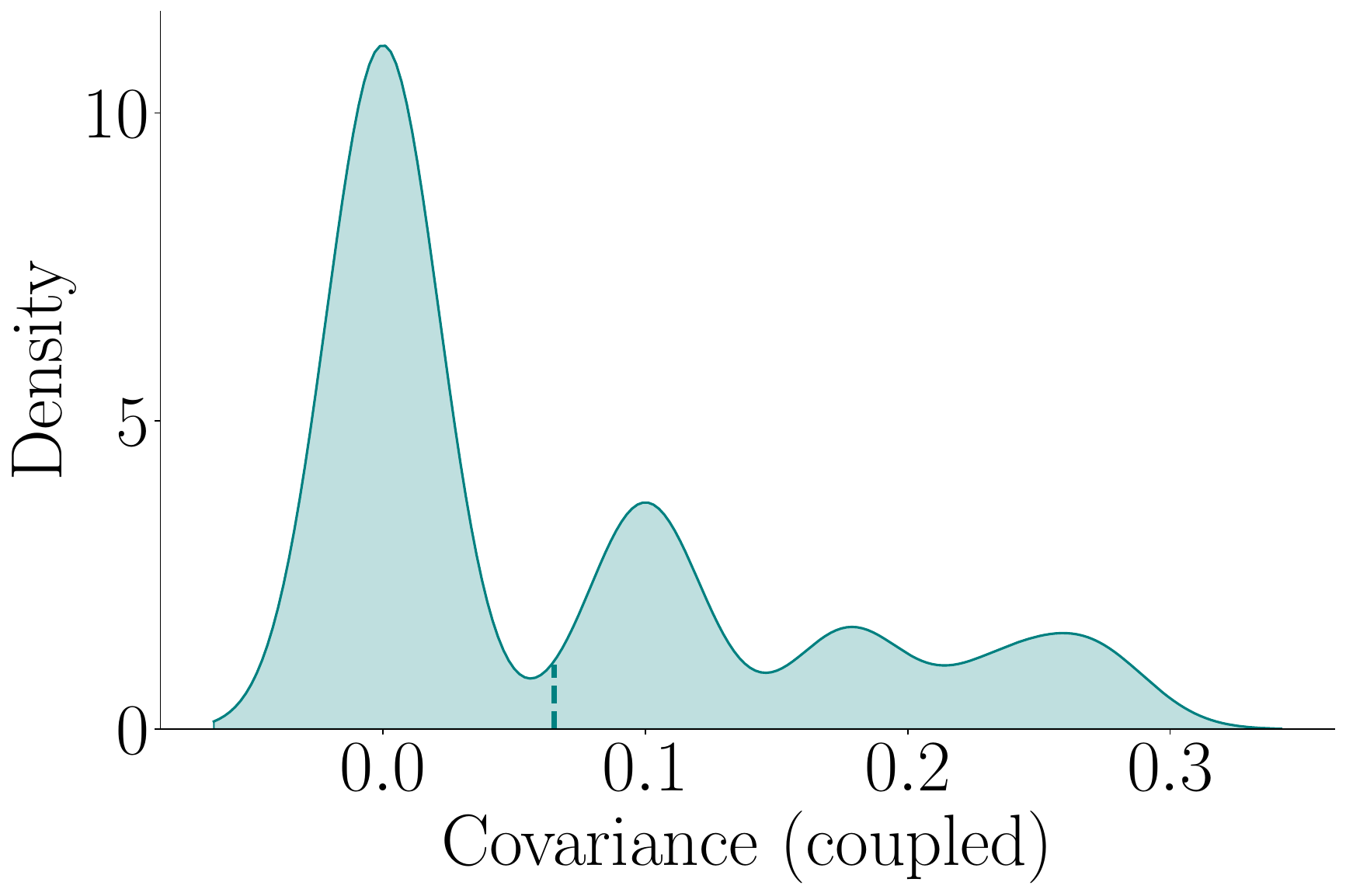} &
    \includegraphics[width=0.23\linewidth]{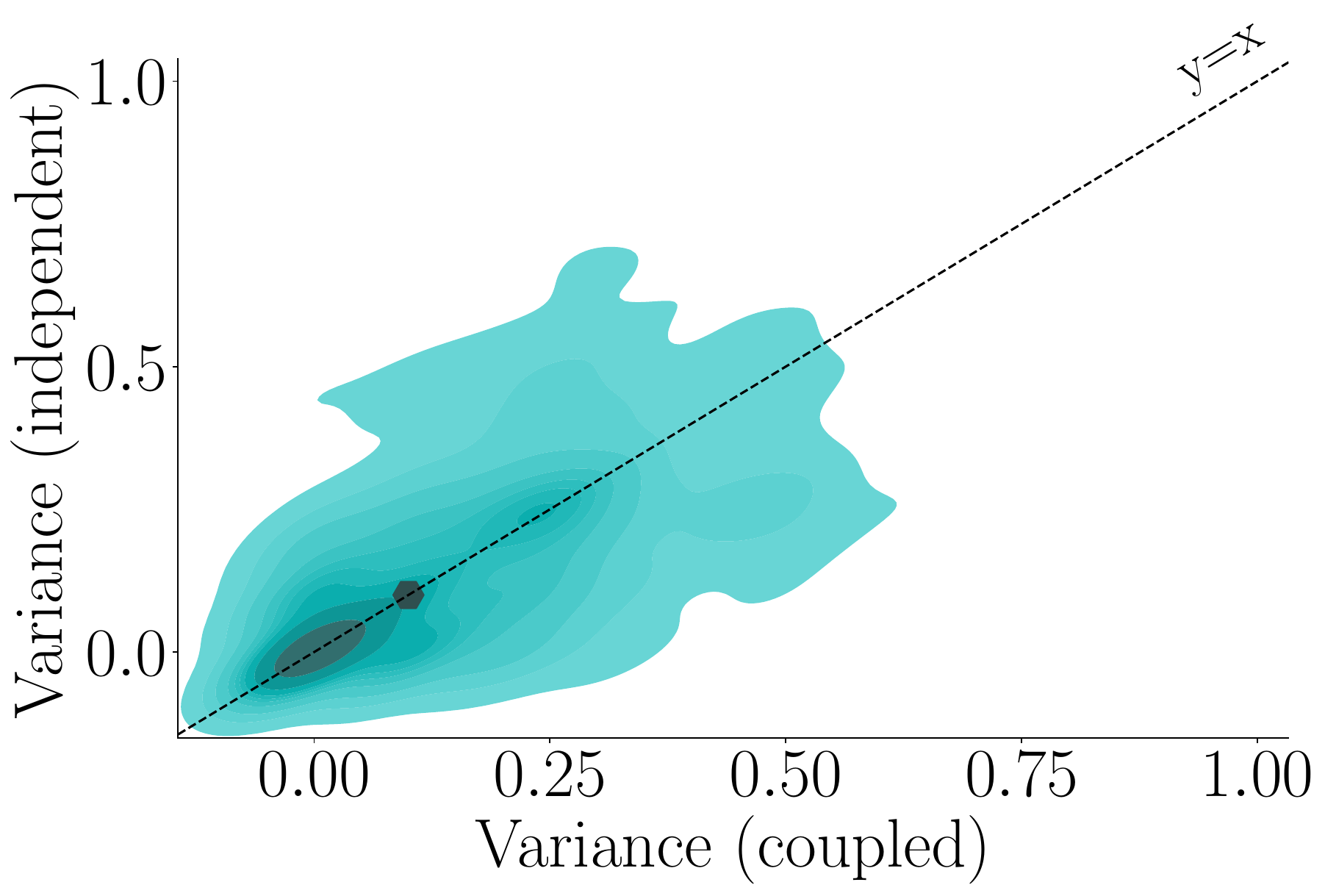} &
    \includegraphics[width=0.23\linewidth]{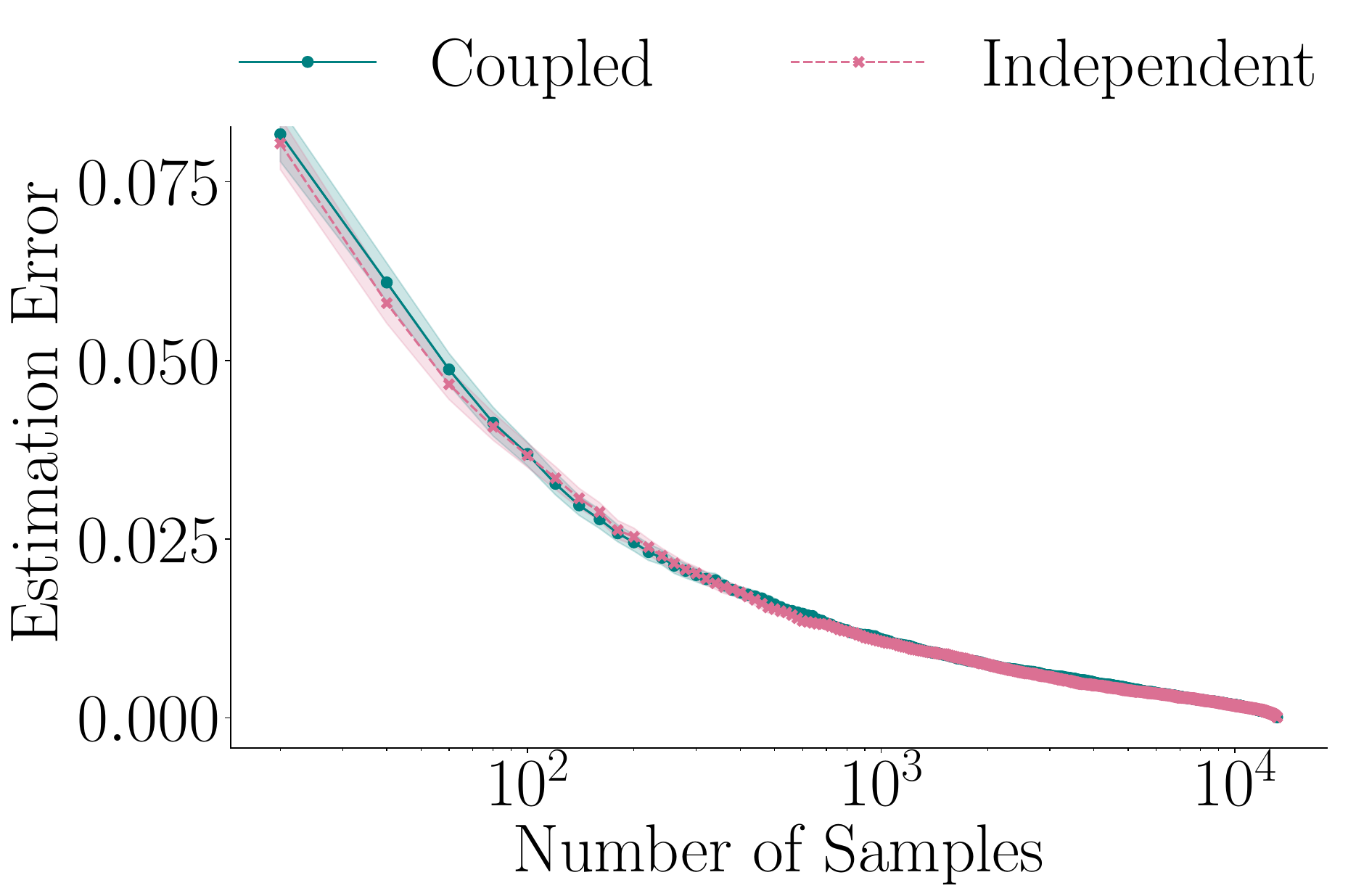} \\ \\
     \multicolumn{3}{c}{\texttt{v0.2} vs. \texttt{v0.1}}\\
    \includegraphics[width=0.23\linewidth]{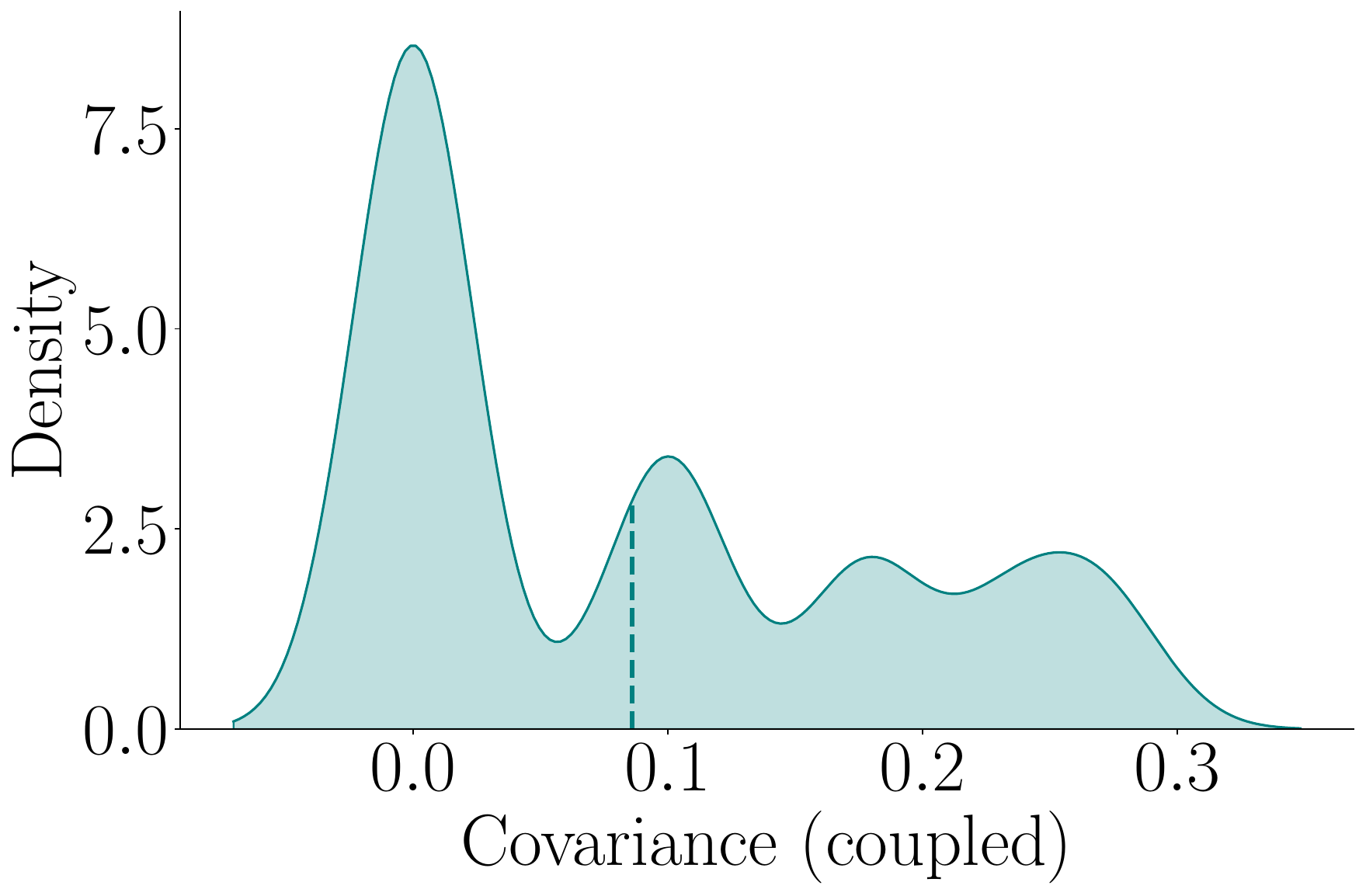} &
    \includegraphics[width=0.23\linewidth]{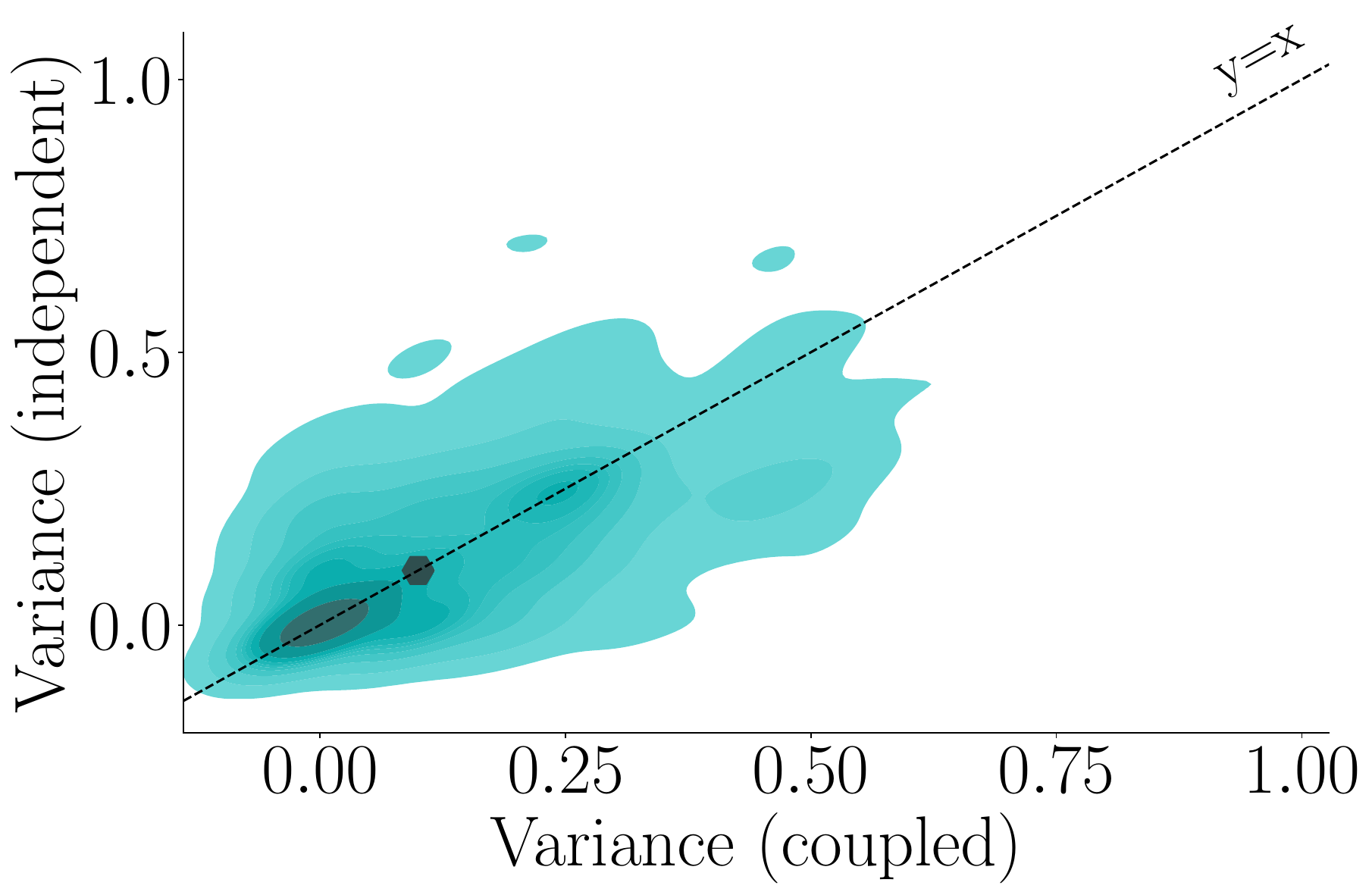} &
    \includegraphics[width=0.23\linewidth]{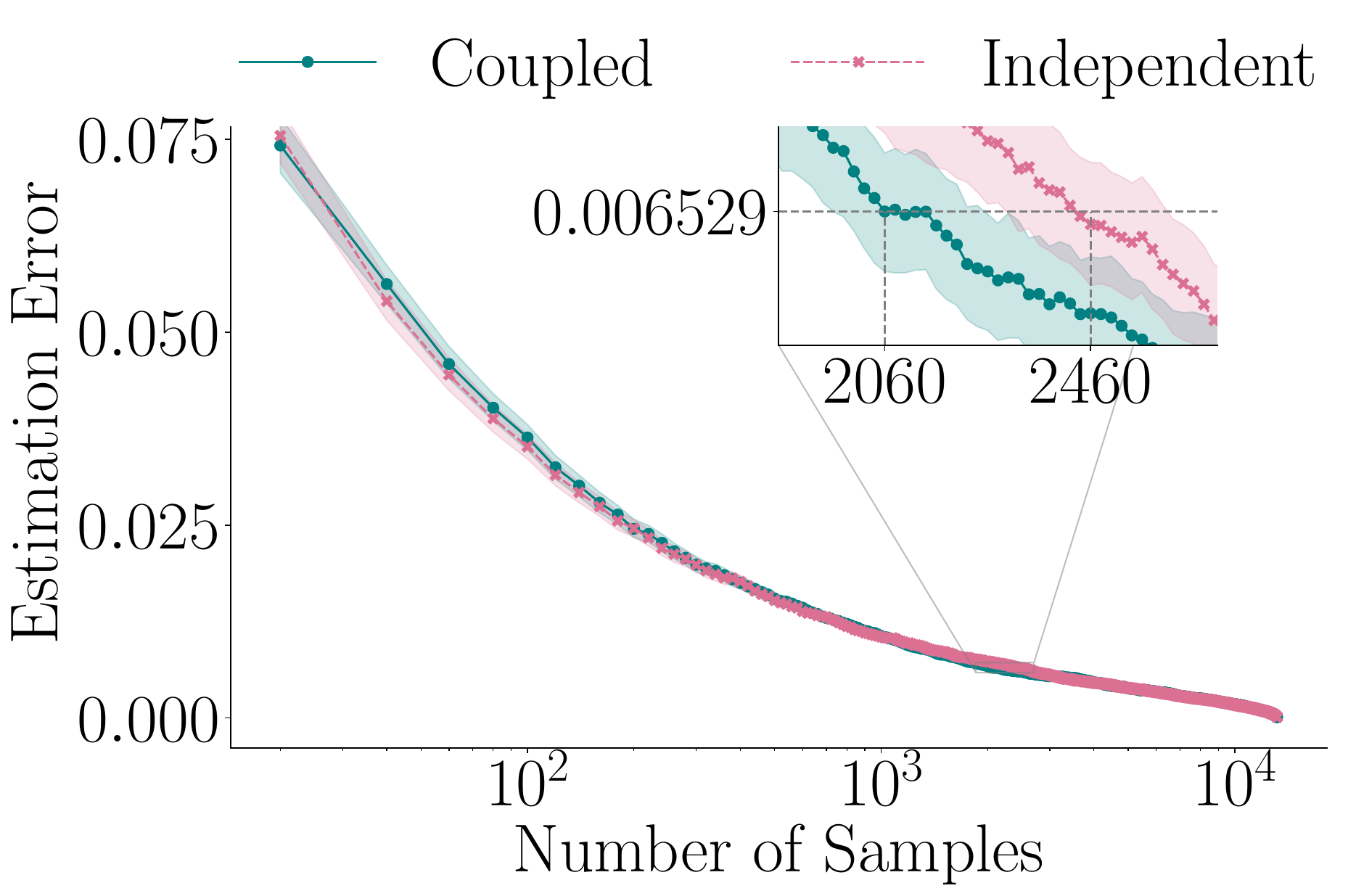} \\ \\
    (a) Score covariance & (b) Variance of the score difference & (c) Estimation error vs. \# samples \\ 
    
\end{tabular}
    \caption{\textbf{Comparison between several pairs of LLMs in the \texttt{Mistral} family on math questions from the GSM8K dataset.}
    Panels in column (a) show the kernel density estimate (KDE) of the covariance between the scores of the two LLMs on each problem under coupled generation; the dashed lines correspond to average values. Panels in column (b) show the KDE of the variance of the difference between the scores of the LLMs on each question under coupled and independent generation; the highlighted points correspond to median values. Panels in column (c) show the absolute error in the estimation of the expected difference between the scores of the LLMs against the number of samples; for each point on the x-axis, we perform $1{,}000$ sub-samplings and shaded areas correspond to $95\%$ confidence intervals.}
    \label{fig:gsm8k-mistral-last}
\end{figure}

\begin{figure}[!!h]
\centering
\begin{tabular}{c c c}
     \multicolumn{3}{c}{\texttt{v0.3} vs. \texttt{v0.3-bnb-8bit}}\\
    \includegraphics[width=0.23\linewidth]{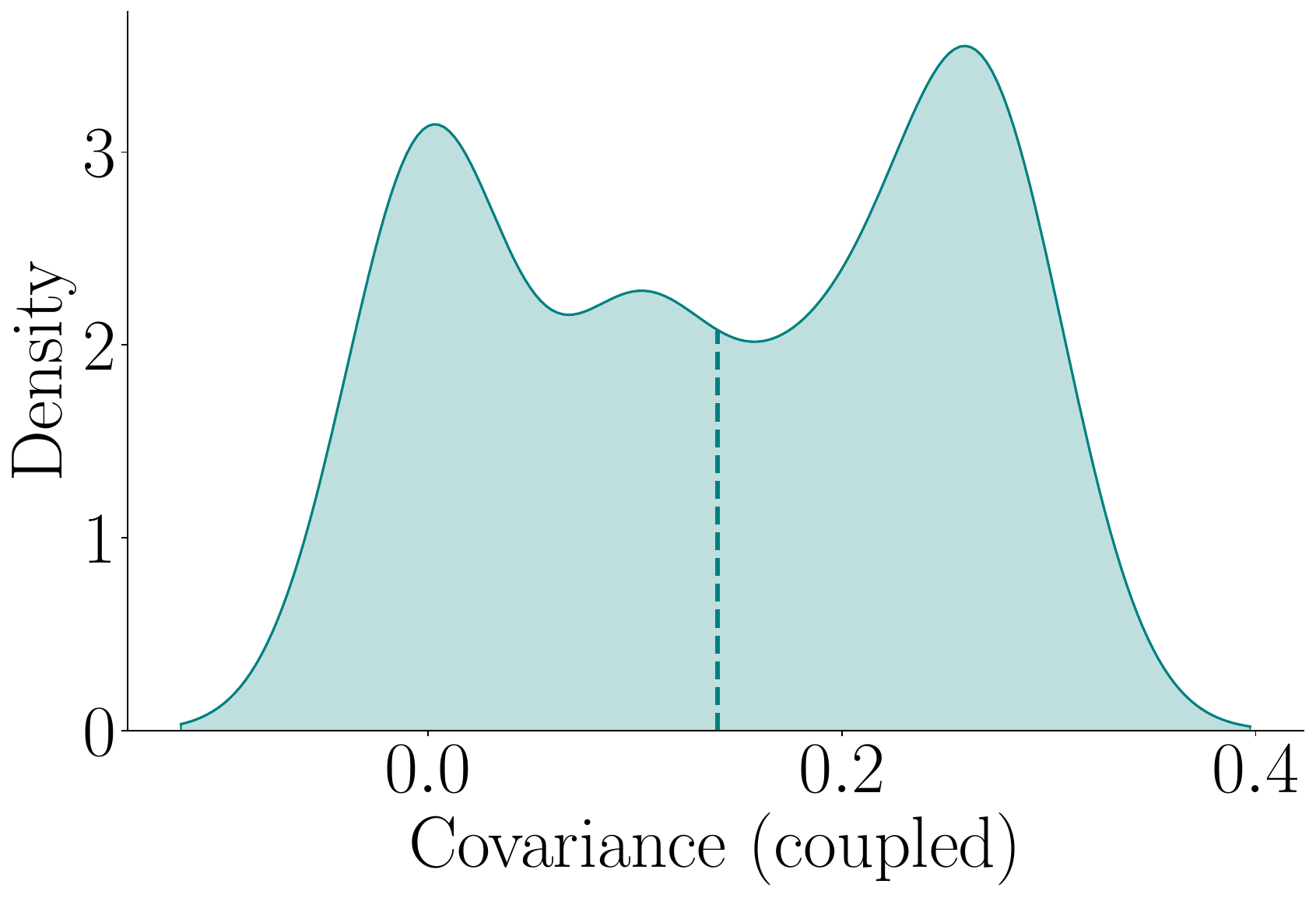} &
    \includegraphics[width=0.23\linewidth]{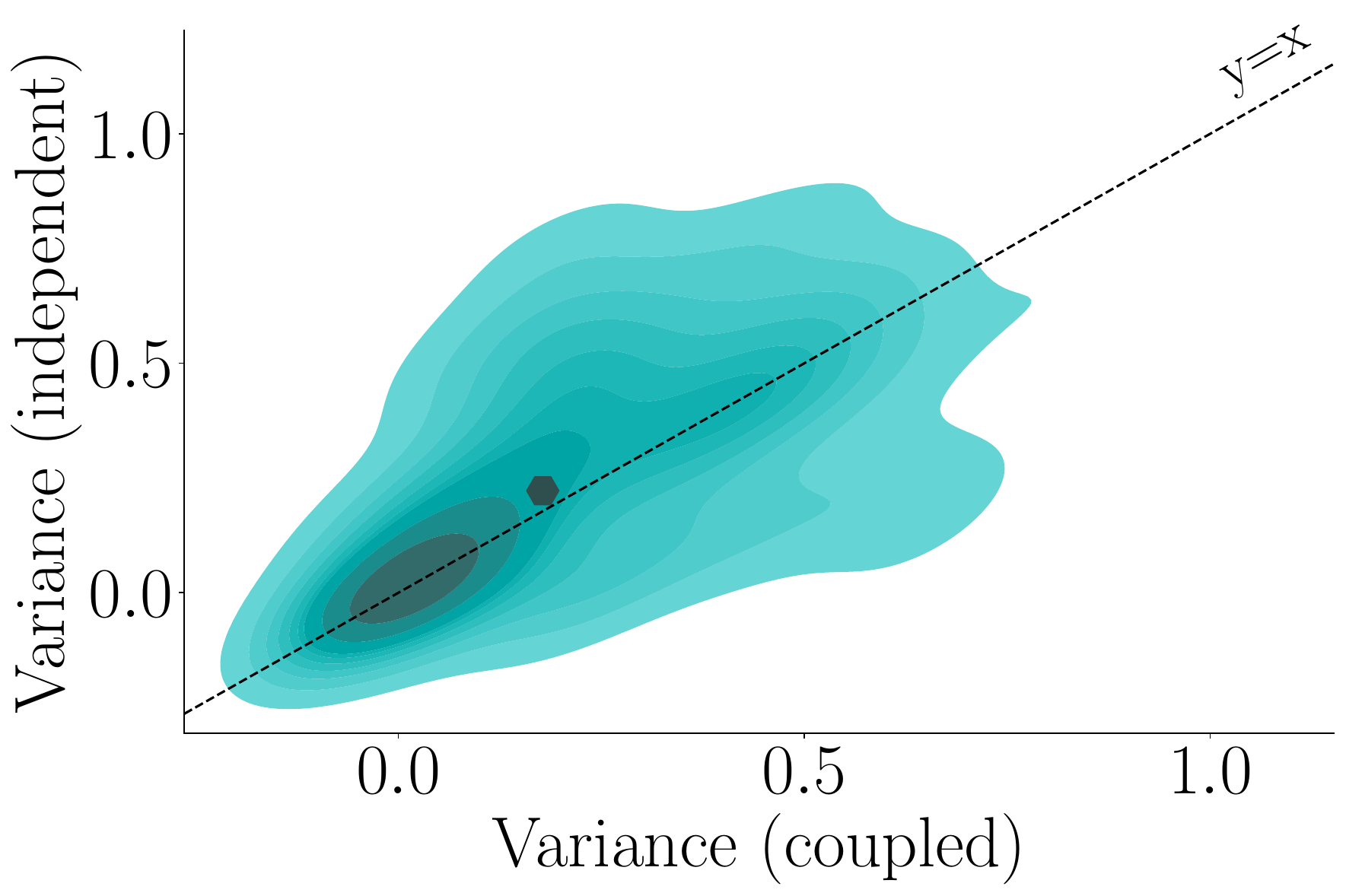} &
    \includegraphics[width=0.23\linewidth]{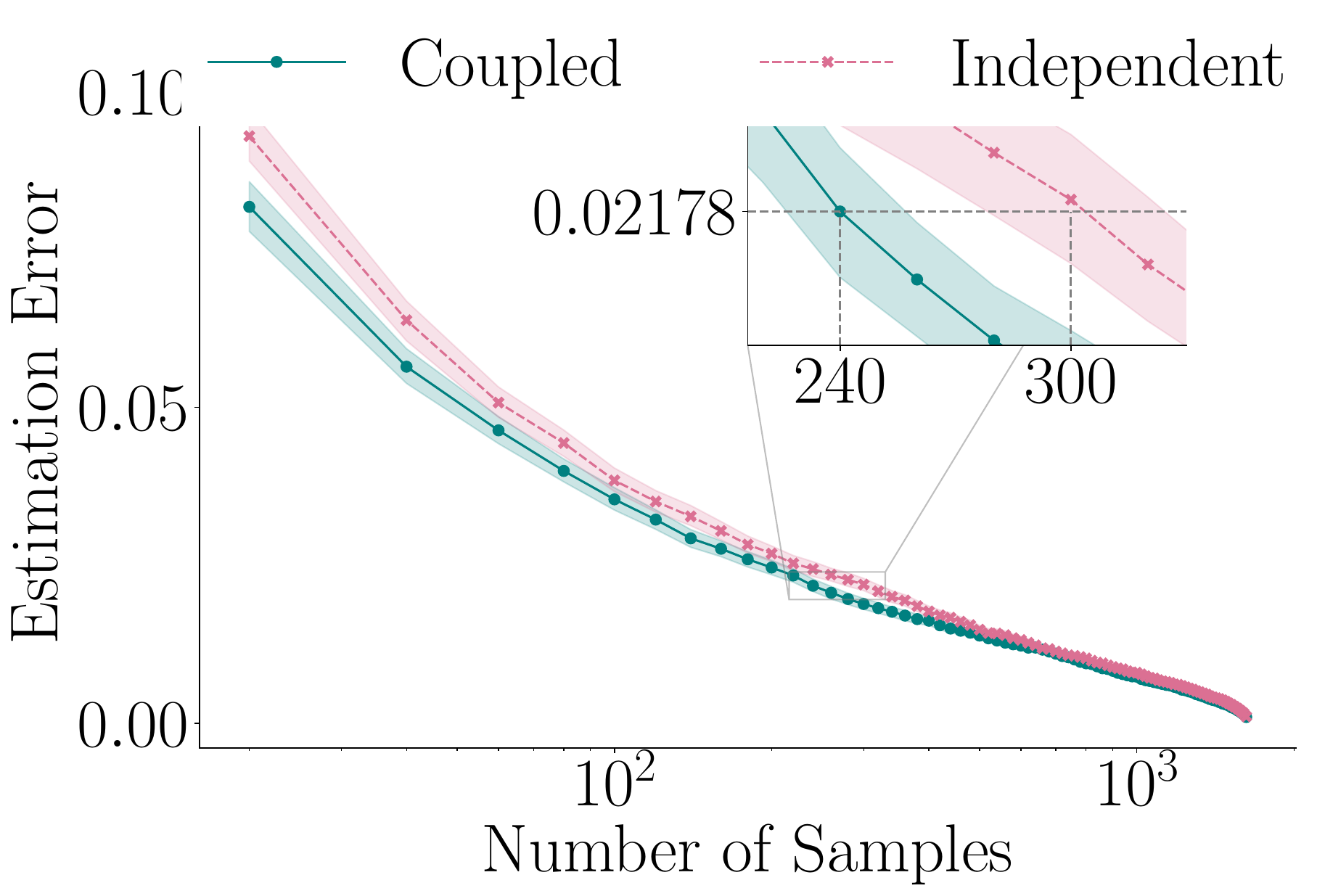} \\ \\
%
    \multicolumn{3}{c}{\texttt{v0.3} vs. \texttt{v0.3-bnb-4bit}}\\
    \includegraphics[width=0.23\linewidth]{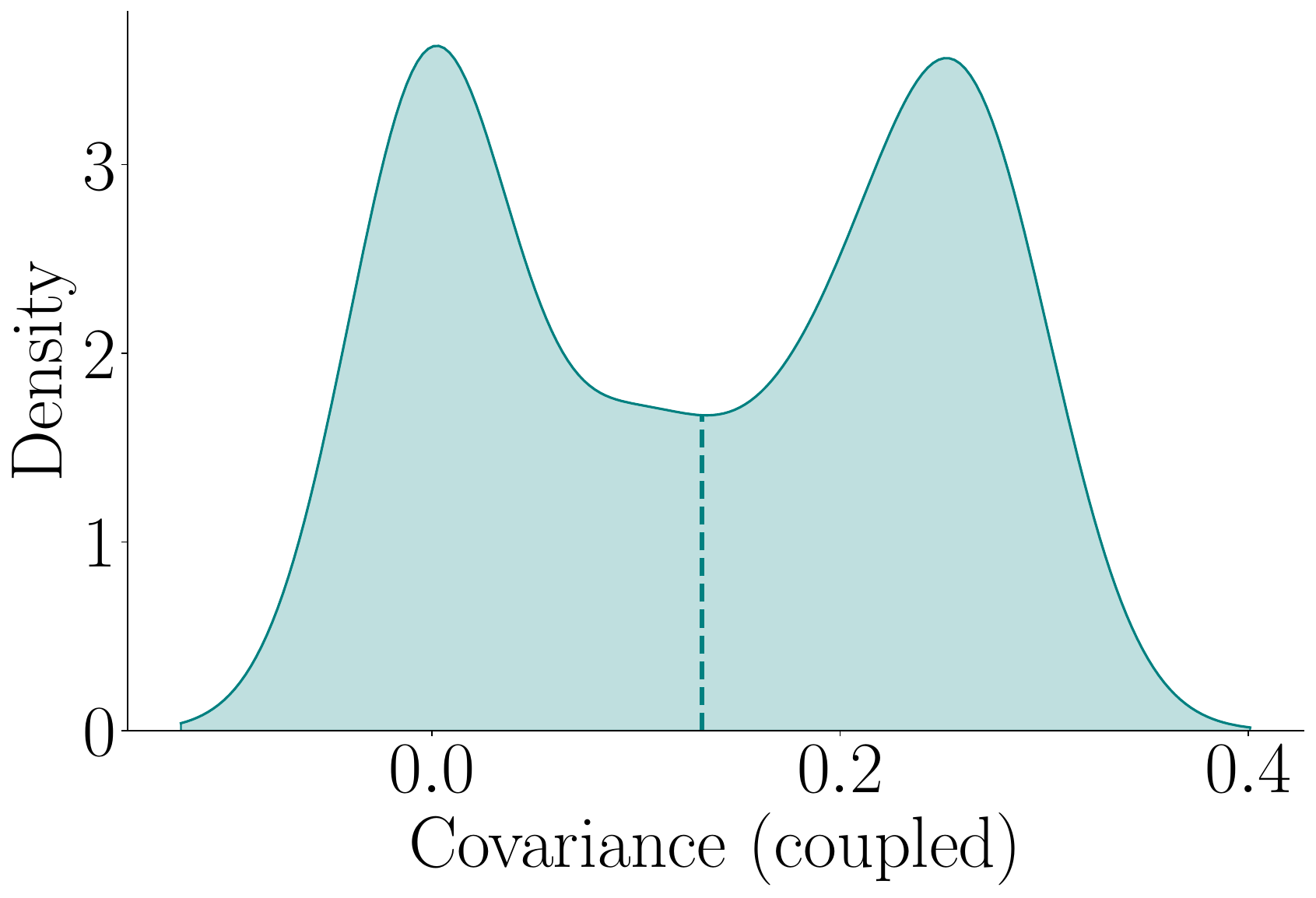} &
    \includegraphics[width=0.23\linewidth]{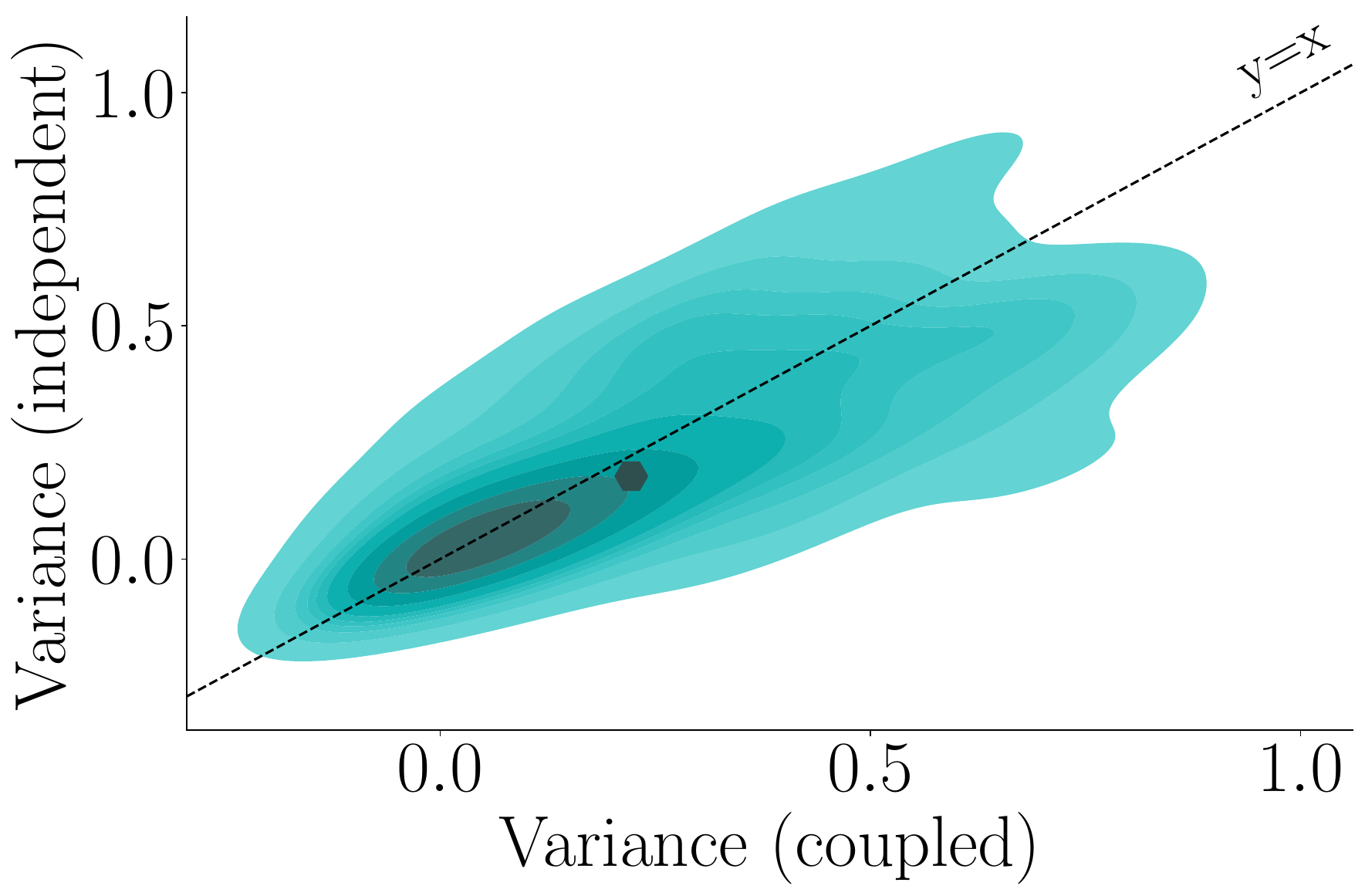} &
    \includegraphics[width=0.23\linewidth]{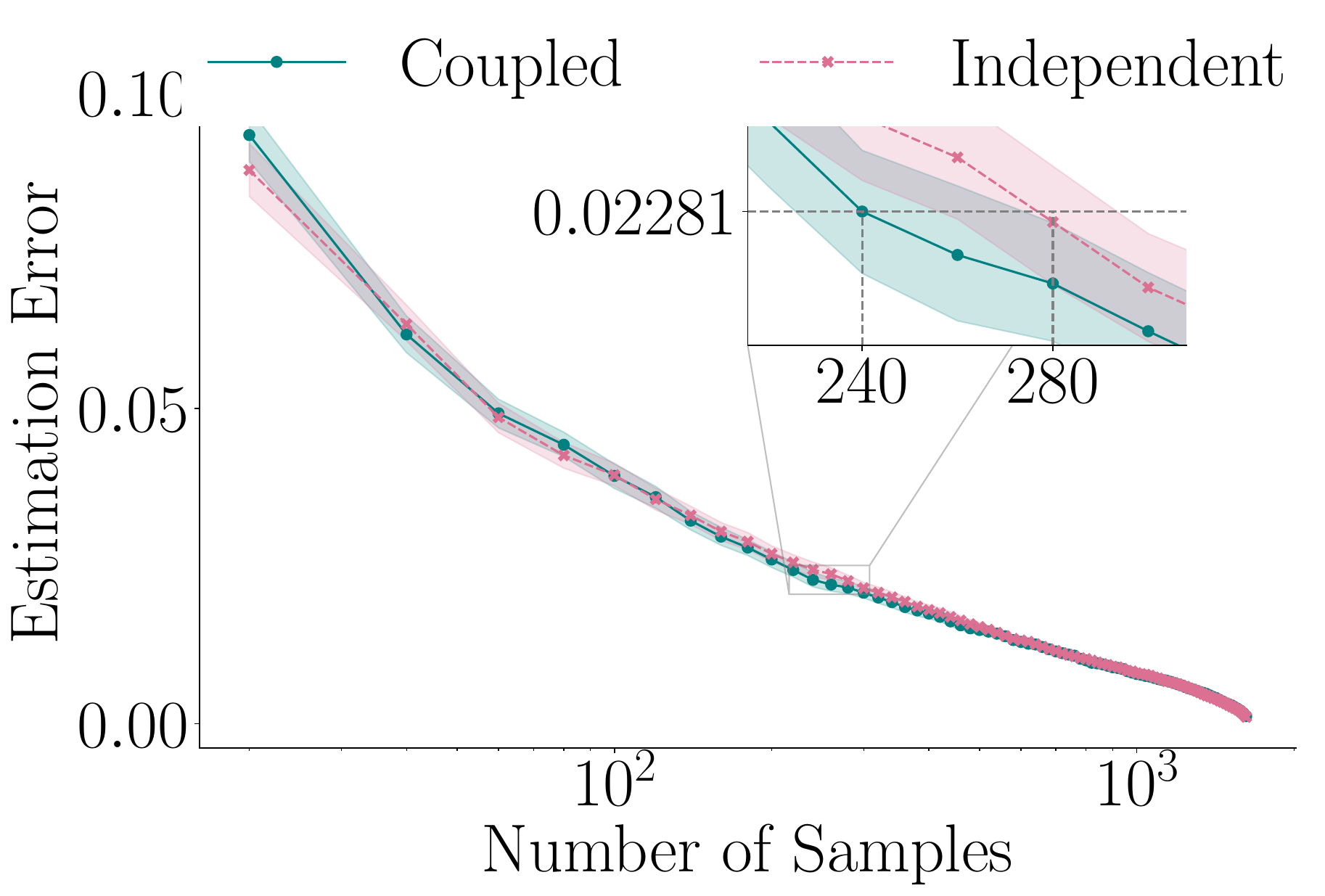} \\ \\
%
     \multicolumn{3}{c}{\texttt{v0.3-bnb-8bit} vs. \texttt{v0.3-bnb-4bit}}\\
    \includegraphics[width=0.23\linewidth]{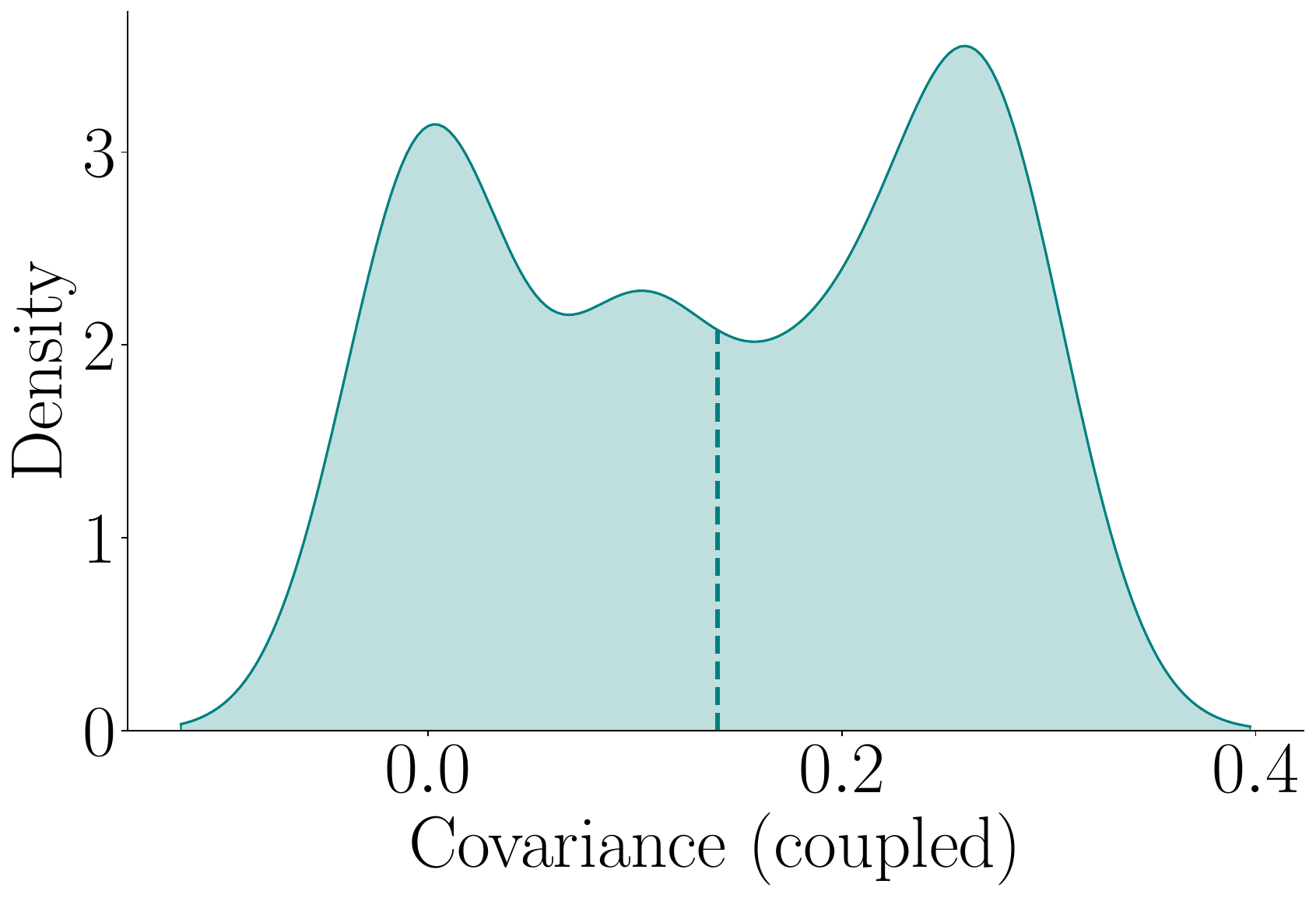} &
    \includegraphics[width=0.23\linewidth]{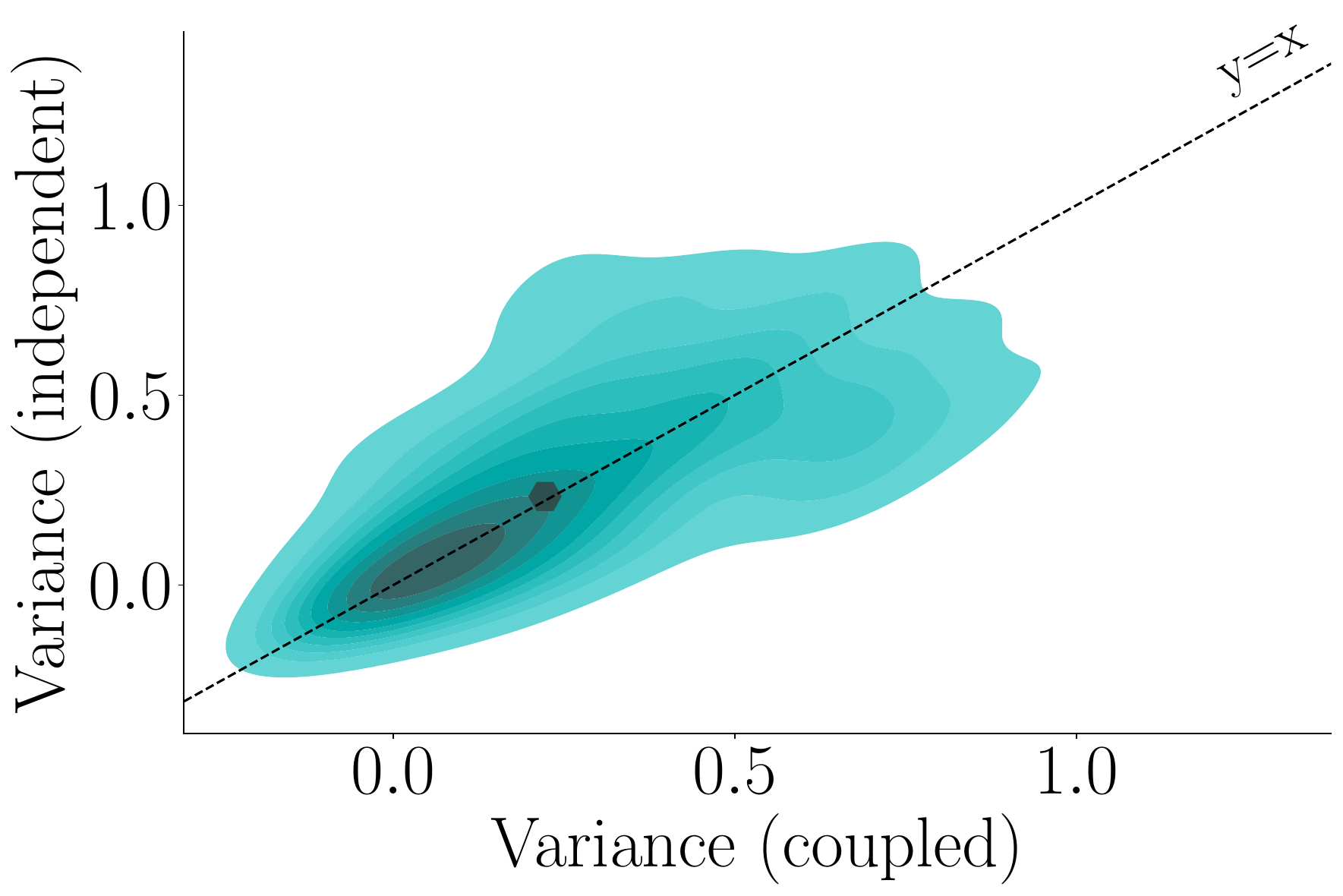} &
    \includegraphics[width=0.23\linewidth]{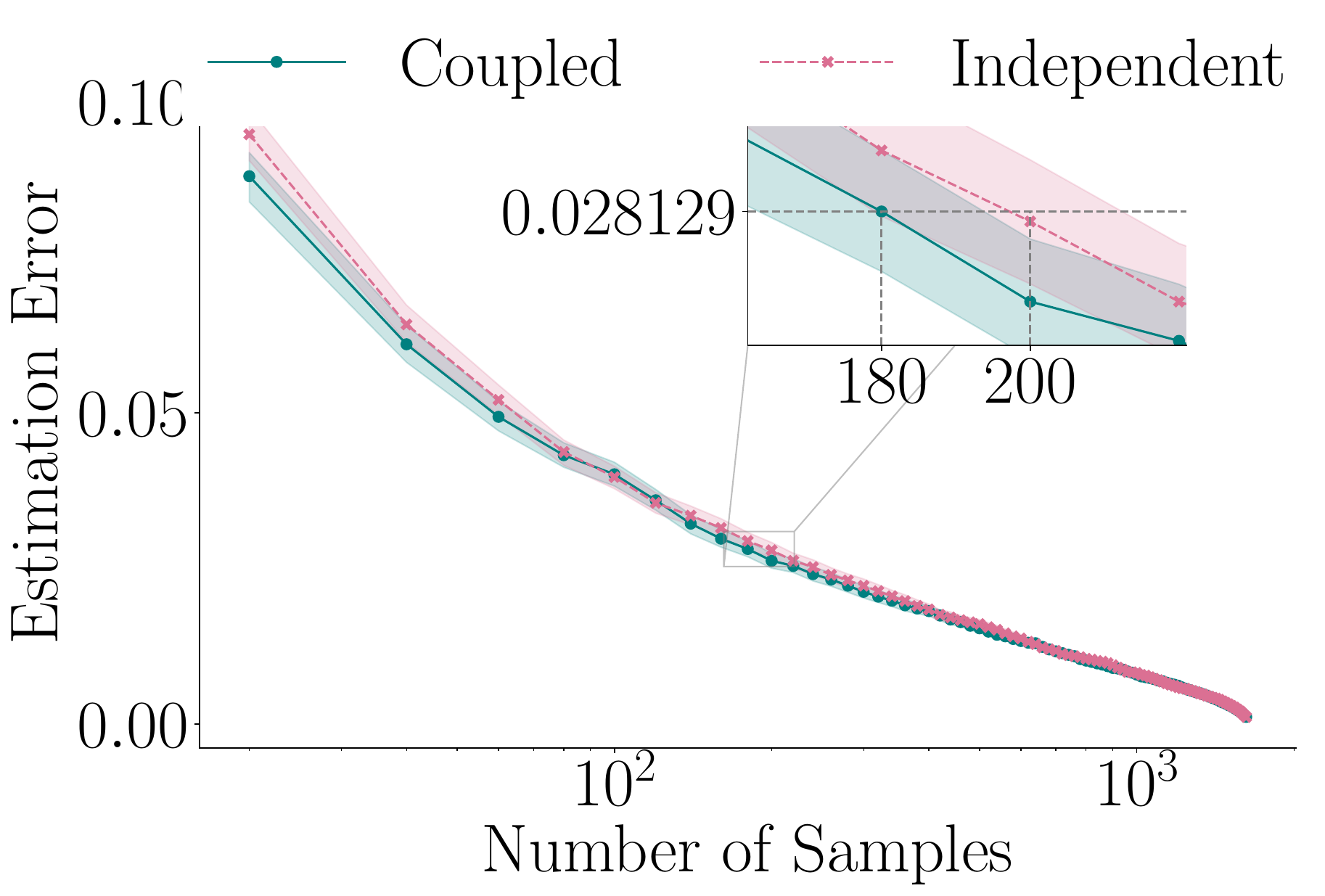} \\ \\
%
  \multicolumn{3}{c}{\texttt{v0.3} vs. \texttt{v0.2}}\\
    \includegraphics[width=0.23\linewidth]{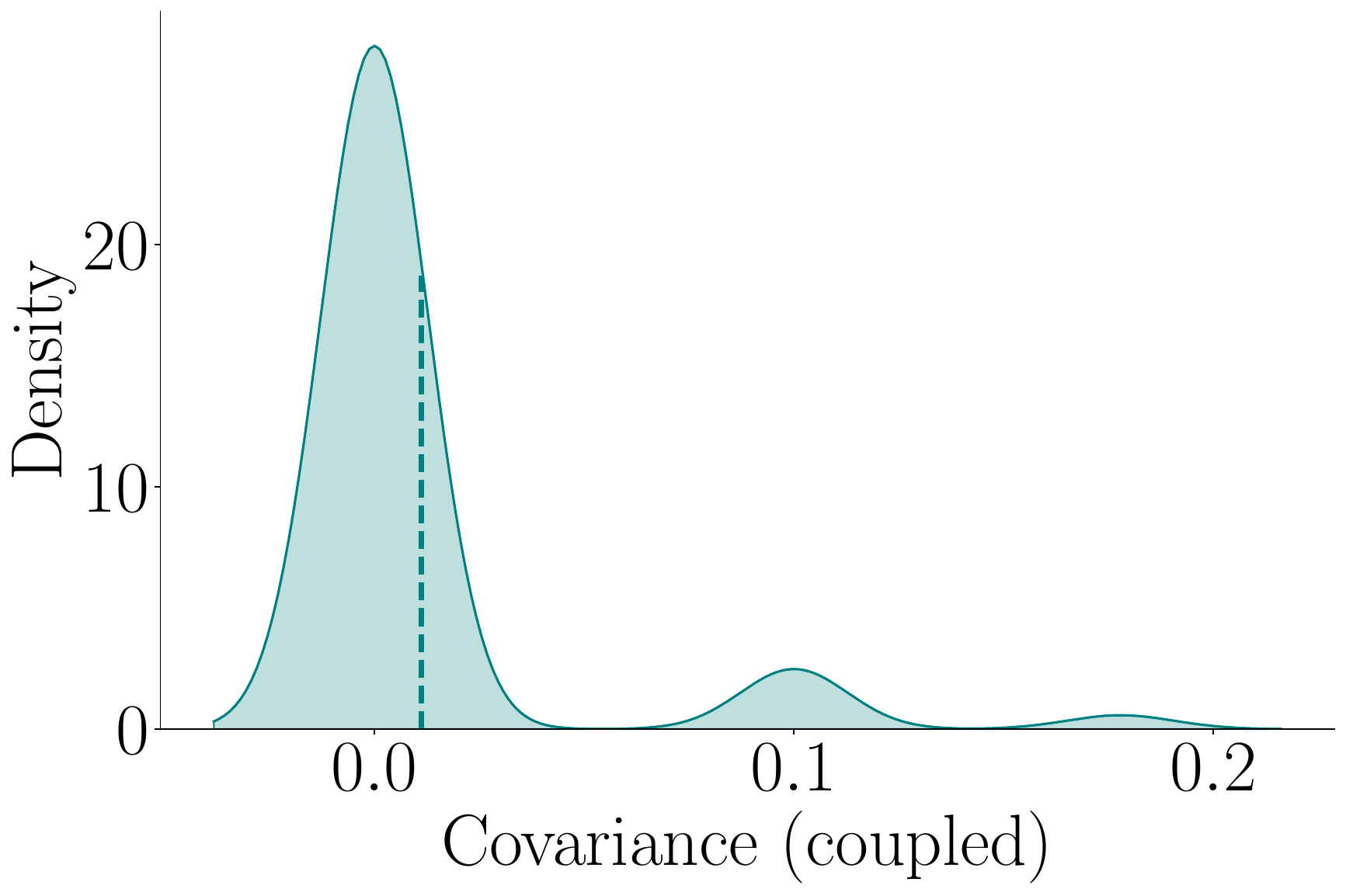} &
    \includegraphics[width=0.23\linewidth]{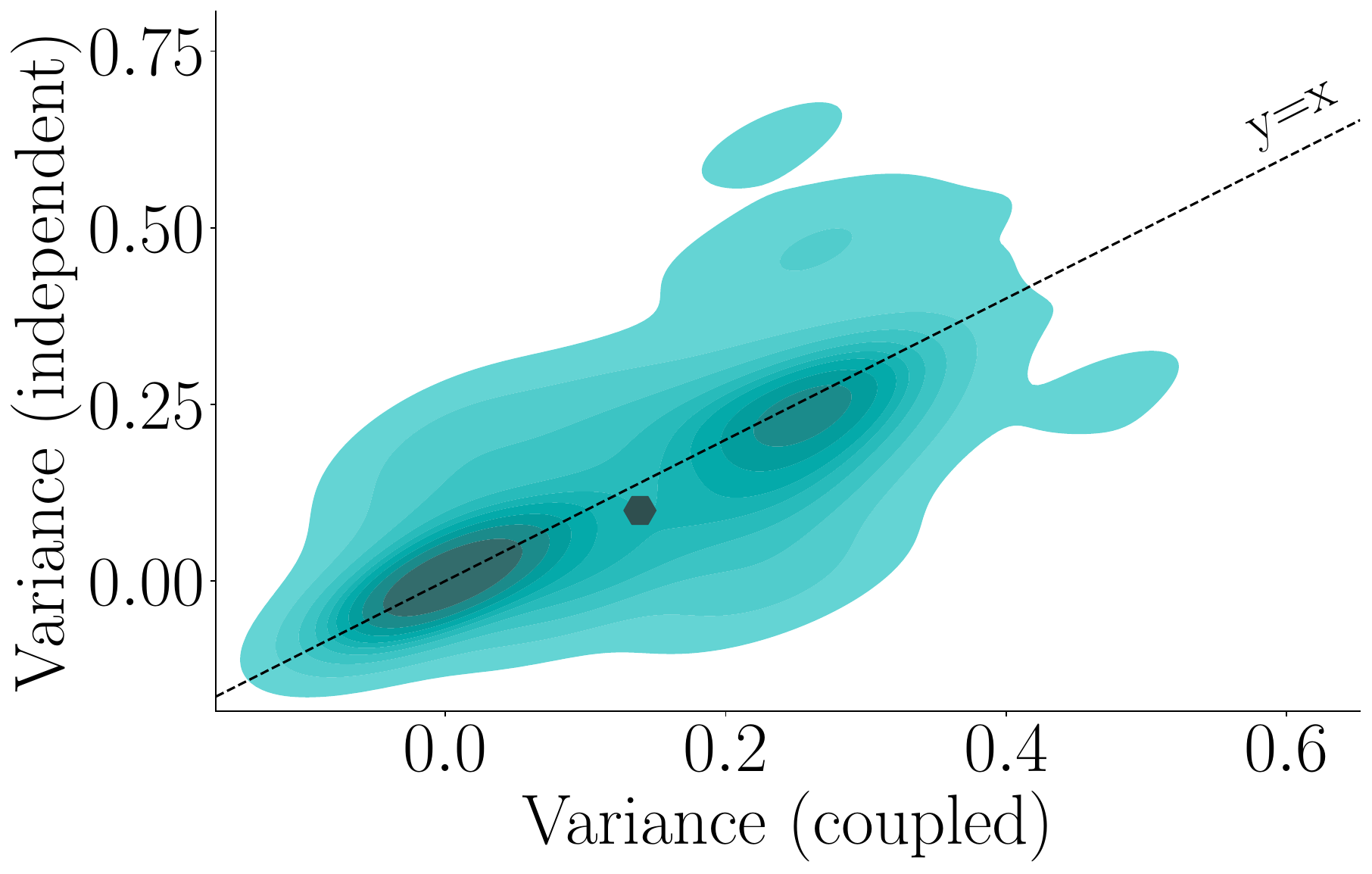} &
    \includegraphics[width=0.23\linewidth]{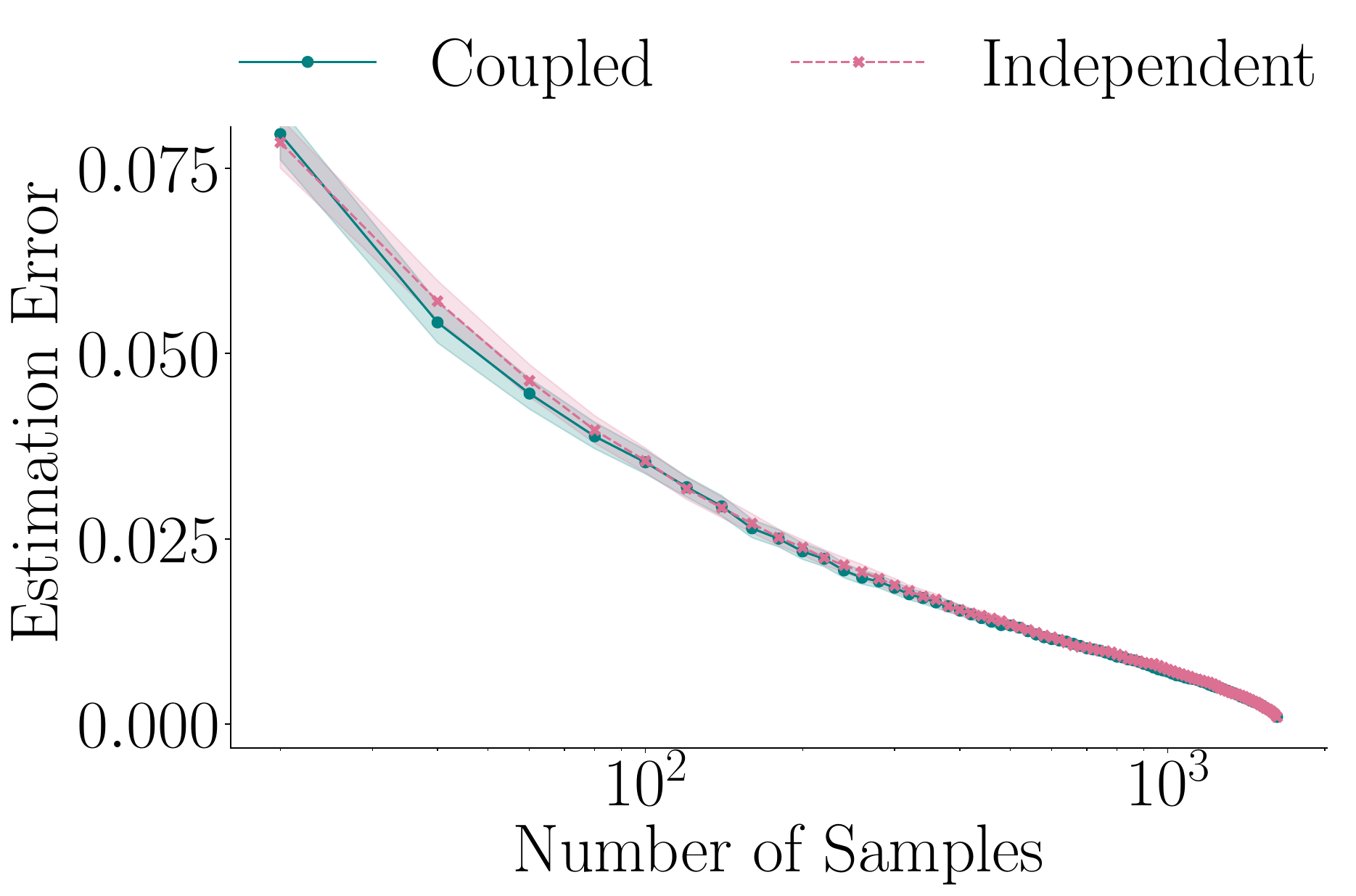} \\ \\

 \multicolumn{3}{c}{\texttt{v0.3} vs. \texttt{v0.1}}\\
    \includegraphics[width=0.23\linewidth]{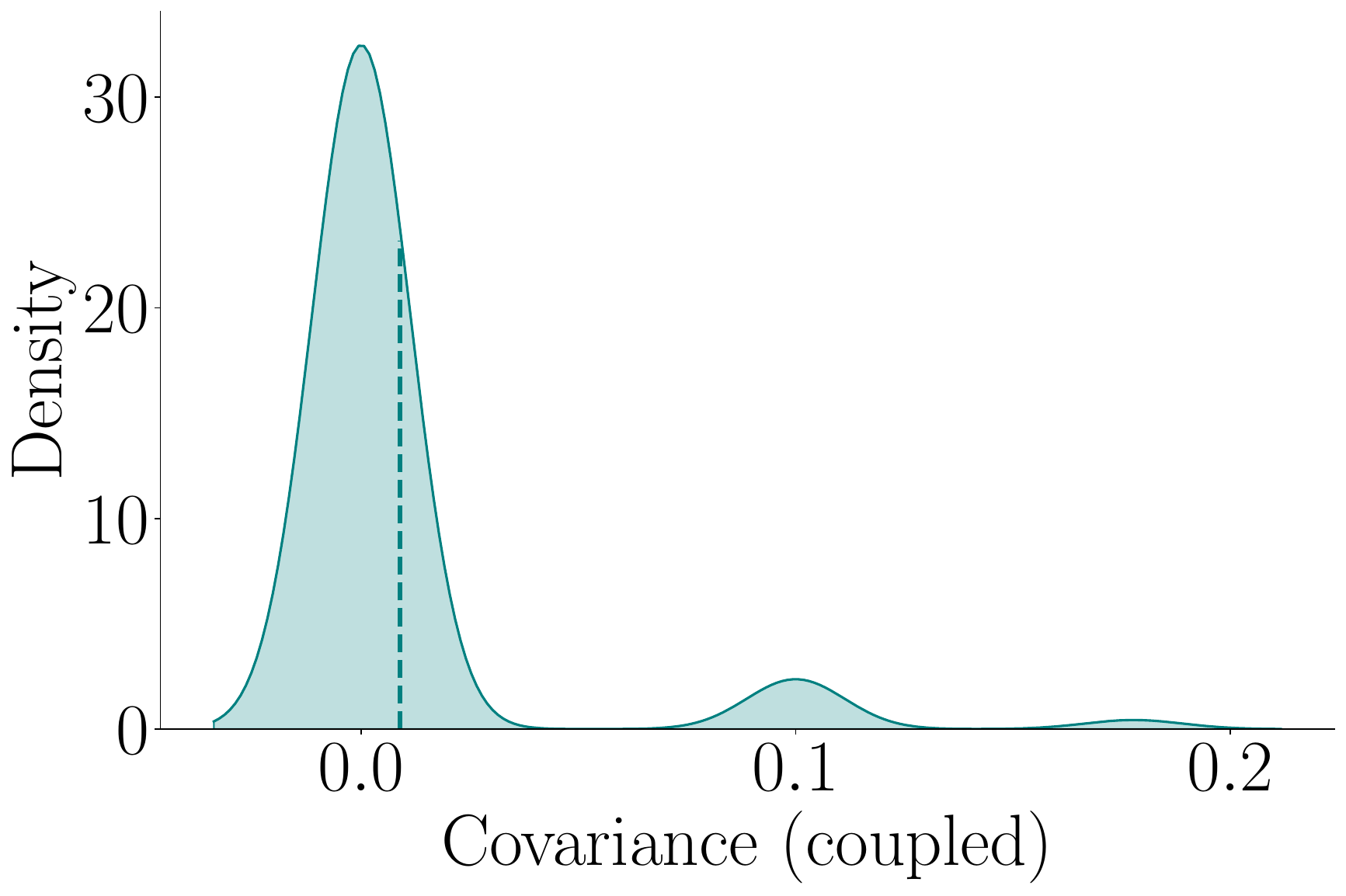} &
    \includegraphics[width=0.23\linewidth]{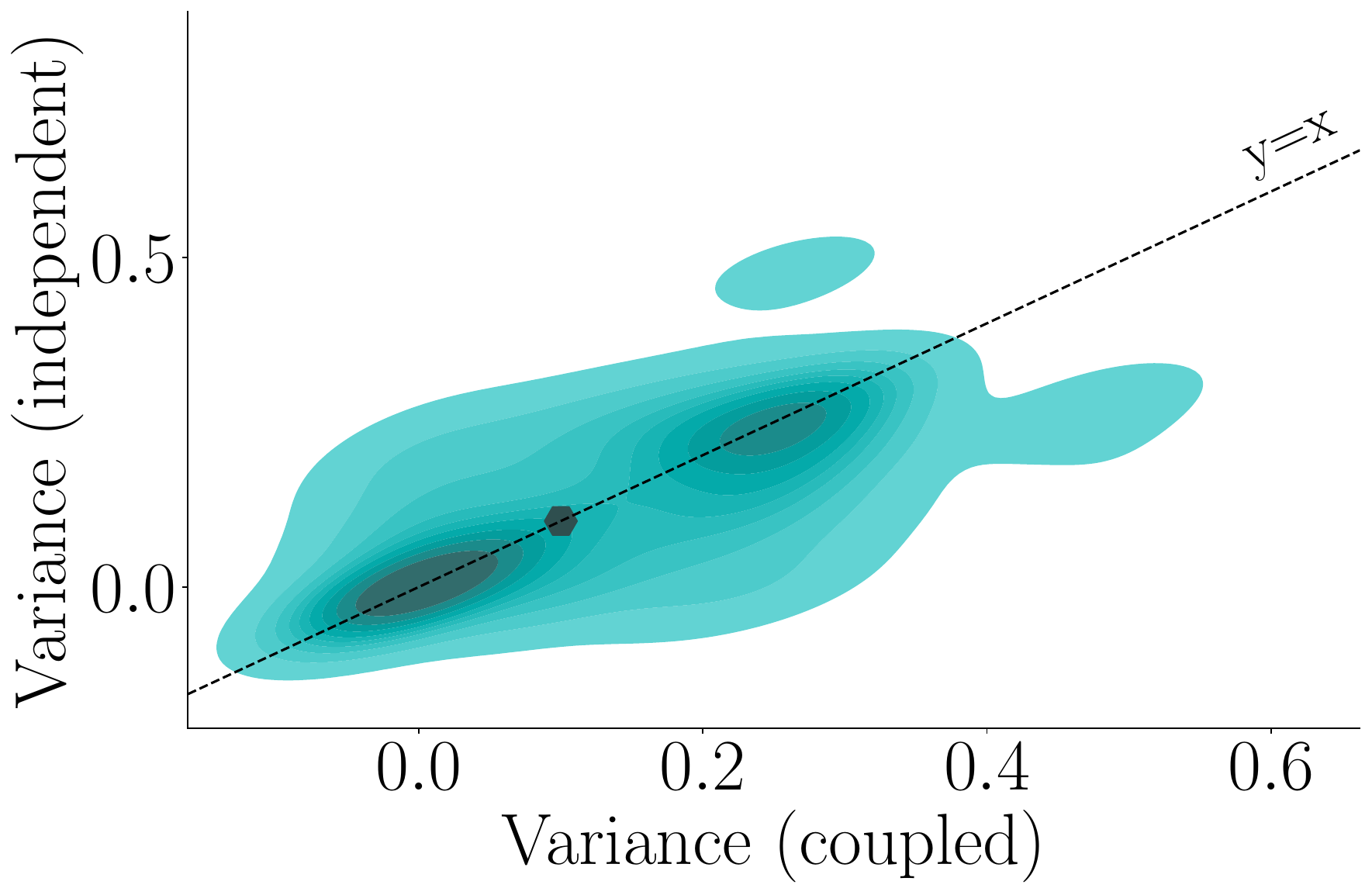} &
    \includegraphics[width=0.23\linewidth]{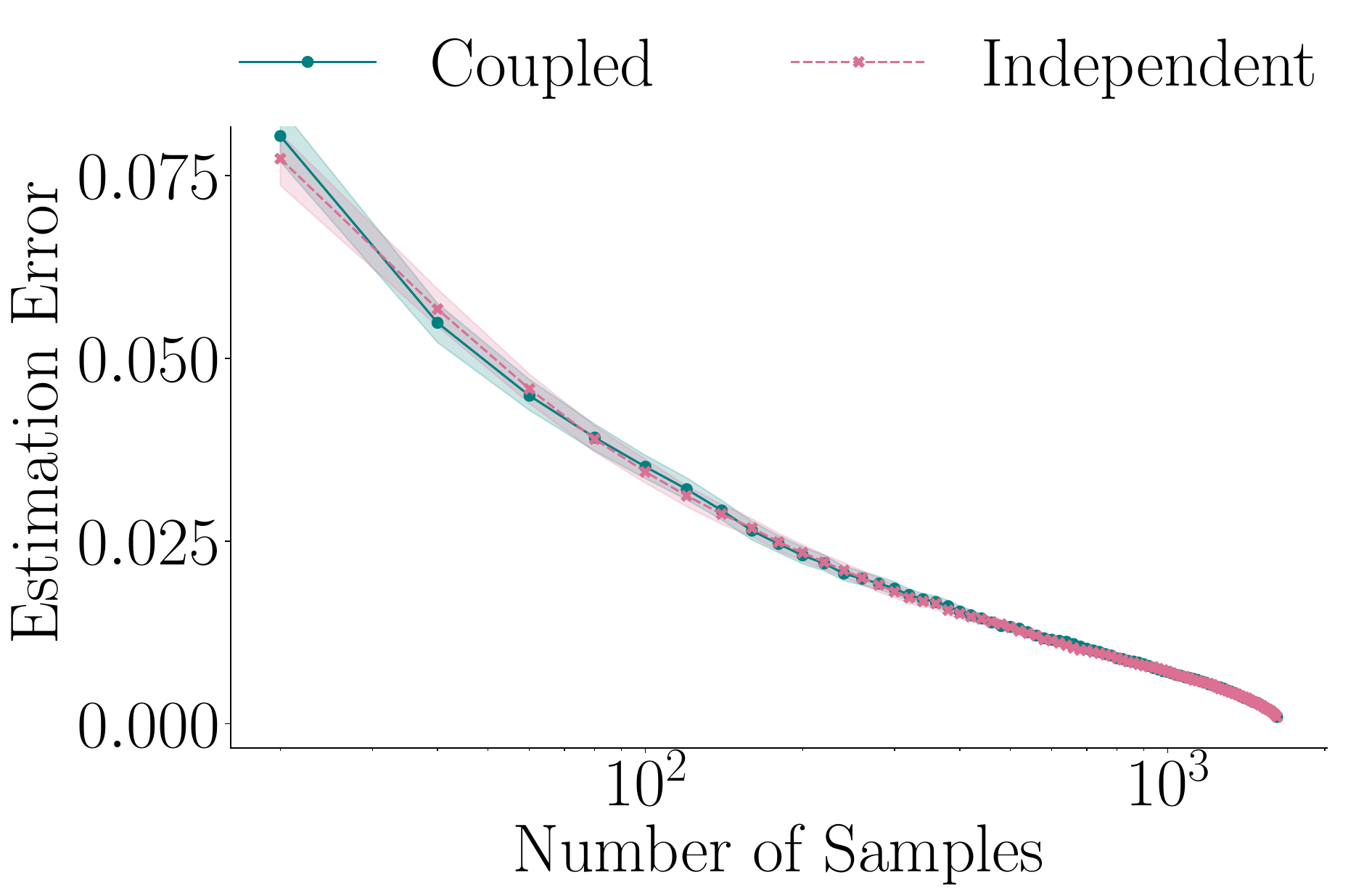} \\ \\
    (a) Score covariance & (b) Variance of the score difference & (c) Estimation error vs. \# samples \\ 
\end{tabular}
    \caption{\textbf{Comparison between several pairs of LLMs in the \texttt{Mistral} family on programming problems from the HumanEval dataset.}
    Panels in column (a) show the kernel density estimate (KDE) of the covariance between the scores of the two LLMs on each problem under coupled generation; the dashed lines correspond to average values. Panels in column (b) show the KDE of the variance of the difference between the scores of the LLMs on each question under coupled and independent generation; the highlighted points correspond to median values. Panels in column (c) show the absolute error in the estimation of the expected difference between the scores of the LLMs against the number of samples; for each point on the x-axis, we perform $1{,}000$ sub-samplings and shaded areas correspond to $95\%$ confidence intervals.}
    \label{fig:human-eval-mistral-first}
\end{figure}

\begin{figure}[h]
\centering
\begin{tabular}{c c c}
    \multicolumn{3}{c}{\texttt{v0.3-bnb-8bit} vs. \texttt{v0.2}}\\
    \includegraphics[width=0.23\linewidth]{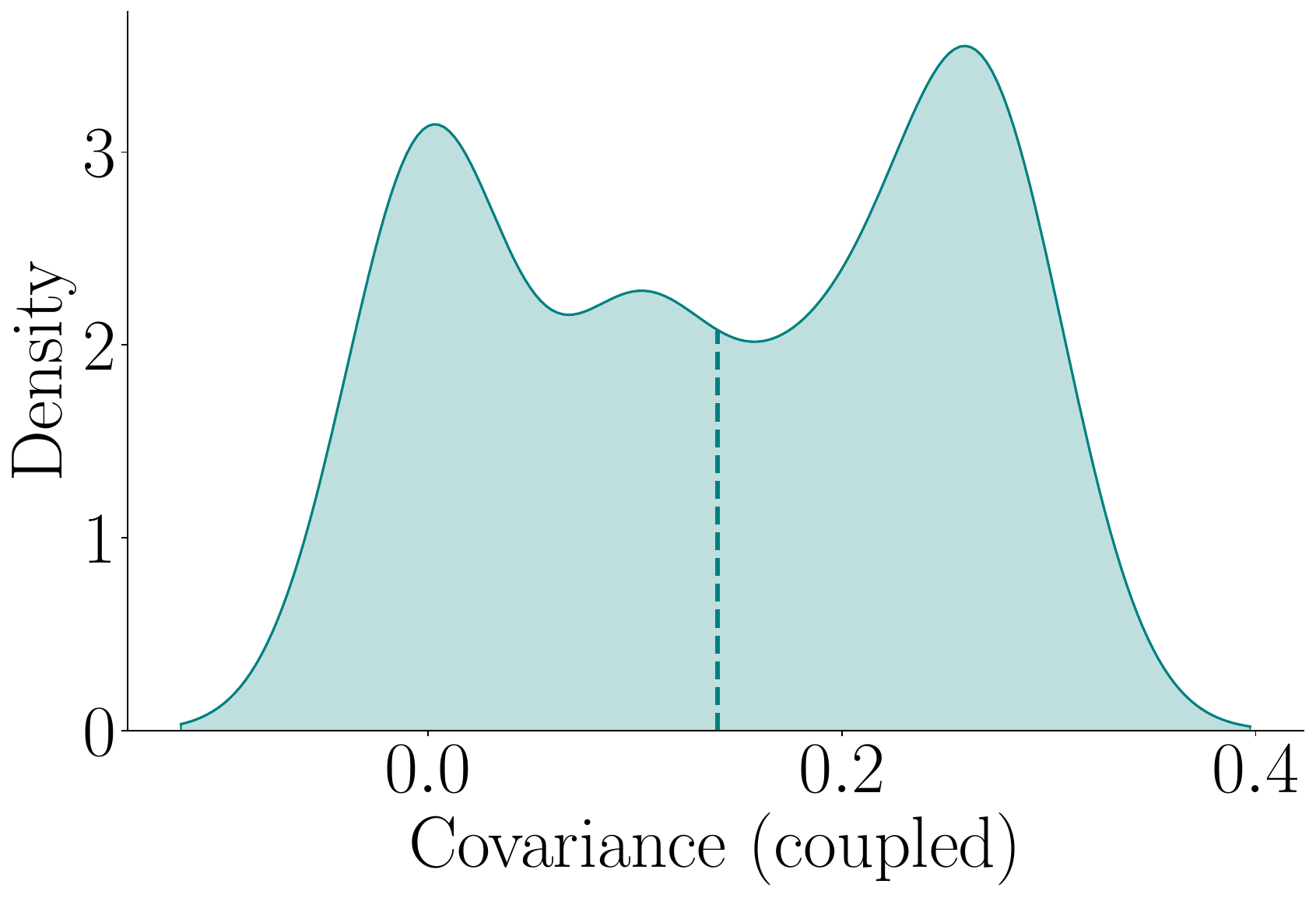} &
    \includegraphics[width=0.23\linewidth]{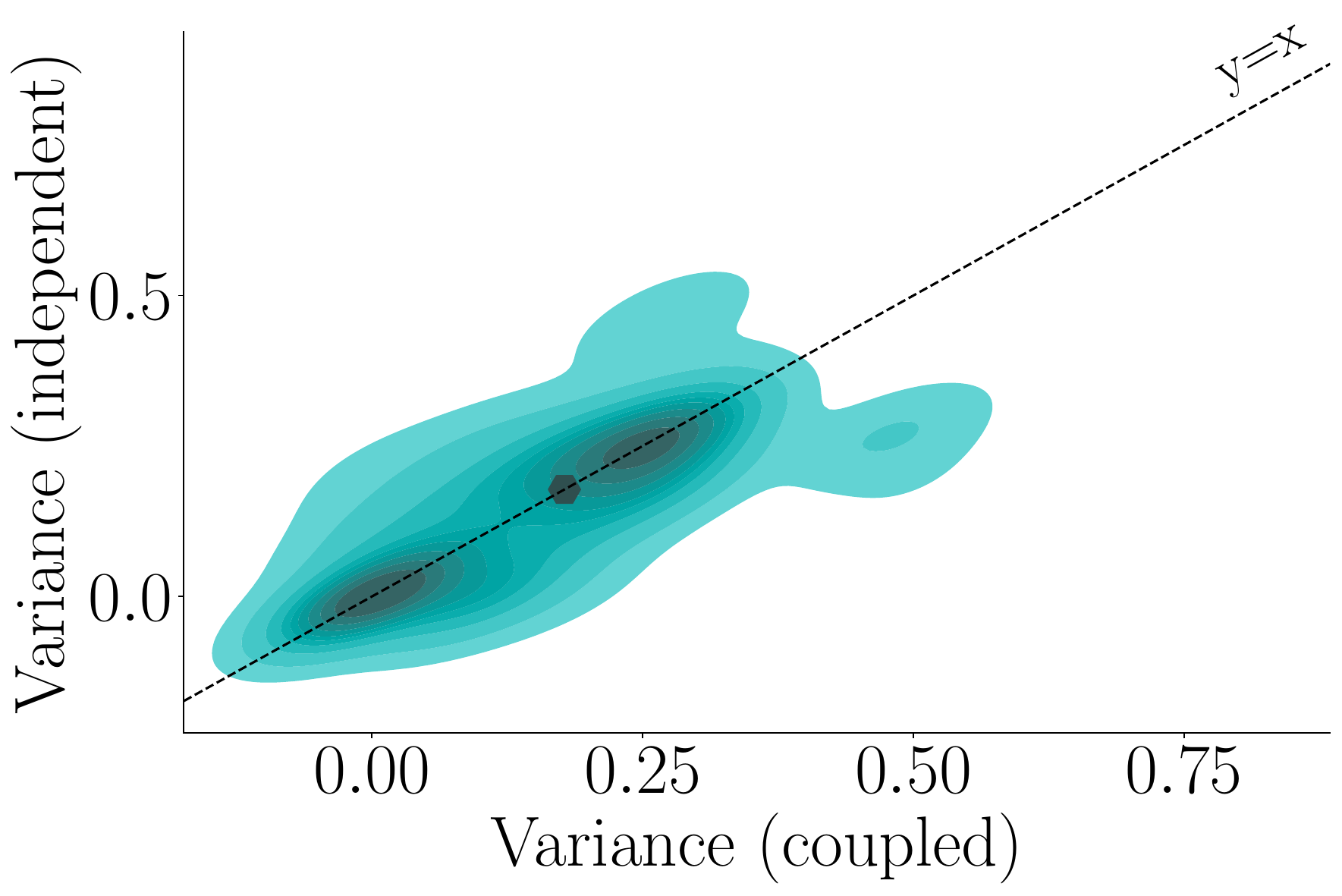} &
    \includegraphics[width=0.23\linewidth]{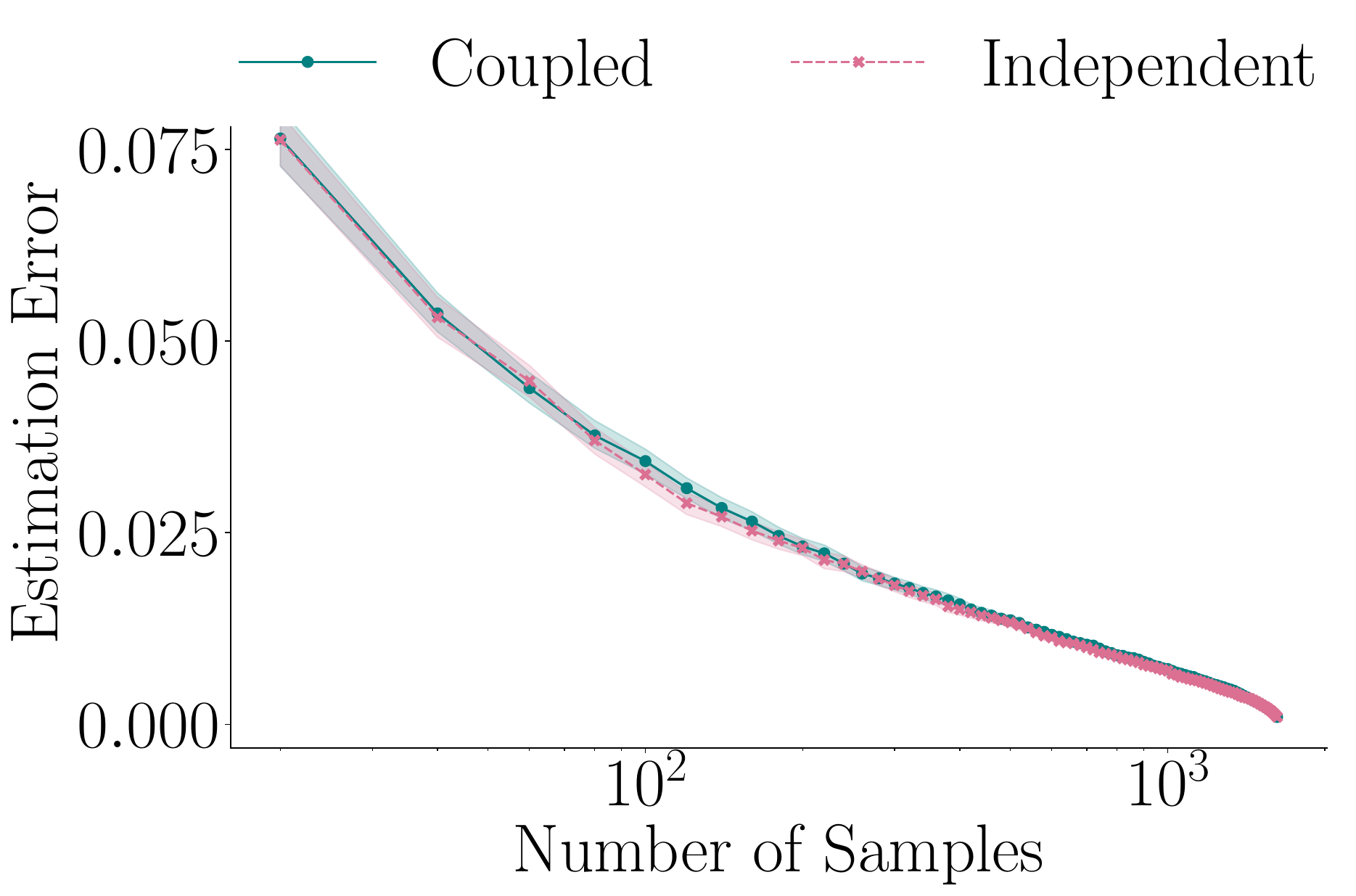} \\ \\
    \multicolumn{3}{c}{\texttt{v0.3-bnb-8bit} vs. \texttt{v0.1}}\\
    \includegraphics[width=0.23\linewidth]{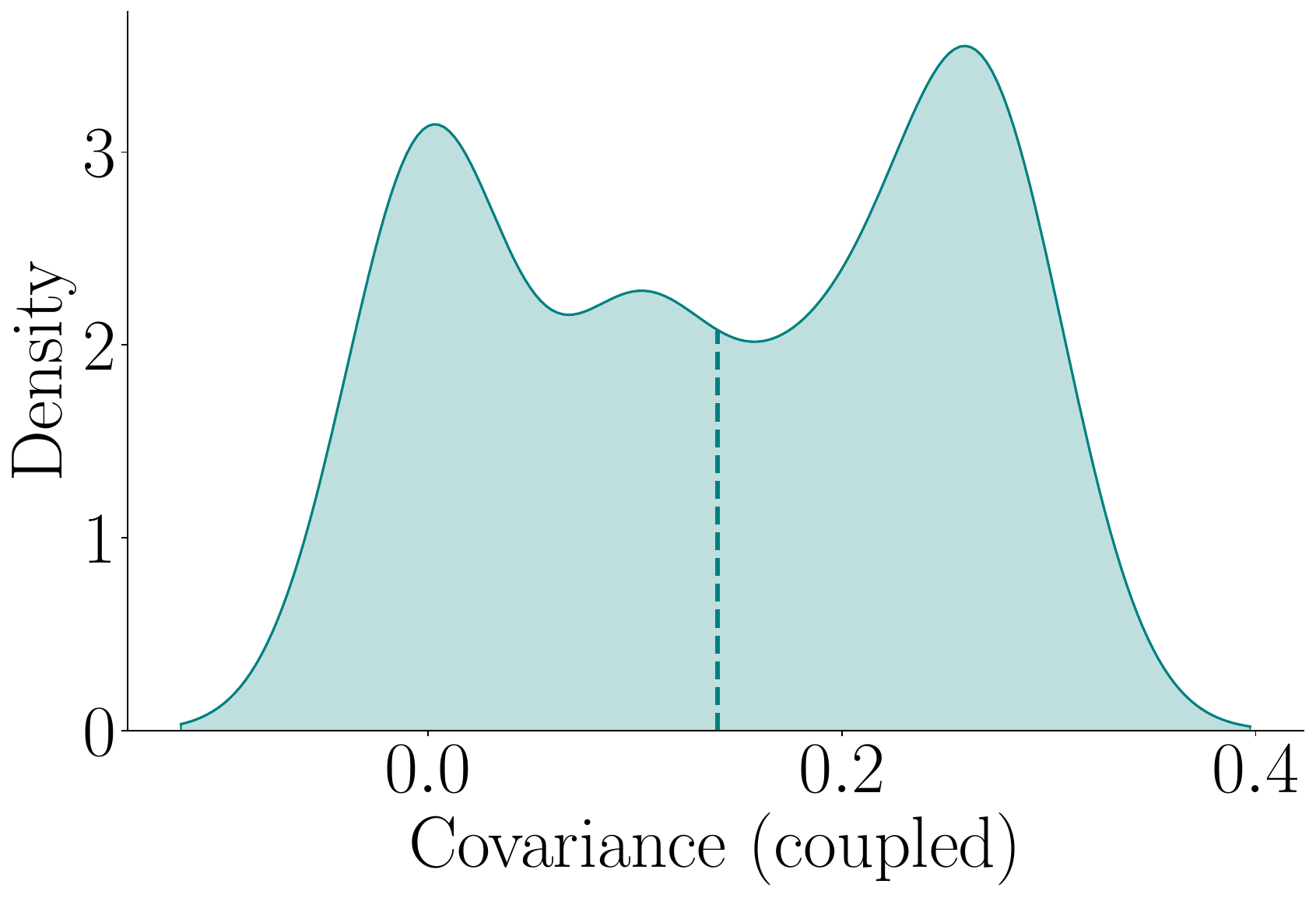} &
    \includegraphics[width=0.23\linewidth]{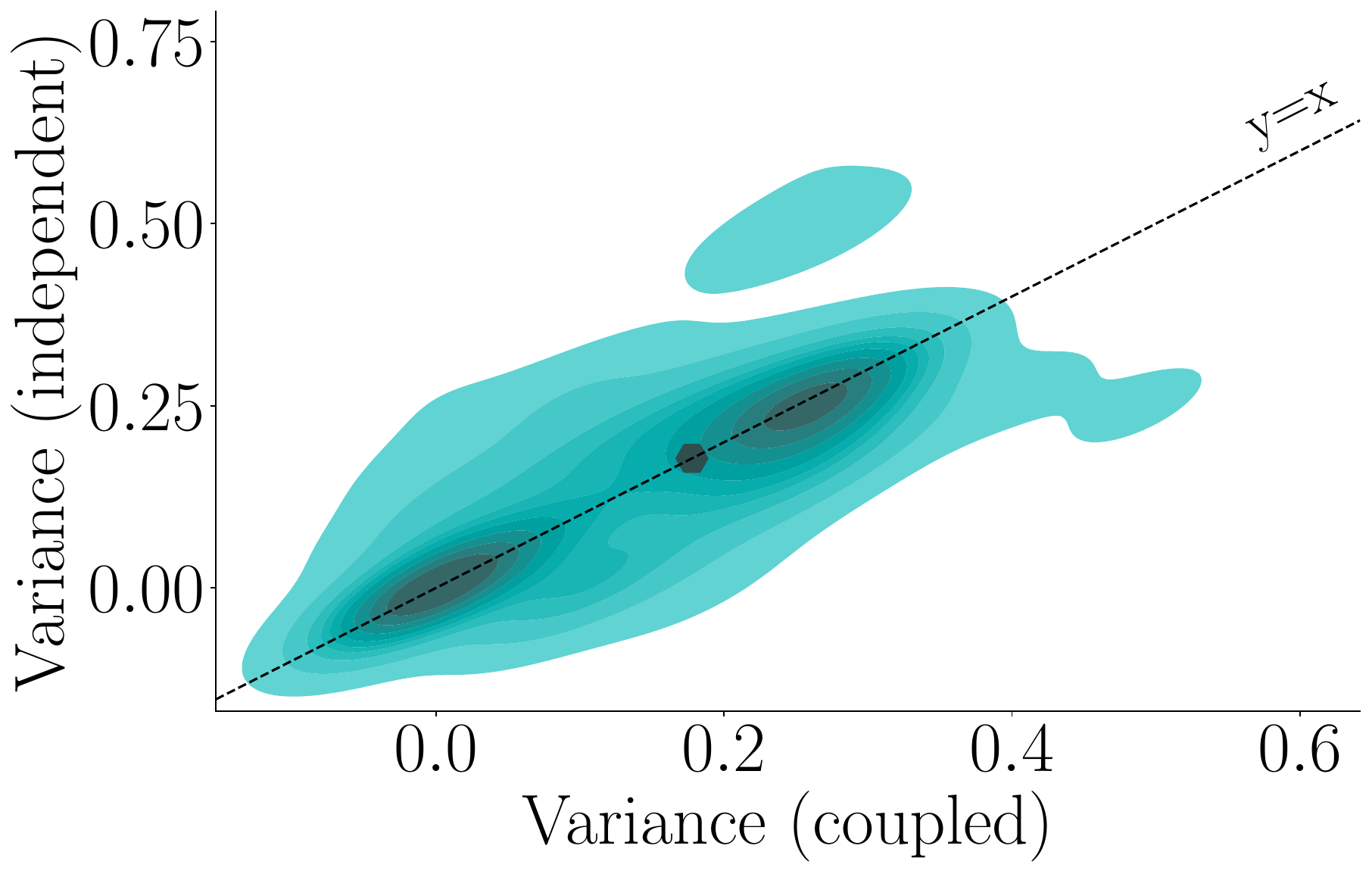} &
    \includegraphics[width=0.23\linewidth]{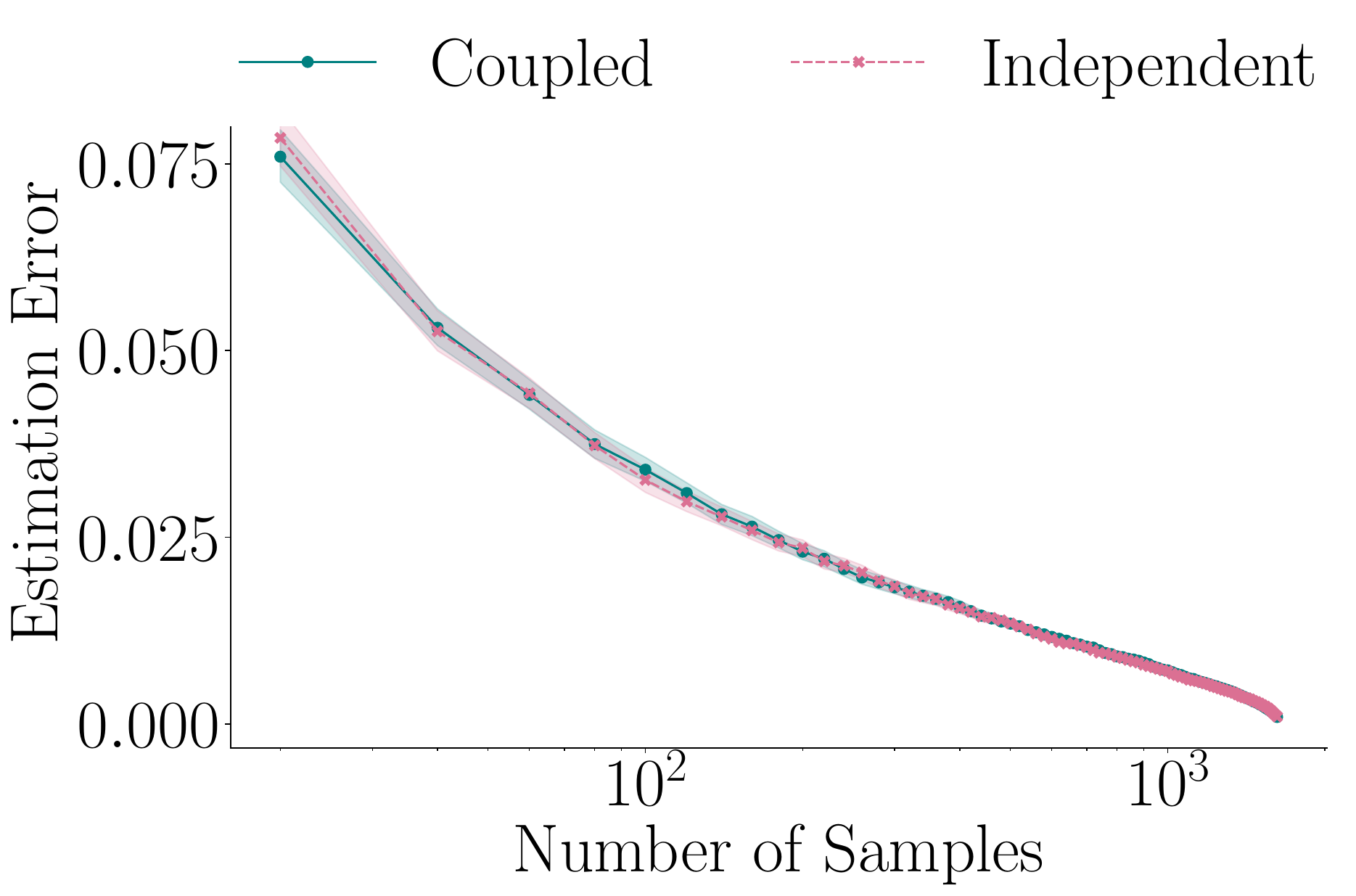} \\ \\
     \multicolumn{3}{c}{\texttt{v0.1} vs. \texttt{v0.3-bnb-4bit}}\\
    \includegraphics[width=0.23\linewidth]{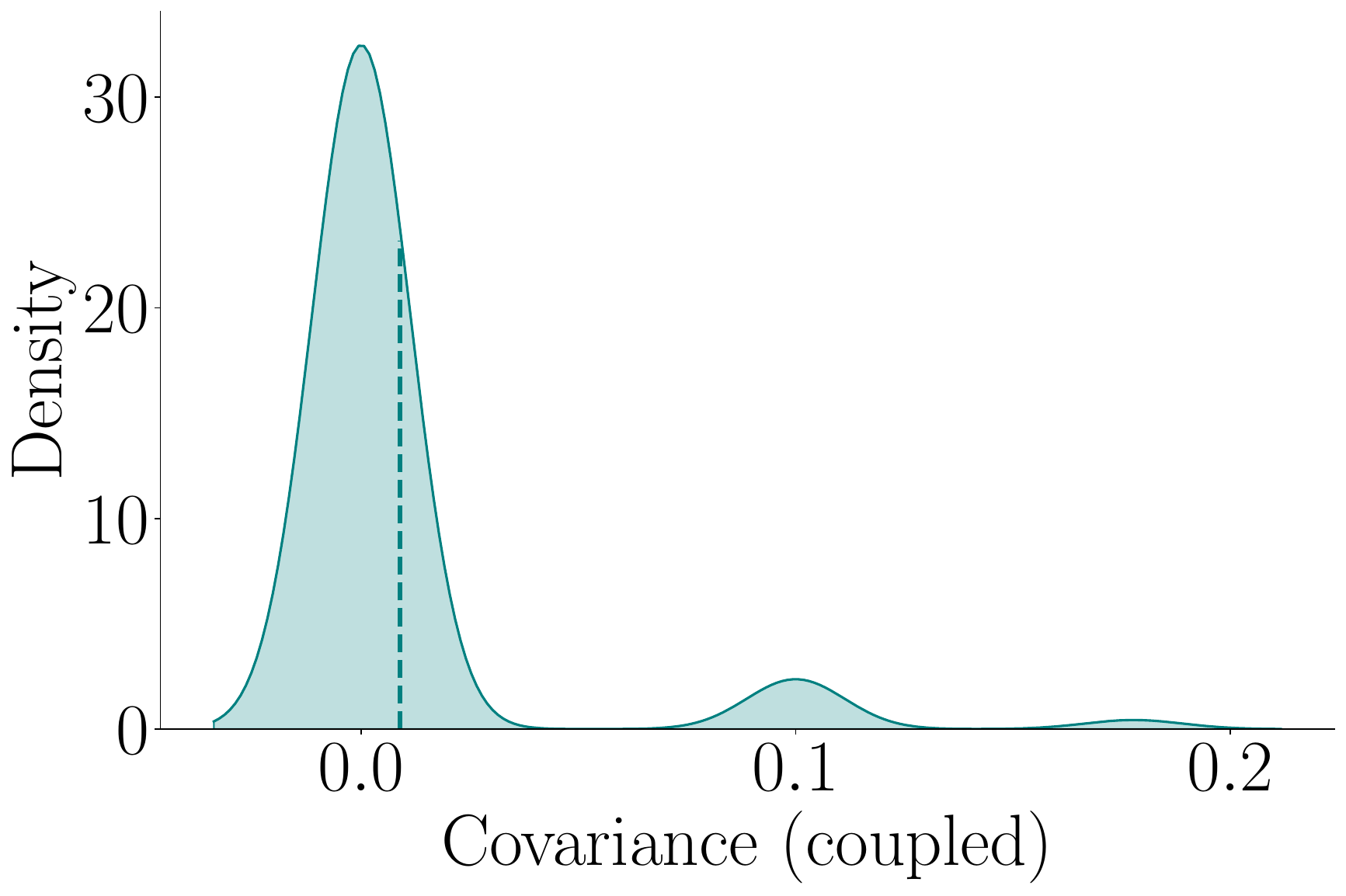} &
    \includegraphics[width=0.23\linewidth]{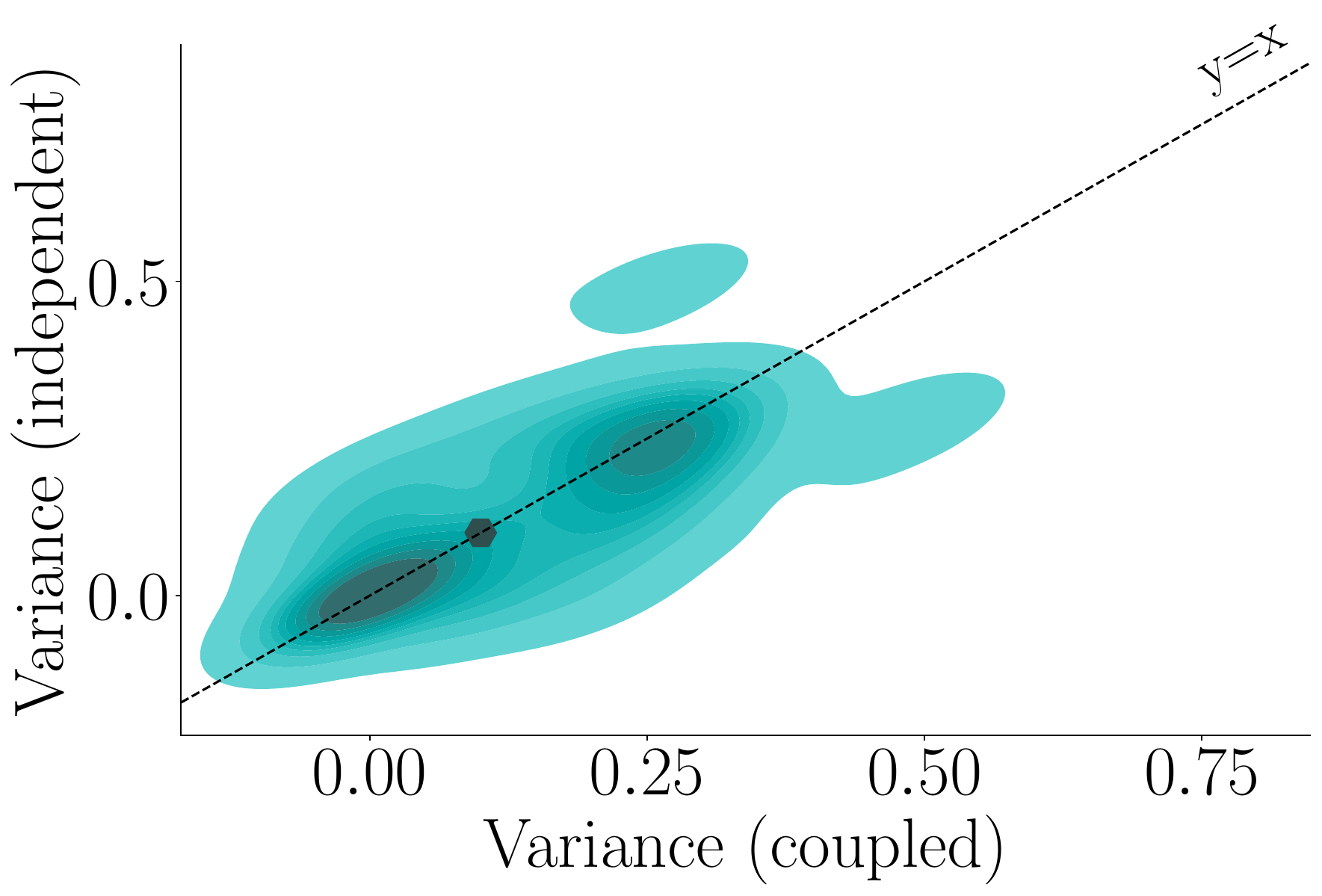} &
    \includegraphics[width=0.23\linewidth]{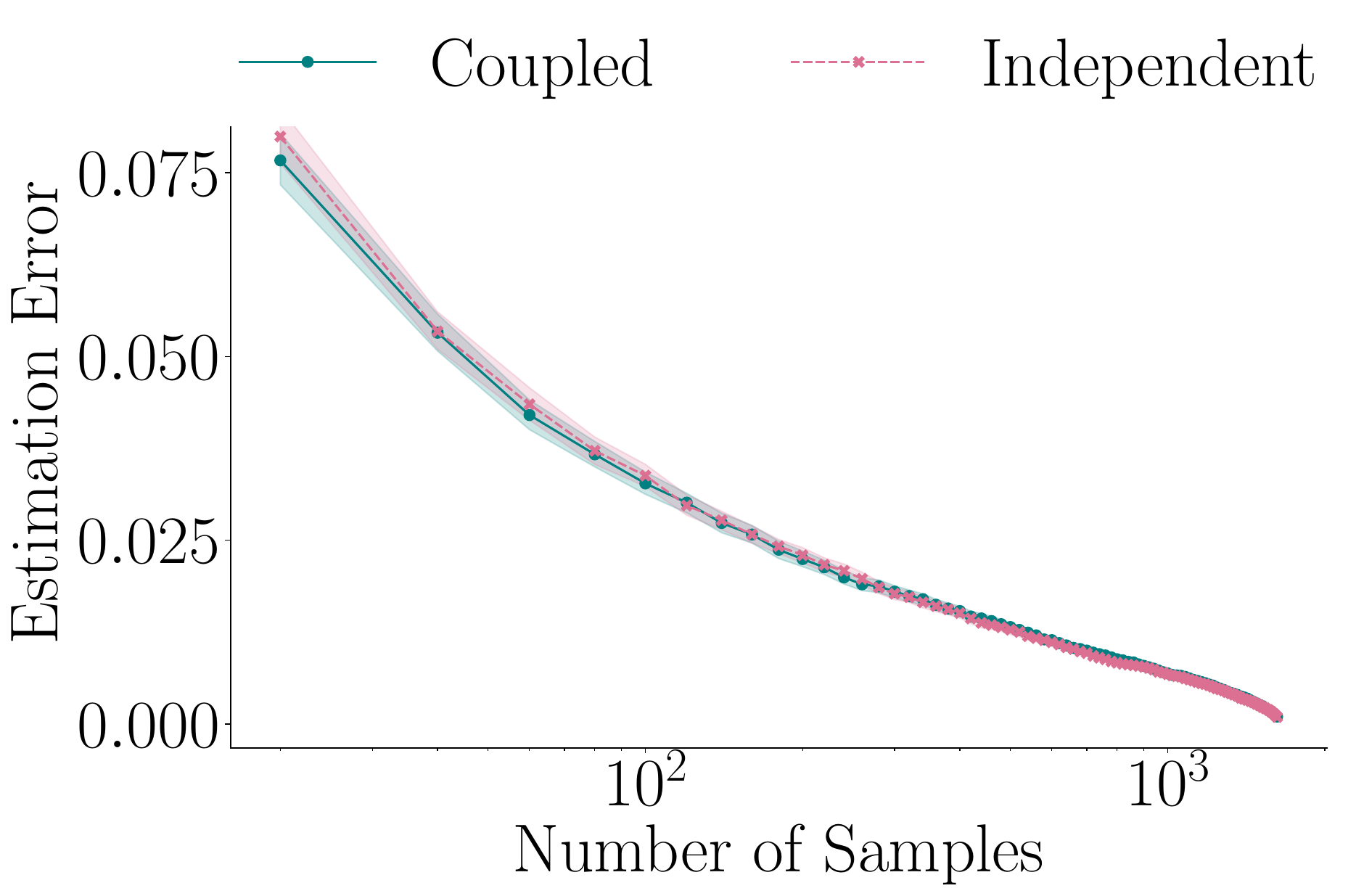} \\ \\
    \multicolumn{3}{c}{\texttt{v0.2} vs. \texttt{v0.3-bnb-4bit}}\\
    \includegraphics[width=0.23\linewidth]{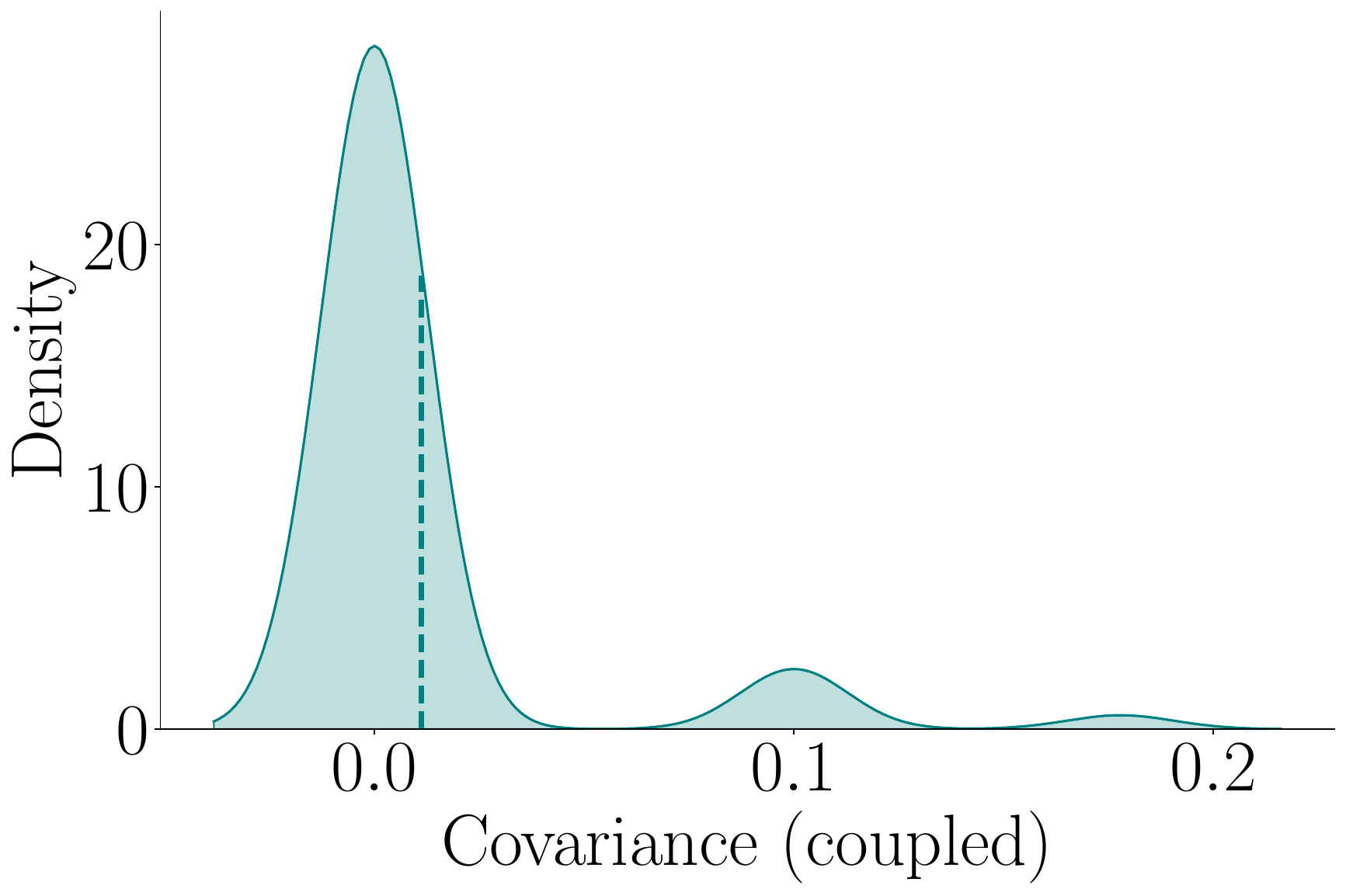} &
    \includegraphics[width=0.23\linewidth]{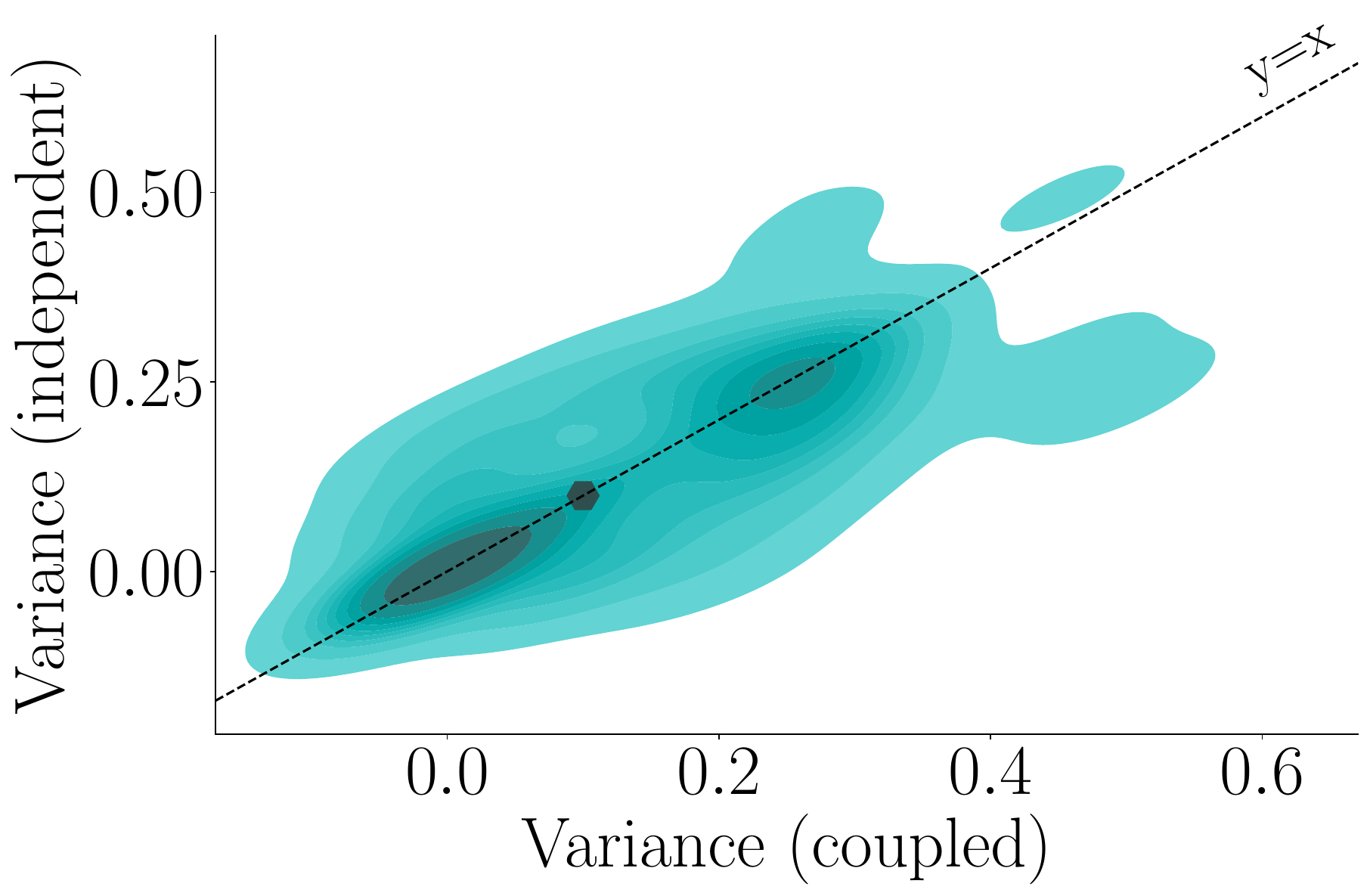} &
    \includegraphics[width=0.23\linewidth]{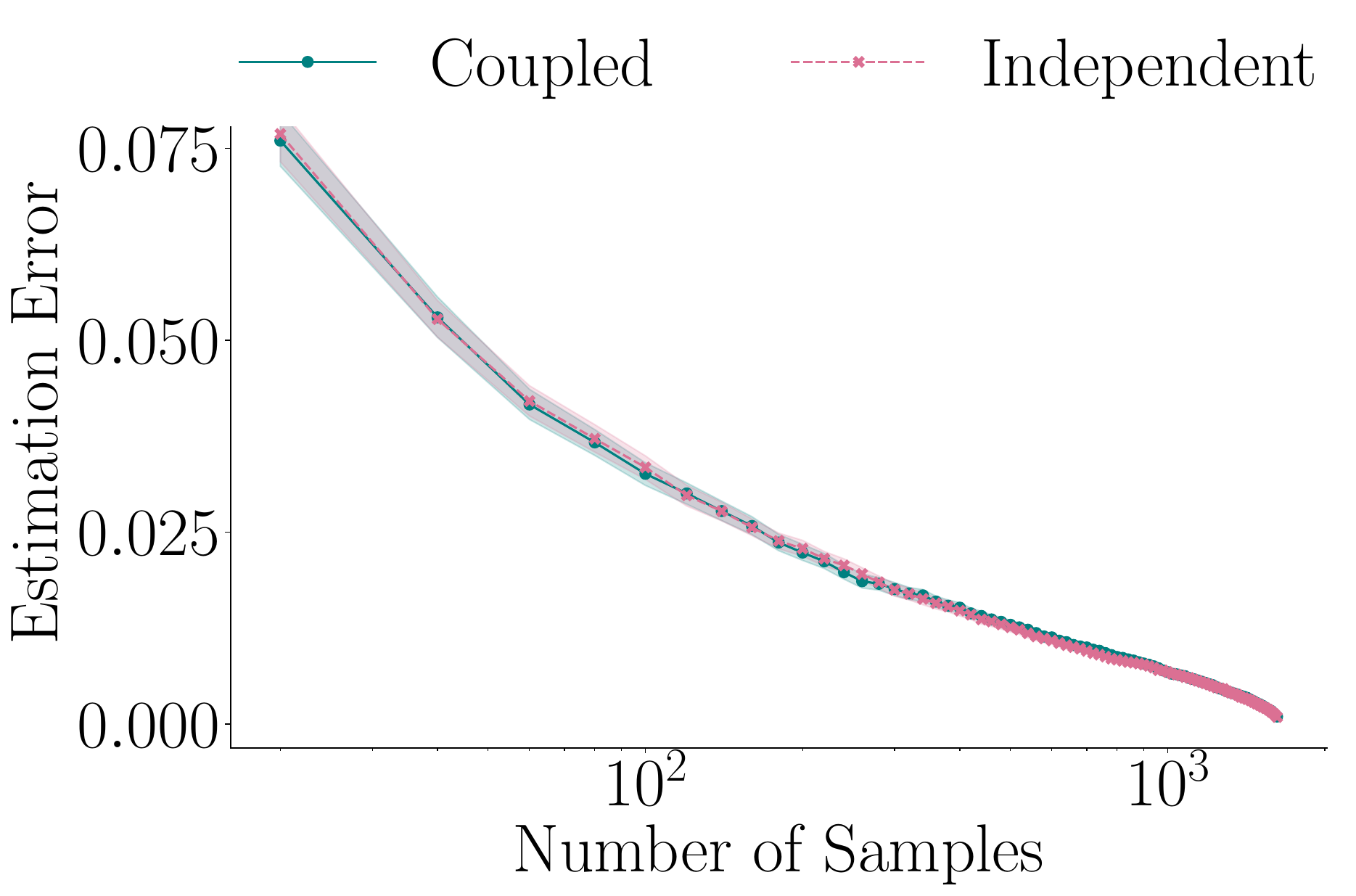} \\ \\
     \multicolumn{3}{c}{\texttt{v0.2} vs. \texttt{v0.1}}\\
    \includegraphics[width=0.23\linewidth]{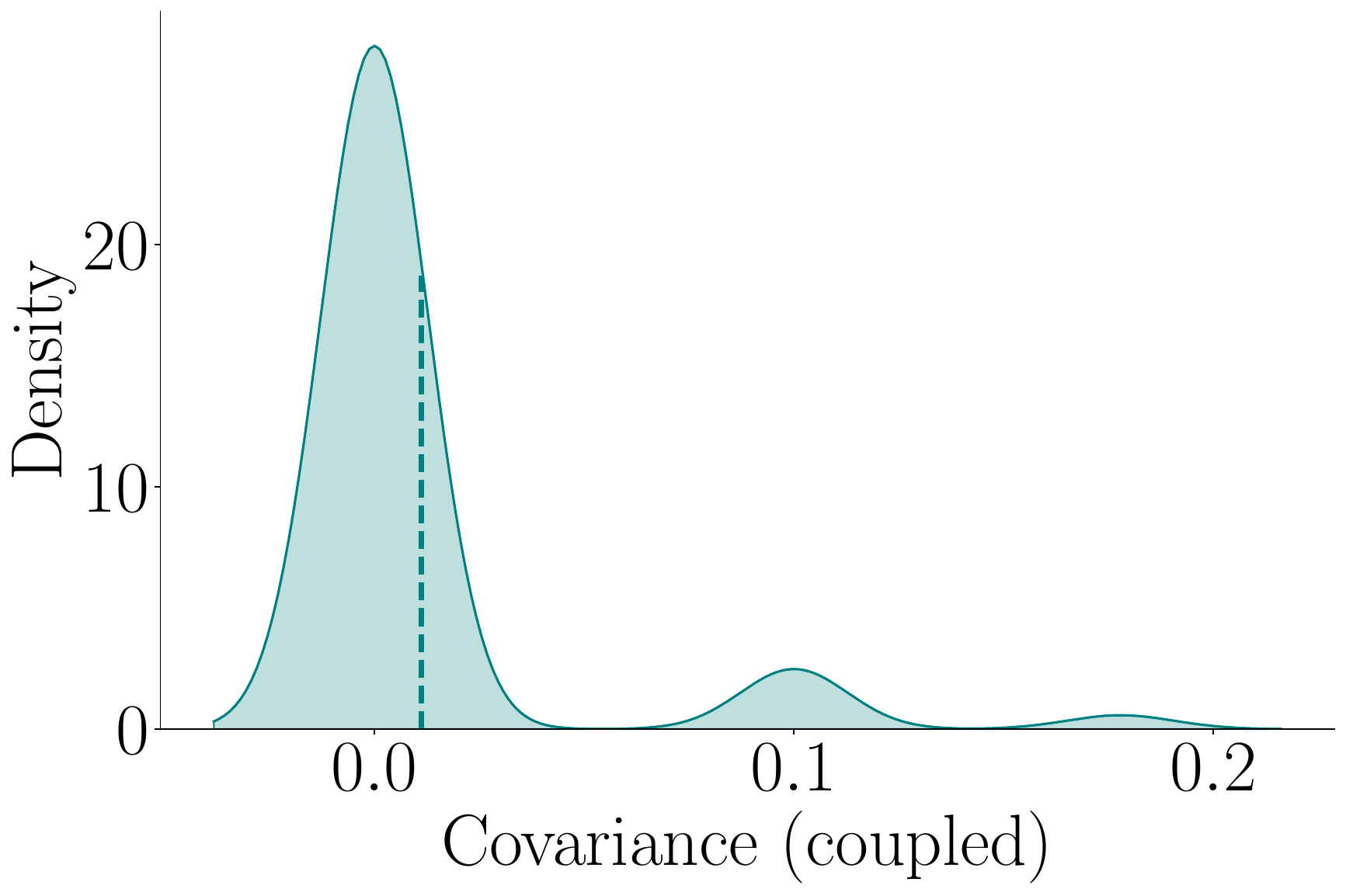} &
    \includegraphics[width=0.23\linewidth]{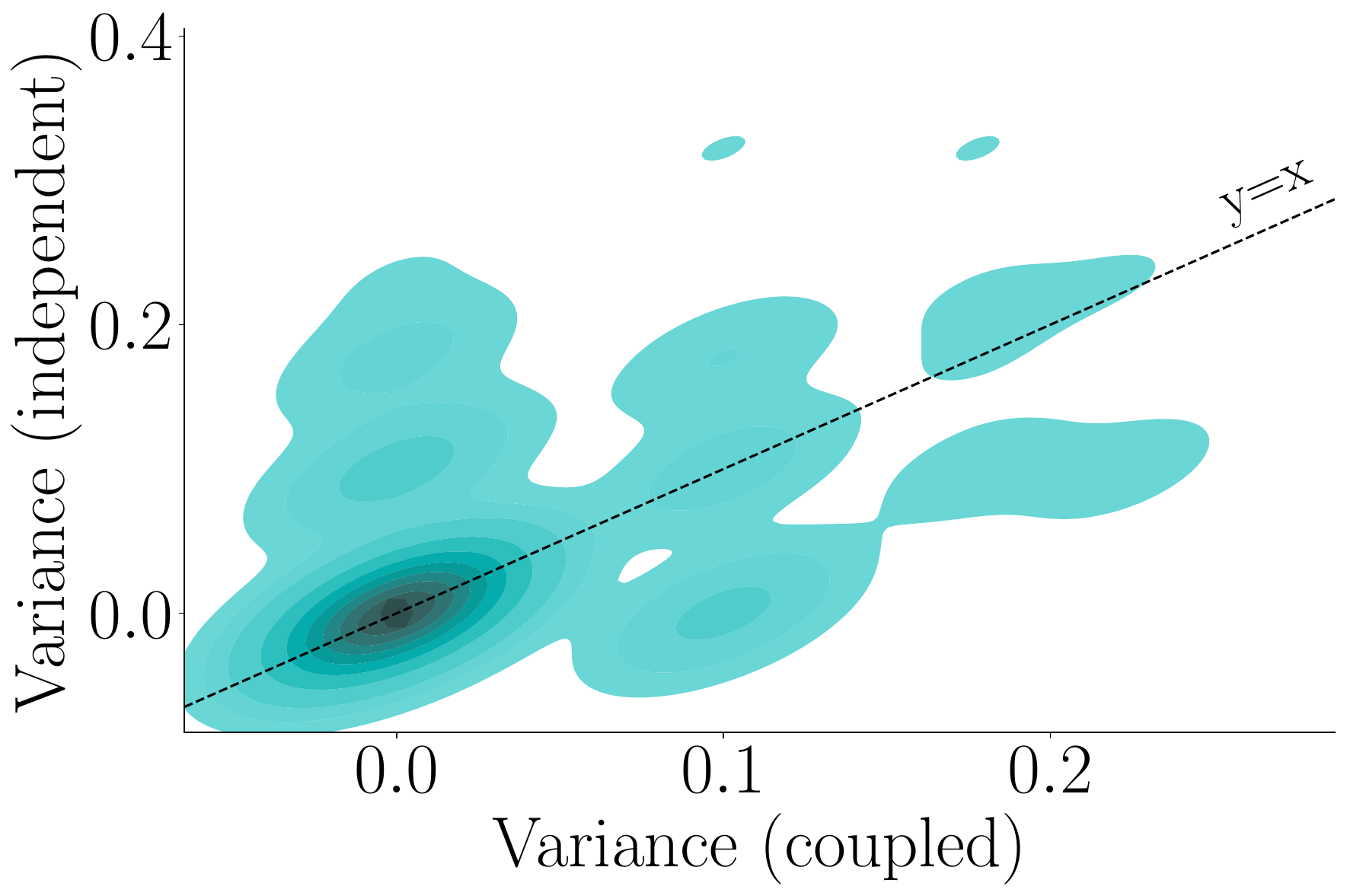} &
    \includegraphics[width=0.23\linewidth]{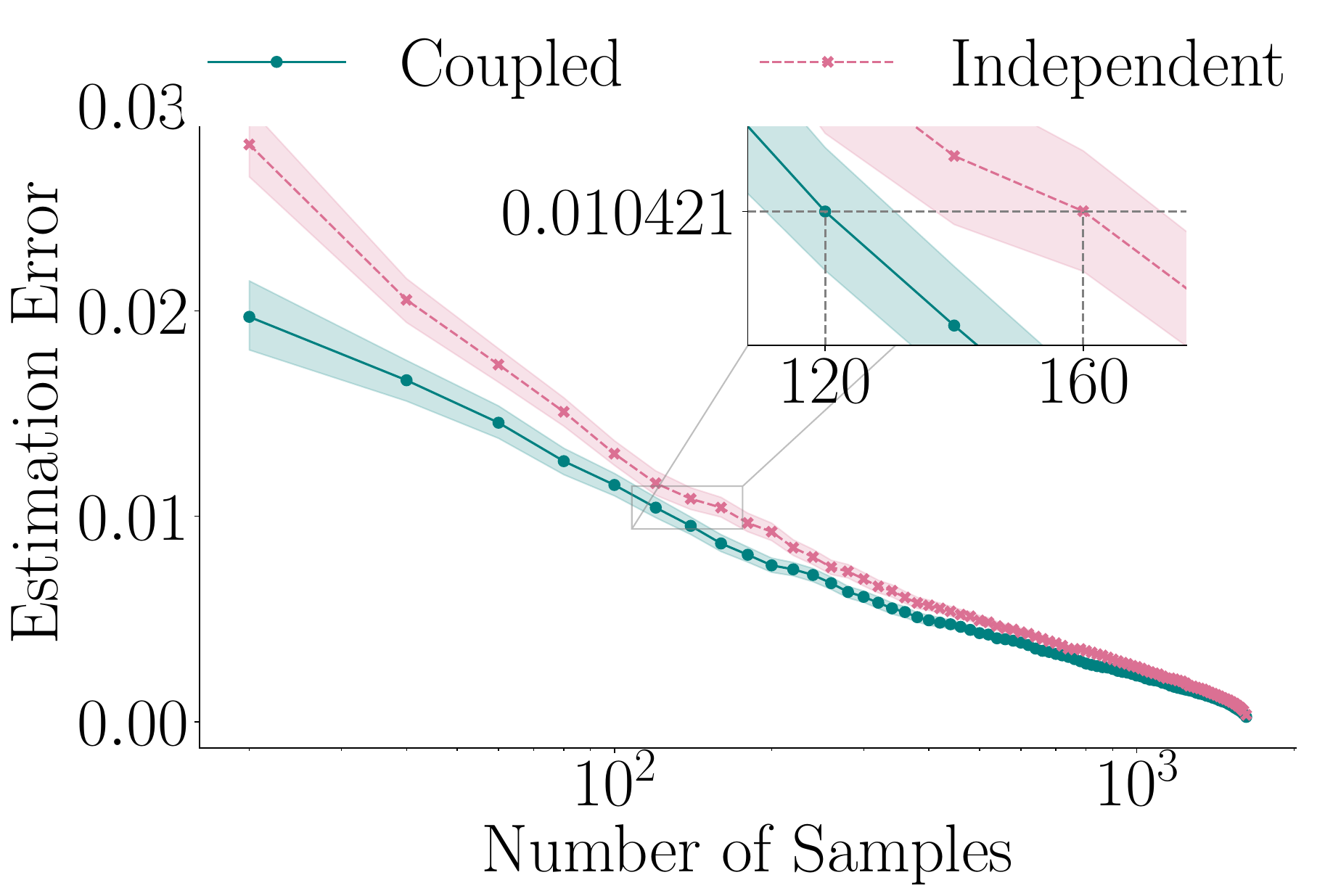} \\ \\
    (a) Score covariance & (b) Variance of the score difference & (c) Estimation error vs. \# samples \\ 
    
\end{tabular}
    \caption{\textbf{Comparison between several pairs of LLMs in the \texttt{Mistral} family on programming problems from the HumanEval dataset.}
    Panels in column (a) show the kernel density estimate (KDE) of the covariance between the scores of the two LLMs on each problem under coupled generation; the dashed lines correspond to average values. Panels in column (b) show the KDE of the variance of the difference between the scores of the LLMs on each question under coupled and independent generation; the highlighted points correspond to median values. Panels in column (c) show the absolute error in the estimation of the expected difference between the scores of the LLMs against the number of samples; for each point on the x-axis, we perform $1{,}000$ sub-samplings and shaded areas correspond to $95\%$ confidence intervals.}
    \label{fig:human-eval-mistral-last}
\end{figure}

Here, we experiment with models from the \texttt{Mistral} family on the GSM8K and HumanEval datasets following the setup described in Appendix~\ref{app:llama-gsm8k-humaneval}. 
Figures~\ref{fig:gsm8k-mistral-first}--\ref{fig:human-eval-mistral-last} show the results for all pairs of models in Table~\ref{tab:mistral-names}, which are qualitatively similar to those in Appendices~\ref{app:llama-gsm8k-humaneval} and~\ref{app:qwen-gsm8k-human-eval}.

\begin{figure}[!!h]
\centering
\begin{tabular}{c c c}
     \multicolumn{3}{c}{\texttt{v0.3} vs. \texttt{v0.3-bnb-8bit}}\\
    \includegraphics[width=0.23\linewidth]{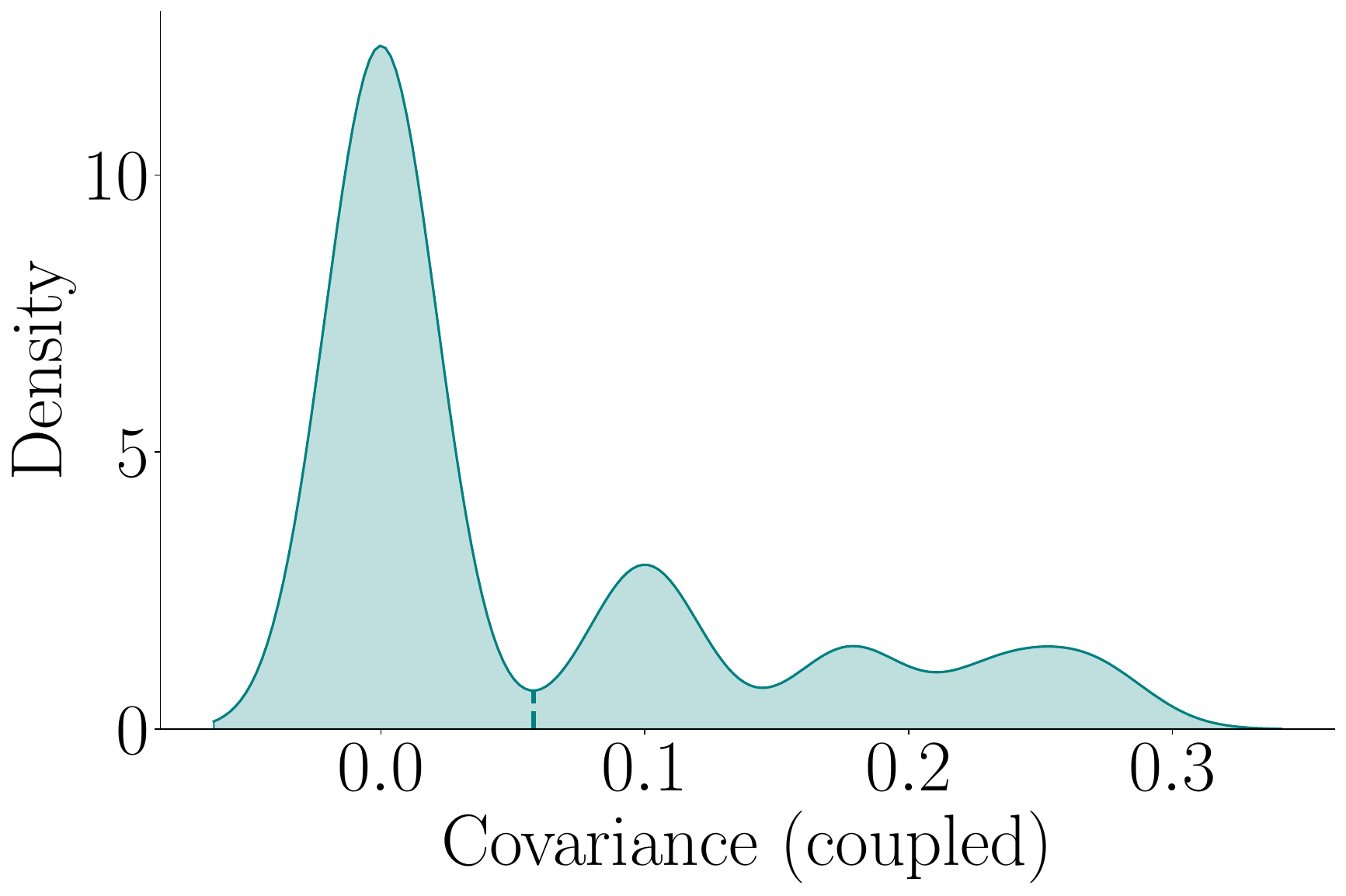} &
    \includegraphics[width=0.23\linewidth]{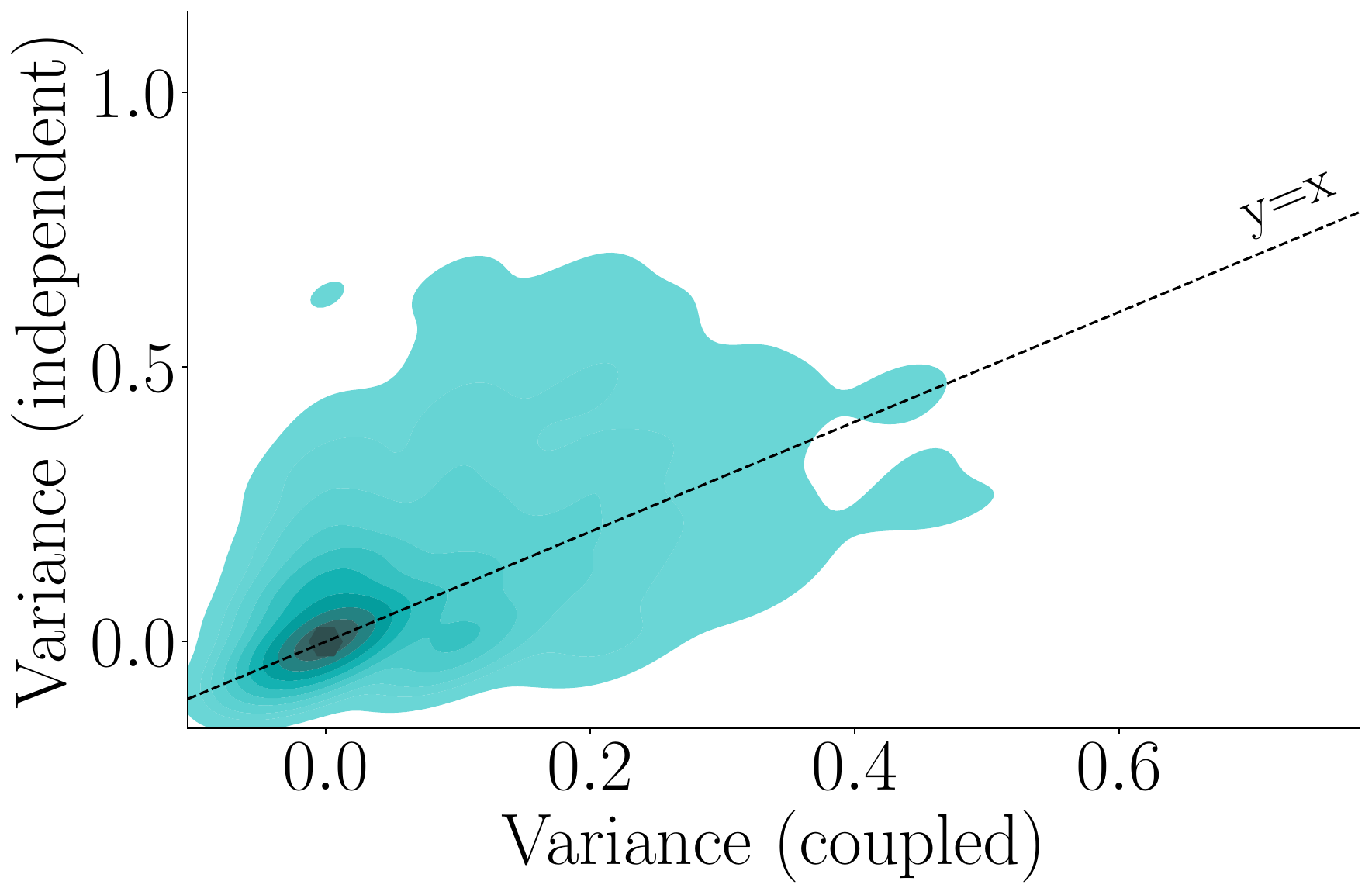} &
    \includegraphics[width=0.23\linewidth]{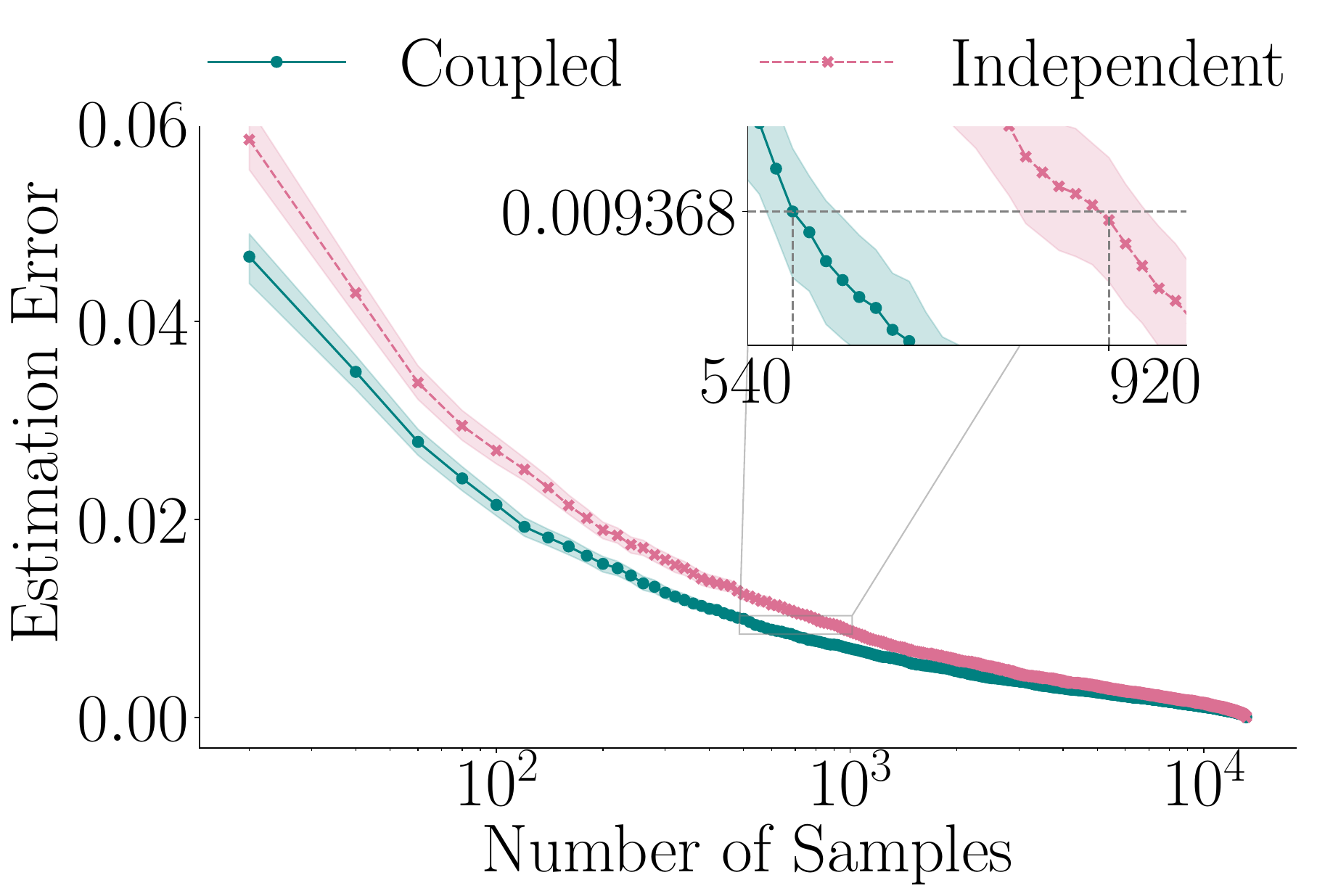} \\ \\
%
    \multicolumn{3}{c}{\texttt{v0.3} vs. \texttt{v0.3-bnb-4bit}}\\
    \includegraphics[width=0.23\linewidth]{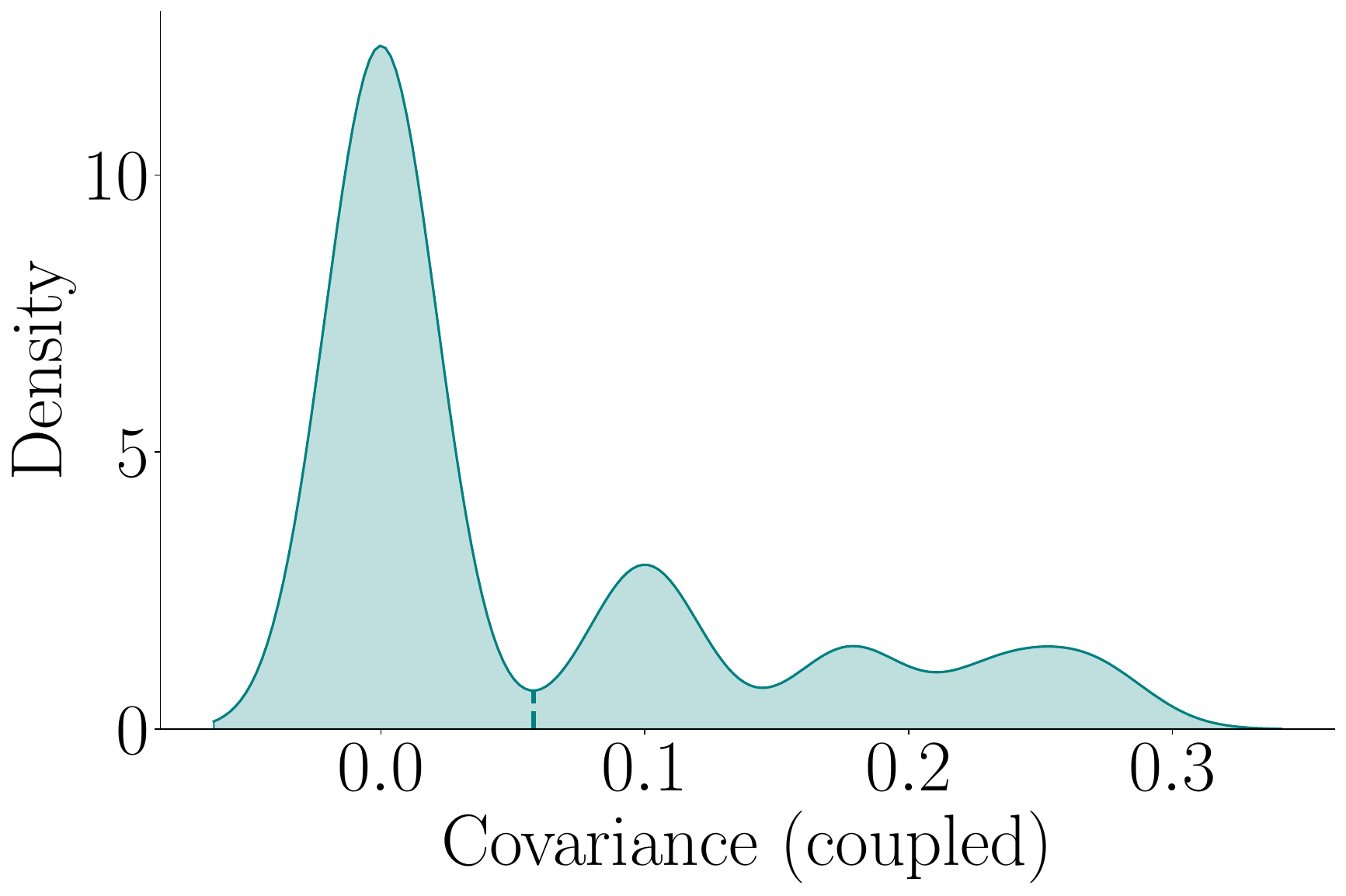} &
    \includegraphics[width=0.23\linewidth]{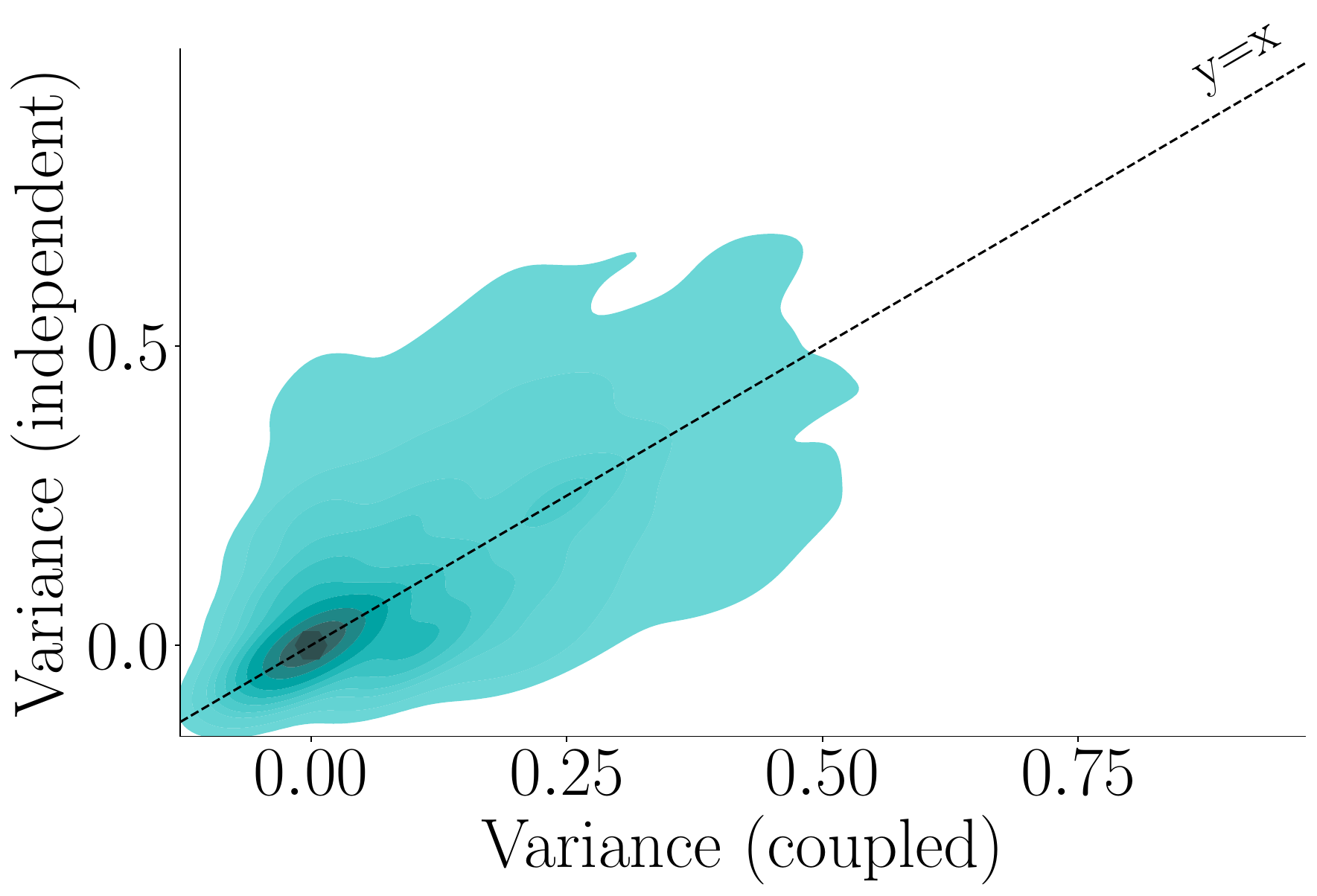} &
    \includegraphics[width=0.23\linewidth]{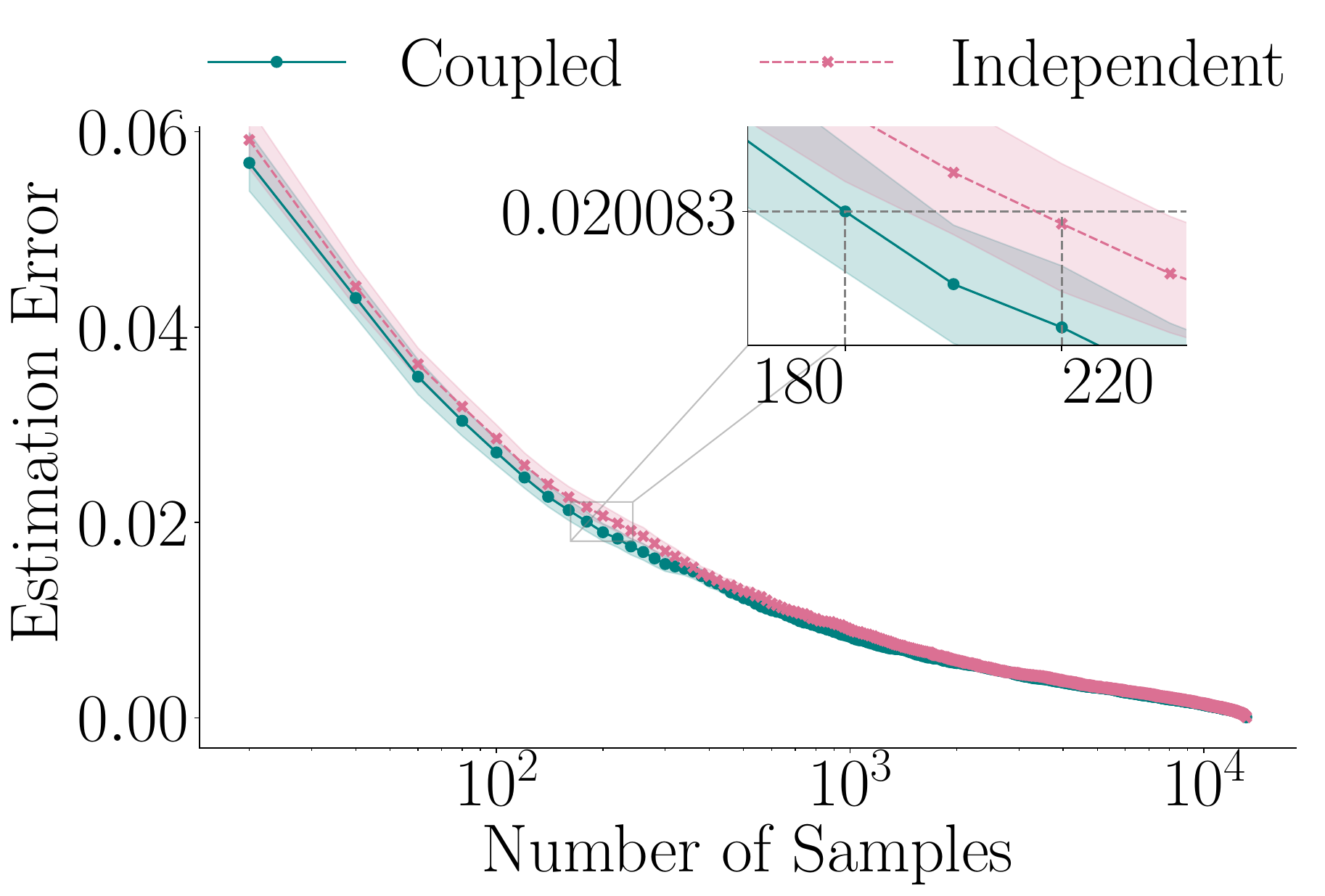} \\ \\
%
     \multicolumn{3}{c}{\texttt{v0.3-bnb-8bit} vs. \texttt{v0.3-bnb-4bit}}\\
    \includegraphics[width=0.23\linewidth]{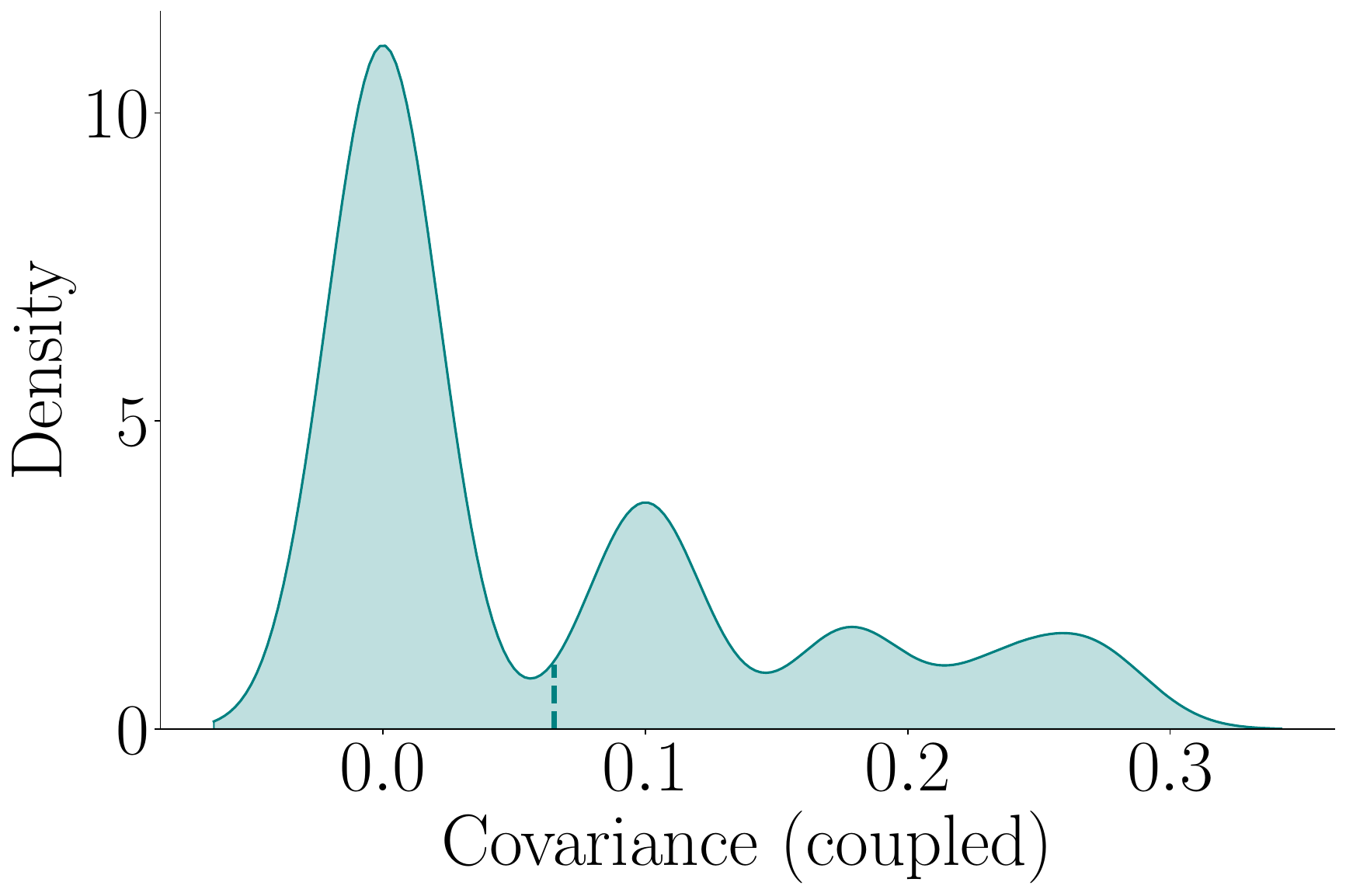} &
    \includegraphics[width=0.23\linewidth]{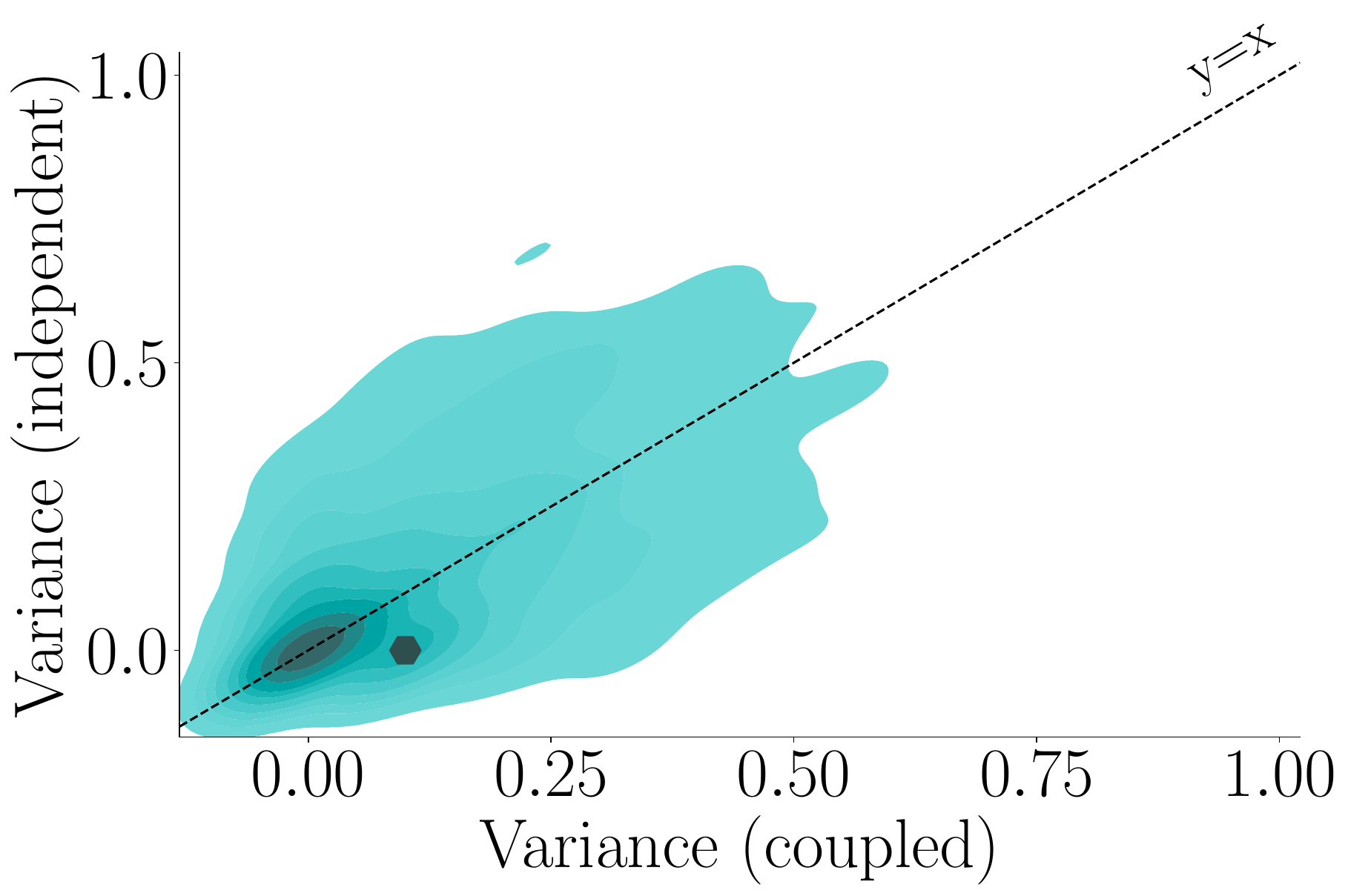} &
    \includegraphics[width=0.23\linewidth]{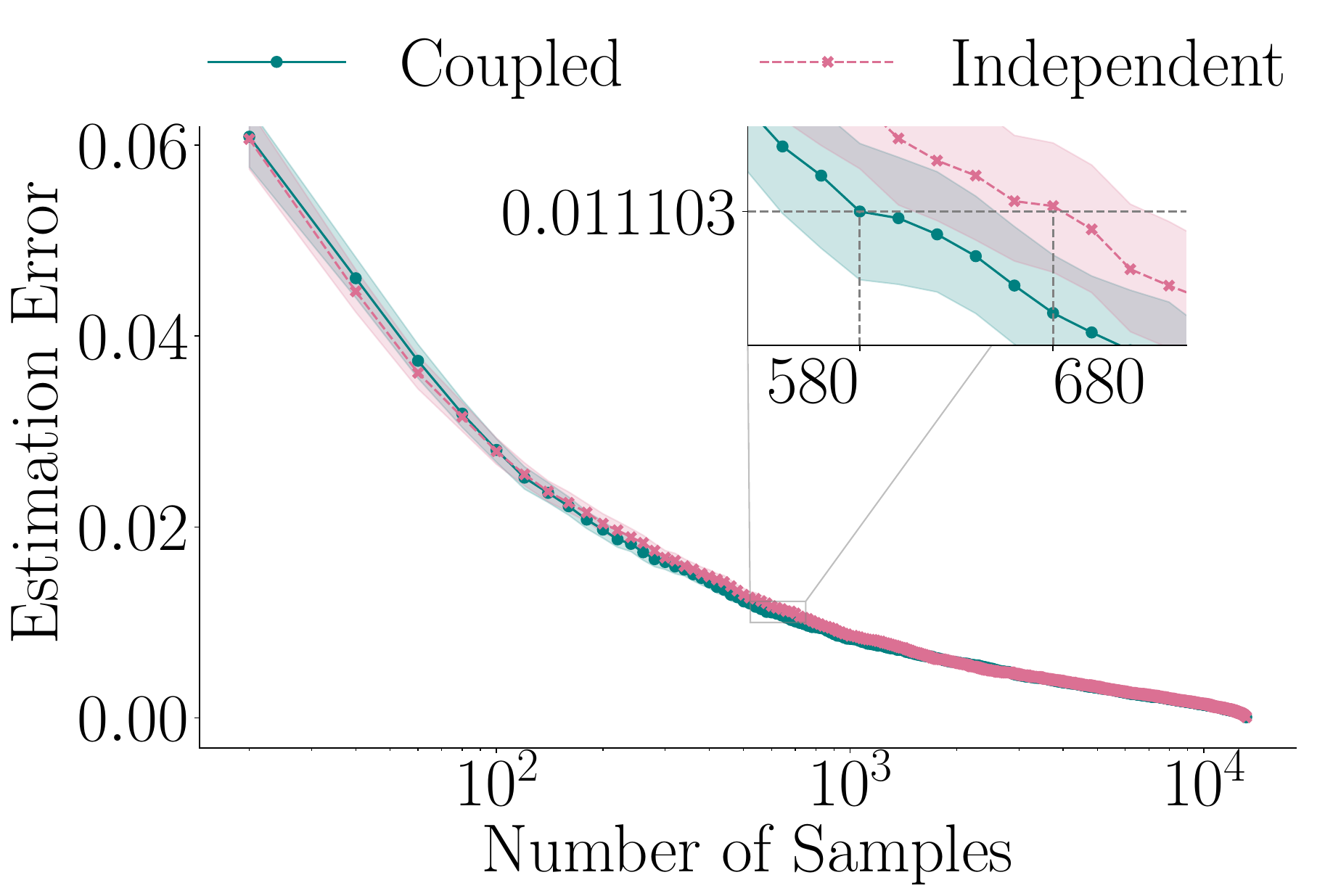} \\ \\
%
  \multicolumn{3}{c}{\texttt{v0.3} vs. \texttt{v0.2}}\\
    \includegraphics[width=0.23\linewidth]{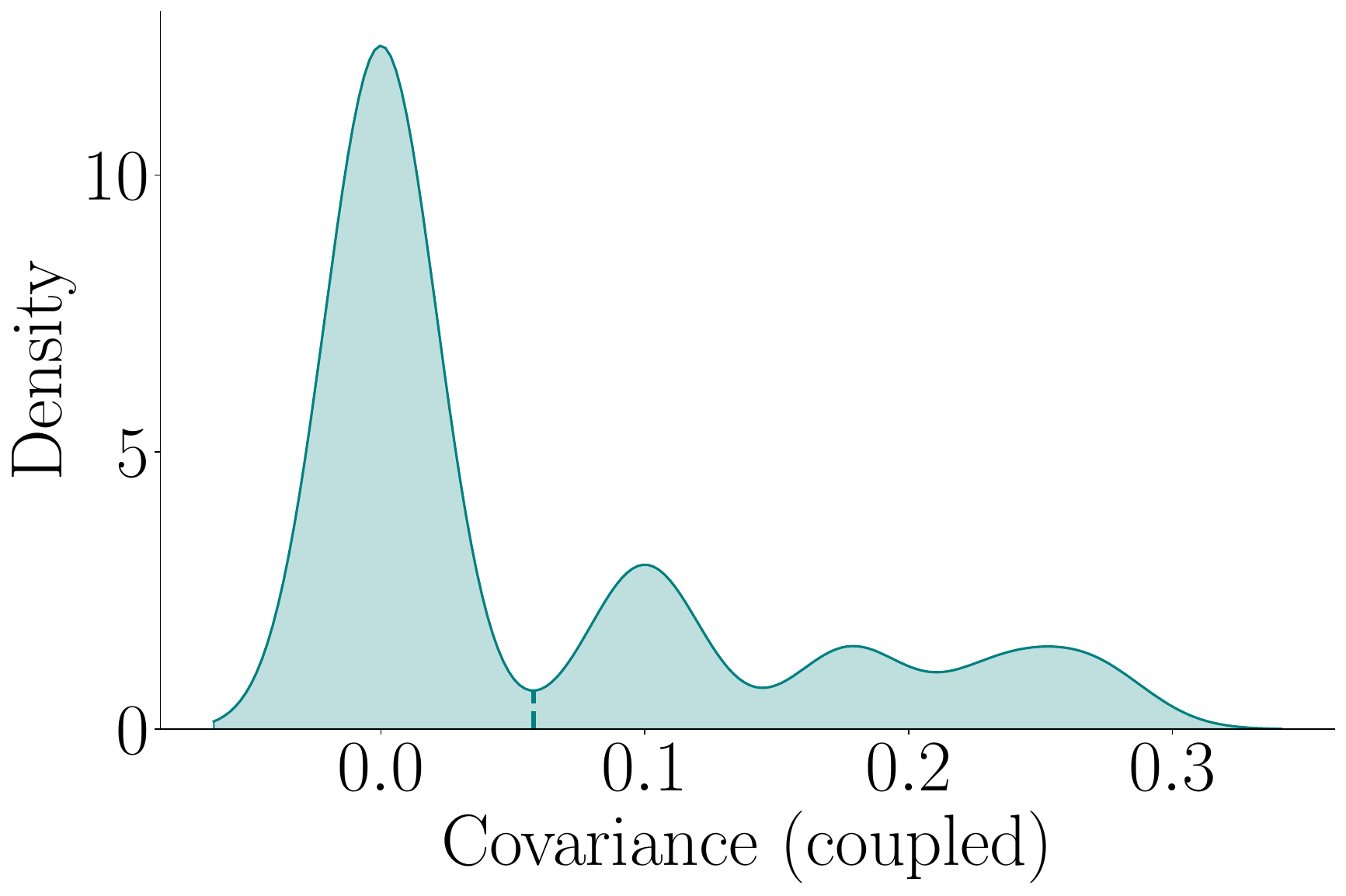} &
    \includegraphics[width=0.23\linewidth]{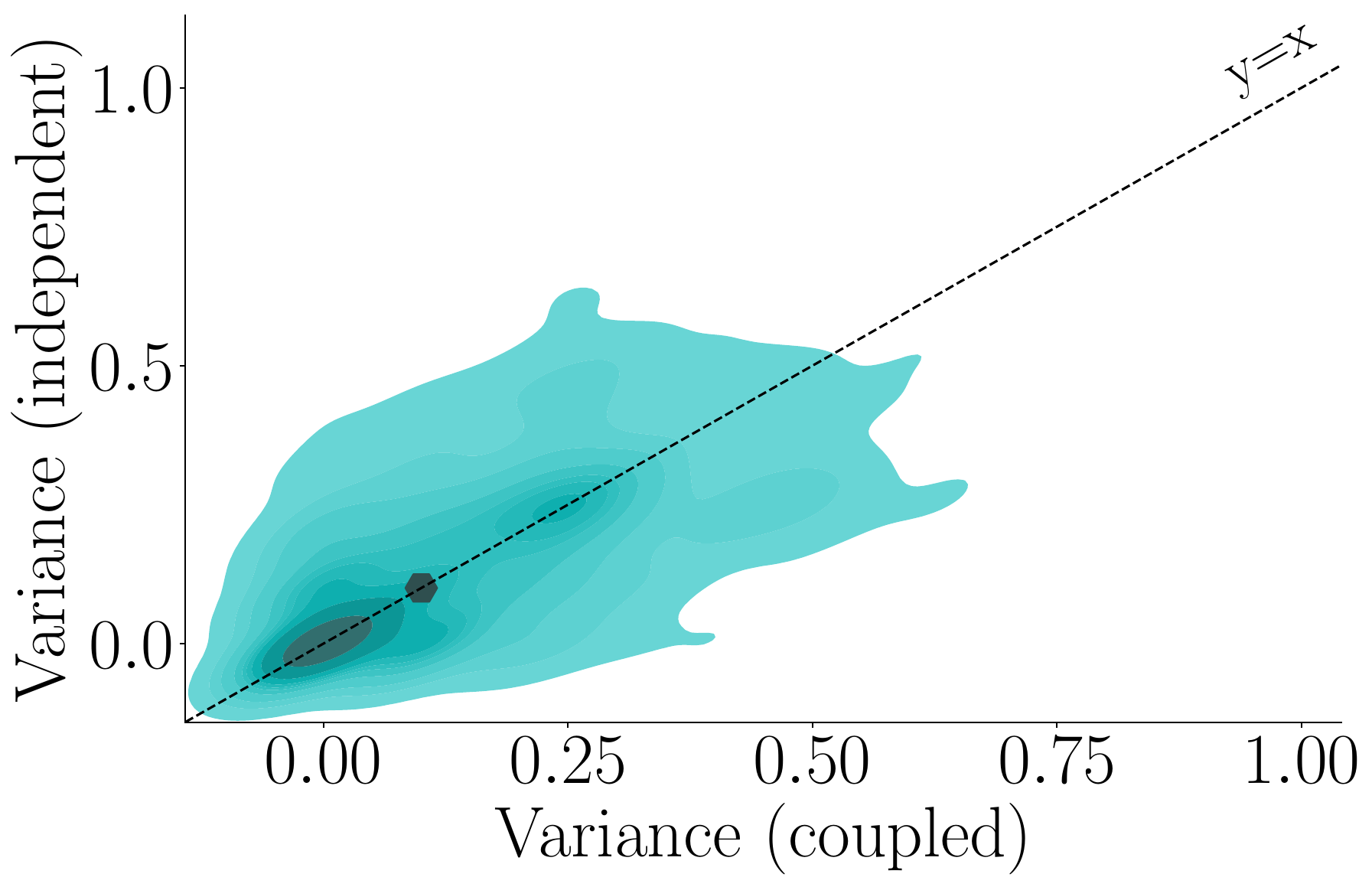} &
    \includegraphics[width=0.23\linewidth]{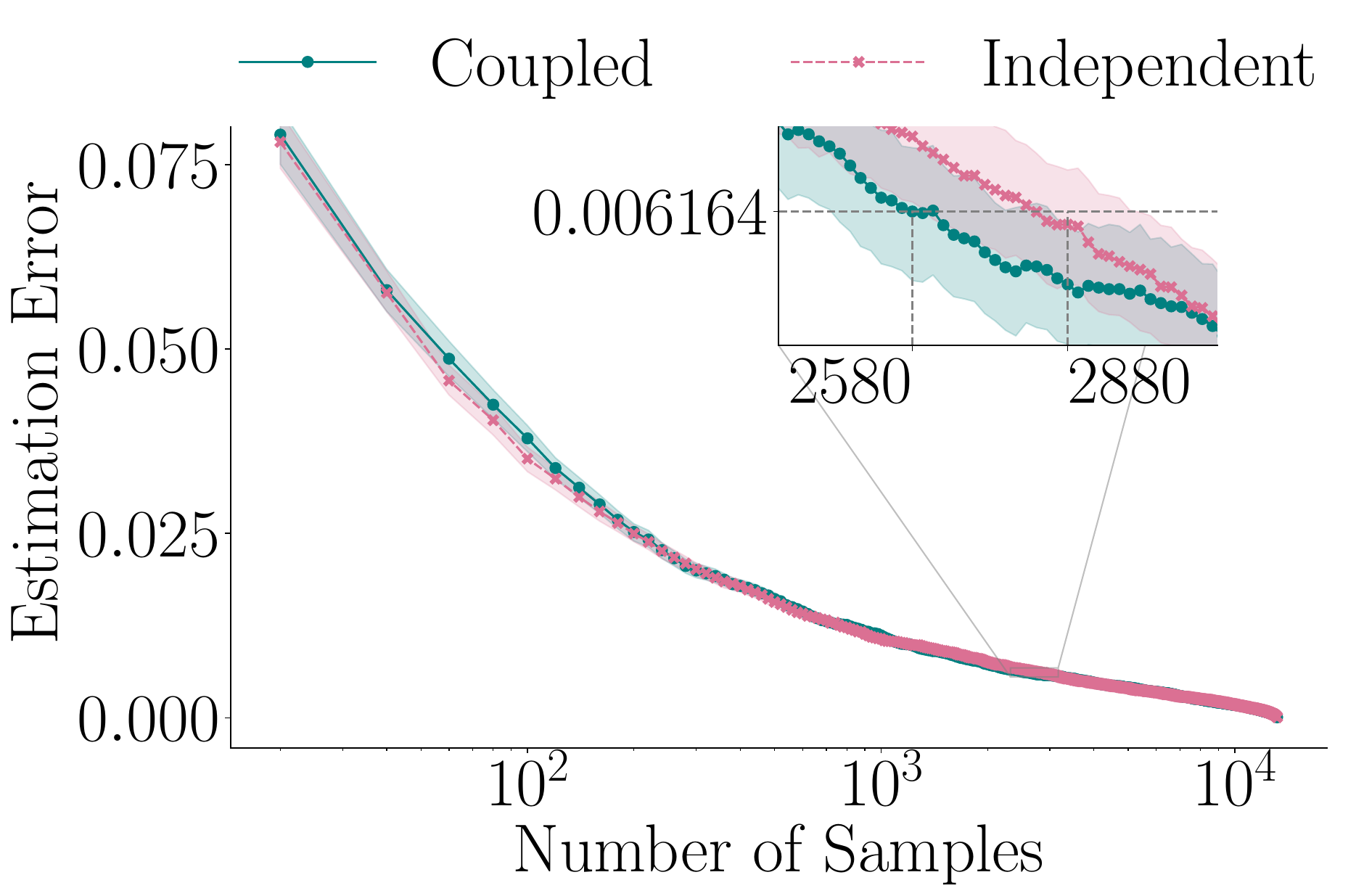} \\ \\

 \multicolumn{3}{c}{\texttt{v0.3} vs. \texttt{v0.1}}\\
    \includegraphics[width=0.23\linewidth]{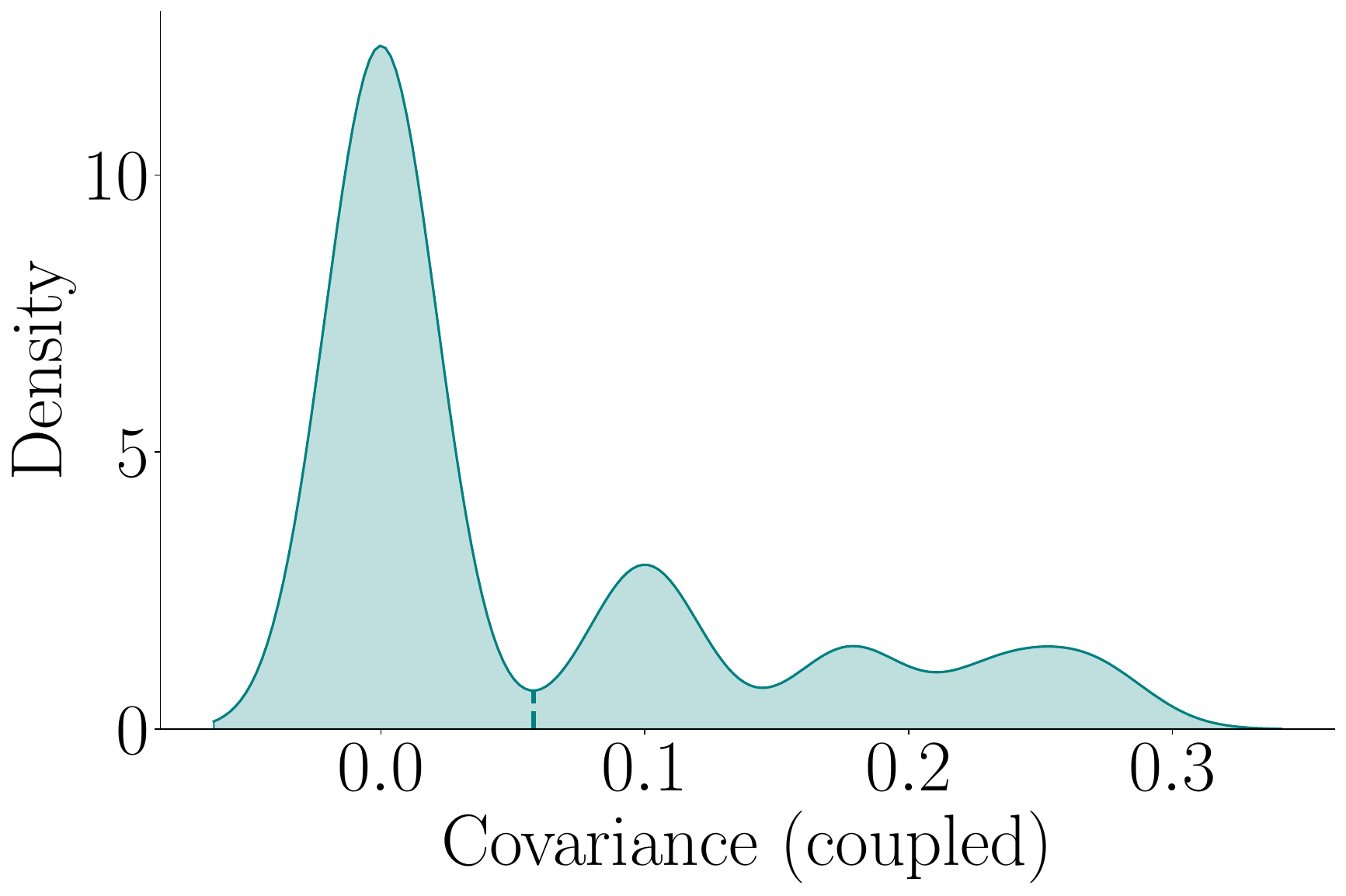} &
    \includegraphics[width=0.23\linewidth]{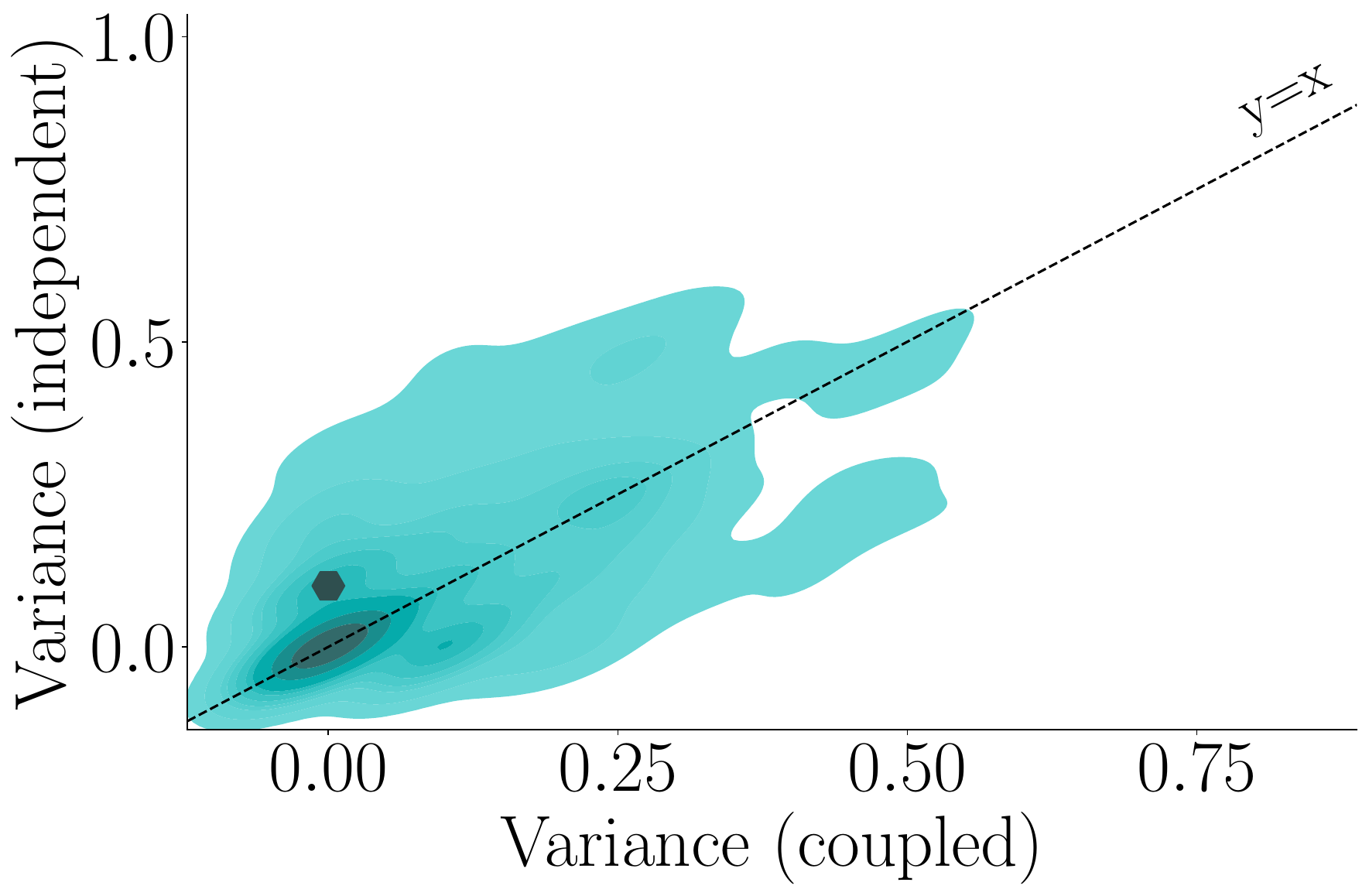} &
    \includegraphics[width=0.23\linewidth]{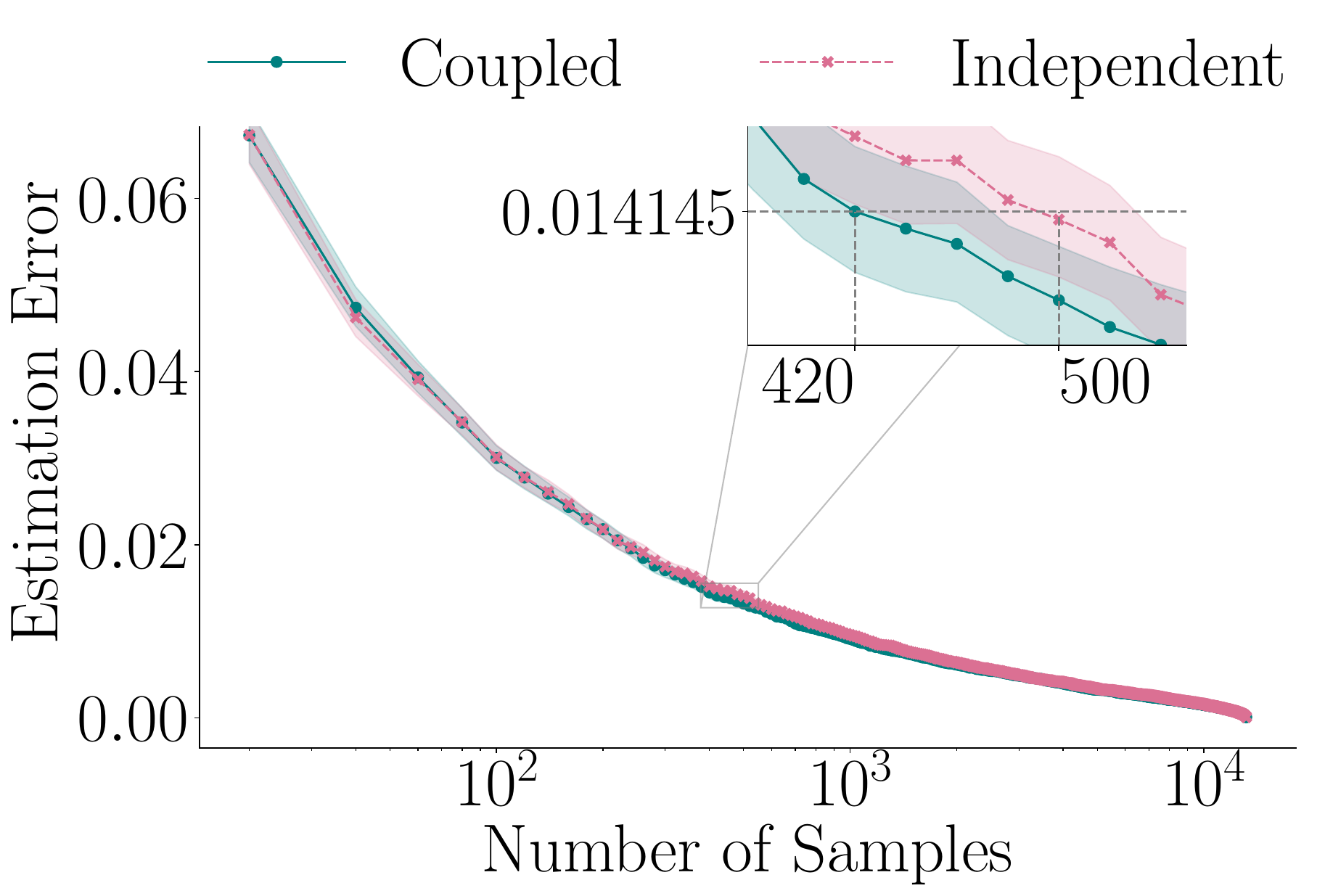} \\ \\
    (a) Score covariance & (b) Variance of the score difference & (c) Estimation error vs. \# samples \\ 
\end{tabular}
    \caption{\textbf{Comparison between several pairs of LLMs in the \texttt{Mistral} family on math questions from the GSM8K dataset.}
    Panels in column (a) show the kernel density estimate (KDE) of the covariance between the scores of the two LLMs on each problem under coupled generation; the dashed lines correspond to average values. Panels in column (b) show the KDE of the variance of the difference between the scores of the LLMs on each question under coupled and independent generation; the highlighted points correspond to median values. Panels in column (c) show the absolute error in the estimation of the expected difference between the scores of the LLMs against the number of samples; for each point on the x-axis, we perform $1{,}000$ sub-samplings and shaded areas correspond to $95\%$ confidence intervals.}
    \label{fig:gsm8k-mistral-first}
\end{figure}
\vspace{-0.2cm}

\vspace{-0.2cm}

\subsection{Pairwise Comparisons}
\begin{figure}[h]
\centering
\includegraphics[width=0.72\linewidth]{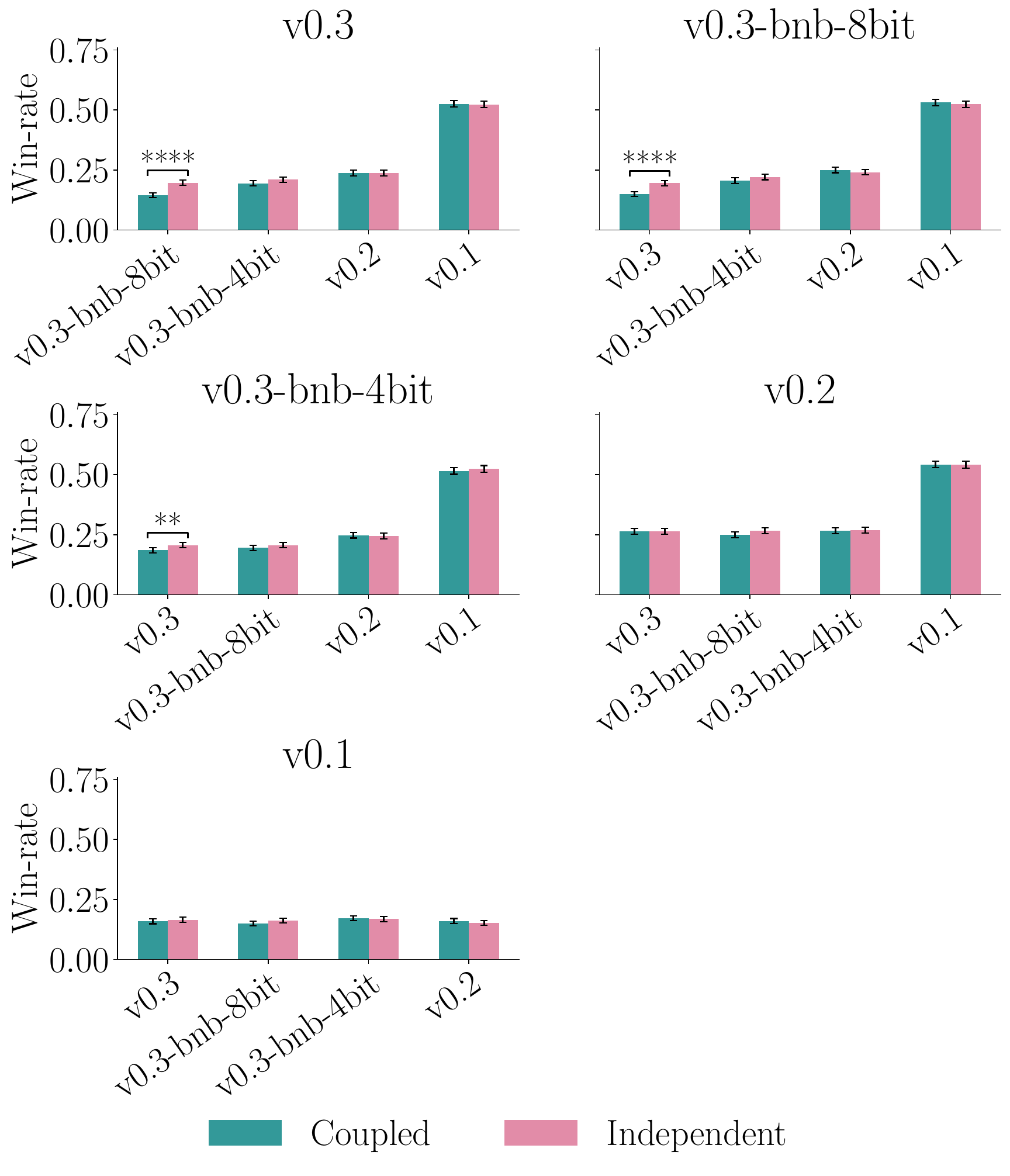}
\caption{
\textbf{Empirical win-rate of each LLM against other LLMs in the \texttt{Mistral} family
on questions from the LMSYS-Chat-1M dataset.} 
Empirical estimate of the win-rate under coupled autoregressive generation as given by Eq.~\ref{eq:coupled-generation-win-rates} and under independent generation generation as given by  Eq.~\ref{eq:independent-generation-win-rates}. 
Each empirical win-rate is computed using pairwise comparisons between the outputs of each LLM and any other LLM over $500$ questions with $10$ (different) random seeds.
The error bars correspond to $95\%$ confidence intervals.
For each pair of empirical win-rates, we conduct a two-tailed test, to test the hypothesis that the empirical win-rates are the same; (\fourstars, \threestars, \twostars, \onestar) indicate $p$-values ($<0.0001$, $<0.001$, $<0.01$, $<0.05$), respectively.
}
\label{fig:lmsys-all-mistral}
\end{figure}

Here, we experiment with models from the \texttt{Mistral} family using pairwise comparisons between their outputs by a strong LLM, when prompted with open-ended questions from the LMSYS Chatbot Arena platform, following the setup described in Section~\ref{sec:pairwise}. 
Table~\ref{tab:lmsys-ranking-mistral} and Figure~\ref{fig:lmsys-all-mistral} summarizes the results, which are qualitatively similar to those in Section~\ref{sec:pairwise}.

\begin{table}[h]
    \centering
    \begin{tabular}{l c c c c}
    \toprule & \multicolumn{2}{c}{Coupled} &  \multicolumn{2}{c}{Independent} \\
    \cmidrule(r{1mm}){2-3} \cmidrule(l{1mm}){4-5}
    {LLM} & Rank & Avg. win-rate 
    & Rank & Avg. win-rate \\ 
    \midrule \texttt{v0.2} & 1 & 0.3310 $\pm$ 0.0026 & 1 & 0.3355 $\pm$ 0.0026 \\ \texttt{v0.3-bnb-4bit} & 2 & 0.2861 $\pm$ 0.0025 & 2 & 0.2953 $\pm$ 0.0025 \\ \texttt{v0.3-bnb-8bit} & 2 & 0.2844 $\pm$ 0.0025 & 2 & 0.2953 $\pm$ 0.0025 \\ \texttt{v0.3} & 4 & 0.2757 $\pm$ 0.0025 & 2 & 0.2918 $\pm$ 0.0025 \\ \texttt{v0.1} & 5 & 0.1598 $\pm$ 0.0020 & 5 & 0.1617 $\pm$ 0.0020 \\ \bottomrule \end{tabular}

    \caption{Average win-rate of each LLM across all other LLMs in the \texttt{Mistral} family ($\pm$ $95\%$ confidence intervals). To derive the rankings, for each LLM, we choose the lowest ranking provided by the method of~\citet{chatzi2024prediction}.}
    \label{tab:lmsys-ranking-mistral}
\end{table}

\end{document}